%% file: main.tex
\documentclass[MAL, biber]{nowfnt}
\usepackage[utf8]{inputenc}

\input{packages}
\input{shortcuts}

\title{The Elements of\\ Differentiable Programming}
\subtitle{}

\maintitleauthorlist{
Mathieu Blondel \\
Google DeepMind \\
mblondel@google.com
\and
Vincent Roulet \\
Google DeepMind  \\
vroulet@google.com
}

\issuesetup
{%
 copyrightowner={Mathieu Blondel and Vincent Roulet},
 pubyear       = 2024,
 }

\usepackage{mwe}

\author[1]{Blondel,Mathieu}
\author[1]{Roulet,Vincent}
\affil[1]{Google DeepMind}

\articledatabox{\nowfntstandardcitation}

\begin{document}

\makeabstracttitle
\begin{abstract}
\input{abstract}
\end{abstract}

\input{toc}

\backmatter
\printbibliography

\end{document}

%% file: packages.tex
\usepackage{amsfonts}
\usepackage{amssymb}
\usepackage{amsmath}
\usepackage{amsthm}
\usepackage{mathtools}
\usepackage{subcaption}
\usepackage{blkarray}

\mathtoolsset{showmanualtags}

\usepackage{booktabs}

\usepackage{enumitem}
\usepackage{graphicx}
\usepackage{bm}
\usepackage{algorithm}
\makeatletter 
 
\@addtoreset{algorithm}{chapter} 
\makeatother
\usepackage[noend]{algpseudocode}
\usepackage{enumitem}
\usepackage[dvipsnames]{xcolor}
\usepackage[capitalize]{cleveref} %

\usepackage{epsdice}

\usepackage{tikz}
\usetikzlibrary{shapes}
\usetikzlibrary{positioning}
\usetikzlibrary{plotmarks}
\usetikzlibrary{patterns}
\usetikzlibrary{intersections,shapes.arrows}
\usetikzlibrary{arrows.meta}
\usetikzlibrary{fit}

\let\originalparagraph\paragraph
\renewcommand{\paragraph}[2][.]{\originalparagraph{#2#1}}

\makeatletter
\renewenvironment{proof}[1][\proofname]{\par
\pushQED{\qed}%
\normalfont \topsep6\p@\@plus6\p@\relax
\trivlist
\item\relax
{\itshape
\color{gray}{#1}\@addpunct{.}}\hspace\labelsep\ignorespaces
\color{gray}
}{%
\popQED\endtrivlist\@endpefalse
}
\makeatother

\usepackage{mdframed}

\definecolor{defcolor}{RGB}{234, 255, 225}  %
\mdfdefinestyle{defsty}{
    backgroundcolor=white,
    rightline=false,
    bottomline=false,
    topline=false,
    linewidth=1pt,
    }
\newenvironment{boxdef}[1]{%
\begin{definition}[#1]\begin{mdframed}[style=defsty]
}{%
\end{mdframed}\end{definition}}

\definecolor{thmcolor}{rgb}{0.94, 0.97, 1.0}  %
\mdfdefinestyle{thmsty}{
    backgroundcolor=white,
    rightline=false,
    bottomline=false,
    topline=false,
    linewidth=1pt,
    }
\newenvironment{boxthm}[1]{%
\begin{theorem}[#1]\begin{mdframed}[style=thmsty]
}{%
\end{mdframed}\end{theorem}}

\newenvironment{boxprop}[1]{%
\begin{proposition}[#1]\begin{mdframed}[style=thmsty]
}{%
\end{mdframed}\end{proposition}}

\crefname{boxcor}{corollary}{corollaries}

\crefname{boxlem}{lemma}{lemmas}

\definecolor{exmcolor}{RGB}{255, 238, 244} %
\mdfdefinestyle{exmsty}{
    backgroundcolor=white,
    rightline=false,
    bottomline=false,
    topline=false,
    linewidth=1pt,
    }
\newenvironment{boxexm}[1]{%
\begin{example}[#1]\begin{mdframed}[style=exmsty]
}{%
\end{mdframed}\end{example}}

\definecolor{remcolor}{RGB}{255,254,224}  %
\mdfdefinestyle{remsty}{
    backgroundcolor=white,
    rightline=false,
    bottomline=false,
    topline=false,
    linewidth=1pt,
    }
\newenvironment{boxrem}[1]{%
\begin{remark}[#1]\begin{mdframed}[style=remsty]
}{%
\end{mdframed}\end{remark}}

\newtheorem*{prop*}{Proposition}

%% file: shortcuts.tex
\newcommand{\EE}{\mathbb{E}}
\newcommand{\FF}{\mathbb{F}}
\newcommand{\II}{\mathbb{I}}
\newcommand{\KK}{\mathbb{K}}
\newcommand{\PP}{\mathbb{P}}
\newcommand{\RR}{\mathbb{R}}
\newcommand{\R}{\mathbb{R}}
\newcommand{\NN}{\mathbb{N}}
\newcommand{\ZZ}{\mathbb{Z}}
\newcommand{\VV}{\mathbb{V}}

\newcommand{\CC}{\mathbb{C}}
\newcommand{\N}{\mathbb{N}}

\newcommand{\indic}{\iota}

\newcommand{\Tr}{\operatorname{tr}}
\newcommand{\diag}{\operatorname{\bf diag}}
\newcommand{\Diag}{\operatorname{\bf Diag}}
\newcommand{\idm}{\operatorname{I}}
\newcommand{\ones}{\operatorname{\mathbf 1}}
\newcommand{\zeros}{{\mathbf{0}}}
\renewcommand{\Vec}{\operatorname{vec}} %

\DeclareMathOperator*{\dom}{dom}
\DeclareMathOperator*{\conv}{conv}
\DeclareMathOperator*{\argmax}{arg\,max}
\DeclareMathOperator*{\argmin}{arg\,min}

\newcommand{\lse}{\mathrm{logsumexp}}
\newcommand{\pow}[1]{^{#1}}
\newcommand{\gn}{\nabla^2_{\mathrm{GN}}}
\newcommand{\fisher}{\nabla^2_{\mathrm{F}}}

\newcommand{\Cov}{\operatorname{Cov}}
\newcommand{\Var}{\operatorname{Var}}
\newcommand{\Gumbel}{\operatorname{Gumbel}}
\newcommand{\Exp}{\operatorname{Exp}}

\newcommand{\logistic}{\mathrm{logistic}}
\newcommand{\sigmoid}{\mathrm{sigmoid}}
\newcommand{\heavistep}{\mathrm{step}}

\newcommand{\parent}{{\operatorname{pa}}}
\newcommand{\rparent}{{\operatorname{random}}}
\newcommand{\dparent}{{\operatorname{determ}}}

\newcommand{\child}{{\operatorname{ch}}}

\newcommand{\sqfn}{\operatorname{sq}} %
\newcommand{\jac}{\bm \partial} %

\newcommand{\eq}{\mathrm{eq}}

\newcommand{\env}{\mathrm{env}}
\newcommand{\envv}{\mathrm{\mathbf{env}}}

\newcommand{\dup}{\mathrm{dup}}

\newcommand{\bv}{{\bm b}}

\newcommand{\dv}{{\bm d}}
\newcommand{\ev}{{\bm e}}

\newcommand{\iv}{{\bm i}}
\newcommand{\kv}{{\bm k}}

\newcommand{\sv}{{\bm s}}
\newcommand{\uv}{{\bm u}}
\newcommand{\vv}{{\bm v}}
\newcommand{\wv}{{\bm w}}
\newcommand{\xv}{{\bm x}}
\newcommand{\yv}{{\bm y}}

\newcommand{\Av}{{\bm A}}

\newcommand{\Cv}{{\bm C}}

\newcommand{\Wv}{{\bm W}}

\newcommand{\Zv}{{\bm Z}}

\renewcommand{\a}{{\bm a}} %
\renewcommand{\b}{{\bm b}} %
\renewcommand{\c}{{\bm c}} %
\renewcommand{\d}{{\bm d}} %
\newcommand{\e}{{\bm e}}

\newcommand{\g}{{\bm g}}
\newcommand{\h}{{\bm h}}
\renewcommand{\l}{{\bm l}} %
\newcommand{\m}{{\bm m}} 
\newcommand{\n}{{\bm n}} 
\newcommand{\p}{{\bm p}}
\newcommand{\q}{{\bm q}}
\renewcommand{\r}{{\bm r}} %
\newcommand{\s}{{\bm s}}
\renewcommand{\t}{{\bm t}} %
\renewcommand{\u}{{\bm u}} %
\renewcommand{\v}{{\bm v}} %
\newcommand{\w}{{\bm w}}
\newcommand{\x}{{\bm x}}
\newcommand{\y}{{\bm y}}
\newcommand{\z}{{\bm z}}

\newcommand{\A}{{\bm A}}
\newcommand{\B}{{\bm B}}
\newcommand{\C}{{\bm C}}
\newcommand{\D}{{\bm D}}
\newcommand{\E}{{\bm E}}

\newcommand{\G}{{\bm G}}
\renewcommand{\H}{{\bm H}}
\newcommand{\J}{{\bm J}}
\newcommand{\K}{{\bm K}}

\newcommand{\M}{{\bm M}}

\newcommand{\Q}{{\bm Q}}
\renewcommand{\S}{{\bm S}}

\newcommand{\U}{{\bm U}}
\newcommand{\V}{{\bm V}}
\newcommand{\W}{{\bm W}}
\newcommand{\X}{{\bm X}}
\newcommand{\Y}{{\bm Y}}
\newcommand{\Z}{{\bm Z}}

\newcommand{\redZ}{{\textcolor{red}{0}}}
\newcommand{\blueO}{{\textcolor{blue}{1}}}

\newcommand{\alphav}{{\bm \alpha}}
\newcommand{\betav}{{\bm \beta}}
\newcommand{\deltav}{{\bm \delta}}
\newcommand{\gammav}{{\bm \gamma}}
\newcommand{\etav}{{\bm \eta}}

\newcommand{\thetav}{{\bm \theta}}
\newcommand{\lambdav}{{\bm \lambda}}
\newcommand{\muv}{{\bm \mu}}

\newcommand{\sigmav}{{\bm \sigma}}
\newcommand{\piv}{{\bm \pi}}

\newcommand{\psiv}{{\bm \psi}}
\newcommand{\rhov}{{\bm \rho}}
\newcommand{\Sigmav}{{\bm \Sigma}}
\newcommand{\omegav}{{\bm \omega}}

\newcommand{\Wt}{\bm{\mathsf W}}
\newcommand{\Vt}{\bm{\mathsf V}}

\def\cA{\mathcal{A}}
\def\cB{\mathcal{B}}
\def\cC{\mathcal{C}}
\def\cD{\mathcal{D}}
\def\cE{\mathcal{E}}
\def\cF{\mathcal{F}}
\def\cG{\mathcal{G}}
\def\cH{\mathcal{H}}

\def\cK{\mathcal{K}}
\def\cL{\mathcal{L}}
\def\cM{\mathcal{M}}
\def\cN{\mathcal{N}}

\def\cP{\mathcal{P}}
\def\cQ{\mathcal{Q}}
\def\cR{\mathcal{R}}
\def\cS{\mathcal{S}}
\def\cT{\mathcal{T}}
\def\cU{\mathcal{U}}
\def\cV{\mathcal{V}}
\def\cW{\mathcal{W}}
\def\cX{\mathcal{X}}
\def\cY{\mathcal{Y}}
\def\cZ{\mathcal{Z}}

\newcommand{\partialfrac}[2]{\frac{\partial #1}{\partial #2}}

\def\wrt{{w.r.t.\ }}
\def\aka{{a.k.a.\ }}
\def\iid{{i.i.d.\ }}

\algdef{SE}[SUBALG]{Indent}{EndIndent}{}{\algorithmicend\ }%
\algtext*{Indent}
\algtext*{EndIndent}

\newcommand{\blue}[1]{\textcolor{blue}{#1}}

\definecolor{lightgray}{gray}{0.7}

\newcommand{\softargmax}{\mathrm{softargmax}}

%% file: abstract.tex
Artificial intelligence has recently experienced remarkable advances, fueled by
large models, vast datasets, accelerated hardware, and, last but not least,
the transformative power of differentiable programming.  This new programming
paradigm enables end-to-end differentiation of complex computer programs
(including those with control flows and data structures), making gradient-based
optimization of program parameters possible.

As an emerging paradigm, differentiable programming builds upon several areas
of computer science and applied mathematics, including automatic
differentiation, graphical models, optimization and statistics.  This book
presents a comprehensive review of the fundamental concepts useful for
differentiable programming.  We adopt two main perspectives, that of
optimization and that of probability, with clear analogies between the two. 

Differentiable programming is not merely the differentiation of
programs, but also the thoughtful design of programs intended for
differentiation. By making programs differentiable, we inherently introduce
probability distributions over their execution, providing a means to quantify
the uncertainty associated with program outputs.

%% file: toc.tex
\input{ack}
\input{source_code}

\input{chapters/notation/notation_main}

\input{chapters/intro/intro_main}

\part{Fundamentals}
\label{part:prelim}

\input{chapters/diff/diff}

\input{chapters/proba_learning/proba_learning}

\part{Differentiable programs}
\label{part:dp}

\input{chapters/neural_nets/neural_nets_main}

\input{chapters/control_flows/control_flows}

\input{chapters/data_struct/data_struct}

\part{Differentiating through programs}
\label{part:cd}

\input{chapters/num_diff/num_diff_main}

\input{chapters/auto_diff/auto_diff}

\input{chapters/higher_order/higher_order}

\input{chapters/graphical_models/graphical_models}

\input{chapters/diff_thru_opt/implicit_diff_main}

\input{chapters/diff_thru_int/grad_est_main} 
\input{chapters/diff_thru_int/ode}

\part{Smoothing programs}
\label{part:smoothing}

\input{chapters/smoothing/smoothing}

\input{chapters/convolution/convolution}

\part{Optimizing differentiable programs}
\label{part:optim}

\input{chapters/optim/basics}

\input{chapters/optim/first_order}
\input{chapters/optim/second_order}

\input{chapters/duality/duality}

%% file: ack.tex
\chapter*{Acknowledgements} 

We thank the following people for sending us feedback, suggestions and typos:
Fabian Pedregosa, 
Kevin Murphy,
Niklas Schmitz,
Nidham Gazagnadou,
Bruno De Backer,
David L\'{o}pez,
Sam Duffield,
Logan Bruns,
Wojciech Stokowiec,
Alex Towell,
John Reid,
Sadish Dhakal,
Fabian Schaipp,
Mahmoud Asem,
Simone Scardapane,
Jonathan Epperlein,
Stephen Fratini,
Po-Hung Yeh,
Adrian Hill,
(add your name here!).
We are particularly indebted to 
Guillaume Gautier,
Archisman Bandyopadhyay,
and
Javier Burroni
for in-depth proofreading.
We also used Gemini extensively for proofreading.

%% file: source_code.tex
\chapter*{Source code}

We provide some Python source code to accompany the book on 
\href{https://github.com/diffprog/code}{github}.

%% file: chapters/notation/notation_main.tex
\chapter*{Notation} 
\label{chap:notation}

\begin{table*}[h]
\caption{Naming conventions}
\centering
\begin{small}
\begin{tabular}{ll}
\toprule
Notation & Description \\ 
\midrule
\addlinespace[0.3em]
$\cX \subseteq \RR^D$ & Input space (e.g., features) \\
$\cY \subseteq \RR^M$ & Output space (e.g., classes) \\
$\cS_k \subseteq \RR^{D_k}$ & Output space on layer or state $k$ \\
$\cW \subseteq \RR^P$ & Weight space \\
$\Lambda \subseteq \RR^Q$ & Hyperparameter space \\
$\Theta \subseteq \RR^R$ & Distribution parameter space, logit space \\
$N$ & Number of training samples \\
$T$ & Number of optimization iterations \\
\midrule
$\x \in \cX$ & Input vector \\
$\y \in \cY$ & Target vector \\
$\s_k \in \cS_k$ & State vector $k$ \\
$\w \in \cW$ & Network (model) weights \\ 
$\lambdav \in \Lambda$ & Hyperparameters \\
$\thetav \in \Theta$ & Distribution parameters, logits \\ 
$\pi \in [0,1]$ & Probability value \\ 
$\piv \in \triangle^M$ & Probability vector \\ 
\bottomrule
\end{tabular}
\end{small}
\end{table*}

\begin{table*}[t]
\caption{Naming conventions (continued)}
\centering
\begin{small}
\begin{tabular}{ll}
\toprule
Notation & Description \\ 
\midrule
\addlinespace[0.3em]
$f$ & Network function \\
$f(\cdot; \x)$ & Network function with $\x$ fixed \\
$L$ & Objective function \\
$\ell$ & Loss function \\
$\kappa$ & Kernel function \\
$\phi$ & Output embedding, sufficient statistic \\
$\mathrm{step}$ & Heaviside step function \\
$\logistic_\sigma$ & Logistic function with temperature $\sigma$ \\
$\logistic$ & Shorthand for $\logistic_1$ \\
\midrule
$p_\thetav$ & Model distribution with parameters $\thetav$ \\
$\rho$ & Data distribution over $\cX \times \cY$ \\
$\rho_\cX$ & Data distribution over $\cX$ \\
$\muv, \sigma^2$ & Mean and variance \\
$Z$ & Random noise variable \\
\bottomrule
\end{tabular}
\end{small}
\end{table*}

%% file: chapters/intro/intro_main.tex
\chapter{Introduction} 
\label{chap:intro} 

\section{What is differentiable programming?}

A computer program is a sequence of elementary instructions for performing a
task. In traditional computer programming, 
the program is typically manually written by a programmer.  
However, for certain tasks, 
particularly those involving intricate patterns and complex decision-making,
such as image recognition or text generation, 
manually writing a program is extremely challenging, if not impossible.

In contrast, modern neural networks offer a different approach. They are
constructed by combining parameterized functional blocks and are trained
directly from data using gradient-based optimization. This end-to-end training
process, where the network learns both feature extraction and task execution
simultaneously, allows neural networks to tackle complex tasks that were
previously considered insurmountable for traditional, hand-coded programs.
This new programming paradigm has been referred to as
``differentiable programming'' or ``software 2.0''.
We give an informal definition below.

\newpage

\begin{boxdef}{Differentiable programming}
Differentiable programming is a programming paradigm in which complex computer
programs (including those with control flows and data structures) can be
differentiated end-to-end automatically, enabling gradient-based optimization of
parameters in the program.
\end{boxdef}

\subsection*{Neural networks as parameterized programs}

\begin{figure}[t]
\begin{center}
\includegraphics[width=0.95\linewidth]{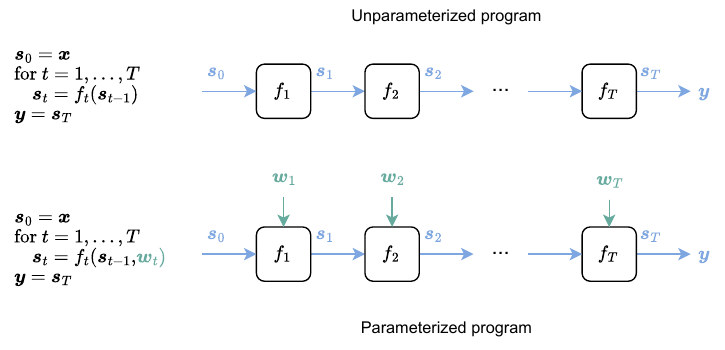}
\end{center}
\caption{
Neural networks can be seen as parameterized programs.
\label{intro:fig:parameterized_programs}}
\end{figure}

In differentiable programming, as in regular computer programming,
a program is defined as the composition of
elementary operations, forming a \textbf{computation graph}. 
The key difference is that,
as illustrated in \cref{intro:fig:parameterized_programs},
the program (such as a neural network)
contains \textbf{parameters} that can be adjusted from data
and can be differentiated end-to-end using \textbf{automatic
differentiation} (autodiff). Typically, it is assumed that the program defines a
\textbf{mathematically valid function} (a.k.a. pure function): the function
should return identical
values for identical arguments and should not have any side effects.
Moreover, the function should have \textbf{well-defined derivatives}, 
ensuring that it
can be used in a gradient-based optimization algorithm.  Therefore,
differentiable programming is not only the art of differentiating through
programs but also of \textbf{designing} meaningful differentiable programs.

\subsection*{Why do we need derivatives?}

Machine learning typically boils down to optimizing a certain objective
function, which is the composition of a loss function and a model (network)
function.
Derivative-free optimization is called \textbf{zero-order optimization}. It only
assumes that we can evaluate the objective function that we wish to optimize.
Unfortunately, it is
known to suffer from the \textbf{curse of dimensionality}, 
i.e., it only scales to low-dimensional problems, 
such as less than $10$ dimensions. Derivative-based
optimization, on the other hand, is much more efficient and can scale to millions
or billions of parameters. Algorithms that use first and second
derivatives are known as \textbf{first-order} and \textbf{second-order}
algorithms, respectively.

\subsection*{Why is autodiff so useful?}

\begin{figure}[t]
\begin{center}
\includegraphics[width=0.60\linewidth]{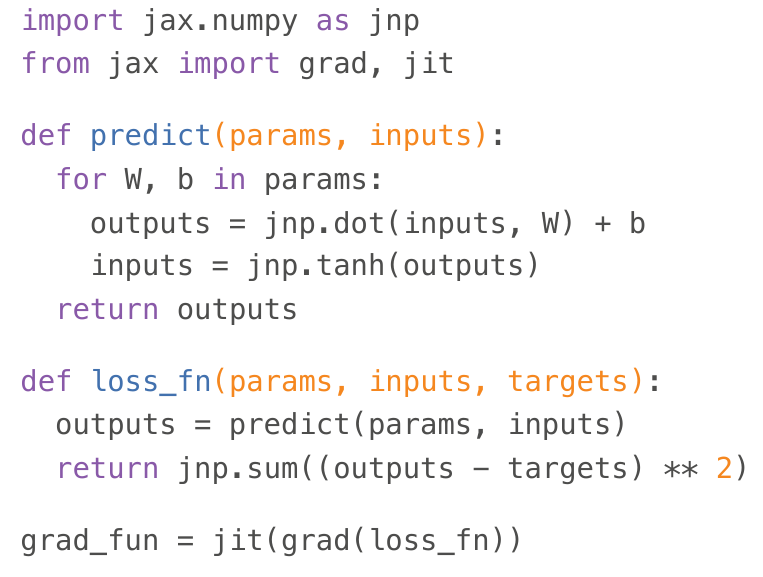}
\end{center}
\caption{
Thanks to automatic differentiation (autodiff), the user can focus on
expressing the forward computation (model), enabling fast experimentation
and alleviating the need for error-prone manual gradient derivation.
\label{intro:fig:code_example}}
\end{figure}

Before the autodiff revolution, researchers and practitioners
needed to manually implement the gradient of the functions they wished to
optimize. Manually deriving gradients can become very tedious for complicated
functions. Moreover, every time the function is changed (for example, for trying
out a new idea), the gradient needs to be
re-derived. Autodiff is a game changer because it allows users
to focus on quickly and creatively experimenting with functions for their tasks.
An example of JAX code \citep{bradbury2018jax} is given in
\cref{intro:fig:code_example}.

\subsection*{Differentiable programming is not just deep learning}

While there is clearly overlap between deep learning and differentiable
programming, their focus is different. Deep learning studies
artificial neural networks composed of multiple layers, able to learn
\textbf{intermediate representations} of the data. 
Neural network architectures have been
proposed with various \textbf{inductive biases}. For example, convolutional neural
networks are designed for images, and transformers are designed for sequences.
On the other hand, differentiable programming studies the
techniques for designing complex programs and differentiating through them. 
It is useful beyond deep learning: for instance in reinforcement
learning, probabilistic programming and scientific computing in general.

\subsection*{Differentiable programming is not just autodiff}

While autodiff is a key ingredient of differentiable programming, it is not
the only one. Differentiable programming is also concerned with the design of
principled differentiable operations. In fact, much research on differentiable
programming has been devoted to making classical computer programming operations
compatible with autodiff. 
As we shall see, many differentiable relaxations can be interpreted in a
probabilistic framework.
A core theme of this book is the interplay between optimization, probability
and differentiation. Differentiation is useful for optimization and conversely,
optimization can be used to design differentiable operators.

\subsection*{Our vision for differentiable programming}

Computer programming offers powerful tools like control flows, data structures,
and standard libraries, enabling users to construct complex programs for solving
intricate problems. 
Our long-term vision is to achieve parity between traditional and differentiable
programming, empowering programmers to seamlessly express differentiable
programs (such as neural networks) using the full suite of tools they are
accustomed to.
However, as discussed earlier,
differentiable programming is not simply a matter of applying automatic
differentiation to existing code. Programs must be designed with
differentiability in mind. This usually comes down to inducing a probability
distribution over the program or its components. While significant work remains
to fully realize this ambitious goal, we hope this book offers a solid
foundation.

\subsection*{Where does ``differentiable programming'' come from?}

While neural networks and autodiff have existed for several decades,
the term ``differentiable programming'' is more recent.
\citet{olah2015neural} discussed analogies between neural network architectures 
and higher-order functions in functional programming, and referred to it as
a ``new kind of programming''.
\citet{dalrymple2016} wrote an essay titled ``differentiable programming'',
recognizing a paradigm where programs learn details through differentiation,
with the expressiveness of functional programming.
\citet{gordon2018} gave a keynote talk titled 
``Some principles of differential programming languages'' at POPL 2018.
The ``differentiable programming'' and ``software 2.0'' terms were popularized,
among others, by \citet{lecun_2018} and \citet{karpathy_2017}. 
From this perspective, autodiff frameworks can be seen, not merely as libraries,
but as domain-specific languages (DSLs) embedded into an existing programming
language, such as Python.  See also \citet{imai2019} for a review.

\section{Book goals and scope}

The present book aims to provide a comprehensive introduction to differentiable
programming with an emphasis on \textbf{core} mathematical tools.
\begin{itemize}
    \item In \cref{part:prelim}, we review \textbf{fundamentals}:
        differentiation and probabilistic learning.

    \item In \cref{part:dp}, we review \textbf{differentiable programs}.
        This includes neural networks, sequence networks and control flows.

    \item In \cref{part:cd}, we review how to \textbf{differentiate through
        programs}.
        This includes automatic differentiation, but also differentiating
        through optimization and integration (in particular, expectations).

    \item In \cref{part:smoothing}, we review \textbf{smoothing programs}. We
        focus on two main techniques: infimal convolution, which comes from the
        world of optimization, and convolution, which comes from the world of
        integration. We also strive to spell out the connections between them.

    \item In \cref{part:optim}, we review \textbf{optimizing programs}: basic
        optimization concepts, first-order
    algorithms, second-order algorithms and duality.

\end{itemize}
Our goal is to present the fundamental techniques useful for differentiable
programming, \textbf{not} to survey how these techniques have been used in
various applications.

\section{Intended audience}

This book is intended to be a graduate-level introduction to
differentiable programming. Our pedagogical choices are made with the machine
learning community in mind. Some familiarity with calculus, linear algebra,
probability theory and machine learning is beneficial.

\section{How to read this book?}

This book does not need to be read linearly chapter by chapter.
When needed, we indicate at the beginning of a chapter what chapters are
recommended to be read as a prerequisite.

\section{Related work}

Differentiable programming builds upon a variety of connected topics.  
We review in this section relevant textbooks, tutorials and software.

Standard textbooks
on backpropagation and automatic differentiation (autodiff) are those of 
\citet{werbos1994roots}
and
\citet{griewank2008evaluating}.
A tutorial with a focus on machine learning is provided by
\citet{baydin2018automatic}.
Automatic differentiation is also reviewed as part of more general textbooks,
such as those of \citet{deisenroth_2020},
\citet{pml1_book} (from a linear algebra perspective) 
and \citet{pml2_book} (from a functional perspective;
autodiff section authored by Roy Frostig).
The present book was also influenced by \citet{peyre_2020}'s textbook on data
science.  The history of reverse-mode autodiff is reviewed by
\citet{griewank2012invented}.

A tutorial on different perspectives of backpropagation is
``There and Back Again: A Tale of Slopes and Expectations''
(\href{https://mml-book.github.io/slopes-expectations.html}{link}),
by Deisenroth and Ong.
A tutorial on
implicit differentiation is
``Deep Implicit Layers - Neural ODEs, Deep Equilibrium Models, and Beyond''
(\href{https://implicit-layers-tutorial.org/}{link}), by
Kolter,  Duvenaud, and Johnson.

The standard reference on inference in graphical models and its connection with
exponential families is that of \citet{wainwright_2008}.
Differentiable programming is also related to probabilistic programming; see,
e.g., \citet{van_2018}.
 
A review of smoothing from the infimal convolution perspective is provided by
\citet{beck_2012_smoothing}.
A standard textbook on convex optimization is that of
\citet{nesterov2018lectures}.
A textbook on first-order optimization methods is that of
\citet{beck_2017}.

Autodiff implementations that accelerated the autodiff revolution in machine
learning are Theano \citep{bergstra_2010} and Autograd \citep{autograd}.
Major modern implementations of autodiff include TensorFlow \citep{abadi_2016},
JAX \citep{bradbury2018jax}, and PyTorch \citep{paszke2019pytorch}.
We in particular acknowledge the JAX team for influencing our view of autodiff.

%% file: chapters/diff/diff.tex
\chapter{Differentiation} \label{chap:diff} 

In this chapter, we review key differentiation concepts.
In particular, we emphasize the fundamental role played by linear maps.

\section{Univariate functions}

\subsection{Derivatives}\label{diff:sec:der_def}

To study functions, we need to capture their
infinitesimal variations around points as defined by the notion of
\textbf{limit}.
\begin{boxdef}{Limit}
  We say that $c \in \RR$ is the \textbf{limit} of $f \colon \RR \to \RR$ as $v
  \in \RR$ approaches $w \in \RR$, denoted
  \[
    \lim_{v \rightarrow w} f(v) = c,
  \]
  if, for any $\varepsilon > 0$, there exists $R > 0$
  such that for any $v \in \RR$ satisfying $0<|v -w| \leq R$, we have 
  $|f(v) - c| \leq \varepsilon$. 
\end{boxdef}
We can also write $f(v) \to c$ as $v \to w$.
Limits are preserved under additions and multiplications. Namely, if
$\lim_{v \rightarrow w} f(v) = c$ and $\lim_{v \rightarrow w} g(v) = d$, then
denoting $(af + bg)(w) \coloneqq af(w) + bg(w)$ for any $a, b \in \RR$ and
$(fg)(w) \coloneqq f(w) g(w)$, we have by definition of the limit,
$\lim_{v \rightarrow w} (af + bg)(v) = ac + bd$
and
$\lim_{v \rightarrow w} (fg)(v) = cd$. 
The preservation of the limit under addition and multiplication by a
scalar is generally referred to as the linearity of the limit, a property that
many definitions in the sequel inherit.

\begin{figure}[t]
  \centering
  \includegraphics[width=0.8\linewidth]{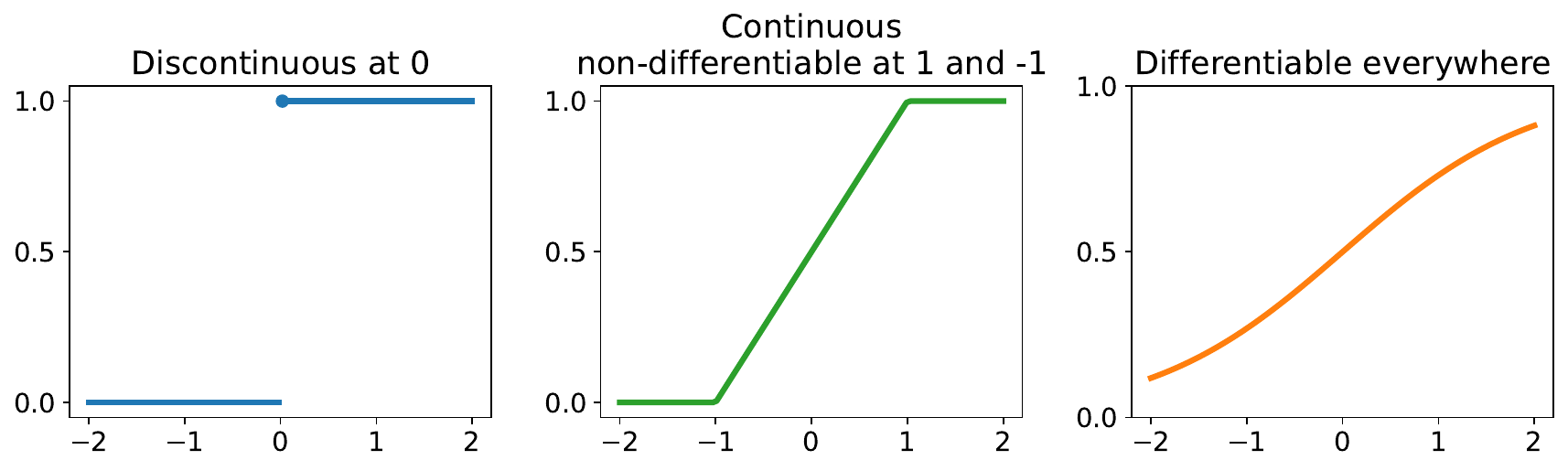}
  \caption{Illustration of discontinuity and non-differentiability.
      {\bf Left.} A discontinuous function, with a jump in function value at $0$. 
      {\bf Center.} A continuous but non-differentiable everywhere function,
    with kinks at $-1$ and $1$. 
    {\bf Right.} A differentiable everywhere function.
  \label{diff:fig:continuity}
  }
\end{figure}

With the notion of limit, we can already delineate a class of ``well-behaved''
functions: functions whose limit at any point equals the value of
the function at that point. Functions satisfying this property are called
\textbf{continuous}.
\begin{boxdef}{Continuous function} \label{diff:def:continuity} A function
$f:\RR \rightarrow \RR$ is continuous at a point $w \in \RR$ if 
\[
\lim_{v \rightarrow w} f(v) = f(w).
\]
A function $f$ is said to be continuous if it is continuous at all points in its
domain.
\end{boxdef}
Although the notion of continuity appears to be a benign assumption, several
functions commonly used in machine learning, such as the Heaviside step function
(displayed in the left panel of~\cref{diff:fig:continuity}), are not continuous
and require special treatment. 
\begin{boxrem}{Little $o$ notation}
In the following, we will make use of Landau's little $o$ notation. We write
\[
  g(v) = o(f(v)) \ \mbox{as} \ v \rightarrow w
\]
if 
\[
  \lim_{v\rightarrow w} \frac{|g(v)|}{|f(v)|} = 0.
\]
That is, the function $f$ dominates $g$ in the limit $v \to w$.
For example, $f$ is continuous at $w$ if and only if 
\[
  f(w+\delta) = f(w) + o(1) \ \mbox{as} \ \delta \rightarrow 0.
\]
\end{boxrem}

\begin{figure}[t]
	\begin{center}
		\includegraphics[width=0.65\linewidth]{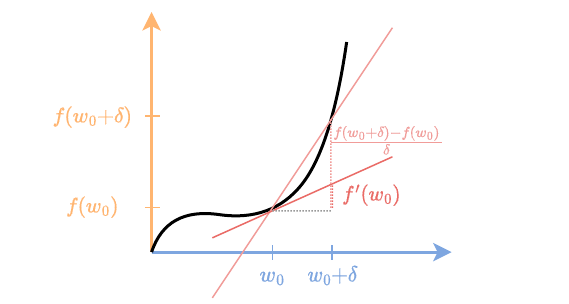}
	\end{center}
	\caption{A function $f$ can be locally approximated around a point $w_0$ by
		a secant, a linear function $w \mapsto aw + b$ with slope $a$ and
		intercept $b$, crossing $f$ at $w_0$ with value $u_0= f(w_0)$ and
		crossing at $w_0 + \delta$ with value $u_\delta=f(w_0+\delta)$. Using
		$u_0 = a w_0 + b$ and $u_\delta = a(w_0 + \delta) + b$, we find that its
		slope is $a = (f(w_0 + \delta) - f(w_0)) / \delta$ and the intercept is
		$b=f(w_0) -aw_0$. The derivative $f'(w)$ of a function $f$ at a point
		$w_0$ is then defined as the limit of the slope $a$ when $\delta \to 0$.
		It is the slope of the tangent of $f$ at $w_0$. The value $f(w)$ of the
		function at $w$ can then be locally approximated around $w_0$ by $w
		\mapsto f'(w_0) w + f(w_0) - f'(w_0) w_0 = f(w_0) + f'(w_0)(w - w_0)$.
		\label{diff:fig:derivative}}
\end{figure}

We now explain derivatives.
Consider a function $f: \RR\rightarrow \RR$. As illustrated in
\cref{diff:fig:derivative}, its value on an interval $[w_0, w_0+\delta]$ can be
approximated by the secant between its values $f(w_0)$ and
$f(w_0+\delta)$, a linear function with slope $(f(w_0+\delta) -
f(w_0))/\delta$. In the limit of an infinitesimal variation $\delta$ around
$w_0$, the secant converges to the \textbf{tangent} of $f$ at $w_0$ and the
resulting slope defines the derivative of $f$ at $w_0$. The definition below
formalizes this intuition.
\begin{boxdef}{Derivative}\label{diff:def:der} The \textbf{derivative} of $f:\RR
	\rightarrow\RR$ at $w\in \RR$ is defined as
	\begin{equation}
	f'(w) \coloneqq \lim_{\delta\rightarrow 0} \frac{f(w+\delta)- f(w)}{\delta},
	\label{diff:eq:derivative_definition}
	\end{equation}
	provided that the limit exists. If $f'(w)$ is well-defined at a particular
$w$, we say that the function $f$ is \textbf{differentiable} at $w$.
If $f$ is differentiable at any $w\in
\RR$, we say that it is \textbf{differentiable everywhere} or differentiable for
short. 
\end{boxdef}

If $f$ is differentiable at a given $w$, then it is necessarily
\textbf{continuous} at $w$ as shown in the following proposition.
Non-differentiability of a continuous function at a given point $w$ is generally
illustrated by a kink, as shown in \cref{diff:fig:continuity}.
\begin{boxprop}{Differentiability implies continuity}
  If $f:\RR \rightarrow \RR$ is
 differentiable at $w\in \RR$, then it is continuous at $w\in \RR$.
 \label{diff:prop:dir_cont}
 \end{boxprop}
 \begin{proof}
   In little $o$ notation, $f$ is differentiable at $w$ if there exists $f'(w)
   \in \RR$, such that 
   \[
     f(w+\delta) = f(w) + f'(w)\delta + o(\delta) \ \mbox{as} \ \delta \rightarrow 0.
   \]
   Since $f'(w)\delta + o(\delta) = o(1)$ as $\delta \rightarrow 0$,
   $f$ is continuous at $w$.
 \end{proof}

In addition to enabling the construction of a linear approximation of $f$ in a
neighborhood of $w$, since it is the slope of the tangent of $f$ at $w$, the
derivative $f'$ informs us about the \textbf{monotonicity} of $f$ around $w$.
If $f'(w)$ is positive, the function is increasing around $w$. Conversely, if
$f'(w)$ is negative, the function is decreasing. Such information can be used to
develop iterative algorithms such as gradient descent,
which seeks to minimize $f$ by computing iterates of
the form $w_{t+1}= w_t - \gamma f'(w_t)$ for $\gamma>0$, thereby moving along
descent directions of $f$ around $w_t$. 

For several elementary functions such as $w^n$, $e^w$, $\ln w$, $\cos w$ or
$\sin w$, their derivatives can be obtained directly by applying the definition
of the derivative in \cref{diff:eq:derivative_definition} as we now illustrate. 
\begin{boxexm}{Derivative of power function}\label{diff:exm:analytic_der}
	Consider $f(w) = w^n$ for $w \in \RR$, $n\in \N\setminus\{0\}$. For any
	$\delta \in \RR$, we have 
	\begin{align*}
		\frac{f(w+\delta) - f(w)}{\delta} 
		& = \frac{(w+\delta)^n - w^n}{\delta} \\
		& = \frac{\sum_{k=0}^{n} \binom{n}{k} \delta^k w^{n-k} - w^n}{\delta} \\
		& =  \sum_{k=1}^{n} \binom{n}{k} \delta^{k-1} w^{n-k} \\
		& = \binom{n}{1}
		w^{n-1}  +  \sum_{k=2}^{n} \binom{n}{k} \delta^{k-1} w^{n-k},
	\end{align*}
	where, in the second line, we used the binomial theorem. Since $
	\binom{n}{1} = n$ and $\lim_{\delta \rightarrow 0} \sum_{k=2}^{n}
	\binom{n}{k} \delta^{k-1} w^{n-k} =0$, we get $f'(w)= n w^{n-1}$.
\end{boxexm}

\begin{boxrem}{Functions on a subset $\cU$ of $\RR$}
For simplicity, we presented the definition of the derivative for a function
defined on the whole set of real numbers $\RR$. If a function
$f:\cU\rightarrow\RR$ is defined on a subset $\cU\subseteq \RR$ of the real numbers,
as it is the case for $f(w)=\sqrt w$ defined on $\cU=\RR_+$, the derivative
of $f$ at $w\in \cU$ is defined by the limit
in~\cref{diff:eq:derivative_definition} provided that the function $f$
is well defined on a neighborhood of $w$, that is, there exists $r>0$ such that
$w+\delta \in \cU$ for any $|\delta|\leq r$.  The function $f$ is then said to
be
\textbf{differentiable everywhere} or differentiable for short if it is
differentiable at any point $w$ in the \textbf{interior} of $\cU$, the set
of points $w\in \cU$ such that $\{w+\delta: |\delta|\leq r\} \subseteq \cU$ for $r$
sufficiently small. For points lying at the boundary of $\cU$ (such as $a$ and $b$
if $\cU=[a, b]$), one may define the right and left derivatives of $f$ at
$a$ and $b$, meaning that the limit is taken by approaching $a$
from the right or $b$ from the left. 
\end{boxrem}

\subsection{Calculus rules}\label{diff:sec:calculus_rules}

For a given $w\in \RR$ and two functions $f:\RR\rightarrow\RR$ and
$g:\RR\rightarrow \RR$, the derivative of elementary operations on $f$ and $g$
such as their sums, products or compositions can easily be derived from the
definition of the derivative, under appropriate conditions on the
differentiability of $f$ and $g$ at $w$. For example, if the
derivatives of $f$ and $g$ exist at $w$, then the derivatives of their weighted
sum or product exist, and satisfy the rules 
\begin{align}
	\forall a, b \in \RR, \ (af+bg)'(w) & = af'(w) +bg'(w) \tag{Linearity}\\
	(fg)'(w) &= f'(w) g(w) + f(w)g'(w),  \tag{Product rule} 
\end{align}
where $(fg)(w) \coloneqq f(w)g(w)$. The linearity can be verified directly from the
linearity of the limits. For the product rule, in little $o$ notation, we have, 
as $\delta\rightarrow 0$,
\begin{align*}
  (fg)(w+\delta) & = (f(w) + f'(w)\delta+ o(\delta))(g(w) + g'(w)\delta + o(\delta)) \\
  & = f(w)g(w) + f'(w)g(w)\delta + f(w)g'(w)\delta + o(\delta),
\end{align*}
hence the result.

If the derivatives of $g$ at $w$ and of $f$ at $g(w)$ exist, then the
derivative of the composition $(f\circ g)(w) \coloneqq f(g(w))$ at $w$ exists 
and is given by 
\begin{align}
\hspace{54pt} (f\circ g)'(w) = f'(g(w)) g'(w).  \tag{Chain rule}
\end{align}
We prove this result more generally in \cref{diff:thm:chain_rule}.
As seen in the sequel, the linearity and the product rule can be seen as byproducts
of the chain rule, making the chain rule the cornerstone of differentiation.

Consider a function that can be
expressed using sums, products or compositions of elementary functions, such as
$f(w) \coloneqq e^w\ln w+\cos w^2$. Its derivative can be computed by applying the
aforementioned rules on the decomposition of $f$ into elementary operations and
functions.

\begin{boxexm}{Applying rules of differentiation}\label{diff:exm:symb_der}
	Consider $f(w) \coloneqq e^w\ln w+\cos w^2$. 
    The derivative of $f$ at $w>0$ can be
	computed step by step as follows, denoting $\sqfn(w) \coloneqq w^2$,
	\begin{align*}
\hspace*{-20pt}	f'(w) & = (\exp \cdot \ln)'(w) + (\cos \circ \sqfn)'(w)  \tag{Linearity}\\
\hspace*{-20pt}	(\exp \cdot \ln)'(w) & = \exp'(w) \cdot \ln(w)  + \exp(w) \cdot \ln'(w)  \tag{Product rule} \\
\hspace*{-20pt}	\quad (\cos \circ \sqfn)'(w) & = \cos'(\sqfn(w)) \sqfn'(w)   \tag{Chain rule} \\
\hspace*{-20pt}	\exp'(w) & = \exp(w),  \hspace{30pt} \ln'(w)  = 1/w, \tag{Elem. func.} \\
\hspace*{-20pt}	\sqfn'(w) & = 2 w, \hspace{44pt} \cos'(w)  =- \sin(w). \hspace{-3pt}  \tag{Elem. func.}
	\end{align*}
We therefore obtain that $f'(w) = e^w \ln w+ e^w/w - 2w\sin w^2$.
\end{boxexm}
Such a process is purely
mechanical and lends itself to an automated procedure, which is the main idea of
automatic differentiation presented in \cref{chap:auto_diff}. 

\subsection{Leibniz's notation}
The notion of derivative was first introduced independently by Newton and
Leibniz in the 18\textsuperscript{th} century~\citep{ball1960short}. The latter
considered derivatives as the quotient of infinitesimal variations. Namely,
denoting $u = f(w)$ a variable depending on $w$ through $f$, Leibniz considered
the derivative of $f$ as the quotient
\[
	f' = \frac{du}{dw} \quad \mbox{with} \quad f'(w) = \left.\frac{du}{dw}\right|_w
\]
where $du$ and $dw$ denote infinitesimal variations of $u$ and $w$ respectively
and the symbol $|_w$ denotes the evaluation of the derivative at a given point
$w$.
This notation simplifies the statement of the chain rule first discovered by
Leibniz~\citep{rodriguez2010semiotic} as we have for $v = g(w)$ and $u = f(v)$
\[
	\frac{du}{dw} = \frac{du}{dv} \cdot \frac{dv}{dw}.
\]
This hints that derivatives are multiplied when considering compositions. At
evaluation, the chain rule in Leibniz notation recovers the formula presented
above as
\[
	\left.\frac{du}{dw}\right|_w 
	= \left.\frac{du}{dv}\right|_{g(w)} \left.\frac{dv}{dw}\right|_w 
	=  f'(g(w)) g'(w) = (f\circ g)'(w).
\]
The ability of Leibniz's notation to capture the chain rule as a mere product of
quotients made it popular throughout the centuries, especially in
mechanics~\citep{ball1960short}. The rationale behind Leibniz's notation,
the concept of ``infinitesimal variations'', was questioned by later
mathematicians for its potential logical issues~\citep{ball1960short}. The
notation $f'(w)$ first introduced by Euler and further popularized by
Lagrange~\citep{cajori1993history} has then taken over in numerous mathematical
textbooks. The concept of infinitesimal variations has been rigorously
defined by using the set of hyperreal numbers. They extend the set of real
numbers by considering each number as a sum of a non-infinitesimal part and an
infinitesimal part~\citep{hewitt1948rings}. The formalism of infinitesimal
variations further underlies the development of automatic differentiation
algorithms through the concept of dual numbers.

\section{Multivariate functions}

\subsection{Directional derivatives}

Let us now consider a function $f:\RR^P \rightarrow \RR$ with multi-dimensional
input $\w \coloneqq (w_1, \ldots, w_P) \in \RR^P$. 
The most important example in machine learning is
a function which, to the parameters $\w \in \RR^P$ of a neural network,
associates a loss value in $\RR$. Variations of $f$ need to be defined along
specific directions, such as the variation $f(\w+\delta\v) {-}f(\w)$ of $f$
around $\w \in \RR^P$ in the direction $\v \in \RR^P$ by an amount $\delta>0$.
This consideration naturally leads to the definition of the directional derivative. 
\begin{boxdef}{Directional derivative}
\label{diff:def:dir_deriv} 
The \textbf{directional derivative} 
of $f$ at $\w$ in the \textbf{direction} $\v$ is given by 
\begin{equation}\nonumber
	\partial f(\w)[\v] 
    \coloneqq \lim_{\delta \rightarrow 0} \frac{f(\w+ \delta \v) - f(\w)}{\delta},
\end{equation}
provided that the limit exists. 
\end{boxdef}

We use the notation $[\v]$ to emphasize that, for a given input $\w$, 
we can see $\v \mapsto \partial f(\w)[\v]$ as a
function.  This is essential to define the differentiability of $f$
(\cref{diff:def:differentiability}) at $\w$ and to later define linear
maps (\cref{diff:sec:jvp_vjp}). 

One example of directional derivative consists in computing the derivative of a
function $f$ at $\w$ in any of the canonical directions 
\begin{equation*}
\e_i 
\coloneqq (0, \ldots, 0, \underbrace{1}_i, 0, \ldots, 0) .
\end{equation*}
This allows us to define the notion of \textbf{partial derivatives}, denoted 
for $i \in [P]$
\begin{equation}\nonumber
	\partial_i f(\w) 
    \coloneqq \partial f(\w)[\ev_i] 
    = \lim_{\delta \rightarrow 0} \frac{f(\w+ \delta \e_i) - f(\w)}{\delta}.
\end{equation}
This is also denoted in Leibniz's notation as $\partial_i f(\w) = \frac{\partial
f(\w)}{\partial w_i}$ or $\partial_i f(\w) = \partial_{w_i} f(\w)$. By moving
along only the $i$\textsuperscript{th} coordinate of the function, the partial
derivative is akin to differentiating the function $\omega_i \mapsto f(w_1,
\ldots, \omega_i, \ldots, w_P)$ around $\omega_i$, letting all other coordinates
fixed at their values $w_i$.

\subsection{Gradients}\label{diff:sec:grad}

We now introduce the gradient vector, which gathers the partial derivatives.
We first recall the definitions of linear map and linear form.
\begin{boxdef}{Linear map, linear form} \label{diff:def:linear}
  A function $l: \RR^P \rightarrow \RR^M$ is a \textbf{linear map} if for any
  $a_1, a_2 \in \RR$, $\v_1, \v_2 \in \RR^D$,
  \[
    l[a_1 \v_1 + a_2 \v_2] = a_1 l[\v_1] + a_2 l[\v_2].
  \]
  A linear map with values in $\RR$, $l: \RR^P \rightarrow \RR$, is called a
  \textbf{linear form}.
\end{boxdef}
Linearity plays a crucial role in the differentiability of a function.
\begin{boxdef}{Differentiability, single-output
  case}\label{diff:def:differentiability} A function $f: \RR^P \rightarrow \RR$
  is \textbf{differentiable} at $\w\in \RR^P$ if its directional derivative is
  defined along any direction, is linear in any direction $\v$, and if
  \[
  \lim_{\|\v\|_2 \rightarrow 0} 
  \frac{|f(\w+ \v) - f(\w) - \partial f(\w)[\v]|}{\|\v\|_2} = 0,
  \]
  where $\|\v\|_2 \coloneqq \sqrt{\langle \v, \v \rangle}$.
\end{boxdef}
We can now introduce the gradient.
\begin{boxdef}{Gradient}\label{diff:def:gradient} 
  The \textbf{gradient} of a differentiable function $f: \RR^P \rightarrow \RR$
  at a point $\w \in \RR^P$ is defined as the vector of partial derivatives
\[
\nabla f(\w) 
\coloneqq \begin{pmatrix} \partial_1 f(\w) \\
  \vdots \\
  \partial_P f(\w) \end{pmatrix}
  = \begin{pmatrix} \partial f(\w)[\e_1] \\
  \vdots \\
  \partial f(\w)[\e_P] \end{pmatrix} \in \RR^P.
\]
By linearity, the directional derivative of $f$ at $\w$ in the direction
$\v=\sum_{i=1}^P v_i \ev_i$ is then given by
\[
\hspace{-1ex}\partial f(\w)[\v] = \sum_{i=1}^P v_i \partial f(\w)[\ev_i] =
\langle \v, \nabla f(\w)\rangle \in \RR.
\]
\end{boxdef}
Here, $\langle \cdot, \cdot \rangle$ denotes the inner product.
We provide its definition in Euclidean spaces in \cref{diff:sec:euc_spaces}.

In the definition above, the fact that the gradient can be used to compute the
directional derivative is a mere consequence of the linearity of $\partial
f(\w)[\v]$ \wrt $\v$. However, in more
abstract cases presented in later sections, the gradient is defined directly
through this property. 

As a simple example, any linear function of the form $f(\w) = \langle \a, \w
\rangle = \sum_{i=1}^P a_i w_i$ is differentiable as we have 
$(\langle \a, \w+\v \rangle - \langle \a, \w \rangle -
\langle \a, \v \rangle)/\|\v\|_2 = 0$ 
for any $\v$ and in particular for $\|\v\| \rightarrow 0$.
Moreover, its gradient is naturally given by $\nabla f(\w) = \a$.

More generally, to show that a function is differentiable and find its gradient,
one approach is to approximate $f(\w + \v)$ around $\v = \zeros$. 
If we can find a vector $\g$ such that 
\[
  f(\w+\v) = f(\w) + \langle \g, \v\rangle + o(\|\v\|_2),
\]
then $f$ is differentiable at $\w$, since $\langle \g, \cdot \rangle$ is linear.
Moreover, $\g$ is then the gradient of $f$ at $\w$.
\begin{boxrem}{Gateaux and Fr\'echet differentiability}
  Multiple definitions of differentiability exist. The one presented in
  \cref{diff:def:differentiability} is that of \textbf{Fr\'echet differentiable}
  functions. Alternatively, if $f:\RR^P \rightarrow \RR$ has well-defined
  directional derivatives along any direction then the function is
  \textbf{Gateaux differentiable}. Note that the existence of directional
  derivatives in any direction is not a sufficient condition for the function
  to be differentiable. In other words, any Fr\'echet differentiable function is
  Gateaux differentiable, but the converse is not true. As a counter-example, one
  can verify that the function $f(x_1, x_2) \coloneqq x_1^3/(x_1^2 + x_2^2)$ is Gateaux
  differentiable at $0$ but not (Fr\'echet) differentiable at $0$ (because the
  directional derivative at 0 is not linear).
  
  Some authors also require Gateaux differentiable functions to have linear
  directional derivatives along any direction. These are still not Fr\'echet
  differentiable functions. Indeed, the limit in
  \cref{diff:def:differentiability} is over any vectors tending to $0$
  (potentially in a pathological way), while directional derivatives look at
  such limits uniquely in terms of a single direction.

  In the remainder of this chapter, 
  all definitions of differentiability are in terms of
  Fr\'echet differentiability.
\end{boxrem}

The next example illustrates how to compute the gradient of the logistic
loss and validates its differentiability.

\begin{boxexm}{Gradient of logistic loss}\label{diff:exm:grad} Consider the
    logistic loss $\ell(\thetav, \y) \coloneqq - \langle \y, \thetav \rangle 
    + \log \sum_{i=1}^M e^{\theta_i}$, 
    that measures the prediction error of the logits
    $\thetav \in \RR^M$ \wrt the correct label $\y \in \{\ev_1, \dots,
    \ev_M\}$. 
    Let us compute the gradient of this loss \wrt $\thetav$ for fixed
    $\y$, i.e., we want to compute the gradient of $f(\thetav) \coloneqq
    \ell(\thetav, \y)$. 
    Let us decompose $f$ as 
    $f = l + \mathrm{logsumexp}$ 
    with
    $l(\thetav) \coloneqq\langle -\y, \thetav\rangle$ 
    and 
    \[
      \mathrm{logsumexp}(\thetav)
      \coloneqq \log \sum_{i=1}^M \exp(\theta_i), 
    \]
    the log-sum-exp function. 
    The function $l$ is linear so differentiable with gradient $\nabla l(\thetav)
    = -\y$. We therefore focus on $\mathrm{logsumexp}$. Denoting $\exp(\thetav) =
    (\exp(\theta_1), \ldots, \exp(\theta_M))$, using that $\exp(x) = 1 + x +
    o(x)$, $\log(1+ x) = x + o(x)$, and denoting $\odot$ the elementwise product, we get
    \begin{align*}
      \mathrm{logsumexp}(\thetav + \v) 
      & = \log\left(\langle \exp(\thetav + \v), \ones \rangle\right) \\
      & = \log\left(\langle \exp(\thetav) \odot \exp(\v), \ones\rangle\right) \\
      & = \log\left(\langle \exp(\thetav) \odot (\ones + \v + o(\|\v\|_2)), \ones\rangle\right) \\
      & = \log\left(\langle \exp(\thetav), \ones\rangle + \langle  \exp(\thetav), \v \rangle + o(\|\v\|_2)\right)\\
      & = \log\left(\langle \exp(\thetav), \ones\rangle \right) 
      + \left\langle \frac{\exp(\thetav)}{\langle \exp(\thetav), \ones\rangle} , \v\right\rangle + o(\|\v\|_2).
    \end{align*}    
    The above decomposition of $\mathrm{logsumexp}(\thetav + \v)$ shows that
    it is differentiable, and that $\nabla \mathrm{logsumexp}(\thetav) =
    \softargmax(\thetav)$, where 
    \begin{align*}
      \softargmax(\thetav) & 
      \coloneqq \left(e^{\theta_1}/\left(\sum_{j=1}^{M}e^{\theta_j}\right), \ldots, e^{\theta_M}/\left(\sum_{j=1}^M
      e^{\theta_j}\right)\right).
    \end{align*}
     Overall, we get that  $\nabla f(\thetav) = -\y + \mathrm{softargmax}(\thetav)$.
\end{boxexm}

\subsubsection*{Linearity of gradients}

The notion of differentiability for multi-input functions naturally inherits
from the linearity of derivatives for single-input functions. For any $u_1,
\ldots, u_M \in \RR$ and any multi-input functions $f_1, \ldots, f_M$
differentiable at $\w$, the function $u_1 f_1 + \ldots + u_M f_M$ is
differentiable at $\w$ and its gradient is 
\[
\nabla (u_1 f_1 + \ldots + u_M f_M)(\w) 
= u_1 \nabla f_1(\w) + \ldots + u_M \nabla f_M(\w).
\]

\subsubsection*{Why is the gradient useful?}

\begin{figure}
	\begin{minipage}{0.48\linewidth}
		\begin{center}
			\includegraphics[width=\linewidth]{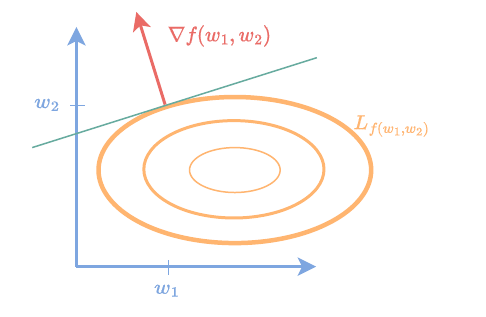}
		\end{center}
		\caption{The gradient of a function $f: \RR^2\rightarrow \RR$ at $(w_1,
	w_2)$ is the normal vector to the tangent space of the level set $L_{f(w_1,
	w_2)} = \{(w_1', w_2'): f(w_1', w_2') = f(w_1, w_2)\}$ and points towards
	points with higher function values.  \label{diff:fig:gradient}}
	\end{minipage}\hspace{10pt}
	\begin{minipage}{0.48\linewidth}
		\begin{center}
			\includegraphics[width=\linewidth]{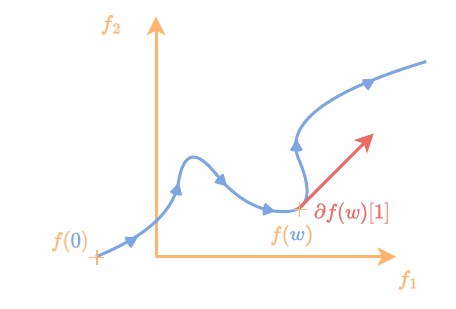}
		\end{center}
		\caption{The directional derivative of a parametric curve $f:\RR
		\rightarrow\RR^2$ at $w$ is the tangent to the curve at the point
		$f(w)\in \RR^2$.\label{diff:fig:directional_deriv}}
	\end{minipage}
\end{figure}

When $f$ is differentiable, we say that $\v$ is an \textbf{ascent direction} of
$f$ from $\w$ if 
\begin{equation*}
    \langle \v, \nabla f(\w) \rangle > 0.
\end{equation*}
Conversely, we say that $\v$ is a \textbf{descent direction} of
$f$ from $\w$ if 
\begin{equation*}
    \langle \v, \nabla f(\w) \rangle < 0.
\end{equation*}
Using this definition, the gradient leads to the \textbf{steepest ascent
direction} of $f$ from $\w$. To see why, we note that
\begin{equation*}
\begin{aligned}
\argmax_{\v\in \RR^P, \|\v\|_2\leq 1} \langle \v, \nabla f(\w) \rangle
&= \argmax_{\v\in \RR^P, \|\v\|_2\leq 1} \partial f(\w)[\v] \\
&= \nabla f(\w)/\|\nabla f(\w)\|_2,
\end{aligned}
\end{equation*}
where we assumed $\nabla f(\w) \neq \mathbf{0}$. The gradient $\nabla f(\w)$ is
orthogonal to the level set of the function (the set of points $\w$ sharing the
same value $f(\w)$) and points towards higher values of $f$, as illustrated in
\cref{diff:fig:gradient}.  
Conversely, the negative gradient $-\nabla f(\w)$ points towards lower values of
$f$.  This observation motivates the development of optimization algorithms such
as gradient descent. It is based on iteratively performing 
the update $\w_{t+1} \coloneqq \w_t -\gamma \nabla f(\w_t)$, for
$\gamma>0$. It therefore seeks for a minimizer of $f$ by moving along the
\textbf{steepest descent direction} around $\w_t$ given, up to a multiplicative
factor, by $-\nabla f(\w_t)$. 
See also \cref{optim:rem:descent_dir} for more details.

\subsection{Jacobians}\label{diff:sec:jac}

Let us now consider a multi-output function $f:\RR^P \rightarrow \RR^M$
defined by $f(\w) \coloneqq (f_1(\w), \dots, f_M(\w))$, where $f_j \colon \RR^P
\to \RR$. A typical example in machine learning is a neural network. The
notion of directional derivative can be extended to such function by defining it
as the vector composed of the coordinate-wise directional derivatives:
\[
\partial f(\w)[\v] 
\coloneqq  \lim_{\delta \rightarrow 0} \frac{f(\w+ \delta \v) - f(\w)}{\delta} 
= \lim_{\delta \rightarrow 0}\begin{pmatrix}
	 \frac{f_1(\w+ \delta \v) - f_1(\w)}{\delta}\\
	\vdots\\
	 \frac{f_M(\w+ \delta \v) - f_M(\w)}{\delta}
	\end{pmatrix} \in \RR^M,
\]
where the limits (provided that they exist) are applied coordinate-wise. The
directional derivative of $f$ in the direction $\v \in \RR^P$ is therefore the
vector that gathers the directional derivative of each $f_j$, i.e., $\partial
f(\w)[\v] = (\partial f_j(\w)[\v])_{j=1}^M$. In particular, we can define the
\textbf{partial derivatives} of $f$ at $\w$ as the vectors 
\[
	\partial_i f(\w) 
	\coloneqq \partial f(\w)[\ev_i] = \begin{pmatrix}
		\partial_i f_1(\w) \\
		\vdots \\
		\partial_i f_M(\w)
	\end{pmatrix}\in \RR^M.
\]
As for the usual definition of the derivative, the directional derivative can
provide a linear approximation of a function around a current input, as
illustrated in \cref{diff:fig:directional_deriv} for a parametric curve
$f:\RR \rightarrow \RR^2$.

Just as in the single-output case, differentiability is defined not only as the
existence of directional derivatives in any direction but also by the linearity
in the chosen direction.
\begin{boxdef}{Differentiability, multi-output
  case}\label{diff:def:differentiability_jac} A function $f:\RR^P\rightarrow
  \RR^M$ is (Fr\'echet) \textbf{differentiable} at a point $\w \in \RR^P$ if its
  directional derivative is defined along any direction, is linear in any
  direction, and, 
  \[
    \lim_{\|\v\|_2 \rightarrow 0} 
    \frac{\|f(\w+ \v) - f(\w) - \partial f(\w)[\v]\|_2}{\|\v\|_2} = 0.
  \]
\end{boxdef}
The partial derivatives of each coordinate's function are gathered in the
\textbf{Jacobian matrix}. 
\begin{boxdef}{Jacobian}
    \label{diff:def:jac} 
  The \textbf{Jacobian} of a differentiable function $f:\RR^P\rightarrow \RR^M$ at $\w$ is
  defined as the matrix gathering partial derivatives of each coordinate's
  function provided they exist,
	\begin{align*}
		\jac f (\w) 
		& \coloneqq \begin{pmatrix}
				\partial_1 f_1(\w) & \ldots & \partial_P f_1(\w) \\
				\vdots & \ddots & \vdots \\
				\partial_1 f_M(\w) & \ldots & \partial_P f_M(\w)
			\end{pmatrix} \in \RR^{M \times P}.
	\end{align*}
	The Jacobian can be represented by stacking columns of partial derivatives or rows
	of gradients,
	\begin{align*}
		\jac f (\w) 
		& = \begin{pmatrix}
				\partial_1 f(\w), \ldots, \partial_P f(\w)
			\end{pmatrix} 
		  = \begin{pmatrix}
			\nabla f_1(\w)^\top \\
			\vdots \\
			\nabla f_M(\w)^\top
			\end{pmatrix} \in \RR^{M \times P}.
	\end{align*}
  By linearity, the directional derivative of $f$ at $\w$ along any input
  direction $\v= \sum_{i=1}^P v_i\ev_i \in \RR^P$ is then given by 
  \[
      \partial f(\w)[\v] 
      = \sum_{i=1}^P v_i \partial_i f(\w) = \jac f(\w) \v \in \RR^M.
  \]
\end{boxdef}
Notice that we use bold $\jac$ to indicate the Jacobian, seen as a matrix. The
Jacobian matrix naturally generalizes the concepts of derivatives and gradients
presented earlier. As for the single input case, to show that a function is
differentiable, one approach is to approximate $f(\w+ \v)$ around $\v=\zeros$.
If we find a linear map $l$ such that 
\[
  f(\w+\v) = f(\w) + l[\v] + o(\|\v\|_2),
\]
then $f$ is differentiable at $\w$. Moreover, if $l$ is represented by matrix
$\J$ such that $l[\v] = \J\v$ then $\J = \jac f(\w)$.

As a simple example, any linear function $f(\w) = \A\w$ for $\A \in
\RR^{M\times P}$ is differentiable, since all its coordinate-wise components are
single-output linear functions, and the Jacobian of $f$ at any $\w$ is given by
$\jac f(\w) = \A$.

\begin{boxrem}{Special cases of the Jacobian}\label{diff:remark:jac_scalar_case} 
For single-output functions $f:\RR^P \rightarrow\RR$, i.e., $M=1$, the Jacobian
matrix reduces to a row vector identified as the \textbf{transpose of the
gradient},
\[
  \jac f(\w) = \nabla f(\w)^\top \in \RR^{1\times P}.
\]
For a single-input function $f: \RR\rightarrow \RR^M$, the Jacobian reduces to
a single column vector of directional derivatives, denoted
\[
  \jac f(w) = f'(w) \coloneqq 
  \begin{pmatrix}
    f_1'(w) \\
    \vdots \\ 
    f_M'(w)
  \end{pmatrix} \in \RR^{M\times 1}.
\]
For a single-input single-output function
$f:\RR\rightarrow\RR$, the Jacobian reduces to the derivative of $f$, i.e.,
\[
  \jac f(w) = f'(w) \in \RR.
\]
\end{boxrem}

The next example illustrates the form of the Jacobian matrix for the
element-wise application of a differentiable function $\sigma$, 
such as the softplus activation (see \cref{neural_nets:sec:activations} for a
review of activation functions).
In this case, the Jacobian takes a simple
diagonal matrix form. As a consequence, the directional derivative
associated with this function is simply given by an element-wise product:
a full matrix-vector product is not needed, as would suggest \cref{diff:def:jac}. 
We will revisit this point in \cref{diff:sec:jvp_vjp}. 

\begin{boxexm}{Jacobian matrix of the softplus activation}
\label{diff:exm:softplus}
	Consider the element-wise application of the softplus defined for
	$\w \in \RR^P$ by
	\[
		f(\w) 
		\coloneqq \begin{pmatrix}
			\sigma(w_1) \\
			\vdots \\
			\sigma(w_P)
		\end{pmatrix}\in \RR^P \quad \mbox{where} \quad \sigma(w) \coloneqq \log(1+e^w).
	\]
	Since $\sigma$ is differentiable, each coordinate of this function is
	differentiable and the overall function is differentiable. The
	$j$\textsuperscript{th} coordinate of $f$ is independent of the
	$i$\textsuperscript{th} coordinate of $\w$ for $i\neq j$, so $\partial_i
	f_j(\w) = 0$ for $i\neq j$. For $i=j$, the result boils down to the
	derivative of $\sigma$ at $w_j$. That is, $\partial_j f_j(\w) =
    \sigma'(w_j)$,
	where $\sigma'(w) = e^{w}/(1+e^w)$. The Jacobian of $f$ is therefore a
	diagonal matrix
	\[
	\jac f(\w) 
    = \diag(\sigma'(w_1), \dots, \sigma'(w_P))
	\coloneqq \begin{pmatrix}
		\sigma'(w_1) & 0 & \ldots & 0 \\
		0 & \ddots & \ddots & \vdots \\
		\vdots & \ddots & \ddots & 0 \\
		0 & \ldots & 0 & \sigma'(w_P)
	\end{pmatrix}.
	\] 
\end{boxexm}

\subsubsection*{Chain rule}

Equipped with a generic definition of differentiability and the associated
objects, gradients and Jacobians, we can now generalize the chain rule,
previously introduced for single-input single-output functions.
\begin{boxprop}{Chain rule}
	\label{diff:thm:chain_rule} Consider $f:\RR^P \rightarrow\RR^M$ and
	$g:\RR^M \rightarrow \RR^R$. If $f$ is differentiable at $\w \in \RR^P$
	and $g$ is differentiable at $f(\w) \in \RR^M$, then the composition $g
	\circ f \colon \RR^P \to \RR^R$ is differentiable at $\w \in \RR^P$ and 
    its Jacobian is given by
	\[
        \underbrace{\jac (g\circ f)(\w)}_{R \times P} 
        = 
        \underbrace{\jac g(f(\w))}_{R \times M} 
        \underbrace{\jac f(\w)}_{M \times P}.
	\]
\end{boxprop}
\begin{proof}
  We progressively approximate $g\circ f(\w +\v)$ using the
  differentiability of $f$ at $\w$ and $g$ at $f(\w)$,
  \begin{align*}
    g(f(\w+\v)) & = g(f(\w) + \jac f(\w)\v + o(\|\v\|)) \\
    & = g(f(\w)) + \jac g(f(\w)) \jac f(\w)\v + o(\|\v\|).
  \end{align*}
  Hence, $g\circ f$ is differentiable at $\w$ with Jacobian $\jac g(f(\w))
  \jac f(\w)$.
\end{proof}
\cref{diff:thm:chain_rule} can be seen as the cornerstone of any derivative
computations. For example, it can be used to rederive the linearity and 
product rules associated to the derivatives of single-input single-output
functions. 

When $g$ is scalar-valued, 
combined with \cref{diff:remark:jac_scalar_case}, we obtain
a simple expression for $\nabla (g \circ f)$.
\begin{boxprop}{Chain rule, scalar-valued case}
	\label{diff:thm:chain_rule_scalar} Consider $f:\RR^P \rightarrow\RR^M$ and
	$g:\RR^M \rightarrow \RR$.
	The gradient of the composition is given by 
	\[
		\nabla 	(g\circ f)(\w) = \jac f(\w)^\top \nabla g(f(\w)).
	\]
\end{boxprop}
This is a very useful identity in machine learning, as we often need to compose
a vector-valued model function and a scalar-valued loss function.  We illustrate
this with linear regression below.
\begin{boxexm}{Linear regression}\label{diff:exm:lin_reg} Consider
	a linear regression of $N$ inputs $\x_1, \ldots, \x_N \in
	\RR^D$ onto $N$ targets $y_1, \ldots, y_N\in \RR$, using a parameter vector 
    $\w \in \RR^{D}$. The loss is defined as the sum of squared residuals,
    $L(\w) \coloneqq \|\X \w -\y\|_2^2 = \sum_{i=1}^N (\langle \x_i, \w \rangle -y_i)^2$ 
    where $\X \coloneqq (\x_1, \ldots, \x_N)^\top \in \RR^{N\times D}$ and 
    $\y \coloneqq (y_1, \ldots, y_N)^\top \in \R^N$.
	
	The function $L$ can be decomposed into a linear mapping $f(\w) \coloneqq \X\w$
	and a squared error $\ell(\p) \coloneqq \|\p- \yv\|_2^2$, 
    so that $L= \ell \circ f$. We can then apply the chain rule 
in \cref{diff:thm:chain_rule_scalar}
    to get 
	\[
	\nabla L(\w) 
	= \jac f(\w)^\top \nabla \ell(f(\w))
	\]
	provided that $f$ and $\ell$ are differentiable at $\w$ and $f(\w)$, respectively.
	
	The function $f$ is linear so differentiable with Jacobian $\jac f(\w)
	= \X$. On the other hand the partial derivatives of $\ell$ are given by
	$\partial_j \ell(\p)= 2(p_j  - y_j)$ for $j \in \{1, \ldots, N\}$. Therefore,
	$\ell$ is differentiable at any $\p$ and its gradient  is $\nabla \ell(\p) =
	2(\p - \yv)$. By combining the two, we then get the gradient of $L$ as 
	\[
	\nabla L(\w) 
	= 2\X^\top (f(\w) - \y) = 2\X^\top(\X \w -\y).
	\] 
\end{boxexm}

\section{Linear maps}
\label{diff:sec:jvp_vjp}

The Jacobian matrix is useful as a representation of the partial derivatives.
However, the core idea underlying the definition of differentiable
functions, as well as their implementation in an autodiff 
framework, lies in the access to two key \textbf{linear maps}. These two maps
encode infinitesimal variations along \textbf{input} or \textbf{output}
directions and are referred to, respectively, as \textbf{Jacobian-vector
product} (JVP) and \textbf{Vector-jacobian product} (VJP).
This section formalizes these notions, in the context of Euclidean spaces.

\subsection{The need for linear maps}
\label{diff:sec:need_linear_maps}

So far, we have focused on functions $f \colon \RR^P \to \RR^M$, that take a
vector as input and produce a vector as output. However, functions that use
matrix or even tensor inputs/outputs are commonplace in neural networks.
For example, consider the function of matrices of the form $f(\W) \coloneqq \W \x$,
where $\x \in \RR^D$ and $\Wv \in \RR^{M\times D}$. This function takes a matrix
as input, not a vector. Of course, a matrix $\W \in \RR^{M\times D}$ can always
be ``flattened'' into a vector $\w \in \RR^{MD}$, by stacking the columns of
$\W$. We denote this operation by $\w = \Vec(\W)$ and its inverse by $\W =
\Vec^{-1}(\w)$. We can then equivalently write $f(\W)$ as $\tilde f(\w) =
f(\Vec^{-1}(\w)) = \Vec^{-1}(\w) \x$, so that the previous framework applies.
However, we will now see that this would be inefficient.

Indeed, the resulting Jacobian of $\tilde f$ at any $\w$ consists in a matrix
of size $\RR^{M \times MD}$, which, after some computations, can be observed to
be mostly filled with zeros. Getting the directional derivative of $f$ at $\W
\in \RR^{M\times D}$ in a direction $\V \in \RR^{M\times D}$ would consist in
(i) vectorizing $\V$ into $\v = \Vec(\V)$, (ii) computing the matrix-vector
product $\jac \tilde f(\w)\v $ at a cost of $M^3D^2$ computations (ignoring the
fact that the Jacobian has many zero entries), (iii) re-shaping the result into a
matrix. 

On the other hand, since $f$ is linear in its matrix input, we can infer that
the directional derivative of $f$
at any $\W \in \RR^{M\times D}$ in any direction $\V\in \RR^{M\times D}$ is
simply given by the function itself applied on $\V$. Namely, we have $\partial
f(\W)[\V] = f(\V) = \V\x$, which is simple to implement and clearly only
requires $MD$ operations. Note that the cost would have been the same, had we
ignored the non-zero entries of $\jac \tilde f(\w)$. The point here is that by
considering the operations associated to the differentiation of a function as
linear maps rather than using the associated representation as a
Jacobian matrix, we can efficiently exploit
the underlying input or output space structure. To that end, we now
recall the main abstractions necessary to extend the previous definitions in the
context of Euclidean spaces.

\subsection{Euclidean spaces}
\label{diff:sec:euc_spaces}

\textbf{Linear spaces}, \aka \textbf{vector spaces}, are spaces equipped with and
closed under an addition rule compatible with multiplication by a scalar (we
limit ourselves to the field of reals). Namely, in a linear space $\cE$, there
exist operations $+$ and $\cdot$, such that for any $\u$, $\v \in \cE$, and
$a \in \RR$, we have $\u +\v \in \cE$ and $a \cdot \u \in \cE$. 

\textbf{Euclidean spaces} are linear spaces
equipped with a basis $\e_1, \ldots, \e_P \in \cE$. Any element
$\v \in \cE$ can be decomposed as $\v = \sum_{i=1}^{P} v_i \e_i$ for some unique
scalars $v_1, \ldots, v_P \in \RR$. A canonical example of Euclidean space is
the set $\RR^P$ of all vectors of size $P$ that we already covered. The set of
matrices $\RR^{P_1 \times P_2}$ of size $P_1\times P_2$ is also naturally a
Euclidean space
generated by the set of canonical matrices $\E_{ij} \in \{0,1\}^{P_1\times P_2}$
for $i \in [P_1], j\in [P_2]$ filled with zero except at the $(i,
j)$\textsuperscript{th} entry filled with one. For example, $\W \in \RR^{P_1
\times P_2}$ can be written $\W = \sum_{i, j=1}^{P_1, P_2} W_{i j} \E_{ij}$.
Euclidean spaces are naturally equipped with a notion of inner product.
\begin{boxdef}{Inner product}\label{diff:def:inner_product}
  An \textbf{inner product} on a linear space $\cE$ is a function  $\langle \cdot, \cdot
  \rangle: \cE\times \cE \rightarrow \R$ that is 
  \begin{itemize}
    \item bilinear: $\x \mapsto
    \langle \x, \w \rangle$ and $\y \mapsto \langle \v, \y\rangle$ are linear for
    any $\w, \v \in \cE$,
    \item symmetric: $\langle \w, \v \rangle = \langle
      \v, \w\rangle$ for any $\w, \v \in \cE$,
    \item positive definite: $\langle \w,  \w\rangle\geq 0$ for any $\w
    \in \cE$, and $\langle \w, \w \rangle = 0$ if and only if $\w=0$.
  \end{itemize}
  An inner product defines a norm $\|\w\| \coloneqq \sqrt{\langle \w, \w
  \rangle}$. 
\end{boxdef}
The norm induced by an inner product defines a distance $\|\w-\v\|$ between $\w,
\v \in \cE$, and therefore a notion of convergence. 

For vectors, where $\cE = \RR^P$, the inner product is the usual one 
$
  \langle \w, \v\rangle = \sum_{i=1}^{P} w_iv_i.
$
For matrices, where $\cE=\RR^{P_1\times P_2}$, the inner product is the
so-called
Frobenius inner product.
It is defined for any 
$\W, \V \in \RR^{P_1 \times P_2}$ by
\[
\langle \W, \V \rangle 
\coloneqq \langle \Vec(\W), \Vec(\V) \rangle
= \sum_{i,j=1}^{P_1, P_2} W_{ij} V_{ij}
= \Tr(\W^\top \V),
\]
where $\Tr(\Z) \coloneqq \sum_{i=1}^P Z_{ii}$ is the trace operator defined for
square matrices $\Zv \in \RR^{P \times P}$. 
For tensors of order $R$, 
which generalize matrices to $\cE=\RR^{P_1\times \ldots \times P_R}$, the inner
product is defined similarly for  $\Wt, \Vt \in \RR^{P_1\times \ldots \times
P_R}$ by
\[
\langle \Wt, \Vt \rangle  
\coloneqq \langle \Vec(\Wt), \Vec(\Vt) \rangle
= \sum_{i_1,\ldots, i_R=1}^{P_1, \ldots, P_R} \Wt_{i_1\ldots i_R} \Vt_{i_1\ldots
i_R},  
\]
where $\Wt_{i_1\ldots i_R}$ is the $(i_1,\ldots, i_R)$\textsuperscript{th} entry
of $\Wt$.

\subsection{Linear maps and their adjoints}

The notion of linear map in \cref{diff:def:linear} naturally extends to
Euclidean spaces. Namely, a function $l:\cE\rightarrow \cF$ from a Euclidean
space $\cE$ onto a Euclidean space $\cF$ is a \textbf{linear map} if for any
$\w, \v\in \cE$ and $a, b \in \RR$, we have $l[a\w + b\v] = a \cdot l[\w] + b
\cdot l[\v]$. When $\cE = \RR^P$ and $\cF=\RR^M$, there always exists a matrix
$\A \in \RR^{M \times P}$ such that $l[\v] = \A \v$. Therefore, we can think of
$\A$ as the ``materialization'' as a matrix of $l$. 
\begin{boxexm}{Linear map} \label{diff:exm:linear_map} 
Consider the linear map $l[\v] \coloneqq (\a \b^\top)\v$, where $\a \in \RR^M$,
$\b \in \RR^P$ and $\v \in \RR^P$. This is a function from $\RR^P$ to $\RR^M$.
We can always materialize the linear map as a matrix $\A \coloneqq \a \b^\top
\in \RR^{M \times P}$ and write $l[\v] = \A \v$. However, applying a linear map
on a vector $\v$ often does not require materializing the corresponding matrix.
In this example, we can simply write $l[\v] = (\b^\top \v) \a$,
since $(\b^\top \v)$ is a scalar value.
This only requires an inner product and an element-wise multiplication,
and is more efficient than materializing $\A$ then computing $\A\v$,
which requires an outer product and a matrix-vector multiplication.
\end{boxexm}
We can define the adjoint operator of a linear map.
\begin{boxdef}{Adjoint operator}
  Given two Euclidean spaces $\cE$ and $\cF$ equipped with inner products
$\langle\cdot, \cdot \rangle_\cE$ and $\langle \cdot, \cdot \rangle_\cF$, the
\textbf{adjoint} of a linear map $l:\cE\rightarrow \cF$ is the unique linear map
$l^*: \cF\rightarrow \cE$ such that for any $\v \in \cE$ and $\u \in \cF$, 
\[
\langle l[\v], \u \rangle_\cF = \langle \v, l^*[\u] \rangle_\cE.     
\]
\end{boxdef}
The adjoint can be thought of as the counterpart of the matrix transpose for
linear maps. When $l[\v] = \A \v$, we have $l^*[\u] = \A^\top \u$ since
\begin{equation*}
\langle l[\v], \u \rangle_\cF 
= \langle \A \v, \u \rangle_\cF
= \langle \v, \Av^\top \u \rangle_\cE
= \langle \v, l^*[\u] \rangle_\cE.     
\end{equation*}

\begin{boxexm}{Adjoint linear map} \label{diff:exm:adjoint_map} 
Using the linear map $l[\v]$ from the previous example, 
we have for all $\u \in \RR^M$ and $\v \in \RR^P$,
\begin{equation*}
\langle \a \b^\top \v, \u \rangle
=
\langle \v, \b \a^\top \u \rangle.
\end{equation*}
Therefore, the adjoint linear map is $l^*[\u] = (\b \a^\top) \u$.
This is a function from $\RR^M$ to $\RR^P$.
It can be materialized as the matrix $\A^\top = \b \a^\top \in \RR^{P \times M}$.
Applying $l^*[\u]$ can be done efficiently as $l^*[\u] = (\a^\top \u) \b$.
\end{boxexm}

\subsection{Jacobian-vector products}

We now define the directional derivative using linear maps, leading to the
notion of Jacobian-vector product (JVP). This can be used to facilitate the
treatment of functions on tensors or for further extensions to
infinite-dimensional spaces. In the following, $\cE$ and $\cF$ denote two
Euclidean spaces equipped with inner products 
$\langle \cdot, \cdot\rangle_\cE$ and $\langle \cdot, \cdot \rangle_\cF$.
We start by defining differentiability in general Euclidean spaces.
\begin{boxdef}{Differentiability in Euclidean spaces}\label{diff:def:differentiability_euclidean}
  A function $f:\cE\rightarrow \cF$ is \textbf{differentiable} at
  a point $\w \in \cE$ if the \textbf{directional derivative} along $\v \in \cE$
  \[
    \partial f(\w)[\v] \coloneqq 
    \lim_{\delta \rightarrow 0} \frac{f(\w+ \delta \v) - f(\w)}{\delta}
  \]
  is well-defined for any $\v \in \cE$, linear in $\v$ and if 
  \[
    \lim_{\|\v\|_\cE \rightarrow 0} 
    \frac{\|f(\w+ \v) - f(\w) - \partial f(\w)[\v]\|_\cF}{\|\v\|_\cE} = 0.
  \]
\end{boxdef}
We can now formally define the Jacobian-vector product.
\begin{boxdef}{Jacobian-vector product}
\label{diff:def:jvp}
  For a differentiable function  $f: \cE \rightarrow \cF$, the \textbf{linear map}
  $\partial f(\w): \cE \to \cF$, mapping $\v$ to $\partial f(\w)[\v]$, is called
  the \textbf{Jacobian-vector product} (JVP).
  From this perspective, the function $\partial f$ is a function from $\cE$ to a
  linear map from $\cE$ to $\cF$.  That is, we have
  \begin{equation*}
  \partial f \colon \cE \to (\cE \to \cF).
  \end{equation*}
\end{boxdef} 
We emphasize again that the directional derivative $\partial f(\w)[\v] \in \cF$
is a value, while the JVP $\v \mapsto \partial f(\w)[\v]$ is a function.
Strictly speaking, $\v$ can belong to any Euclidean space $\cE$ and does not
need to be limited to a vector, as the JVP acronym would suggest.
We adopt the name JVP, as it is now standard.

\subsubsection*{Recovering the gradient}

Previously, we saw that for differentiable functions with vector input and
scalar output, the directional derivative is equal to the inner
product between the direction and the gradient. The same applies when
considering differentiable functions from a Euclidean space with single outputs,
except that the gradient is now an element of the input space and the inner
product is the one associated with the input space. 

\begin{boxprop}{Gradient}\label{diff:prop:grad_euclidean} 
  
  If a function $f:\cE \rightarrow \RR$ is differentiable at $\w \in \cE$, then
  there exists $\nabla f(\w) \in \cE$, called the \textbf{gradient} of $f$ at $\w$
  such that the directional derivative of $f$ at $\w$ along any input direction
  $\v \in \cE$ is given by 
    \[
      \partial f(\w)[\v] 
      = \langle \nabla f(\w), \v\rangle_\cE.
    \]
\end{boxprop}
In Euclidean spaces, the existence of the gradient can simply be shown by
decomposing the partial derivative along a basis of $\cE$. Such a definition
generalizes to infinite-dimensional (e.g., Hilbert spaces) spaces as discussed
in \cref{diff:sec:extension_non_euc}.

\subsection{Vector-Jacobian products}

Consider a function $f \colon \RR^P \to \RR^M$.
Instead of variations of $f$ along an \textbf{input} direction 
$\v \in \RR^P$, we may also consider the variations of $f$ along an 
\textbf{output} direction $\u \in \RR^M$, namely, computing the gradient 
$\nabla \langle \u, f \rangle(\w)$
of the scalar-valued function
\begin{equation*}
\langle \u, f \rangle(\w) \coloneqq \langle \u, f(\w) \rangle \in \RR.
\end{equation*}
Equivalently, we may compute the gradients $\nabla f_j(\w)$ of each
coordinate function $f_j \coloneqq \langle \ev_j, f \rangle$ at $\w$, 
where $\ev_j$ is the
$j$\textsuperscript{th} canonical vector in $\RR^M$. 
The infinitesimal variations of $f$ at $\w$ along any output direction
$\u= \sum_{j=1}^M u_j\ev_j \in \RR^M$ are given by 
\[
\nabla \langle \u, f \rangle(\w) 
= \sum_{j=1}^M u_j \nabla f_j(\w) = \jac f(\w)^\top \u \in \RR^P,
\]
where $\jac f(\w)^\top \in \RR^{P \times M}$ is the Jacobian's transpose.
Using the definition of derivative as a limit, we may also write
for $i \in [P]$
\begin{equation*}
\nabla_i \langle \u, f \rangle(\w) 
= [\jac f(\w)^\top \u]_i
= \lim_{\delta \rightarrow 0} \frac{\langle \u, f(\w + \delta \e_i)
-f(\w) \rangle}{\delta},
\end{equation*}
where $\ev_i$ is the $i$\textsuperscript{th} canonical vector in $\RR^P$. 

For generic Euclidean spaces $\cE$ and $\cF$, the counterpart of the transpose
is the adjoint operator, leading to the notion of vector-Jacobian product.
\begin{boxprop}{Vector-Jacobian product}\label{diff:prop:vjp}
  If a function $f:\cE \rightarrow \cF$ is differentiable at $\w \in \cE$, then its
  infinitesimal variation along an \textbf{output} direction $\u \in \cF$ is given by the
  \textbf{adjoint map}  $\partial f(\w)^* \colon \cF \to \cE$ of the JVP, called the
  \textbf{vector-Jacobian product} (VJP). It satisfies
  \begin{align*}
    \nabla \langle \u, f\rangle_\cF (\w) = \partial f(\w)^*[\u],
  \end{align*}
  where we denoted $\langle \u, f\rangle_\cF(\w) \coloneqq \langle \u,
  f(\w)\rangle_\cF$. 
  The function $\partial f(\cdot)^*$ is a function from $\cE$ 
  to a linear map from $\cF$ to $\cE$. That is, we have
  \begin{equation*}
  \partial
  f(\cdot)^* \colon \cE \to (\cF \to \cE).
  \end{equation*}
\end{boxprop}
\begin{proof}
  The chain rule presented in~\cref{diff:thm:chain_rule} naturally generalizes
  to Euclidean spaces (see \cref{diff:thm:chain_rule_jvp_vjp}). Since $\langle
  \u, \cdot\rangle_\cF$ is linear, its directional derivative is itself. Therefore,
  the
  directional derivative of $\langle \u, f\rangle_\cF$ is 
  \begin{align*}
    \partial (\langle \u, f\rangle_\cF)(\w)[\v] 
    & = \langle \u, \partial f(\w)[\v]\rangle_\cF \\
    & = \langle \partial f(\w)^*[\u], \v \rangle_\cE.
  \end{align*}
  As this is true for any $\v \in \cE$, $\partial f(\w)^*[\u]$ is the gradient
  of $\langle \u, f\rangle_\cF$ per \cref{diff:prop:grad_euclidean}.
\end{proof}

\begin{figure}[t]
\begin{center}
\includegraphics[width=0.70\linewidth]{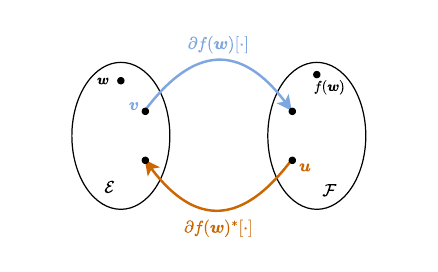}
\end{center}
\caption{Jacobian-vector product (JVP) $\v \mapsto \partial f(\w)[\v]$ and
vector-Jacobian product (VJP) $\u \mapsto \partial f(\w)^*[\u]$, seen as
linear maps.
\label{diff:fig:jvp_vjp_recap}}
\end{figure}

We illustrate the JVP and VJP linear maps in \cref{diff:fig:jvp_vjp_recap}.

\subsection{Chain rule using linear maps}

\begin{figure}[t]
\begin{center}
\includegraphics[width=0.90\linewidth]{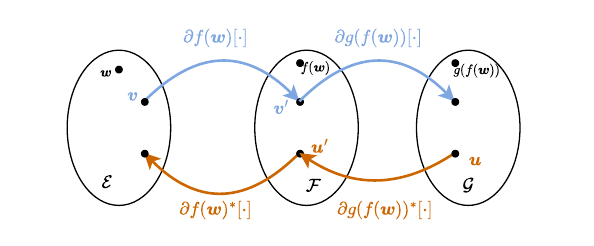}
\end{center}
\caption{Chain rule using JVP and VJP linear maps.
\label{diff:fig:chain_rule_recap}}
\end{figure}

The chain rule presented before in terms of Jacobian matrices can readily be
formulated to take advantage of the implementations of
the JVP and VJP as linear maps.
\begin{boxprop}{Chain rule, general case}\label{diff:thm:chain_rule_jvp_vjp} 
  Consider $f:\cE
  \rightarrow \cF$ and $g:\cF \rightarrow \cG$, where $\cE$, $\cF$ and $\cG$ are
  Euclidean spaces.
  If $f$ is differentiable at $\w \in \cE$ and $g$ is differentiable at
  $f(\w) \in \cF$, then the composition $g \circ f$ is differentiable at
  $\w \in \cE$. Its JVP is given for all $\v \in \cE$ by
  \[
    \partial (g\circ f)(\w)[\v] 
    = \partial g(f(\w))[\partial f(\w)[\v]]
  \]
  and its VJP is given for all $\u \in \cG$ by
  \[
	\partial (g\circ f)(\w)^*[\u] 
    = \partial f(\w)^*[\partial g(f(\w))^*[\u]].
  \]
\end{boxprop} 
The proof follows the one of~\cref{diff:thm:chain_rule}.
This is illustrated in \cref{diff:fig:chain_rule_recap}.
When the last function is scalar-valued, which is often the case in machine
learning, we obtain the following simplified result.
\begin{boxprop}{Chain rule, scalar case}\label{diff:thm:chain_rule_grad_vjp}
    Consider $f:\cE \rightarrow \cF$ and $g: \cF \rightarrow \RR$, 
    the gradient of the composition is given by 
  \[
    \nabla 	(g\circ f)(\w) = \partial f(\w)^* [\nabla g(f(\w))].
  \]
\end{boxprop} 

\subsection{Functions of multiple inputs (fan-in)}

Oftentimes, the inputs of a function do not belong to only one Euclidean space 
but to a product of them. 
An example is $f(\x, \W) \coloneqq \W \x$, which is defined on $\cE \coloneqq
\RR^D \times \RR^{M \times D}$. In such a case, it is convenient to generalize
the notion of partial derivatives to handle blocks of inputs. 

Consider a function $f(\w_1, \ldots, \w_S)$ defined on 
$\cE \coloneqq \cE_1 \times \ldots \times \cE_S$,
where $\w_i \in \cE_i$. 
We denote the partial derivative with respect to the $i$\textsuperscript{th} input
$\w_i$ along $\v_i \in \cE_i$ as 
$\partial_i f(\w_1, \ldots, \w_S)[\v_i]$. 
Equipped with this notation, we can analyze how JVPs
or VJPs are decomposed along several inputs.
\begin{boxprop}{Multiple inputs} 
\label{diff:prop:multiple_inputs}
Consider a differentiable function of the form
$f(\w) = f(\w_1, \dots, \w_S)$ 
with signature
$f \colon \cE \to \cF$,
where 
$\w \coloneqq (\w_1, \dots, \w_S) \in \cE$
and
$\cE \coloneqq \cE_1 \times \dots \times \cE_S$.
Then the JVP with the input direction 
$\v = (\v_1, \dots, \v_S) \in \cE$ 
is given by
\begin{align*}
\partial f(\w)[\v]
&= \partial f(\w_1, \dots, \w_S)[\v_1, \dots, \v_S] \in \cF \\
&= \sum_{i=1}^S \partial_i f(\w_1, \dots, \w_S)[\v_i].
\end{align*}
The VJP with the output direction $\u \in \cF$ is given by
\begin{align*}
\partial f(\w)^*[\u] 
&= \partial f(\w_1, \dots, \w_S)^*[\u] \in \cE \\
&= (\partial_1 f(\w_1, \dots, \w_S)^*[\u], \dots, 
\partial_S f(\w_1, \dots, \w_S)^*[\u]).
\end{align*}
\end{boxprop}

\begin{boxexm}{Matrix-vector product} \label{diff:exm:multiple_inputs} 
Consider $f(\x, \W) \coloneqq \W \x$, where $\W \in \RR^{M \times D}$ and $\x
\in \RR^D$. This corresponds to setting $\cE \coloneqq \cE_1 \times \cE_2
\coloneqq \RR^D \times \RR^{M \times D}$ and $\cF \coloneqq \RR^M$. For the JVP,
letting $\v \in \cE_1$ and $\V \in \cE_2$, we obtain
\begin{equation*}
\partial f(\x, \W)[\v, \V] = \W \v + \V \x \in \cF.
\end{equation*}
We can also access the individual JVPs as
\begin{equation*}
\begin{aligned}
    \partial_1 f(\x, \W)[\v] &= \W \v \in \cF, \\
    \partial_2 f(\x, \W)[\V] &= \V \x \in \cF.
\end{aligned}
\end{equation*}
For the VJP, letting $\u \in \cF$, we obtain
\begin{equation*}
\partial f(\x, \W)^*[\u] = (\W^\top \u, \u \x^\top)
\in \cE.
\end{equation*}
We can access the individual VJPs by
\begin{equation*}
\begin{aligned}
    \partial_1 f(\x, \W)^*[\u] &= \W^\top \u \in \cE_1, \\
    \partial_2 f(\x, \W)^*[\u] &= \u \x^\top \in \cE_2.
\end{aligned}
\end{equation*}
\end{boxexm}

\begin{boxrem}{Nested inputs} 
It is sometimes convenient to group inputs into meaningful parts.
For instance, if the input is naturally broken down into two parts $\x = (\x_1,
\x_2)$, where $\x_1$ is a text part and $\x_2$ is an image part, and the
network parameters are naturally grouped into three layers $\w = (\w_1, \w_2,
\w_3)$, we can write $f(\x, \w) = f((\x_1, \x_2), (\w_1, \w_2, \w_3))$.
This is mostly a convenience and we can again reduce it to a function of a
single input, thanks to the linear map perspective in Euclidean spaces.
\end{boxrem}

\begin{boxrem}{Hiding away inputs}
It will often be convenient to ignore inputs when differentiating. We use the
semicolon for this purpose. For instance, a function of the form $L(\w; \x, \y)$
(notice the semicolon) has signature $L \colon \cW \to \RR$ because we treat
$\x$ and $\y$ as constants. Therefore, the gradient is $\nabla L(\w; \x, \y) \in
\cW$. On the other hand, the function $L(\w, \x, \y)$ (notice the comma) has
signature $L \colon \cW \times \cX \times \cY \to \RR$ so its gradient is
$\nabla L(\w, \x, \y) \in \cW \times \cX \times \cY$.  If we need to
access partial gradients, we use indexing, e.g., $\nabla_1 L(\w, \x, \y) \in
\cW$ or $\nabla_\w L(\w, \x, \y) \in \cW$ when there is no ambiguity.
\end{boxrem}

\subsection{Functions with multiple outputs (fan-out)}

Similarly, it is often convenient to deal with functions that have multiple outputs.
\begin{boxprop}{Multiple outputs} 
\label{diff:prop:multiple_outputs}
Consider a differentiable function of the form 
$f(\w) \coloneqq (f_1(\w), \dots, f_T(\w))$,
with signatures 
$f \colon \cE \to \cF$
and
$f_i \colon \cE \to \cF_i$,
where 
$\cF \coloneqq \cF_1 \times \dots \times \cF_T$.
Then the JVP with the input direction $\v \in \cE$ 
is given by
\begin{align*}
\partial f(\w)[\v] = (\partial f_1(\w)[\v], \dots, \partial f_T(\w)[\v]) \in \cF.
\end{align*}
The VJP with the output direction $\u = (\u_1, \dots, \u_T) \in \cF$ is
\begin{align*}
\partial f(\w)^*[\u] 
&= \partial f(\w)^*[\u_1, \dots, \u_T] \in \cE \\
&= \sum_{i=1}^T \partial f_i(\w)^*[\u_i].
\end{align*}
\end{boxprop}
Combined with the chain rule, we obtain that the Jacobian of
\begin{equation*}
h(\w) \coloneqq g(f(\w)) = g(f_1(\w), \dots, f_T(\w))
\end{equation*}
is $\partial h(\w) 
=  \sum_{i=1}^T \partial_i g(f(\w)) \circ \partial f_i(\w)$ and
therefore the JVP is
\begin{equation*}
	\partial h(\w) [\v]
	=  \sum_{i=1}^T \partial_i g(f(\w)) [\partial f_i(\w)[\v]].
\end{equation*}

\subsection{Extensions to non-Euclidean linear spaces}
\label{diff:sec:extension_non_euc}

So far, we focused on Euclidean spaces, i.e., linear spaces with a finite basis.
However, the notions studied earlier can be generalized to more generic spaces. 

For example, \textbf{directional derivatives} (see
\cref{diff:def:differentiability_euclidean}) can be defined in any linear space
equipped with a norm and complete with respect to this norm. Such spaces are
called \textbf{Banach spaces}. Completeness is a technical assumption that
requires that any Cauchy sequence converges (a Cauchy sequence is a sequence
whose elements become arbitrarily close to each other as the sequence
progresses). A function $f: \cE \rightarrow \cF$ defined from a Banach space $\cE$
onto a Banach space $\cF$ is then called \textbf{Gateaux differentiable} if its
directional derivative is defined along any direction (where limits are defined
\wrt the norm in $\cF$). Some authors also require the directional derivative to
be linear to define a Gateaux differentiable function.

\textbf{Fr\'echet differentiability} can also naturally be generalized to Banach
spaces. The only difference is that, in generic Banach spaces, the linear map
$l$ satisfying \cref{diff:def:differentiability_euclidean} must be continuous,
i.e., there must exist $C>0$, such that $l[\v] \leq C \|\v\|$, where $\|\cdot\|$
is the norm in the Banach space $\cE$.

The definitions of gradient and VJPs require in addition a notion of
inner product. They can be defined in \textbf{Hilbert spaces}, that is, linear
spaces equipped with an inner product and complete with respect to the norm
induced by the inner product (they could also be defined in a Banach space by
considering operations in the dual space, see,
e.g.~\citep{clarke2008nonsmooth}). The existence of the gradient is ensured by
\textbf{Riesz's representation theorem} which states that any continuous linear
form in a Hilbert space can be represented by the inner product with a vector.
Since for a differentiable function $f: \cE \rightarrow \RR$, the JVP $\partial
f(\w): \cE \rightarrow \RR$ is a linear form, Riesz's representation theorem
ensures the existence of the gradient as the element $g \in \cE$ such that
$\partial f(\w) \v = \langle g, \v\rangle$ for any $\v \in \cE$. The VJP is also
well-defined as the adjoint of the JVP \wrt the inner product of the Hilbert
space.

As an example, the space of squared integrable functions on $\RR$ is a Hilbert
space equipped with the inner product
$
\langle a, b \rangle  \coloneqq \int a(x)b(x) dx.
$
Here, we cannot find a finite number of functions that can express all possible
functions on $\RR$. Therefore, this space is not a mere Euclidean space. Nevertheless,
we can consider functions on this Hilbert space (called \textbf{functionals} to
distinguish them from the elements of the space). The associated directional
derivatives and gradients, can be defined and are called respectively,
\textbf{functional derivative} and \textbf{functional gradient}, see, e.g.,
\citet{frigyik2008introduction} and references therein.

\section{Second-order differentiation}
\label{diff:sec:second_derivative}

\subsection{Second derivatives}

\begin{figure}
  \centering
  \includegraphics{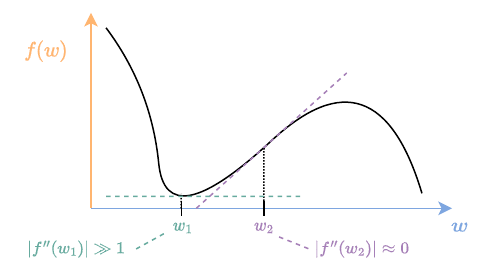}
  \caption{Points at which the second derivative is small are points along which
  the function is well approximated by its tangent line. On the other hand,
  points with large second derivative tend to be badly approximated by the
  tangent line. \label{higher:fig:second_der}}
\end{figure}

For a single-input, single-output differentiable function $f:\RR \rightarrow
\RR$, its derivative at any point is itself a function $f': \RR \rightarrow
\RR$. We may then consider the derivative of the derivative at any point: the
\textbf{second derivative}.
\begin{boxdef}{Second derivative}
  The \textbf{second derivative} $f^{(2)}(w)$ of a differentiable function $f:
  \RR \rightarrow \RR$ at $w \in \RR$ is defined as the derivative of $f'$ at
  $w$, that is, 
  \[
    f^{(2)}(w) 
    \coloneqq \lim_{\delta \rightarrow 0} \frac{f'(w+\delta) - f'(w)}{\delta},
  \]
  provided that the limit is well-defined. If the second derivative of a
  function $f$ is well-defined at $w$, the function is said to be \textbf{twice
  differentiable} at $w$.
  The second derivative is also denoted $f''$.
\end{boxdef}
If $f$ has a small second derivative at a given $w$, the derivative around $w$
is almost constant. That is, the function behaves like a line around $w$, as
illustrated in \cref{higher:fig:second_der}. Hence, the second derivative is
usually interpreted as the \textbf{curvature} of the function at a given point.

\subsection{Second directional derivatives}

For a multi-input function $f:\RR^P \rightarrow \RR$, we saw that the
directional derivative encodes infinitesimal variations of $f$ along a given
direction. To analyze the second derivative, the curvature of the
function at a given point $\w$, we can consider the variations along a pair of
directions, as defined below.
\begin{boxdef}{Second directional derivative}
  The \textbf{second directional derivative} of $f:\RR^P \rightarrow \RR$ at
  $\w\in \RR^P$ along $\v, \v' \in \RR^P$ is defined as the directional
  derivative of $\w \mapsto \partial f(\w)[\v]$ along $\v'$, that is,
  \[
    \partial^2 f(\w)[\v, \v'] 
    \coloneqq \lim_{\delta \rightarrow 0} 
    \frac{\partial f(\w + \delta \v')[\v] - \partial f(\w)[\v]}{\delta},
  \]
  provided that $\partial f(\w)[\v]$ is well-defined around $\w$ and that the
  limit exists.
\end{boxdef}

Of particular interest are the variations of a function around the canonical
directions: the \textbf{second partial derivatives}, defined as
\[
  \partial_{i j}^2 f(\w) \coloneqq \partial^2 f(\w)[\e_i, \e_j] 
\]
for $\e_i$, $\e_j$ the $i$\textsuperscript{th} and $j$\textsuperscript{th}
canonical directions in $\RR^P$, respectively. In Leibniz notation, the second
partial derivatives are denoted 
\begin{equation*}
\partial_{i j}^2 f(\w) = \frac{\partial^2
f(\w)}{\partial w_i \partial w_j}.
\end{equation*}

\subsection{Hessians}

For a multi-input function, twice differentiability is simply defined as the
differentiability of any directional derivative $\partial f(\w)[\v]$ \wrt $\w$. 
\begin{boxdef}{Twice differentiability}\label{diff:def:twice_differentiability}
  A function $f: \RR^P \rightarrow \RR$ is twice differentiable at $\w
  \in \RR^P$ if it is differentiable and $\partial f: \RR^P \rightarrow (\RR^P
  \rightarrow\RR)$ is also differentiable at $\w$.
\end{boxdef}
As a result, the
second directional derivative is a bilinear form.
\begin{boxdef}{Bilinear map, bilinear form}\label{diff:def:bilinear}
  A function $b: \RR^P \times \RR^P \rightarrow \RR^M$ is a \textbf{bilinear
  map} if $b[\v, \cdot]: \RR^P \rightarrow \RR$ is linear for any $\v$ and
  $b[\cdot, \v']$ is linear for any $\v'$. That is,
  \[
    b[\v, \v'] = \sum_{i=1}^P v_i b[\e_i, \v'] = \sum_{i=1}^{P}
    \sum_{j=1}^P v_iv_j' b[\e_i, \e_j],
  \]
  for $\v = \sum_{i=1}^P v_i \e_i$ and $\v' = \sum_{i=1}^P v_i' \e_i$. A
  bilinear map with values in $\RR$, $b: \RR^P \times \RR^P \rightarrow \RR$, is
  called a \textbf{bilinear form}.
\end{boxdef}
The second partial derivatives are gathered in the \textbf{Hessian}
and the second directional derivatives can be computed from it.
\begin{boxdef}{Hessian}
  The \textbf{Hessian} of a twice differentiable function $f:\RR^P \rightarrow
  \RR$ at $\w$ is the $P \times P$ matrix gathering all second partial
  derivatives,
  \[
    \nabla^2 f(\w) \coloneqq \begin{pmatrix}
      \partial_{11} f(\w) & \ldots & \partial_{1P} f(\w) \\
      \vdots & \ddots & \vdots \\
      \partial_{P1} f(\w) & \ldots & \partial_{PP} f(\w)
  \end{pmatrix} \in \RR^{P \times P},
  \]
  provided that all second partial derivatives are well-defined. 

  The second directional derivative at $\w$ is bilinear in any direction $\v =
  \sum_{i=1}^P v_i \e_i$ and $\v' = \sum_{i=1}^P
  v_i' \e_i$. Therefore,
  \[
    \partial^2 f(\w)[\v, \v'] 
    = \sum_{i, j = 1}^{P} v_i v_j'\partial^2 f(\w)[\e_i, \e_j]
    = \langle \v, \nabla^2 f(\w) \v'\rangle.
  \] 
\end{boxdef}
Given the gradient of $f$, the Hessian is equivalent to the transpose of the
Jacobian of the gradient. By slightly generalizing the notation $\nabla$ to
denote the transpose of the Jacobian of a function (which matches its definition
for single-output functions), we have that the Hessian can be expressed as
$\nabla^2 f(\w) = \nabla (\nabla f)(\w)$, which justifies its notation.

Similarly as for the differentiability of a function $f$, twice
differentiability of $f$ at $\w$ is equivalent to having the second partial
derivatives not only defined but also continuous in a neighborhood of $\w$.
Remarkably, by requiring twice differentiability, i.e., continuous second
partial derivatives, the Hessian is guaranteed to be symmetric
\citep{schwarz1873communication}.
\begin{boxprop}{Symmetry of the Hessian} \label{diff:thm:symmetry_hessian} 
  If a function $f:\RR^P \rightarrow \RR$ is twice differentiable at $\w$, then
  its Hessian $\nabla^2 f(\w)$ is symmetric, that is, $\partial_{i j}^2 f(\w) =
  \partial_{j i}^2 f(\w)$ for any $i, j \in \{1, \ldots P\}$. 
\label{diff:prop:symmetry_hessian}
\end{boxprop}
The symmetry of the Hessian means that it can alternatively be written as
$\nabla^2 f(\w) = (\partial_{ji}^2f(\w))_{i, j=1}^P = \jac (\nabla f)(\w)$, 
i.e.,
the Jacobian of the gradient of $f$.

\subsection{Hessian-vector products}

Similarly to the Jacobian, we can exploit the formal definition of the Hessian
as a bilinear form to extend its definition to Euclidean spaces. In particular,
we can define the notion of Hessian-vector product.
\begin{boxdef}{Hessian-vector product}\label{diff:def:hvp}
  If a function $f: \cE \rightarrow \R$ defined on a Euclidean space $\cE$ with
  inner product $\langle\cdot, \cdot\rangle$, is twice differentiable at $\w \in
  \cE$, then for any $\v \in \cE$, there exists $\v \mapsto \nabla^2 f(\w)[\v]$,
  called the \textbf{Hessian-vector product} (HVP) of $f$ at $\w$ along $\v$,
  such that for any $\v' \in \cE$,
  \[
    \partial^2 f(\w)[\v, \v'] = \langle \v', \nabla^2 f(\w)[\v]\rangle.
  \]
  In particular for $\cE = \RR^P$, the HVP is 
  $\nabla^2 f(\w)[\v]= (\partial^2 f(\w)[\v, \e_i])_{i=1}^P$.
\end{boxdef}
From an autodiff point of view, the HVP can be implemented in four different
ways, as explained in \cref{higher:sec:hvp}.

\subsection{Second-order Jacobians}

The previous definitions naturally extend to multi-output functions $f: \cE
\rightarrow \cF$, where $f \coloneqq (f_1, \dots, f_M)$, $f_j \colon \cE \to
\cF_j$ and $\cF \coloneqq \cF_1 \times \dots \times \cF_M$. 
The second directional derivative is defined by
gathering the second derivatives of each coordinate's function. 
That is, for $\w, \v, \v' \in \cE$,
\[
\partial^2 f(\w)[\v, \v'] = (\partial^2 f_j(\w)[\v, \v'])_{j=1}^M \in \cF.
\]
The function $f$ is twice differentiable if and only if all its coordinates are
twice differentiable. The second directional derivative is then a {\bf bilinear
map}. We can then compute second directional derivatives as
\[
  \partial^2 f(\w)[\v, \v'] 
  = \sum_{i,j=1}^P v_i v_j' \partial^2 f(\w)[\e_i, \e_j]
  = (\langle \v, \nabla^2 f_j(\w) \v'\rangle)_{j=1}^M.
\]
When $\cE = \RR^P$ and $\cF_j = \RR$, so that $\cF = \RR^M$,
the bilinear map can be materialized as a tensor
\[
\partial^2 f(\w) 
= (\partial^2 f(\w)[\e_i, \e_j])_{i,j=1}^P
\in \RR^{M \times P \times P},
\]
the ``second-order Jacobian'' of $f$.
However, similarly to the Hessian, it is usually more convenient to apply the
bilinear map to prescribed vectors $\v$ and $\v'$ than to materialize the second
partial derivatives as a tensor.

\section{Higher-order differentiation}
\label{diff:sec:higher}

\subsection{Higher-order derivatives}

Derivatives can be extended to any order. Formally, the $n$\textsuperscript{th}
derivative can be defined inductively as follows for a single-input,
single-output function.

\begin{boxdef}{$n$\textsuperscript{th} order derivative} The
  $n$\textsuperscript{th} derivative $f^{(n)}$ of a function $f:\RR
  \rightarrow \RR$ at $w \in \RR$ is defined as
  \[
    f^{(n)}(w) 
    \coloneqq (f^{(n-1)})'(w)
    = \lim_{\delta \rightarrow 0} \frac{f^{(n-1)}(w+\delta) - f^{(n-1)}(w)}{\delta}
  \]
  provided that $f^{(n-1)}$ is differentiable around $w$ and that the limit
  exists. In such a case, the function is said to be $n$ times differentiable
  at $w$.
\end{boxdef}

\subsection{Higher-order directional derivatives}

For a multi-input function $f$, we can naturally extend the notion of
directional derivative as follows.
\begin{boxdef}{$n$\textsuperscript{th} order directional derivative} The 
  $n$\textsuperscript{th} directional derivative of $f: \RR^P \rightarrow \R$
  at $\w \in \RR^P$ along $\v_1, \ldots, \v_n$ is defined as 
  \begin{align*}
    & \partial^n f(\w)[\v_1, \ldots, \v_n] \\
    & = \partial (\partial^{n-1}f(\w)[\v_1, \ldots, \v_{n-1}])[\v_n] \\
    & = \lim_{\delta \rightarrow 0} 
    \frac{
      \partial f(\w + \delta \v_n)[\v_1, \ldots, \v_{n-1}] 
      - \partial f(\w)[\v_1, \ldots, \v_{n-1}]
    }{
      \delta
    }
  \end{align*}
\end{boxdef}
A multi-input function $f$ is $n$-times differentiable if it is $n-1$
differentiable and its $n-1$ directional derivative along any direction is
differentiable. As a consequence the $n$\textsuperscript{th} directional
derivative is a \textbf{multilinear form}. 
\begin{boxdef}{Multilinear map, multilinear form}
A function $c: \otimes_{i=1}^n \RR^P
\rightarrow \RR^M$ is a {\bf multilinear map} if it is linear in each coordinate
given all others fixed, that is, if $\v_j \mapsto c[\v_1,\ldots,\v_j,\ldots,
\v_n]$ is linear in $\v_j$ for any $j \in [n]$. 
It is a \textbf{multilinear form} if it has values in $\RR$.
\end{boxdef}
The $n$\textsuperscript{th}
order directional derivative is then given by
\[
  \partial^n f(\w)[\v_1, \ldots, \v_n] 
  = \sum_{i_1, \ldots, i_n=1}^P v_{1,i_1} \ldots v_{n, i_n} 
  \partial^n f(\w)[\e_{i_1}, \ldots, \e_{i_n}].
\]
The $n$\textsuperscript{th} order partial derivatives can be materialized as an
$n$\textsuperscript{th} order tensor 
\[
\nabla^n f(\w) = (\partial^n f(\w)[\e_{i_1}, \ldots,
\e_{i_n}])_{i_1, \ldots, i_n=1}^P \in \RR^{P \times \ldots \times P}.
\]

\subsection{Higher-order Jacobians}

All above definitions extend directly to the case of multi-output functions
$f:\cE \rightarrow \cF$, where $\cF \coloneqq \cF_1 \times \dots \times \cF_M$.  
The $n$\textsuperscript{th} directional derivatives are then
\[
  \partial^n f(\w)[\v_1, \ldots, \v_n] 
  = (\partial^n f_j(\w)[\v_1, \ldots, \v_n])_{j=1}^M.
\]
The function $f$ is then $n$ times differentiable if it is $n-1$
differentiable and its $n-1$ directional derivative along any direction is
differentiable. As a consequence, the $n$\textsuperscript{th} directional
derivative is a {\bf multilinear map}. 
The $n$\textsuperscript{th} directional derivative can be decomposed into 
partial derivatives as
\[\partial^n
f(\w)[\v_1, \ldots, \v_n] = \sum_{i_1, \ldots, i_n=1}^P v_{1,i_1} \ldots v_{n,
i_n} \partial^n f(\w)[\e_{i_1}, \ldots, \e_{i_n}].
\]
When $\cE = \RR^P$ and $\cF = \RR^M$, the 
$n$\textsuperscript{th} order partial derivatives
can be materialized by an
$(n+1)$\textsuperscript{th} order tensor
\[
\partial^n f(\w) = (\partial^n f_j(\w)[\e_{i_1}, \ldots, \e_{i_n}])_{j=1, i_1,
\ldots, i_n=1}^{M, P, \ldots, P} \in
\RR^{M \times P \times \ldots \times P}.
\]

\subsection{Taylor expansions}
\label{diff:sec:taylor_exp}

With Landau's little $o$ notation, we have seen that if a function is
differentiable, it is approximated by a linear function in $\v$,
\[
  f(\w+\v) 
  = f(\w) 
  + \langle \nabla f(\w), \v\rangle
  + o(\|\v\|_2).
\]
Such an expansion of the function up to its first derivative is called the
\textbf{first-order Taylor expansion} of $f$ around $\w$.

If the function $f$ is twice differentiable, we can approximate it by a
quadratic in $\v$, leading to the \textbf{second-order Taylor expansion} of $f$
around $\w$,
\[
  f(\w+\v) 
  = f(\w) 
  + \langle \nabla f(\w), \v \rangle
  + \frac{1}{2}\langle \v, \nabla^2 f(\w) \v\rangle
  + o(\|\v\|_2^2).
\]
Compared to the first-order Taylor approximation, it is naturally more accurate
around $\w$, as reflected by the fact that $\|\v\|_2^3
\leq \|\v\|_2^2$ for $\|\v\|_2 \le 1$. 

More generally, we can build the \textbf{$\n$\textsuperscript{th} order Taylor
expansion} of a $n$ times differentiable function $f \colon \RR^P \to \RR^M$
around $\w \in \RR^P$ by
\begin{align*}
f(\w+\v) 
& = f(\w) 
+ \partial f(\w)[\v] 
+ \frac{1}{2} \partial^2 f(\w)[\v, \v] 
+ \ldots \\
& + \frac{1}{n!} \partial^n
f(\w)[\underbrace{\v, \ldots, \v}_{n \, \mathrm{times}}]
+ o(\|\v\|_2^{n}).
\end{align*}
Note that, using the change of variable $\w' = \w + \v \iff \v = \w'
- \w$, it is often convenient to write the $n$\textsuperscript{th} Taylor
expansion of $f(\w')$ around $\w$ as
\begin{equation*}
f(\w') 
= f(\w) + \sum_{j=1}^{n} \frac{1}{j!} \partial^j 
f(\w)[\underbrace{\w' - \w, \ldots, \w' - \w}_{j \, \mathrm{times}}]
+ o(\|\w' - \w\|_2^{n}).
\end{equation*}
Taylor expansions will prove useful in \cref{chap:finite_diff} for computing
derivatives by finite differences.

\section{Differential geometry}

In this chapter, we progressively generalized the notion of derivative from real
numbers to vectors and variables living in a linear space (\aka vector space),
either finite dimensional or infinite dimensional. We can further
generalize these notions by considering a local notion of linearity. This is
formalized by smooth manifolds in \textbf{differential geometry}, whose
terminology is commonly adopted in the automatic differentiation literature and
software. In this section, we give a
brief overview of derivatives on smooth manifolds (simply referred to as
manifolds), and refer to \citet{boumal2023introduction} for a complete
introduction.

\subsection{Differentiability on manifolds}

Essentially, a manifold is a set that can be locally approximated by a Euclidean
space. The most common example is a sphere like the Earth. Seen from the Moon,
the Earth is not a plane, but locally, at a human level, it can be seen as a
flat surface. Euclidean spaces are also trivial examples of manifolds. 
A formal
characterization of the sphere as a manifold is presented in
\cref{diff:exm:embedded_manifolds}. For now, we may think of a ``manifold'' as
some set (e.g., the sphere) contained in some ambient Euclidean space; note
however that manifolds can be defined generally without being contained in a
Euclidean space \citep[Chapter 8]{boumal2023introduction}.
Differentiability in manifolds is simply inherited from the notion of
differentiability in the ambient Euclidean space.
\begin{boxdef}{Differentiability of restricted functions}
  Let $\cM$ and $\cN$ be manifolds.
  A function $f: \mathcal{M} \rightarrow \mathcal{N}$ defined from $\mathcal{M}
  \subseteq \cE$ to $\mathcal{N} \subseteq \cF$, with $\cE$ and $\cF$ Euclidean spaces, 
  is differentiable if $f$ is the restriction of a differentiable function 
  $\bar f:\cE \rightarrow \cF$, so that $f$ coincides with $\bar f$ on
  $\mathcal{M}$.
\end{boxdef}
Our objective is to formalize the directional derivatives and gradients for
functions defined on manifolds. This formalization leads to the definitions of
tangent spaces and cotangent spaces, and the associated generalizations of JVP
and VJP operators as pushforward and pullback operators, respectively.

\subsection{Tangent spaces and pushforward operators}

\begin{figure}
  \centering
  \includegraphics[width=0.7\linewidth]{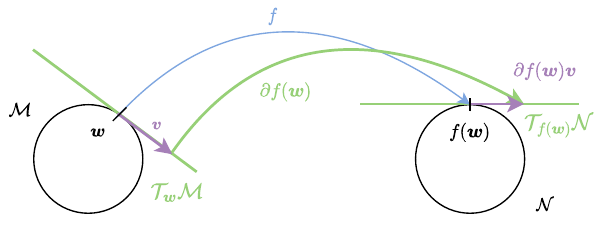}
  \caption{A differentiable function $f$ defined from a sphere $\cM$ to a
  sphere $\cN$ defines a push-forward operator that maps tangent vectors
  (derivatives of functions on the sphere passing through $\w$) in the tangent space
  $\cT_{\w} \cM$ to tangent vectors of $\cN$ at $f(\w)$
  in the tangent space $\cT_{f(\w)}\cN$.
    \label{diff:fig:sphere}
  }
\end{figure}

To generalize the notion of directional derivatives of a function $f$, the one
property we want to preserve is the chain rule. Rather than starting from the
variations of $f$ at a given point along a direction, we start with the
variations of $f$ along curves. Namely, on a manifold like the sphere
$\cS^P$ in $\RR^P$, we can look at curves $\alpha: \R \rightarrow
\cS^P$ passing through $\w \in \cS^P$ at time $0$, that is, $\alpha(0) =
\w$. 
For single-input functions like $\alpha$,
we denote for simplicity 
$\alpha'(0) \coloneqq (\alpha_1'(0), \ldots, \alpha_P'(0))$.
The directional derivative of a function $f$ must typically serve to define the
derivative of $f\circ \alpha$ at $0$, such that $(f\circ \alpha)'(0) =
\partial f(\w)[\alpha'(0)]$. 
In the case of the sphere, as illustrated in \cref{diff:fig:sphere}, the derivative
$\alpha'(0)$ of a curve $\alpha$ passing through a point $\w$ is always
\textbf{tangent} to the sphere at $\w$. The tangent plane to the sphere at $\w$
then captures all possible relevant vectors to pass to the JVP we are building.
To define the directional derivative of a function $f$ on a manifold, we
therefore restrict ourselves to an operator defined on the \textbf{tangent space}
$\cT_\w\cM$, whose definition below is simplified for our purposes.
\begin{boxdef}{Tangent space}\label{diff:def:tangent}
  The \textbf{tangent space} of a manifold $\cM$ at $\w \in \cM$ is
  defined as 
  \[
    \cT_\w \mathcal{M} \coloneqq \{\v = \alpha'(0) 
    \ \mbox{for any} \ 
    \alpha: \RR \rightarrow \cM
    \ \mbox{differentiable s.t.} \
    \alpha(0) = \w
    \}.
  \]
\end{boxdef}
In the case of the sphere in~\cref{diff:fig:sphere}, the tangent space is a
plane, that is, a Euclidean space. This property is generally true: tangent
spaces are Euclidean spaces, enabling us to define directional
derivatives as linear maps. Now, if $f$ is differentiable and goes
from a manifold $\cM$ to a manifold $\cN$, then $f\circ \alpha$ is a
differentiable curve in $\cN$. Therefore, $(f\circ \alpha)'(0)$ is the
derivative of a curve passing through $f(\w)$ at $0$ and is tangent to
$\cN$ at $f(\w)$. Hence, the directional derivative of $f: \cM
\rightarrow \cN$ at $\w$ can be defined as a function from the tangent space
$\cT_\w \mathcal{M}$ of $\cM$ at $\w$ onto the tangent space $\cT_{f(\w)}
\mathcal{N}$ of $\cN$ at $f(\w)$. Overall, we built the directional
derivative (JVP) by considering how a composition of $f$ with any curve
$\alpha$ pushes forward the derivative of $\alpha$ into the derivative of
$f\circ \alpha$. The resulting JVP is called a \textbf{pushforward
operator} in differentiable geometry.
\begin{boxdef}{Pushforward operator}\label{diff:def:push-forward}
  Given two manifolds $\cM$ and $\cN$, the \textbf{pushforward operator} of a
  differentiable function $f: \mathcal{M} \rightarrow \mathcal{N}$ at $\w \in
  \mathcal{M}$ is the linear map $\partial f(\w): \mathcal{T}_\w \mathcal{M} \rightarrow
  \mathcal{T}_{f(\w)} \mathcal{N}$ defined by 
  \[
    \partial f(\w)[\v] \coloneqq (f\circ \alpha)'(0),
  \]
  for any $\v \in \cT_\w \cM$ such that $\v = \alpha'(0)$,
  where $\alpha: \RR \rightarrow \cM$ is a differentiable curve passing
  through $\w$ at $0$, i.e., $\alpha(0) = \w$.
\end{boxdef}

\subsection{Cotangent spaces and pullback operators}

To generalize the JVP, we composed $f: \cM \rightarrow \cN$ with any
single-input function $\alpha: \RR\rightarrow \cM$ giving values on the
manifold. The derivative of any such $\alpha$ is then pushed forward from
$\cT_\w \mathcal{M}$ to $\cT_{f(\w)} \mathcal{N}$ by the action of $f$. 
To define the VJP, we take a symmetric approach. We consider all
single-output differentiable functions $\beta: \cN \rightarrow \RR$ defined on
$\y \in \cN$ with $\y=f(\w)$ for some $\w \in \cM$. We then want to pull
back the derivatives of $\beta$ when precomposing it by $f$. Therefore,
the space on which the VJP acts is the space of directional derivatives of any
$\beta: \cN \rightarrow \RR$ at $\y$, defining the \textbf{cotangent space}.
\begin{boxdef}{Cotangent space}\label{diff:def:cotangent}
  The \textbf{cotangent space} of a manifold $\cN$ at $\y \in \cN$ is defined as 
  \begin{align*}
    \cT_\y^* \mathcal{N} & = \{
      u = \partial \beta(\y)
    \ \mbox{for any} \ \beta: \cN \rightarrow \RR \ \mbox{differentiable}
    \} \\
    & = \{u: \cT_\y \mathcal{N} \rightarrow \RR \ \mbox{for any linear map} \ u \},
  \end{align*}
\end{boxdef}
Note that elements of the cotangent space are linear mappings, not vectors. This
distinction is important to define the pullback operator as an operator on
functions as done in measure theory. From a linear algebra viewpoint, the
cotangent space is exactly the \textbf{dual space} of $\cT_\y\mathcal{N}$, that
is, the set of linear maps from $\cT_\y \mathcal{N}$ to $\RR$, called
\textbf{linear forms}. As $\cT_\y \mathcal{N}$ is a Euclidean space, its dual
space $\cT_\y^* \mathcal{N}$ is also a Euclidean space. The \textbf{pullback}
operator is then defined as the operator that gives access to directional
derivatives of $\beta \circ f$ given the directional derivative of $\beta$
at $f(\w)$. 
\begin{boxdef}{Pullback operator}\label{diff:def:pull-back} Given two manifolds
  $\cM$ and $\cN$, the \textbf{pullback operator} of a differentiable function
  $f: \mathcal{M} \rightarrow \mathcal{N}$ at $\w \in \mathcal{M}$ is the
  linear map $\partial f(\w)^\star: \mathcal{T}_{f(\w)}^*\mathcal{N}
  \rightarrow \mathcal{T}_{\w}^* \mathcal{M}$ defined by 
  \[
    \partial f(\w)^\star u \coloneqq \partial (\beta \circ f)(\w),
  \]
  for any $u \in \cT_{f(\w)} \mathcal{N}^*$ such that $\partial \beta(
  f(\w)) = u$, for a differentiable function $\beta: \cN \rightarrow \RR$.
\end{boxdef}
Contrary to the pushforward operator that acts on vectors, the pullback operator
acts on linear forms. Hence, the slight difference in notation between $\partial
f(\w)^\star$ and $\partial f(\w)^*$, the adjoint operator of
$\partial f(\w)$. To properly define the adjoint operator $\partial 
f(\w)^*$, we need a notion of inner product. Since tangent spaces are Euclidean
spaces, we can define an inner product $\langle \cdot, \cdot \rangle_\w$ for
each $\cT_\w \cM$ and $\w\in \cM$, making $\cM$ a \textbf{Riemannian manifold}.
Equipped with these inner products, the cotangent space can be identified with
the tangent space, and we can define gradients.
\begin{boxdef}{Gradients in Riemannian manifolds}
  Let $\cM$ be a Riemannian manifold equipped with inner products $\langle
  \cdot, \cdot\rangle_\w $. For any \textbf{cotangent vector} $u \in \cT_\w^*
  \mathcal{M}$, with $\w \in \cM$, there exists a unique \textbf{tangent vector}
  $\u \in \cT_\w \mathcal{M}$ such that 
  \[
    \forall \v \in \cT_\w \mathcal{M}, \ u[\v] = \langle \u, \v\rangle_\w.
  \]
  In particular for any differentiable function $f: \cM \rightarrow \RR$, we can
  define the \textbf{gradient} of $f$ as the unique tangent vector $\nabla
  f(\w) \in \cT_\w \mathcal{M}$ such that 
  \[
    \forall \v \in \cT_\w \mathcal{M}, \ \partial f(\w)[\v] = \langle \nabla f(\w), \v\rangle.
  \]
\end{boxdef}
Therefore, rather than pulling back directional derivatives, we can pull back
gradients. The corresponding operator is then naturally the adjoint $\partial
f(\w)^*$ of the pushforward operator. Namely, given two Riemannian manifolds
$\cM$ and $\cN$, and a differentiable function $f: \mathcal{M} \rightarrow
\mathcal{N}$, we have
\[
    \left(\partial f(\w)^\star u\right)[\v] = \langle \partial f(\w)^*[\u], \v\rangle 
  \ \mbox{for any} \ \v \in \cT_\w \mathcal{M}
\]
for $u = \langle \cdot, \u\rangle \in \cT_{f(\w)}^* \mathcal{N}$ represented by $\u \in
\cT_{f(\w)} \mathcal{N}$.

\begin{table}
  \centering
  \begin{tabular}{c|c|c}
    Function & $f$ & $\cM \rightarrow \cN$ \\
    \midrule
    Push-forward & 
    $\partial f(\w)$ 
    & $\cT_\w\mathcal{M} \rightarrow \cT_{f(\w)} \mathcal{N}$ \\
    \midrule
    Pullback & 
    $\partial f(\w)^\star$ 
    & $\cT_{f(\w)}^*\mathcal{N} \rightarrow \cT_\w^* \mathcal{M}$  \\
    \midrule
    Adjoint of pushforward & 
    $\partial f(\w)^*$ 
    & $\cT_{f(\w)} \mathcal{N} \rightarrow \cT_\w \mathcal{M}$ 
  \end{tabular}
  \caption{For a differentiable function $f$ defined from a manifold $\cM$
  onto a manifold $\cN$, the JVP is generalized with the notion of pushforward
  $\partial f(\w)$. The counterpart of the pushforward is the pullback
  operation $\partial f(\w)^\star$ that acts on linear forms in the
  tangent spaces. For Riemannian manifolds, the pullback operation can be
  identified with the adjoint operator $\partial f(\w)^*$ of the
  pushforward operator as any linear form is represented by a vector.}
\end{table}

\begin{boxexm}{The sphere as a manifold}\label{diff:exm:embedded_manifolds}

The sphere $\cS^P$ in $\RR^P$ is defined as the set of points $\w\in \RR^P$,
satisfying $c(\w) \coloneqq \langle \w, \w \rangle - 1 = 0$, with JVP $\partial
c(\w)[\v] = 2\langle \w, \v\rangle$. 

Let us first understand why we may locally see the sphere $\cS^P$ as a space of
dimension $P-1$. Take the hemisphere $\cH = \{\w \in \cS^P: w_P \geq 0\}$, it
can be entirely described by using the first $P-1$ coordinates of each point,
the last one is simply given by $w_P= \sqrt{1- \sum_{i=1}^{P-1}w_i^2}$.
Formally, for any vector $\v = (v_1, \ldots, v_{P-1})$ of $P-1$ coordinates such
that $\sum_{i=1}^{P-1} v_i^2 \leq 1$, we can define $\psi_1(\v) \coloneqq
\sqrt{1- \sum_{i=1}^{P-1} v_i^2}$ and $\psiv(\v) \coloneqq (v_1, \ldots,
v_{P-1}, \psi_1(\v))$. We have then $\langle \psiv(\v), \psiv(\v)\rangle = 1$,
that is $c(\psiv(\v)) = 0$. We can then describe the hemisphere $\cH$ as the
image of the unit ball $B = \{\v = (v_1, \ldots, v_{P-1}): \sum_{i=1}^{P-1}
v_i^2 \leq 1\}$ in dimension $P-1$ through the mapping $\psi$, that is, $\cH =
\psi(B)$. In particular, any open set on the hemisphere can be described as the
mapping of an open set in $\RR^{P-1}$. The reasoning applies to all hemispheres
with different mappings, hence the local aspect of this viewpoint.

The tangent space can be naturally characterized in terms of the constraining
function $c$. Namely, the curve $\alpha: \RR \rightarrow \cS^P$ such that
$\alpha(0) =\w$ satisfies for any $\delta\in \RR$, $c(\alpha(\delta)) =
\zeros$.
Hence, differentiating the implicit equation, we have
\[
  (c\circ \alpha)'(0) = \partial c(\w)[\alpha'(0)].
\]
That is, $\alpha'(0)$ is in the null space of $\partial c(\w)$, denoted
\[
  \mathrm{Null}(\partial c(\w))
\coloneqq \{\v \in \RR^P: \partial c(\w)[\v] =0\}.
\]
The tangent space of $\cS^P$ at $\w$ is then 
\begin{align*}
  \cT_\w \mathcal{M} & = \mathrm{Null}(2 \langle \w, \cdot\rangle)  \\
  & = \{ \v \in \RR^P: \langle \w, \v\rangle = 0\}
\end{align*}
We naturally recover that the tangent space is a Euclidean space of dimension
$P-1$, defined as the set of points orthogonal to $\w$.
\end{boxexm}

\section{Generalized derivatives}

While we largely focus on differentiable functions in this book, it is important
to characterize non-differentiable functions. We distinguish here two cases:
continuous functions and non-continuous functions. For the former case, there
exist generalizations of the notion of directional derivative, gradient and
Jacobian, presented below. For non-continuous functions, even if derivatives
exist almost everywhere, they may be uninformative. For example,
piecewise-constant functions, encountered in e.g. control flows
(\cref{chap:cf}), are almost everywhere differentiable but with zero
derivatives. In such cases, surrogate functions can be defined to ensure the
differentiability of a program (\cref{part:smoothing}).

\subsection{Rademacher's theorem}
\label{diff:sec:rademacher}

We first recall the definition of (locally) Lipschitz continuous function.
\begin{boxdef}{(Locally) Lipschitz continuous function}\label{diff:def:local_lip}
  A function $f: \cE \rightarrow \cF$, is Lipschitz continuous if there exists
  $C\geq 0$ such that for any $\x, \y \in \cE$,
  \[
    \|f(\x)- f(\y)\| \leq C\|\x-\y\|.
  \]
  A function $f: \cE \rightarrow \cF$ is locally Lipschitz continuous if for any
  $\x \in \cE$, there exists a neighborhood $\cU$ of $\x$ such that $f$
  restricted to $\cU$ is Lipschitz continuous.
\end{boxdef}
Rademacher's theorem~\citep{rademacher1919partielle} 
then ensures that $f$ is differentiable almost everywhere.
\begin{boxprop}{Rademacher's theorem}\label{diff:thm:rademacher}
Let $\cE$ and $\cF$ denote Euclidean spaces. 
  If $f: \cE \rightarrow \cF$ is locally Lipschitz-continuous, then $f$ is almost
  everywhere differentiable, that is, the set of points in $\cE$ at which $f$ is
  not differentiable is of (Lebesgue) measure zero.
\end{boxprop}
See also~\citet{morrey2009multiple} for a standard proof.

\subsection{Clarke derivatives}

Rademacher's theorem hints that the definitions of directional derivatives,
gradients and Jacobians may be generalized to locally Lipschitz continuous
functions. This is what \citet{clarke1975generalized} did in his seminal work,
which laid the foundation of \textbf{nonsmooth analysis}. 
The first building block is a
notion of generalized directional derivative.
\begin{boxdef}{Clarke generalized directional
derivative}\label{diff:def:clarke_diff}
  The \textbf{Clarke generalized directional derivative} of a locally Lipschitz
  continuous function $f: \cE \rightarrow \RR$ at $\w \in \cE$ in the direction $\v
  \in \cE$ is 
  \[
    \partial_C f(\w)[\v] \coloneqq \limsup_{\substack{\u \rightarrow \w \\ \delta \searrow 0}} 
    \frac{f(\u + \delta \v) - f(\u)}{\delta},
  \]
  provided that the limit exists, where $\delta \searrow 0$ means that $\delta$
  approaches $0$ by non-negative values, and where the limit superior is defined as
  \[
    \limsup_{\x \rightarrow \a} f(\x)
    \coloneqq \lim_{\varepsilon \rightarrow 0}
    \sup\{f(\x) : \x \in B(\a, \varepsilon)\setminus\{\a\}\}
  \]
  for $B(\a, \varepsilon) \coloneqq \{\x \in \cE: \|\x-\a\|\leq \varepsilon \}$ the ball
  centered at $\a$ of radius $\varepsilon$.
\end{boxdef}
There are two differences with the usual definition of a directional derivative: (i)
we consider slopes of the function in a neighborhood of the point rather than
at the given point, (ii) we take a limit superior rather than a usual limit. The
first point is rather natural in the light of Rademacher's theorem: we can
properly characterize variations on points where the function is differentiable,
therefore we may take the limits of these slopes as a candidate slope for the point
of interest. The second point is more technical but essential: it allows us to
characterize the directional derivative as the supremum of some linear
forms~\citep{clarke2008nonsmooth}. These linear forms in turn define a set of
generalized gradients~\citep[Chapter 2]{clarke2008nonsmooth}.
\begin{boxdef}{Clarke generalized gradient}\label{diff:def:clarke_gradient}
  A \textbf{Clarke generalized gradient} of a locally Lipschitz function $f:
  \cE\rightarrow \RR$ at $\w \in \cE$ is a point $\g \in \cE$ such that 
$\forall \v \in \cE$
  \[
    \partial_C f(\w)[\v] \geq \langle \g, \v \rangle.
  \]
  The set of Clarke generalized gradients is called the \textbf{Clarke
  subdifferential} of $f$ at $\w$.
\end{boxdef}
\cref{diff:def:clarke_diff} and~\cref{diff:def:clarke_gradient} can be used in
non-Euclidean spaces, such as Banach or Hilbert
spaces~\citep{clarke2008nonsmooth}. In Euclidean spaces, the Clarke generalized
gradients can be characterized more simply thanks to Rademacher's
theorem~\citep[Theorem 8.1]{clarke2008nonsmooth}. Namely, as shown below, they can be defined as
a convex combination of limits of gradients of $f$ evaluated at a sequence in
$\cE\setminus \Omega$ that converges to $\w$. 
\begin{boxprop}{Characterization of Clarke generalized gradients}
  \label{diff:prop:clarke_gradient} Let $f: \cE\rightarrow \RR$ be a locally
  Lipschitz continuous and denote $\Omega$ the set of points at which $f$ is not
  differentiable (\cref{diff:thm:rademacher}). An element $\g \in \cE$ is a Clarke
  generalized gradient of $f$ at $\w \in \cE$ if and only if 
  \begin{align*}
    \g & \in \conv\left(\left\{
      \lim_{n \rightarrow +\infty} \nabla f(\v_n) \colon
      (\v_n)_{n=1}^{+\infty} \ \mbox{s.t.} \ \v_n \in \cE\setminus \Omega,  \ 
      \v_n  \underset{n \rightarrow + \infty}{\rightarrow} \w
    \right\}\right).
  \end{align*}
\end{boxprop}
In the above, the convex hull of a set $S\subseteq \cE$, the set of
convex combinations of elements of $S$, is denoted
\begin{equation*}
  \conv(S) \coloneqq \{\lambda_1 \s_1 + \ldots + \lambda_m \s_m \
  \colon m \in \NN, \ \lambda_i \geq 0, \sum_{i=1}^{m} \lambda_i = 1, \s_i
  \in S\}.
\end{equation*}

The Jacobian of a function $f: \cE \rightarrow \cF$ between two Euclidean spaces
can be generalized similarly~\citep[Section 3.3]{clarke2008nonsmooth}.
\begin{boxdef}{Clarke generalized Jacobian}
  Let $f: \cE\rightarrow \cF$ be locally Lipschitz continuous and denote $\Omega$
  the set of points at which $f$ is not differentiable
  (\cref{diff:thm:rademacher}). A \textbf{Clarke generalized Jacobian} of $f$
  at $\w \in \cE$ is an element $\mathbf{J}$ of
  \[
    \conv\left(\left\{
      \lim_{n \rightarrow +\infty} \jac f(\v_n) \colon
      (\v_n)_{n=1}^{+\infty} \ \mbox{s.t.} \ \v_n \in \cE\setminus \Omega,  \ 
      \v_n  \underset{n \rightarrow + \infty}{\rightarrow} \w
    \right\}\right).
  \]
\end{boxdef}
For a continuously differentiable function $f: \cE \rightarrow \cF$ or $f:\cE
\rightarrow \RR$, there is a unique generalized gradient, recovering the usual
gradient~\citep[Proposition 3.1, page 78]{clarke2008nonsmooth}. The chain rule
can be generalized to these objects~\citep{clarke2008nonsmooth}.
Recently,~\citet{bolte2020mathematical, bolte2022complexity} further generalized
Clarke gradients through the definition of conservative gradients to define
automatic differentiation schemes for nonsmooth functions.

\section{Summary}

\begin{itemize}

\item The usual definition of \textbf{derivatives} of real-valued univariate functions
extends to multivariate functions $f \colon \RR^P \to \RR$ through the notion of
\textbf{directional derivative} $\partial f(\w)[\v]$ at $\w \in \RR^P$ in the
direction $\v \in \RR^P$.  

\item To take advantage of the representation of $\w =
\sum_{j=1}^P w_j \e_j$ using the canonical bases $\{\e_1, \dots, \e_P\}$, the
definition of \textbf{differentiable} functions requires the \textbf{linearity}
of the directional derivative \wrt the direction $\v$. 

\item This requirement gives
rise to the notion of \textbf{gradient} $\nabla f(\w) \in \RR^P$, the vector
that gathers the partial derivatives and further defines the \textbf{steepest
ascent direction} at $\w$.  

\item For vector-input vector-output functions $f \colon
\RR^P \to \RR^M$, the directional derivative leads to the definition of
\textbf{Jacobian matrix} $\jac f(\w) \in \RR^{M \times P}$, the matrix which
gathers all partial derivatives (notice that we use bold $\jac$). 
The \textbf{chain rule} is then the \textbf{product} of Jacobian matrices.

\item These notions can be extended to general Euclidean spaces, such as the spaces of
matrices or tensors.  For functions of the form $f \colon \cE \to \R$, the
gradient is $\nabla f(\w) \in \cE$.  More generally, for functions of the form $f
\colon \cE \to \cF$, the Jacobian $\partial f(\w)$ can be seen as a \textbf{linear
map} (notice the non-bold $\partial$).  The directional derivative at $\w \in \cE$
naturally defines a linear map $l[\vv] = \partial f(\w)[\vv]$, where
$\partial f(\w) \colon \cE \to \cF$ is called the \textbf{Jacobian vector product}
(JVP) and captures the infinitesimal variation at $\w \in \cE$ along the
\textbf{input} direction $\v \in \cE$.

\item Its \textbf{adjoint} $\partial f(\w)^* \colon \cF \to \cE$ defines another linear
map $l[\uv] = \partial f(\w)^*[\uv]$ called the \textbf{vector Jacobian
product} (VJP) and captures the infinitesimal variation at $\w \in \cE$ along the
\textbf{output} direction $\u \in \cF$.  The \textbf{chain rule} is then the
\textbf{composition} of these linear maps.  

\item For the particular case when we
compose a scalar-valued function $\ell$ (such as a loss function) with a
vector-valued function $f$ (such as a network function), the gradient is given
by $\nabla (\ell \circ f)(\w) = \partial f(\w)^* \nabla \ell(f(\w))$.  This
is why being able to apply the adjoint to a gradient, which as we shall see can
be done with reverse-mode autodiff, is so pervasive in machine learning.

\item The definitions of JVP and VJP operators can further be generalized in the
context of differentiable geometry. In that framework, the JVP amounts to the
\textbf{pushforward} operator that acts on \textbf{tangent vectors}. The VJP
amounts to the \textbf{pullback} operator that acts on \textbf{cotangent
vectors}.

\item We also saw that the \textbf{Hessian matrix} of a function $f(\w)$ from $\RR^P$
to $\RR$ is denoted $\nabla^2 f(\w) \in \RR^{P \times P}$. 
It is symmetric if the second partial
derivatives are continuous. Seen as linear map, the Hessian leads to the notion
of \textbf{Hessian-vector product} (HVP), which we saw can be reduced to the JVP
or the VJP of $\nabla f(\w)$.

\item The main take-away message of this chapter is that computing the directional
derivative or the gradient of compositions of functions \textbf{does not}
require computing intermediate Jacobians but only to evaluate linear maps (JVPs
or VJPs) associated with these intermediate functions.  The goal of automatic
differentiation, presented in~\cref{chap:auto_diff}, is precisely to provide an
efficient implementation of these maps for \textbf{computation chains} or
more generally for \textbf{computation graphs}.

\end{itemize}

%% file: chapters/proba_learning/proba_learning.tex
\chapter{Probabilistic learning}
\label{chap:proba_learn}

In this chapter,
we review how to perform probabilistic learning.
We also introduce exponential family distributions,
as they play a key role in this book.

\section{Probability distributions}\label{proba_learn:sec:dist}

\subsection{Discrete probability distributions}

A discrete probability distribution over a set $\cY$ 
is specified by its
\textbf{probability mass function} (PMF) $p \colon \cY \to [0,1]$.
The probability of $\y \in \cY$ is then defined by
\begin{equation*}
\PP(Y = \y) \coloneqq p(\y),
\end{equation*}
where $Y$ denotes a random variable.
When $Y$ follows a distribution $p$, we write $Y \sim p$
(with some abuse of notation, we use the same letter $p$ to denote the
distribution and the PMF).
The \textbf{expectation} of $\phi(Y)$, where $Y \sim p$ and
$\phi \colon \cY \to \RR^M$, is then
\begin{equation*}
\EE[\phi(Y)] = \sum_{\y \in \cY} p(\y) \phi(\y),
\end{equation*}
its \textbf{variance} (for one-dimensional variables) is
\begin{equation*}
\VV[\phi(Y)] 
= \EE[(\phi(Y) - \EE[\phi(Y)])^2] 
= \sum_{\y \in \cY} p(\y) (\phi(\y) - \EE[\phi(Y)])^2
\end{equation*}
and its \textbf{mode} is
\begin{equation*}
\argmax_{\y \in \cY} p(\y). 
\end{equation*}
The \textbf{Kullback-Leibler} (KL) divergence (also known as relative
entropy) between two discrete distributions over $\cY$, with
associated PMFs $p$ and $q$, is the statistical ``distance'' defined by
\begin{equation*}
\mathrm{KL}(p, q)
\coloneqq \sum_{\y \in \cY} p(\y) \log\left(\frac{p(\y)}{q(\y)}\right)
= \EE_{Y \sim p} \log\left(\frac{p(Y)}{q(Y)}\right).
\end{equation*}

\subsection{Continuous probability distributions}

A continuous probability distribution over $\cY$ is specified by its
\textbf{probability density function} (PDF) $p \colon \cY \to \RR_+$. The
probability of 
$\cA \subseteq \cY$ is then
\begin{equation*}
\PP(Y \in \cA) = \int_\cA p(y) dy.
\end{equation*}
The expectation, variance and KL divergence are defined
analogously to the discrete setting, simply replacing $\sum_{\y \in \cY}$ with
$\int_{\cY}$. Specifically, the expectation of $\phi(Y)$ is
\begin{equation*}
\EE[\phi(Y)] = \int_{\cY} p(y) \phi(y) dy,
\end{equation*}
the variance is
\begin{equation*}
\VV[\phi(Y)] 
= \EE[(\phi(Y) - \EE[\phi(Y)])^2] 
= \int_{\cY} p(y) (\phi(y) - \EE[\phi(Y)])^2 dy
\end{equation*}
and the KL divergence is
\begin{equation*}
\mathrm{KL}(p, q)
\coloneqq \int_\cY p(y) \log\left(\frac{p(y)}{q(y)}\right) dy
= \EE_{Y \sim p} \log\left(\frac{p(Y)}{q(Y)}\right).
\end{equation*}
The mode is defined as the argmax of the PDF.

When $\cY = \RR$, we can also define the \textbf{cumulative distribution
function} (CDF)
\begin{equation*}
\PP(Y \le b) = \int_{-\infty}^b p(y) dy.
\end{equation*}
The probability of $Y$ lying in the semi-closed interval $(a, b]$ is
then
\begin{equation*}
\PP(a < Y \le b) = \PP(Y \le b) - \PP(Y \le a).
\end{equation*}

\section{Maximum likelihood estimation}
\label{proba_learn:sec:mle}

\subsection{Negative log-likelihood}

We saw that a probability distribution over $\cY$ is specified by
$p(\y)$, which is called the probability mass function (PMF) for
discrete variables or the probability density function (PDF) for continuous
variables. In practice, the true distribution $p$ generating the data is unknown
and we wish to approximate it with a distribution $p_\lambdav$, with parameters
$\lambdav \in \Lambda$.
Given a finite set of \iid observations $\y_1, \dots, \y_N$,
how do we fit $\lambdav \in \Lambda$ to the data?
This can be done by maximizing the \textbf{likelihood} of the data, i.e.,
we seek to solve
\begin{equation*}
\widehat \lambdav_N \coloneqq
\argmax_{\lambdav \in \Lambda} \prod_{i=1}^N p_{\lambdav}(\y_i).
\end{equation*}
This is known as \textbf{maximum likelihood estimation} (MLE).
Because the $\log$ function is monotonically increasing,
this is equivalent to minimizing the \textbf{negative log-likelihood},
i.e., we have
\begin{equation*}
\widehat \lambdav_N
= \argmin_{\lambdav \in \Lambda} -\sum_{i=1}^N \log p_{\lambdav}(\y_i).
\end{equation*}

\begin{boxexm}{MLE for the normal distribution}
Suppose we set $p_\lambdav$ to the normal distribution with parameters
$\lambdav = (\mu, \sigma)$, i.e.,
\begin{equation*}
p_\lambdav(y) \coloneqq \frac{1}{\sigma \sqrt{2 \pi}} 
\exp\left(-\frac{1}{2} \left(\frac{y - \mu}{\sigma}\right)^2\right).
\end{equation*}
Then, given observations $y_1, \dots, y_N$, 
the MLEs for $\mu$ and $\sigma^2$ are the sample mean and the sample
variance, respectively.
\end{boxexm}

\subsection{Consistency \wrt the Kullback-Leibler divergence}

It is well-known that the MLE is \textbf{consistent},
in the sense of the Kullback-Leibler divergence.
That is, denoting the true distribution $p$ and
\begin{equation*}
\lambdav_\infty \coloneqq 
\argmin_{\lambdav \in \Lambda} \mathrm{KL}(p, p_\lambdav)
\end{equation*}
where
\begin{equation*}
\mathrm{KL}(p, p_\lambdav)
=
\EE_{Y \sim p} \log\left(\frac{p(Y)}{p_\lambdav(Y)}\right),
\end{equation*}
then $\widehat \lambdav_N \to \lambdav_\infty$ in expectation over the
observations, as $N \to \infty$. This can be seen by using
\begin{equation*}
\mathrm{KL}(p, p_\lambdav)
\approx
\frac{1}{N} \sum_{i=1}^N \log\left(\frac{p(\y_i)}{p_\lambdav(\y_i)}\right)
=
\frac{1}{N} \sum_{i=1}^N \left(\log p(\y_i) - \log p_\lambdav(\y_i)\right)
\end{equation*}
and the law of large numbers.

\section{Probabilistic supervised learning}
\label{proba_learn:sec:cpd}

\subsection{Conditional probability distributions}
\label{proba_learn:sec:conditional_proba}

Many times in machine learning, instead of a probability $\PP(Y = \y)$ for some
$\y \in \cY$,
we wish to define a conditional probability $\PP(Y = \y | X = \x)$,
for some input $\x \in \cX$.
This can be achieved by reduction to an unconditional probability distribution,
\begin{equation*}
\PP(Y = \y | X = \x) \coloneqq p_\lambdav(\y)
\end{equation*}
where
\begin{equation*}
\lambdav \coloneqq f(\x, \w)
\end{equation*}
and $f$ is a \textbf{model function} with \textbf{model parameters} 
$\w \in \cW$.
That is, rather than being a deterministic function from $\cX$ to $\cY$,
$f$ is a function from $\cX$ to $\Lambda$, the set
of permissible \textbf{distribution parameters}
of the output distribution associated with the input.

We emphasize that $\lambdav$ could be a single parameter or a collection of
parameters. For instance, in the Bernoulli distribution, $\lambda = \pi$,
while in the univariate normal distribution, $\lambdav = (\mu, \sigma)$.

In \cref{proba_learn:sec:exp_family} and throughout this book,
we will also use the notation $p_\thetav$ instead of $p_\lambdav$
when $\thetav$ are the canonical parameters of an exponential family
distribution.

\subsection{Inference}

The main advantage of this probabilistic approach is that our prediction model
is much richer than if we just learned a function from $\cX$ to $\cY$.  
We now have access to the whole distribution over possible
outcomes in $\cY$ and can compute various statistics:
\begin{itemize}
\item Probability: $\PP(Y = \y | X = \x)$ 
    or $\PP(Y \in \cA | X=\x)$,

\item Expectation: $\EE[\phi(Y)|X=\x]$ for some function $\phi$,

\item Variance: $\VV[\phi(Y)|X=\x]$,

\item Mode: $\argmax_{\y \in \cY} p_\lambdav(\y)$.

\end{itemize}
We now review probability distributions useful for 
binary classification,
multiclass classification,
regression,
multivariate regression,
and integer regression.
In the following, 
to make the notation more lightweight,
we omit the dependence on $\x$.

\subsection{Binary classification}

For \textbf{binary outcomes}, where $\cY = \{0,1\}$, 
we can use a \textbf{Bernoulli distribution} with parameter
\begin{equation*}
\lambda \coloneqq \pi \in [0,1].
\end{equation*}
When a random variable $Y$ is distributed according to a Bernoulli distribution
with parameter $\pi$, we write
\begin{equation*}
Y \sim \mathrm{Bernoulli}(\pi). 
\end{equation*}
The PMF of this distribution is
\begin{equation*}
p_\pi(y) \coloneqq
\begin{cases}
\pi &\mbox{if } y = 1 \\
1 - \pi &\mbox{if } y = 0
\end{cases}.
\end{equation*}
The Bernoulli distribution is a \textbf{binomial distribution}
with a single trial.
Since $y \in \{0, 1\}$, the PMF can be rewritten as
\begin{equation*}
p_\pi(y) = \pi^y (1-\pi)^{1 -y}.
\end{equation*}
The mean is 
\begin{equation*}
\EE[Y] = \pi = \PP(Y = 1)
\end{equation*}
and the variance is
\begin{equation*}
\VV[Y] = \pi (1-\pi) = \PP(Y=1)\PP(Y=0).
\end{equation*}

\begin{figure*}
\centering
\includegraphics[scale=0.32]{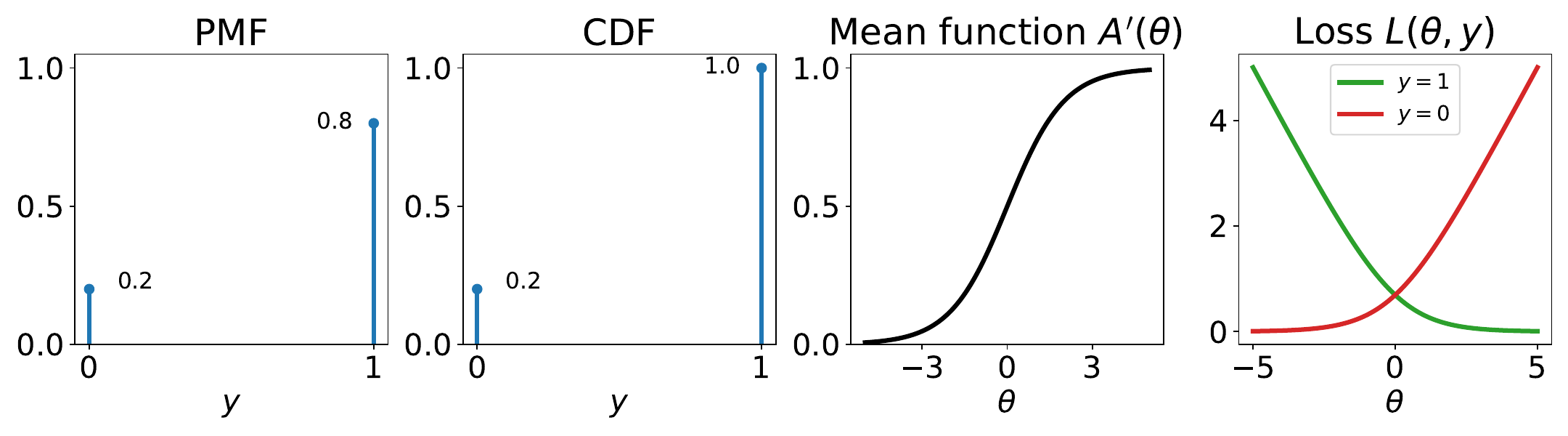}
\caption{
The {\bf Bernoulli distribution}, 
whose PMF and CDF are here illustrated with parameter $\pi = 0.8$.
Its mean function is 
$\pi = A'(\theta) = \mathrm{logistic}(\theta) = \frac{1}{1 + \exp(-\theta)}$, 
where $\theta$ is for instance the output of a neural network.
The negative log-likelihood leads to the {\bf logistic loss},
$L(\theta, y) = \mathrm{softplus}(\theta) - \theta y = \log (1 +
\exp(\theta)) - \theta y$.
The loss curve is shown for $y \in \{0,1\}$.
}
\label{proba_learn:fig:bernoulli}
\end{figure*}

\subsubsection*{Parameterization using a sigmoid}

Since the parameter $\pi$ of a Bernoulli distribution needs to belong to
$[0,1]$, we typically use a \textbf{sigmoid function} 
(\cref{neural_nets:sec:sigmoids}), such as a
\textbf{logistic function} as the output layer:
\begin{equation*}
\pi \coloneqq f(\x, \w) \coloneqq \logistic(g(\x, \w)),
\end{equation*}
where $g \colon \cX \times \cW \to \RR$ is for example a neural network 
and 
\begin{equation*}
\mathrm{logistic}(a) \coloneqq \frac{1}{1 + \exp(-a)} \in (0,1).
\end{equation*}
When $g$ is linear in $\w$, 
this is known as \textbf{binary logistic regression}.

\begin{boxrem}{Link with the logistic distribution}\label{proba_learn:rem:logistic}
The logistic distribution with mean and scale parameters $\mu$ and $\sigma$ 
is a \textbf{continuous} probability distribution with PDF
\begin{equation*}
p_{\mu,\sigma}(u) \coloneqq \frac{1}{\sigma} p_{0,1}\left(\frac{u - \mu}{\sigma}\right)
\end{equation*}
where
\begin{equation*}
p_{0,1}(z) \coloneqq \frac{\exp(-z)}{(1 + \exp(-z))^2}.
\end{equation*}
If a random variable $U$ follows a logistic distribution with parameters $\mu$
and $\sigma$, we write $U \sim \mathrm{Logistic}(\mu, \sigma)$.
The CDF of $U \sim \mathrm{Logistic}(\mu, \sigma)$ is 
\begin{equation*}
\PP(U \le u) 
= \int_{-\infty}^u p_{\mu,\sigma}(t) dt
= \mathrm{logistic}\left(\frac{u - \mu}{\sigma}\right).
\end{equation*}
Therefore, if 
\begin{equation*}
U \sim \mathrm{Logistic}(\mu, \sigma)
\end{equation*}
and
\begin{equation*}
Y \sim \mathrm{Bernoulli}\left(\logistic\left(\frac{u -
    \mu}{\sigma}\right)\right),
\end{equation*}
then
\begin{equation*}
\PP(Y = 1)  = \PP(U \le u).
\end{equation*}
Here, $U$ can be interpreted as a latent continuous variable
and $u$ as a threshold.
\label{proba_learn:rem:link_with_logistic_dist}
\end{boxrem}

\subsection{Multiclass classification}

For \textbf{categorical outcomes} with $M$ possible choices,
where $\cY = [M]$,
we can use a \textbf{categorical distribution} with
parameters
\begin{equation*}
\lambdav \coloneqq \piv \in \triangle^M, 
\end{equation*}
where we define the probability simplex
\begin{equation*}
\triangle^M 
\coloneqq \{\piv \in \RR_+^M \colon \langle \piv, \ones \rangle = 1 \},
\end{equation*}
i.e., the set of valid discrete probability distributions.
When $Y$ follows a categorical distribution with parameter $\piv$,
we write
\begin{equation*}
Y \sim \mathrm{Categorical}(\piv).
\end{equation*}
The PMF of the categorical distribution is
\begin{equation*}
p_\piv(y) \coloneqq \langle \piv, \phi(y) \rangle = \pi_y,
\end{equation*}
where 
\begin{equation*}
\phi(y) \coloneqq \e_y
\end{equation*}
is the standard basis vector for the coordinate $y \in [M]$.

Since $Y$ is a categorical variable, it does not make sense to compute the
expectation of $Y$ but we can compute that of $\phi(Y) = \e_Y$,
\begin{equation*}
\EE_{Y \sim p_\piv}[\phi(Y)] = \piv.
\end{equation*}
Therefore,
as was also the case for the Bernoulli distribution,
the mean and the probability distribution (represented by the vector
$\piv$) are the same in this case.

\begin{figure*}
\centering
\includegraphics[scale=0.32]{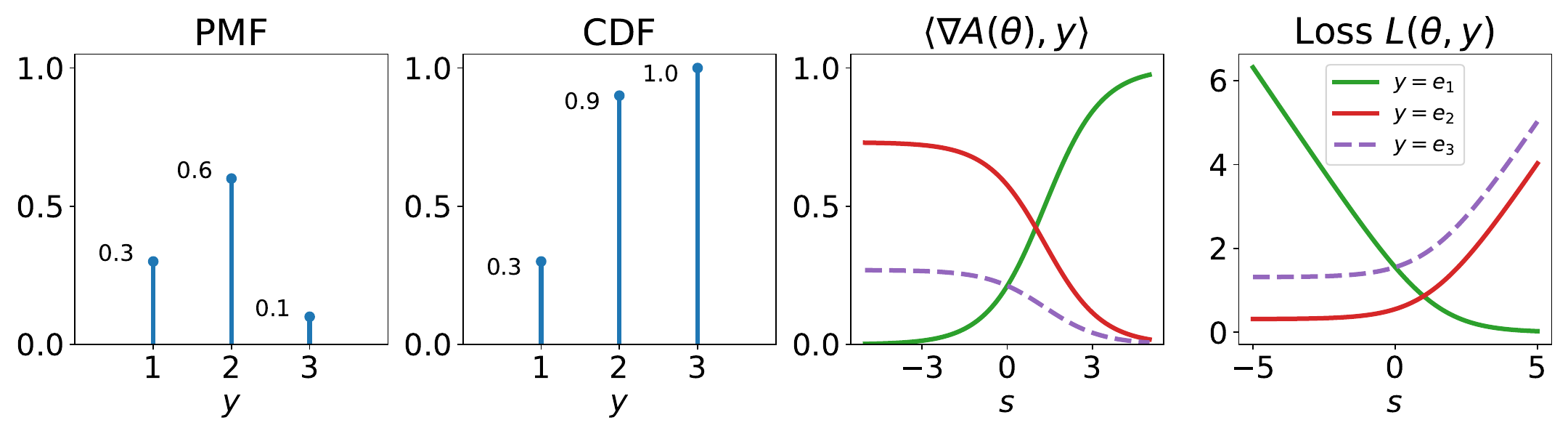}
\caption{
The {\bf categorical distribution}, 
whose PMF and CDF are here illustrated with parameter $\piv = (0.3, 0.6, 0.1)$.
Its mean function is 
$\piv = \nabla A(\thetav) = \mathrm{softargmax}(\thetav)$, 
where $\thetav \in \RR^M$ is for instance the output of a neural network.
Here, for illustration purpose, we choose to set $\thetav = (s, 1, 0)$ and vary
only $s$. Since the mean function $\nabla A(\thetav)$ belongs to $\RR^3$, 
we choose to display
$\langle \nabla A(\thetav), \e_i \rangle = \nabla A(\thetav)_i$, 
for $i \in \{1,2,3\}$.
The negative log-likelihood leads to the {\bf logistic loss},
$L(\thetav, \y) = \mathrm{logsumexp}(\thetav) - \langle \thetav, \y \rangle$.
The loss curve is shown for $\y \in \{\e_1,\e_2,\e_3\}$, again with 
$\thetav=(s,1,0)$ and varying $s$.
}
\label{proba_learn:fig:categorical}
\end{figure*}

\subsubsection*{Parameterization using a softargmax}

Since the parameter vector $\piv$ of a categorical distribution needs to belong
to $\triangle^M$, we typically use a softargmax as the output layer:
\begin{equation*}
\piv \coloneqq f(\x, \w) \coloneqq 
\softargmax(g(\x, \w)),
\end{equation*}
where $g \colon \cX \times \cW \to \RR^M$ is for example a neural network
and
\begin{equation*}
\softargmax(\u) \coloneqq \frac{\exp(\u)}{\sum_j \exp(u_j)}
\in \mathrm{relint}(\triangle^M).
\end{equation*}
The output of the softargmax is in the relative interior of $\triangle^M$,
$\mathrm{relint}(\triangle^M) = \triangle^M \cap \RR^M_{>0}$. That is, the
produced probabilities are always strictly positive. The categorical
distribution is a \textbf{multinomial distribution} with a single trial.
When $g$ is linear in $\w$, 
this is therefore known as \textbf{multiclass} or 
\textbf{multinomial logistic regression},
though strictly speaking a multinomial distribution could use
more than one trial.

\subsection{Regression}

For \textbf{real outcomes}, where $\cY = \RR$, we can use, among other choices,
a \textbf{normal distribution} with parameters
\begin{equation*}
\lambdav \coloneqq (\mu, \sigma), 
\end{equation*}
where $\mu \in \RR$ is the mean parameter and $\sigma \in \RR_+$ is the standard
deviation parameter.
When $Y$ follows a normal distribution with parameters $(\mu, \sigma)$,
we write
\begin{equation*}
Y \sim \mathrm{Normal}(\mu, \sigma).
\end{equation*}
The PDF is
\begin{equation*}
p_{\mu,\sigma}(y) \coloneqq 
\frac{1}{\sigma \sqrt{2\pi} }
\exp\left(-\frac{1}{2}\frac{(y-\mu)^2}{\sigma^2}\right).
\end{equation*}
The expectation is
\begin{equation*}
\EE_{Y \sim p_{\mu,\sigma}}[Y] = \mu.
\end{equation*}
One advantage of the probabilistic perspective
is that we are not limited to
predicting the mean. We can also compute the CDF
\begin{equation*}
\PP(Y \le y) = \frac{1}{2}\left[
    1 + \mathrm{erf}\left(\frac{y - \mu}{\sigma \sqrt{2}}\right)
\right],
\end{equation*}
where we used the \textbf{error function}
\begin{equation*}
\operatorname{erf}(z) 
\coloneqq \frac{2}{\sqrt\pi}\int_0^z e^{-t^2}\,\mathrm{d}t.
\end{equation*}
This function is available in most scientific computing libraries,
such as SciPy \citep{scipy2020}. 
We can also write
\begin{equation*}
\PP(Y \le y) = \Phi\left(\frac{y-\mu}{\sigma}\right)
\end{equation*}
where
\begin{equation}
\Phi(z) \coloneqq \frac{1}{2}\left[
    1 + \mathrm{erf}\left(\frac{z}{\sqrt{2}}\right)
\right]
\label{proba_learn:eq:normal_cdf}
\end{equation}
is the CDF of the standard Gaussian distribution (with zero mean and unit
variance).
From the CDF, we also easily obtain
\begin{equation*}
\PP(a < Y \le b)
=
\frac{1}{2} \left[
    \mathrm{erf}\left(\frac{b - \mu}{\sigma \sqrt{2}}\right)
    - \mathrm{erf}\left(\frac{a - \mu}{\sigma \sqrt{2}}\right)
\right].
\end{equation*}

\begin{figure*}
\centering
\includegraphics[scale=0.32]{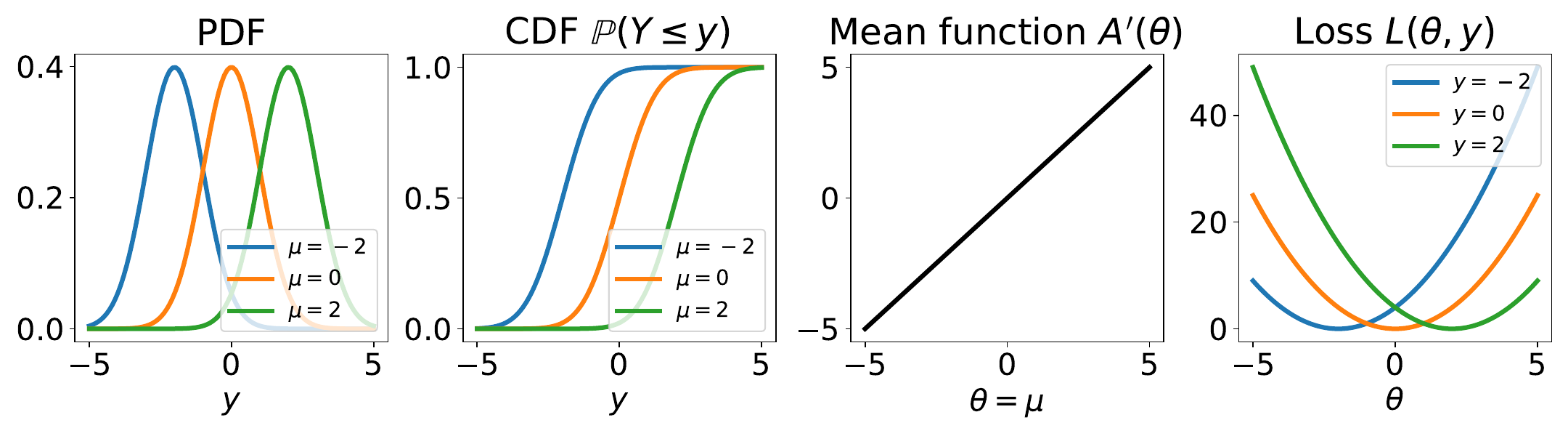}
\caption{
The {\bf Gaussian distribution}, with mean parameter $\mu$ and variance
$\sigma^2=1$.
Its mean function is $\mu = A'(\theta) = \theta$, 
where $\theta$ is for instance the output of a neural network.
The negative log-likelihood leads to the {\bf squared loss},
$L(\theta, y) = (y - \theta)^2$.
The loss curve is shown for $y \in \{-2, 0, 2\}$.
}
\label{proba_learn:fig:gaussian}
\end{figure*}

\subsubsection*{Parameterization}

Typically, in regression, the mean is output by a model,
while the standard deviation $\sigma$ is kept fixed (typically set to $1$). 
Since $\mu$ is unconstrained, we can simply set
\begin{equation*}
\mu \coloneqq f(\x, \w) \in \RR,
\end{equation*}
where $f \colon \cX \times \cW \to \RR$ is for example a neural network.
That is, the output of $f$ is the mean of the distribution, 
\begin{equation*}
\EE_{Y \sim p_{\mu,1}}[Y] = \mu = f(\x, \w).
\end{equation*}
We can also use $\mu$ to predict $\PP(Y \le y)$ or $\PP(a < Y \le b)$, as shown
above.

\subsection{Multivariate regression}

More generally, for \textbf{multivariate outcomes}, where $\cY = \RR^M$, we can
use a \textbf{multivariate normal distribution} with parameters
\begin{equation*}
\lambdav \coloneqq (\muv, \Sigmav), 
\end{equation*}
where $\muv \in \RR^M$ is the mean
and
$\Sigmav \in \RR^{M \times M}$ is the covariance matrix. 
When $Y$ follows a multivariate normal distribution with parameters $(\muv,
\Sigmav)$, we write
\begin{equation*}
Y \sim \mathrm{Normal}(\muv, \Sigmav).
\end{equation*}
The PDF is
\begin{equation*}
    p_{\muv,\Sigmav}(\y) \coloneqq 
    \frac{1}{\sqrt{(2 \pi)^M |\Sigmav|}}
\exp\left(-\frac{1}{2} \langle \y - \muv, 
\Sigmav^{-1} (\y - \muv) \rangle\right).
\end{equation*}
Using a diagonal covariance matrix is equivalent to using 
$M$ independent normal distributions for each $Y_j$,
for $j \in [M]$.
The expectation is
\begin{equation*}
    \EE_{Y \sim p_{\muv,\Sigmav}}[Y] = \muv.
\end{equation*}

\subsubsection*{Parameterization}

Typically, in multivariate regression, the mean is output by a model, while the
covariance matrix is kept fixed
(typically set to the identity matrix).
Since $\muv$ is again unconstrained, we can simply set
\begin{equation*}
\muv \coloneqq f(\x, \w) \in \RR^M.
\end{equation*}
More generally, we can parametrize the function $f$ so as to output both the
mean $\muv$ and the covariance matrix $\Sigmav$, i.e.,
\begin{equation*}
(\muv, \Sigmav) \coloneqq f(\x, \w) \in \RR^M \times \RR^{M \times M}.
\end{equation*}
The function $f$ must be designed such that $\Sigmav$ is symmetric and positive
semi-definite. This is easy to achieve for instance by parametrizing
$\Sigmav = \S \S^\top$ for some matrix $\S$.

\subsection{Integer regression}

For \textbf{integer outcomes}, where $\cY = \NN$, we can use, among other
choices, a \textbf{Poisson distribution} with mean parameter
$\lambda > 0$. 
When $Y$ follows a Poisson distribution with parameter $\lambda$, we write
\begin{equation*}
Y \sim \mathrm{Poisson}(\lambda).
\end{equation*}
The PMF is
\begin{equation*}
\PP(Y = y) =
p_\lambda(y) \coloneqq \frac{\lambda^y \exp(-\lambda)}{y!}.
\end{equation*}
It is the probability of $y$ events occurring in an interval of time.  The
Poisson distribution is frequently used when there is a large number of possible
events, each of which is rare.

The CDF is
\begin{equation*}
\PP(Y \le y) = \sum_{i=0}^y \PP(Y = i).
\end{equation*}
The Poisson distribution implies that the \textbf{index of dispersion} (the
ratio between variance and mean) is $1$, since
\begin{equation*}
\EE[Y] = \VV[Y] = \lambda. 
\end{equation*}
When this assumption is inappropriate, one can use generalized Poisson
distributions \citep{satterthwaite1942generalized}.

\begin{figure*}
\centering
\includegraphics[scale=0.32]{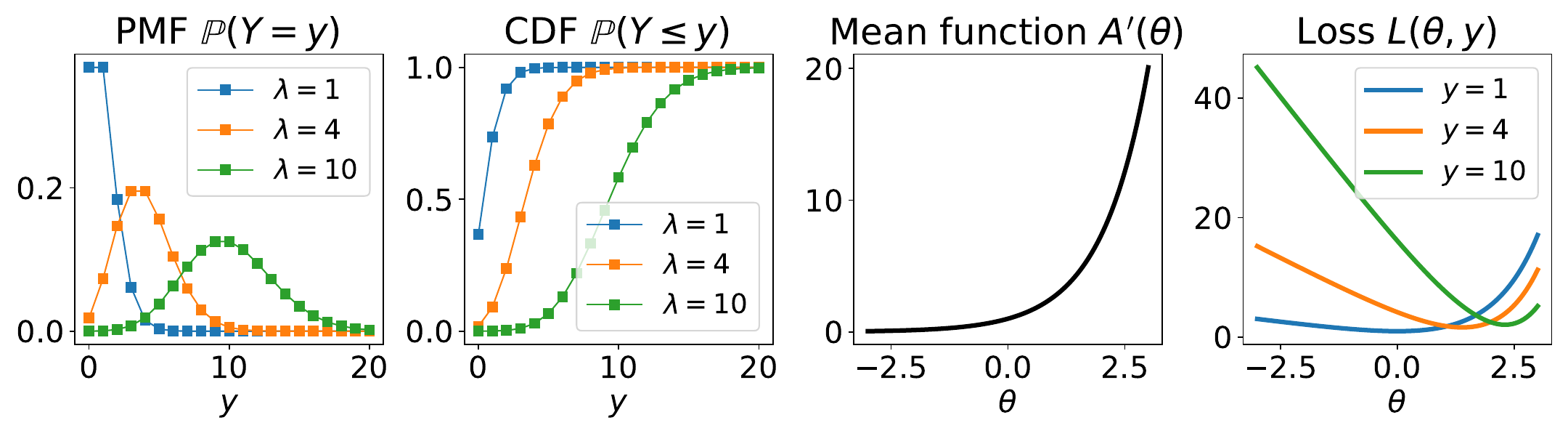}
\caption{
The {\bf Poisson distribution}, with mean parameter $\lambda$.
For the PMF and the CDF, the lines between markers are shown
for visual aid: the Poisson distribution does not assign probability mass to
non-integer values.
Its mean function is $\lambda = A'(\theta) = \exp(\theta)$, 
where $\theta$ is for instance the output of a neural network.
The negative log-likelihood leads to the {\bf Poisson loss},
$L(\theta, y) = -\log p_\lambda(y)
= -y \theta + \exp(\theta) + \log(y!)$, 
which is a convex function of $\theta$.
The loss curve is shown for $y \in \{1, 4, 10\}$.
}
\label{proba_learn:fig:poisson}
\end{figure*}

\subsubsection*{Parameterization using an exponential}

Since the parameter $\lambda$ of a Poisson distribution
needs to be strictly positive, we typically
use an exponential function as output layer:
\begin{equation*}
\lambda \coloneqq f(\x, \w) \coloneqq \exp(g(\x, \w)) > 0,
\end{equation*}
where $g \colon \cX \times \cW \to \RR$.

\newpage

\subsection{Loss functions}

We now discuss how to learn the model parameters $\w \in \cW$ from 
input-output pairs $(\x_1, \y_1), \dots, (\x_N, \y_N)$.

\subsubsection*{Deterministic vs. probabilistic approaches}

In a deterministic approach,
if we used a mapping $f \colon \cX \times \cW \to \cY$,
we could formulate an objective function of the form
\begin{equation*}
L(\w) \coloneqq \frac{1}{N} \sum_{i=1}^N \ell(f(\x_i, \w), \y_i),
\end{equation*}
where $\ell \colon \cY \times \cY \to \RR$ is a loss function.
Unfortunately, $L(\w)$ would be typically \textbf{discontinuous} if $\cY$ is a
discrete output space (as is the case in classification), 
making optimization difficult.

In contrast, in the probabilistic approach, we use a mapping
$f \colon \cX \times \cW \to \Lambda$ to distribution parameters,
and we can formulate an objective of the form
\begin{equation*}
L(\w) \coloneqq \frac{1}{N} \sum_{i=1}^N \ell(f(\x_i, \w), \y_i),
\end{equation*}
where $\ell \colon \Lambda \times \cY \to \RR$.
This is typically a \textbf{continuous} objective, since $\Lambda$ is typically
a continuous set and $p_\lambdav$ varies continuously \wrt $\lambdav$ even if
$p_\lambdav$ is a distribution over a discrete set $\cY$.  In other words, the
probabilistic approach is not only powerful for the inference it allows us to do
(probability, expectation, variance, mode), but also because it allows us to
formulate a continuous and typically differentiable objective function!

\begin{figure*}[p]
\centering
\includegraphics[scale=1.1]{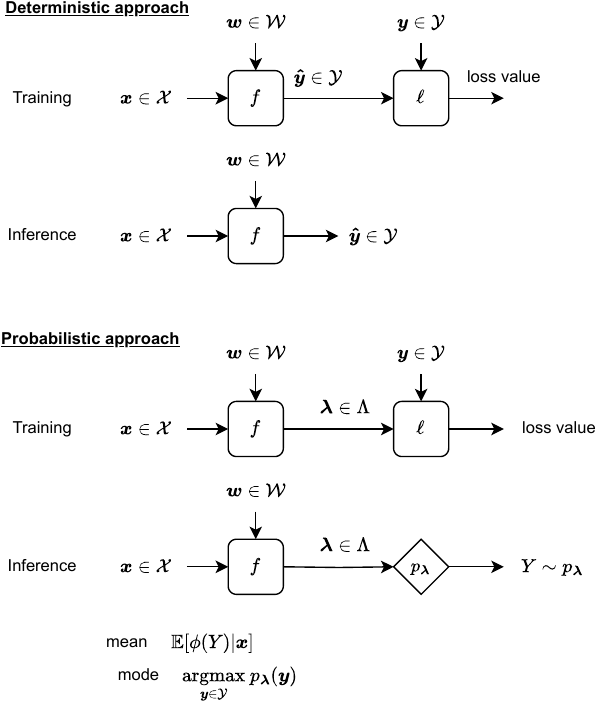}
\caption{Summary of deterministic vs. probabilistic approaches.
}
\label{proba_learn:fig:deter_vs_proba}
\end{figure*}

This is summarized in \cref{proba_learn:fig:deter_vs_proba}.

\newpage

\subsubsection*{Negative log-likelihood}

In the conditional setting
briefly reviewed in \cref{proba_learn:sec:conditional_proba},
we can use maximum likelihood estimation (MLE) 
to estimate the model parameters $\w \in \cW$ of $f$.
Given a set of input-output pairs $(\x_1, \y_1), \dots, (\x_N, \y_N)$,
we choose the model parameters that maximize the \textbf{likelihood} of the
data,
\begin{equation*}
\widehat \w_N \coloneqq
\argmax_{\w \in \cW} \prod_{i=1}^N p_{\lambdav_i}(\y_i),
\end{equation*}
where
\begin{equation*}
\lambdav_i \coloneqq f(\x_i, \w).
\end{equation*}
Again, this is equivalent to minimizing the \textbf{negative log-likelihood},
\begin{equation*}
\widehat \w_N
= \argmin_{\w \in \cW} -\sum_{i=1}^N \log p_{\lambdav_i}(\y_i).
\end{equation*}
In the notation above, this corresponds to defining the loss function
\begin{equation*}
\ell(\lambdav_i, \y_i) \coloneqq -\log p_{\lambdav_i}(\y_i). 
\end{equation*}

\subsubsection*{Recovering well-known loss functions}

Interestingly, MLE allows us to recover several popular loss functions.
\begin{itemize}

\item For the Bernoulli distribution with parameter
$\lambda_i = \pi_i = \mathrm{logistic}(\theta_i)$, where
$\theta_i \coloneqq g(\x_i, \w)$, we have
\begin{align*}
-\log p_{\lambda_i}(y_i) 
&= -\left[y_i \log \pi_i + (1-y_i) \log (1-\pi_i)\right] \\
&= \log(1 + \exp(\theta_i)) - y_i \theta_i \\
&= \mathrm{softplus}(\theta_i) - y_i \theta_i,
\end{align*}
which is the \textbf{binary logistic loss} function.

\item For the categorical distribution with parameters
$\lambdav_i = \piv_i = \mathrm{softargmax}(\thetav_i)$,
where $\thetav_i \coloneqq g(\x_i, \w)$, we have
\begin{align*}
-\log p_{\lambdav_i}(y_i) 
&= -\log \pi_{i,y_i} \\
&= \log \sum_{j=1}^M \exp(\theta_{i,j}) - \theta_{i,y_i} \\
&= \mathrm{logsumexp}(\thetav_i) - \langle \thetav_i, \e_{y_i} \rangle,
\end{align*}
which is the \textbf{multiclass logistic loss} function,
also known as \textbf{cross-entropy loss}.

\item For the normal distribution 
with mean $\lambda_i = \mu_i = f(\x_i, \w)$ and fixed variance $\sigma_i^2$,
we have
\begin{equation*}
-\log p_{\lambda_i}(y_i) = 
\frac{1}{2\sigma_i^2} (y_i - \mu_i)^2
+\frac{1}{2} \log \sigma_i^2 + \frac{1}{2} \log(2\pi),
\end{equation*}
which is, up to a constant and with unit variance, the \textbf{squared loss}
function.

\item For the Poisson distribution 
with mean $\lambda_i = \exp(\theta_i)$,
where $\theta_i \coloneqq g(\x_i, \w)$,
we have
\begin{align*}
-\log p_{\lambda_i}(y_i) 
&= -y_i \log(\lambda_i) + \lambda_i + \log(y_i!) \\
&= -y_i \theta_i + \exp(\theta_i) + \log(y_i!)
\end{align*}
which is the \textbf{Poisson loss} function.
The loss function is convex \wrt $\lambda_i$ and $\theta_i$
for $y_i \ge 0$.

\end{itemize}

\section{Exponential family distributions}
\label{proba_learn:sec:exp_family}

\subsection{Definition}

The exponential family
is a class of probability distributions, whose PMF or PDF
can be written in the form
\begin{align*}
p_\thetav(\y) 
&= \frac{h(\y) \exp\left[\langle \thetav, \phi(\y)
\rangle\right]}{\exp(A(\thetav))} \\
&= h(\y) \exp\left[\langle \thetav, \phi(\y) \rangle - 
A(\thetav)\right],
\end{align*}
where $\thetav$ are the \textbf{natural} or \textbf{canonical parameters} of the
distribution.
The function $h$ is known as the base measure.
The function $\phi$ is the \textbf{sufficient statistic}: it holds all the
information about $\y$ and is used to embed $\y$ in a vector space.
The function $A$ is the \textbf{log-partition} or \textbf{log-normalizer} (see
below for details).
All the distributions we reviewed in \cref{proba_learn:sec:cpd} belong to the
exponential family. With some abuse of notation, we use $p_\lambdav$ for the
distribution in \textbf{original form} and $p_\thetav$ for the distribution in
\textbf{exponential family form}.  As we will see, we can go from $\thetav$ to
$\lambdav$ and vice versa. We illustrate how to rewrite a distribution in
exponential family form below.

\begin{boxexm}{Bernoulli distribution}
The PMF of the Bernoulli distribution
with parameter $\lambda = \pi$ equals
\begin{align*}
p_\lambda(y)
&\coloneqq \pi^y (1-\pi)^{1-y} \\
&= \exp(\log(\pi^y (1-\pi)^{1-y})) \\
&= \exp(y \log(\pi) + (1-y) \log(1-\pi)) \\
&= \exp(\log(\pi/(1-\pi)) y + \log(1 - \pi)) \\
&= \exp(\theta y - \log(1 + \exp(\theta))) \\
&= \exp(\theta y - \mathrm{softplus}(\theta)) \\
&\eqqcolon p_\theta(y).
\end{align*}
Therefore, Bernoulli distributions belong to the exponential family,
with natural parameter $\theta = \mathrm{logit}(\pi) \coloneqq
\log(\pi/(1+\pi))$.
Conversely, we have $\pi = \mathrm{logistic}(\theta) = \frac{1}{1 +
\exp(-\theta)}$.
\end{boxexm}
We rewrite the previously-described distributions in exponential family form 
in \Cref{proba_learn:tab:exponential_family}.
This list is non-exhaustive: there are many more distributions in the
exponential family! \citep{barndorff2014information}

\begin{table}[p]
\caption{Examples of distributions in the exponential family.}
\begin{tabular}{lcc}
\toprule
& Bernoulli & Categorical \\
\midrule
$\cY$ & $\{0,1\}$ & $[M]$ \\
$\lambdav$
& $\pi = \mathrm{logistic}(\theta)$ 
& $\piv = \mathrm{softargmax}(\thetav)$ \\
$\thetav$ 
& $\mathrm{logit}(\pi)$
& $\log \piv + A(\thetav)$ \\
$\phi(y)$ & $y$ & $\e_y$ \\
$A(\thetav)$
& $\mathrm{softplus}(\theta)$ 
& $\mathrm{logsumexp}(\thetav)$ \\
$h(y)$ & 1 & 1 \\[0.5em]
\toprule
       & Normal (location only) & Normal (location-scale) \\
\midrule
$\cY$ & $\RR$ & $\RR$ \\
$\lambdav$ 
& $\mu = \theta \sigma$
& $(\mu, \sigma^2) = (\frac{-\theta_1}{2\theta_2}, \frac{-1}{2\theta_2})$ \\
$\thetav$
& $\frac{\mu}{\sigma}$
& $(\frac{\mu}{\sigma^2}, \frac{-1}{2\sigma^2})$ \\
$\phi(y)$ & $\frac{y}{\sigma}$ & $(y, y^2)$ \\
$A(\thetav)$
& $\frac{\theta^2}{2} = \frac{\mu^2}{2\sigma^2}$
& $\frac{-\theta_1^2}{4\theta_2} - \frac{1}{2}
        \log(-2\theta_2) = \frac{\mu^2}{2\sigma^2} + \log \sigma$ \\
$h(y)$
& $\frac{\exp(\frac{-y^2}{2\sigma^2})}{\sqrt{2\pi}\sigma}$
& $\frac{1}{\sqrt{2\pi}}$ \\[0.5em]
\toprule
       & Multivariate normal & Poisson \\
\midrule
$\cY$ & $\RR^M$ & $\NN$ \\
$\lambdav$
& $(\muv, \Sigmav) = (-\frac{1}{2} \thetav_2^{-1} \thetav_1, -\frac{1}{2}
\thetav_2^{-1})$
& $\lambda = \exp(\theta)$ \\
$\thetav$
& $(\Sigmav^{-1} \muv, -\frac{1}{2} \Sigmav^{-1})$
& $\log \lambda$ \\
$\phi(\y)$ & $(\y, \y\y^\top)$ & $y$ \\
$A(\thetav)$ 
& $-\frac{1}{4} \thetav_1^\top \thetav_2^{-1} \thetav_1 - \frac{1}{2} \log |-2\thetav_2|$
& $\exp(\theta)$ \\
& $= \frac{1}{2} \muv^\top \Sigmav^{-1} \muv + \frac{1}{2} \log |\Sigmav|$
& \\
$h(\y)$ & $(2\pi)^{-M/2}$ & $1/y!$ \\
\bottomrule
\end{tabular}
\label{proba_learn:tab:exponential_family}
\end{table}

\subsection{The log-partition function}

We focus our discussion in this section on
discrete variables for simplicity.
However, the results stated in this section also hold for continuous variables
\citep{wainwright_2008}.
The log-partition function $A$ is the logarithm of the distribution's
normalization factor,
\begin{equation*}
A(\thetav) 
\coloneqq
\log \sum_{\y \in \cY} h(\y) \exp\left[\langle \thetav, \phi(\y)
\rangle\right].
\end{equation*}
We denote the set of valid parameters
\begin{equation*}
\Theta \coloneqq \{ \thetav \in \RR^M \colon A(\thetav) < +\infty\} \subseteq
\RR^M. 
\end{equation*}
We can conveniently rewrite $A(\thetav)$ as
\begin{equation*}
A(\thetav) = \mathrm{logsumexp}(B(\thetav)) 
\coloneqq \log \sum_{\y \in \cY} \exp([B(\thetav)]_\y), 
\end{equation*}
where we defined the \textbf{affine map}
\begin{equation*}
B(\thetav) 
\coloneqq (\langle \thetav, \phi(\y) \rangle + \log h(\y))_{\y \in \cY}.
\end{equation*}
Since $A(\thetav)$ is the composition of
$\mathrm{logsumexp}$, a convex function, and of $B$, an affine map,
we immediately obtain the following proposition.
\begin{boxprop}{Convexity of the log-partition}
$A(\thetav)$ is a convex function.
\label{proba_learn:prop:convexity_A}
\end{boxprop}

A major property of the log-partition function is that its gradient
coincides with the expectation of $\phi(Y)$ according to $p_\thetav$.
\begin{boxprop}{Gradient of the log-partition}
\begin{align*}
\muv(\thetav) \coloneqq 
\nabla A(\thetav)
= \EE_{Y \sim p_\thetav}[\phi(Y)] \in \cM.
\end{align*}
\label{proba_learn:prop:grad_A}
\end{boxprop}
\begin{proof}
The result follows directly from
\begin{equation*}
\nabla A(\thetav)
= \partial B(\thetav)^* \nabla \mathrm{logsumexp}(B(\thetav))
= (\phi(\y))_{\y \in \cY} ~ \mathrm{softmax}(B(\thetav)). 
\end{equation*}
\end{proof}
The gradient $\nabla A(\thetav)$ is therefore often called the \textbf{mean
function}. 
The set of achievable means $\muv(\thetav)$ is defined by
\begin{equation*}
\cM
\coloneqq \mathrm{conv}(\phi(\cY))
\coloneqq \{ \EE_{p}[\phi(Y)] \colon p \in \cP(\cY)\},
\end{equation*}
where 
$\mathrm{conv}(\cS)$ is the convex hull of $\cS$
and
$\cP(\cY)$ is the set of valid probability distributions over $\cY$.

Similarly, the Hessian $\nabla^2 A(\thetav)$ coincides with the covariance matrix
of $\phi(Y)$ according to $p_\thetav$ \citep[Chapter 3]{wainwright_2008}.  

When the exponential family is \textbf{minimal}, which means that the parameters
$\thetav$ uniquely identify the distribution, it is known that $\nabla A$ is a
one-to-one mapping from $\Theta$ to $\cM$. That is,
$\muv(\thetav) = \nabla A(\thetav)$ and $\thetav = (\nabla A)^{-1}(\muv(\thetav))$.

\subsection{Maximum entropy principle}

Suppose we observe the empirical mean $\widehat \muv = \frac{1}{N}
\sum_{i=1}^N \phi(\y_i) \in \cM$ of some observations
$\y_1,\dots,\y_N$.
How do we find a probability distribution achieving this
mean? Clearly, such a distribution may not be unique.
One way to choose among all possible distributions is by using the
\textbf{maximum entropy principle}.
Let us define the Shannon entropy by
\begin{equation*}
H(p) \coloneqq -\sum_{\y \in \cY} p(\y) \log p(\y)
\end{equation*}
for discrete variables and by
\begin{equation*}
H(p) \coloneqq -\int_{\cY} p(y) \log p(y) dy
\end{equation*}
for continuous variables. This captures the level of ``uncertainty'' in $p$,
i.e., it is maximized when the distribution is uniform. 
Then, the \textbf{maximum entropy distribution} satisfying the first-order
moment condition (i.e., whose expectation matches the empirical mean) is
\begin{equation*}
p^\star \coloneqq
\argmax_{p \in \cP(\cY)} H(p)
\quad \text{s.t.} \quad
\EE_{Y \sim p}[\phi(Y)] = \widehat \muv.
\end{equation*}
It can be shown that the maximum entropy distribution is necessarily in the
exponential family with sufficient statistic defined by $\phi$ and its 
canonical parameters $\thetav$ 
coincide with the \textbf{Lagrange multipliers} of the above constraint
\citep[Section 3.1]{wainwright_2008}.

\subsection{Maximum likelihood estimation}

As in \cref{proba_learn:sec:mle}, 
to fit the parameters $\thetav \in \Theta$ of an exponential
family distribution to some \iid observations
$\y_1, \dots, \y_N$, we can use the MLE principle, i.e.,
\begin{equation*}
\widehat \thetav_N
= \argmax_{\thetav \in \Theta} \prod_{i=1}^N p_\thetav(\y_i)
= \argmin_{\thetav \in \Theta} -\sum_{i=1}^N \log p_\thetav(\y_i).
\end{equation*}
Fortunately, for exponential family distributions, the log probability/density
enjoys a particularly simple form.
\begin{boxprop}{Negative log-likelihood}
The negative log-likelihood of an exponential family distribution is
\begin{equation*}
-\log p_\thetav(\y)
= A(\thetav) - \langle \thetav, \phi(\y) \rangle - \log h(\y).
\end{equation*}
Its gradient is
\begin{equation*}
-\nabla_\thetav \log p_\thetav(\y)
= \nabla A(\thetav) - \phi(\y)
= \EE_{Y \sim p_\thetav}[\phi(Y)] - \phi(\y)
\end{equation*}
and its Hessian is
\begin{equation*}
-\nabla^2_\thetav \log p_\thetav(\y)
= \nabla^2 A(\thetav),
\end{equation*}
which is \textbf{independent} of $\y$.
\label{proba_learn:prop:negative_ll_exp_family}
\end{boxprop}
It follows from \cref{proba_learn:prop:convexity_A} that 
$\thetav \mapsto -\log p_\thetav(\y)$ is \textbf{convex}.
Interestingly, we see that the gradient is the \textbf{residual} between the
expectation of $\phi(Y)$ according to the model and the observed $\phi(\y)$.
Therefore, the negative log-likelihood of an exponential family distribution can
be seen as performing first moment matching.

\subsection{Probabilistic learning with exponential families}

In the supervised probabilistic learning setting, 
we wish to estimate a conditional distribution of the form
$p_\w(\y \mid \x)$. 
Given a model function $f$, such as a neural network,
a common approach for defining such a conditional distribution
is by reduction to the unconditional setting,
\begin{equation*}
p_\w(\y \mid \x) \coloneqq p_\thetav(\y)
\quad \text{where} \quad
\thetav \coloneqq f(\x, \w).
\end{equation*}
In other words, the role of $f$ is to produce the parameters of $p_\thetav$
given some input $\x$. It is a function from $\cX \times \cW$ to $\Theta$.
Note that $f$ must be designed such that it produces an output in 
\begin{equation*}
\Theta \coloneqq \{ \thetav \in \RR^M \colon A(\thetav) < +\infty\}.
\end{equation*}
Many times, $\Theta$ will be the entire $\RR^M$ but this is not always the case.
For instance, as we previously discussed, for a multivariate normal
distribution, where $\thetav = (\muv, \Sigmav) = f(\x, \w)$, we need to
ensure that $\Sigmav$ is a positive semidefinite matrix.  

\subsubsection*{Training}

Given input-output pairs $\{(\x_i, \y_i)\}_{i=1}^N$,
we then seek to find the parameters $\w$ of $f(\x, \w)$ by minimizing
the negative log-likelihood
\begin{equation*}
\argmin_{\w \in \cW}
-\sum_{i=1}^N \log p_{\thetav_i}(\y_i)
=
\argmin_{\w \in \cW}
\sum_{i=1}^N A(\thetav_i) - \langle \thetav_i, \phi(\y_i) \rangle
\end{equation*}
where $\thetav_i \coloneqq f(\x_i, \w)$.
While $-\log p_\thetav(\y)$ is a convex function of $\thetav$ for exponential
family distributions, we emphasize that
$-\log p_{f(\x,\w)}(\y)$ is typically a nonconvex function of 
$\w$, when $f$ is a \textbf{nonlinear} function, such as a neural network.

\begin{figure*}[t]
\centering
\includegraphics[scale=1.1]{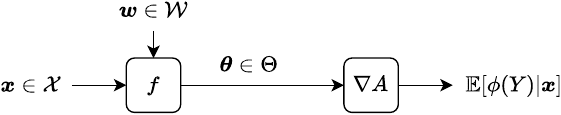}
\caption{Generalized linear models.}
\label{proba_learn:fig:glm}
\end{figure*}

\subsubsection*{Inference}

Once we have found $\w$ by minimizing the objective function above,
there are several possible strategies to perform inference for a new input $\x$.
\begin{itemize}
\item \textbf{Expectation.} 
When the goal is to compute the expectation of $\phi(Y)$, we can use
$\nabla A(f(\x, \w))$. That is, we compute the distribution parameters
associated with $\x$ by $\thetav = f(\x, \w)$ and then we compute the mean by
$\muv = \nabla A(\thetav)$. 
When $f$ is linear in $\w$, the composition $\nabla A \circ f$ is called
a \textbf{generalized linear model}.
This is illustrated in \cref{proba_learn:fig:glm}.

\item \textbf{Probability.}
When the goal is to compute the probability of a certain $\y$,
we can compute the distribution parameters associated with $\x$ by $\thetav = f(\x,
\w)$ and then we can compute $\PP(Y = \y | X = \x) = p_\thetav(\y)$. 
In the particular case of the categorical distribution (of which the Bernoulli
distribution is a special case), we point out again that the mean and the
probability vector coincide:
\begin{equation*}
\muv = \piv = \nabla A(\thetav) = \softargmax(\thetav) \in \triangle^M.
\end{equation*}

\item \textbf{Other statistics.} When the goal is to compute other quantities,
such as the variance or the CDF, we can convert the natural parameters
$\thetav$ to the original distribution parameters $\lambdav$ (see
\cref{proba_learn:tab:exponential_family} for examples).
Then, we can use established formulas for the distribution in original form,
to compute the desired quantities.
\end{itemize}

\section{Summary}

\begin{itemize}

\item We reviewed \textbf{discrete} and
\textbf{continuous} probability distributions. 

\item We saw how to fit distribution
parameters to data using the \textbf{maximum likelihood estimation} (MLE)
principle and saw its connection with the \textbf{Kullback-Leibler divergence}.

\item Instead of designing a model function from the input space $\cX$
to the output space $\cY$, we saw that we can perform \textbf{probabilistic
supervised learning} by designing a model function from $\cX$ to
\textbf{distribution parameters} $\Lambda$.

\item Leveraging the so-obtained parametric \textbf{conditional distribution} then
allowed us to compute, not only output probabilities, but also various
statistics such as the mean and the variance of the outputs.  

\item We reviewed the \textbf{exponential family}, 
a principled generalization of numerous distributions,
which we saw is tightly connected with the \textbf{maximum entropy principle}.

\item Importantly, the approaches described in this chapter produce perfectly valid
\textbf{computation graphs}, meaning that we can combine them with neural
networks and we can use automatic differentiation, to compute their derivatives.
    
\end{itemize}

%% file: chapters/neural_nets/neural_nets_main.tex
\chapter{Parameterized programs} 
\label{chap:neural_nets} 

Neural networks can be thought of as parameterized programs: programs with
learnable parameters. In this chapter, we begin by reviewing how to represent
programs mathematically. We then review several key neural network architectures
and components.

\section{Representing computer programs}

\subsection{Computation chains}
\label{neural_nets:sec:comput_chains} 

To begin with, we consider simple programs that apply a
\textbf{sequence} of functions $f_1,\dots,f_K$ to an input $\s_0 \in
\cS_0$.  We call such programs \textbf{computation chains}.
For example, an image may go through a sequence of transformations such as
cropping, rotation, normalization, and so on.
In neural networks, the transformations are typically parameterized, 
and the parameters are learned, leading to feedforward networks,
presented in \cref{neural_nets:sec:ff}.
Another example of a sequence of functions is a for loop, presented
in~\cref{cf:sec:for_loops}.

\begin{figure}[t]
	\begin{center}
		\includegraphics[width=\linewidth]{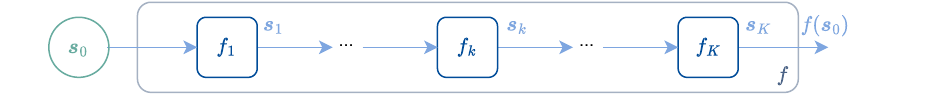}
		\caption{A computation chain is a sequence of function compositions. 
		In the graph above, each intermediate node represents a single function. The
		first node represents the input, the last node the output. Edges
		represent the dependencies of the functions with respect to previous
		outputs or to the initial input. \label{neural_nets:fig:chain}}
	\end{center}
\end{figure}

Formally, a computation chain can be written as
\begin{align}
	\s_0 &\in \cS_0\nonumber \\
	\s_1 &\coloneqq f_1(\s_0) \in \cS_1 \nonumber\\ 
		 & \hspace{6pt} \vdots \nonumber\\
	\s_K &\coloneqq f_K(\s_{K-1}) \in \cS_K \nonumber \\
	f(\s_0) &\coloneqq \s_K. 
  \label{neural_nets:eq:chain}
\end{align}
Here, $\s_0$ is the \textbf{input}, $\sv_k \in \cS_k$ is an intermediate
\textbf{state} of the program, and $\s_K \in \cS_K$ is the final
\textbf{output}. 
Of course, the domain
(input space) of $f_k$ must be compatible with the image (output space) of
$f_{k-1}$.
That is, we should have $f_k \colon \cS_{k-1} \to \cS_k$.
We can write a computation chain equivalently as
\begin{align*}
  f(\s_0)
  &= (f_K \circ \dots \circ f_2 \circ f_1)(\s_0) \\
  &= f_K(\dots f_2(f_1(\s_0))).
\end{align*}
A computation chain can be represented by a directed graph, shown
in~\cref{neural_nets:fig:chain}. 
The edges in the chain define a \textbf{total order}.
The order is total,
since two nodes are necessarily linked to each other by a path.

\subsection{Directed acyclic graphs}

In generic programs, intermediate functions may depend, not only on the previous
function output, but on the outputs of several
different functions.  Such dependencies are best expressed using graphs.

A \textbf{directed graph} $\cG = (\cV, \cE)$ is defined by a set of
\textbf{vertices} or \textbf{nodes} $\cV$ and a set of \textbf{edges} $\cE$
defining directed dependencies between vertices. An edge $(i,j) \in \cE$
is an 
ordered pair of vertices $i \in \cV$ and $j \in \cV$. It is also denoted $i
\rightarrow j$, to indicate that $j$ depends on $i$. For representing inputs and
outputs, it will be convenient to use
\textbf{incoming half-edges} $\rightarrow j$ and \textbf{outgoing half-edges} $i
\rightarrow$.

In a graph $\cG= (\cV, \cE)$, the \textbf{parents} of a vertex $j$ are the set of
nodes pointing to $j$, denoted $\parent(j) \coloneqq \{i: i \rightarrow j\}$.
The \textbf{children} of a vertex $i$ are the set of nodes $i$ is pointing to,
that is, $\child(i) \coloneqq \{j \colon i \rightarrow j\}$. Vertices without
parents are called \textbf{roots} and vertices without children are called
\textbf{leaves}. 

A \textbf{path} from $i$ to $j$ is defined by a sequence of vertices $j_1,
\ldots, j_m$, potentially empty, such that $i \rightarrow j_1 \rightarrow \ldots
\rightarrow j_m \rightarrow j$. An \textbf{acyclic} graph is a graph such that
there exists no vertex $i$ with a path from $i$ to $i$. 
A \textbf{directed acyclic graph} (DAG) is a graph that is both directed and
acyclic.

The edges of a DAG define a \textbf{partial order} of the vertices, denoted
$i \preceq j$ if there exists a path from $i$ to $j$. The order is partial,
since two vertices may not necessarily be linked to each other by a path.
Nevertheless, we can define a total order called a \textbf{topological
order}: any order such that $i \leq j$ if and only if there is no path from
$j$ to $i$. 

\begin{figure}
  \centering
  \includegraphics[width=0.4\linewidth]{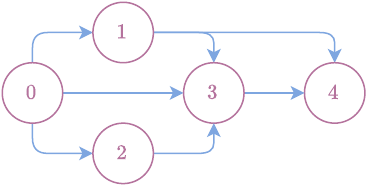}~
  \caption{%
  Example of a \textbf{directed acyclic graph}. Here the nodes are $\cV = \{0,
  1, 2, 3, 4\}$, the edges are $\cE = \{(0, 1), (0, 2), (0, 3), (1, 3), (2, 3),
  (1, 4), (3, 4)\}$. Parents of the node $3$ are $\parent(3) = \{0, 1, 2\}$.
  Children of node $1$ are $\child(1) = \{3, 4\}$. There is a unique root, $0$,
  and a unique leaf, $4$; $0 \rightarrow 3 \rightarrow 4$ is a path from $0$ to
  $4$. This is an acyclic graph since there is no cycle (i.e., a path from a
  node to itself). We can order nodes $0$ and $3$ as $0 \leq 3$ since there is
  no path from $3$ to $0$. Similarly, we can order $1$ and $2$ as $1 \leq 2$
  since there is no path from $2$ to $1$. Two possible topological orders of
  the nodes are $(0, 1, 2, 3, 4)$ and $(0, 2, 1, 3, 4)$.
  \label{neural_nets:fig:graph}
  }
\end{figure}

\subsection{Computer programs as DAGs}\label{neural_nets:sec:comput_graphs}

We assume that a program defines a
mathematically valid function (a.k.a. pure function): the program
should return identical
values for identical arguments and should not have any side effects.
We also assume that the program halts, i.e., that it terminates in a
\textbf{finite} number of steps.
As such a program is made of a finite number of intermediate functions and
intermediate variables,
the dependencies between functions and variables can be expressed using a 
directed acyclic graph (DAG).
Without loss of generality, we make the following simplifying assumptions:
\begin{enumerate}
    \item There is a single input $\s_0 \in \cS_0$.
    \item There is a single output $\s_K \in \cS_K$.
    \item Each intermediate function $f_k$ in the program outputs a single
        variable $\s_k \in \cS_k$. 
\end{enumerate}
We number the nodes as $\cV \coloneqq \{0, 1, \dots, K\}$.
Node $0$ is the root, corresponding to the input $\s_0 \in \cS_0$.
Node $K$ is the leaf, corresponding to the final output $\s_K \in \cS_K$. 
Because of the third assumption above,
apart from $\s_0$, each variable $\s_k$ is in \textbf{bijection} with a function
$f_k$. Therefore, node $0$ represents the input $\s_0$, and
nodes $1, \dots, K$ represent both a function $f_k$ and an output variable
$\s_k$.

\begin{figure}[t]
  \centering
  \includegraphics[width=\linewidth]{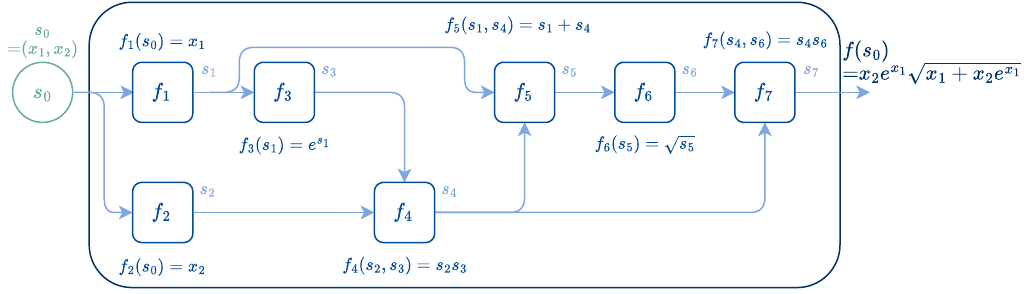}
  \caption{Representation of $f(x_1, x_2) = x_2e^{x_1}\sqrt{x_1 + x_2 e^{x_1}}$
  as a DAG, with functions and variables as nodes. Edges indicate function and
  variable dependencies.
  The function $f$ is decomposed as $8$ elementary functions in topological
  order.
  \label{neural_nets:fig:simple_graph}}
\end{figure}

Edges in the DAG represent dependencies.
The parents $i_1, \dots, i_{p_k} \coloneqq
\parent(k)$ of node $k$, where $p_k \coloneqq |\parent(k)|$, indicate the
variables 
$\s_{\parent(k)} \coloneqq \s_{i_1}, \dots, \s_{i_{p_k}}$
that the function $f_k$ needs to perform its computation.
Put differently, the parents $i_1, \dots, i_{p_k}$ indicate the functions 
$f_{i_1}, \ldots, f_{i_{p_k}}$ that need to be 
evaluated, prior to evaluating $f_k$. 
An example of a computation graph in our formalism is presented
in~\cref{neural_nets:fig:simple_graph}. 

\subsubsection*{Executing a program}

To execute a program, we need to ensure that we evaluate the intermediate
functions in the correct order. 
Therefore, we assume that the nodes $0, 1, \dots, K$ are in a topological order
(if this is not the case, we need to perform a topological sort first). 
We can then execute a program by evaluating for $k \in [K]$
\begin{align*}
\s_k 
\coloneqq f_k(\s_{\parent(k)})
\coloneqq f_k(\s_{i_1}, \dots, \s_{i_{p_k}}) \in \cS_k.
\end{align*}
Note that we can either view $f_k$ as a single-input function of 
$\s_{\parent(k)}$, which is a tuple of elements,
or as a multi-input function of
$\s_{i_1}, \dots, \s_{i_{p_k}}$.
The two views are essentially equivalent.

The procedure for executing a program is summarized in
\cref{neural_nets:algo:comp_graph}.

\begin{algorithm}[t]\caption{Executing a program}
  \label{neural_nets:algo:comp_graph}
      \begin{algorithmic}[1]
          \Statex {\bf Functions:} $f_1, \dots, f_K$ in topological order
          \Statex {\bf Input:} input $\s_0 \in \cS_0$
          \For{$k \coloneqq 1, \ldots, K$}
          \State Retrieve parent nodes 
          $(i_1, \dots, i_{p_k}) \coloneqq \parent(k)$
          \State Compute 
$\s_k \coloneqq f_k(\s_{\parent(k)}) \coloneqq f_k(\s_{i_1}, \ldots,
\s_{i_{p_k}})$ 
          \EndFor
          \State {\bf Output:} $f(\s_0) \coloneqq \s_K$
      \end{algorithmic}
  \end{algorithm}

\subsubsection*{Dealing with multiple program inputs or outputs}

When a program has multiple inputs, we can always group them into 
$\s_0 \in \cS_0$ 
as 
$\s_0 = (\s_{0,1}, \dots, \s_{0,N_0})$
with
$\cS_0 = (\cS_{0,1} \times \dots \times \cS_{0,N_0})$, 
since later functions can always filter out
what elements of $\s_0$ they need.
Likewise, if an intermediate function $f_k$ has multiple outputs, we can always
group them as a single output 
$\s_k = (\s_{k,1},\dots,\s_{k,N_k})$
with
$\cS_k = (\cS_{k,1} \times \dots \times \cS_{k,N_k})$,
since later functions can filter out the
elements of $\s_k$ that they need.

\subsubsection*{Alternative representation: bipartite graphs}

\begin{figure}[t]
  \centering

\includegraphics[width=0.45\linewidth]{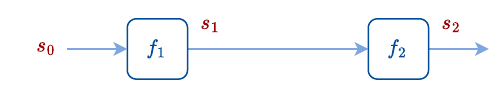}
\hspace{1em}
\includegraphics[width=0.45\linewidth]{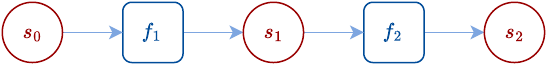}
\caption{Two possible representations of a program. 
\textbf{Left:} Functions and output variables are represented by the same nodes.
\textbf{Right:} functions and variables are represented by a disjoint set of
nodes.}
\label{neural_nets:fig:revisiting_chains}
\end{figure}

In our formalism, because a function $f_k$ always has a single output $\s_k$,
a node $k$ can be seen as representing both the variable $\s_k$ and the function
$f_k$.
Alternatively,
as shown in \cref{neural_nets:fig:revisiting_chains},
we can represent variables and functions as separate nodes, that is,
using a \textbf{bipartite graph}. 
This formalism is akin to \textbf{factor graphs}~\citep{frey1997factor,
loeliger2004introduction} used in probabilistic modeling, but with directed
edges. 
One advantage of this formalism is that it allows functions to explicitly
have multiple outputs.
We focus on our formalism for simplicity.

\subsection{Arithmetic circuits}
\label{neural_nets:sec:arithmetic_circuits}

Arithmetic circuits are one of the simplest examples of computation graph,
originating from \textbf{computational complexity theory}.
Formally, an arithmetic circuit over a field
$\FF$, such as the reals $\RR$, is a directed acyclic graph (DAG) 
whose root nodes are elements of $\FF$ and whose functions $f_k$ are 
either $+$ or $\times$. The latter are often called \textbf{gates}.
Contrary to the general computation graph case, because each $f_k$ is either $+$
or $\times$, it is important to allow the graph to have several root nodes.
Root nodes can be either variables or constants, and should belong to $\FF$.

Arithmetic circuits can be used to compute \textbf{polynomials}.
There are potentially multiple arithmetic circuits for representing a
given polynomial. One important question is then to find the most efficient
arithmetic circuit for computing a given polynomial. To compare arithmetic
circuits representing the same polynomial, an intuitive notion of complexity is
the \textbf{circuit size}, as defined below.
\begin{boxdef}{Circuit and polynomial sizes}
\label{neural_nets:def:circuit_size}
The size $S(\cC)$ of a circuit $\cC$ is the number of edges in the directed
acyclic graph representing $\cC$.
The size $S(f)$ of a polynomial $f$ is the smallest $S(\cC)$ among all $\cC$
representing $f$.
\end{boxdef}
For more information on arithmetic circuits, we refer the reader
to the monograph of \citet{chen_2011}.

\section{Feedforward networks}\label{neural_nets:sec:ff}

A feedforward network can be seen as a computation chain with
\textbf{parameterized} functions $f_k$,
\begin{align*}
  \sv_0 &\coloneqq \xv \\
  \sv_1 &\coloneqq f_1(\sv_0, \wv_1) \\
  \sv_2 &\coloneqq f_2(\sv_1, \wv_2) \\
  &\vdots \\
  \sv_K &\coloneqq f_K(\sv_{K-1}, \wv_K),
\end{align*}
for a given input $\x \in \cX$ and \textbf{learnable parameters}
$\w_1, \ldots, \w_K \in \cW_1 \times \dots \times \cW_K$. 
Each function $f_k$ is called a \textbf{layer}
and each $\sv_k \in \cS_k$ can be seen as an \textbf{intermediate
representation} of the input $\xv$. The dimensionality of 
$\cS_k$ is known as the
\textbf{width} (or number of hidden units) of layer $k$. A feedforward network
defines a function $\s_K \eqqcolon f(\xv, \wv)$ from $\cX \times \cW$ to
$\cS_K$, where $\wv \coloneqq (\wv_1, \dots, \wv_K) \in \cW \coloneqq \cW_1
\times \ldots \times \cW_K $. 

Given such a parameterized program, we can learn the parameters by adjusting
$\wv$ to fit some data. For instance, given a dataset of $(\xv_i, \yv_i)$ pairs,
we may minimize the squared loss $\|\yv_i - f(\x_i, \w)\|_2^2$ on average over
the data, w.r.t. $\wv$. The minimization of such a loss requires accessing its
gradients with respect to $\w$.

\section{Multilayer perceptrons}\label{neural_nets:sec:mlp}

\subsection{Combining affine layers and activations}

In the previous section, we did not specify how to parametrize the feedforward
network. A typical parametrization, called the multilayer perceptron (MLP), uses
\textbf{fully-connected} (also called \textbf{dense}) layers of the form
\begin{equation*}
\sv_k 
= f_k(\sv_{k-1}, \wv_k) 
\coloneqq a_k(\W_k \sv_{k-1} + \bv_k),
\end{equation*}
where we defined the tuple $\wv_k \coloneqq (\W_k, \bv_k)$ and 
where we assumed that
$\W_k$ and $\bv_k$ are a matrix and a vector of appropriate size. We can further
decompose the layer into two functions. The function $\sv \mapsto \W_k \sv +
\bv_k$ is called an affine layer. The function $\vv \mapsto a_k(\vv)$ is a
parameter-free \textbf{nonlinearity}, often called an \textbf{activation
function} (see \Cref{neural_nets:sec:activations}).
The value
$\alphav_k \coloneqq \W_k \sv_{k-1} + \bv_k$
is often called the \textbf{pre-activation value}
and the value
$a_k(\alphav_k)$
is the \textbf{activation value}.

More generally, we may replace the matrix-vector product $\W_k \sv_{k-1}$ by any
parametrized linear function of $\sv_{k-1}$. For example, 
\textbf{convolutional layers}
use the convolution of an input $\sv_{k-1}$ with some filters $W_k$,
seen as a linear map.

\begin{boxrem}{Dealing with multiple inputs}
Sometimes, it is necessary to deal with networks of multiple inputs. For
example, suppose we want to design a function $g(\x_1, \x_2, \w_g)$, where
$\x_1 \in \cX_1$ and $\x_2 \in \cX_2$. A simple way to do so is to use
the concatenation $\x \coloneqq \x_1 \oplus \x_2 \in \cX_1 \oplus \cX_2$ as
input to a network $f(\x, \w_f)$. Alternatively, instead of concatenating $\x_1$
and $\x_2$ at the input layer, they can be concatenated after having been
through one or more hidden layers.
\end{boxrem}

\subsection{Link with generalized linear models}

When the depth is $K=1$ (only one layer), the output of an MLP is
\begin{equation*}
\s_1 = a_1(\W_1 \xv + \bv_1).
\end{equation*}
This is called a \textbf{generalized linear model} (GLM);
see \cref{proba_learn:sec:exp_family}.
Therefore, MLPs include GLMs as a special case. 
In particular, when $a_1$ is the
softargmax (see \Cref{neural_nets:sec:activations}), we obtain (multiclass)
logistic regression. For general depth $K$, the output of an MLP is
\begin{equation*}
\s_K = a_K(\W_K \sv_{K-1} + \bv_K).
\end{equation*}
This can be seen as a GLM on top of \textbf{learned representation} $\sv_{K-1}$
of the input $\xv$. This is the main appeal of MLPs: they learn the feature
representation and the output model at the same time! We will see that MLPs can
also be used as subcomponents in other architectures.

\section{Activation functions}
\label{neural_nets:sec:activations}

As we saw in \Cref{neural_nets:sec:mlp}, feedforward networks typically use an
activation function $a_k$ at each layer. In this section, we present various
nonlinearities from scalar to scalar or from vector to scalar. We also present
probability mappings that can be used as such activations.

\subsection{ReLU and softplus}

Many activations are \textbf{scalar-to-scalar} functions,
but they can also be applied to vectors in an element-wise fashion.
The \textbf{ReLU} (rectified linear unit) is a popular nonlinearity defined as
the \textbf{non-negative part} of its input
\begin{equation*}
\mathrm{relu}(u) 
\coloneqq \max(u, 0) 
= \begin{cases} 
u, & u \ge 0 \\ 
0, & u < 0
\end{cases}.
\end{equation*}
It is a piecewise linear function and includes a kink at $u=0$.
A multilayer perceptron with ReLU activations
is called a \textbf{rectifier neural network}. The layers take the form
\begin{equation*}
\sv_k = \mathrm{relu}(\Av_k \sv_{k-1} + \bv_k),
\end{equation*}
where the ReLU is applied element-wise. The ReLU can be replaced with a smooth
approximation (i.e., without kinks), called the \textbf{softplus}
\begin{equation*}
    \mathrm{softplus}(u) \coloneqq \log(1 + e^u).
\end{equation*}
Unlike the ReLU, it is always strictly positive. 
Other smoothed variants of the ReLU are possible, 
see \cref{smoothing:sec:relu}.

\subsection{Max pooling and log-sum-exp}
\label{neural_nets:sec:vector_to_scalar}

Many activations are \textbf{vector-to-scalar} functions: 
they \textbf{reduce} vectors to a scalar value. 
This scalar value can be seen as a statistic, ``summarizing'' the vector. 

\subsubsection*{Max pooling}

An example of vector-to-scalar reduction is the maximum value, 
also known as \textbf{max pooling}. 
Given a vector $\u \in \R^M$, it is defined as
\begin{equation*}
\max(\u) \coloneqq \max_{j \in [M]} u_j.
\end{equation*}

\begin{figure}[t]
\centering
\includegraphics[scale=0.45]{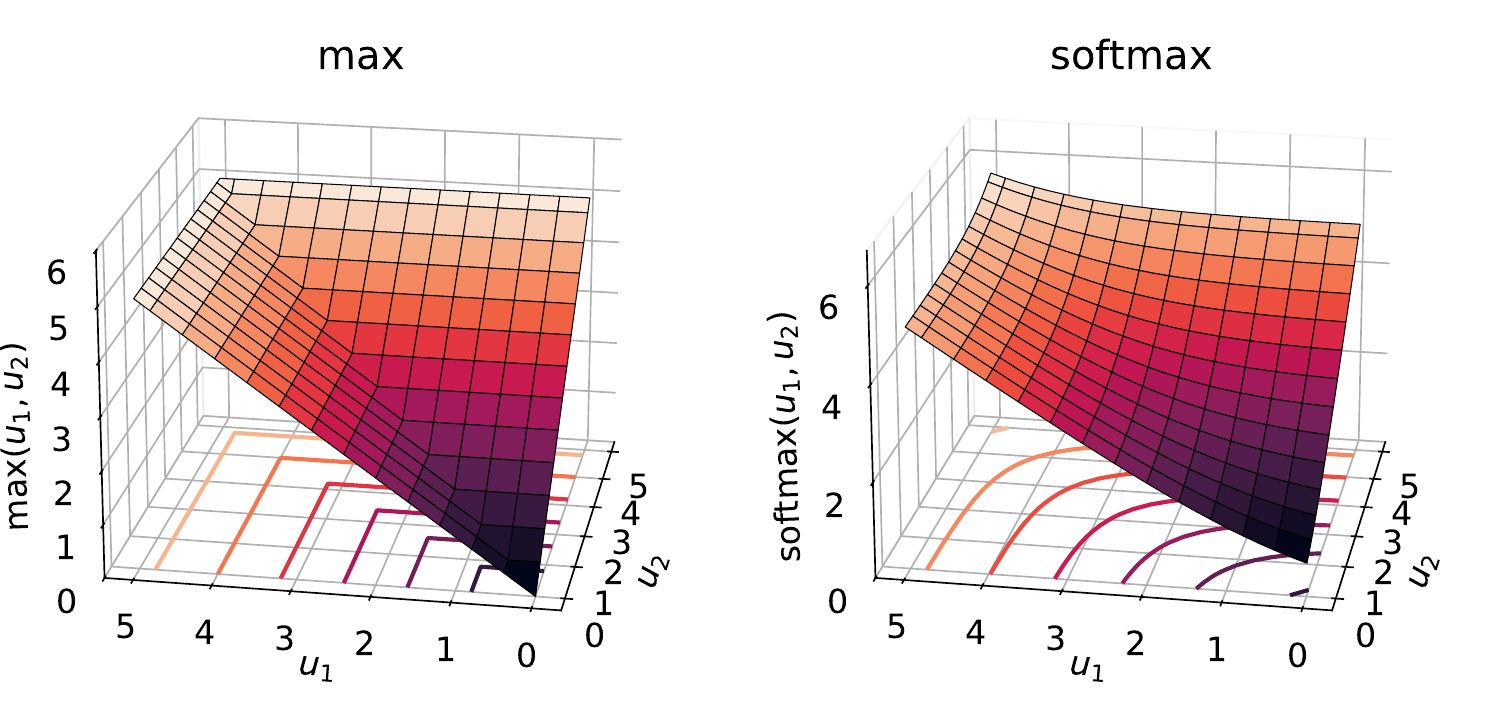}
\caption{The maximum operator is a piecewise linear function. The log-sum-exp
(\aka softmax) is a smoothed approximation.}
\label{neural_net:fig:max_softmax_3d}
\end{figure}

\subsubsection*{Log-sum-exp as a soft maximum}

Another example of vector-to-scalar reduction is the \textbf{log-sum-exp},
\begin{equation*}
\lse(\u) \coloneqq 
\mathrm{softmax}(\u) \coloneqq 
\log \sum_{j=1}^M e^{u_j}.
\end{equation*}
As illustrated in \cref{neural_net:fig:max_softmax_3d},
it is known to behave like a \textbf{soft maximum}.
The log-sum-exp can be seen as a generalization
of the softplus, as we have for all $u \in \RR$
\begin{equation*}
\lse((u, 0)) = \mathrm{softplus}(u).
\end{equation*}
A numerically stable implementation of the log-sum-exp is given
by
\begin{equation*}
\lse(\u) = \lse(\u - c\ones) + c,
\end{equation*}
where $c \coloneqq \max_{j \in [M]} u_j$.

More generally, we can introduce a temperature parameter $\gamma > 0$
\begin{equation*}
\lse_\gamma(\u) = \gamma \cdot \lse(\u / \gamma).
\end{equation*}
It can be shown that for all $\u \in \RR^M$,
\begin{equation*}
    \max(\u) \le \lse_\gamma(\u) \le \max(\u) + \gamma \cdot \log(M).
\end{equation*}
Therefore, $\lse_\gamma(\u) \to \max(\u)$ as $\gamma \to 0$.
Other definitions of soft maximum are possible; 
see \cref{smoothing:sec:max_argmax}.

\subsubsection*{Log-sum-exp as a log-domain sum}

Besides its use as a soft maximum, the log-sum-exp often arises for computing
sums in the log domain. Indeed, suppose we want to compute
$s \coloneqq \sum_{i=1}^M u_i$, where $u_i > 0$. 
If we define $\tilde{u}_i \coloneqq \log u_i$ and
$\tilde{s} \coloneqq \log s$, we then have 
\begin{equation*}
\tilde{s} = \log \sum_{i=1}^M \exp(\tilde{u}_i).
\end{equation*}
Written differently, we have the identity
\begin{equation*}
\log\left(\sum_{i=1}^M u_i\right) = \lse(\log(\u)).
\end{equation*}
We can therefore see the log-sum-exp as the sum counterpart
of the identity for products 
\begin{equation*}
\log\left(\prod_{i=1}^M u_i \right) = \sum_{i=1}^M \log(u_i).
\end{equation*}
As an example, we use the log-sum-exp to perform the forward-backward algorithm
in the log-domain in \cref{gm:sec:forward_backward_chain}.

\subsection{Sigmoids: binary step and logistic functions}
\label{neural_nets:sec:sigmoids}

Oftentimes, we want to map a real value to a number in $[0,1]$, that
can represent the probability of an event. For that purpose, we generally use
\textbf{sigmoids}. A sigmoid is a function with a characteristic ``S''-shaped
curve. These functions are \textbf{scalar-to-scalar} probability mappings:
they are used to squash real values to $[0,1]$. 

\subsubsection*{Binary step function}

An example is the \textbf{binary step function}, 
also known as \textbf{Heaviside step function}, 
\begin{equation*}
\mathrm{step}(u) \coloneqq
\begin{cases} 
1, & u \ge 0 \\ 
0, & u < 0
\end{cases}.
\end{equation*}
It is a mapping from $\R$ to $\{0, 1\}$. Unfortunately, it has a discontinuity:
a jump in its graph at $u=0$.
Moreover, because the function is constant at
all other points, it has zero derivative at these points, which makes it
difficult to use as part of a neural network trained with backpropagation.

\subsubsection*{Logistic function}

A better sigmoid is the \textbf{logistic function}, 
which is a mapping from $\R$ to $(0, 1)$ and is defined as
\begin{align*}
\mathrm{logistic}(u) 
&\coloneqq \frac{1}{1 + e^{-u}}  \\
&= \frac{e^u}{1 + e^u} \\
&= \frac{1}{2} + \frac{1}{2} \tanh\left(\frac{u}{2}\right).
\end{align*}
It maps $(-\infty, 0)$ to $(0, 0.5)$, $[0, +\infty)$ to $[0.5, 1)$ and it
satisfies $\mathrm{logistic}(0) = 0.5$. It can therefore be seen as mapping from
real values to probability values. The logistic can be seen as a differentiable
approximation to the discontinuous binary step function $\mathrm{step}(u)$.
The logistic function can be shown to be the derivative of softplus, i.e.,
for all $u \in \RR$
\begin{equation*}
\mathrm{softplus}'(u) = \mathrm{logistic}(u).
\end{equation*}
Two important properties of the logistic function are that
for all $u \in \RR$
\begin{equation*}
    \mathrm{logistic}(-u) = 1 - \mathrm{logistic}(u)
\end{equation*}
and
\begin{align*}
\mathrm{logistic}'(u) 
&= \mathrm{logistic}(u) \cdot \mathrm{logistic}(-u) \\
&= \mathrm{logistic}(u) \cdot (1 - \mathrm{logistic}(u)).
\end{align*}
Other sigmoids are possible; see \cref{inf_conv:sec:sigmoid}.

\subsection{Probability mappings: argmax and softargmax}

It is often useful to transform a real vector into a vector of probabilities.
This is a mapping from $\R^M$ to the probability simplex, defined by
\begin{equation*}
\triangle^M \coloneqq 
\left\{\piv \in \R^M \colon  \forall j \in [M],\ \pi_j \geq 0,\
\sum_{j=1}^M \pi_j= 1\right\}.
\end{equation*}
Two examples of such \textbf{vector-to-vector} probability mappings
are the argmax and the softargmax.

\subsubsection*{Argmax}

The argmax operator is defined by
\begin{equation*}
\mathrm{argmax}(\u) \coloneqq \phi\left(\argmax_{j \in [M]} u_j\right)
\in \{\e_1, \dots, \e_M\},
\end{equation*}
where $\phi(j)$ denotes the one-hot encoding of an integer $j \in
[M]$, that is, 
\begin{equation*}
\phi(j) 
\coloneqq (0, \ldots, 0, \underbrace{1}_j, 0, \ldots, 0) 
= \e_j \in \{0,1\}^M.
\end{equation*}
This mapping puts all the probability mass onto a single coordinate (in case of
ties, we pick a single coordinate arbitrarily). Unfortunately, this mapping is a
discontinuous function. 

\subsubsection*{Softargmax}

As a differentiable everywhere relaxation, 
we can use the
\textbf{softargmax} defined by
\begin{equation*}
\mathrm{softargmax}(\u) \coloneqq \frac{\exp(\u)}{\sum_{j=1}^M \exp(u_j)}
\in \mathrm{relint}(\triangle^M).
\end{equation*}
This operator is commonly known in the literature as
\textit{softmax} but this is a misnomer: this operator
really defines a differentiable relaxation of the argmax. 
The output of the softargmax belongs to the relative interior of the 
probability simplex $\mathrm{relint}(\triangle^M) = \{\piv \in \triangle^M
\colon \piv > \zeros\}$, meaning
that it can never reach the borders of the simplex. 
If we denote $\piv =
\mathrm{softargmax}(\u)$, this means that $\pi_j \in (0,1)$,
that is,
$\pi_j$ can never be exactly $0$ or $1$.
The $\mathrm{softargmax}$ is the gradient of log-sum-exp,
\begin{equation*}
\nabla \mathrm{logsumexp}(\u) = \mathrm{softargmax}(\u). 
\end{equation*}
The $\mathrm{softargmax}$ can be seen as a generalization of the logistic
function, as we have for all $u \in \RR$
\begin{equation*}
[\mathrm{softargmax}((u, 0))]_1 = \mathrm{logistic}(u).
\end{equation*}

\begin{boxrem}{Degrees of freedom and invertibility of softargmax}
The softargmax operator satisfies the property for all 
$\u \in \RR^M$ and $c \in \RR$
\begin{equation*}
\piv \coloneqq \mathrm{softargmax}(\u) = \mathrm{softargmax}(\u + c \ones).
\end{equation*}
This means that the softargmax operator has $M-1$ degrees of freedom
and is a non-invertible function.
However, due to the above property, without loss of generality,
we can impose $\u^\top \ones = \sum_{i=1}^M u_i = 0$
(if this is not the case, we simply do $u_i \leftarrow u_i - \bar{u}$,
where $\bar{u} \coloneqq \frac{1}{M} \sum_{j=1}^M u_j$).
Using this constraint together with
\begin{equation*}
\log \pi_i = u_i - \log \sum_{j=1}^M \exp(u_j),
\end{equation*}
we then obtain
\begin{equation*}
\sum_{i=1}^M \log \pi_i = - M \log \sum_{j=1}^M \exp(u_j)
\end{equation*}
so that
\begin{equation*}
u_i =
[\mathrm{softargmax}^{-1}(\piv)]_i = 
\log \pi_i - \frac{1}{M} \sum_{j=1}^M \log \pi_j.
\end{equation*}
\end{boxrem}

\begin{figure}[t]
\centering
\includegraphics[scale=2.5]{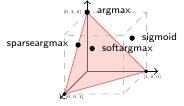}
\caption{The argmax, the softargmax and the sparseargmax (see
    \cref{inf_conv:sec:relaxed_argmax}) are mappings from $\RR^M$ to the
    probability simplex $\triangle^M \coloneqq \{\piv \in \RR^M_+ \colon \langle
    \piv, \ones \rangle =1\}$. More precisely, the argmax maps to vertices of
    the probability simplex, the set of one-hot vectors $\{\e_1, \dots, \e_M\}$,
    the softargmax maps to the relative interior $\mathrm{relint}(\triangle^M) =
    \{\piv \in \triangle^M \colon \piv > \zeros\}$ and the sparseargmax maps to
    the entire probability simplex $\triangle^M$, including sparse probability
vectors.  Sigmoids applied element-wise return vectors in the hypercube
$[0,1]^M$, whose vertices are the $2^M$ vectors in $\{0,1\}^M$.}
\label{neural_net:fig:simplex}
\end{figure}

\section{Normalization layers}
\label{neural_nets:sec:normalization}

Intermediate states in neural networks, such as activations, can often attain a
wide range of different values, potentially making it difficult for gradient
descent to converge.  To remedy this issue, we can introduce normalization
layers at suitable locations in the network.  In this section, we present the
two most popular ones: batch normalization and layer normalization.

\subsection{Batch normalization}

Suppose we are given a batch of intermediate variables (such as activations or
states)
$\s_1,\dots,\s_B \in \RR^D$,
where $\s_i \coloneqq (s_{i,1},\dots, s_{i,D})$,
obtained by applying some function to $B$ samples drawn from the
training set (we omit the dependency on the layer index $k$ for clarity).
In batch normalization \citep{ioffe2015batch},
we normalize the values by
calculating the \textbf{standard score} (\aka \textbf{z-score})
across batch samples,
\begin{align*}
    \mu_{\blue{j}} &\coloneqq \frac{1}{B}
    \sum_{\blue{i=1}}^{\blue{B}} s_{i,j} 
\quad \forall j \in [D] \\
\sigma_{\blue{j}}^2 &\coloneqq \frac{1}{B} \sum_{\blue{i=1}}^{\blue{B}} (s_{i,j} - \mu_j)^2 
\quad \forall j \in [D]\\
\widehat{s}_{i,j} &\coloneqq \frac{s_{i,j} - \mu_j}{\sigma_j}
\quad i \in [B], j \in [D].
\end{align*}
Here, the means $\mu_j$ and the standard deviations $\sigma_j$ are 
computed for each feature $j \in [D]$ across samples $i \in [B]$ in the batch.
In practice, we often add a small value $\varepsilon > 0$ to $\sigma_j$ 
to avoid division by zero or numerical instabilities. 
Moreover, we often rescale the values as
\begin{equation*}
\widetilde{s}_{i,j}
\coloneqq
\beta_j + \gamma_j \widehat{s}_{i,j}
\quad i \in [B], j \in [D],
\end{equation*}
where the means
$\betav \coloneqq (\beta_1, \dots, \beta_D)$
and standard deviations
$\gammav \coloneqq (\gamma_1, \dots, \gamma_D)$
are learnable parameters.

One issue with batch normalization is that the means $\mu_j$ and standard
deviations $\sigma_j$
cannot be computed at inference time, as there is no notion of 
training batch (a single sample would lead to a variance of $0$). 
To address this issue, we can estimate means
$\widehat{\mu}_j$ and standard deviations $\widehat{\sigma}_j$ 
across the whole training set during the course of training,
usually using a \textbf{running average}. 
A practical batch normalization implementation therefore needs to maintain $D$
mean and standard deviation statistics, so as to be able to use them at
inference time. 

\subsection{Layer normalization}
\label{neural_nets:sec:layer_norm}

As an alternative, in layer normalization \citep{ba2016layer},
we instead standardize the values by summing across features,
\begin{align*}
\mu_{\blue{i}} &\coloneqq \frac{1}{D} \sum_{\blue{j=1}}^{\blue{D}} s_{i,j} 
\quad \forall i \in [B] \\
\sigma_{\blue{i}}^2 &\coloneqq \frac{1}{D} \sum_{\blue{j=1}}^{\blue{D}} (s_{i,j} - \mu_i)^2 
\quad \forall i \in [B] \\
\widehat{s}_{i,j} &\coloneqq \frac{s_{i,j} - \mu_i}{\sigma_i}
\quad i \in [B], j \in [D].
\end{align*}
This time, the means $\mu_i$ and the standard deviations $\sigma_i$ are 
computed for each sample $i \in [B]$ across features $j \in [D]$
(as before, we often add a small value $\varepsilon$ to $\sigma_i$).
Similarly to batch normalization, we often rescale the values as
\begin{equation*}
\widetilde{s}_{i,j}
\coloneqq
\beta_j + \gamma_j \widehat{s}_{i,j}
\quad i \in [B], j \in [D],
\end{equation*}
where the means
$\betav \coloneqq (\beta_1, \dots, \beta_D)$
and standard deviations
$\gammav \coloneqq (\gamma_1, \dots, \gamma_D)$
are learnable parameters.
A key advantage of layer normalization compared to batch normalization is that
it is well defined at inference time, 
since it is applied on a per-sample basis and does not rely
on the notion of a training batch.
As a result, we can view layer normalization as a function that to any $\s_i \in
\RR^D$ (regardless of whether it is part of a batch or not) associates
\begin{equation*}
\widetilde{\s_i} \coloneqq \mathrm{LayerNorm}(\s_i).
\end{equation*}

\section{Residual neural networks}\label{neural_nets:sec:residual}

We now discuss another feedforward network parametrization: residual neural
networks. Consider a feedforward network with $K+1$ layers $f_1$, $\dots$,
$f_K$, $f_{K+1}$. Surely, as long as $f_{K+1}$ can exactly represent the
identity function, the set of functions that this feedforward network can
express should be a superset of the functions that $f_1$, $\dots$, $f_K$ can
express. In other words, depth should in theory not hurt the expressive power of
feedforward networks. Unfortunately, the assumption that each $f_k$ can
exactly represent the identity function may not hold in practice. This means
that deeper networks can sometimes be more difficult to train than shallower
ones, making the accuracy saturate or degrade as a function of depth. 

The key idea
of residual neural networks \citep{he_2016}
is to design layers $f_k$, called \textbf{residual blocks},
that make it easier to represent the identity function. Formally, a residual
block takes the form
\begin{equation*}
\sv_k 
= f_k(\sv_{k-1}, \wv_k) 
\coloneqq \sv_{k-1} + h_k(\sv_{k-1}, \wv_k).
\end{equation*}
The function $h_k$ is called \textbf{residual}, since it models the difference
$\sv_k - \sv_{k-1}$. The addition with $\sv_{k-1}$ is often called a
\textbf{skip connection}. As long as it is easy to adjust $\wv_k$ so that
$h_k(\sv_{k-1}, \wv_k) = \zeros$, $f_k$ can freely become the identity
function. For instance, if we use
\begin{equation*}
h_k(\sv_{k-1}, \wv_k) 
\coloneqq \Cv_k a_k(\W_k \sv_{k-1} + \bv_k) + \dv_k,
\end{equation*}
where $\wv_k \coloneqq (\W_k, \bv_k, \Cv_k, \dv_k)$, 
it suffices to set $\Cv_k$ and
$\dv_k$ to a zero matrix and vector. Residual blocks are known to remedy the
so-called vanishing gradient problem.

Many papers and software packages include an additional activation and
instead define the residual block as
\begin{equation*}
\sv_k 
= f_k(\sv_{k-1}, \wv_k) 
\coloneqq a_k(\sv_{k-1} + h_k(\sv_{k-1}, \wv_k)),
\end{equation*}
where $a_k$ is typically chosen to be the ReLU activation. Whether to
include this additional activation or not is essentially a modelling choice. In
practice, residual blocks may also include additional operations such as batch
norm and convolutional layers.

\section{Recurrent neural networks}
\label{seq_net:sec:rnn}

\begin{figure}
  \centering
  \begin{subfigure}[b]{0.45\linewidth}
      \centering
      \includegraphics[width=0.9\linewidth]{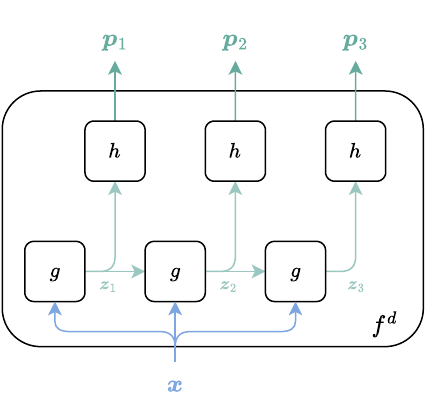}
      \caption{One to many (decoder)}
      \label{neural_nets:fig:one_to_many}
  \end{subfigure}
  \hfill
  \begin{subfigure}[b]{0.46\linewidth}
      \centering
      \includegraphics[width=\linewidth]{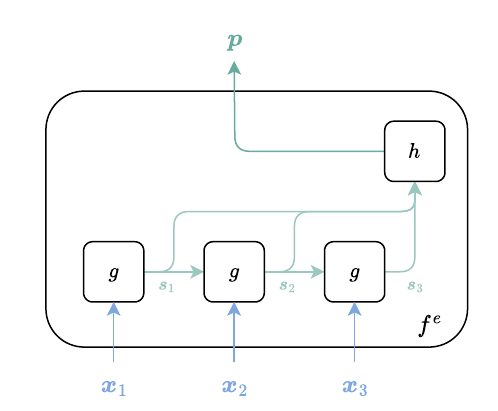}
      \caption{Many to one (encoder)}
      \label{neural_nets:fig:many_to_one}
  \end{subfigure}
  \hfill
  \begin{subfigure}[b]{0.45\linewidth}
    \centering
    \includegraphics[width=\linewidth]{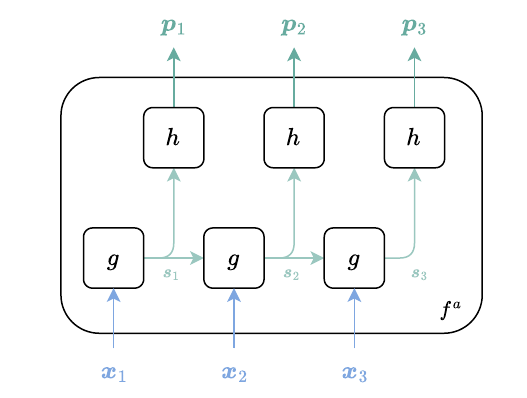}
    \caption{Sequence to sequence aligned}
    \label{neural_nets:fig:seq_to_seq_aligned}
  \end{subfigure}
  \hfill
  \begin{subfigure}[b]{0.45\linewidth}
    \centering
    \includegraphics[width=\linewidth]{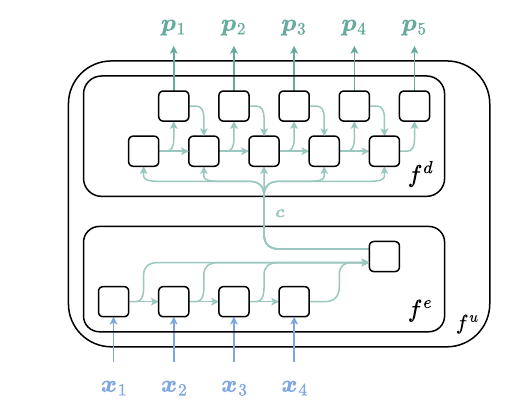}
    \caption{Sequence to sequence unaligned}
    \label{neural_nets:fig:seq_to_seq_unaligned}
  \end{subfigure}
  \caption{Recurrent neural network architectures}
  \label{neural_nets:fig:rnns}
\end{figure}

Recurrent neural networks (RNNs) are a class of neural networks that operate on
sequences of vectors, either as input, output or both. Their actual
parametrization depends on the setup but the core idea is to maintain a
\textbf{state vector} that is updated from step to step by a recursive
function that uses \textbf{shared parameters} across steps. Unrolling this
recursion defines a valid computational graph, as we will see in
\cref{chap:auto_diff}. We distinguish between the following setups illustrated
in \cref{neural_nets:fig:rnns}:
\begin{itemize}
\item Vector to sequence (one to many): \\
    $f^d \colon \R^D \times \R^P \to \R^{L \times M}$
\item Sequence to vector (many to one): \\
    $f^e \colon \R^{L \times D} \times \R^P \to \R^M$
\item Sequence to sequence (many to many, aligned): \\
    $f^a \colon \R^{L \times D} \times \R^P \to \R^{L \times M}$
\item Sequence to sequence (many to many, unaligned): \\
    $f^u \colon \R^{L \times D} \times \R^P \to \R^{L' \times M}$
\end{itemize}
where $L$ stands for length.
Note that we use the same number of parameters $P$ for each setup for notational
convenience, but this of course does not need to be the case.
Throughout this section,
we use the notation 
$\p_{1:L} \coloneqq (\p_1, \dots, \p_L)$
for a sequence of $L$ vectors.

\subsection{Vector to sequence}

In this setting, we define a \textbf{decoder} function $\p_{1:L} = f^d(\xv,
\wv)$ from an \textbf{input vector} 
$\x \in \R^D$ and parameters $\w \in \R^P$ to an
\textbf{output sequence} $\p_{1:L} \in \R^{L \times M}$. 
This is for instance useful for
image caption generation, where a sentence (a sequence of word embeddings) is
generated from an image (a vector of pixels). Formally, we may define $\p_{1:L}
\coloneqq f^d(\x, \w)$ through the recursion
\begin{equation*}
\begin{aligned}
    \z_l &\coloneqq g(\x, \z_{l-1}, \w_g) \quad l \in [L] \\
    \p_l &\coloneqq h(\z_l, \w_h) \quad l \in [L].
\end{aligned}
\end{equation*}
where $\wv \coloneqq (\w_g, \w_h, \z_0)$. The goal of $g$ is to update the
current \textbf{decoder state} $\z_l$ given the input $\x$, 
and the previous decoder state $\z_{l-1}$. 
The goal of $h$ is to generate the output $\p_l$ given the
current decoder state $\z_l$. Importantly, the parameters of $g$ and $h$ are
\textbf{shared across steps}. Typically, $g$ and $h$ are parametrized using
one-hidden-layer MLPs. Note that $g$ has multiple inputs; we discuss how to
deal with such cases in \cref{neural_nets:sec:mlp}.

\subsection{Sequence to vector}

In this setting, we define an \textbf{encoder} function $\p = f^e(\x_{1:L},
\wv)$ from an \textbf{input sequence} 
$\x_{1:L} \in \R^{L \times D}$ and parameters $\w
\in \R^P$ to an \textbf{output vector} $\p \in \R^M$. 
This is for instance useful for sequence classification, 
such as sentiment analysis. Formally, we may define $\p
\coloneqq f^e(\x_{1:L}, \w)$ using the recursion
\begin{equation*}
\begin{aligned}
\s_l &\coloneqq \gamma(\x_l, \s_{l-1}, \w_\gamma) \quad l \in [L] \\
\p &= \mathrm{pooling}(\s_{1:L})
\end{aligned}
\end{equation*}
where $\w \coloneqq (\w_\gamma, \s_0)$. The goal of $\gamma$ is similar to $g$,
except that it updates \textbf{encoder states} 
and does not take previous predictions as
input. 
The \textbf{pooling} function is typically parameter-less.
Its goal is to reduce a sequence to a vector. 
Examples include using the last state, the average of states and
the coordinate-wise maximum of states.

\subsection{Sequence to sequence (aligned)}

In this setting, we define a function $\p_{1:L} = f^a(\x_{1:L}, \wv)$ from an
\textbf{input sequence} $\x_{1:L} \in \R^{L \times D}$ 
and parameters $\w \in \R^P$ to an
\textbf{output sequence} $\p_{1:L} \in \R^{L \times M}$, 
which we assume to be of the \textbf{same length}. 
An example of application is part-of-speech tagging, where the goal is
to assign each word $\x_l$ to a part-of-speech (noun, verb, adjective, etc).
Formally, we may define $\p_{1:L} = f^a(\x_{1:L}, \w)$ as
\begin{equation*}
\begin{aligned}
\s_l &\coloneqq \gamma(\x_l, \s_{l-1}, \w_\gamma) \quad l \in [L] \\
\p_l &= h(\s_l, \w_h) \quad l \in [L] \\
\end{aligned}
\end{equation*}
where $\wv \coloneqq (\wv_\gamma, \wv_h, \sv_0)$. The functions $\gamma$ and
$h$ are similar as before.

\subsection{Sequence to sequence (unaligned)}

In this setting, we define a function 
$\p_{1:L'} = f^u(\x_{1:L}, \wv)$ 
from an \textbf{input sequence} $\x_{1:L} \in \R^{L \times D}$ 
and parameters $\w \in \R^P$ to an output sequence 
$\p_{1:L'} \in \R^{L' \times M}$, which potentially has a
\textbf{different length}. 
An example of application is machine translation, where the
sentences in the source and target languages do not necessarily have the same
length. Typically, 
$\p_{1:L'} = f^u(\x_{1:L}, \w)$ is defined as the following two steps
\begin{equation*}
\begin{aligned}
\c &\coloneqq f^e(\x_{1:L}, \w_e) \\
\p_{1:L'} &\coloneqq f^d(\c, \w_d)
\end{aligned}
\end{equation*}
where $\w \coloneqq (\w_e, \w_d)$, and where we reused the previously-defined
encoder $f^e$ and decoder $f^d$. 
Putting the two steps together, we obtain
\begin{equation*}
\begin{aligned}
\s_l &\coloneqq \gamma(\x_l, \s_{l-1}, \w_\gamma) \quad l \in [L] \\
\c &= \mathrm{pooling}(\s_{1:L}) \\
\z_l &\coloneqq g(\c, \p_{l-1}, \z_{l-1}, \w_g) \quad l \in [L'] \\
\p_l &\coloneqq h(\z_l, \w_h) \quad l \in [L'].
\end{aligned}
\end{equation*}

This architecture is aptly named the \textbf{encoder-decoder} architecture. Note
that we denoted the length of the target sequence as $L'$. However, in practice,
the target length can be input dependent and is often not known ahead of time.
To deal with this issue, the vocabulary (of size $D$ in our notation) is
typically augmented with an ``end of sequence'' (EOS) token so that, at
inference time, we know when to stop generating the output sequence. One
disadvantage of this encoder-decoder architecture, however, is that all the
information about the input sequence is contained in the \textbf{context
vector} $\c$,
which can therefore become a \textbf{bottleneck}. 

\section{Transformers}
\label{neural_nets:sec:transformers} 

Transformers \citep{vaswani2017attention} are one of the most successful recent
developments in deep learning. In this section, we review Transformers,
component by component.

\subsection{Attention}

The goal of an attention layer is to map a sequence of inputs
$\v_1,\dots,\v_L \in \RR^{D_v}$
to a sequence of outputs
$\u_1,\dots,\u_L \in \RR^{D_v}$,
of same dimension.
A natural idea is to define each output element $\u_i$ 
using a linear combination of the input elements,
\begin{equation*}
\u_i \coloneqq \sum_{j=1}^L a_{i,j} \v_j \in \RR^{D_v}.
\end{equation*}
We typically use a convex combination:
we assume that the combination weights are such that
$\a_i \coloneqq (a_{i,1},\dots,a_{i,L}) \in \triangle^L$.
This ensures that increasing a coefficient $a_{i,j}$ is made at the expense
of decreasing another coefficient $a_{i,j'}$, for $j \neq j'$.
Let us form the matrices 
$\V \in \RR^{L \times D_v}$ 
and
$\U \in \RR^{L \times D_v}$
by stacking
$\v_1,\dots,\v_L$
and
$\u_1,\dots,\u_L$,
seen as row vectors.
Let us also form the \textbf{attention matrix}
$\A \in \triangle^{L \times L}$
gathering the entries $a_{i,j}$,
where $\triangle^{L \times L}$ denotes the set of row-wise stochastic $L \times
L$ matrices.
Then, we can rewrite attention succinctly as
\begin{equation*}
\U = \A \V \in \RR^{L \times D_v}. 
\end{equation*}
Following \cref{ds:dict},
we can view attention as a \textbf{soft dictionary lookup}.
From this perspective, 
the matrix $\V \in \RR^{L \times D_v}$ plays the role of dictionary values,
and the attention matrix can be defined as
\begin{equation*}
\A \coloneqq \mathrm{softargmax}(\Q \K^\top) \in \triangle^{L \times L},
\end{equation*}
where $\Q \in \RR^{L \times D_k}$ and $\K \in \RR^{L \times D_k}$ play the roles
of queries and dictionary keys, respectively.
Intuitively, $\Q\K^\top \in \RR^{L \times L}$ can be seen as a similarity matrix
containing the similarities between queries and dictionary keys.
Putting everything together, we can define attention as
\begin{equation*}
\mathrm{Attention}(\Q, \K, \V) \coloneqq
\mathrm{softargmax}\left(\Q \K^\top\right)\V
\in \RR^{L \times D_v}.
\end{equation*}
In \textbf{masked attention}, we additionally incorporate a mask 
$\M \in \RR^{L \times L}$,
\begin{equation*}
\mathrm{Attention}(\Q, \K, \V; \M) \coloneqq
\mathrm{softargmax}\left((\Q \K^\top) \circ \M\right)\V
\in \RR^{L \times D_v},
\end{equation*}
where $\circ$ denotes the Hadamard product (element-wise multiplication).
The mask can be used to force some attention weights $a_{i,j}$ to be zero, by
setting the corresponding mask entry $m_{i,j}$ to $-\infty$.
For instance, in decoder-only architectures
(\cref{neural_nets:sec:decoder_only_transformer}), the mask will prove useful to
define \textbf{causal attention} for autoregressive models.

\begin{boxrem}{Scaled attention}
Practical implementations often use \textbf{scaled attention},
where a factor of $\frac{1}{\sqrt{D_k}}$ is used within the soft-argmax,
in order to reduce the variance \citep[Footnote 4]{vaswani2017attention}.
We omit this detail for clarity.
\end{boxrem}

\subsection{Self-attention}

Suppose we are given a sequence of feature vectors
$\x_1, \dots, \x_L \in \RR^D$,
that we gather as a matrix
$\X \in \RR^{L \times D}$.
To define the attention weights $a_{i,j}$, 
a natural idea is to consider the similarity between $\x_i$ and $\x_j$, as
measured by the inner product $\langle \x_i, \x_j \rangle$.
To ensure that $\a_i \coloneqq (a_{i,1},\dots,a_{i,L}) \in \triangle^L$,
we can then define
\begin{equation*}
a_{i,j} \coloneqq \frac{\exp(\langle \x_i, \x_j \rangle)}{
\sum_{j'=1}^L \exp(\langle \x_i, \x_{j'} \rangle)}
\in (0,1).
\end{equation*}
In matrix notation, this can be written
\begin{equation*}
\A \coloneqq \mathrm{softargmax}(\X \X^\top) \in \triangle^{L \times L},
\end{equation*}
where $\mathrm{softargmax}$ is applied in a row-wise fashion.
The matrix $\X \X^\top \in \RR^{L \times L}$ is the \textbf{Gram matrix}
associated with the row vectors $\x_1, \dots, \x_L$.
In the notation of the previous section, this corresponds to using
$\mathrm{Attention}(\Q, \K, \V)$ with $\Q = \K = \V = \X$. In other words,
the elements of the sequence ``pay attention'' to each other.
This is called \textbf{self-attention}.

So far, the formulation we described is parameter-free. 
To give more expressive power to the self-attention layer,
\citet{vaswani2017attention} proposed instead to define $\Q$, $\K$ and $\V$ by
projecting $\X$ as
\begin{align*}
    \Q &\coloneqq \X \W^Q \in \RR^{L \times D_k} \\
    \K &\coloneqq \X \W^K  \in \RR^{L \times D_k}\\
    \V &\coloneqq \X \W^V \in \RR^{L \times D_v},
\end{align*}
using the learned weight matrices 
\begin{align*}
    \W^Q &\in \RR^{D \times D_k} \\
    \W^K &\in \RR^{D \times D_k} \\
    \W^V &\in \RR^{D \times D_v}.
\end{align*}
Importantly, the size of the weight matrices is independent of the length $L$ of
the sequences. This allows self-attention to work with sequences of arbitrary
length.

\subsection{Multi-head attention}
\label{neural_net:sec:multi_head_attention}

In order to be able to capture multiple patterns of attention,
\citet{vaswani2017attention} found it beneficial to define 
$H$ attention heads
\begin{equation*}
\Y_i \coloneqq \mathrm{Attention}(\Q_i, \K_i, \V_i; \M)
\in \RR^{L \times D_v},
\end{equation*}
where
\begin{align*}
    \Q_i &\coloneqq \X \W^Q_i \in \RR^{L \times D_k} \\
    \K_i &\coloneqq \X \W^K_i \in \RR^{L \times D_k} \\
    \V_i &\coloneqq \X \W^V_i \in \RR^{L \times D_v},
\end{align*}
using the learned weight matrices
\begin{align*}
    \W^Q_i &\in \RR^{D \times D_k} \\
    \W^K_i &\in \RR^{D \times D_k} \\
    \W^V_i &\in \RR^{D \times D_v}.
\end{align*}
Let us denote the concatenation of the $H$ attention heads by
\begin{equation*}
\mathrm{Concat}(\Y_1, \dots, \Y_H) \in \RR^{L \times H D_v}.
\end{equation*}
\citet{vaswani2017attention} then define multi-head attention as
\begin{align*}
\mathrm{MultiheadAttention}(\X; \M)
&\coloneqq
\mathrm{Concat}(\Y_1, \dots, \Y_H) \W^O  \\
&= \sum_{i=1}^H \mathrm{Attention}(\X \W^Q_i, \X \W_i^K, \X \W_i^V) \W_i^O \\
&\in \RR^{L \times D}
\end{align*}
where
$\W^O \in \RR^{H D_v \times D}$
is a learned matrix.
We summarize the procedure in \cref{net:algo:multihead_attention}.
We use a for loop for the sake of clarity; a GPU-friendly implementation would
instead be based on a single matrix multiplication.

\begin{algorithm}[t]\caption{Multi-head attention with $H$ attention heads
\label{net:algo:multihead_attention}}
\begin{algorithmic}[1]
\Statex{\bf Input:} $\X \in \RR^{L \times D}$, 
optional mask $\M \in \RR^{L \times L}$
\Statex{\bf Parameters:} $\{\W_i^Q\}_{i=1}^H$, $\{\W_i^K\}_{i=1}^H$, 
$\{\W_i^V\}_{i=1}^H$, $\W^O$
\For {$i \coloneqq 1, \ldots, H$} 
\State $\Q_i \coloneqq \X \W^Q_i \in \RR^{L \times D_k}$
\State $\K_i \coloneqq \X \W^K_i \in \RR^{L \times D_k}$
\State $\V_i \coloneqq \X \W^V_i \in \RR^{L \times D_v}$
\State $\Y_i \coloneqq \mathrm{Attention}(\Q_i, \K_i, \V_i; \M) 
\in \RR^{L \times D_v}$
\EndFor 
\State $\Y \coloneqq \mathrm{Concat}(\Y_1, \dots, \Y_H) \W^O$
\Statex {\bf Output:} $\mathrm{MultiheadAttention}(\X; \M) 
\coloneqq \Y \in \RR^{L \times D}$
\end{algorithmic}
\end{algorithm}

\subsection{Transformer layer}

To improve training efficiency,
we can introduce residual connections and a first layer normalization 
(\cref{neural_nets:sec:layer_norm}),
to define
\begin{equation*}
\Z \coloneqq \mathrm{LayerNorm}_1(\mathrm{MultiheadAttention}(\X; \M) + \X) 
\in \RR^{L \times D}.
\end{equation*}
To improve expressiveness, a Transformer layer further uses an MLP and a second
layer normalization,
\begin{equation*}
\X \leftarrow \mathrm{LayerNorm}_2(\mathrm{MLP}(\Z) + \Z) \in \RR^{L \times D}.
\end{equation*}
The subscripts in $\mathrm{LayerNorm}$ are used to emphasize that the two layers
use their own parameters. In addition, normalization is applied in an
element-wise (token-wise) fashion.
The MLP block typically uses a single hidden layer with an activation function
$\sigma$, such as a GELU (Gaussian error linear unit); 
see \cref{net:algo:mlp_transformer}. Because of the use of residual connections,
the input and output dimensions of the MLP block must be the same.

\begin{algorithm}[t]\caption{Transformer's MLP block 
\label{net:algo:mlp_transformer}}
\begin{algorithmic}[1]
\Statex{\bf Input:} $\Z \in \RR^{L \times D}$
\Statex{\bf Parameters:} $\W_1 \in \RR^{D \times D_1}, \W_2 \in \RR^{D_1 \times
D}$ 
\State $\Z \leftarrow \Z \W_1 \in \RR^{L \times D_1}$
\State $\Z \leftarrow \sigma(\Z) \in \RR^{L \times D_1}$
\State $\Z \leftarrow \Z \W_2 \in \RR^{L \times D}$
\Statex {\bf Output:} $\Z \in \RR^{L \times D}$
\end{algorithmic}
\end{algorithm}

\begin{boxrem}{Post-normalization vs. pre-normalization}
Our description of the Transformer layer follows the original formulation
\citep{vaswani2017attention}, which uses post-normalization. Some
implementations instead rely on pre-normalization, that is,
\begin{align*}
\Z &\coloneqq \mathrm{MultiheadAttention}(\mathrm{LayerNorm}_1(\X); \M) + \X 
\in \RR^{L \times D} \\
\X &\leftarrow \mathrm{MLP}(\mathrm{LayerNorm}_2(\Z)) + \Z \in \RR^{L \times D}.
\end{align*}
\end{boxrem}

\subsection{Transformer block}
\label{neural_net:sec:transformer_block}

A Transformer layer can be iterated $K$ times (each time with different
parameters) to define a Transformer block of depth $K$. 
Keeping the dependency on parameters implicit,
we can see a Transformer, with an optional mask $\M \in \RR^{L \times L}$,
as a function 
\begin{equation*}
\X \mapsto \mathrm{Transformer}(\X; \M) 
= \mathrm{Transformer}(\x_1, \dots, \x_L; \M) 
\end{equation*}
from $\RR^{L \times D}$ to $\RR^{L \times D}$. 
The Transformer takes a sequence $\X \in \RR^{L \times D}$ and uses the
inter-dependencies between sequence elements to produce a representation of that
sequence.
We summarize the procedure in \cref{net:algo:transformer}.
The $\mathrm{MultiheadAttention}$, $\mathrm{LayerNorm}$ and $\mathrm{MLP}$
blocks are indexed by $k$ to emphasize that their parameters are different for
each iteration.

\subsubsection*{Number of parameters and computational complexity}

The Transformer layer can be seen as a function from $\RR^{L \times D}$ to
$\RR^{L \times D}$. 
To offer the same function signature,
a standard multi-layer perceptron would have needed
$O(L^2 D^2)$ parameters and a forward pass through the network would have had a
time complexity of $O(L^2 D^2)$ as well. In contrast, a Transformer layer has
$O(D^2)$ parameters. Self-attention has a complexity of $O(L^2 D)$ and the final
MLP layer has a complexity of $O(L D^2)$.

\begin{algorithm}[t]\caption{Transformer block of depth $K$
\label{net:algo:transformer}}
\begin{algorithmic}[1]
\Statex{\bf Input:} $\X \in \RR^{L \times D}$,
optional mask $\M \in \RR^{L \times L}$
\Statex{\bf Parameters:} Multi-head attention parameters, MLP parameters and
layer norm parameters (each depth $k$ uses different parameters)
\For {$k \coloneqq 1, \ldots, K$} 
\State $\Y \coloneqq \mathrm{MultiheadAttention}_k(\X; \M) \in \RR^{L \times D}$
\State $\Z \coloneqq \mathrm{LayerNorm}_{k,1}(\Y + \X) \in \RR^{L \times D}$
\Comment{Residual connection}
\State $\X \leftarrow \mathrm{LayerNorm}_{k,2}(\mathrm{MLP}_k(\Z) + \Z) 
\in \RR^{L \times D}$
\Comment{MLP layer}
\EndFor 
\Statex {\bf Output:} $\X \in \RR^{L \times D}$
\end{algorithmic}
\end{algorithm}

\subsection{Token encoding}
\label{neural_nets:sec:input_encoding} 

Text can be represented as a sequence $x_1,\dots,x_L$ of \textbf{discrete}
symbols, often called tokens. Here, each token $x_i \in [M]$, where $M$ is the
vocabulary size, can correspond to words, subwords, or even individual
characters, depending on the tokenization procedure.
To obtain a sequence of \textbf{continuous} vectors $\x_1, \dots, \x_L$,
where $\x_i \in \RR^D$,
it is common to use an \textbf{embedding layer}. This layer transforms $x_i \in
[M]$ into $\x_i \in \RR^D$ using the linear projection
\begin{equation*}
\x_i \coloneqq \W^E \e_{x_i},
\end{equation*}
where $\e_j \in \RR^M$ is the one-hot encoding of $j \in [M]$ and $\W^E \in
\RR^{D \times M}$ is a learnable embedding matrix. The column $\W^E_{:, j} \in
\RR^D$ can be interpreted as the continuous representation of token $j \in [V]$.
Usually, the vocabulary includes special tokens such as ``BOS''
(beginning of sequence), ``EOS'' (end of sequence) and ``PAD'' (padding token,
to ensure that the sequence is of length $L$).

\subsection{Positional encoding}
\label{neural_net:sec:positional_encoding}

Due to their parameterization, Transformers are \textbf{equivariant} with
respect to permutations: permuting the input sequence and applying the
Transformer is equivalent to applying the Transformer and permuting the output
sequence. Formally, for any permutation matrix $\cP$ of size 
$L \times L$ and input sequence $\x_1, \dots, \x_L$ seen as a matrix $\X$,
we have
\begin{equation*}
\mathrm{Transformer}(\cP \X) = \cP \mathrm{Transformer}(\X).
\end{equation*}
This means that Transformers treat an ordered sequence as a \textbf{multi-set} 
(a modification of the concept of a set that allows for multiple instances 
of its elements).
To leverage order information in a Transformer, several approaches are possible
\citep{dufter2022position}: adding positional encoding (either absolute or
relative), modifying the attention matrix and pre-processing the input 
sequence with an RNN. To add positional encoding, one typically adds a vector
$\p_i \in \RR^D$ representing the position $i \in [L]$ to the corresponding
element $\x_i \in \RR^D$,
\begin{equation*}
\x_i \leftarrow \x_i + \p_i.
\end{equation*}
An ideal positional encoding should work with any sequence length.

\subsubsection*{Learned positional encoding}

Similarly to the token encoding, a simple way to define a positional encoding is
to use an embedding layer.
Each position $i \in [L]$ is transformed into a position vector $\p_i \in \RR^D$
by
\begin{equation*}
\p_i \coloneqq \W^P \e_i
\end{equation*}
where $\e_i \in \RR^L$ is the one-hot encoding of $i \in [L]$
and $\W^P \in \RR^{D \times L}$ is a learnable embedding matrix.
The column $\W^P_{:, i} \in \RR^D$ can be interpreted as the continuous
representation of position $i \in [L]$.

\subsubsection*{Sinusoidal positional encoding}

\begin{figure}[t]
\centering
\includegraphics[scale=0.4]{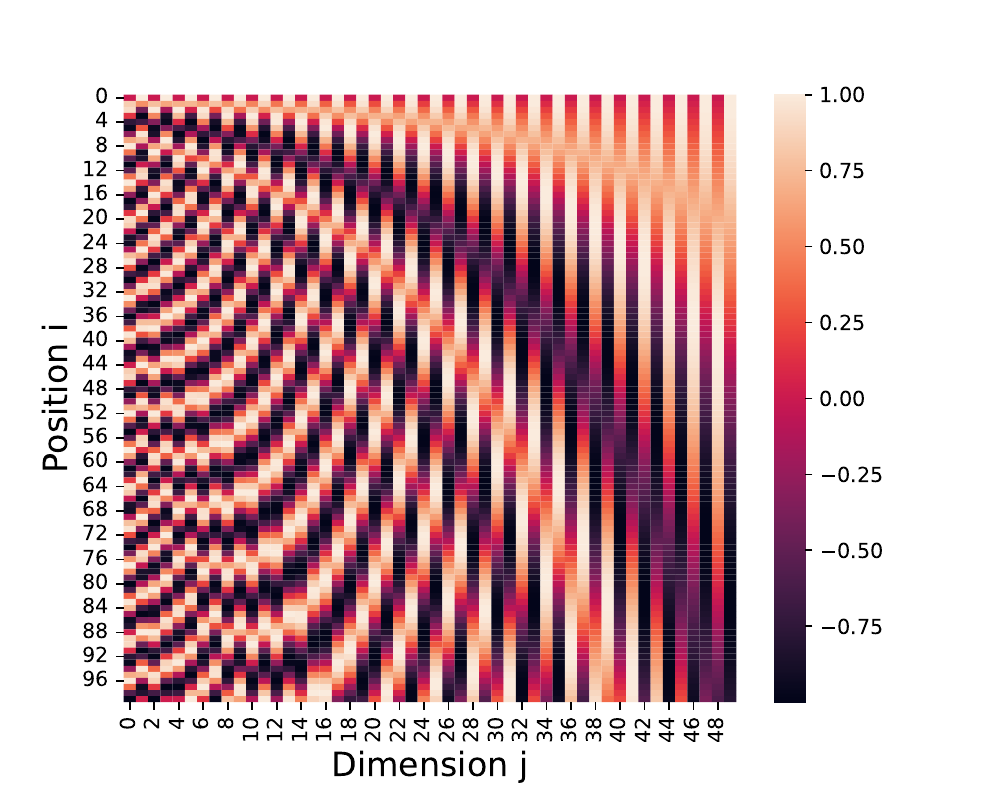}
\caption{Heatmap of a positional encoding matrix $\bm{P} \in \RR^{L \times D}$,
with $L=100$, $D=50$, and $N=30$. The values belong to the range $[-1,1]$.}
\label{neural_net:fig:positional_encoding_matrix}
\end{figure}

Instead of learning the positional encoding, we can define it in a heuristic
manner. For instance,
\citet{vaswani2017attention} proposed the \textbf{sinusoidal positional
encoding}
$\p_i \coloneqq (p_{i,1}, \dots, \p_{i,D})$,
defined by
\begin{equation*}
p_{i,j} \coloneqq 
\begin{cases}
\sin(\omega_j \cdot i) &\mbox{ if } j \text{ is even} \\
\cos(\omega_{j-1} \cdot i) &\mbox{ if } j \text{ is odd}
\end{cases} \in [-1,1],
\end{equation*}
where
\begin{equation*}
\omega_j 
\coloneqq \frac{1}{N^{\frac{j}{D}}} 
= N^{-\frac{j}{D}}
\end{equation*}
and where $N$ is a constant, set to $N \coloneqq 10000$ by the authors.
We can gather the values in a position encoding matrix $\bm{P} \in \RR^{L \times
D}$, as depicted in \cref{neural_net:fig:positional_encoding_matrix}.

\begin{figure}[t]
\centering
\includegraphics[scale=0.4]{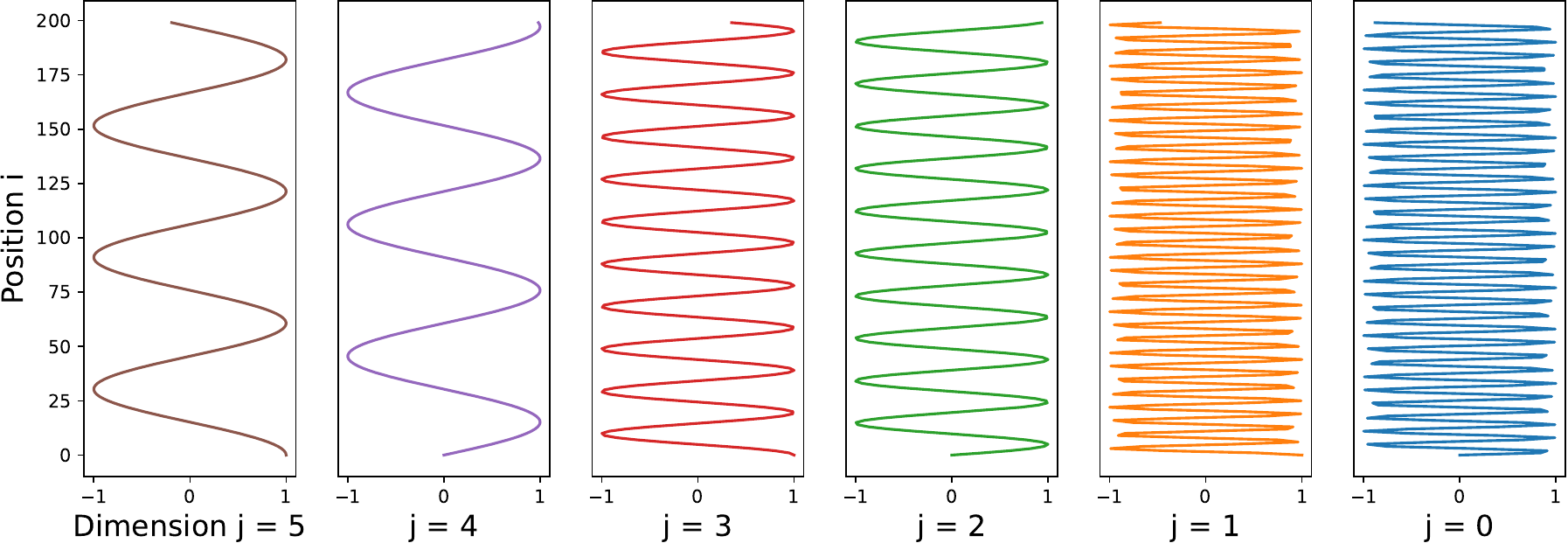}
\caption{Using a sinusoidal positional encoding, each coordinate $j \in [D]$ is
associated with a sine or cosine wave. In analogy with a binary encoding, in
which lower bits alternate between $0$ and $1$ more frequently than higher bits,
the wave associated with $j$
oscillates more frequently when $j$ is small.
}
\label{neural_net:fig:positional_encoding_waves}
\end{figure}

On first sight, sinusoidal positional encoding may seem a bit mysterious.
To gain some intuition, it is useful to compare it to a discrete
\textbf{binary encoding} of integers,
\begin{align*}
    7 &\leftrightarrow \blueO \blueO \blueO \\
    6 &\leftrightarrow \blueO \blueO \redZ \\
    5 &\leftrightarrow \blueO \redZ \blueO \\
    4 &\leftrightarrow \blueO \redZ \redZ \\
    3 &\leftrightarrow \redZ \blueO \blueO \\
    2 &\leftrightarrow \redZ \blueO \redZ \\
    1 &\leftrightarrow \redZ \redZ \blueO \\
    0 &\leftrightarrow \redZ \redZ \redZ.
\end{align*}
We see that the bits alternate more frequently between $\redZ$ and $\blueO$
as we go from right to left.
The sinusoidal positional encoding achieves a similar behavior, but is
continuous.
Indeed, each coordinate $j \in [D]$ is associated with a sine wave (when $j$ is
even) or a cosine wave (when $j$ is odd).
The \textbf{wavelength} or \textbf{period} of the wave associated with $j$ is
$\frac{2 \pi}{\omega_j}$ and its \textbf{frequency} is
$\frac{\omega_j}{2 \pi}$. Since $\omega_j$ is a decreasing function of $j$,
we see that the waves oscillate more frequently when $j$ is small.
This behavior is illustrated in \cref{neural_net:fig:positional_encoding_waves}.

Stacking sine and cosine waves has been used for creating random Fourier
features \citep{rahimi2007random,sutherland2015error}.
These use waves of random frequencies, while sinusoidal positional encoding uses
waves of increasing frequency. These approaches therefore mainly differ in the
way we choose wave frequencies.

\subsubsection*{Recovering relative positional information}

An important property of the sinusoidal positional encoding is that
$\p_{i + \delta}$ can be represented as
a linear function of $\p_i$, for any offset $\delta$
\citep{vaswani2017attention,zhang2021dive}. 
The model should therefore be able
to easily learn to attend by relative positions.
Indeed, using the trigonometric identities for angle sums
\begin{align*}
\cos(\alpha + \beta) &= \cos(\alpha) \cos(\beta) - \sin(\alpha) \sin(\beta) \\
\sin(\alpha + \beta) &= \sin(\alpha) \cos(\beta) + \cos(\alpha) \sin(\beta),
\end{align*}
we have for $j$ even
\begin{align*}
\begin{pmatrix}
p_{i+\delta,j+1} \\
p_{i+\delta,j}
\end{pmatrix}
&= 
\begin{pmatrix}
\cos(\omega_j (i + \delta)) \\
\sin(\omega_j (i + \delta))
\end{pmatrix} \\
&=
\begin{pmatrix}
\cos(\omega_j \cdot i) \cos(\omega_j \cdot \delta) -
\sin(\omega_j \cdot i) \sin(\omega_j \cdot \delta) \\
\sin(\omega_j \cdot i) \cos(\omega_j \cdot \delta) + 
\cos(\omega_j \cdot i) \sin(\omega_j \cdot \delta) 
\end{pmatrix} \\
&=
\begin{pmatrix}
\cos(\omega_j \cdot \delta) & -\sin(\omega_j \cdot \delta) \\
\sin(\omega_j \cdot \delta) & \cos(\omega_j \cdot \delta)
\end{pmatrix}
\begin{pmatrix}
\cos(\omega_j \cdot i) \\
\sin(\omega_j \cdot i) 
\end{pmatrix} \\
&=
\begin{pmatrix}
\cos(\omega_j \cdot \delta) & -\sin(\omega_j \cdot \delta) \\
\sin(\omega_j \cdot \delta) & \cos(\omega_j \cdot \delta)
\end{pmatrix}
\begin{pmatrix}
p_{i,j+1} \\
p_{i,j}
\end{pmatrix}
\end{align*}
Therefore, $(p_{i,j+1}, p_{i,j})$ can be linearly projected to
$(p_{i+\delta,j+1}, p_{i+\delta,j})$, by applying a \textbf{rotation matrix}.
This allows a Transformer to easily learn to attend by relative positions.

\subsubsection*{Rotary positional encoding (RoPE)}

Another popular approach for taking into account absolute positional information
in self-attention is RoPE \citep{su2024roformer},
which stands for rotary positional encoding. For simplicity, we briefly explain
the idea with a single attention head and $D_v = D_k = D$. Suppose we have a
sequence $\x_1, \dots, \x_L \in \RR^D$ already encoded with token encoding. 
We then define the query and key vectors as
\begin{align*}
    \q_m &\coloneqq \bm{R}_D(\theta, m) \W^Q \x_m \quad m \in [L] \\
    \bm{k}_n &\coloneqq \bm{R}_D(\theta, n) \W^K \x_n \quad n \in [L],
\end{align*}
where $\bm{R}_D(\theta, m) \in \RR^{D \times D}$ is a rotation matrix.
The main intuition is that we are rotating the $D$-dimensional vector
by an angle which is a multiple of the position $m$.
In contrast to the learned and sinusoidal positional encodings,
the weight matrices $\W^Q$ and $\W^K$ are applied \textbf{before} applying RoPE,
not after.

When $D=2$, $\bm{R}_D(\theta,m) = \bm{R}_2(\theta,m)$ is defined as the rotation
matrix
\begin{equation*}
\bm{R}_2(\theta, m) \coloneqq
\begin{pmatrix}
    \cos(m\theta) & -\sin(m\theta) \\
    \sin(m\theta) & \cos(m \theta)
\end{pmatrix}
\end{equation*}
for some angle $\theta \in \RR$ and position $m \in [L]$.
Using Euler's formula
\begin{equation*}
e^{im\theta} = \cos(m \theta) + i \sin(m \theta),
\end{equation*}
we note that
$\v' \coloneqq \bm{R}_2(\theta, m) \v$ for $\v = (v_1, v_2) \in \RR^2$ and
$\v' = (v'_1, v'_2) \in \RR^2$ is 
equivalent to $z' \coloneqq z e^{i m \theta}$ 
for $z \coloneqq v_1 + i v_2 \in \CC$ and $z'= v'_1 + i v'_2 \in \CC$.
See \citet{su2024roformer} for a detailed derivation of RoPE's formula.

When $D>2$, assuming $D$ is even, $\bm{R}_D(\theta, m)$ is defined as
\begin{equation*}
\bm{R}_D(\theta, m)
\coloneqq
\begin{pmatrix}
\bm{R}_2(\theta_1, m) & & & \\
& \bm{R}_2(\theta_2, m) & & \\
         & & \ddots & \\
         & & & \bm{R}_2(\theta_{D/2}, m)
\end{pmatrix},
\end{equation*}
where $\theta_j \coloneqq 10000^{-2(j-1)/D}$.
In practice, we never materialize $\bm{R}_D(\theta, m)$ as a matrix, since it
would be very sparse, but rather view it as a
linear map applied to an arbitrary vector $\v \in \RR^D$,
\begin{equation*}
\bm{R}_D(\theta, m)\v =
\begin{pmatrix}
v_1 \\
v_2 \\
v_3 \\
v_4 \\
\vdots \\
v_{D-1} \\
v_D
\end{pmatrix}
\otimes
\begin{pmatrix}
\cos(m \theta_1) \\
\cos(m \theta_1) \\
\cos(m \theta_2) \\
\cos(m \theta_2) \\
\vdots \\
\cos(m \theta_{D/2}) \\
\cos(m \theta_{D/2})
\end{pmatrix}
+
\begin{pmatrix}
-v_2 \\
v_1 \\
-v_4 \\
v_3 \\
\vdots \\
-v_D \\
v_{D-1}
\end{pmatrix}
\otimes
\begin{pmatrix}
\sin(m \theta_1) \\
\sin(m \theta_1) \\
\sin(m \theta_2) \\
\sin(m \theta_2) \\
\vdots \\
\sin(m \theta_{D/2}) \\
\sin(m \theta_{D/2})
\end{pmatrix}.
\end{equation*}
Once we computed $\Q$ and $\K$, 
we can use $\mathrm{Attention}(\Q, \K, \V; \M)$ as usual.
For multi-head attention, we simply apply the same approach for each attention
head.

As analyzed by \citet{su2024roformer}, RoPE satisfies several valuable
properties. In particular, it is flexible \wrt the sequence length $L$ and 
despite being an absolute encoding, it manages to also capture relative distance
between tokens. It therefore has the merits of both absolute and relative
positional encodings. 

\subsection{Decoder-only architectures}
\label{neural_nets:sec:decoder_only_transformer} 

\subsubsection*{Defining a language model}

Transformers were originally developed to create encoder-decoder architectures
for machine translation \citep{vaswani2017attention}. However, in the context of
language modelling, Transformers are now routinely used for decoder-only
\textbf{autoregressive} architectures.
Suppose we are given a sequence of discrete symbols $x_1, \dots, x_L \in [M]$
(if a sequence has less than $L$ elements, we can use padding symbols).
The goal of (unconditional) language models is to create a function producing the joint
probability,
\begin{equation*}
(x_1, \dots, x_L) \mapsto \PP(X_1 = x_1, \dots, X_L = x_L).
\end{equation*}
Using the chain rule of probability (\cref{gm:sec:chain_rule_proba}),
without loss of generality, we can write
\begin{equation*}
\PP\left(X_1 = x_1, \ldots, X_L = x_L\right) 
= \prod_{k=1}^L \mathbb P(X_k = x_k \mid X_1 = x_1, \dots, X_{k-1}=x_{k-1}).
\end{equation*}
If the maximum length $L$ is very large, it is more practical to only consider a
context window of size $N$,
\begin{equation*}
\PP\left(X_1 = x_1, \ldots, X_L = x_L\right) 
\coloneqq
\prod_{k=1}^L
\PP(X_k = x_k \mid X_{k-N}=x_{k-N}, \dots, X_{k-1} = x_{k-1}).
\end{equation*}
This amounts to defining a \textbf{higher-order Markov chain}
(\cref{gm:sec:higher_order_Markov_chain}).
Using this approach, creating a language model then 
boils down to defining a function
\begin{equation*}
(x_{k-N}, \dots, x_{k-1}, x_k) 
\mapsto 
\PP(X_k = x_k \mid X_{k-N}=x_{k-N}, \dots, X_{k-1} = x_{k-1}).
\end{equation*}
Let us assume that the sequence of discrete symbols 
$x_1, \dots, x_L \in [M]$
has been mapped to a sequence of continuous vectors
$\x_1, \dots, \x_L \in \RR^D$
using token encoding
(\cref{neural_nets:sec:input_encoding})
and positional encoding
(\cref{neural_net:sec:positional_encoding}).
Using a causal Transformer block, we can obtain a representation of the current
context,
\begin{equation*}
(\x_{k-N}, \dots, \x_{k-1}) 
\leftarrow
\mathrm{Transformer}(\x_{k-N}, \dots, \x_{k-1}; \M_N).
\end{equation*}
To ensure that the Transformer only relies on past tokens to predict the current
token, the mask matrix $\M_N$ is set to a matrix of size $N \times N$, with the
elements of the lower part set to $1$, and the elements of the upper part set to
$-\infty$.  To reduce this to logits in $\RR^M$, we usually use the last element
of the context $\x_{k-1} \in \RR^D$ and apply a linear model, to obtain
\begin{equation*}
\thetav_k \coloneqq \W^D \x_{k-1} \in \RR^M, 
\end{equation*}
where $\W^D \in \RR^{M \times D}$ is a learned ``disembedding'' matrix.
To obtain a valid probability distribution, we apply a soft-argmax
on the logits $\thetav_k$,
\begin{equation*}
\piv_k \coloneqq \mathrm{softargmax}(\thetav_k) \in \triangle^M.
\end{equation*}
Finally, we can now define
\begin{equation*}
\PP(X_k = x_k \mid X_{k-N}=x_{k-N}, \dots, X_{k-1} = x_{k-1})
\coloneqq
[\piv_k]_{x_k}.
\end{equation*}

\begin{boxrem}{Weight tying}
Instead of learning a separate disembedding matrix $\W^D \in \RR^{M \times D}$, 
a frequently used technique is to set $\W^D \coloneqq (\W^E)^\top$, where
$\W^E \in \RR^{D \times M}$ is the embedding matrix from
\cref{neural_nets:sec:input_encoding}. This weight tying reduces the number of
parameters to learn and is shown to work well in practice
\citep{press2016using}.
\end{boxrem}

\subsubsection*{Training}

Let us gather the Transformer parameters (multi-head attention, MLP, layer norm)
as $\w \in \RR^P$. The decoder-only Transformer with a context window of size
$N$ then defines a parametric probability distribution
\begin{equation*}
p_\w(x_1, \ldots, x_L) 
\coloneqq
\prod_{k=1}^L
p_\w(x_k \mid x_{k-N}, \dots, x_{k-1}).
\end{equation*}
Given a corpus of sequences $\cD$, we usually seek the parameters $\w \in \RR^P$
by maximizing the log-likelihood,
\begin{equation*}
\EE_{x \sim \cD} \left[\log p_\w(x_1, \dots, x_L)\right]
=
\EE_{x \sim \cD} \left[\sum_{k=1}^L
\log p_\w(x_k \mid x_{k-N}, \dots, x_{k-1})
\right].
\end{equation*}
This amounts to using a logistic (cross-entropy) loss in a token-wise fashion.
Importantly, the context $x_{k-N}, \dots, x_{k-1}$ used to predict the next
token $x_k$ always comes from the data, not from the tokens generated by the
model. This is called \textbf{teacher forcing}
and makes Transformer training highly parallelizable.
The objective is usually solved
approximately using stochastic gradient algorithms. To maximize GPU utilization,
sequences of variable lengths are usually packed together in order to form
batches.

\subsubsection*{Sampling}

A decoder-only Transformer with a context window of size $N$ can be seen as
forming a higher-order Markov chain, which is a special case of Bayesian network.
To generate a sequence from the model, we can therefore use \textbf{ancestral
sampling} (\cref{gm:sec:ancestral_sampling}), 
\begin{align*}
    X_0 &\coloneqq x_0 \\
    X_1 &\sim p_\w(\cdot \mid X_0) \\
    X_2 &\sim p_\w(\cdot \mid X_0, X_1) \\
        &\vdots \\
    X_k &\sim p_\w(\cdot \mid X_{k-N}, \dots, X_{k-1}),
\end{align*}
where $x_0$ denotes the beginning-of-sequence (BOS) token.
The sampling stops when the end-of-sequence (EOS) token is generated or when a
maximum length is reached. The generated sequences are \iid.

Because sampling happens one token at a time, it is highly sequential.
The Transformer block is called sequentially as 
\begin{align*}
&\mathrm{Transformer}(X_0; \M_1) \\
&\mathrm{Transformer}(X_0, X_1; \M_2) \\
&\dots \\
&\mathrm{Transformer}(X_{k-1}, \dots X_{k-N}; \M_N).
\end{align*}
To avoid repeating the same computations again and again, 
assuming that a causal mask is used,
the past key and value
matrices used in multi-head attention
(\cref{neural_net:sec:multi_head_attention}) are usually stored in the so-called
\textbf{KV cache}.

\subsection{Encoder-only architectures}

Encoder-only Transformers can be used for learning to represent sequences. 
The most prominent example is BERT 
\citep{devlin2019bert}, which stands for
bidirectional encoder representations from transformers.
BERT uses \cref{net:algo:transformer} without causal mask (hence
``bidirectional'' in the acronym).
Discrete sequence elements are transformed to vectors using token encoding,
positional encoding and potentially segment encoding.
For reasons that will become clear below,
the first token of every sequence is always a special classification token
\texttt{[CLS]}. 
Training is broken down into two phases: pretraining and finetuning.

\subsubsection*{Pretraining}

Since pre-training is performed on an unlabeled corpus, it is necessary to
synthetically generate prediction tasks.

In masked prediction, a percentage (typically 15\%) of the input tokens are
randomly masked. The model is then trained to predict the original vocabulary ID
of the masked tokens, given the context provided by the unmasked tokens. This
forces the model to learn a rich, bidirectional representation of the language.
The masking strategy involves:
\begin{itemize}
\item 80\% of the time, replacing the chosen token with \texttt{[MASK]},
\item 10\% of the time, replacing the chosen token with a random token from the
    vocabulary,
\item 10\% of the time, keeping the chosen token unchanged.
\end{itemize}

In next sentence prediction, the model is trained to understand the relationship
between two sentences. For each training example, BERT is given two sentences,
$A$ and $B$. 50\% of the time, B is the actual next sentence that follows A in
the original document (labeled IsNext). The other 50\% of the time, $B$ is a
random sentence from the corpus (labeled NotNext).
A special \texttt{[CLS]} token is prepended to the input sequence, and its final
hidden state is used to predict whether sentence $B$ follows sentence A. The
\texttt{[SEP]} token separates the two sentences.

\subsubsection*{Finetuning}

After pre-training on a massive corpus, the pretrained BERT model can be
finetuned for various downstream NLP tasks (e.g., text classification, question
answering, named entity recognition) by adding a small, task-specific output
layer on top of the pre-trained Transformer encoder. The entire model, including
the pretrained weights, is then finetuned on the labeled data for the specific
task. For classification tasks, the final hidden state corresponding to the
class token \texttt{[CLS]} is used. For tagging tasks (token-level
classification) such as named entity recognition (NER) or part-of-speech (POS)
tagging, the final hidden state of each token is used. For question answering
tasks, which usually involve finding a span (start and end token) within a given
text, this is treated as two token-level classification problems: one for the
start token and one for the end token.

\subsection{Encoder-decoder architectures}

The encoder-decoder architecture was the original architecture proposed in the
seminal Transformer paper \citep{vaswani2017attention}. It can be used for
sequence-to-sequence tasks, such as machine translation, summarization and
question answering. We denote the input sequence by $\X$ and the output sequence
by $\Y$.

\subsubsection*{Encoder}

The role of the encoder is to produce a rich representation $\H_{\mathrm{enc}}$
of the input sequence $\X$, which serves as a context for the subsequent
decoding phase.
As in encoder-only architectures, this is achieved by using
\cref{net:algo:transformer} without causal mask.
The multi-head self-attention layer allows the encoder to weigh the importance
of different tokens in the input sequence relative to each other. For each token
in the input, it computes a representation that incorporates information from
all other tokens in the same sequence.
The encoder block uses its own parameters for each sub-component
(multi-head attention, layer norm, MLP). In particular,
multi-head attention uses parameters
$\{\W_i^Q\}_{i=1}^H$, $\{\W_i^K\}_{i=1}^H$, $\{\W_i^V\}_{i=1}^H$ and $\W^O$.

\subsubsection*{Decoder}

The role of the decoder is to generate the output sequence $\Y$ one token at a
time, based on the encoded representation $\H_{\mathrm{enc}}$ from the encoder
and the previously generated tokens $\Y_{1:t-1}$. The decoder differs in the way
attention is applied.
First, we apply causal multi-head self-attention to obtain a representation
$\H_{\mathrm{dec}}$ corresponding to the previously generated tokens
$\Y_{1:t-1}$.  Second, we apply multi-head \textbf{cross-attention}. The key
difference with multi-head self-attention is that the key, query and value
matrices are defined as
\begin{align*}
    \Q^{\mathrm{dec}}_i &\coloneqq \H_\mathrm{dec} {\W'}_i^Q \\
    \K^{\mathrm{enc}}_i &\coloneqq \H_\mathrm{enc} {\W'}_i^K \\
    \V^{\mathrm{enc}}_i &\coloneqq \H_\mathrm{enc} {\W'}_i^V,
\end{align*}
where $\{{\W'}_i^Q\}_{i=1}^H$, $\{{\W'}_i^K\}_{i=1}^H$ and
$\{{\W'}_i^V\}_{i=1}^H$ are the weight matrices used for the decoder.
Subsequently, MLP and layer norm layers are applied, similarly as before.

\subsubsection*{Differences with decoder-only architectures}

An encoder-decoder architecture provides a dedicated component for understanding
the input and another for generating the output. It allows for a bidirectional
understanding of the input.  It is therefore best suited for
sequence-to-sequence (seq2seq) tasks where the input and output sequences may
have different domains or modalities.

Decoder-only models, on the other hand, are streamlined for generating text by
continuously predicting the next token, leveraging a single, powerful
autoregressive mechanism.
That being said, due to their simplicity, decoder-only architectures are now
increasingly being used even for seq2seq tasks, by prepending the input (prompt)
to the context. This means that the input is subject to the causal mask, but
the performance remains remarkably strong.

\section{Summary}

\begin{itemize}

\item Programs can be mathematically represented as a directed acyclic graph. 

\item Neural networks are parameterized programs.

\item Feedforward networks are parameterized computation chains.

\item Multilayer perceptrons (MLPs), residual neural
networks (ResNets) and convolutional neural networks (CNNs) are all particular
parametrizations of feedforward networks. 

\item Transformer blocks are designed to process sequences of variable-length,
    but they are equivariant to permutations and therefore require positional
    encoding, in order to leverage positional information.

\end{itemize}

%% file: chapters/control_flows/control_flows.tex
\chapter{Control flows}
\label{chap:cf}

Control flows, such as conditionals or loops,
are an essential part of computer programming, 
as they allow us to express complex programs.
It is therefore natural to ask whether these constructs can be included in a
differentiable program.
This is what we study in this chapter.

\section{Comparison operators}

\begin{figure}[t]
  \centering
  \includegraphics[width=0.8\linewidth]{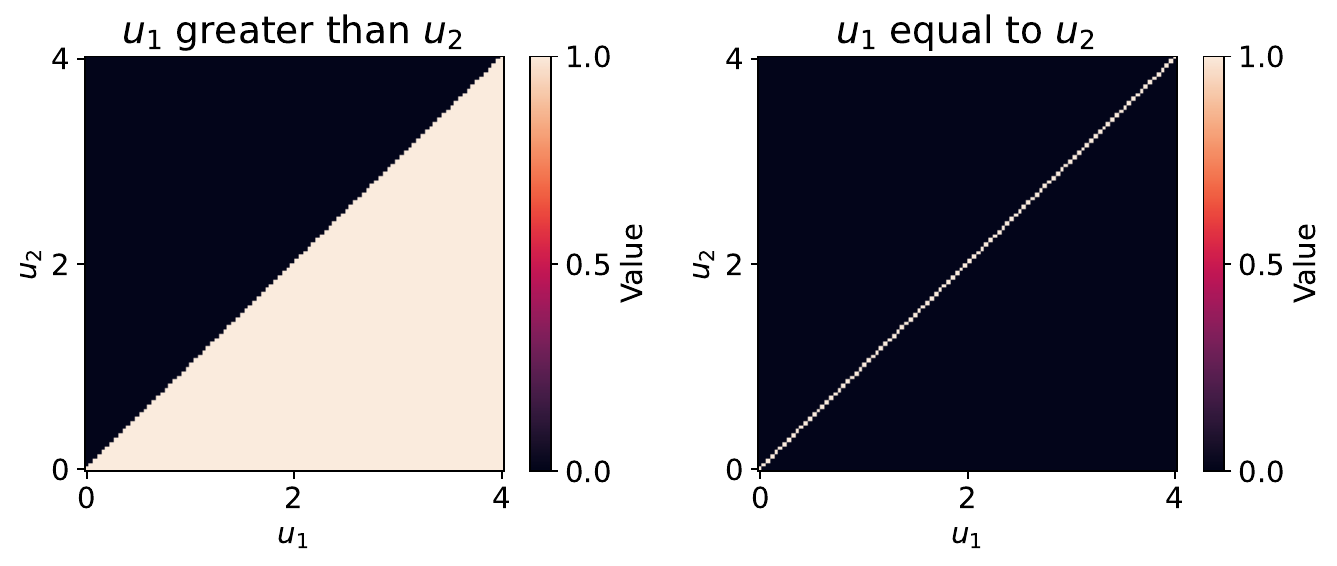}
  \includegraphics[width=0.8\linewidth]{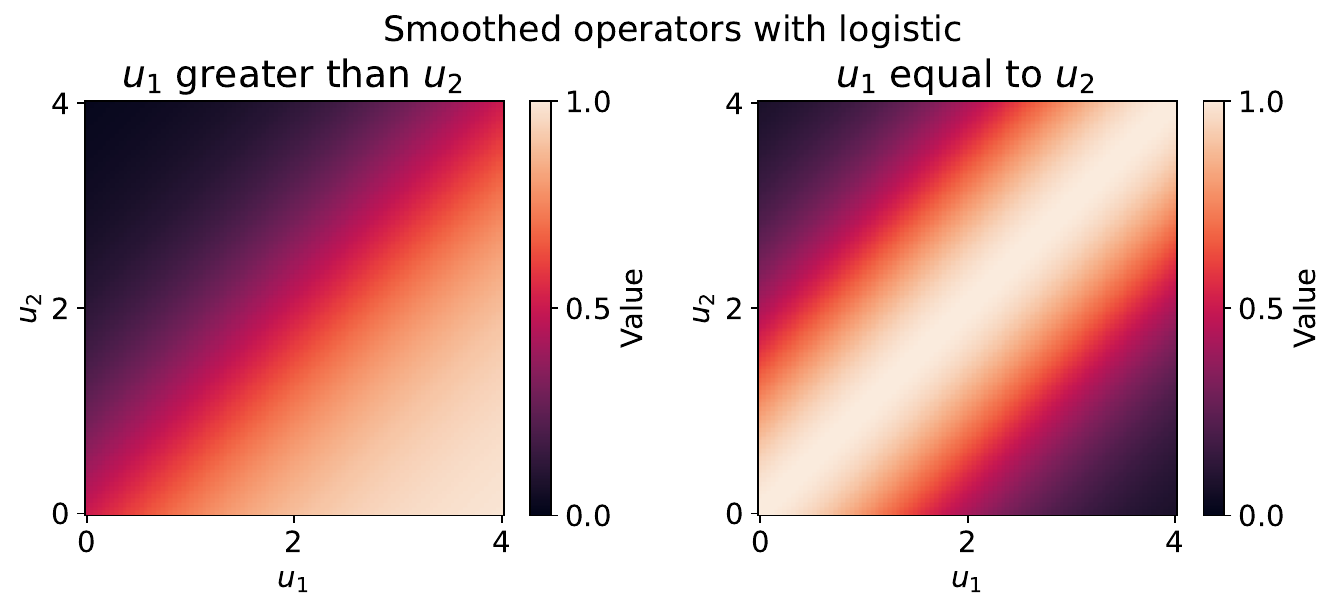}
  \caption{
    The greater than and equal to operators are discontinuous functions,
    leading to black or white pictures. They can be smoothed with 
    appropriate approximations of the Heaviside step function.
  \label{cf:fig:hard_compa_ops}
  }
\end{figure}

Control flows rely on \textbf{comparison operators}, 
\aka \textbf{relational operators}.
Formally, we can define a comparison operator
$\pi = \mathrm{op}(u_1, u_2)$
as a function from $u_1 \in \RR$ and $u_2 \in \RR$ to $\pi \in \{0,1\}$.
The binary (Boolean) output $\pi$ can then be used within a conditional
statement (see \cref{cf:sec:if_else}, \cref{cf:sec:else_if}) 
to decide whether to execute one branch or another.
We define the following operators, illustrated in \cref{cf:fig:hard_compa_ops}:
\begin{itemize}
    \item \textbf{greater than:}
        \begin{align*}
\mathrm{gt}(u_1, u_2)
&\coloneqq
\begin{cases}
1 &\mbox{ if } u_1 \ge u_2 \\
0 &\mbox{ otherwise }
\end{cases} \\
&= \mathrm{step}(u_1 - u_2)
        \end{align*}
    \item \textbf{less than:}
        \begin{align*}
\mathrm{lt}(u_1, u_2)
&\coloneqq
\begin{cases}
1 &\mbox{ if } u_1 \le u_2 \\
0 &\mbox{ otherwise }
\end{cases} \\
&= \mathrm{step}(u_2 - u_1)
        \end{align*}
    \item \textbf{equal:} 
        \begin{align*}
\mathrm{eq}(u_1, u_2)
&\coloneqq
\begin{cases}
1 &\mbox{ if } |u_1 - u_2| = 0 \\
0 &\mbox{ otherwise }
\end{cases} \\
&= \mathrm{gt}(u_2, u_1) \cdot \mathrm{gt}(u_1, u_2) \\
&= \mathrm{step}(u_2 - u_1) \cdot \mathrm{step}(u_1 - u_2) \\
\end{align*}
    \item \textbf{not equal:} 
\begin{align*}
\mathrm{neq}(u_1, u_2)
&\coloneqq
\begin{cases}
1 &\mbox{ if } |u_1 - u_2| > 0 \\
0 &\mbox{ otherwise }
\end{cases} \\
&= 1 - \mathrm{eq}(u_1, u_2) \\
&= 1 - \mathrm{step}(u_2 - u_1) \cdot \mathrm{step}(u_1 - u_2),
        \end{align*}
\end{itemize}
where
$\mathrm{step} \colon \RR \to \{0,1\}$ is the \textbf{Heaviside step function}
\begin{equation*}
\mathrm{step}(u)
\coloneqq \begin{cases}
    1 &\mbox{ if } u \ge 0 \\
    0 &\mbox{ otherwise}
\end{cases}.
\end{equation*}
The Heaviside step function is piecewise constant.
At $u=0$, the function is discontinuous. 
At $u \neq 0$, it is continuous and has \textbf{null derivative}.
Since the comparison operators we presented are all expressed in terms of
the $\mathrm{step}$ function, they are all continuous and differentiable almost
everywhere, with null derivative. Therefore, while their derivatives are
well-defined almost everywhere, they are \textbf{uninformative} 
and prevent gradient backpropagation.

\section{Soft inequality operators}
\label{cf:sec:soft_cmp}

\subsection{Heuristic definition}

To obtain a \textbf{continuous relaxation} of inequality operators,
we can heuristically replace the step function in the expression of
``greater than'' and ``less than'' by a
sigmoid function $\mathrm{sigmoid}_\sigma$, where $\sigma > 0$ is a scaling
parameter. Such a sigmoid function should satisfy the following properties:
\begin{itemize}
    \item $\mathrm{sigmoid}_\sigma(-u) = 1 - \mathrm{sigmoid}_\sigma(u)$,
    \item $\lim_{u \to \infty} \mathrm{sigmoid}_\sigma(u) = 1$,
    \item $\lim_{u \to -\infty} \mathrm{sigmoid}_\sigma(u) = 0$,
    \item $\mathrm{sigmoid}_\sigma(0) = \frac{1}{2}$.
\end{itemize}
Two examples of sigmoids satisfying the aforementioned properties are the
logistic function (the CDF of the standard logistic distribution)
\[
\mathrm{sigmoid}_\sigma(u) 
\coloneqq \logistic_\sigma(u) \coloneqq \frac{1}{1+ e^{-u/\sigma}} \in (0, 1)
\]
and the standard Gaussian's CDF, defined in \cref{proba_learn:eq:normal_cdf},
\begin{equation*}
\mathrm{sigmoid}_\sigma(u) \coloneqq \Phi(u/\sigma).
\end{equation*}
We may then define the soft ``greater than'' 
\begin{align*}
\mathrm{gt}(\mu_1, \mu_2) 
&= \mathrm{step}(\mu_1 - \mu_2) \\
&\approx \sigmoid_\sigma(\mu_1 - \mu_2) \\
&\eqqcolon \mathrm{gt}_\sigma(\mu_1, \mu_2)
\end{align*}
and the soft ``less than''
\begin{align*}
\mathrm{lt}(\mu_1, \mu_2) 
&= \mathrm{step}(\mu_2-\mu_1) \\
&\approx \sigmoid_\sigma(\mu_2 - \mu_1) \\
&\eqqcolon \mathrm{lt}_\sigma(\mu_1, \mu_2) \\
&= 1 - \sigmoid_\sigma(\mu_1 - \mu_2) \\
&= 1 - \mathrm{gt}_\sigma(\mu_1, \mu_2).
\end{align*}
In the limit, we have that $\sigmoid_\sigma(\mu_1 - \mu_2) \to 1$ when $\mu_1 -
\mu_2 \to \infty$.  In the limit, $\sigmoid_\sigma$ therefore outputs a value
of $1$ if $\mu_1$ and $\mu_2$ are infinitely apart.
Besides the logistic function and the standard Gaussian's CDF, other sigmoid
functions are possible, as discussed in \cref{inf_conv:sec:sigmoid}.  In
particular, with sparse sigmoids, there exists a finite value $\tau$ such that
$\mu_1 - \mu_2 \ge \tau \Longrightarrow \sigmoid_\sigma(\mu_1 - \mu_2) = 1$.

\subsection{Stochastic process perspective}

When the sigmoid used to replace the step function is the logistic function or
the standard Gaussian's CDF,
we can revisit the previous heuristic definition of
$\mathrm{gt}_\sigma(\mu_1,\mu_2)$ 
and
$\mathrm{lt}_\sigma(\mu_1,\mu_2)$ 
from a more formal perspective.
Indeed, to real values $\mu_1 \in \RR$ and $\mu_2 \in \RR$, 
we can associate random variables
\begin{equation*}
U_1 \sim p_{\mu_1,\sigma_1}
\quad \text{and} \quad
U_2 \sim p_{\mu_2,\sigma_2},
\end{equation*}
thereby forming a \textbf{stochastic process} (we assume that $\sigma_1$ and
$\sigma_2$ are fixed). 
Alternatively, we can also write
\begin{equation*}
(U_1, U_2)\sim p_{\mu_1,\sigma_1}\otimes p_{\mu_2,\sigma_2},
\end{equation*}
where for two distributions $p_1$ and $p_2$, we denote by $p_1 \otimes p_2$
their outer product $(p_1 \otimes p_2)(u_1, u_2) \coloneqq p_1(u_1)p_2(u_2)$.
We can then define
\begin{align*}
\mathrm{gt}_\sigma(\mu_1, \mu_2) 
&= \EE\left[\mathrm{gt}(U_1, U_2)\right] \\
&= \EE\left[\mathrm{step}(U_1 - U_2)\right] \\
&= \PP(U_1 - U_2 \ge 0) \\
&= \PP(U_2 - U_1 \le 0) \\
&= F_{U_2 - U_1}(0),
\end{align*}
where 
$F_X$ is the \textbf{cumulative distribution function} (CDF)
of the random variable $X$,
and $\sigma$ is a function of $\sigma_1$ and $\sigma_2$.
Similarly, we obtain
\begin{align*}
\mathrm{lt}_\sigma(\mu_1, \mu_2) 
&= \EE\left[\mathrm{lt}(U_1, U_2)\right] \\
&= \EE\left[\mathrm{step}(U_2 - U_1)\right] \\
&= \PP(U_1 - U_2 \le 0) \\
&= F_{U_1 - U_2}(0).
\end{align*}
We see that the soft inequality operators are based
on the CDF of the \textbf{difference} between $U_1$ and $U_2$.

For location-scale family distributions
(\cref{grad_est:sec:location_scale_transform}),
from a perturbation perspective, we can also define noise variables 
$Z_1 \sim p_{0,1}$ and $Z_2 \sim p_{0,1}$ such that
$U_1 = \mu_1 + \sigma_1 Z_1$
and
$U_2 = \mu_2 + \sigma_2 Z_2$ (\cref{grad_est:sec:location_scale_transform}).
We then have 
\begin{align*}
\mathrm{gt}_\sigma(\mu_1, \mu_2) 
&= \EE \left[\mathrm{gt}(\mu_1 + \sigma_1 Z_1, \mu_2 + \sigma_2 Z_2)\right] \\
\mathrm{lt}_\sigma(\mu_1, \mu_2) 
&= \EE \left[\mathrm{lt}(\mu_1 + \sigma_1 Z_1, \mu_2 + \sigma_2 Z_2)\right].
\end{align*}

\subsubsection*{Gaussian case}

When 
$U_1 \sim \mathrm{Normal}(\mu_1, \sigma_1^2)$
and
$U_2 \sim \mathrm{Normal}(\mu_2, \sigma_2^2)$,
we have
\begin{equation}
\label{cf:eq:diff_normal}
U_1 - U_2 \sim \mathrm{Normal}(\mu_1 - \mu_2, \sigma_1^2 + \sigma_2^2).
\end{equation}
Denoting $\Phi$ the standard Gaussian's CDF, we then obtain
\begin{align*}
\mathrm{gt}_\sigma(\mu_1, \mu_2) 
&= F_{U_2 - U_1}(0) \\
&= \Phi\left(\frac{0 - (\mu_2 - \mu_1)}{\sigma}\right) \\
&= \Phi\left(\frac{\mu_1-\mu_2}{\sigma}\right) \\
\mathrm{lt}_\sigma(\mu_1, \mu_2) 
&= F_{U_1 - U_2}(0) \\
&= \Phi\left(\frac{0 - (\mu_1 - \mu_2)}{\sigma}\right) \\
&= \Phi\left(\frac{\mu_2-\mu_1}{\sigma}\right),
\end{align*}
where $\sigma \coloneqq \sqrt{\sigma_1^2 + \sigma_2^2}$.
The corresponding distribution for $Z_1$ and $Z_2$ is \textbf{Gaussian noise}.

\subsubsection*{Logistic case}
When 
$U_1 \sim \mathrm{Gumbel}(\mu_1, \sigma)$
and
$U_2 \sim \mathrm{Gumbel}(\mu_2, \sigma)$,
we have
\begin{equation}
\label{cf:eq:diff_logistic}
U_1 - U_2 \sim \mathrm{Logistic}(\mu_1 - \mu_2, \sigma).
\end{equation}
We then obtain (see also~\cref{perturb:prop:gumbel_binary}) 
\begin{align*}
\mathrm{gt}_\sigma(\mu_1, \mu_2) 
&= \mathrm{logistic}\left(\frac{\mu_1-\mu_2}{\sigma}\right) \\
\mathrm{lt}_\sigma(\mu_1, \mu_2) 
&= \mathrm{logistic}\left(\frac{\mu_2-\mu_1}{\sigma}\right).
\end{align*}
The corresponding distribution for $Z_1$ and $Z_2$ is \textbf{Gumbel noise}.

\subsubsection*{Recovering hard inequality operators}

We easily recover the ``hard'' inequality operator by
\[
\mathrm{gt}(\mu_1, \mu_2) = \EE\left[\mathrm{gt}(U_1, U_2)\right],
\]
where 
$U_i \sim \delta_{\mu_i}$
and
where $\delta_{\mu_i}$ is the \textbf{delta distribution} that assigns a
probability of $1$ to $\mu_i$.

\section{Soft equality operators}

\subsection{Heuristic definition}

The equality operator $\mathrm{eq}(\mu_1,\mu_2)$ can be seen as an extreme kind
of \textbf{similarity function} between numbers, that can only output the values
$0$ or $1$. 
To define soft equality operators, a natural idea is therefore to replace the
equality operator by a more general similarity function. A similarity function
should achieve its maximum at $\mu_1 = \mu_2$ and it should decrease 
as $\mu_1$ and $\mu_2$ move apart. A common family of
similarity functions are \textbf{kernels}.
Briefly, a kernel $k(\mu_1, \mu_2)$ can be seen as the inner product 
\[
k(\mu_1, \mu_2) \coloneqq \langle \phi(\mu_1), \phi(\mu_2) \rangle
\]
between the embeddings $\phi(\mu_1)$ and $\phi(\mu_2)$ of $\mu_1$ and $\mu_2$ in
some (potentially infinite-dimensional) space $\cH$,
a reproducing kernel Hilbert space to be precise; see
\citet{scholkopf2002learning,shawe2004kernel} for an in-depth review of kernels. 
To obtain a similarity measure between $0$ and $1$ approximating the equality
operator, we can normalize to obtain
\begin{align*}
  \mathrm{eq}(\mu_1, \mu_2) 
  & \approx \frac{k(\mu_1,\mu_2)}
  {\sqrt{k(\mu_1,\mu_1)k(\mu_2,\mu_2)}} \nonumber\\
  & = \frac{\langle \phi(\mu_1), \phi(\mu_2) \rangle}
  {\|\phi(\mu_1)\|\|\phi(\mu_2)\|},
\end{align*}
where $\|\phi(\mu)\| \coloneqq \sqrt{\langle \phi(\mu), \phi(\mu)\rangle} =
\sqrt{\kappa(\mu, \mu)}$.
This is the \textbf{cosine similarity} between $\phi(\mu_1)$ and
$\phi(\mu_2)$.

A particular kind of kernel is \textbf{isotropic} kernels of the form
\[
k(\mu_1, \mu_2) \coloneqq \kappa(\mu_1 - \mu_2),
\]
that depend only on the difference between inputs.
When the kernel has a scale parameter $\sigma > 0$, 
we use the notation $\kappa_\sigma$.  
We can then define a soft equality operator as
\begin{align*}
\mathrm{eq}(\mu_1, \mu_2) 
\approx 
\mathrm{eq}_\sigma(\mu_1, \mu_2) 
\coloneqq
\frac{\kappa_\sigma(\mu_1 - \mu_2)}{\kappa_\sigma(0)}.
\end{align*}
Several isotropic kernels can be chosen such as the \textbf{Gaussian kernel}
\begin{equation*}
\kappa_\sigma(t) \coloneqq \exp\left(-\frac{t^2}{2\sigma^2}\right)
\end{equation*}
or the \textbf{logistic kernel}
\begin{equation*}
\kappa_\sigma(t) 
\coloneqq \mathrm{sech}^2\left(\frac{t}{2\sigma}\right),
\end{equation*}
where we defined the hyperbolic secant
\begin{equation*}
\mathrm{sech}(u) \coloneqq 2 / (\exp(u) + \exp(-u)). 
\end{equation*}
As their names suggest, these kernels arise naturally from a
probabilistic perspective, that we present below. 

\begin{figure*}
\centering
\includegraphics[scale=0.35]{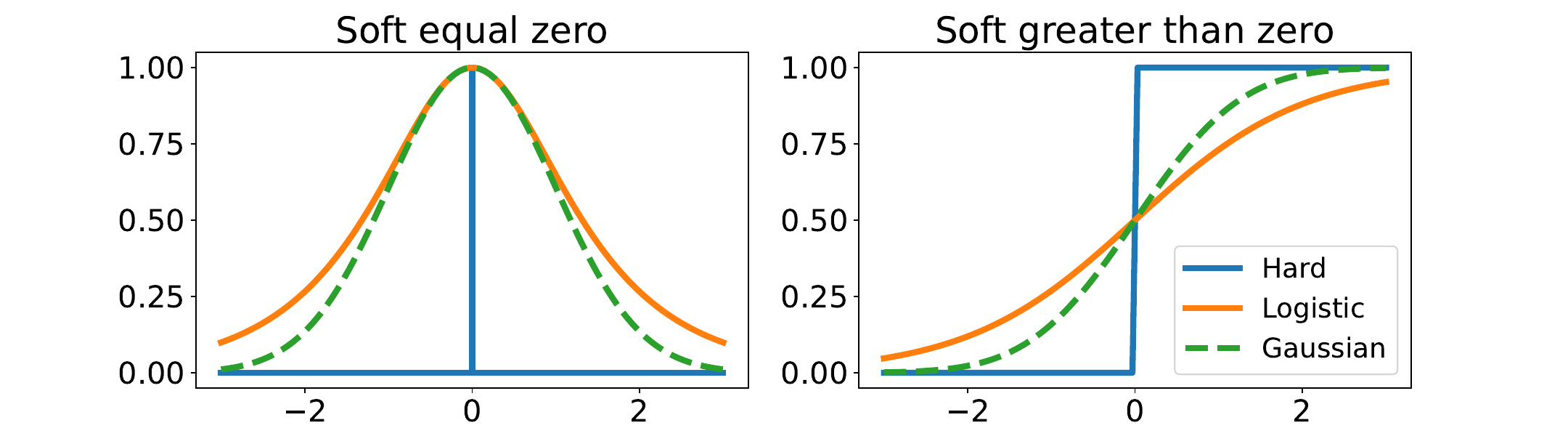}
\caption{Soft equality and soft greater than operators 
can be defined as normalized kernels (PDF) and as CDF functions, respectively.}
\label{cf:fig:soft_cmp}
\end{figure*}

The soft equality operators
obtained with these kernels are illustrated in \cref{cf:fig:soft_cmp}. 
Intuitively, we replaced a bar located at $\mu_1=\mu_2$ with a 
bump function.
The soft
equality operator obtained with the logistic kernel coincides with the
expression \citet{petersen_2021} arrive at (see their Eq.\ 9), in a different
manner.

\subsection{Stochastic process perspective}

We again adopt the stochastic process perspective, 
in which we associate random variables
\begin{equation*}
U_1 \sim p_{\mu_1,\sigma}
\quad \text{and} \quad
U_2 \sim p_{\mu_2,\sigma}
\end{equation*}
to real values $\mu_1 \in \RR$ and $\mu_2 \in \RR$.
However, to handle the equality operator,
we cannot simply use the expectation of
$\mathrm{eq}(U_1, U_2)$ since
\begin{equation*}
\EE[\mathrm{eq}(U_1, U_2)] 
= \PP(U_1=U_2)= 0,
\end{equation*}
$U_1$ and $U_2$ being independent continuous variables.
While we cannot use the probability of $U_1=U_2$, or equivalently of
$U_1-U_2=0$, we can consider using the probability density function (PDF)
$f_{U_1-U_2}$ of $U_1-U_2$ evaluated at $0$.
To ensure that the maximum is achieved at $0$ with value $1$,
we can normalize the PDF to define
\begin{equation*}
\mathrm{eq}_\sigma(\mu_1,\mu_2) 
= \frac{f_{U_1 - U_2}(0)}{f_{U_0}(0)},
\end{equation*}
where $U_0 \sim p_{0, \sigma}$ (we assume that $f_{U_0}$ attains its
maximum at $0$).
It is well-known that the PDF of the sum of two random variables is
the \textbf{convolution} of their respective PDFs. We therefore have
\begin{align*}
f_{U_1-U_2}(t)
&= (f_{U_1} \ast f_{-U_2})(t) \\
&= \int_{-\infty}^\infty f_{U_1}(\tau) f_{-U_2}(t - \tau) d\tau.
\end{align*}
In particular, with $t=0$, if $f_X$ is the PDF of a location-scale family
distributed random variable, we obtain
\begin{align*}
f_{U_1-U_2}(0)
&= (f_{U_1} \ast f_{-U_2})(0) \\
&= \int_{-\infty}^\infty f_{U_1}(\tau) f_{-U_2}(-\tau) d\tau \\
&= \int_{-\infty}^\infty f_{U_1}(\tau) f_{U_2}(\tau) d\tau \\
&\coloneqq \langle f_{U_1}, f_{U_2} \rangle \\
&\coloneqq \kappa(\mu_1,\mu_2).
\end{align*}
We indeed recover an inner product and therefore a kernel.

\subsubsection*{CDF and PDF of absolute difference}

While $\PP(U_1=U_2) = 0$, we can also consider using
$\PP(|U_1-U_2| \le \varepsilon) = F_{|U_1-U_2|}(\varepsilon)$ as an alternative
notion of soft equality.
For any random variable $X$, we have
\begin{align*}
F_{|X|}(x)
&= \PP(|X| \le x) \\
&= \PP(-x \le X \le x) \\
&= \PP(X \le x) - \PP(X \le -x) \\
&= F_X(x) -  F_X(-x).
\end{align*}
Therefore,
\begin{equation*}
\PP(|U_1-U_2| \le \varepsilon)  
= F_{U_1-U_2}(\varepsilon) - F_{U_1-U_2}(-\varepsilon).
\end{equation*}
We can also derive the PDF of $|X|$ as
\begin{align*}
f_{|X|}(x) 
&= F'_X(x) - F'_{X}(-x) \\
&= f_X(x) + f_{X}(-x)
\end{align*}
and in particular
\begin{equation*}
f_{|X|}(0) = 2 f_X(0).
\end{equation*}
Therefore
\begin{equation*}
f_{|U_1-U_2|}(0) = 2 f_{U_1-U_2}(0),
\end{equation*}
further justifying using the PDF of $U_1 - U_2$ evaluated at $0$.
When $X$ follows a normal distribution,
$|X|$ follows the so-called folded normal distribution.

\subsubsection*{Gaussian case}

When 
$U_1 \sim \mathrm{Normal}(\mu_1, \sigma_1^2)$
and
$U_2 \sim \mathrm{Normal}(\mu_2, \sigma_2^2)$,
we obtain from \cref{cf:eq:diff_normal}
\begin{equation*}
f_{U_1 - U_2}(t) 
= \frac{1}{\sqrt{2\pi}} \exp\left(-\frac{(t - (\mu_1 -\mu_2))^2}{2(\sigma_1^2 +
\sigma_2^2)}\right)
\end{equation*}
so that
\begin{equation*}
\mathrm{eq}_\sigma(\mu_1,\mu_2)
= 
\exp\left(-\frac{(\mu_1 -\mu_2)^2}{2(\sigma_1^2 + \sigma_2^2)}\right)
\in [0,1].
\end{equation*}
We indeed recover $\kappa_\sigma(\mu_1 - \mu_2) / \kappa_\sigma(0)$, where
$\kappa_\sigma$ is the Gaussian kernel with $\sigma = \sqrt{\sigma_1^2 +
\sigma_2^2}$.
For the CDF of the absolute difference, we obtain
\begin{equation*}
\PP(|U_1-U_2| \le \varepsilon)
= 
\Phi\left(\frac{\varepsilon-(\mu_1-\mu_2)}{\sigma}\right)
-
\Phi\left(\frac{-\varepsilon-(\mu_1-\mu_2)}{\sigma}\right).
\end{equation*}

\subsubsection*{Logistic case}

When 
$U_1 \sim \mathrm{Gumbel}(\mu_1, \sigma)$
and
$U_2 \sim \mathrm{Gumbel}(\mu_2, \sigma)$,
recalling that 
\begin{equation*}
\mathrm{sech}(u) \coloneqq 2 / (\exp(u) + \exp(-u)),
\end{equation*}
we obtain from \cref{cf:eq:diff_logistic}
\begin{equation*}
f_{U_1 - U_2}(t) 
= \frac{1}{4\sigma} 
\mathrm{sech}^2\left(\frac{t - (\mu_1 - \mu_2)}{2 \sigma}\right)
\end{equation*}
so that
\begin{equation*}
\mathrm{eq}_\sigma(\mu_1,\mu_2)
= \mathrm{sech}^2\left(\frac{\mu_1 - \mu_2}{2 \sigma}\right)
\in [0,1].
\end{equation*}
We indeed recover $\kappa_\sigma(\mu_1 - \mu_2) / \kappa_\sigma(0)$, where
$\kappa_\sigma$ is the logistic kernel with $\sigma = \sigma_1 = \sigma_2$.

\subsection{Gaussian process perspective}

The previous approach relied on mapping $\mu_1$ and $\mu_2$ to two
\textbf{independent} random variables 
$U_1 \sim p_{\mu_1,\sigma_1}$ 
and
$U_2 \sim p_{\mu_2,\sigma_2}$ (we assume that $\sigma_1$ and $\sigma_2$ are
fixed).
Instead, we can consider mapping $\mu_1$ and $\mu_2$ to two \textbf{dependent}
random variables $U_1$ and $U_2$, 
whose covariance depends on the similarity between $\mu_1$ and $\mu_2$.
We can do so by using a \textbf{Gaussian process}~\citep{hida1993gaussian}. 

A Gaussian process on $\RR$ is a stochastic process
$\{U_\mu: \mu\in \RR\}$ indexed by $\mu \in \RR$ such that any subset
of $K$ random variables $(U_{\mu_1}, \dots, U_{\mu_K})$ associated with 
$(\mu_1, \ldots, \mu_K) \in \RR^K$ is a multivariate Gaussian random variable. 
The Gaussian
process is characterized by the mean function
$\mu \mapsto \EE[U_{\mu}]$,
and its covariance function
$(\mu_i,\mu_j) \mapsto \Cov(U_{\mu_i}, U_{\mu_j})$.
For the mean function, we may simply choose $\EE[U_{\mu}]=\mu$. 
For the covariance
function, we need to ensure that the
variance of any combination of random variables in the Gaussian process is
non-negative. This property is satisfied by kernel functions.
We can therefore define 
\[
  \Cov(U_{\mu_i}, U_{\mu_j}) \coloneqq k(\mu_i, \mu_j),
\]
for some kernel $k$.
Equipped with such a mapping from real numbers to random variables, we need a
measure of similarity between random variables. A natural choice is their
\textbf{correlation}
\[
  \mathrm{corr}(U_{\mu_i}, U_{\mu_j}) 
  \coloneqq \frac{\Cov(U_{\mu_i}, U_{\mu_j})}{\sqrt{\Var(U_{\mu_i})\Var(U_{\mu_j})}}
  \in [-1, 1].
\]
We therefore obtain
\begin{align*}
\mathrm{corr}(U_{\mu_i}, U_{\mu_j}) 
&= \frac{k(\mu_i,\mu_j)}{\sqrt{k(\mu_i,\mu_i)k(\mu_j,\mu_j)}} \\
&= \frac{\langle \phi(\mu_i),\phi(\mu_j)
\rangle}{\|\phi(\mu_i)\|\|\phi(\mu_j)\|},
\end{align*}
which coincides with the \textbf{cosine similarity} measure we saw before.
In the particular case $K=2$ and when
$k_\sigma(\mu_1, \mu_2) = \kappa(\mu_1 - \mu_2)$,
we then recover the previous heuristically defined soft equality operator
\[
  \mathrm{eq}_\sigma(\mu_1, \mu_2)
  = \mathrm{corr}(U_{\mu_1}, U_{\mu_2})
  = \frac{\kappa(\mu_1 -\mu_2)}{\kappa(0)}.
\]

\section{Logical operators}

Logical operators can be used to perform Boolean algebra.
Formally, we can define them as functions from
$\{0,1\} \times \{0,1\}$ to $\{0,1\}$.
The 
\textbf{and} (logical conjunction \aka logical product), 
\textbf{or} (logical disjunction \aka logical addition)
and \textbf{not} (logical negation \aka logical complement)
operators, for example, are defined by
\begin{align*}
\mathrm{and}(\pi, \pi')
&\coloneqq
\begin{cases}
1 &\mbox{ if } \pi = \pi' = 1 \\
0 &\mbox{ otherwise }
\end{cases} \\
\mathrm{or}(\pi, \pi')
&\coloneqq
\begin{cases}
    1 &\mbox{ if } 1 \in \{\pi, \pi'\} \\
0 &\mbox{ otherwise }
\end{cases} \\
\mathrm{not}(\pi)
&\coloneqq
\begin{cases}
0 &\mbox{ if } \pi = 1 \\
1 &\mbox{ if } \pi = 0
\end{cases}.
\end{align*}
Classical properties of these operators include
\begin{itemize}
    \item Commutativity:
        \begin{align*}
        \mathrm{and}(\pi, \pi') = \mathrm{and}(\pi', \pi) \\
        \mathrm{or}(\pi, \pi') = \mathrm{or}(\pi', \pi)
        \end{align*}
    \item Associativity:
        \begin{align*}
            \mathrm{and}(\pi, \mathrm{and}(\pi', \pi'')) &=
            \mathrm{and}(\mathrm{and}(\pi, \pi'), \pi'') \\
            \mathrm{or}(\pi, \mathrm{or}(\pi', \pi'')) &=
            \mathrm{or}(\mathrm{or}(\pi, \pi'), \pi'')
        \end{align*}

    \item Distributivity of \textbf{and} over \textbf{or}:
    \begin{equation*}
        \mathrm{and}(\pi, \mathrm{or}(\pi', \pi''))
        =
        \mathrm{or}(\mathrm{and}(\pi, \pi'), \mathrm{and}(\pi, \pi''))
    \end{equation*}

    \item Neutral element:
    \begin{align*}
        \mathrm{and}(\pi, 1) &= \pi \\
        \mathrm{or}(\pi, 0) &= \pi 
    \end{align*}

    \item De Morgan's laws:
    \begin{align*}
        \mathrm{not}(\mathrm{or}(\pi, \pi'))
        &=
        \mathrm{and}(\mathrm{not}(\pi), \mathrm{not}(\pi')) \\
        \mathrm{not}(\mathrm{and}(\pi, \pi'))
        &=
        \mathrm{or}(\mathrm{not}(\pi), \mathrm{not}(\pi')).
    \end{align*}
\end{itemize}

More generally, 
for a binary vector $\piv = (\pi_1, \dots, \pi_K) \in \{0,1\}^K$,
we can define $\textbf{all}$ (universal quantification, $\forall$) 
and $\textbf{any}$ (existential quantification, $\exists$) operators, 
which are functions from $\{0,1\}^K$ to $\{0,1\}$, as
\begin{equation*}
\mathrm{all}(\piv) \coloneqq
\begin{cases}
1 &\mbox{ if } \pi_1 = \dots = \pi_K = 1 \\
0 &\mbox{ otherwise }
\end{cases}
\end{equation*}
and
\begin{equation*}
\mathrm{any}(\piv) \coloneqq
\begin{cases}
1 &\mbox{ if } 1 \in \{\pi_1, \dots, \pi_K\} \\
0 &\mbox{ otherwise }
\end{cases}.
\end{equation*}

\section{Continuous extensions of logical operators}

\subsection{Probabilistic continuous extension}

\begin{figure}[t]
\centering
\includegraphics[width=0.8\linewidth]{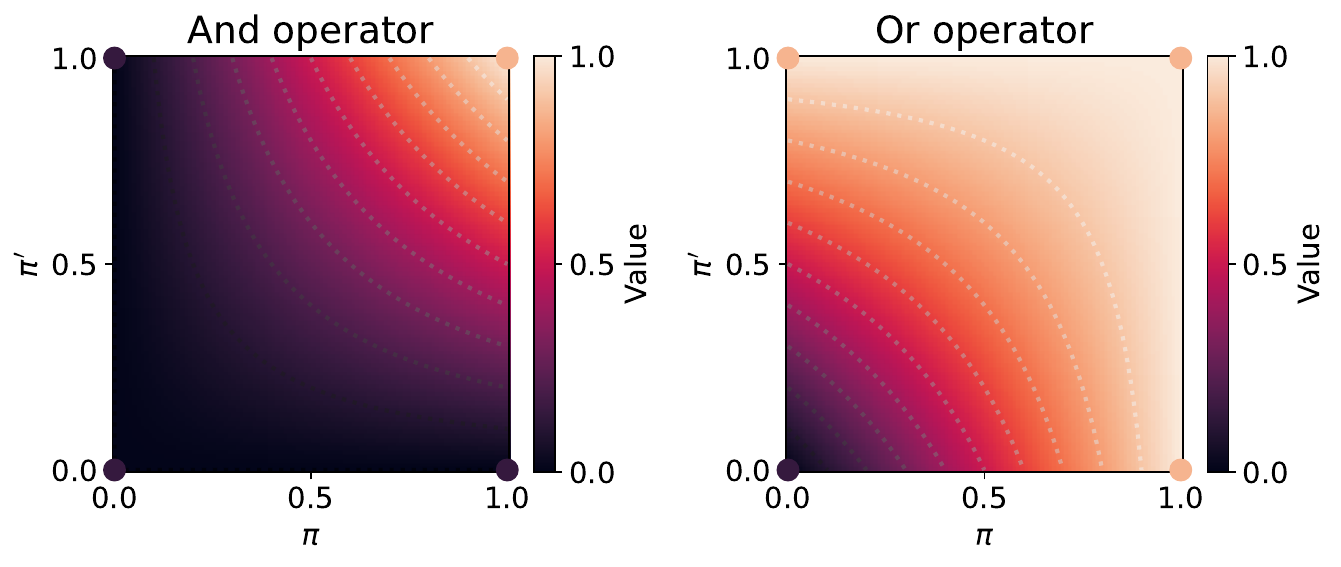}
\caption{The Boolean \textbf{and} and \textbf{or} operators are functions from
$\{0,1\} \times \{0,1\}$ to $\{0,1\}$ (corners in the figure)
but their continuous extensions
$\mathrm{and}(\pi, \pi') \coloneqq \pi \cdot \pi'$
as well as
$\mathrm{or}(\pi, \pi') \coloneqq \pi + \pi' - \pi \cdot \pi'$
define a function from $[0,1] \times [0,1]$ to $[0,1]$.}
\label{cf:fig:and_or}
\end{figure}

We can equivalently write the \textbf{and}, \textbf{or} and \textbf{not} 
operators as
\begin{align*}
\mathrm{and}(\pi, \pi') &= \pi \cdot \pi' \\
\mathrm{or}(\pi, \pi') &= \pi + \pi' - \pi \cdot \pi' \\
\mathrm{not}(\pi) &= 1 - \pi.
\end{align*}
These are \textbf{extensions} of the previous definitions:
we can use them as functions from $[0, 1] \times [0,1] \to [0, 1]$,
as illustrated in \cref{cf:fig:and_or}.
This means that we can use the soft comparison operators
defined in \cref{cf:sec:soft_cmp} to obtain $\pi, \pi' \in [0,1]$.
Likewise, we can define continuous extensions of \textbf{all} and \textbf{any},
which are functions from $[0,1]^K$ to $[0,1]$, as
\begin{align*}
    \mathrm{all}(\piv) &= \prod_{i=1}^K \pi_i \\
    \mathrm{any}(\piv) &= 1 - \prod_{i=1}^K (1 - \pi_i). 
\end{align*}

From a probabilistic perspective,
if we let $Y$ and $Y'$ be two independent random variables distributed 
according to \textbf{Bernoulli distributions} 
with parameters $\pi$ and $\pi'$, then
\begin{align*}
\mathrm{and}(\pi, \pi') &= \PP(Y = 1 \cap Y' = 1) 
                        = \PP(Y = 1) \cdot \PP(Y' = 1) \\
\mathrm{or}(\pi, \pi') &= \PP(Y = 1 \cup Y' = 1) \\
                       &= \PP(Y=1) + \PP(Y'=1) - \PP(Y = 1 \cap Y' = 1) \\
                       &= \PP(Y=1) + \PP(Y'=1) - \PP(Y = 1) \PP(Y' = 1) \\
\mathrm{not}(\pi) &= \PP(Y \neq 1) = 1 - \PP(Y = 1).
\end{align*}
In probability theory, these correspond to the product rule of two independent
variables, the addition rule, and the complement rule.

Likewise, if we let $Y = (Y_1, \dots, Y_K) \in \{0,1\}^K$ 
be a random variable distributed
according to a \textbf{multivariate Bernoulli distribution} with parameters
$\piv = (\pi_1, \dots, \pi_K)$,
then
\begin{align*}
\textrm{all}(\piv) 
&= \PP(Y_1 = 1 \cap \dots \cap Y_K = 1) \\
&= \prod_{i=1}^K \PP(Y_i = 1) \\
\textrm{any}(\piv) 
&= \PP(Y_1 = 1 \cup \dots \cup Y_K = 1) \\
&= 1 - \PP(\neg(Y_1 = 1 \cup \dots \cup Y_K = 1)) \\
&= 1 - \PP(Y_1 \neq 1 \cap \dots \cap Y_K \neq 1) \\
&= 1 - \prod_{i=1}^K (1 - \PP(Y_i=1)).
\end{align*}
These are the chain rule of probability and the addition rule of probability for
$K$ independent variables.

\subsection{Triangular norms and co-norms}
\label{cf:sec:t_norms}

More generally,
in the \textbf{fuzzy logic} literature \citep{klir_1995,jayaram_2008},
the concepts of triangular norms and co-norms have been introduced 
to provide continuous relaxations of the \textbf{and} and \textbf{or} operators,
respectively.
\begin{boxdef}{Triangular norms and conorms}
A triangular norm, \aka t-norm,
is a function from $[0,1] \times [0,1]$ to $[0,1]$
which is commutative, associative, neutral \wrt $1$
and is monotone, meaning that $t(\pi, \pi') \le t(\tau, \tau')$ 
for all $\pi \le \tau$ and $\pi' \le \tau'$.
A triangular conorm, \aka t-conorm, is defined similarly but is neutral \wrt
$0$.
\end{boxdef}
The previously-defined probabilistic extensions of $\textbf{and}$ and
\textbf{or} are examples of triangular norms and conorms.
More examples are given in \cref{cf:tab:t_norms}.
Thanks to the associative property of these operators,
we can generalize them to vectors $\piv \in [0,1]^K$
to define continuous extensions of the \textbf{all} and \textbf{any} operators,
as shown in \cref{cf:tab:aggreg}.
For more examples and analysis, see for instance 
\citet[Chapters 2 and 3]{van_krieken_thesis}.

\begin{figure}[t]
  \centering
  \includegraphics[width=0.49\linewidth]{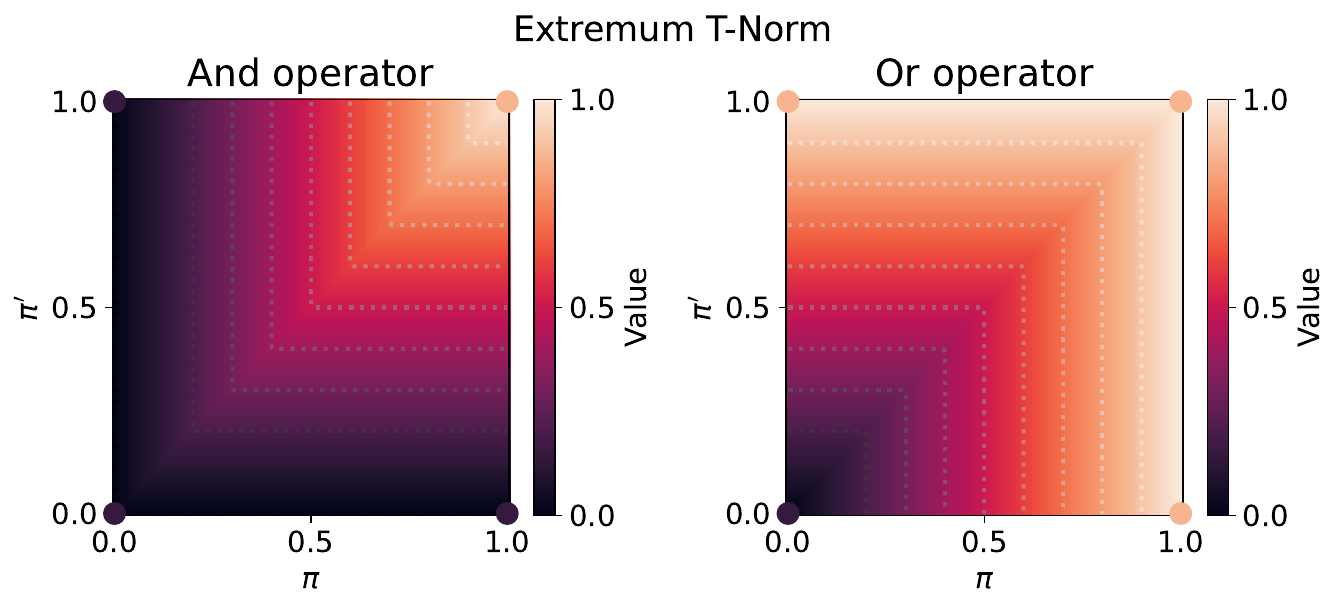}
  \includegraphics[width=0.49\linewidth]{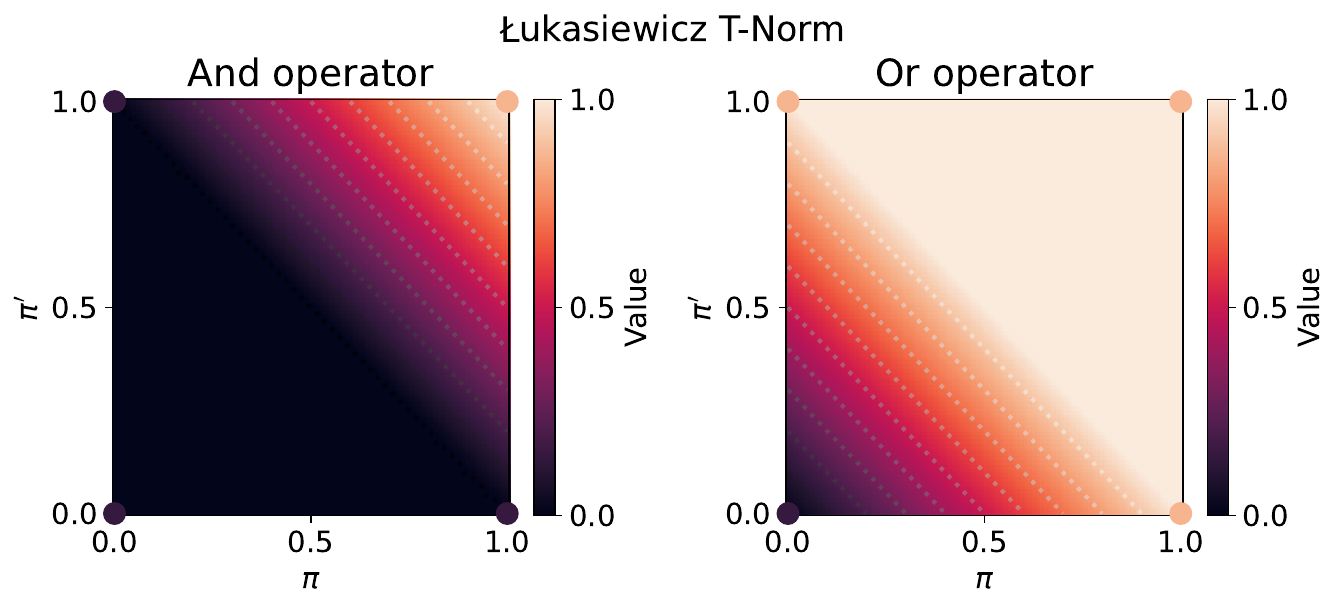}
  \caption{Alternative relaxations of the Boolean \textbf{and} and \textbf{or} operators using 
  triangular norms (t-norms).
  \label{cf:fig:and_or_t_norm}
  }
\end{figure}

\begin{table}[t]
\caption{Examples of triangular norms and conorms,
which are continuous relaxations of the \textbf{and} and \textbf{or} operators,
respectively. More instances can be obtained by smoothing out the min and max
operators.}
\centering
\begin{small}
\begin{tabular}{lcc}
\toprule
~ & t-norm (relaxed and) & t-conorm (relaxed or) \\
\midrule
Probabilistic & $\pi \cdot \pi'$ & $\pi + \pi' - \pi \cdot \pi'$ \\
Extremum & $\min(\pi, \pi')$ & $\max(\pi, \pi')$ \\
{\L}ukasiewicz & $\max(\pi + \pi' - 1, 0)$ & $\min(\pi + \pi', 1)$ \\
\bottomrule
\end{tabular}
\end{small}
\label{cf:tab:t_norms}
\end{table}

\begin{table}[t]
\caption{Continuous extensions of the \textbf{all} and \textbf{any} operators.}
\centering
\begin{small}
\begin{tabular}{lcc}
\toprule
~ & All ($\forall$) & Any ($\exists$) \\
\midrule
Probabilistic & $\prod_{i=1}^K \pi_i$ & $1 - \prod_{i=1}^K (1 - \pi_i)$ \\
Extremum & $\min(\pi_1, \dots, \pi_K)$ & $\max(\pi_1, \dots, \pi_K)$ \\
{\L}ukasiewicz & $\max(\sum_{i=1}^K \pi_i - (K-1), 0)$ & $\min(\sum_{i=1}^K
\pi_i, 1)$ \\
\bottomrule
\end{tabular}
\end{small}
\label{cf:tab:aggreg}
\end{table}

\section{If-else statements}
\label{cf:sec:if_else}

An if-else statement executes different code depending on a condition.
Formally, we can define the 
$\mathrm{ifelse} \colon \{0,1\} \times \cV \times \cV \to \cV$
function by
\begin{align}
\mathrm{ifelse}(\pi, \v_1, \v_0)
&\coloneqq 
\begin{cases}
\v_1 &\mbox{ if } \pi = 1 \\
\v_0 &\mbox{ if } \pi = 0 
\end{cases} \label{cf:eq:ifelse} \\
&= \pi \cdot \v_1 + (1 - \pi) \cdot \v_0. \nonumber
\end{align}
The $\pi$ variable is called the \textbf{predicate}.
It is a \textbf{binary} (Boolean) variable,
making the function $\mathrm{ifelse}$ undefined if $\pi \not \in \{0,1\}$.
The function is therefore discontinuous and nondifferentiable \wrt $\pi \in
\{0,1\}$.
On the other hand, $\v_0 \in \cV$ and $\v_1 \in \cV$,
which correspond to the false and true \textbf{branches},
can be \textbf{continuous} variables.
If $\pi=1$, the function is linear \wrt $\v_1$ and 
constant \wrt $\v_0$.
Conversely, if $\pi=0$, the function is linear \wrt $\v_0$
and constant \wrt $\v_1$.
We now discuss how to differentiate through $\mathrm{ifelse}$.

\subsection{Differentiating through branch variables}

For $\pi \in \{0,1\}$ fixed, 
$\mathrm{ifelse}(\pi, \v_1, \v_0)$
is a valid function \wrt $\v_1 \in \cV$ and $\v_0 \in \cV$, and can therefore be
used as a node in a computational graph (\cref{auto_diff:sec:graphs}).
Due to the linearity \wrt $\v_1$ and $\v_0$, we obtain
that the Jacobians \wrt $\v_1$ and $\v_0$ are
\begin{align*}
\partial_{\v_0} \mathrm{ifelse}(\pi, \v_1, \v_0)
&\coloneqq 
\begin{cases}
0 &\mbox{ if } \pi = 1 \\
I &\mbox{ if } \pi = 0 
\end{cases} \\
&= (1-\pi) \cdot I
\end{align*}
and
\begin{align*}
\partial_{\v_1} \mathrm{ifelse}(\pi, \v_1, \v_0)
&\coloneqq 
\begin{cases}
I &\mbox{ if } \pi = 1 \\
0 &\mbox{ if } \pi = 0 
\end{cases} \\
&= \pi \cdot I,
\end{align*}
where $I$ is the identity matrix of appropriate size.
Most of the time, if-else statements are composed with other functions.
Let
$g_1 \colon \cU_1 \to \cV$
and
$g_0 \colon \cU_0 \to \cV$
be differentiable functions.
We then define 
$\v_1 \coloneqq g_1(\u_1)$ 
and 
$\v_0 \coloneqq g_0(\u_0)$, 
where 
$\u_1 \in \cU_1$
and
$\u_0 \in \cU_0$.
The composition of $\mathrm{ifelse}$, $g_1$ and $g_0$ is then
the function
$f \colon \{0,1\} \times \cU_1 \times \cU_0 \to \cV$
defined by
\begin{align*}
f(\pi, \u_1, \u_0) 
&\coloneqq \mathrm{ifelse}(\pi, g_1(\u_1), g_0(\u_0)) \\
&= \pi \cdot g_1(\u_1) + (1-\pi) \cdot g_0(\u_0).
\end{align*}
We obtain that the Jacobians are
\begin{equation*}
\partial_{\u_1} f(\pi, \u_1, \u_0) 
= \pi \cdot \partial g_1(\u_1)
\end{equation*}
and
\begin{equation*}
\partial_{\u_0} f(\pi, \u_1, \u_0) 
= (1-\pi) \cdot \partial g_0(\u_0).
\end{equation*}
As long as $g_1$ and $g_0$ are differentiable functions, 
we can therefore differentiate through the branch variables 
$\u_1$ and $\u_0$ without any issue.
More problematic is the predicate variable $\pi$, as we now discuss.

\subsection{Differentiating through predicate variables}

The predicate variable $\pi$ is binary and therefore cannot be
differentiated directly.
However, $\pi$ can be the output of a comparison operator.
For example, suppose we want to express the function
$f_h \colon \RR \times \cU_1 \times \cU_0 \to \cV$
defined by
\begin{equation*}
f_h(p, \u_1, \u_0)
\coloneqq
\begin{cases}
g_1(\u_1) &\mbox{ if } p \ge 0 \\
g_0(\u_0) &\mbox{ otherwise }
\end{cases}.
\end{equation*}
Using our notation, this can be rewritten as
\begin{align*}
f_h(p, \u_1, \u_0) 
&\coloneqq \mathrm{ifelse}(\mathrm{gt}(p, 0), g_1(\u_1), g_0(\u_0)) \\
&= \mathrm{ifelse}(\heavistep(p), g_1(\u_1), g_0(\u_0)) \\
&= \heavistep(p) g_1(\u_1) + (1 - \heavistep(p)) g_0(\u_0).
\end{align*}
The Heaviside step function has a \textbf{discontinuity} at $p=0$,
but it is continuous and differentiable with derivative $\heavistep'(p) = 0$
for all $p \neq 0$. The function $f_h$ therefore has \textbf{null derivative}
\wrt $p \neq 0$,
\begin{align*}
\partial_p f_h(p, \u_1, \u_0) 
&= \partial_1 f_h(p, \u_1, \u_0) \\
&= \heavistep'(p)(g_1(\u_1) - g_0(\u_0)) \\
&= \zeros.
\end{align*}
In other words, while $f_h$ has \textbf{well-defined} derivatives \wrt $p$ for
$p \neq 0$, the derivatives are \textbf{uninformative}.
As another example, let us now consider the function
\begin{equation*}
g_h(\u_1, \u_0) \coloneqq f_h(t(\u_1), \u_1, \u_0),
\end{equation*}
for some differentiable function $t$. 
This time, $\u_1$ influences both the predicate and the true branch.
Then, using \cref{diff:prop:multiple_inputs}, we obtain
\begin{align*}
\partial_{\u_1} g_h(\u_1, \u_0) 
&= \partial t(\u_1) \partial_1 f_h(t(\u_1), \u_1, \u_0)
+ \partial_2 f_h(t(\u_1), \u_1, \u_0) \\
&= \partial_2 f_h(t(\u_1), \u_1, \u_0).
\end{align*}
In other words, the derivatives of the predicate $t(\u_1)$ do not influence
the derivatives of $g_h$.

\subsection{Continuous relaxations}

Fortunately, we recall that
\begin{equation*}
\mathrm{ifelse}(\pi, \v_1, \v_0)
= \pi \cdot \v_1 + (1 - \pi) \cdot \v_0.
\end{equation*}
This function is perfectly well-defined, even if $\pi \in [0,1]$, instead of
$\pi \in \{0,1\}$. That is, this definition is an \textbf{extension}
of \cref{cf:eq:ifelse} from the discrete set $\{0,1\}$ to 
the continuous unit segment $[0,1]$.
We saw that
\begin{equation*}
\mathrm{gt}(a, b) 
\approx \mathrm{gt}_\sigma(a, b) 
\coloneqq \sigmoid_\sigma(a - b) \in [0,1],
\end{equation*}
where we use $\sigmoid_\sigma$ to denote a differentiable S-shaped function
mapping $\RR$ to $[0,1]$.
For instance, we can use the logistic function or the standard Gaussian's CDF.
If we now define
\begin{align}
f_s(p, \u_1, \u_0) 
&\coloneqq \mathrm{ifelse}(\mathrm{gt}_\sigma(p, 0), g_1(\u_1), g_0(\u_0)) \nonumber \\
&= \mathrm{ifelse}(\sigmoid_\sigma(p), g_1(\u_1), g_0(\u_0)) \nonumber \\
&= \sigmoid_\sigma(p) g_1(\u_1) + (1 - \sigmoid_\sigma(p)) g_0(\u_0), 
\label{cf:eq:soft_ifelse}
\end{align}
the Jacobian becomes
\begin{equation*}
\partial_p f_s(p, \u_1, \u_0)    
= \sigmoid'_\sigma(p)(g_1(\u_1) - g_0(\u_0)).
\end{equation*}
If $\sigmoid_\sigma = \logistic(\cdot / \sigma)$ or $\sigmoid_\sigma =
\Phi(\cdot/\sigma)$,
the Jacobian is \textbf{non-null everywhere}, allowing gradients to
backpropagate through the computational graph. This is an example of smoothing
as studied in \cref{part:smoothing}.

\begin{figure}[t]
  \includegraphics[width=\linewidth]{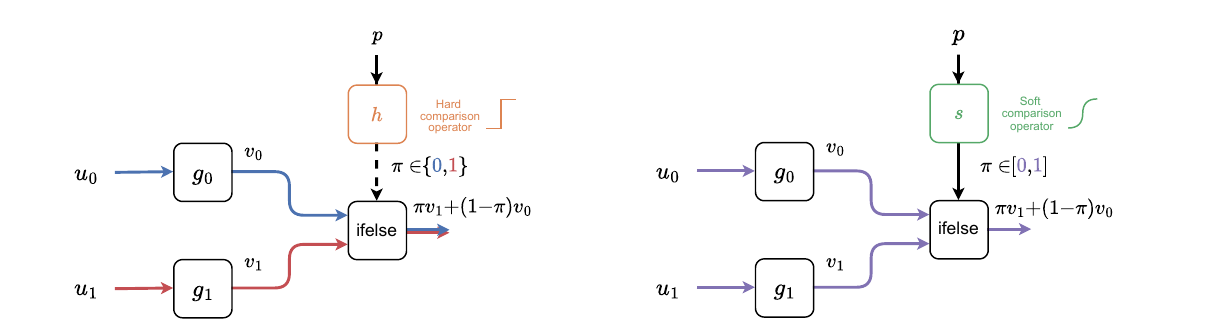}
  \caption{Computation graphs of programs using if-else statements with either
    hard or soft comparison operators. By using a hard comparison operator (step
    function, left panel) the predicate $\pi$ is a discrete variable
    (represented by a dashed line). Depending on the value ($0$ or $1$) of the
    predicate $\pi$, only one branch (red or blue) contributes to the output.
    Derivatives along a path of continuous variables (dense lines) can be
    computed. However, discrete variables such as the predicate prevent the
    propagation of meaningful derivatives. By using a soft comparison operator
    (sigmoid, right panel), the predicate is a continuous variable and
    derivatives with respect to the input $p$ can be taken. In this case both
    branches (corresponding to $g_0$ and $g_1$) contribute to the output
    and therefore need to be evaluated.
    \label{cf:fig:if_else_flows}
  }
\end{figure}

\subsubsection*{Probabilistic perspective}

\begin{figure}[t]
  \centering
  \includegraphics[scale=0.4]{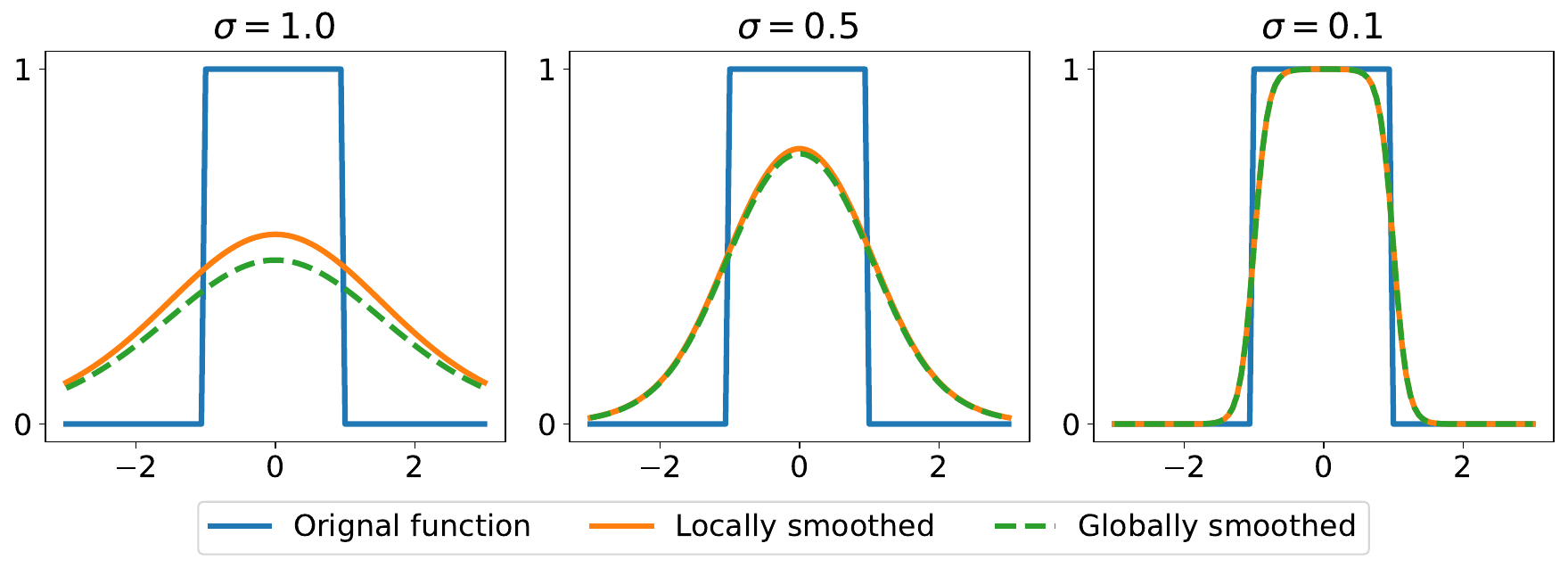}
  \caption{Global versus local smoothing approaches on a gate function $f(x)
      \coloneqq 1$ if $x \in [-1, 1]$, and $f(x) \coloneqq 0$ otherwise.
In our notation, we can write 
$f(x) = \mathrm{ifelse}(\mathrm{and}(\mathrm{gt}(x, -1), \mathrm{lt}(x, 1)), 1, 0)$.
  A local approach smoothes out $\mathrm{gt}$ and $\mathrm{lt}$ separately.
  A global approach uses the expectation of the whole program,
  see~\cref{cf:rem:global_vs_local}. 
  We observe that, though the approaches differ for large $\sigma$, they quickly
  coincide for smaller $\sigma$.}
  \label{cf:fig:global_vs_local_smoothing}
\end{figure}

From a probabilistic perspective,
we can view \cref{cf:eq:soft_ifelse} 
as the expectation of $g_i(\u_i)$, where $i \in \{0,1\}$ 
is a binary random variable
distributed according to a \textbf{Bernoulli distribution} with parameter 
$\pi = \sigmoid_\sigma(p)$:
\begin{equation*}
f_s(p, \u_1, \u_0)
= \EE_{i \sim \mathrm{Bernoulli}(\sigmoid_\sigma(p))}\left[g_i(\u_i)\right].
\end{equation*}
Taking the expectation over the two possibles branches makes the function
differentiable with respect to $p$, since $\sigmoid_\sigma(p)$ is differentiable.
Of course, this comes at the cost of evaluating both branches, instead of a
single one. The probabilistic perspective suggests that we can also compute
the variance if needed as
\begin{align*}
&\VV_{i \sim \mathrm{Bernoulli}(\sigmoid_\sigma(p))}\left[g_i(\u_i)\right] \\
=& 
\EE_{i \sim \mathrm{Bernoulli}(\sigmoid_\sigma(p))}
\left[(f_s(p, \u_1, \u_0) - g_i(\u_i))^2 \right].
\end{align*}
The probabilistic viewpoint also suggests different scales at which a smoothing
can be defined as illustrated in \cref{cf:fig:global_vs_local_smoothing}.

Another perspective \citep{petersen_2021} is based on the 
\textbf{logistic distribution}.
Indeed, if $P$ is a random variable following a logistic distribution
with mean $p$ and scale $1$,
we saw in \cref{proba_learn:rem:logistic} that the CDF is
$\PP(P \le 0) = \logistic(-p) = 1 - \logistic(p)$
and therefore
\begin{align*}
f_s(p, \u_1, \u_0)
&= \mathrm{ifelse}(\logistic(p), g_1(\u_1), g_0(\u_0)) \\
&= \logistic(p) g_1(\u_1) + (1 - \logistic(p)) g_0(\u_0) \\
&= \PP(P > 0) \cdot g_1(\u_1) + \PP(P \le 0) \cdot g_0(\u_0).
\end{align*}

\begin{boxrem}{Global versus local smoothing}\label{cf:rem:global_vs_local}
  Consider the function
  \begin{align*}
    f(x, y, z) \coloneqq
    \begin{cases}
      y & \mbox{if} \ a \leq x \leq b \\
      z & \mbox{otherwise}
    \end{cases}.
  \end{align*}
  The derivatives \wrt $y$ and $z$ are well-defined. 
  The derivative \wrt $x$ on
  the other hand is not well-defined since it involves comparison operators and
  the logical operator $\mathrm{and}$. Using our notation, we can 
  rewrite the function as 
  \begin{align*}
    f(x, y, z) & = 
    \mathrm{ifelse}(\mathrm{and}(\mathrm{gt}(x, a), \mathrm{lt}(x, b)), y, z).
  \end{align*}
  A local smoothing approach consists in replacing $\mathrm{gt}$ and
  $\mathrm{lt}$ by $\mathrm{gt}_\sigma$ and $\mathrm{lt}_\sigma$ locally in the
  program:
  \begin{align*}
  f^{\mathrm{loc}}_\sigma(x, y, z) &\coloneqq
    \mathrm{ifelse}(\mathrm{and}(\mathrm{gt}_\sigma(x, a), \mathrm{lt}_\sigma(x,
    b)), y, z) \\
                                   &= \pi_a \pi_b y + (1-\pi_a\pi_b)z
  \end{align*}
  where
  \begin{align*}
    \pi_a &\coloneqq \mathrm{sigmoid}_\sigma(x-a) \\
    \pi_b &\coloneqq \mathrm{sigmoid}_\sigma(b-x),
\end{align*}
for any sigmoid function $\sigmoid_\sigma$.
A global smoothing approach instead uses the expectation of the entire
program
  \begin{align*}
  f^{\mathrm{glob}}_\sigma(x, y, z) &\coloneqq
  \EE_Z[\mathrm{ifelse}(\mathrm{and}(\mathrm{gt}(x+\sigma Z, a), \mathrm{lt}(x +
  \sigma Z, b)), y, z)] \\
                                    &=\mathrm{ifelse}(\pi, y, z)
  \end{align*}
  where
  \begin{align*}
    \pi &\coloneqq \EE_Z[\mathrm{and}(\mathrm{gt}(x + \sigma Z, a), \mathrm{lt}(x +\sigma Z, b))] \\
    & = \PP(a\leq x+\sigma Z\leq b) \\
    & =  \sigmoid_\sigma(b - x) - \sigmoid_\sigma(a-x)\\
    & = \pi_b - \pi_a,
  \end{align*}
  for $\sigmoid_\sigma$ the CDF of $\sigma Z$.
  We therefore obtain
  \begin{equation*}
  f^{\mathrm{glob}}_\sigma(x, y, z) 
  =(\pi_b - \pi_a) y + (1 - (\pi_b - \pi_a)) z.
  \end{equation*}
  The difference 
  stems from the fact that the local approach smoothes out 
  $a \le x$ and $x \le b$ independently
  (treating $\ones_{X \geq a}$ and $\ones_{X\leq b}$ as independent random
  variables), while the global approach smoothes out
  $a \le x \le b$ simultaneously.
  In practice, both approaches approximate the original function well as $\sigma
  \rightarrow 0$ and coincide for $\sigma$ sufficiently small as illustrated in~
  \cref{cf:fig:global_vs_local_smoothing}.
\end{boxrem}

\section{Else-if statements}
\label{cf:sec:else_if}

In the previous section, we focused on if-else statements:
conditionals with only two branches. We now generalize our study
to conditionals including else-if statements, that have $K$ branches.

\subsection{Encoding $K$ branches}

For conditionals with only $2$ branches,
we encoded the branch that the conditional needs to take using the binary
variable $\pi \in \{0,1\}$.
For conditionals with $K$ branches,
we need a way to encode which of the $K$ branches the conditional needs to take.
To do so, we can use a vector $\piv \in \{\e_1, \dots, \e_K\}$,
where $\e_i$ denotes the standard basis vector (\aka one-hot vector)
\begin{equation*}
\e_i \coloneqq (0, \dots, \underbrace{1}_i, \dots, 0),
\end{equation*}
a vector with a single one in the coordinate $i$ and $K-1$ zeros.
The vector $\e_i$ is the encoding of a \textbf{categorical variable}
$i \in [K]$.

\subsubsection*{Combining booleans}

To form such a vector $\piv \in \{\e_1,\dots,\e_K\}$, 
we can combine the previously-defined 
comparison and logical operators to define $\piv = (\pi_1,\dots,\pi_K)$.
However, we need to ensure that only one $\pi_i$ is non-zero.
We give an example in \cref{cf:exm:st}.

\subsubsection*{Argmax and argmin operators}

Another way to form $\piv$ is to use
the \textbf{argmax} and \textbf{argmin} operators
\begin{align*}
\mathrm{argmax}(\p)
&\coloneqq
\argmax_{\piv \in \{\e_1, \dots, \e_K\}}
\langle \piv, \p \rangle \\
\mathrm{argmin}(\p)
&\coloneqq
\argmin_{\piv \in \{\e_1, \dots, \e_K\}}
\langle \piv, \p \rangle
= \mathrm{argmax}(-\p).
\end{align*}
They can be seen as a natural generalization of the greater than
and less than operators.
In case of ties, we break them arbitrarily.

\subsection{Conditionals}

We can now express a conditional statement as the function \\
$\mathrm{cond} \colon \{\e_1, \dots, \e_K\} 
\times \cV^K \to \cV$
defined by
\begin{align}
\mathrm{cond}(\piv, \v_1, \dots, \v_K)
&\coloneqq 
\begin{cases}
\v_1 &\mbox{ if } \piv = \e_1 \\
\vdots & \\
\v_K &\mbox{ if } \piv = \e_K
\end{cases} \label{cf:eq:cond} \\
&= \sum_{i=1}^K \pi_i \v_i \nonumber.
\end{align}
Similarly to the $\mathrm{ifelse}$ function,
the $\mathrm{cond}$ function is discontinuous and nondifferentiable
\wrt $\piv \in \{\e_1,\dots,\e_K\}$. 
However, given $\piv = \e_i$ fixed for some $i$,
the function is linear in $\v_i$ and constant in $\v_j$ for $j \neq i$.
We illustrate how to express a simple example, using this formalism.

\begin{figure}[t]
\centering
\includegraphics[scale=0.4]{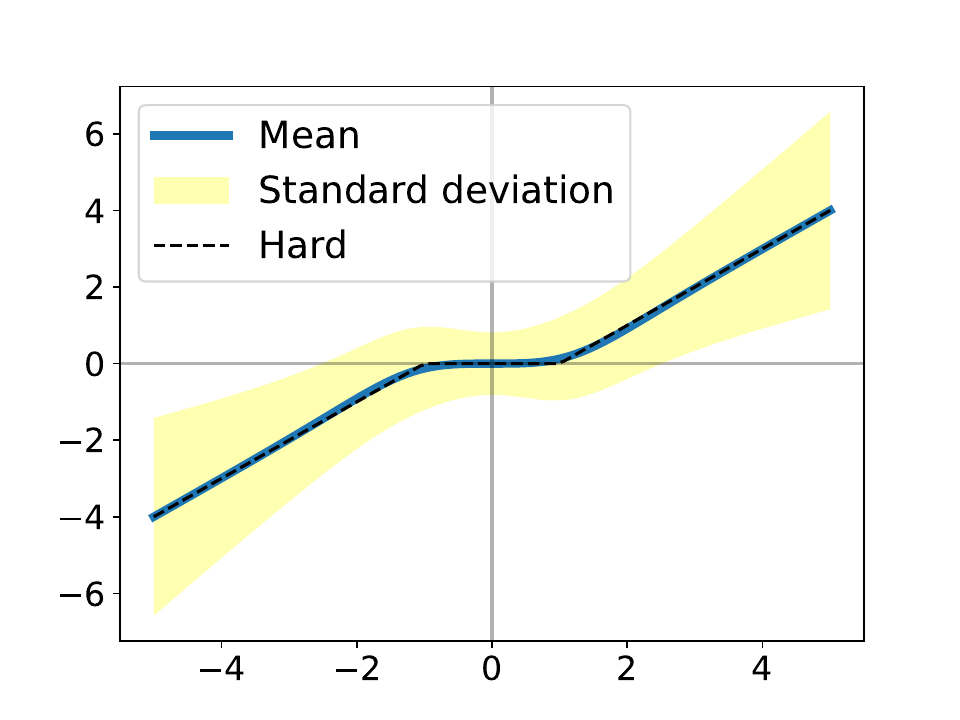}
\caption{A conditional with three branches: 
the soft-thresholding operator
(see \cref{cf:exm:st}). 
It is a piecewise linear function (dotted black line).
Using a softargmax, we can induce a categorical probability distribution over 
the three branches. The expected value (blue line)
can be seen as a smoothed out version of the operator. 
The induced distribution allows us to also compute the standard deviation.
}
\label{cf:fig:st}
\end{figure}

\begin{boxexm}{Soft-thresholding operator}
The soft-thresholding operator (see also \cref{optim:sec:prox_grad})
is a commonly-used operator to promote sparsity. It is defined by
\begin{equation*}
\mathrm{SoftThreshold}(u, \lambda) \coloneqq
\begin{cases}
0 &\mbox{ if } |u| \le \lambda \\
u - \lambda &\mbox{ if } u \ge \lambda \\
u + \lambda &\mbox{ if } u \le -\lambda
\end{cases}.
\end{equation*}
To express it in our formalism,
we can define $\piv \in \{\e_1, \e_2, \e_3\}$ 
using comparison operators as
\begin{align*}
\piv 
&\coloneqq (\mathrm{lt}(|u|, \lambda), 
\mathrm{gt}(u, \lambda), 
\mathrm{lt}(u, -\lambda)) \\
&= (\heavistep(\lambda - |u|),
\heavistep(u - \lambda), \heavistep(-u - \lambda)).
\end{align*}
Equivalently, we can also define $\piv$ using an argmax operator as
\begin{equation*}
\piv \coloneqq \mathrm{argmax}((\lambda - |u|, u - \lambda, -u - \lambda)).
\end{equation*}
In case of ties, which happens at $|u|=\lambda$,
we keep only one non-zero coordinate in $\piv$.
We can then rewrite the operator as
\begin{equation*}
\mathrm{SoftThreshold}(u, \lambda) = \mathrm{cond}(\piv, 0, u - \lambda, u + \lambda).
\end{equation*}
As we will see, replacing $\mathrm{argmax}$ with $\mathrm{softargmax}$ induces a
categorical distribution over the three possible branches. The mean value can be
seen as a smoothed out version of the operator, and we can also compute the
standard deviation, as illustrated in \cref{cf:fig:st}.  
\label{cf:exm:st}
\end{boxexm}

\subsection{Differentiating through branch variables}

For $\piv$ fixed,
$\mathrm{cond}(\piv, \v_1, \dots, \v_K)$
is a valid function \wrt $\v_i$, and can therefore again
be used as a node in a computational graph.
Due to the linearity \wrt $\v_i$, we obtain
that the Jacobian \wrt $\v_i$ is
\begin{align*}
\partial_{\v_i} \mathrm{cond}(\piv, \v_1, \dots, \v_K)
&\coloneqq 
\begin{cases}
I &\mbox{ if } \piv = \e_i \\
\zeros &\mbox{ if } \piv \neq \e_i 
\end{cases}.
\end{align*}
Let $g_i \colon \cU_i \to \cV$ be a differentiable function
and $\u_i \in \cU_i$.
If we define the composition
\begin{equation*}
f(\piv, \u_1, \dots, \u_K)
\coloneqq
\mathrm{cond}(\piv, g_1(\u_1), \dots, g_K(\u_K)),
\end{equation*}
we then obtain that the Jacobian \wrt $\u_i$ is
\begin{align*}
\partial_{\u_i}
f(\piv, \u_1, \dots, \u_K)
&\coloneqq 
\begin{cases}
\partial g_i(\u_i) &\mbox{ if } \piv = \e_i \\
\zeros &\mbox{ if } \piv \neq \e_i 
\end{cases}.
\end{align*}
As long as the $g_i$ functions are differentiable, 
we can therefore differentiate through the branch variables 
$\u_i$ for $\piv$ fixed.

\subsection{Differentiating through predicate variables}

As we saw,
$\piv$ can be obtained by combining comparison and logical operators, or it can
be obtained by argmax and argmin operators.
We illustrate here why these operators are problematic.
For example, 
suppose we want to express the function
\begin{equation*}
f_a(\p, \u_1, \dots, \u_K) \coloneqq
\begin{cases}
\v_1 &\mbox{ if } \p = \e_1 \\
\vdots & \\
\v_K &\mbox{ if } \p = \e_K
\end{cases}.
\end{equation*}
In our notation, this can be expressed as
\begin{equation*}
f_a(\p, \u_1, \dots, \u_K)
\coloneqq
\mathrm{cond}(\mathrm{argmax}(\p), g_1(\u_1), \dots, g_K(\u_K)),
\end{equation*}
As for the ifelse case, the Jacobian \wrt $\p$ is null almost everywhere,
\begin{equation*}
\partial_\p f_a(\p, \u_1, \dots, \u_K) = \zeros.
\end{equation*}

\subsection{Continuous relaxations}

Similarly to the Heaviside step function, the argmax and argmin functions are
piecewise constant, with discontinuities in case of ties. Their Jacobians are
zero almost everywhere, and undefined in case of ties. Therefore, while their
Jacobian is well-defined almost everywhere, they are uninformative and prevent
gradient backpropagation. We can replace the argmax with a \textbf{softargmax}
\begin{equation*}
\mathrm{softargmax}(\p) \coloneqq 
\frac{\exp(\p)}{\sum_{i=1}^K \exp(p_i)} \in \triangle^K
\end{equation*}
and similarly
\begin{equation*}
\mathrm{softargmin}(\p) \coloneqq \mathrm{softargmax}(-\p) \in \triangle^K.
\end{equation*}
Other relaxations of the argmax are possible, 
as discussed in \cref{inf_conv:sec:relaxed_argmax}.
See also \cref{perturb:sec:argmax} for the perturbation perspective.

Fortunately, the definition
\begin{equation*}
\mathrm{cond}(\piv, \v_1, \dots, \v_K)
= \sum_{i=1}^K \pi_i \v_i
\end{equation*}
is perfectly valid if we use
$\piv \in \triangle^K$ 
instead of
$\piv \in \{\e_1,\dots,\e_K\}$,
and can therefore be seen as an \textbf{extension} of \cref{cf:eq:cond}.
If we now define
\begin{align}
f_s(\p, \u_1, \dots, \u_K) 
&\coloneqq \mathrm{cond}(\softargmax(\p), g_1(\u_1), \dots, g_K(\u_K)) 
\nonumber \\
&= \sum_{i=1}^K [\softargmax(\p)]_i \cdot g_i(\u_i),
\label{cf:eq:soft_cond}
\end{align}
the Jacobian becomes
\begin{equation*}
\partial_\p f_s(\p, \u_1, \dots, \u_K)    
= 
\partial \softargmax(\p)(g_1(\u_1), \dots, g_K(\u_K)),
\end{equation*}
which is \textbf{non-null everywhere}, allowing gradients to backpropagate
through the computational graph.

\subsubsection*{Probabilistic perspective}

From a probabilistic perspective,
we can view \cref{cf:eq:soft_cond} 
as the expectation of $g_i(\u_i)$, where $i \in [K]$ 
is a categorical random variable
distributed according to a \textbf{categorical distribution} with parameter 
$\piv = \softargmax(\p)$:
\begin{equation*}
f_s(\p, \u_1, \dots, \u_K)
= \EE_{i \sim \mathrm{Categorical}(\softargmax(\p))}\left[g_i(\u_i)\right].
\end{equation*}
Taking the expectation over the $K$ possible branches makes the function
differentiable with respect to $\p$, at the cost of
evaluating all branches, instead of a single one.
Similarly as for the if-else case, we can compute the variance if needed as
\begin{align*}
& \VV_{i \sim \mathrm{Categorical}(\softargmax(\p))}\left[g_i(\u_i)\right] \\
=& 
\EE_{i \sim \mathrm{Categorical}(\softargmax(\p))}
\left[(f_s(\p, \u_1, \dots, \u_K) - g_i(\u_i))^2 \right].
\end{align*}
This is illustrated in \cref{cf:fig:st}.  

\section{For loops}
\label{cf:sec:for_loops}

\begin{figure}
	\begin{minipage}{0.48\linewidth}
    \vspace*{30pt}
		\begin{center}
      \includegraphics[width=\linewidth]{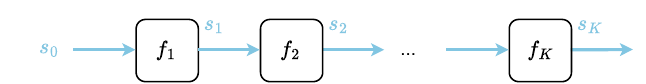}
		\end{center}
    \vspace*{20pt}
    \caption{A \textbf{for loop} forms a computation chain.
        A feed forward network can be seen as a parameterized for loop,
        where each function $f_k$ depends on some parameters $\w_k$.
    \label{cf:fig:for_loop}
    }
  \end{minipage}~
	\begin{minipage}{0.48\linewidth}
		\begin{center}
      \includegraphics[width=\linewidth]{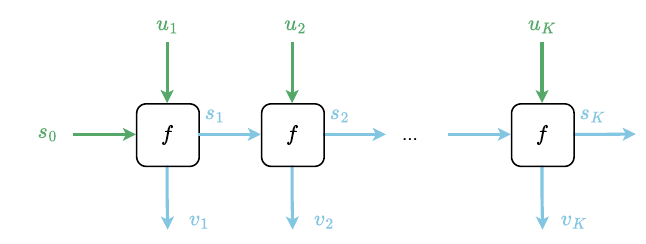}
		\end{center}
        \caption{Computation graph of the \textbf{scan function}.
    Sequence-to-sequence RNNs can be seen as a parameterized scan function.
  \label{cf:fig:scan}
  }
	\end{minipage}
\end{figure}

For loops are a control flow for sequentially calling a fixed number $K$ of 
functions, reusing the output from the previous iteration.
In full generality, a for loop can be written as follows.
\begin{algorithm}[H]
\caption{$\r = \mathrm{forloop}(\s_0)$}
\begin{algorithmic}
\For {$k \coloneqq 1, \ldots, K$}
\State $\s_k \coloneqq f_k(\s_{k-1})$
\EndFor
\State $\r \coloneqq \s_K$
\end{algorithmic}
\end{algorithm}
As illustrated in \cref{cf:fig:for_loop},
this defines a computation chain. Assuming the functions $f_k$ are all
differentiable, this defines a valid computation graph, we can therefore use
automatic differentiation to differentiate $\mathrm{forloop}$ \wrt its input
$\s_0$. Feedforward networks, reviewed in \cref{neural_nets:sec:ff}, can be seen
as \textbf{parameterized for loops}, i.e.,
\begin{equation*}
f_k(\s_{k-1}) \coloneqq g_k(\s_{k-1}, \w_k),
\end{equation*}
for some differentiable function $g_k$.

\begin{boxexm}{Unrolled gradient descent}
Suppose we want to minimize \wrt $\w$ the function
\begin{equation*}
L(\w, \lambda) 
\coloneqq 
\frac{1}{N} \sum_{i=1}^N \ell(h(\x_i, \w), \y_i) + \frac{\lambda}{2}
\|\w\|^2_2.
\end{equation*}
Given an initialization $\w_0$, 
gradient descent (\cref{optim:sec:gd}) performs iterations of the
form
\begin{equation*}
\w_k 
= f(\w_{k-1}, \gamma_k, \lambda)
\coloneqq \w_{k-1} - \gamma_k \nabla_1 L(\w_{k-1}, \lambda).
\end{equation*}
Gradient descent can therefore be expressed as
a for loop with
\begin{equation*}
f_k(\w_{k-1}) \coloneqq f(\w_{k-1}, \gamma_k, \lambda).
\end{equation*}
This means that we can differentiate through the iterations of 
gradient descent, as long as $f$ is differentiable, meaning that $L$ is twice
differentiable.
This is useful for instance to perform gradient-based optimization
of the hyperparameters $\gamma_k$ or $\lambda$.
This is a special case of bilevel optimization; see also \cref{chap:imp_diff}.
\end{boxexm}

\begin{boxexm}{Bubble sort}
Bubble sort is a simple sorting algorithm that works by repeatedly swapping
elements if necessary. Mathematically, swapping two elements $i$ and $j$
can be written as a function from $\RR^N \times [N] \times [N]$ to $\RR^N$
defined by
\begin{equation*}
\mathrm{swap}(\v, i, j) 
\coloneqq \v + (v_j - v_i) \e_i + (v_i - v_j) \e_j.
\end{equation*}
We can then write bubble sort as
\begin{align*}
&\text{\bf for} ~ i \coloneqq 1,\dots,N ~ \text{\bf do} \\
&\quad \text{\bf for} ~ j \coloneqq 1, \dots, N - i - 1 ~ \text{\bf do} \\
&\quad \quad \v' \coloneqq \mathrm{swap}(\v, j, j+1) \\
&\quad \quad \pi \coloneqq \heavistep(v_j - v_{j+1}) \\
&\quad \quad \v \leftarrow \mathrm{ifelse}(\pi, \v', \v)
\end{align*}
Replacing the Heaviside step function with the logistic function
gives a smoothed version of the algorithm.
\end{boxexm}

\section{Scan functions}

Scan is a higher-order function
(meaning a function of a function)
originating from functional programming.
It is useful to perform an operation $f$ on individual elements $\u_k$
while carrying the result $\s_k$ of that operation to the next iteration.
\begin{algorithm}[H]
\caption{$\r = \mathrm{scan}(\s_0, \u_1, \dots, \u_K)$}
\begin{algorithmic}
\For {$k \coloneqq 1, \ldots, K$}
\State $\s_k, \v_k \coloneqq f(\s_{k-1}, \u_k)$
\EndFor
\State $\r \coloneqq (\s_K, \v_1, \dots, \v_K)$
\end{algorithmic}
\end{algorithm}

As illustrated in \cref{cf:fig:scan},
this again defines a valid computational graph
and can be differentiated through using autodiff,
assuming the function $f$ is differentiable.
Sequence-to-sequence RNNs,
reviewed in \cref{seq_net:sec:rnn},
can be seen as a \textbf{parameterized scan}.
An advantage of this abstraction is that
parallel scan algorithms have been studied extensively 
in computer science \citep{blelloch_1989,sengupta_2010}.

\begin{boxexm}{Prefix sum}
Scan can be seen as a generalization of the \textbf{prefix sum}
(\aka \textbf{cumulative sum}) from the addition to any binary operation.
Indeed, a prefix sum amounts to perform 
\begin{align*}
\v_1 &\coloneqq \u_1 \\
\v_2 &\coloneqq \u_1 + \u_2 \\
\v_3 &\coloneqq \u_1 + \u_2 + \u_3 \\
     &\vdots
\end{align*}
which can be expressed as a scan by defining
\begin{align*}
    \v_k &\coloneqq \s_{k-1} + \u_k \\
    f(\s_{k-1}, \u_k) &\coloneqq (\v_k, \v_k)
\end{align*}
starting from $\s_0 = \zeros$
($\s_K$ and $\v_K$ are redundant in this case).
\end{boxexm}

\section{While loops}
\label{cf:sec:while_loops}

\subsection{While loops as cyclic graphs}

A while loop is a control flow used to repeatedly perform an operation,
reusing the output of the previous iteration,
until a certain condition is met.
Suppose 
$f \colon \cS \to \{0,1\}$
is a function to determine whether to stop ($\pi=1$) or continue ($\pi=0$) and
$g \colon \cS \to \cS$
is a function for performing an operation.
Then, without loss of generality, a while loop can be written as follows.
\begin{algorithm}[H]
\caption{$\r = \mathrm{whileloop}(\s)$}
\begin{algorithmic}
\State $\pi \leftarrow f(\s)$
\While {$\pi = 0$}
\State $\s \leftarrow g(\s)$
\State $\pi \leftarrow f(\s)$
\EndWhile 
\State $\r \coloneqq \s$
\end{algorithmic}
\end{algorithm}

This definition is somewhat cyclic, as we used the \textbf{while} keyword.
However, we can equivalently rewrite the algorithm recursively.
\begin{algorithm}[H]
\caption{$\r = \mathrm{whileloop}(\s)$
\label{cf:algo:while}
}
\begin{algorithmic}
\State $\pi \coloneqq f(\s)$
\If{$\pi = 0$}
\State $\r \coloneqq \s$
\Else
\State $\r \coloneqq \mathrm{whileloop}(g(\s))$
\EndIf
\end{algorithmic}
\end{algorithm}

\begin{figure}[t]
  \centering
  \includegraphics[width=0.5\linewidth]{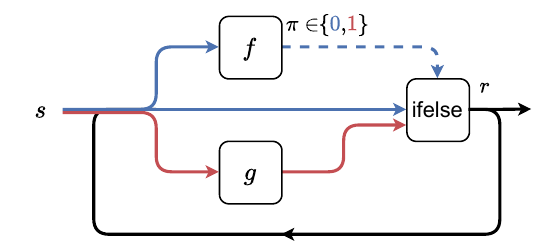}
  \caption{ A while loop can be represented as a cyclic graph. 
      The while loop stops if $\pi = 1$ and performs another iteration $\s
      \leftarrow g(\s)$, $\pi \leftarrow f(\s)$ if $\pi = 0$.
  \label{cf:fig:rolled_while_loop}
  }
\end{figure}

Unlike for loops and scan, the number of iterations of while loops is not known
ahead of time, and may even be infinite. In this respect, a while loop can be
seen as a \textbf{cyclic graph}, as illustrated
in~\cref{cf:fig:rolled_while_loop}.

\subsubsection*{Importance of lazy evaluation}

We can also implement \cref{cf:algo:while}
in terms of the $\mathrm{ifelse}$ function defined in \cref{cf:sec:if_else}
as
\begin{align*}
\r 
&\coloneqq \mathrm{ifelse}(f(\s), \mathrm{whileloop}(g(\s)), \s) \\
&=  f(\s) \cdot \mathrm{whileloop}(g(\s)) + (1 - f(\s)) \cdot \s.
\end{align*}
However, to avoid an infinite recursion,
it is crucial that $\mathrm{ifelse}$ supports \textbf{lazy evaluation}.
That is, $\mathrm{whileloop}(g(\s))$ in the definition above
should be evaluated if and only if $\pi = f(\s) = 1$.
In other words, the fact that $f(\s) \in \{0,1\}$ is crucial to ensure
that the recursion is well-defined.

\subsection{Unrolled while loops}

\begin{figure}[t]
  \centering
  \includegraphics[width=\linewidth]{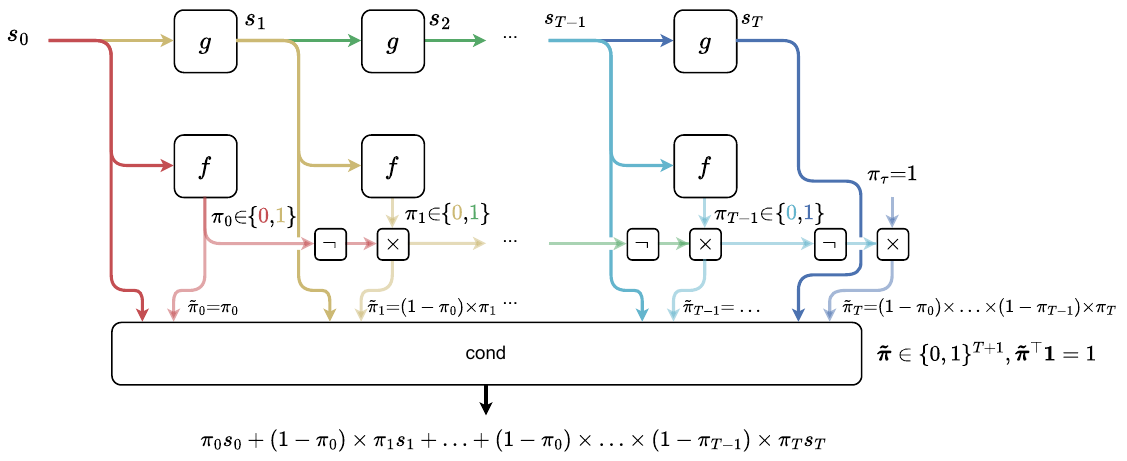}
  \caption{Computation graph of an unrolled truncated while loop.
    As in \cref{cf:fig:if_else_flows}, we depict
    continuous variables in dense lines and discrete variables in dashed lines.
    The output of a while loop with at most $T$ iterations can be written as a
    conditional with $T+1$ branches, $\mathrm{cond}(\tilde{\piv}, \s_0, \dots,
    \s_T) = \sum_{t=0}^T \tilde{\pi}_t \s_t$.
    \label{cf:fig:unrolled_while_loop}
  }
\end{figure}

To avoid the issues with unbounded while loops, 
we can enforce that a while loop stops after $T$ iterations, 
i.e., we can truncate the while loop.
Unrolling \cref{cf:algo:while} gives (here with $T=3$)
\begin{align*}
    &\pi_0 \coloneqq f(\s_0) \\
    &\text{\bf{if}} ~ \pi_0 = 1 ~ \textbf{\bf{then}} \\
    &\quad \r \coloneqq \s_0 \\
    &\text{\bf{else}} \\
    &\quad \s_1 \coloneqq g(\s_0), \pi_1 \coloneqq f(\s_1) \\
    &\quad \text{\bf{if}} ~ \pi_1 = 1 ~ \textbf{\bf{then}} \\
    &\quad \quad \r \coloneqq \s_1 \\
    &\quad \text{\bf{else}} \\
    &\quad \quad \s_2 \coloneqq g(\s_1), \pi_2 \coloneqq f(\s_2) \\
    &\quad \quad \text{\bf{if}} ~ \pi_2 = 1 ~ \textbf{\bf{then}} \\
    &\quad \quad \quad \r \coloneqq \s_2 \\
    &\quad \quad \text{\bf{else}} \\
    &\quad \quad \quad \r \coloneqq \s_3 \coloneqq g(\s_2) \\
\end{align*}
Using the $\mathrm{ifelse}$ function, we can rewrite it as
\begin{align*}
\r = \mathrm{ifelse}(&\pi_0, \\
                     &\s_0, \\
                     &\mathrm{ifelse}(\pi_1, \\
                     & \hspace{1cm}   \s_1,  \\
                     & \hspace{1cm}   \mathrm{ifelse}(\pi_2, \\
                     & \hspace{2cm}                   \s_2, \\
                     & \hspace{2cm}                   \s_3)))
\end{align*}
which is itself equivalent to
\begin{align*}
\r 
&= \pi_0 \s_0 + (1-\pi_0) \left[ \pi_1 \s_1 + (1-\pi_1) \left[ \pi_2 \s_2 +
(1-\pi_2) \s_3 \right] \right] \\
&= \pi_0 \s_0 + 
(1-\pi_0) \pi_1 \s_1 +
(1-\pi_0) (1- \pi_1) \pi_2 \s_2 +
(1-\pi_0) (1- \pi_1) (1 - \pi_2) \s_3.
\end{align*}
More generally, for $T \in \NN$, the formula is
\begin{align*}
\r 
& = \sum_{i=0}^T \left((1-\pi_0) \dots (1-\pi_{i-1})\right) \pi_i \s_i  \\
& = \sum_{i=0}^T  \left(\prod_{j=0}^{i-1} (1 - \pi_j)\right) \pi_i \s_i,
\end{align*}
where we defined
\begin{align*}
  \s_i 
  & \coloneqq g(\s_{i-1}) \coloneqq g^{i}(\s_0) 
  \coloneqq \underbrace{g \circ \dots \circ g}_{i \text{ times}}(\s_0) \in \cS \\
  \pi_i & \coloneqq f(\s_i) \in \{0,1\}.
\end{align*}
See also \citep{petersen_2021}.
If we further define the shorthand notation
\begin{align*}
    \tilde{\pi}_0 &\coloneqq \pi_0\\
    \tilde{\pi}_i &\coloneqq \left(\prod_{j=0}^{i-1}(1 - \pi_j)\right)\pi_i
    \quad i \in \{1,\dots,T\},
\end{align*}
so that $\tilde{\piv} \coloneqq (\tilde{\pi}_0,
\tilde{\pi}_1,\dots,\tilde{\pi}_T) \in \triangle^{T+1}$
is a discrete probability distribution containing the probabilities to stop at
each of the $T$ iterations,
we can rewrite the output of a truncated while using a conditional,
\begin{equation*}
\r 
= \mathrm{cond}(\tilde{\piv}, \s_0, \s_1, \dots, \s_T)
= \sum_{t=0}^T \tilde{\pi}_t \s_t.
\end{equation*}
This is illustrated in \cref{cf:fig:unrolled_while_loop}.

\begin{boxexm}{Computing the square root using Newton's method}\label{cf:exm:square_root}
Computing the square root $\sqrt{x}$ of a real number $x > 0$ 
can be cast as a root finding problem, 
which we can solve using Newton's method. 
Starting from an initialization $s_0$, the iterations read
\begin{equation*}
s_{i+1} 
\coloneqq g(\s_i) 
\coloneqq \frac{1}{2} \left(s_i + \frac{x}{s_i}\right).
\end{equation*}
To measure the error on iteration $i$, we can define
\begin{equation*}
\varepsilon(s_i) \coloneqq \frac{1}{2}(s_i^2 - x)^2.
\end{equation*}
As a stopping criterion, we can then use
\begin{align*}
    \pi_i &\coloneqq 
\begin{cases}
    1 &\mbox{ if } \varepsilon(s_i) \le \tau \\
    0 &\mbox{ otherwise }
\end{cases} \\
          &= \heavistep(\tau - \varepsilon(s_i)),
\end{align*}
where $0 < \tau \ll 1$ is an error tolerance 
and 
$\heavistep$ is the Heaviside step function.
\end{boxexm}

\subsection{Markov chain perspective}

Given the function $g \colon \cS \to \cS$ and the initialization $\s_0 \in \cS$, 
a while loop can only go through a discrete set
of
values $\s_0, \s_1, \s_2, \ldots$ 
defined by $\s_i = g(\s_{i-1})$. 
This set is potentially countably infinite if the while loop is unbounded,
and finite if the while loop is guaranteed to stop.
Whether the loop
moves from the state $\s_i$ to the state $\s_{i+1}$, or stays at $\s_i$, is
determined by the stopping criterion $\pi_i\in\{0, 1\}$. To model the state of
the while loop, we can then consider a \textbf{Markov chain} with a discrete
space $\{\s_0, \s_1, \s_2,\dots \}$, which we can always identify with $\{0,1,2,\dots
\}$, with transition probabilities
\begin{equation*}
\PP(S_{t+1} = \s_i | S_t = \s_j)
= p_{i,j} \coloneqq
\begin{cases}
\pi_i &\mbox{ if } i = j \\
(1-\pi_i) &\mbox{ if } i = j + 1 \\
0 &\mbox{ otherwise }
\end{cases},
\end{equation*}
and initial state $S_0 = \s_0$. Here, $S_t$ is the value at iteration $t$ of the
loop. Note that since $\pi_i \in \{0, 1\}$, the $p_{i,j}$ values
are ``degenerate'' probabilities. However, this framework lets us generalize
to a smooth version of the while loop naturally. To illustrate the framework, if
the while loop stops at $T=3$, the transition probabilities can be cast as a
matrix 
\begin{equation*}
\mathbf{P}
\coloneqq (p_{i,j})_{i,j=0}^T
\coloneqq
\begin{blockarray}{ccccc}
& \s_0 & \s_1 & \s_2 & \s_3 \\
\begin{block}{c(cccc)}
    \s_0 & 0 & 1 & 0 & 0 \\
    \s_1 & 0 & 0 & 1 & 0 \\
    \s_2 & 0 & 0 & 0 & 1 \\
    \s_3 & 0 & 0 & 0 & 1 \\
\end{block}
\end{blockarray}.
\end{equation*}
The output $\r$ of the while-loop is determined by the time at which
the state stays at the same value
\[
I = \min\{i \in \{1, 2, \ldots\} \ \mbox{s.t.} \  S_i = S_{i-1}\}.
\]
Note that $I$ itself is a random variable,
as it is defined by the $S_i$ variables. 
It is called a \textbf{stopping time}. 
The output of the chain is then 
\begin{align*}
  \r & = \EE[S_I] \\
  & = \sum_{i=1}^{+\infty} \PP(I= i) \EE[S_i | I = i]  \\
  & = \sum_{i=1}^{+\infty} \PP(I= i) \s_{i-1} \\
  & = \sum_{i=1}^{+\infty} \prod_{j=0}^{i-2}(1-\pi_j) \pi_{i-1}\s_{i-1} \\
  & = \sum_{i=0}^{+\infty} \prod_{j=0}^{i-1}(1-\pi_j) \pi_i\s_i.
\end{align*}
Because the stopping time is not known ahead of time,
the sum over $i$ goes from $0$ to $\infty$.
However, if we enforce in the stopping criterion that the while loop runs no
longer than $T$ iterations, by setting
\[
\pi_i \coloneqq \mbox{or}(f(\s_i), \mathrm{eq}(i, T)) \in \{0,1\},
\] 
we then naturally recover the expression 
found by unrolling the while loop before,
\begin{align*}
  \r & = \EE[S_I] = \sum_{i=0}^{T} \prod_{j=0}^{i-1}(1-\pi_j) \pi_i\s_i.
\end{align*}
For example, with $T=3$, the transition probability matrix is
\begin{equation*}
\mathbf{P} =
\begin{blockarray}{ccccc}
& \s_0 & \s_1 & \s_2 & \s_3 \\
\begin{block}{c(cccc)}
    \s_0 & \pi_0 & 1-\pi_0 & 0 & 0 \\
    \s_1 & 0 & \pi_1 & 1-\pi_1 & 0 \\
    \s_2 & 0 & 0 & \pi_2 & 1 - \pi_2 \\
    \s_3 & 0 & 0 & 0 & 1 \\
\end{block}
\end{blockarray}.
\end{equation*}

\subsubsection*{Smoothed while loops}

With the help of this framework, we can backpropagate even through the while
loop's stopping criterion, provided that we smooth out the predicate. 
For example, we saw that the stopping criterion in
\cref{cf:exm:square_root} is
$f(\s_i) = \heavistep(\tau - \varepsilon(\s_i))$
and therefore
\[
\pi_i \coloneqq \mbox{or}(f(\s_i), \mathrm{eq}(i, T)) \in \{0,1\}.
\] 
Due to the $\heavistep$ function, the derivative of the while loop with respect
to $\tau$ will always be $0$, just like it was the case for if-else statements. 
If we change the stopping criterion to
$f(\s_i) = \sigmoid(\tau - \varepsilon(\s_i))$,
we then have
(recall that $\mathrm{or}$ is well defined on $[0,1] \times [0,1]$)
\[
\pi_i \coloneqq \mathrm{or}(f(\s_i), \mathrm{eq}(i, T)) \in [0,1].
\] 
With $\sigmoid$, we obtain more informative derivatives. In particular, with
$\sigmoid = \logistic$, the derivatives \wrt $\tau$ are always non-zero.
The smoothed output is expressed as before as the expectation
\begin{equation*}
\r = \EE[S_I] = \sum_{i=0}^{T} \prod_{j=0}^{i-1}(1-\pi_j) \pi_i\s_i
\end{equation*}
where
\begin{equation*}
\pi_i \coloneqq \sigmoid(\tau - \varepsilon(\s_i)).
\end{equation*}
Instead of enforcing a number $T$ of iterations, it is also possible to stop
when the probability of stopping becomes high enough \citep{petersen_2021}, 
assuming that the probability of stopping converges to $1$.

\section{Summary}

\begin{itemize}
\item For conditionals, we saw that differentiating through the branch variables is
not problematic given a fixed predicate.

\item However, for the predicate variable, we saw that a
differentiable relaxation is required to avoid null derivatives.

\item We introduced soft comparison operators in a principled manner, using
a stochastic process perspective, as well as the continuous extension of logical
operators.

\item For loops and scan define valid computational graphs, 
as their number of iterations is fixed ahead of time.
Feedforward networks and RNNs can be seen as parameterized for loops and scan, 
respectively.

\item Unlike for loops and scan, the number of iterations of while loops
is not known ahead of time and may even be infinite.
However, unrolled while loops define valid directed acyclic graphs.
We defined a principled way to differentiate through the stopping criterion of a
while loop, thanks to a Markov chain perspective.

\end{itemize}

%% file: chapters/data_struct/data_struct.tex
\chapter{Data structures}
\label{chap:ds}

In computer science, a data structure is a specialized format for organizing,
storing and accessing data.  Mathematically, a data structure forms a so-called
algebraic structure: it consists of a set and the functions to operate on that
set.  In this chapter, we review how to incorporate data structures into
differentiable programs, with a focus on lists and dictionaries.

\section{Lists}
\label{ds:lists}

A list is an ordered sequence of elements.
We restrict ourselves to lists whose elements all belong to the same 
value space $\cV$.  
Formally, we denote a list of fixed length $K$ with values in $\cV$ by
a $K$-tuple
\begin{equation*}
\l \coloneqq (\l_1, \dots, \l_K) \in \cL_{K}(\cV) 
\end{equation*}
where each $\l_i \in \cV$ and where
\begin{equation*}
\cL_K(\cV) \coloneqq \cV^K = \underbrace{\cV \times \dots \times
\cV}_{K\text{ times}}.
\end{equation*}

\subsection{Basic operations}

\subsubsection*{Getting values}

We first present how to retrieve values from a list $\l \in \cL_K(\cV)$.
We define the function 
$\mathrm{list.get} \colon \cL_K(\cV) \times [K] \to \cV$ as
\begin{equation*}
\mathrm{list.get}(\l, i) \coloneqq \l_i.
\end{equation*}
The function is continuous and differentiable in $\l \in \cL_K(\cV)$ 
but not in $i \in [K]$, as it is a discrete variable.
In the particular case $\cV = \RR$, $\cL_K(\cV)$ is equivalent to $\RR^K$ and
we can therefore write
\begin{equation*}
\mathrm{list.get}(\l, i) = \langle \l, \e_i \rangle,
\end{equation*}
where $\{\e_1,\dots,\e_K\}$ is the standard basis of $\RR^K$.

\subsubsection*{Setting values}

We now present how to replace values from a list $\l \in \cL_K(\cV)$.
We define the function
$\mathrm{list.set} \colon \cL_K(\cV) \times [K] \times \cV \to \cL_K(\cV)$ 
as
\begin{equation*}
[\mathrm{list.set}(\l, i, \v)]_j
\coloneqq
\begin{cases}
    \v &\mbox{ if } i = j \\
    \l_j &\mbox{ if } i \neq j
\end{cases},
\end{equation*}
for $j \in [K]$.
In the functional programming spirit,
the function returns the \textbf{whole} new list, 
even though a single element has been
modified.  Again, the function is continuous and differentiable in $\l \in
\cL_K(\cV)$ and $\v \in \cV$ but not in $i \in [K]$.
In the particular case $\cV = \RR$, 
given a list $\l = (l_1,\dots,l_K)$,
we can set the value $v \in \RR$ by
\begin{equation*}
\mathrm{list.set}(\l, i, v) = \l + (v - l_i) \e_i.
\end{equation*}
That is, we subtract the old value $l_i$ and add the new value $v$ at the
location $i \in [K]$ in $\l$.

\subsubsection*{Implementation}

A fixed-length list can be implemented as an \textbf{array},
which enables $O(1)$ random access to individual elements.
The hardware counterpart of an array is random access memory (RAM),
in which memory can be retrieved by address (location).

\subsection{Operations on variable-length lists}

So far, we focused on lists of fixed length $K$.
We now turn our attention to
variable-length lists, whose size can decrease or increase
over time. 
In addition to the $\mathrm{list.get}$ and $\mathrm{list.set}$ 
functions, they support functions that can change the size of a list.

\subsubsection*{Initializing lists}

In order to initialize a list, we define
$\mathrm{list.init} \colon \cV \to \cL_1(\cV)$
as
\begin{equation*}
\mathrm{list.init}(\v) \coloneqq (\v),
\end{equation*}
where we used $(\v)$ to denote a $1$-tuple.

\subsubsection*{Pushing values}

In order to add new values either to the left or to the right, we define
$\mathrm{list.pushLeft} \colon \cL_K(\cV) \times \cV \to \cL_{K+1}(\cV)$ 
as
\begin{equation*}
\mathrm{list.pushLeft}(\l, \v) \coloneqq (\v, \l_1, \dots, \l_K).
\end{equation*}
and
$\mathrm{list.pushRight} \colon \cL_K(\cV) \times \cV \to \cL_{K+1}(\cV)$ 
as
\begin{equation*}
\mathrm{list.pushRight}(\l, \v) \coloneqq (\l_1, \dots, \l_K, \v).
\end{equation*}

\subsubsection*{Popping values}

In order to remove values either from the left or from the right, we define
$\mathrm{list.popLeft} \colon \cL_K(\cV) \to \cL_{K-1}(\cV) \times \cV$ 
as
\begin{equation*}
\mathrm{list.popLeft}(\l) \coloneqq (\l_2, \dots, \l_{K}), \l_1
\end{equation*}
and
$\mathrm{list.popRight} \colon \cL_K(\cV) \to \cL_{K-1}(\cV) \times \cV$ 
as
\begin{equation*}
\mathrm{list.popRight}(\l) \coloneqq (\l_1, \dots, \l_{K-1}), \l_K.
\end{equation*}
The set $\cL_0(\cV)$ is a singleton which contains the empty list.

\subsubsection*{Inserting values}

The pushLeft and pushRight functions can only insert values at the beginning 
and at the end of a list, respectively.
We now study the insert function, whose goal is to be able to add a 
new value at an arbitrary location, shifting all values to the right and
increasing the list size by $1$.
We define the function 
$\mathrm{list.insert}: \cL_K(\cV) \times [K+1] \times \cV \rightarrow
\cL_{K+1}(\cV)$ as
\begin{equation*}
[\mathrm{list.insert}(\l, i, \v)]_j
\coloneqq
\begin{cases}
\l_j & \mbox{if} \ j < i \\
\v & \mbox{if} \ j = i \\
\l_{j-1} & \mbox{if} \ j > i
\end{cases},
\end{equation*}
for $j \in [K+1]$.
As for the $\mathrm{list.set}$ function,
$\mathrm{list.insert}$ is readily continuous and differentiable in $\l$ and
$\v$, but not in $i$, as it is a discrete variable.
As special cases, we naturally recover
\begin{align*}
\mathrm{list.insert}(\l, 1, \v) & = \mathrm{list.pushLeft}(\l, \v), \\
\mathrm{list.insert}(\l, K+1, \v) & = \mathrm{list.pushRight}(\l, \v).
\end{align*}

\subsubsection*{Differentiability}

The $\mathrm{list.init}$, $\mathrm{list.push}$ and $\mathrm{list.pop}$ functions
are readily continuous and differentiable with
respect to their arguments (a continuous relaxation is not needed).
As for the $\mathrm{list.set}$ function, the $\mathrm{list.insert}$ function is
continuous and differentiable in $\l$ and $\v$, but not in $i$.

\subsubsection*{Implementation}

Under the hood, a variable-length list can be implemented as a
linked list or as a dynamic array. A linked list gives $O(K)$ random access
while a dynamic array allows $O(1)$ random access, at the cost of memory
reallocations.

\subsubsection*{Stacks and queues}

The $\mathrm{list.pushRight}$ and $\mathrm{list.popRight}$ functions can be used
to implement a \textbf{stack} (last in first out \aka LIFO behavior).
The $\mathrm{list.pushLeft}$ and $\mathrm{list.popRight}$ functions can be used
to implement a \textbf{queue} (first in first out \aka FIFO behavior).

\subsection{Continuous relaxations using soft indexing}

\subsubsection*{Getting values}

\begin{figure}[t]
  \centering
  \includegraphics[width=\linewidth]{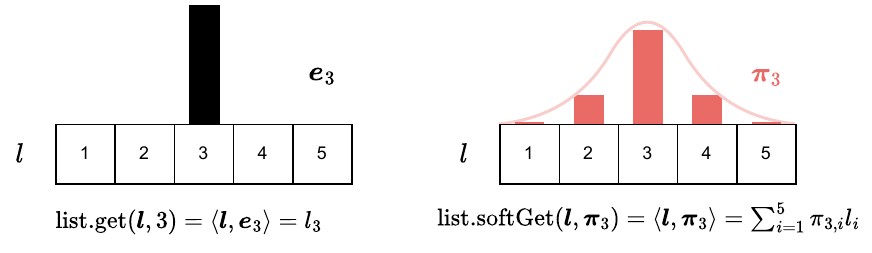}
  \caption{The $\mathrm{list.get}(\l, i)$ function is continuous and
      differentiable in $\l$ but not in $i$. 
      Its relaxation $\mathrm{list.softGet}(\l, \piv_i)$ is
      differentiable in both $\l$ and $\piv_i$.
      When $\cV=\RR$,
      $\mathrm{list.softGet}(\l, \piv_i)$ can be seen as taking the inner
      product between the list $\l$ and the probability distribution $\piv_i$,
      instead of the delta distribution (canonical vector) $\e_i$.
  \label{data_struct:fig:list_get}
  }
\end{figure}

In order to be able to differentiate $\mathrm{list.get}$ \wrt indexing,
a natural idea is to replace the integer index $i \in [K]$ by a distribution
$\piv_i \in \triangle^K$, which we can interpret as a \textbf{soft index}.
An integer index $i \in [K]$ is then equivalent to a 
\textbf{delta distribution} 
$\piv_i \in \{\e_1, \dots, \e_K\}$.
We define the continuous relaxation
$\mathrm{list.softGet} \colon \cL_K(\cV) \times \triangle^K \to \conv(\cV)$
as
\begin{align*}
\mathrm{list.softGet}(\l, \piv_i)
&\coloneqq \sum_{j=1}^K \pi_{i,j} \l_j \\
&= \mathrm{cond}(\piv_i, \l_1, \dots, \l_K) \\
&= \EE_{I \sim \mathrm{Categorical}(\piv_i)}[\l_I],
\end{align*}
where $\mathrm{cond}$ is studied in \cref{cf:sec:else_if}.
In the particular case $\cV = \RR$, we obtain
\begin{equation*}
\mathrm{list.softGet}(\l, i) = \langle \l, \piv_i \rangle.
\end{equation*}
This is illustrated in \cref{data_struct:fig:list_get}. 

The choice of the distribution $\piv_i = (\pi_{i,1}, \dots, \pi_{i,K})$ encodes
the importance of the elements $(\l_1,\dots,\l_K)$ \wrt $\l_i$.
If we consider that the smaller $|i-j|$ is, 
the more related $\l_i$ and $\l_j$ are,
then it makes sense to define a distribution centered around $i$ (i.e., such
that the mode of the distribution is achieved at $i$).
For example, limiting ourselves to the \textbf{neighbors}
$\l_{i-1}$ and $\l_{i+1}$ 
(i.e., a window of size $1$), we can define the sparse distribution
\begin{equation*}
\piv_i \coloneqq 
\frac{1}{4} \cdot \e_{i-1} + \frac{1}{2} \e_i + \frac{1}{4} \cdot \e_{i+1} \in
\triangle^K.
\end{equation*}
In this particular case, the continuous relaxation of the $\mathrm{list.get}$
function can then be expressed as a \textbf{discrete convolution},
\begin{equation*}
\mathrm{list.softGet}(\l, \piv_i) 
= \left(\mathrm{list.get}(\l, \cdot) \ast \kappa\right)(i)
= \sum_{j=-\infty}^\infty \mathrm{list.get}(\l, i - j) \kappa(j),
\end{equation*}
where $\kappa(-1) \coloneqq \frac{1}{4}$, $\kappa(1) \coloneqq \frac{1}{4}$,
$\kappa(0) \coloneqq \frac{1}{2}$, and $\kappa(j) \coloneqq 0$ for $j \not \in
\{-1,0,1\}$. 
Assuming $\cV = \RR^M$, 
the computational complexity of $\mathrm{list.softGet}$ is 
$O(M \cdot |\mathrm{supp}(\piv_i)|)$.

\subsubsection*{Setting values}

To differentiate \wrt indexing,
we can define the continuous relaxation
$\mathrm{list.softSet} \colon \cL_K(\cV) \times \triangle^K \times \cV \to 
\cL_K(\conv(\cV))$ 
as
\begin{align*}
\left[\mathrm{list.softSet}(\l, \piv_i, \v)\right]_j
&\coloneqq 
\EE[\mathrm{list.set}(\l, I, \v)]_j \\
&= \PP(I=j) \v + \PP(I \neq j) \l_j \\
&= \pi_{i,j} \v + (1 - \pi_{i,j}) \l_j,
\end{align*}
where $j \in [K]$ and $I \sim \mathrm{Categorical}(\piv_i)$.
Equivalently, we can write
\begin{align*}
\mathrm{list.softSet}(\l, \piv_i, \v)
&= (\pi_{i,1} \v + (1-\pi_{i,1}) \l_1,
\dots,
\pi_{i,K} \v + (1-\pi_{i,K}) \l_K) \\
&=
(\mathrm{ifelse}(\pi_{i,1}, \v, \l_1),
\dots,
\mathrm{ifelse}(\pi_{i,K}, \v, \l_K)),
\end{align*}
where $\mathrm{ifelse}$ is studied in \cref{cf:sec:if_else}.
Since
\begin{equation*}
\mathrm{ifelse}(\pi, \u_1, \u_0)
= \EE_{I \sim \mathrm{Bernoulli}(\pi)}[\u_I],
\end{equation*}
this relaxation amounts to using an element-wise expectation.
As a result, the list output by $\mathrm{list.softSet}$
takes values in $\conv(\cV)$ instead of $\cV$.
Note however that when $\cV = \RR^M$, then $\conv(\cV) = \RR^M$ as well.

\subsubsection*{Inserting values}

To differentiate value insertion \wrt indexing,
we can define the continuous relaxation
$\mathrm{list.softInsert} \colon \cL_K(\cV) \times \triangle^{K+1} \times \cV
\to \cL_{K+1}(\conv(\cV))$ as
\begin{align*}
[\mathrm{list.softInsert}(\l, \piv_i, \v)]_j 
&\coloneqq \EE[\mathrm{list.insert}(\l, I, \v)] \\
& = \PP(I > j) \l_j + \PP(I = j) \v + \PP(I < j) \l_{j-1},
\end{align*}
where $I \sim \mathrm{Categorical}(\piv_i)$.
The three necessary probabilities can easily be calculated
for $j \in [K+1]$ by
\begin{align*}
    \PP(I > j) &= 
\begin{cases}
    0 &\mbox{ if } j = K+1 \\
    \sum_{k=j+1}^{K+1} \pi_{i, k} &\mbox{ otherwise }
\end{cases} \\
    \PP(I = j) &= \pi_{i,j} \quad  \\
    \PP(I < j) &= 
\begin{cases}
    0 &\mbox{ if } j = 1 \\
    \sum_{k=1}^{j-1} \pi_{i, k} &\mbox{ otherwise }
\end{cases}.
\end{align*}

\subsubsection*{Multi-dimensional indexing}

In multi-dimensional lists (arrays or tensors),
each element $\l_\iv \in \cV$ of a list 
$\l \in \cL_{K_1,\dots,K_T}(\cV)$
can now be indexed by a multivariate integer 
$\iv = (i_1, \dots, i_T) \in [K_1] \times \dots \times [K_T]$, 
where $T \in \NN$ is the number of axes of $\l$.
We can always flatten a multi-dimensional list into a uni-dimensional list
by replacing the multi-dimensional index 
$\iv \in [K_1] \times \dots \times [K_T]$ 
by a flat index 
$i \in [K_1 \dots K_T]$.
The converse operation, converting a flat uni-dimensional array into a
multi-dimensional array, is also possible.
Therefore, there is a \textbf{bijection} between $[K]$ and 
$[K_1] \times \dots \times [K_T]$ 
for
$K \coloneqq K_1 \dots K_T$.

This means that the previous discussion on soft indexing in the uni-dimensional
setting readily applies to the multi-dimensional setting.
All it takes is the ability to define a probability distribution $\piv_\iv \in
\triangle^{K_1 \times \dots \times K_T}$. For example, when working with images,
we can define a probability distribution putting probability mass only on the
neighboring pixels of pixel $\iv$, a standard approach in image processing.
Another simple approach is to use a product of axis-wise probability
distributions.

\section{Dictionaries}
\label{ds:dict}

A dictionary (\aka associative array or map) is an unordered list of
\textbf{key-value pairs}, such that each possible key appears at most once in
the list.  We denote the set of keys by $\cK$ and the set of values by $\cV$
(both being potentially infinite).
We can then define the set of dictionaries of size $L$ from $\cK$ to $\cV$ by
\begin{equation*}
\cD_L(\cK,\cV) 
\coloneqq \cL_L(\cK \times \cV)
= (\cK \times \cV)^L
\end{equation*}
and one such dictionary by
\begin{equation*}
\d \coloneqq ((\kv_1,\v_1), \dots, (\kv_L,\v_L)) \in \cD_L(\cK,\cV).
\end{equation*}

\subsection{Basic operations}

\subsubsection*{Getting values}

The goal of the $\mathrm{dict.get}$ function is to retrieve the value
associated with a key, assuming that the dictionary contains this key.
Formally, we define the 
$\mathrm{dict.get} \colon \cD_L(\cK,\cV) \times \cK \to \cV \cup \{\infty\}$
function as
\begin{equation*}
\mathrm{dict.get}(\d, \kv)
\coloneqq 
\begin{cases}
    \v_i &\mbox{ if } \exists i \in [L] \text{ s.t. } \kv = \kv_i \\
    \infty & \mbox{ if } \kv \not \in \{\kv_1,\dots,\kv_L\}
\end{cases}.
\end{equation*}
The function is continuous and differentiable in the dictionary $\d$, 
but not in the key $\kv$.
Equivalently, we can write the function as
\begin{align*}
\mathrm{dict.get}(\d, \kv) \coloneqq 
\frac{\sum_{i=1}^L \eq(\kv, \kv_i) \v_i}
{\sum_{i=1}^L \eq(\kv, \kv_i)}.
\end{align*}
The denominator encodes the fact that the function is undefined if no key in the
dictionary $\d$ matches the key $\kv$.
Assuming $\kv \in \{\kv_1,\dots,\kv_L\}$ and $\cV = \RR^M$, we can also write
\begin{equation*} 
\mathrm{dict.get}(\d, \kv) = \v_i
\quad \text{where} \quad
i = \argmax_{j \in [L]} \|\kv - \kv_j\|_2,
\end{equation*}
which shows that we can see $\mathrm{dict.get}$ as a \textbf{nearest neighbor
search}.

\subsubsection*{Setting values}

The goal of the $\mathrm{dict.set}$ function is to replace the value associated
with an existing key.
Formally, we define the
$\mathrm{dict.set}: \cD_L(\cK, \cV) \times \cK \times \cV 
\rightarrow \cD_L(\cK, \cV)$ function as 
\begin{equation*}
  (\mathrm{dict.set}(\d, \kv, \v))_i \coloneqq
  \begin{cases}
    (\kv_i, \v) & \mbox{if} \ \kv_i = \kv \\
    (\kv_i, \v_i) & \mbox{if} \ \kv_i \neq \kv
  \end{cases}.
\end{equation*}
The function leaves the dictionary unchanged if no key in the
dictionary matches the input key $\kv$.
The function is continuous and differentiable in $\d$ and $\v$, 
but not in $\kv$.

\subsubsection*{Implementation}

While we view dictionaries as lists of key-value pairs,
in practice, a dictionary (\aka associative array) is often implemented using a
hash table or search trees.
The hardware counterpart of a dictionary is called content-addressable memory
(CAM), \aka associative memory.

\subsection{Continuous relaxation using kernel regression}

\begin{figure}
  \centering
  \includegraphics[width=0.6\linewidth]{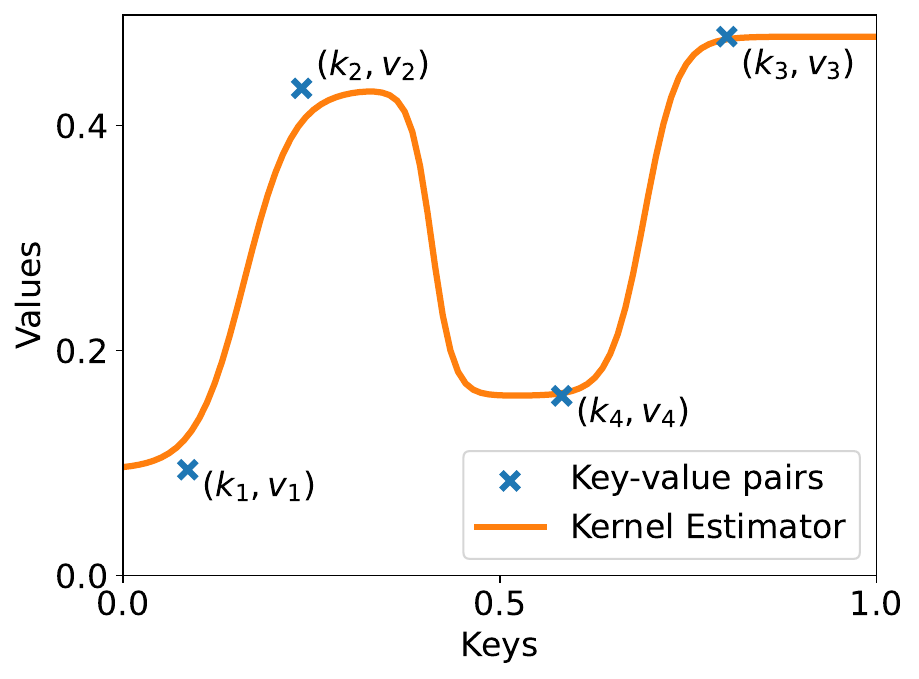}
  \caption{Given a set of key-value pairs $(\kv_i,\v_i) \in \cK \times \cV$
      defining a dictionary $\d$, 
  we can estimate a continuous mapping from $\cK$ to $\cV$ using
  Nadaraya–Watson kernel regression (here, illustrated with $\cK = \cV = \RR$).
  When keys are normalized to have unit norm, this recovers softargmax attention
  from Transformers.
  \label{data_struct:fig:dict_kernel_estim}}
\end{figure}

A dictionary can be seen as a (potentially non-injective) function that
associates a value $\v$ to each key $\kv$. To obtain a continuous relaxation of
the operations associated to a dictionary, 
we can adopt a probabilistic perspective of the mapping from keys to values. 
We can view keys and values as two continuous random variables $K$ and $V$. 
We can express the conditional PDF $f(\v|\kv )$ of $V|K$ in terms
of the joint PDF $f(\kv, \v)$ of $(K,V)$ and the marginal PDF $f(\kv)$ of $K$ as
\begin{equation*}
f(\v|\kv) = \frac{f(\kv, \v)}{f(\kv)}.
\end{equation*}
Integrating, we obtain the \textbf{conditional expectation}
\begin{align*}
\EE[V | K=\kv] 
= \int_{\cV} f(\v|\kv) \v d\v
= \int_{\cV} \frac{f(\kv, \v)}{f(\kv)} \v d\v.
\end{align*}
This is the \textbf{Bayes predictor}, in the sense that
$\EE[V|K]$ is the minimizer of
$\EE[(h(K) - V)^2]$ over the space of measurable functions $h \colon \cK \to
\cV$.
Using a sample of $L$ input-output pairs $(\kv_i, \v_i)$, corresponding to
key-value pairs in our case, \textbf{Nadaraya–Watson kernel regression}
estimates the joint PDF and the marginal PDF using \textbf{kernel density
estimation} (KDE). Using a product of isotropic kernels $\kappa_\sigma$ and
$\rho_\sigma$ for key-value pairs, we can define
\begin{equation*}
\widehat{f}_\sigma(\kv, \v) 
\coloneqq \frac{1}{L} \sum_{i=1}^L \kappa_\sigma(\kv - \kv_i) 
\rho_\sigma(\v - \v_i).
\end{equation*}
The corresponding marginal distribution on the keys is then given as
\begin{align*}
\widehat{f}_\sigma(\kv) 
&\coloneqq 
\int_{\cV} \widehat{f}_\sigma(\kv, \v)d\v\\
&=
\frac{1}{L} \sum_{i=1}^L \kappa_\sigma(\kv - \kv_i) \int_{\cV} \rho_\sigma(\v -
\v_i) d\v\\
&=\frac{1}{L} \sum_{i=1}^L \kappa_\sigma(\kv - \kv_i).
\end{align*}
Replacing $f$ with $\widehat{f}_\sigma$, we obtain the following estimator of
the conditional expectation
\begin{align*}
\widehat{\EE}[V|K=\kv]
&\coloneqq \int_{\cV} \frac{\widehat{f}_\sigma(\kv, \v)}{\widehat{f}_\sigma(\kv)} \v d\v \\
&= \int_{\cV} \frac{\frac{1}{L} \sum_{i=1}^L \kappa_\sigma(\kv - \kv_i) 
\rho_\sigma(\v - \v_i)}{\frac{1}{L} \sum_{i=1}^L \kappa_\sigma(\kv - \kv_i)} \v d\v \\
&=  \frac{\sum_{i=1}^L \kappa_\sigma(\kv - \kv_i)
\int_{\cV} \rho_\sigma(\v - \v_i) \v d\v}{\sum_{i=1}^L \kappa_\sigma(\kv -
\kv_i)} \\
&=  \frac{\sum_{i=1}^L \kappa_\sigma(\kv - \kv_i) \v_i}{\sum_{i=1}^L
\kappa_\sigma(\kv - \kv_i)}.
\end{align*}
In the above, we assumed that
$\rho_\sigma(\v - \v_i) = p_{\v_i,\sigma}(\v)$, 
where $p_{\v_i,\sigma}(\v)$ is the PDF of a distribution whose mean is
$\v_i$, so that
\begin{equation*}
\int_{\cV} \rho_\sigma(\v - \v_i)\v d\v 
= \EE_{V \sim p_{\v_i,\sigma}}[V] 
= \v_i.
\end{equation*}
Given a dictionary $\d = ((\kv_1,\v_1),\dots,(\kv_L,\v_L))$,
we can therefore define
the $\mathrm{dict.softGet} \colon \cD_L(\cK, \cV) \times \cK \to \conv(\cV)$ 
function as
\begin{equation*}
\mathrm{dict.softGet}(\d, \kv) 
\coloneqq \frac{\sum_{i=1}^L \kappa_\sigma(\kv - \kv_i) \v_i}{
\sum_{i=1}^L \kappa_\sigma(\kv - \kv_i)}.
\end{equation*}
This kernel regression perspective on dictionaries was previously pointed out by
\citet{zhang2021dive}. It is illustrated in
\cref{data_struct:fig:dict_kernel_estim} with $\cK = \cV = \RR$.

\subsection{Discrete probability distribution perspective}

\begin{figure}[t]
  \centering
  \includegraphics[width=0.75\linewidth]{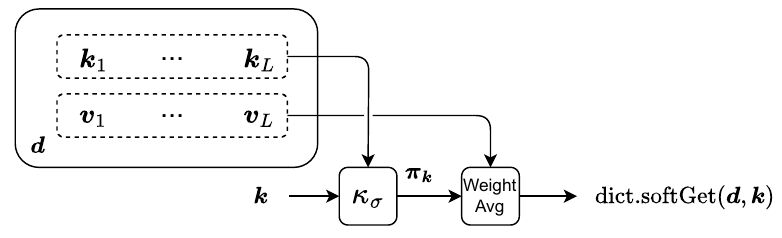}
  \caption{
    Computation graph of the $\mathrm{dict.softGet}$ function.
    We can use a kernel $\kappa_\sigma$ to produce a discrete probability
    distribution $\piv_\kv = (\pi_{\kv,1},\dots,\pi_{\kv,L}) \in \triangle^L$, 
    that captures the affinity between
    the dictionary keys $(\kv_1,\dots,\kv_L)$ and the input key $\kv$.
    The $\mathrm{dict.softGet}$ function can then merely be seen as a convex
    combination (weighted average) of values $(\v_1,\dots,\v_L)$ 
    using the probability values $(\pi_{\kv,1},\dots,\pi_{\kv,L})$ as weights.
    \label{data_struct:fig:dict_get}
  }
\end{figure}

While the set of possible keys $\cK$ is potentially infinite,
the set of keys $\{\kv_1,\dots,\kv_L\} \subset \cK$ associated with a particular
dictionary $\d = ((\kv_1,\v_1),\dots,(\kv_L,\v_L))$ is finite.
To a particular key $\kv$, we can therefore associate a discrete probability
distribution
$\piv_\kv = (\pi_{\kv,1},\dots,\pi_{\kv,L}) \in \triangle^L$ 
over the keys $(\kv_1,\dots,\kv_L)$ of $\d$, defined by
\begin{equation*}
\pi_{\kv,i} \coloneqq
\frac{\kappa_\sigma(\kv - \kv_i)}{\sum_{j=1}^L \kappa_\sigma(\kv - \kv_j)}
\quad \forall i \in [L].
\end{equation*}
This distribution captures the affinity between the input key $\kv$
and the keys $(\kv_1,\dots,\kv_L)$ of dictionary $\d$.
As illustrated in~\cref{data_struct:fig:dict_get},
we obtain
\begin{align*}
\mathrm{dict.softGet}(\d, \kv) 
&= \EE_{i \sim \mathrm{Categorical}(\piv_\kv)}[\v_i] \\
&= \sum_{i=1}^L \pi_{\kv,i} \v_i.
\end{align*}
In the limit $\sigma \to
0$, we recover
\begin{equation*}
\mathrm{dict.get}(\d, \kv) =
\frac{\sum_{i=1}^L \mathrm{eq}(\kv,\kv_i) \v_i}{\sum_{i=1}^L
\mathrm{eq}(\kv,\kv_i)}.
\end{equation*}

While the $\mathrm{dict.get}$ function is using a mapping from
keys $\kv \in \{\kv_1,\dots,\kv_L\}$ to integer indices $[L]$,
the $\mathrm{dict.softGet}$ function is using a mapping
from keys $\kv \in \{\kv_1,\dots,\kv_L\}$ 
to distributions $\piv_{\kv} \in \triangle^L$.
This perspective allows us to reuse the soft functions we developed for lists in
\cref{ds:lists}.
For example, we can softly replace the value associated with key $\kv$ by
performing
\begin{equation*}
\mathrm{list.softSet}(\d, \piv_\kv, (\kv, \v)).
\end{equation*}
Unlike $\mathrm{dict.set}$, the function is differentiable \wrt 
the distribution $\piv_\kv$.

\subsection{Link with attention in Transformers}

In the case when $\kappa_\sigma$ is the Gaussian kernel, assuming that the keys
are normalized to have unit norm (which is often the case in practical
implementations~\citep{schlag2021linear, dehghani2023scaling}), we obtain
\begin{align*}
\kappa_\sigma(\kv - \kv_i)
&= \exp(-\|\kv - \kv_i\|^2_2 / (2\sigma^2)) \\
&= \exp(-(\|\kv\|^2_2 + \|\kv_i\|^2_2) / (2\sigma^2))
\exp(\langle \kv, \kv_i \rangle / \sigma^2) \\
&= \exp(-1/\sigma^2)
\exp(\langle \kv, \kv_i \rangle / \sigma^2)
\end{align*}
so that
\begin{align*}
\pi_{\kv,i}
&= \frac{\kappa_\sigma(\kv - \kv_i)}{\sum_{j=1}^L \kappa_\sigma(\kv - \kv_j)}\\
&= \frac{\exp(\langle \kv, \kv_i \rangle / \sigma^2)}{
\sum_{j=1}^L \exp(\langle \kv, \kv_j \rangle / \sigma^2)}.
\end{align*}
We recognize the softargmax operator.
Given a dictionary \\
$\d = ((\kv_1,\v_1), \dots, (\kv_L,\v_L))$,
we thus recover attention from Transformers \citep{vaswani2017attention} as
\begin{align*}
\mathrm{dict.softGet}(\d, \kv) 
&= \sum_{i=1}^L \pi_{\kv,i} \v_i \\
&= \sum_{i=1}^L \frac{\exp(\langle \kv, \kv_i \rangle / \sigma^2) \v_i}{
\sum_{j=1}^L \exp(\langle \kv, \kv_j \rangle / \sigma^2)}.
\end{align*}
Transformers can therefore be interpreted as relying on a differentiable
dictionary mechanism.
Besides Transformers, content-based memory addressing is also used in neural
Turing machines \citep{graves2014neural}.

\section{Summary}

\begin{itemize}
    \item Operations on lists are continuous and differentiable \wrt the list,
        but not \wrt the integer index. Similarly, operations on dictionaries
        are continuous and differentiable \wrt the dictionary, but not \wrt the
        input key.

    \item Similarly to the way we handled the predicate in conditionals,
        we can replace the integer index (respectively the key) with a
        probability distribution over the indices (respectively the keys).

    \item This allows us to obtain a probabilistic relaxation of operations on
        lists. In particular, the relaxation for $\mathrm{list.get}$ amounts to
        performing a convolution. The relaxation for $\mathrm{dict.get}$
        amounts to computing a conditional expectation using kernel regression.

    \item When using a Gaussian kernel with keys normalized to unit norm,
        we recover softargmax attention from Transformers.
\end{itemize}

%% file: chapters/num_diff/num_diff_main.tex
\chapter{Finite differences}
\label{chap:finite_diff}

One of the simplest ways to numerically compute derivatives is to use
finite differences, 
which approximate the infinitesimal definition of derivatives. 
Finite differences only require \textbf{function evaluations}, and
can therefore work with blackbox functions: they
ignore the compositional structure of functions.
Without loss of generality, our exposition focuses on
computing directional derivatives $\partial f(\w)[\v]$, for a function $f
\colon \cE \to \cF$, evaluated at $\w \in \cE$, in the direction $\v \in
\cE$.

\section{Forward differences}

From \cref{diff:def:dir_deriv} and \cref{diff:def:jvp},
the directional derivative and more generally the JVP are defined as a limit,
\begin{equation*}
\partial f(\w)[\v] 
\coloneqq 
\lim_{\delta \rightarrow 0} \frac{f(\w + \delta\v) -f(\w)}{\delta}. 
\end{equation*}
This suggests that we can approximate the directional derivative and the JVP
using
\begin{equation*}
\partial f(\w)[\v] 
\approx
\frac{f(\w + \delta\v) - f(\w)}{\delta}, 
\end{equation*}
for some $0 < \delta \ll 1$. 
This formula is called a \textbf{forward difference}.

To measure the error incurred by this approximation,
we use Landau's $O$ notation.
\begin{boxrem}{Big $O$ notation}
  For two functions $g$ and $f$, we write
  \[
    g(\delta) = O(f(\delta)) \ \text{as} \ \delta \to \gamma,
  \]
  if the ratio $|g(\delta)|/|f(\delta)|$ is bounded as $\delta \to \gamma$.
  For example, we write $\delta \cos(\delta) = O(\delta)$ as $\delta \to 0$.
  In the following, unless otherwise specified,
  we compare limits as $\delta \to 0$.
We observe that $g(\delta) = O(\delta^k)$ implies $g(\delta) = o(\delta^{k-1})$.
However, Landau's $O$ notation provides finer practical interpretations than
Landau's $o$ notation. For example, if $g(\delta) = O(\delta^2)$, then
we know its upper bound scales quadratically, meaning that reducing $\delta$ by
a factor of 2 reduces this bound by a factor of 4. Knowing only that 
$g(\delta) = o(\delta)$ does not allow us to make such a concrete statement, 
because it merely dictates that $g(\delta)$ decays strictly faster than $\delta$,
without providing a specific bounding rate or scaling factor for that decay.
For numerical analysis, Landau's $O$ notation is therefore preferable.
\end{boxrem}

From the Taylor expansion in \cref{diff:sec:taylor_exp},
we indeed have
\begin{equation*}
f(\w+\delta\v) - f(\w)
=
\delta \partial f(\w)[\v] 
+ \frac{\delta^2}{2} \partial^2 f(\w)[\v, \v]
+ \frac{\delta^3}{3!} \partial^3 f(\w)[\v, \v, \v]
+ \dots
\end{equation*}
so that
\begin{align*}
\frac{f(\w+\delta\v) - f(\w)}{\delta}
&=
\partial f(\w)[\v] 
+ \frac{\delta}{2} \partial^2 f(\w)[\v, \v]
+ \frac{\delta^2}{3!} \partial^3 f(\w)[\v, \v, \v]
+ \dots \\
&=
\partial f(\w)[\v] + O(\delta).
\end{align*}
The error incurred by choosing a finite rather than infinitesimal $\delta$ in
the forward difference formula is called the \textbf{truncation error}. The
Taylor approximation above shows that this error is of the order of
$O(\delta)$.

\begin{figure}
	\begin{center}
		\includegraphics[width=0.6\linewidth]{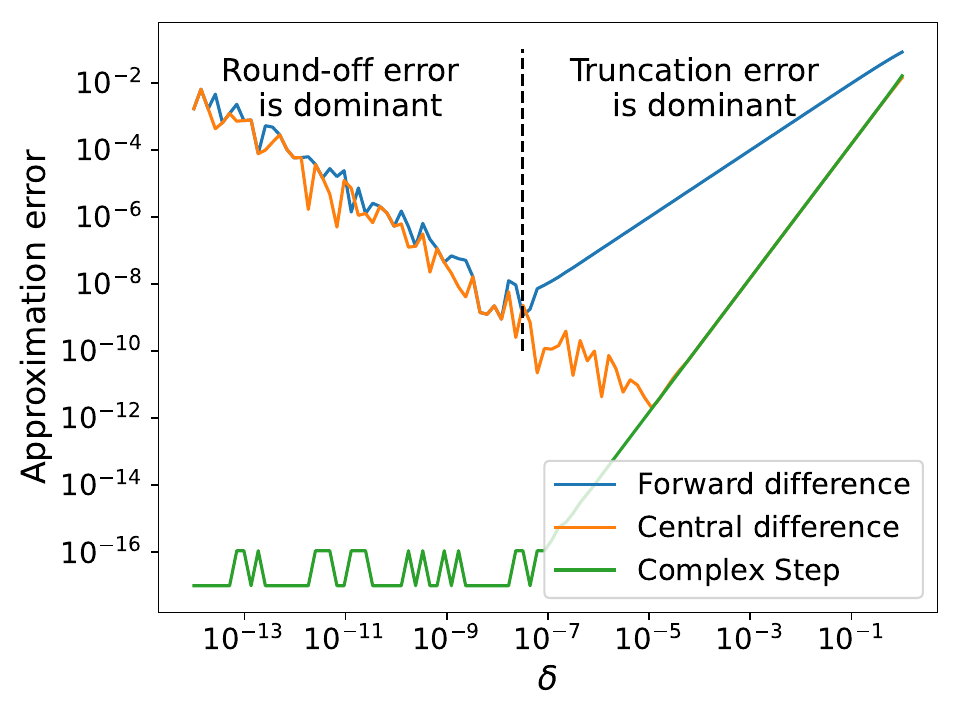}
            \caption{Numerical differentiation of
                $f(x) \coloneqq \mathrm{softplus}(x) = \log(1+\exp(x))$,
                to approximate $f'(x) = \mathrm{logistic}(x)$ at $x=1$.
The forward and central difference methods
        induce both truncation error (for large $\delta$) and round-off error
    (for small $\delta$). 
    The complex step method enjoys smaller round-off error.
\label{num_diff:fig:approx_error}}
	\end{center}
\end{figure}

However, we cannot choose a too small value of $\delta$, because the evaluation
of the function $f$ on a computer rounds the value of $f$ to machine
precision. Mathematically, a scalar-valued function $f$ 
evaluated on a computer becomes a function $\tilde{f}$ such that
$\tilde{f}(\w) \approx [f(\w)/\varepsilon] \varepsilon$, 
where $[f(\w)/\varepsilon]$
denotes the closest integer of $f(\w)/\varepsilon \in \R$ and $\varepsilon$ is
the machine precision, i.e., the smallest non-zero real number encoded by the
machine. This means that the difference $f(\w+\delta\v) - f(\w)$ evaluated on
a computer is prone to \textbf{round-off error} of the order of
$O(\varepsilon)$. 
We illustrate the trade-off between truncation and round-off errors in
\cref{num_diff:fig:approx_error}.

\section{Backward differences}

As an alternative, we can approximate the directional derivative and the JVP by
\begin{equation*}
\partial f(\w)[\v] 
\approx
\frac{f(\w) - f(\w - \delta\v)}{\delta}, 
\end{equation*}
for some $0 < \delta \ll 1$. This formula is called a \textbf{backward
difference}. From the Taylor expansion in \cref{diff:sec:taylor_exp}, we
easily verify that $(f(\w) - f(\w-\delta\v))/\delta = \partial
f(\w)[\v] + O(\delta)$, so that the truncation error is the same as for the
forward difference.

\section{Central differences}

Rather than using an asymmetric formula to approximate the derivative,
as in forward and backward differences,
we can use a symmetric formula
\begin{equation*}
\partial f(\w)[\v] 
\approx
\frac{f(\w + \delta\v) - f(\w-\delta\v)}{2\delta}, 
\end{equation*}
for some $0 < \delta \ll 1$. 
This formula is called a \textbf{central difference}.
From the Taylor expansion in \cref{diff:sec:taylor_exp},
we have
\begin{align*}
f(\w+\delta\v) - f(\w-\delta\v)
=& 
2\delta \partial f(\w)[\v] 
+ \frac{2\delta^3}{3!} \partial^3 f(\w)[\v, \v, \v] \\
 &+ \frac{2\delta^5}{5!} \partial^5 f(\w)[\v, \dots, \v] + \dots
\end{align*}
so that
\begin{align*}
\frac{f(\w+\delta\v) - f(\w-\delta\v)}{2\delta}
=& 
\partial f(\w)[\v] 
+ \frac{\delta^2}{3!} \partial^3 f(\w)[\v, \v, \v] + \ldots\\
=& \partial f(\w)[\v] + O(\delta^2).
\end{align*}
We see that the terms corresponding to derivatives of \textbf{even order}
canceled out, allowing the formula to achieve $O(\delta^2)$ truncation error.
For any $\delta<1$, the truncation error of the central difference is much
smaller than the one of the forward or backward differences as confirmed
empirically in \cref{num_diff:fig:approx_error}. 

\section{Higher-accuracy finite differences}

The truncation error can be further reduced by making use of additional function
evaluations. One can generalize the forward difference scheme by a formula of
the form 
\[
  \partial f(\w)[\v] \approx \sum_{t=0}^p \frac{a_t}{\delta} f(\w + t\delta \v)
\]
requiring $p+1$ evaluations. To select the $a_t$ and reach a truncation error of
order $O(\delta^p)$, we can use a Taylor expansion on each term of the sum
to get 
\begin{align*}
  \sum_{t=0}^p \frac{a_t}{\delta} f(\w + t\delta \v) 
  & = \sum_{t=0}^p a_t \sum_{j=0}^p 
  \frac{t^j\delta^{j-1}}{j!} \partial^j f(\w)[\v, \ldots, \v] + O(\delta^p).
\end{align*}
By grouping the terms in the sum for each order of derivative, we obtain a set
of $p+1$ equations to be satisfied by the $p+1$ coefficients $a_0, \ldots, a_p$,
that is, 
\begin{align*}
  a_0 + a_1 + \ldots + a_p & = 0 \\
  a_1 + 2 a_2 + \ldots + p a_p & = 1 \\
  a_1 + 2^j a_2 + \ldots + p^j a_p & = 0 \quad \forall j \in \{2, \ldots, p\}.
\end{align*}
This system of equations can be solved analytically to derive the coefficients. 
Backward differences can be generalized similarly by using
$\partial f(\w)[\v] \approx \sum_{t=0}^p \frac{a_t}{\delta} f(\w - t\delta
\v)$.
Similarly, the central difference scheme can be generalized by using 
\[
  \partial f(\w)[\v] \approx \sum_{t=-p}^p \frac{a_t}{\delta} f(\w + t\delta \v),
\]
to reach a truncation error of order $O(\delta^{2p})$. Solving for the coefficients
$a_{-p}, \ldots, a_p$ as above reveals that $a_0 = 0$. Therefore, only $2p$
evaluations are necessary.

\section{Higher-order finite differences}

To approximate higher order derivatives, we can follow a similar reasoning.
Namely, we can generalize the forward difference scheme to
approximate the derivative of order $k$ by 
\[
  \partial^k f(\w)[\v, \ldots, \v] 
  \approx \sum_{t=0}^p \frac{a_t}{\delta^k}f(\w + t \delta \v).
\]
As before, we can expand the terms in the sum. For the approximation to capture
only the $k$\textsuperscript{th} derivative, we now require the coefficients
$a_t$ to satisfy
\begin{align*}
  0^j a_0 + 1^j a_1 + 2^j a_2 + \ldots + p^j a_p & = 0 \quad \forall j \in \{0, \ldots, k-1\}. \\
  0^k a_0 + 1^k a_1 + 2^k a_2 + \ldots + p^k a_p & = k! \\
  0^j a_0 + 1^j a_1 + 2^j a_2 + \ldots + p^j a_p & = 0 \quad \forall j \in \{k+1, \ldots, p\}.
\end{align*}
With the resulting coefficients, we obtain a truncation error of order
$O(\delta^{p - k +1})$, while making $p+1$ evaluations. For example, for
$p=k=2$, we can approximate the second-order derivative as
\[
  \partial^2 f(\w)[\v, \v] 
  \approx \frac{-(3/2)f(\w) +2 f(\w+ \delta \v) - (1/2) f(\w + 2\delta \v)}{\delta^2},
\]
with a truncation error of order $O(\delta)$.

The central difference scheme can be generalized similarly by
\[
  \partial^k f(\w)[\v, \ldots, \v]
  \approx \sum_{t=-p}^p \frac{a_t}{\delta^k} f(\w + t\delta \v),
\]
to reach truncation errors of order $O(\delta^{2p +2 - 2\lceil (k+1)/2\rceil})$.
For example, for $k=2$, $p=1$, we obtain the second-order central difference
\begin{equation*}
  \partial^2 f(\w)[\v, \v] 
  \approx \frac{f(\w+\delta\v) + f(\w-\delta\v) - 2f(\w)}{\delta^2}.
\end{equation*}
By using a Taylor expansion we see that, this time, the terms corresponding to
derivatives of \textbf{odd order} cancel out and the truncation error is
$O(\delta^2)$ while requiring $3$ evaluations.

\section{Complex-step derivatives}

Suppose $f$ is well defined on $\CC^P$, 
the space of $P$-dimensional complex numbers.
Let us denote the imaginary unit by $i = \sqrt{-1}$.
Then, the Taylor expansion of $f$ reads
\begin{align*}
f(\w+(i \delta)\v) 
=&
f(\w)
+
(i\delta) \partial f(\w)[\v] 
+ \frac{(i\delta)^2}{2} \partial^2 f(\w)[\v, \v] \\
 &+ \frac{(i\delta)^3}{3!} \partial^3 f(\w)[\v, \v, \v]
+ \dots \\
=&
f(\w)
+
(i\delta) \partial f(\w)[\v] 
- \frac{\delta^2}{2} \partial^2 f(\w)[\v, \v] \\
 &- \frac{i\delta^3}{3!} \partial^3 f(\w)[\v, \v, \v]
+ \dots \quad .
\end{align*}
We see that the real part corresponds to even-degree terms and the
imaginary part corresponds to odd-degree terms.
We therefore obtain
\begin{equation*}
\operatorname{Re}(f(\w+(i \delta)\v))
= f(\w) + O(\delta^2)
\end{equation*}
and
\begin{equation*}
\operatorname{Im}\left(\frac{f(\w+(i \delta)\v)}{\delta}\right)
= \partial f(\w)[\v] + O(\delta^2).
\end{equation*}
This suggests that we can compute directional derivatives using
the approximation
\begin{equation*}
\partial f(\w)[\v]
\approx
\operatorname{Im}\left(\frac{f(\w+(i \delta)\v)}{\delta}\right),
\end{equation*}
for $0 < \delta \ll 1$.
This is called the \textbf{complex-step derivative} approximation
\citep{squire_1998,martins_2003}.

Contrary to forward, backward and central differences, we see that only a
\textbf{single function call} is necessary. A function call on complex numbers
may take roughly twice the cost of a function call on real numbers. However,
thanks to the fact that a difference of functions is no longer needed, the
complex-step derivative approximation usually enjoys smaller round-off error as
illustrated in \cref{num_diff:fig:approx_error}. That said, one drawback of
the method is that all elementary operations within the program implementing the
function $f$ must be well-defined on complex numbers, e.g., using overloading.

\section{Complexity}

We now discuss the computational complexity in terms of function evaluations of
finite differences and complex-step derivatives.
For concreteness, as this is the most common use case in machine learning, 
we discuss the case of a single $M=1$ output, i.e.,
we want to differentiate a function $f \colon \RR^P \to \RR$.
Whether we use forward, backward or central differences, 
the computational complexity of computing the directional derivative 
$\partial f(\w)[\v]$ in any direction $\v$ amounts to two calls to $f$.
For computing the gradient $\nabla f(\w)$, we can use (see
\cref{diff:def:gradient}) that
\begin{equation*}
[\nabla f(\w)]_j 
= \langle \nabla f(\w), \e_j \rangle = \partial f(\w)[\e_j],
\end{equation*}
for $j \in [P]$.
For forward and backward differences, 
we therefore need $P+1$ function calls to
compute the gradient,
while we need $2P$ function calls for central differences.
For the complex step approximation, we need $P$ complex function calls.
We summarize the complexities in \cref{finite_diff:tab:computational_cost}.

\begin{table}[t]
\caption{Computational complexity in number of function evaluations for
computing the directional derivative and the gradient of a function $f \colon
\RR^P \to \RR$ by finite differences and complex step derivatives.}
\begin{center}
\begin{tabular}{lcc}
\toprule
& Directional derivative & Gradient \\
\midrule
Forward difference & $2$ & $P+1$ \\
Backward difference & $2$ & $P+1$ \\
Central difference & $2$ & $2P$ \\
Complex step & $1$ & $P$ \\
\bottomrule
\end{tabular}
\end{center}
\label{finite_diff:tab:computational_cost}
\end{table}

\section{Summary}

\begin{itemize}

\item Finite differences are a simple way to numerically compute
derivatives using only function evaluations.

\item Central differences achieve smaller truncation error
than forward and backward differences.
It is possible to achieve smaller truncation error,
at the cost of more function evaluations.

\item Complex-step derivatives achieve smaller round-off error 
than central differences but require the function and the program
implementing it to be well-defined on complex numbers.

\item However, whatever the method used,
finite differences require a number of function calls that is
proportional to the number of dimensions. They are therefore seldom used in
machine learning, where there can be millions or billions of dimensions.
The main use cases of finite differences are therefore 
i) for blackbox functions of low dimension 
and 
ii) for test purposes (e.g., checking that a gradient function is correctly
implemented).

\item For modern machine learning, the main workhorse is automatic differentiation, as
it leverages the compositional structure of functions.
This is what we study in the next chapter.

\end{itemize}

%% file: chapters/auto_diff/auto_diff.tex
\chapter{Automatic differentiation}\label{chap:auto_diff}

In \cref{chap:diff}, we reviewed the fundamentals of differentiation and
stressed the importance of two linear maps: the Jacobian-vector product (JVP)
and its adjoint, the vector-Jacobian product (VJP).
In this chapter, we review \textbf{forward-mode} and \textbf{reverse-mode}
autodiff using these two linear maps. We start with \textbf{computation chains}
and then generalize to feedforward networks and general \textbf{computation
graphs}. We also review checkpointing, reversible layers and randomized
estimators.

\section{Computation chains}\label{auto_diff:sec:chains}

To begin with, consider a \textbf{computation chain}
(\cref{neural_nets:sec:comput_chains}) representing a function $f \colon \cS_0
\to \cS_K$ expressed as a sequence of compositions $f \coloneqq f_K \circ \dots
\circ f_1$, where $f_k \colon \cS_{k-1} \to \cS_k$. The computation of $f$ can
be unrolled into a sequence of operations 
\begin{align}
	\s_0 &\in \cS_0\nonumber \\
	\s_1 &\coloneqq f_1(\s_0) \in \cS_1 \nonumber\\ 
		 & \hspace{6pt} \vdots \nonumber\\
	\s_K &\coloneqq f_K(\s_{K-1}) \in \cS_K \nonumber\\
	f(\s_0) &\coloneqq \s_K. \label{auto_diff:eq:chain}
\end{align}
Our goal is to compute the variations of $f$
around a given input $\s_0$. In a feedforward network, this amounts
to estimating the influence of a given input $\s_0$ for fixed parameters (we
will see how to estimate the variations w.r.t. parameters $\w$ in the sequel). 

\paragraph{Jacobian matrix}

We first consider the computation of the full Jacobian $\jac f(\s_0)$, 
seen as a \textbf{matrix}, as the notation $\bm{\partial}$ indicates.
Following \cref{diff:thm:chain_rule}, we have
\begin{equation}
    \jac f(\s_0) = 
    \jac f_K(\s_{K-1}) 
    \ldots 
    \jac f_1(\s_0),
    \label{auto_diff:eq:chain_full_jac}
\end{equation}
where $\jac f_k(\s_{k-1})$ are the Jacobians of the
intermediate functions computed at $\s_0, \ldots, \s_K$, as defined
in~\cref{auto_diff:eq:chain}. 
We may also want to compute the transpose of the Jacobian
\begin{equation*}
    \jac f(\s_0)^\top = 
    \jac f_1(\s_0)^\top
    \ldots 
    \jac f_K(\s_{K-1})^\top.
\end{equation*}
In both cases, the main drawback of this approach is
computational: computing the full $\jac f(\s_0)$ requires materializing the
intermediate Jacobians in memory and performing matrix-matrix
multiplications. However, in practice, computing the full Jacobian is rarely
needed. Indeed, oftentimes, we only need to right-multiply or
left-multiply with $\jac f(\s_0)$. This gives rise to forward-mode and
reverse-mode autodiff, respectively.

\subsection{Forward-mode}

\begin{figure}[t]
	\begin{center}
		\includegraphics[width=\linewidth]{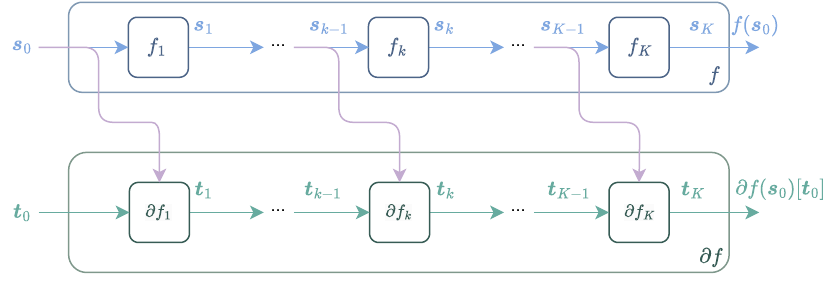}
		\caption{Forward-mode autodiff for a computation chain. For readability,
        we denoted the intermediate JVP as a function of two variables
        $\partial f_k: 
        \s_{k-1}, \t_{k-1} \mapsto 
        \partial f_k(\s_{k-1})[\t_{k-1}]$ 
        with $\partial f_k(\s_{k-1})[\t_{k-1}]= \t_k$.
        \label{auto_diff:fig:chain_jvp}}
	\end{center}
\end{figure}

We now interpret the Jacobian $\partial f(\s_0)$ as a \textbf{linear map}, as the
non-bold $\partial$ indicates.
Following \cref{diff:thm:chain_rule_jvp_vjp}, $\partial f(\s_0)$ is the
composition of the intermediate linear maps,
\[
\partial f(\s_0) 
= 
\partial f_K(\s_{K-1}) 
\circ 
\ldots 
\circ 
\partial f_1(\s_0).
\]
Evaluating $\partial f(\s_0)$ on an \textbf{input direction} $\v \in \cS_0$
can be decomposed, like the function
\cref{auto_diff:eq:chain} itself, into intermediate computations
\begin{align*}
	\t_0 & \coloneqq \v \\
	\t_1 & \coloneqq \partial f_1(\s_0)[\t_0] \\
	& \hspace{6pt} \vdots\\
	\t_K & \coloneqq \partial f_K(\s_{K-1})[\t_{K-1}]\\
	\partial f(\s_0)[\v] &\coloneqq \t_K. 
\end{align*}
Each intermediate $\partial f_k(\s_{k-1})[\t_{k-1}]$ amounts to a
Jacobian-vector product (JVP) and can be performed in a \textbf{forward} manner,
alongside the computation of the intermediate states $\s_k$. 
This can also be seen as
multiplying the matrix defined in \cref{auto_diff:eq:chain_full_jac} with a
vector, from \textbf{right to left}. This is illustrated in
\cref{auto_diff:fig:chain_jvp} and the procedure is summarized in
\cref{auto_diff:algo:forward_chain}. 

\begin{algorithm}\caption{Forward-mode autodiff for computation chains
  \label{auto_diff:algo:forward_chain}}
\begin{algorithmic}[1]
\Statex{\bf Functions:} 
$f \coloneqq f_K \circ \ldots \circ f_1$
\Statex{\bf Inputs:} input $\s_0 \in \cS_0$, input direction $\v
\in \cS_0$
\State Initialize $\t_0 \coloneqq \v$
\For {$k \coloneqq 1, \ldots, K$} 
\State Compute $\s_k \coloneqq f_k(\s_{k-1}) \in \cS_k$ \label{auto_diff:line:forward_computations}
\State Compute $\t_k \coloneqq \partial f_k(\s_{k-1})[\t_{k-1}] \in \cS_k$
\EndFor 
\Statex {\bf Outputs:} 
$f(\s_0) \coloneqq \s_K$, $\partial f(\s_0)[\v] = \t_K$
\end{algorithmic}
\end{algorithm}

\paragraph{Computational complexity}

The JVP follows exactly the computations of $f$, with an additional variable
$\t_k$ being propagated. If we consider that computing $\partial f_k$ is
roughly as
costly as computing $f_k$, then computing a JVP has roughly twice the
computational cost of $f$.  See \cref{auto_diff:sec:complexity_dag} for a more
general and more formal
statement.

\paragraph{Memory usage}

The memory usage of a program at a given evaluation step is the number of
variables that need to be stored in memory to ensure the execution of all
remaining steps. The memory cost of a program is then the maximal memory usage
over all evaluation steps. For our purposes, we analyze the memory usage and
memory cost by examining the given program. Formal definitions of operations on
memory such as read, write, delete and associated memory costs are presented by
\citet[Chapter 4]{griewank2008evaluating}. 

For example, to execute the chain $f = f_K \circ \dots \circ f_1$, 
at each step $k$, we only
need to have access to $\s_{k-1}$ to execute the rest of the program. As we
compute $\s_k$, we can delete $\s_{k-1}$ from memory and replace it by $\s_k$.
Therefore, the memory cost associated with the evaluation of $f$ is just the
maximal dimension of the $\s_k$ variables.

\begin{figure}[h]
	\begin{center}
		\includegraphics[width=0.5\linewidth]{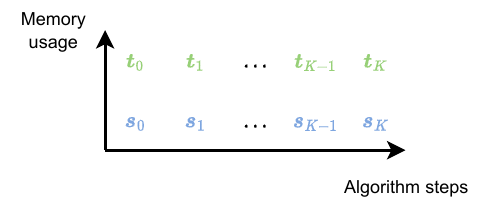}
		\caption{Memory usage of forward-mode autodiff for a computation chain. 
        Here $\t_0 = \v$, $\s_K = f(\s_0)$, $\t_K = \partial f(\s_0)[\v]$.
        \label{auto_diff:fig:chain_jvp_memory}}
	\end{center}
\end{figure}

For forward mode autodiff, as we follow the computations of $f$, at each step
$k$, we only need to have access to $\s_{k-1}$ and $\t_{k-1}$ to execute the
rest of the program. The memory used by $\s_{k-1}$ and $\t_{k-1}$ can directly
be used for $\s_k, \t_k$ once they are computed. The memory usage associated
with
the JVP is summarized in~\cref{auto_diff:fig:chain_jvp_memory}. Overall
the memory cost of the JVP is then exactly twice the memory cost of the function
itself.

\subsection{Reverse-mode}

\begin{figure}
	\begin{center}
		\includegraphics[width=\linewidth]{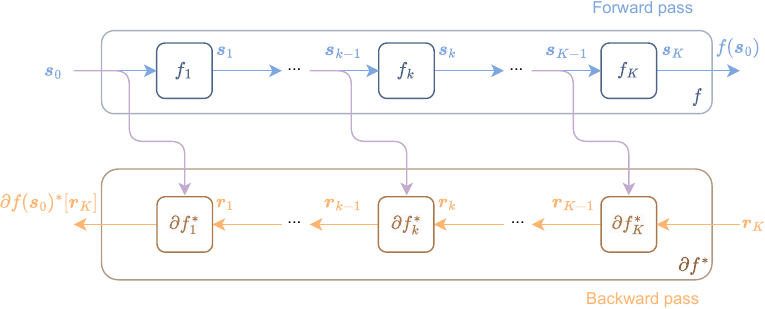}
		\caption{Reverse mode of automatic differentiation for a computation
            chain.
            For readability, we denoted the intermediate VJPs as 
            functions of two variables 
            $\partial f_k^* : 
            (\s_{k-1}, \r_k) \mapsto
            \partial f_k(\s_{k-1})^*[\r_k]$,
            with $\partial f_k(\s_{k-1})^*[\r_k]= \r_{k-1}$.
            \label{auto_diff:fig:chain_vjp} }
	\end{center}
\end{figure}

In machine learning, most functions whose gradient we need to compute take the
form $\ell \circ f$, where $\ell$ is a scalar-valued loss function and $f$ is
a network. As seen in \cref{diff:thm:chain_rule_scalar}, the gradient takes the
form
\begin{equation*}
\nabla (\ell \circ f)(\s_0) = \partial f(\s_0)^*[\nabla \ell(f(\s_0))].
\end{equation*}
This motivates the need for applying the \textbf{adjoint} 
$\partial f(\s_0)^*$ to 
$\nabla \ell(f(\s_0)) \in \cS_K$ and more generally to any
\textbf{output direction} $\u \in \cS_K$.
From \cref{diff:thm:chain_rule_grad_vjp}, we have
\[
\partial f(\s_0)^* 
= 
\partial f_1(\s_0)^* 
\circ 
\ldots 
\circ 
\partial f_K(\s_{K-1})^*.
\]
Evaluating 
$\partial f(\s_0)^*$
on an output direction $\u \in \cS_K$ is decomposed as 
\begin{align*}
	\r_K & \coloneqq \u \\
	\r_{K-1} &\coloneqq \partial f_K(\s_{K-1})^*[\r_K] \\
	& \hspace{6pt} \vdots \\
	\r_0 &\coloneqq \partial f_1(\s_0)^*[\r_1] \\
	\partial f(\s_0)^*[\u] &\coloneqq \r_0.
\end{align*}
Each intermediate adjoint $\partial f_k(\s_{k-1})^*$ amounts to a
vector-Jacobian product (VJP). The key difference with the forward mode is that
the procedure runs \textbf{backward} through the chain, hence the name
\textbf{reverse mode} autodiff. This can also be seen as multiplying
\cref{auto_diff:eq:chain_full_jac} from \textbf{left to right}. The procedure is
illustrated in \cref{auto_diff:fig:chain_vjp} and summarized in
\cref{auto_diff:algo:reverse_chain}. 

\begin{algorithm}\caption{Reverse-mode autodiff for computation chains
\label{auto_diff:algo:reverse_chain}}
\begin{algorithmic}[1]
    \Statex{\bf Functions:} $f \coloneqq f_K \circ \ldots \circ f_1$, 
	\Statex{\bf Inputs:} input $\s_0 \in \cS_0$, output direction $\u \in \cS_K$
        \For {$k\coloneqq 1, \ldots, K$} \Comment{Forward pass} \label{auto_diff:line:reverse_mode_forward_start}
		    \State Compute $\s_k \coloneqq f_k(\s_{k-1}) \in \cS_k$
		\EndFor \label{auto_diff:line:reverse_mode_forward_end}
	\State Initialize $\r_K \coloneqq \u$. 
    \For{$k\coloneqq K, \ldots, 1$} \Comment{Backward pass} \label{auto_diff:line:reverse_mode_backward_start}
		\State Compute $\r_{k-1} \coloneqq \partial f_k(\s_{k-1})^*[\r_k] \in
        \cS_{k-1}$
	\EndFor \label{auto_diff:line:reverse_mode_backward_end}
    \Statex{\bf Outputs:} $f(\s_0) \coloneqq \s_K$, 
    $\partial f(\s_0)^*[\u] = \r_0$
\end{algorithmic}
\end{algorithm}

\paragraph{Computational complexity}

In terms of number of operations, the VJP simply passes two times through the
chain, once forward, then backward. If we consider the intermediate VJPs to be
roughly as costly as the intermediate functions themselves, 
the VJP amounts to just twice the cost of the
original function, just as the JVP. See \cref{auto_diff:sec:complexity_dag} for a
more generic and formal statement.

\paragraph{Memory usage}

\begin{figure}
  \begin{center}
    \includegraphics[width=\linewidth]{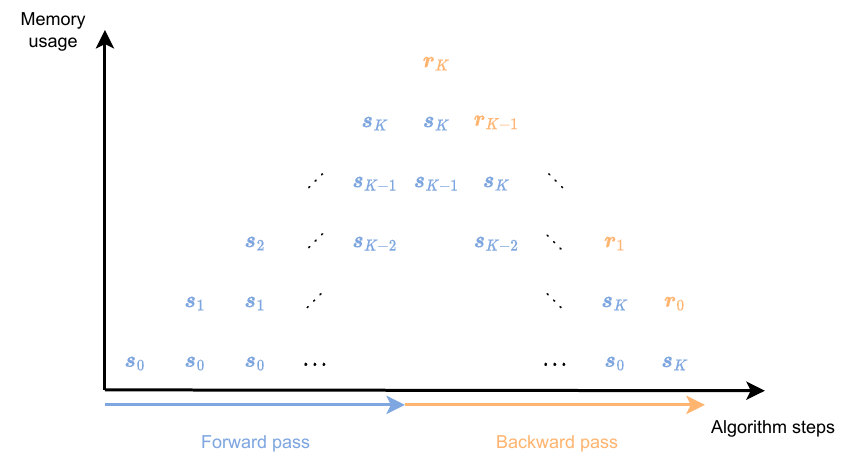}
  \end{center}
  \caption{Memory usage of reverse mode autodiff for a 
  computation chain.\label{auto_diff:fig:chain_vjp_memory}}
\end{figure}

Recall that the memory usage of a program at a given evaluation step is the
number of variables that need to be stored in memory to ensure the execution of
the remaining steps. If we inspect \cref{auto_diff:algo:backprop}, 
to execute all backward steps, that is the loop in
line~\ref{auto_diff:line:reverse_mode_backward_start}, we need to have access to
all the intermediate inputs $\s_0, \ldots, \s_{K-1}$. Therefore, 
the memory cost of
reverse-mode autodiff is proportional to the length of the chain $K$.
\cref{auto_diff:fig:chain_vjp_memory} illustrates the memory usage during reverse
mode autodiff. It grows linearly until the end of the forward pass and then
progressively decreases until it outputs the value of the function and the VJP.
The memory cost can be mitigated by means of checkpointing techniques presented
in \cref{auto_diff:sec:checkpointing}.

\paragraph{Decoupled function and VJP evaluations}

The additional memory cost of reverse mode autodiff comes with some
advantages. If we need to compute $\partial f(\s_0)^*[\u_i]$ for $n$ different
output directions $\u_i$, we only need to compute and store once the
intermediate computations $\s_k$ and then make $n$ calls to the backward pass.
In other words, by storing in memory the intermediate computations $\s_k$,
we may instantiate a \textbf{VJP operator}, which we may apply to any $\u$
through the backward pass. Formally, the forward and backward passes can be
decoupled as
\begin{equation*}
\mathrm{forward}(f, \s_0)
\coloneqq (f(\s_0), \partial f(\s_0)^*)
\end{equation*}
where
\begin{equation*}
\partial f(\s_0)^*[\u] 
\coloneqq \mathrm{backward}(\u; \s_0, \ldots, \s_{K-1}).
\end{equation*}
In functional programming terminology, the VJP $\partial f(\s_0)^*$ is a
\textbf{closure}, as it contains the intermediate computations $\s_0, \dots,
\s_{K-1}$. The same can be done for the JVP
$\partial f(\s_0)$ if we want to apply it to multiple directions $\v_i$.

\begin{boxexm}{Multilayer perceptron with fixed parameters
    \label{auto_diff:exm:mlp}} 
    As a running example, consider a
    multilayer perceptron (MLP) with one hidden layer and (for now) given fixed
    weights.
	As presented in \cref{chap:neural_nets}, an MLP can be decomposed as
	\begin{align*}
		\s_0 & = \x \\
		\s_1 & = f_1(\s_0) = \sigma(\A_1 \s_0 + \b_1) \\
		\s_2 & = f_2(\s_1) = \A_2 \s_1 + \b_2 \\
		f(\x) & = \s_2,
	\end{align*}
	for $\A_1, \A_2, \b_1, \b_2$ some fixed parameters and $\sigma$ an
	activation function such as the softplus activation function $\sigma(x) =
	\log(1+e^x)$ with derivative $\sigma'(x) = e^x/(1+e^x)$. 
	
	Evaluating the JVP of $f$ on an input $\x$ along a direction $\v$
	can then be decomposed as
	\begin{align*}
		\t_0 & = \v \\
		\t_1 & = \sigma'(\A_1 \s_0 + \b_1) \odot (\A_1 \t_0) \\
		\t_2 & = \A_2 \t_1 \\
		\partial f(\x)[\v] & = \t_2,
	\end{align*}
    where we used in the second line the JVP of element-wise functions as in
    \cref{diff:exm:jvp_vjp_sep_func}.

	Evaluating the VJP of $f$ at $\x$ requires evaluating the intermediate
    VJPs at the stored activations
	\begin{align*}
		\r_2 & = \u \\
		\r_1 & = \partial f_2(\s_1)^*[\r_2] 
			   = \A_2^\top \r_2 \\
		\r_0 & = \partial f_1(\s_0)^*[\r_1] 
		       = \A_1^\top (\sigma'(\A_1 \s_0 + \b_1) \odot \r_1) \\
		\partial f(\x)^*[\u] & = \r_0.
	\end{align*}
\end{boxexm}

\subsection{Complexity of computing entire Jacobians}

In this section, we analyze the time and space complexities of forward-mode and
reverse-mode autodiff for computing the \textbf{entire}
Jacobian matrix $\jac f(\s_0)$
of a computation chain $f = f_K \circ \dots \circ f_1$,
where $f_k \colon \cS_{k-1} \to \cS_k$.  
We assume $\cS_k \subseteq \RR^{D_k}$, $D_K = M$ and $D_0 = D$.
Therefore, we have $f \colon \RR^D \to \RR^M$
and $\jac f(\s_0) \in \RR^{M \times D}$.

\subsubsection*{Complexity of forward-mode autodiff}

Using \cref{diff:def:jac}, we find that we can extract each \textbf{column}
$[\jac f(\s_0)]_{:,j} \in \RR^M$ of the Jacobian matrix, for $j \in [D]$, 
by multiplying with the standard basis vector $\e_j \in \RR^D$:
\begin{align*}
    [\jac f(\s_0)]_{:, 1} &= \partial f(\s_0)[\e_1] \\
                     &\vdots \\
    [\jac f(\s_0)]_{:, D} &= \partial f(\s_0)[\e_D].
\end{align*}
Computing the full Jacobian matrix therefore requires $D$ JVPs with vectors in
$\RR^D$.
Assuming each $f_k$ in the chain composition has the form 
$f_k: \R^{D_{k-1}} \rightarrow \R^{D_k}$, 
seen as a matrix, $\jac f_k(\s_{k-1})$ has size $D_{k} \times
D_{k-1}$.  Therefore, the computational cost of $D$ JVPs is
$O\left(D \sum_{k=1}^K D_k D_{k-1}\right)$.
The memory cost is $O(\max_{k \in [K]} D_k)$, since we can release intermediate
computations after each layer is processed.
Setting $D_1 = \dots = D_{K-1} = D$ for simplicity and using $D_K=M$, we obtain
that the computational cost of computing $D$ JVPs and therefore of computing the
full Jacobian matrix by forward-mode autodiff is $O(MD^2 + KD^3)$.
The memory cost is $O(\max\{D, M\})$. If a function has a single input ($D=1$),
then the forward mode computes the entire Jacobian at once, which reduces to a
single directional derivative. 

\subsubsection*{Complexity of reverse-mode autodiff}

Using \cref{diff:def:jac}, we find that we can extract each \textbf{row} of the
Jacobian matrix $[\jac f(\s_0)]_{i} \in \RR^D$,
for $i \in [M]$, by multiplying with the standard basis vector $\e_i \in \RR^M$:
\begin{align*}
[\jac f(\s_0)]_{1} &= \partial f(\s_0)^*[\e_1] \\
               &\vdots \\
[\jac f(\s_0)]_{M} &= \partial f(\s_0)^*[\e_M].
\end{align*}
Computing the full Jacobian matrix therefore requires $M$ VJPs with vectors in
$\RR^M$. Assuming as before that each $f_k$ in the chain composition has the
form $f_k: \R^{D_{k-1}} \rightarrow \R^{D_k}$, the
computational cost of $M$ VJPs is $O\left(M \sum_{k=1}^K D_k D_{k-1}\right)$.
However, the memory cost is $O(\sum_{k=1}^{K} D_k)$, as we need to store the
intermediate computations for each of the $K$ layers. Setting $D_0 = \dots =
D_{K-1} = D$ for simplicity and using $D_K=M$, we obtain that the computational
cost of computing $M$ VJPs and therefore of computing the full Jacobian matrix
by reverse-mode autodiff is $O(M^2D + KMD^2)$. The memory cost is $O(KD + M)$.
If the function has a single output ($M=1$), reverse-mode autodiff computes the
entire Jacobian at once, which reduces to the gradient.

\subsubsection*{When to use forward-mode vs.\ reverse-mode autodiff?}

We summarize the time and space complexities in
\cref{autodiff:tab:chain_complexity}. Generally, if $M<D$, reverse-mode is more
advantageous at the price of some memory cost. If $M\geq D$, forward
mode is more advantageous.

\begin{table}[t]
  \begin{center}
  \begin{tabular}{c|c|c}
    & Forward-mode & Reverse-mode \\
    \hline
    Time & $O(MD^2 + KD^3)$ & $O(M^2D + KMD^2)$ \\
    Space & $O(\max\{M,D\})$ & $O(KD + M)$ \\
  \end{tabular}
  \caption{
    Time and space complexities of forward-mode and reverse-mode autodiff
    for computing the full Jacobian
    of a chain of functions $f = f_K \circ \dots \circ f_1$, 
    where $f_k \colon \RR^D \to \RR^D$ 
    if $k = 1, \dots, K-1$ and $f_K \colon \RR^D \to \RR^M$.
    We assume $\partial f_k$ is a dense linear operator.
    Forward mode requires $D$ JVPs. Reverse mode requires $M$ VJPs.
    \label{autodiff:tab:chain_complexity}
  }
  \end{center}
\end{table}

\section{Feedforward networks}

\begin{figure}
    \begin{center}
        \includegraphics[width=\linewidth]{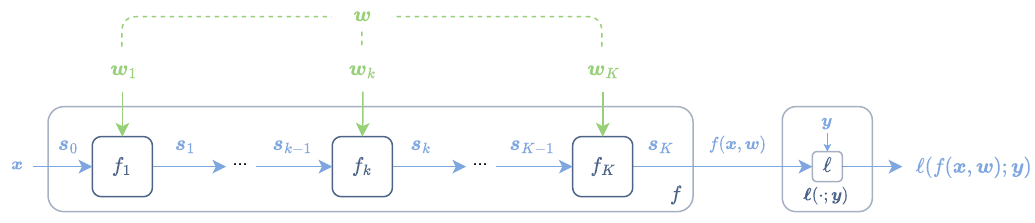}
    \end{center}
    \caption{Computation graph of an MLP as a function of its
    parameters.\label{auto_diff:fig:mlp_graph}}
\end{figure}

In the previous section, we derived forward-mode autodiff and reverse-mode
autodiff for computation chains with an input $\s_0 \in \cS_0$. 
In this section,
we now derive reverse-mode autodiff for feedforward networks, in which each
layer $f_k$ is now allowed to depend explicitly on some additional parameters
$\w_k \in \cW_k$. The recursion is
\begin{align*}
	\s_0 & \coloneqq \x \in \cS_0 \\
	\s_1 & \coloneqq f_1(\s_0, \w_1) \in \cS_1 \\ 
		 & \hspace{6pt} \vdots \\
	\s_K & \coloneqq f_K(\s_{K-1}, \w_K) \in \cS_K \\
	f(\x, \w) & \coloneqq \s_K,
\end{align*}
where $\cS_0 = \cX$ and $\w = (\w_1, \dots, \w_K) \in \cW_1 \times \dots
\times \cW_K$. Each $f_k$ is now a function of two arguments. The first
argument depends on the previous layer, but the second argument does not. This
is illustrated in \cref{auto_diff:fig:mlp_graph}.
We now explain how to differentiate a feedforward network.

\subsection{Computing the adjoint}

The function has the form $f \colon \cE \to \cF$, 
where
$\cE \coloneqq \cX \times (\cW_1 \times \dots \times \cW_K)$
and
$\cF \coloneqq \cS_K$.
From \cref{diff:sec:jvp_vjp},
we know that the VJP has the form
$\partial f(\x, \w)^* \colon \cF \to \cE$.
Therefore, we want to be able to compute
$\partial f(\x, \w)^*[\u] \in \cE$ for any $\u \in \cF$.

Fortunately, the backward recursion is only a slight modification of 
the computation chain case. Indeed, since
$f_k \colon \cE_k \to \cF_k$,
where 
$\cE_k \coloneqq \cS_{k-1} \times \cW_k$
and
$\cF_k \coloneqq \cS_k$,
the intermediate VJPs have the form
$\partial f_k(\s_{k-1}, \w_k)^* \colon \cF_k \to \cE_k$.
We therefore arrive at the recursion
\begin{align*}
    \r_K &= \u \in \cS_K \\
    (\r_{K-1}, \g_K) &= \partial f_K(\s_{K-1}, \w_K)^* [\r_K] \in \cS_{K-1}
    \times \cW_K \\
                     &\hspace{6pt} \vdots \\
    (\r_0, \g_1) &= \partial f_1(\s_0, \w_1)^* [\r_1] \in \cS_0 \times \cW_1.
\end{align*}
The final output is
\begin{equation*}
\partial f(\x, \w)^*[\u] = (\r_0, (\g_1, \dots, \g_K)).
\end{equation*}

\subsection{Computing the gradient}

We often compose a network with a loss function
\begin{equation*}
L(\w; \x, \y) \coloneqq \ell(f(\x, \w); \y) \in \RR.
\end{equation*}
From \cref{diff:thm:chain_rule_grad_vjp},
the gradient is given by
\begin{equation*}
\nabla L(\w; \x, \y) 
= (\g_1, \dots, \g_K) \in \cW_1 \times \dots \times \cW_K
\end{equation*}
where
\begin{equation*}
\partial f(\x, \w)^*[\u] = (\r_0, (\g_1, \dots, \g_K)),
\end{equation*}
with $\u = \nabla \ell(f(\x, \w); \y) \in \cS_K$.
The output $\r_0 \in \cS_0$, where $\cS_0 = \cX$,
corresponds to the gradient w.r.t. $\x \in \cX$ and is
typically not needed, except in generative modeling settings. 
The full procedure is summarized in
\cref{auto_diff:algo:backprop}.

\begin{figure}
    \begin{center}
        \includegraphics[width=\linewidth]{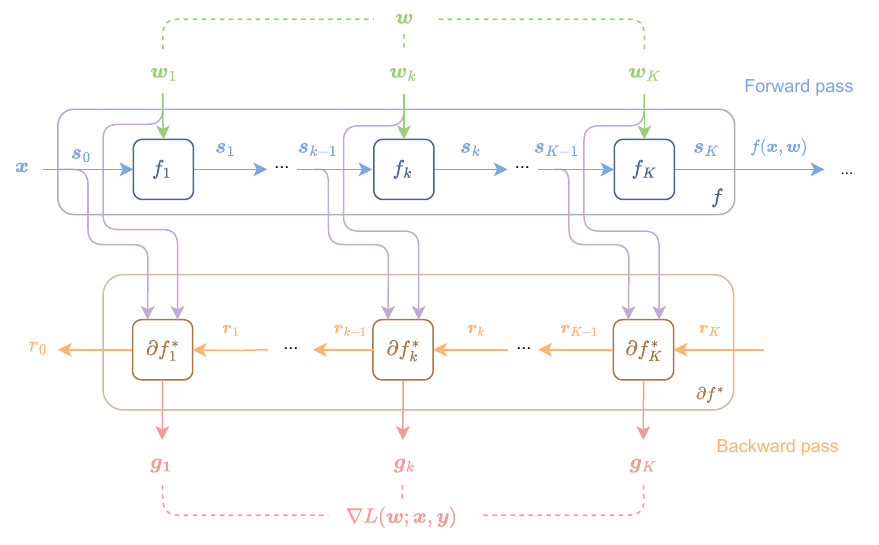}
    \end{center}
    \caption{Reverse mode of automatic differentiation, a.k.a., gradient
    back-propagation to compute the gradient of the loss of an MLP on an input
    label pair. For readability, we denoted the intermediate VJPs as 
    functions of three variables
    $\partial f_k^*: 
    (\s_{k-1}, \w_{k-1}, \r_k) \mapsto 
    \partial f_k(\s_{k-1}, \w_k)^*[\r_k]$
    with $\partial f_k(\s_{k-1}, \w_k)^*[\r_k] = (\r_{k-1}, \g_k)$.
    \label{auto_diff:fig:mlp_vjp}}
\end{figure}

\begin{algorithm}\caption{Gradient back-propagation for feedforward networks 
    \label{auto_diff:algo:backprop}}
    \begin{algorithmic}[1]
        \Statex {\bf Functions:} $f_1, \dots, f_K$ in sequential order 
        \Statex {\bf Inputs:} data point $(\x,\y) \in \cX \times \cY$
        \Statex \hspace{3.8em} parameters $\w= (\w_1, \ldots, \w_K) \in \cW_1
        \times \dots \times \cW_K$
        \State Initialize $\s_0 \coloneqq \x$ \Comment{Forward pass}
        \For{$k \coloneqq 1, \ldots, K$} 
        \State Compute and store $\s_k \coloneqq f_k(\s_{k-1}, \w_k) \in \cS_k$ 
        \EndFor
        \State Compute $\ell(\s_K; \y)$ and $\u \coloneqq \nabla \ell(\s_K; \y)
        \in \cS_K$
        \State Initialize $\r_K \coloneqq \u \in \cS_K$  \Comment{Backward pass}
        \For{$k \coloneqq K, \ldots, 1$}
        \State Compute
        $(\r_{k-1}, \g_k) \coloneqq \partial f_k(\s_{k-1}, \w_k)^*[\r_k] \in
        \cS_{k-1} \times \cW_k$
        \EndFor
        \State {\bf Outputs:} 
        $L(\w; \x, \y) \coloneqq \ell(\s_K; \y)$,
        $\nabla L(\w; \x, \y) = (\g_1, \ldots, \g_K)$
    \end{algorithmic}
\end{algorithm}

\newpage
\section{Computation graphs}\label{auto_diff:sec:graphs}

In the previous sections, we reviewed autodiff for computation chains and
its extension to feedforward networks. In this section, we review its
generalization to computation graphs, 
introduced in \cref{neural_nets:sec:comput_graphs}.
Our formalism assumes, without loss of generality, that functions can take
multiple inputs but only produce a single output, as we can always group
multiple outputs as a tuple. This enables a one-to-one correspondence between
output variables $\s_k$ and functions $f_k$.

\subsection{Forward mode}

\begin{figure}[p]
  \centering
  \includegraphics[width=\linewidth]{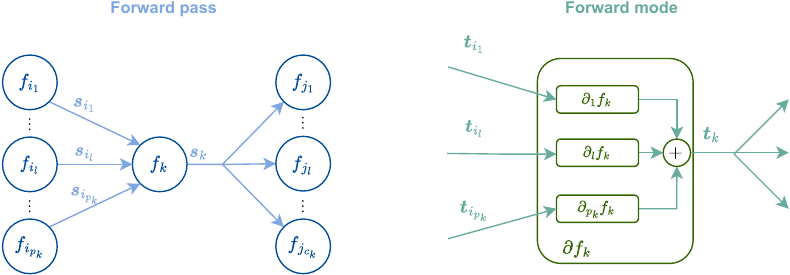}
  \caption{\label{auto_diff:fig:comp_graph_forward_mode}
  Forward mode automatic differentiation in a computation graph.
  }
\end{figure}

The forward mode corresponds to computing the JVP of the program in an input
direction $\v \in \cS_0$. 
In the case of a \textbf{computation chain} 
$f \coloneqq f_K \circ \dots \circ f_1$,
each $f_k$ takes a \textbf{single} input
$\s_{k-1} \in \cS_{k-1}$ 
and
each $\partial f_k(\s_{k-1})$ 
takes a single input direction $\t_{k-1} \in \cS_{k-1}$.
The forward pass on iteration $k \in [K]$ is then,
starting from $\t_0 \coloneqq \v$,
\begin{align*}
\s_k &\coloneqq f_k(\s_{k-1}) \in \cS_k \\
\t_k &\coloneqq \partial f_k(\s_{k-1})[\t_{k-1}] \in \cS_k.
\end{align*}
In the case of a \textbf{computation graph}, 
specified by functions $f_1, \dots, f_K$ in topological order,
each $f_k$ may now take \textbf{multiple} inputs
$(\s_{i_1}, \dots, \s_{i_{p_k}}) \in \cS_{i_1} \times \dots \times \cS_{i_{p_k}}$,
and each $\partial f_k(\s_{i_1}, \dots, \s_{i_{p_k}})$ 
takes as many input directions 
$(\t_{i_1}, \dots, \t_{i_{p_k}}) \in \cS_{i_1} \times \dots \times \cS_{i_{p_k}}$,
where $(i_1, \ldots, i_{p_k}) \coloneqq \parent(k)$ are the parents of $f_k$. 
The forward pass on step $k \in [K]$ then computes both intermediate inputs and
directions as
\begin{align*}
\s_k &\coloneqq f_k(\s_{i_1}, \dots, \s_{i_{p_k}}) \in \cS_k \\
\t_k &\coloneqq 
\partial f_k(\s_{i_1}, \dots, \s_{i_{p_k}})[\t_{i_1}, \dots, \t_{i_{p_k}}] \in
\cS_k.
\end{align*}
Let us recall that $\partial_i f_k$ means that we differentiate $f_k$ \wrt
its $i$\textsuperscript{th} argument.
Using the fan-in rule in \cref{diff:prop:multiple_inputs}, 
we obtain that the derivatives are propagated as
\begin{equation*}
\t_k 
= \sum_{j=1}^{p_k}
\partial_{j} f_k(\s_{i_1}, \dots, \s_{i_{p_k}})[\t_{i_j}] \in \cS_k.
\end{equation*}
The final output is $\partial f(\s_0)[\v] = \t_K$.
The resulting generic forward-mode procedure is summarized in
\cref{auto_diff:fig:comp_graph_forward_mode}
and in
\cref{auto_diff:algo:jvp_graph}. 
Intuitively, the algorithm consists in computing and summing intermediate JVPs
along the forward pass.  Although not explicitly mentioned, we can release
$\s_k$ and $\t_k$ from memory when no child node depends on node $k$.

\begin{algorithm}[p]\caption{Forward-mode autodiff for computation graphs}
\label{auto_diff:algo:jvp_graph}
    \begin{algorithmic}[1]
        \Statex {\bf Functions:} $f_1, \dots, f_K$ in topological order
        \Statex {\bf Inputs:} input $\s_0 \in \cS_0$, input direction $\v \in
        \cS_0$
        \State Initialize $\t_0 \coloneqq \v$
        \For{$k \coloneqq 1, \ldots, K$} \Comment{Forward pass}
        \State Retrieve parent nodes $(i_1, \dots, i_{p_k}) \coloneqq \parent(k)$
        \State Compute 
          $\s_k \coloneqq f_k(\s_{i_1}, \dots, \s_{i_{p_k}}) \in \cS_k$ 
        \State Compute 
\begin{align*}
\t_k &\coloneqq 
\partial f_k(\s_{i_1}, \dots, \s_{i_{p_k}})[\t_{i_1}, \dots, \t_{i_{p_k}}] \\
& = \sum_{j=1}^{p_k} \partial_{j} f_k(\s_{i_1}, \dots, \s_{i_{p_k}})[\t_{i_j}]
\in \cS_k.
\end{align*}
        \EndFor
        \State {\bf Outputs:} $f(\s_0) \coloneqq \s_K \in \cS_K$, 
        $\partial f(\s_0)[\v] = \t_K \in \cS_K$
    \end{algorithmic}
\end{algorithm}

\newpage

\subsection{Reverse mode}

The reverse mode corresponds to computing the VJP of the program in an output
direction $\u \in \cS_K$. In the case of a \textbf{computation chain} $f
\coloneqq f_K \circ \dots \circ f_1$, each intermediate variable $\s_k$ is
\textbf{used only once} to compute the next variable $\s_{k+1} \coloneqq
f_{k+1}(\s_k)$. To compute the VJP of the chain, it then suffices to
reverse the order of the operations and to use the corresponding VJPs.
Since each function takes a single input, each VJP $\partial
f_k(\s_k)^*[\r_k]$ produces a \textbf{single} output direction $\r_{k-1} \in
\cS_k$. 
A backward pass therefore computes from $k \coloneqq K$ to $k \coloneqq 1$, 
starting from $\r_K \coloneqq \u$,
\begin{equation*}
\r_{k-1} \coloneqq \partial f_k(\s_{k-1})^*[\r_k] \in \cS_{k-1}.
\end{equation*}
The final output is $\r_0 \coloneqq \partial f(\s_0)^*[\u]$.

In \textbf{computation graphs}, intermediate variables may be used more than
once, since a node $k$ may have several children nodes. This complicates
reversing the computation graph. 
To circumvent this issue, 
following the formalism of Roy Frostig \citep[Section 6.2.2.2]{pml2_book},
we may formally
distinguish between the output $\s_k$ of $f_k$ and the inputs $\s_{k \rightarrow
j}$ of $f_j$ for $j \in \child(k)$, the children of $f_k$.
We do so by introducing an operation that simply duplicates $\s_k$, 
\begin{align*}
\dup(\s_k) & 
\coloneqq (\s_k, \ldots, \s_k).
\end{align*}
The tuple output by $\mathrm{dup}$ is of length $|\child(k)|$.
The forward pass can then be formally rewritten as, for $k \in [K]$,
\begin{equation*}
\s_k \coloneqq f_k(\s_{i_1 \rightarrow k}, \ldots, \s_{i_{p_k} \rightarrow k})
\end{equation*}
where $(i_1, \dots, i_{p_k}) \coloneqq \parent(k)$
followed by
\begin{equation*}
(\s_{k \rightarrow j_1}, \ldots, \s_{k \rightarrow j_{c_k}}) \coloneqq
\dup(\s_k),
\end{equation*}
where $(j_1, \ldots, j_{c_k}) \coloneqq \child(k)$. The benefit of this approach
is that each duplicated output is input only once to the subsequent child
functions. Thanks to this, we can now associate to each variable, $\s_{i
\rightarrow j}$ or $\s_k$, a single corresponding intermediate variable, $\r_{i
\rightarrow j}$ or $\r_k$, in the backward pass. The reverse mode can then
simply be written by going through the VJPs of the functions $f_k$ and the VJP
of $\dup$ in reverse order.

For the VJP of the $\mathrm{dup}$ operation,
let us denote
the intermediate variables in the backward pass associated with
$\s_{k \rightarrow j_1}, \ldots, \s_{k \rightarrow j_{c_k}}$
by
$\r_{k \rightarrow j_1}, \dots, \r_{k \rightarrow j_{c_k}}$.
Following the fan-out rule in \cref{diff:prop:multiple_outputs}, the VJP of
the $\dup$ operation on iteration $k$ is  
\begin{align*}
  \r_k & 
  = \dup(\s_k)^*[\r_{k \rightarrow j_1}, \dots, \r_{k \rightarrow j_{c_k}}] \\
   & = \sum_{i=1}^{c_k} \dup_i(\s_k)^*[\r_{k \rightarrow {j_i}}]\\
   & = \sum_{i=1}^{c_k} \r_{k \rightarrow {j_i}}.
\end{align*}
In the last line, we used that $\dup_i(\s) = \s$ by definition of the
duplication operation so $\partial\dup_i(\s) = \idm$ and $\partial\dup_i(\s)^* =
\idm$. The VJP of $\dup$ justifies why, if an intermediate value $\s_k$ is
used by later functions $f_{j_1},\dots, f_{j_{c_k}}$, for $(j_1,\dots,j_{c_k})
\coloneqq \child(k)$, the derivatives with respect to $\s_k$ need to sum all the
variations through the $f_j$ functions into the variable $\r_k$. 

\begin{figure}[t]
  \centering
  \includegraphics[width=\linewidth]{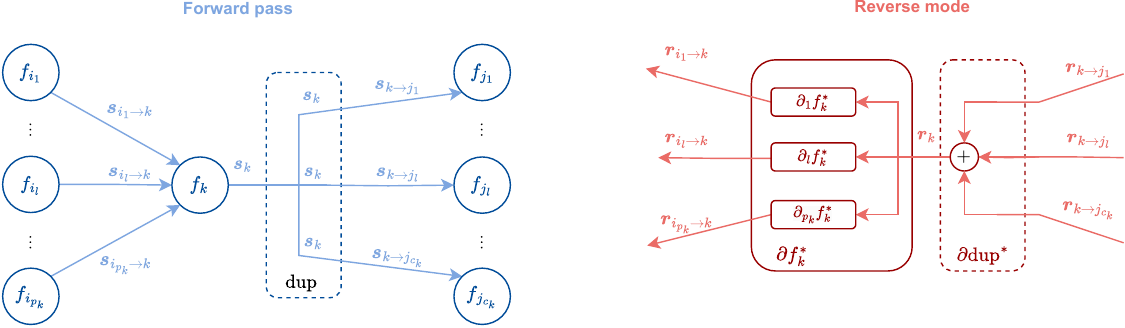}
  \caption{\label{auto_diff:fig:comp_graph_reverse_mode}
  Reverse mode automatic differentiation in a computation graph.
  }
\end{figure}

For the VJP of the $f_k$ functions, following the fan-in rule in
\cref{diff:prop:multiple_inputs}, the VJP returns \textbf{multiple} output
variations,
\begin{equation*}
\r_{i_1 \rightarrow k}, \dots, \r_{i_{p_k}  \rightarrow k} 
= \partial f_k(\s_{i_1 \rightarrow k}, \ldots, \s_{i_{p_k} \rightarrow k})^*[\r_k],
\end{equation*}
where $(i_1, \dots, i_{p_k}) \coloneqq \parent(k)$,
and where, for $j \in \{1, \dots, p_k\}$,
\begin{equation*}
\r_{i_j \rightarrow k} \coloneqq \partial_j f_k(\s_{i_1 \rightarrow k}, \ldots,
\s_{i_{p_k} \rightarrow k})^*[\r_k] \in \cS_{i_j}.
\end{equation*}
The overall formalism is summarized in
\cref{auto_diff:fig:comp_graph_reverse_mode}.

\subsubsection*{Implementation}

The duplication operation $\dup$ is just a formalism to mathematically derive
the reverse mode. In
practice, the variables $\s_k$ are not duplicated but accessed several times
during the backward pass. The variables $\r_k$ are computed by accumulating $\r_{k
\rightarrow j}$ into $\r_k$ each time a variable $\r_{k \rightarrow j}$ is
computed. Therefore, for each $k \in [K]$, we can compute the VJP and perform
the in-place updates,
\begin{align*}
\r_{i_1 \rightarrow k}, \dots, \r_{i_{p_k}  \rightarrow k} 
& = \partial f_k(\s_{i_1}, \ldots, \s_{i_{p_k}})^*[\r_k] \\
\r_{i_j} & \leftarrow  \r_{i_j} + \r_{i_j \rightarrow k}
\quad \forall j \in \{1, \dots, p_k\}.
\end{align*}
The topological ordering ensures that $\r_k$ has been fully computed
when we reach $f_k$. The resulting generic reverse-mode procedure is
presented in \cref{auto_diff:algo:vjp_graph}.

\begin{algorithm}[t]\caption{Reverse-mode autodiff for computation graphs}
  \label{auto_diff:algo:vjp_graph}
      \begin{algorithmic}[1]
          \Statex {\bf Functions:} $f_1, \dots, f_K$ in topological order
          \Statex {\bf Inputs:} input $\s_0 \in \cS_0$, output direction $\u \in
          \cS_K$
          \For{$k \coloneqq 1, \ldots, K$} \Comment{Forward pass}
          \State Retrieve parent nodes $(i_1, \dots, i_{p_k}) \coloneqq \parent(k)$
          \State Compute 
          $\s_k \coloneqq f_k(\s_{i_1}, \ldots, \s_{i_{p_k}}) \in \cS_k$ 
          \State Instantiate VJP
          $l_k \coloneqq \partial f_k(\s_{i_1}, \ldots, \s_{i_{p_k}})^*$ 
          \EndFor
          \State Initialize $\r_K \coloneqq \u$, 
          $\r_k \leftarrow 0 ~ \forall k  \in \{0, \ldots, K-1\}$ 
          \Comment{Backward pass}
          \For{$k \coloneqq K, \ldots, 1$}
          \State Retrieve parent nodes $(i_1, \dots, i_{p_k}) \coloneqq \parent(k)$ 
          \State Compute $\r_{i_1 \rightarrow k}, \dots, \r_{i_{p_k} \rightarrow k} \coloneqq 
          l_k[\r_k]$
          \State Compute $\r_{i_j} \leftarrow \r_{i_j} + \r_{i_j \rightarrow k}
          \in \cS_{i_j} ~ \forall j \in \{1, \dots, p_k\}$
          \EndFor
          \State {\bf Outputs:} $f(\s_0) \coloneqq \s_K \in \cS_K$, 
          $\partial f(\s_0)^*[\u] = \r_0 \in \cS_0$
      \end{algorithmic}
  \end{algorithm}

\begin{figure}[t]
  \includegraphics[width=\linewidth]{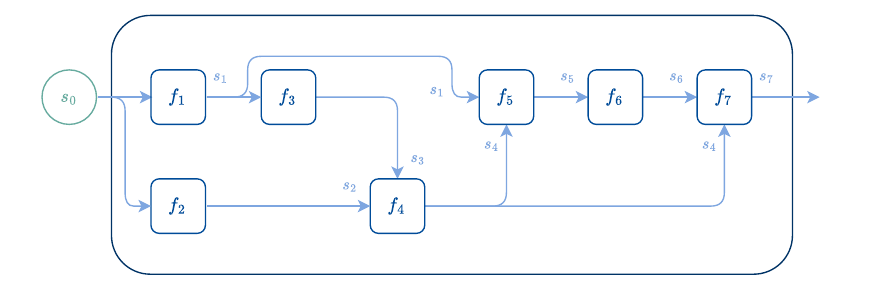}
  \caption{
Same computation graph as \cref{neural_nets:fig:simple_graph}
but without the actual definition of each $f_k$, for simplicity.
\label{auto_diff:fig:graph_comput_simplified}}
\end{figure}

\begin{boxexm}{Example of forward and reverse modes}
\label{auto_diff:exm:forward_and_reverse_modes}
We use \cref{auto_diff:fig:graph_comput_simplified} to illustrate the forward and
reverse modes in a computation graph. Let us assume that the intermediate
variables $\s_1,\dots,\s_7$ have readily been computed.
The \textbf{forward mode} corresponds to
\begin{align*}
\t_0 &\coloneqq \v \\
\t_1 &\coloneqq \partial f_1(\s_0)[\t_0] \\
\t_2 &\coloneqq \partial f_2(\s_0)[\t_0] \\
\t_3 &\coloneqq \partial f_3(\s_1)[\t_1] \\
\t_4 &\coloneqq \partial f_4(\s_2, \s_3)[\t_2, \t_3]
     = \partial_1 f_4(\s_2, \s_3)[\t_2] + 
       \partial_2 f_4(\s_2, \s_3)[\t_3] \\
\t_5 &\coloneqq \partial f_5(\s_1, \s_4)[\t_1, \t_4]
     = \partial_1 f_5(\s_1, \s_4)[\t_1] +
       \partial_2 f_5(\s_1, \s_4)[\t_4] \\
\t_6 &\coloneqq \partial f_6(\s_5)[\t_5] \\
\t_7 &\coloneqq \partial f_7(\s_4, \s_6)[\t_4, \t_6]
      = \partial_1 f_7(\s_4, \s_6)[\t_4] +
        \partial_2 f_7(\s_4, \s_6)[\t_6].
\end{align*}
The \textbf{reverse mode} corresponds to
\begin{align*}
(\r_{4 \to 7}, \r_{6 \to 7}) 
&\coloneqq \partial f_7(\s_4, \s_6)^*[\r_7]
    \quad &&\r_7 \coloneqq \u \\
\r_{5 \to 6} 
&\coloneqq \partial f_6(\s_5)^*[\r_6]
    \quad &&\r_6 \coloneqq \r_{6 \to 7} \\
(\r_{1 \to 5}, \r_{4 \to 5})
&\coloneqq \partial f_5(\s_1, \s_4)^*[\r_5]
    \quad &&\r_5 \coloneqq \r_{5 \to 6} \\
(\r_{2 \to 4}, \r_{3 \to 4})
&\coloneqq \partial f_4(\s_2, \s_3)^*[\r_4]
    \quad &&\r_4 \coloneqq \r_{4 \to 5} + \r_{4 \to 7} \\
\r_{1 \to 3}
&\coloneqq \partial f_3(\s_1)^*[\r_3]
    \quad &&\r_3 \coloneqq \r_{3 \to 4} \\
\r_{0 \to 2}
&\coloneqq \partial f_2(\s_0)^*[\r_2]
    \quad &&\r_2 \coloneqq \r_{2 \to 4} \\
\r_{0 \to 1}
&\coloneqq \partial f_1(\s_0)^*[\r_1]
    \quad &&\r_1 \coloneqq \r_{1 \to 3} + \r_{1 \to 5} \\
& && \r_0 \coloneqq \r_{0 \to 1} + \r_{0 \to 2}.
\end{align*}
\end{boxexm}

\subsection{Complexity, the Baur-Strassen theorem}
\label{auto_diff:sec:complexity_dag}

For computing the gradient of a function $f \colon \cE \to \RR$ represented by
a computation graph, we saw that the reverse mode is more efficient than the
forward mode. As we previously stated, assuming that the elementary functions
$f_k$ in the DAG and their VJP have roughly the same computational complexity, 
then $f$ and $\nabla f$ have roughly the same computational complexity.
This fact is crucial and is the pillar on which modern machine learning relies:
it allows us to optimize high-dimensional functions by
gradient descent.

For arithmetic circuits, reviewed in \cref{neural_nets:sec:arithmetic_circuits},
this crucial fact is made more precise in the celebrated Baur-Strassen
theorem \citep{baur1983complexity}.
If $f$ is a polynomial, then so is its gradient $\nabla f$.
The theorem gives an upper bound on the size of the best circuit for computing
$\nabla f$ from the size of the best circuit for computing $f$.
\begin{boxprop}{Baur-Strassen's theorem}
For any polynomial \\ $f \colon \cE \to \RR$, we have 
\begin{equation*}
S(\nabla f) \le 5 \cdot S(f),
\end{equation*}
where the size $S(f)$ of a polynomial $f$ 
is defined in \cref{neural_nets:def:circuit_size}.
\end{boxprop}
A simpler proof by backward induction was given by \citet{morgenstern_1985}.
See also the proof of Theorem 9.10 in \citet{chen_2011}.
For general computation graphs, 
that have more primitive functions than just $+$ and $\times$,
a similar result can be obtained;
see, e.g., \citep[Theorem 2]{bolte2022complexity}.

\section{Implementation}

\subsection{Primitive functions}

An autodiff system implements a set $\cA$ of 
primitive or elementary functions, which serve as building blocks for creating
other functions, by function composition.
For instance, we saw that in arithmetic circuits
(\cref{neural_nets:sec:arithmetic_circuits}),
$\cA = \{+, \times\}$. More generally, $\cA$ may contain all
the necessary functions for expressing programs. 
We emphasize, however, that $\cA$
is not necessarily restricted to low-level functions such as log and exp, but
may also contain higher-level functions.  For instance, even though the
log-sum-exp can be expressed as the composition of elementary operations (log,
sum, exp), it is usually included as a primitive on its own, both because it is
a very commonly-used building block, but also for numerical stability reasons.

\subsection{Closure under function composition}

Each function $f_k$ in a computation graph belongs to a set $\cF$, the class of
functions supported by the system. A desirable property of an autodiff
implementation is that the set $\cF$ is \textbf{closed} under function
composition, 
meaning that if $f \in \cF$ and $g \in \cF$, then $f \circ g \in \cF$.
This means that composed functions can themselves be used for
composing new functions. This property is also crucial for supporting
higher-order differentiation (\cref{chap:higher}) and automatic linear
transposition (\cref{auto_diff:sec:auto_linear_transposition}).
When $f_k$ is a composition of elementary functions in $\cA$, then $f_k$
itself is a \textbf{nested} DAG.
However, we can always \textbf{inline} each composite function, such
that all functions in the DAG belong to $\cA$.

\subsection{Examples of JVPs and VJPs}

An autodiff system must implement for each $f \in \cA$ its JVP for
supporting the forward mode, and its VJP for supporting the reverse mode.
We give a couple of examples.
We start with the JVP and VJP of linear functions.
\begin{boxexm}{JVP and VJP of linear functions}
\label{diff:exm:vjp_lin_mat}
	Consider the matrix-vector product
	$f(\W) = \W\x \in \RR^M$, where $\x \in
	\RR^D$ is fixed and $\W \in \RR^{M\times D}$. As
	already mentioned in \cref{diff:sec:need_linear_maps}, 
    the JVP of $f$ at $\W \in \RR^{M \times D}$ along an input
	direction $\V \in \RR^{M \times D}$ is simply 
	\begin{equation*}
	\partial f(\W)[\V] = f(\V) =
	\V\x \in \RR^M.
	\end{equation*}
	To find the associated VJP, we note that for any $\u \in \RR^M$ and $\V \in
	\RR^{M\times D}$, we must have $\langle \partial f(\W)[\V], \u \rangle =
	\langle \V, \partial f(\W)^*[\u] \rangle$. Using the properties of the
	trace, we have
	\begin{equation*}
	\langle \partial f(\W)[\V], \u \rangle 
	= \langle \V \x, \u \rangle 
	= \Tr(\x^\top \V^\top \u )
	= \langle \V, \u \x^\top \rangle.
	\end{equation*}
	Therefore, we find that the VJP is given by 
	\[\partial f(\W)^*[\u] = \u \x^\top \in \RR^{M \times D}.\]
	Similarly, consider now a matrix-matrix product $f(\W) = \W\X$, where
	$\W \in \RR^{M \times D}$ and where $\X \in \RR^{D \times N}$ is fixed. 
	The JVP at $\W \in \RR^{M \times D}$ along an input direction $\V \in \RR^{M
	\times D}$ is simply
	\begin{equation*}
	\partial f(\W)[\V] = f(\V) = \V\X \in \RR^{M\times N}.
	\end{equation*}
	The VJP along the output direction $\U \in \RR^{M\times N}$
	is
	\[
	\partial f(\W)^*[\U] = \U \X^\top \in \RR^{M \times D}.
	\]
\end{boxexm}
Another simple example is element-wise separable functions.
\begin{boxexm}{JVP and VJP of separable function}\label{diff:exm:jvp_vjp_sep_func}
Consider the function $f(\w) \coloneqq (g_1(w_1), \dots, g_P(w_P))$,
where each $g_i \colon \RR \to \RR$ has a derivative $g_i'$.
The Jacobian matrix is then a diagonal matrix
\begin{equation*}
\jac f(\w) = \diag(g_1'(w_1), \dots, g_P'(w_P)) \in \RR^{P \times P}.
\end{equation*}
In this case, the JVP and VJP are actually the same
\begin{equation*}
\partial f(\w)[\v] 
= \partial f(\w)^*[\v]
= (g_1'(w_1), \dots, g_P'(w_P)) \odot \v,
\end{equation*}
where $\odot$ indicates element-wise multiplication.
\end{boxexm}

\subsection{Automatic linear transposition}
\label{auto_diff:sec:auto_linear_transposition}

On first sight, if we want to support both forward and reverse modes,
it appears like we need to implement both the JVP and the VJP for each
primitive operation $f \in \cA$. 
Fortunately, there exists a way to recover VJPs from JVPs, and vice-versa.

We saw in \cref{diff:sec:jvp_vjp} that if $l(\w)$ is a linear map, then
its JVP is $\partial l(\w)[\v] = l(\v)$ (independent of $\w$). 
Conversely, the VJP is 
$\partial l(\w)^*[\u] = l^*(\u)$, where $l^*$ is the adjoint operator of $l$ 
(again, independent of $\w$).

Let us define $l(\u; \w) \coloneqq \partial f(\w)^*[\u]$, i.e.,
the VJP of $f$ in the output direction $\u$.
Since $l(\u; \w)$ is linear in $\u$,
we can apply the reasoning above to
compute its VJP
\begin{equation*}
\partial l(\u; \w)^*[\v] 
= l^*(\v; \w)
= \partial f(\w)^{**}[\v]
= \partial f(\w)[\v],
\end{equation*}
which is independent of $\u$.
In words, the VJP of a VJP is the corresponding JVP!
This means that we can implement forward-mode autodiff even if we only have
access to VJPs. As an illustration and sanity check, we give the following
example.
\begin{boxexm}{Automatic transpose of ``dot''}
If we define \\
$f(\x, \W) \coloneqq \W \x$,
from \cref{diff:exm:vjp_lin_mat}, we know that
\begin{align*}
\partial f(\x, \W)^*[\u]
&= (\W^\top \u, \u \x^\top) \\
&= (f(\u, \W^\top), f(\x^\top, \u)) \\
&\eqqcolon l(\u; \x, \W).
\end{align*}
Using \cref{diff:prop:multiple_outputs}, we obtain
\begin{align*}
\partial l(\u; \x, \W)^*[\v, \V]
&= f(\v, \W) + f(\x, \V) \\
&= \W \v + \V \x \\
&= \partial f(\x, \W)[\v, \V].
\end{align*}
\end{boxexm}
The other direction, automatically creating a VJP from a JVP, 
is also possible but is more technical and relies on the notion of
\textbf{partial evaluation} \citep{frostig_2021,radul_2022}.

\section{Checkpointing}\label{auto_diff:sec:checkpointing}

We saw that forward-mode autodiff can release intermediate computations from
memory along the way, while reverse-mode autodiff needs to cache all of them.
This means that the memory complexity of reverse-mode autodiff, in its standard
form, grows linearly with the number of nodes in the computation graph. A
commonly-used technique to circumvent this issue is checkpointing, which
trades-off computation time for better memory usage. Checkpointing works by
selectively storing only a subset of the intermediate values, called
\textbf{checkpoints}, and by recomputing others on the fly. The specific choice of
the checkpoint locations in the computation graph determines the
memory-computation trade-off.  While it is possible to heuristically set
checkpoints at user-specified locations, it is also possible to perform a
checkpointing strategy algorithmically, as studied in depth by
\citet{griewank1992achieving,griewank2008evaluating}. In this section, we review
two divide-and-conquer algorithms: recursive halving and dynamic programming. 
Our exposition focuses on computation chains $f = f_K
\circ \ldots \circ f_1$, with $f_k: \RR^D \rightarrow \RR^D$ for simplicity.

\begin{figure}[t]
  \includegraphics[width=\linewidth]{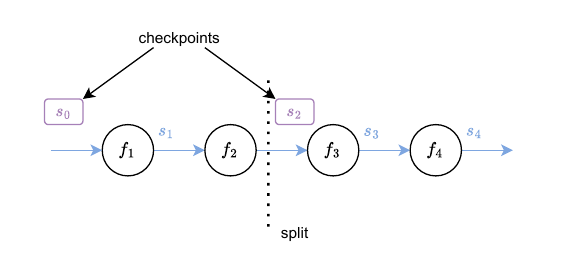}
  \caption{Checkpointing trades-off computation time for better memory usage by
selectively storing only a subset of the intermediate values, called
checkpoints, and by recomputing others on the fly. Recursive halving and dynamic
programming are two divide-and-conquer strategies to select the checkpoint
locations.
\label{auto_diff:fig:checkpointing}}
\end{figure}

\paragraph{Computational and memory complexities at two extremes}

Let $\cC(K)$ be the number of calls to the individual functions $f_k$ (we
ignore the cost of computing the intermediate VJPs) and $\cM(K)$ be the number
of function inputs cached, when performing reverse-mode autodiff on a chain $f
= f_K \circ \ldots \circ f_1$. On one extreme, if we store all intermediate
computations, as done in \cref{auto_diff:algo:reverse_chain}, to compute only
the VJP $\partial f(\s_0)^*[\u]$, we have
\[
\cC(K) = K-1 \quad \text{and} \quad \cM(K) = K.
\]
This is optimal \wrt computational complexity, but suboptimal \wrt memory. On
the other extreme, if we only store the initial input, as done in
\cref{auto_diff:algo:reverse_full_recompute}, then we have
\[
\mathcal{C}(K) = K(K-1)/2 \quad \text{and} \quad \mathcal{M}(K) = 1.
\]
This is optimal \wrt memory but leads to a computational complexity that is
quadratic in $K$. 

\begin{algorithm}[t]\caption{Reverse-mode autodiff with constant memory \\
  $\texttt{vjp\_full\_recompute}(f_K \circ \ldots \circ f_1, \s_0, \u)
  \coloneqq \partial (f_K \circ \ldots \circ f_1)(\s_0)^*[\u]$
  \label{auto_diff:algo:reverse_full_recompute}}
\begin{algorithmic}[1]
\Statex{\bf Inputs:} Chain $f_K \circ \ldots \circ f_1$, input
$\s_0 \in \cS_0$, output direction $\u \in \cS_K$
\If {$K=1$}
\State {\bf return} $\partial f_1(\s_0)^*[\u]$
\Else
\State Set $\r_K = \u$
\For {$k \coloneqq K, \ldots, 1$}
  \State Compute $\s_{k-1} = (f_{k-1} \circ \ldots  \circ f_1)(\s_0)$
  \State Compute $\r_{k-1} = \partial f_k(\s_{k-1})^*[\r_k]$
\EndFor
\State {\bf return:} $\r_0$ 
\EndIf
\end{algorithmic}
\end{algorithm}

\subsection{Recursive halving}

As a first step towards obtaining a better computation-memory trade-off,
we may split the chain $\s_K = f_K \circ \dots \circ
f_1(\s_0)$ as
\begin{align*}
    \s_{K/2} &= f_{K/2} \circ \ldots \circ f_1(\s_0) \\
    \s_{K} &= f_{K} \circ \ldots \circ f_{K/2+1}(\s_{K/2}),
\end{align*}
for $K$ even. Then, rather than recomputing all intermediate computations $\s_k$
from the input $\s_0$ as in \cref{auto_diff:algo:reverse_full_recompute}, we can
store $\s_{K/2}$ and recompute $\s_{k}$ for $k>K/2$ starting from $\s_{K/2}$.
Formally, this strategy amounts to the following steps.
\begin{enumerate}
  \item Compute $\s_{K/2} = f_{K/2} \circ \ldots \circ f_1(\s_0)$ 
  \item Compute $\r_{K/2} = \mbox{\texttt{vjp\_full\_recompute}}(f_K \circ \ldots
      \circ f_{K/2 +1}, \s_{K/2}, \u)$ 
  \item Compute $\r_0 = \mbox{\texttt{vjp\_full\_recompute}}(f_{K/2} \circ \ldots
      \circ f_1, \s_0, \r_{K/2})$
\end{enumerate}
At the expense of having to store the additional checkpoint $\s_{K/2}$,
this already roughly halves the computational complexity compared
to~\cref{auto_diff:algo:reverse_full_recompute}. 

\begin{algorithm}[t]\caption{Reverse-mode autodiff with recursive halving \\
  $\texttt{vjp\_halving}(f_K \circ \ldots \circ f_1, \s_0, \u) \coloneqq 
  \partial (f_K \circ \ldots \circ f_1)(\s_0)^*[\u]$
  \label{auto_diff:algo:binary_schedule}}
\begin{algorithmic}[1]
\Statex{\bf Functions:} Chain $f_K \circ \ldots \circ f_1$
\Statex{\bf Inputs:} input $\s_0 \in \cS_0$, output direction $\u \in \cS_K$
\If {$K=1$}
\State {\bf return} $\partial f_1(\s_0)^*[\u]$
\Else
\State Compute $\s_{K/2} = f_{K/2} \circ \ldots \circ f_1(\s_0)$ \label{auto_diff:line:binary_forward}
\State Compute $\r_{K/2} = \mbox{\texttt{vjp\_halving}}(f_K \circ \ldots \circ f_{K/2 +1}, \s_{K/2}, \u)$ 
\label{auto_diff:line:binary_second_half}
\State Compute $\r_0 = \mbox{\texttt{vjp\_halving}}(f_{K/2} \circ \ldots
\circ f_1, \s_0, \r_{K/2})$
\label{auto_diff:line:binary_first_half}
\State {\bf return:} $\r_0$
\EndIf
\end{algorithmic}
\end{algorithm}

\begin{figure}[t]
  \includegraphics[width=\linewidth]{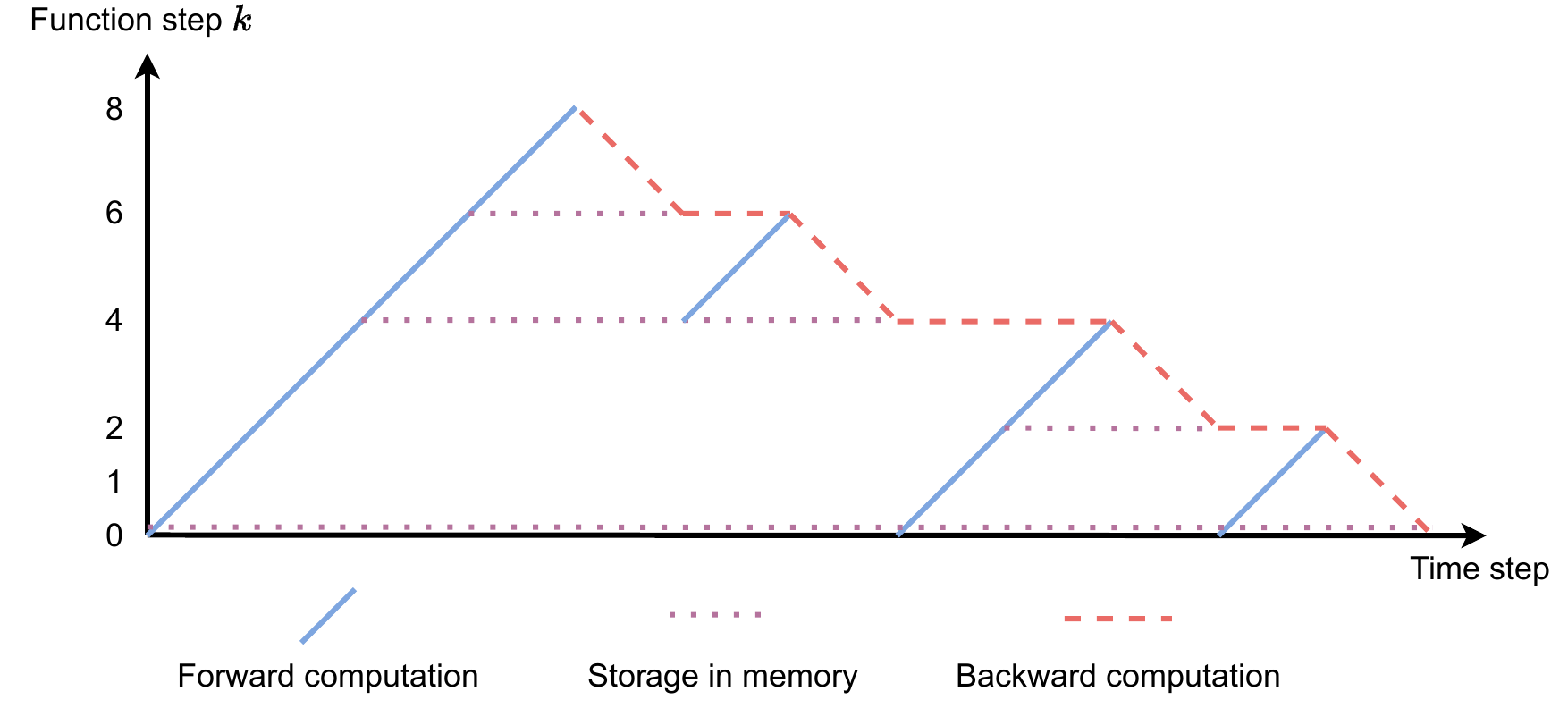}
  \caption{Illustration of checkpointing with recursive halving, for a chain of $8$
    functions. The chain is first fully evaluated while storing some
    computations as checkpoints in memory. 
    Then, during the backward pass,
    we recompute some intermediate values from the
    latest checkpoint available. 
    In contrast, vanilla reverse-mode autodiff (with full caching of the
    intermediate computations) would lead to a simple triangle shape. 
\label{auto_diff:fig:recursive_halving}}
\end{figure}

We can then apply this reasoning recursively, as formalized in
~\cref{auto_diff:algo:binary_schedule}. The algorithm is known as
\textbf{recursive binary schedule} \citep{griewank2003mathematical} and
illustrated in~\cref{auto_diff:fig:recursive_halving}. In terms of number of
function evaluations $\mathcal{C}(K)$, for $K$ even, we make $K/2$ function
calls, and we call the procedure recursively twice, that is,
\begin{align*}
    \cC(K) & = 2\cC(K/2) + K/2.
\end{align*}
If the chain is of length $1$, we directly use the VJP, so $\cC(1) = 0$.
Hence, the number of function calls, if $K$ is a power of $2$, is
\[
\cC(K) = \frac{K}{2}\log_2 K.
\]
In terms of memory usage, \cref{auto_diff:algo:binary_schedule} uses $\s_0$ not
only at line~\ref{auto_diff:line:binary_forward} but also at line
\ref{auto_diff:line:binary_first_half}. So when the algorithm is called
recursively on the second half of the chain at
line~\ref{auto_diff:line:binary_second_half}, one memory slot is taken by $\s_0$.
This line is called recursively until the chain is reduced to a single function.
At that point, the total number of memory slots used is equal to the number of
times we split the function in half, that is, $\log_2 K$ for $K$ a power of $2$.
On the other hand, the input $\s_0$ is no longer used after
line~\ref{auto_diff:line:binary_first_half} of
\cref{auto_diff:algo:binary_schedule}. At that line, the memory slot taken by
$\s_0$ can be consumed by the recursive call on the first-half. In other words,
calling the algorithm recursively on the first half does not incur extra memory
cost. So if $K$ is a power of $2$, the memory cost of \cref{auto_diff:algo:binary_schedule} is
\[
\cM(K) = \log_2 K.
\]

\subsection{Dynamic programming}

Recursive halving requires $\log_2 K$ memory slots for a chain of length $K$.
However, as illustrated in \cref{auto_diff:fig:recursive_halving}, at a
given time step, all memory slots may not be exploited. 

To optimize the approach, we observe that recursive halving is just one instance
of a program that splits the chain and calls itself recursively on each part. In
other words, it is a form of \textbf{divide-and-conquer} algorithm. Rather than
splitting the chain in half, we may consider splitting the chain at some index
$l$. One split is used to reverse the computations from $l+1$ to $K$ by a
recursive call that consumes one memory slot. The other split is used on a
recursive call that reverses the computations from $0$ to $l$. That second call
does not require an additional memory slot, as it can use directly the memory
slot used by the original input $\s_0$. To split the chain in such two parts, we
need $l$ intermediate computations to go from $\s_0$ to $\s_l$. The
computational complexity $\mathcal{C}(k, s)$, counted as the number of function
evaluations, for a chain of length $k$ with $s$ memory slots then satisfies the
recurrence
\[
\mathcal{C}(k, s) = \mathcal{C}(k-l, s-1) + \mathcal{C}(l, s) + l,
\]
for all $l \in \{1, \dots, k-1\}$.
By simply taking $l = k/2$, we recover exactly the computational complexity of
recursive halving. To refine the latter, we may split the chain by
selecting $l$ to minimize the complexity. An optimal scheme must satisfy the
recursive equation, 
\begin{equation}\label{auto_diff:eq:checkpointing_dyn_prog}
  \mathcal{C}^*(k, s) \coloneqq
  \min_{1\leq l\leq k-1} \{\mathcal{C}^*(k-l, s-1) + \mathcal{C}^*(l, s) + l\}.
\end{equation}
Note that $\mathcal{C}^*(K, S)$ can be computed from $ \mathcal{C}^*(k, s)$ for
$k=1, \ldots, K-1$, $s=1, \ldots, S-1$. This suggests a \textbf{dynamic
programming} approach to find an optimal scheme algorithmically.
For a chain of length $k=1$, the cost is null as we directly reverse the
computation, so $\mathcal{C}^*(1, s) \coloneqq 0$. On the other hand for a
memory $s=1$, there is only one possible scheme that saves only the initial
input as in~\cref{auto_diff:algo:reverse_full_recompute}, so $ \mathcal{C}^*(k,
1) \coloneqq (k(k-1))/2$. The values $\mathcal{C}^*(k, s)$ can then be computed
incrementally from $k=1$ to $K$ and $s=1$ to $S$
using~\cref{auto_diff:eq:checkpointing_dyn_prog}. The optimal splits can be
recorded along the way as
\begin{equation*}
l^*(k, s) \coloneqq \argmin_{1\leq l\leq k-1}
\{\mathcal{C}^*(k-l, s-1) + \mathcal{C}^*(l, s) + l\}. 
\end{equation*}
The optimal split for $K, S$ can then be found by \textbf{backtracking} 
the optimal splits along both branches corresponding to $\mathcal{C}^*(k-l,
s-1)$ and $\mathcal{C}^*(l, s)$. 
As the final output consists in traversing a binary tree, it was called
\textbf{treeverse}~\citep{griewank1992achieving}. Note that the dynamic
programming procedure is generic and could a priori incorporate varying
computational costs for the intermediate functions $f_k$.

\subsubsection*{Analytical formula}

It turns out that we can also find an optimal scheme \emph{analytically}. This
scheme was found by~\citet{griewank1992achieving}, following the analysis of
optimal inversions of sequential programs by divide-and-conquer algorithms done
by~\citet{grimm1996optimal}; see also~\citet[Section
6]{griewank2003mathematical} for a simple proof. 
The main idea consists in considering the number of times an evaluation step
$f_k$ is repeated. As we split the chain at $l$, all steps from $1$ to $l$ will
be repeated at least once. In other words, treating the second half of the chain
incurs one memory cost, while treating the first half of the chain incurs one
repetition cost. \citet{griewank1992achieving} shows that for fixed $K$, $S$, we
can find the minimal number of repetitions analytically and build the
corresponding scheme with simple formulas for the optimal splits.

Compared to the dynamic programming approach, it means that we do not need to
compute the pointers $l^*(k, s)$, and we can use a simple formula to set $l^*(k,
s)$. We still need to traverse the corresponding binary tree given $K, S$ and
$l^*(k, s)$ to obtain the schedules. Note that such an optimal scheme does not
take into account varying computational costs for the functions $f_k$.

\subsection{Online checkpointing}

The optimal scheme presented above requires knowing the total number of nodes in
the computation graph ahead of time. However, when differentiating through for
example a while loop (\cref{cf:sec:while_loops}), this is not the case. To
circumvent this issue, online checkpointing schemes have been developed and
proven to be nearly optimal~\citep{stumm2010new, wang2009minimal}. These schemes
start by defining a set of $S$ checkpoints with the first $S$ computations, then
these checkpoints are rewritten dynamically as the computations keep going. Once
the computations terminate, the optimal approach presented above for a fixed
length is applied on the set of checkpoints recorded.

\section{Reversible layers}\label{auto_diff:sec:reversible}

\subsection{General case}

The memory requirements of reverse-mode autodiff can be completely alleviated
when the functions $f_k$ are invertible (meaning that $f_k^{-1}$ exists) 
and when $f_k^{-1}$ is easily accessible. 
In that case, rather than storing the intermediate computations $\s_{k-1}$,
necessary to compute the VJP $\r_k \mapsto \partial f_k(\s_{k-1})^*[\r_k]$,
one can compute them on the fly during the backward pass from $\s_k$ using 
$\s_{k-1} = f_k^{-1}(\s_k)$.
We summarize the procedure for the case of computation chains in
\cref{auto_diff:algo:reversible}. Compared to vanilla reverse-mode autodiff in
\cref{auto_diff:algo:reverse_chain}, the algorithm has
optimal memory complexity, as we can release $\s_k$ and $\r_k$ as we go.

\begin{algorithm}[t]
\caption{Reverse-mode autodiff for reversible
chains.\label{auto_diff:algo:reversible}}
  \begin{algorithmic}[1]
    \Statex{\bf Functions:} $f \coloneqq f_K \circ \ldots \circ f_1$, 
    with each $f_k$ invertible
    \Statex{\bf Inputs:} input $\s_0 \in \cS_0$, output direction $\u \in \cS_K$
    \State Compute $\s_K = f_K \circ \ldots \circ f_1(\s_0)$
    \For {$k \coloneqq K, \ldots, 1$}
      \State Compute $\s_{k-1} = f_k^{-1}(\s_k)$
      \State Compute $\r_{k-1} 
      =  \partial f_k(\s_{k-1})^*[\r_k]$
    \EndFor
    \Statex{\bf Outputs:} $f(\s_0) \coloneqq \s_K$, 
    $\partial f(\s_0)^*[\u] = \r_0$
  \end{algorithmic}
\end{algorithm}

In practice, $f_k^{-1}$ often does not exist or may not be easily accessible.
However, network architectures can be constructed to be easily invertible by
design.
Examples include
reversible residual networks \citep{gomez_2017},
orthonormal RNNs \citep{helfrich2018orthogonal},
neural ODEs (\cref{chap:ode}),
and momentum residual neural networks \citep{sander_2021}; see also references
therein.

\subsection{Case of orthonormal JVPs}

When the JVP of each $f_k$ is an orthonormal linear mapping, i.e.,
\begin{equation*}
\partial f_k(\s_{k-1})^{-1} = \partial f_k(\s_{k-1})^*,
\end{equation*}
it is easy to check that the VJP of $f = f_K \circ \ldots \circ f_1$
is equal to the JVP of $f^{-1} = f_1^{-1} \circ \ldots \circ f_K^{-1}$, that
is
\begin{equation*}
\partial f(\s_0)^*[\u] = \partial f^{-1}(\s_K)[\u].
\end{equation*}
In other words, in the case of orthonormal JVPs, reverse-mode autodiff of $f$
coincides with forward-mode autodiff of $f^{-1}$!

\section{Randomized forward-mode gradient estimator}
\label{auto_diff:sec:randomized_forward}

Forward-mode autodiff does not require storing intermediate activations.
However, for a function $f \colon \RR^P \to \RR$, computing the gradient
$\nabla f$ using forward-mode autodiff requires $P$ JVPs, which is intractable
if $P$ is large. Can we approximate $\nabla f$ with fewer JVPs? The following
proposition gives an \textbf{unbiased estimator} of 
$\nabla f$ that only involves JVPs. 
\begin{boxprop}{Unbiased forward-mode estimator of the gradient}
Let $f \colon \RR^P \to \RR$ be a differentiable function. Then,
\begin{align*}
\nabla f(\muv)
&= \EE_{Z \sim p} \left[\partial f(\muv)[Z] Z \right] \\
&= \EE_{Z \sim p} \left[\langle \nabla f(\muv), Z \rangle Z \right],
\end{align*}
where $p \coloneqq \mathrm{Normal}(0,1)^P$ is the isotropic Gaussian
distribution.
\end{boxprop}
This estimator is for instance used by \citet{baydin_2022}.
It can be seen as the \textbf{zero-temperature limit} of the
gradient of a perturbed function, estimated by the score-function estimator
(SFE); see \cref{pert:sec:zero_temp_limit}.

In practice, the expectation above can be approximated by drawing $M$
noise vectors $\z_1,\dots,\z_M$, and averaging $\langle \nabla
f(\muv), \z_i \rangle \z_i$ over $i \in [M]$.

A word of caution: while this estimator can be useful for example when we do
not want to store the intermediate activations for memory reasons, this of
course comes at the cost of increasing the variance, which influences the
convergence rate of SGD, as seen in \cref{optim:sec:sgd}.

\begin{figure}[t]
    \centering
  \includegraphics[width=0.65\linewidth]{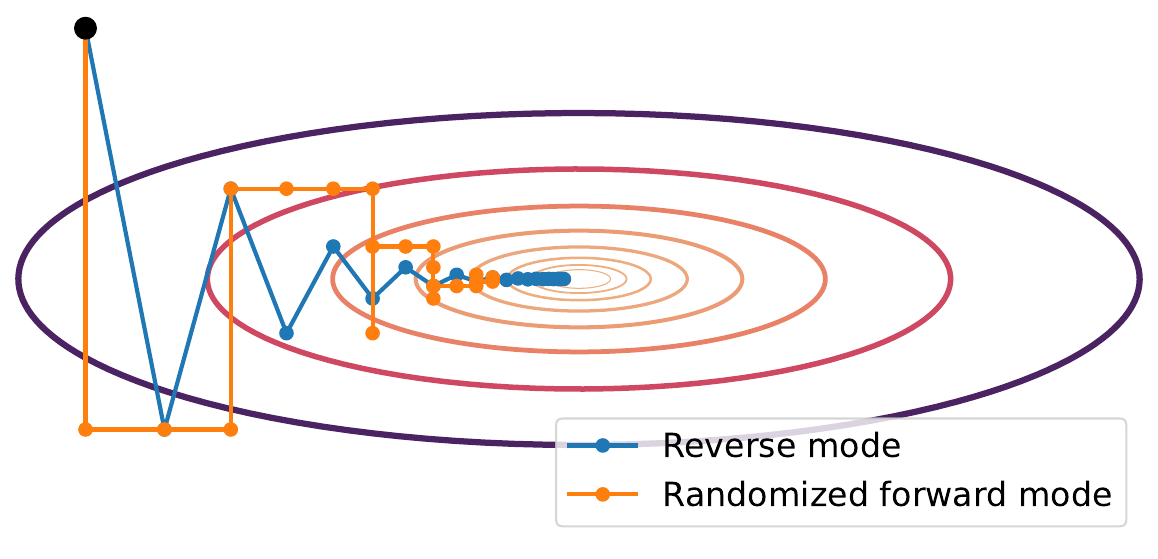}
  \caption{The randomized forward-mode gradient estimator only requires forward
      passes, but it suffers from high variance, even more so in high dimension.
\label{auto_diff:fig:randomized_forward_mode}}
\end{figure}

\section{Summary}

\begin{itemize}

\item Computer programs can be seen as directed acyclic graphs, where nodes
correspond to the output of intermediate operations in the program, and edges
represent the dependencies of current operations on past operations.  

\item Automatic
differentiation (autodiff) for a function $f \colon \RR^P \to \RR^M$ has two
main modes: forward mode and reverse mode. 

\item The forward mode:
i) uses JVPs, ii) builds the Jacobian one column at a time, iii) is efficient
for tall Jacobians ($M \ge P$), iv) need not store intermediate computations.

\item The reverse mode: i) uses VJPs, ii) builds the Jacobian one row at a time,
    iii) is efficient for wide Jacobians ($P \ge M$), iv) needs to store
    intermediate computations, in order to be computationally optimal.  

\item To trade computational efficiency for better memory efficiency,
we can use checkpointing techniques.

\item The complexity of computing the gradient of a function $f \colon
\RR^P \to \RR$ using the reverse mode is at most a constant time bigger than
that of evaluating the function itself. This is the Baur-Strassen
theorem, in arithmetic circuits. This astonishing result is one of the pillars
of modern machine learning.

\end{itemize}

%% file: chapters/higher_order/higher_order.tex
\chapter{Second-order automatic differentiation}
\label{chap:higher}

We review in this chapter how to perform automatic differentiation
for second-order derivatives.

\section{Hessian-vector products}
\label{higher:sec:hvp}

We consider in this section a function $f \colon \cE \to \RR$.
Similarly to the Jacobian, for most purposes, we do not need access to the full
Hessian but rather to the Hessian-vector product (HVP) $\nabla^2 f(\w) [\v]$ 
at $\w \in \cE$, in a direction $\v \in \cE$, as defined in~\cref{diff:def:hvp}.
The latter can be computed in four different ways, depending on how we combine
the two main modes of autodiff.

\subsection{Four possible methods}\label{higher:ssec:formulas}

An HVP can be computed in four different ways.

\begin{enumerate}

\item {\bf Reverse on reverse:}
The Hessian can be seen as the transposed Jacobian of the gradient, hence the
HVP can be computed as the \textbf{VJP of the gradient},
\[
  \nabla^2 f(\w)[\v] = \partial (\nabla f)(\w)^*[\v].
\]

\item {\bf Forward on reverse:}
Owing to its symmetry (see \cref{diff:prop:symmetry_hessian}), the Hessian can
also be seen as the Jacobian of the gradient, hence the HVP can be computed as
the \textbf{JVP of the gradient},
\[
  \nabla^2 f(\w)[\v] = \partial (\nabla f)(\w)[\v].
\]

\item {\bf Reverse on forward:}
Recall that for any function $g \colon \cE \to \cE$, 
the VJP can equivalently be defined as the gradient along an output direction
$\v \in \cE$, that is, 
\begin{equation*}
\partial g(\w)^*[\v] = \nabla \langle g, \v\rangle(\w),
\end{equation*}
where we recall the shorthand 
$\langle g, \v \rangle(\w) \coloneqq \langle \v, g(\w) \rangle$,
so that $\langle g, \v \rangle$ is a function of $\w$.
In our case, we can therefore rewrite the reverse-on-reverse approach as
\begin{equation*}
\partial (\nabla f)(\w)^*[\v] = \nabla \langle \nabla f, \v\rangle(\w).
\end{equation*}
We know that
$
\langle \nabla f, \v\rangle(\w) =
\langle \nabla f(\w), \v\rangle = 
\partial f(\w)[\v]$ is the JVP of $f$ at $\w$
along $\v$. 
Therefore, we can also compute the HVP as the \textbf{gradient of
the JVP} of $f$ at $\w$ along $\v$,
\[
  \nabla^2 f(\w)[\v] 
  = \nabla (\partial f(\cdot)[\v])(\w),
\]
where we use the notation
$(\partial f(\cdot)[\v])(\w) \coloneqq \partial f(\w)[\v] $ 
to insist on the fact that it is a function of $\w$.

\item {\bf Forward on forward:}
Finally, we can use the definition of the HVP in~\cref{diff:def:hvp} as a
vector of second partial derivatives along $\v$ and each canonical
direction. That is, assuming $\cE = \RR^P$,
we can compute the \textbf{JVP of the JVP} $P$ times:
\begin{equation*}
\nabla^2 f(\w)[\v] = (\partial^2 f(\w)[\v, \e_i])_{i=1}^P.
\end{equation*}

\end{enumerate}
The four different ways of computing the HVP
are summarized in \cref{higher:tab:hvp}.

\begin{table}[t]
  \begin{center}
  \begin{tabular}{c|c}
    Method & 
    Computation \\
    \hline
    Reverse on reverse (VJP of gradient)  & 
    $\partial (\nabla f)(\w)^*[\v]$ \\
    \hline 
    Forward on reverse (JVP of gradient) & 
    $\partial (\nabla f)(\w)[\v]$ \\
    \hline
    Reverse on forward (gradient of JVP) & 
    $\nabla (\partial f(\cdot)[\v])(\w)$ \\
    \hline 
    Forward on forward (JVPs of JVPs) & 
    $(\partial^2 f(\w)[\v, \e_i])_{i=1}^P$ \\
  \end{tabular}
  \caption{
    Four different ways of computing the HVP $\nabla^2 f(\w)[\v]$.
\label{higher:tab:hvp}
  }
  \end{center}
\end{table}

\subsection{Complexity}

To get a sense of the computational and memory complexity of the four
approaches, we consider a chain of functions $f \coloneqq f_K \circ \dots \circ
f_1$ as done in \cref{auto_diff:sec:chains}. To simplify our analysis, we assume
$f_k \colon \RR^P \to \RR^P$ for $k \in \{1, \dots, K-1\}$ and $f_K \colon \RR^P
\to \RR$.

\begin{figure}
  \includegraphics[width=\linewidth]{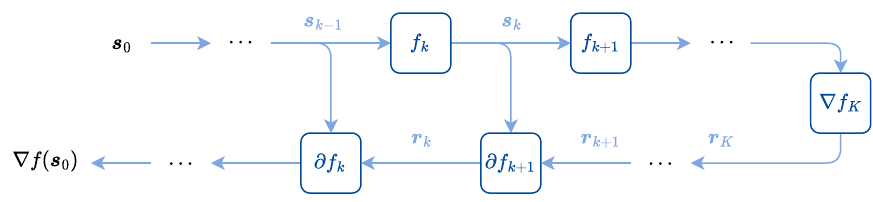}
  \caption{Computation graph corresponding to reverse mode autodiff for
  evaluating the gradient of $f=f_K \circ \ldots \circ f_1$. While $f$ 
  is a simple chain, $\nabla f$ is a DAG. \label{higher:fig:rev}}
\end{figure}

The computation graph of the reverse mode is illustrated in
\cref{higher:fig:rev}. While $f = f_K \circ \dots \circ f_1$ would be
represented by a simple chain, the computational graph of $\nabla f$ is no
longer a chain: it is a DAG.
This is due to the computations of $\partial f_k(\s_{k-1})^*[\r_k]$, where both
$\s_{k-1}$ and $\r_k$ depend on $\s_0$. 

\begin{figure}
  \includegraphics[width=\linewidth]{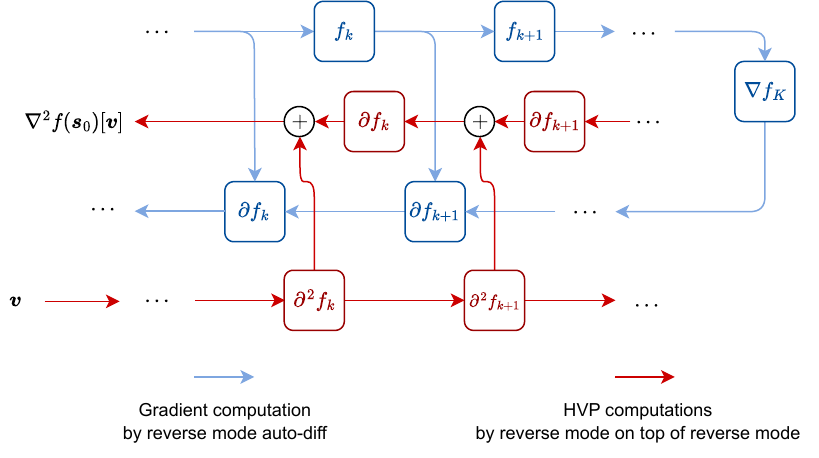}
  \caption{Computation graph for computing the HVP $\nabla^2 f(\w)[\v]$ by
  using reverse mode on top of reverse mode.
  As the computation graph of $\nabla f$ induces fan-in operations $\s_{k-1},
  \r_k \mapsto \partial f_k(\s_{k-1})^*[\r_k]$, the reverse mode applied on
  $\nabla f$ induces branching of the computations at each such node.
  \label{higher:fig:rev_on_rev}}
\end{figure}

\begin{figure}
  \includegraphics[width=\linewidth]{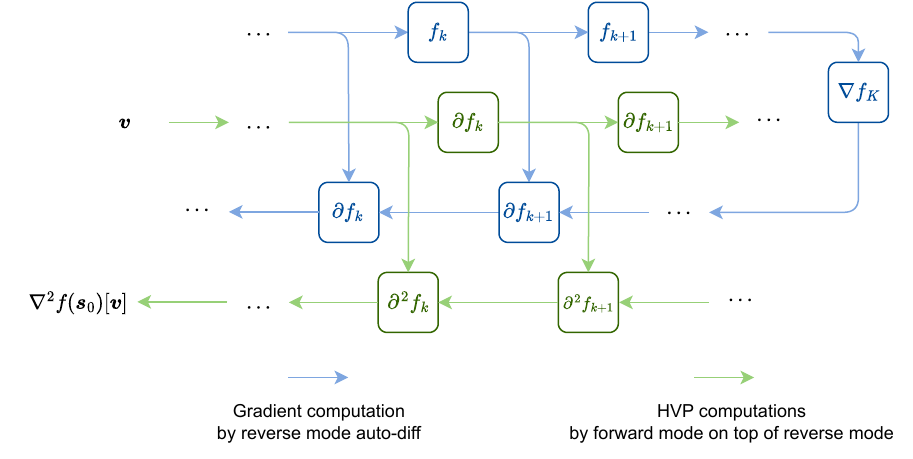}
  \caption{Computation graph for computing the HVP $\nabla^2 f(\w)[\v]$ by
  using forward mode on top of reverse mode.
  The forward mode naturally follows the computations done for the gradient,
  except that it passes through the derivatives of the intermediate operations.
  \label{higher:fig:for_on_rev}}
\end{figure}

We illustrate the computation graphs of reverse-on-reverse and
forward-on-reverse in \cref{higher:fig:rev_on_rev} and
\cref{higher:fig:for_on_rev} respectively. By applying reverse mode on reverse
mode, at each fan-in operation $\s_{k-1}, \r_k \mapsto \partial
f_k(\s_{k-1})^*[\r_k]$,
the reverse mode on $\nabla f$ branches out into two paths that are later merged
by a sum. By applying forward mode on top of reverse mode, the flow of
computations simply follows that of $\nabla f$.

With this in mind, following a similar calculation as for
\cref{autodiff:tab:chain_complexity}, we obtain the following results. We assume
that each $\partial f_k(\s_{k-1})$ is a dense linear operator, so that its
application has the same cost as a matrix-vector multiplication. For the memory
complexity, we consider that the inputs of each operation are saved to compute
the required derivatives in the backward passes.

\begin{enumerate}
\item {\bf Reverse on reverse:}
$O(KP^2)$ time and $O(KP)$ space.

\item {\bf Forward on reverse:}
$O(KP^2)$ time and $O(KP)$ space.

\item {\bf Reverse on forward:}
$O(KP^2)$ time and $O(KP)$ space.

\item {\bf Forward on forward:}
$O(KP^3)$ time and $O(3P)$ space for the $P$ JVPs with $\e_1,\dots,\e_P$.

\end{enumerate}

We see that, for chains of functions, ``reverse on reverse'', ``forward on
reverse'' and ``reverse on forward'' all have similar time complexities up to
some constant factors. Using reverse mode on top of reverse mode requires
storing the information backpropagated, i.e., the $\r_k$ (resp. the information
forwarded, i.e., the $\t_k$ in \cref{auto_diff:fig:chain_jvp}), to perform the
final reverse pass. By using forward mode on top of reverse mode, this
additional cost is not incurred, making it slightly less memory expensive. In
addition, reverse mode on top of reverse mode induces a few additional
summations due to the branching and merge operations depicted in
\cref{higher:fig:rev_on_rev}. The same holds when using reverse on top of
forward as we cannot avoid fan-in operations (this time of the form $\s_{k-1},
\t_{k-1} \mapsto \partial f_k(\s_{k-1})[\t_{k-1}]$).
Unfortunately, ``forward on forward'' is prohibitively expensive. 

To summarize,
among the four approaches presented to compute HVPs, the forward-on-reverse
mode is a priori the most preferable in terms of computational and memory
complexities.
Note, however, that computations of higher derivatives
can benefit from dedicated autodiff implementations such as Taylor mode
autodiff,
that do not merely compose forward and reverse modes. 
For general
functions $f$, it is reasonable to benchmark the first three methods to
determine which method is the best for the function at hand.

\section{Gauss-Newton matrix}
\label{higher:sec:gnvp}

\subsection{An approximation of the Hessian}

The Hessian matrix $\nabla^2 L(\w)$ of a function $L \colon \cW \to \RR$ is 
often used to construct a quadratic approximation of $L(\w)$,
\begin{equation*}
L(\w + \v) \approx L(\w) + \langle \nabla L(\w), \v \rangle
+ \frac{1}{2} \langle \v, \nabla^2 L(\w)[\v] \rangle.
\end{equation*}
Unfortunately, when $L$ is nonconvex, $\nabla^2 L(\w)$ is typically an
\textbf{indefinite matrix}, which means that the above approximation is a
\textbf{nonconvex quadratic} \wrt $\v$. 
For instance, if $L = \ell \circ f$ with $\ell$ convex, then $L$ is convex if
$f$ is linear, but it is typically nonconvex if $f$ is nonlinear.
The (generalized) Gauss-Newton matrix is a principled alternative to
the Hessian, which is defined for $L \coloneqq \ell \circ f$.
\begin{boxdef}{Gauss-Newton matrix}
  Given a differentiable function $f: \cW \rightarrow \cM$ and a twice
  differentiable function $\ell:\cM \rightarrow \RR$, the (generalized)
  Gauss-Newton matrix of the composition $L = \ell \circ f$ evaluated
  at a point $\w \in \cW$ is defined as 
  \begin{equation*}
      \gn (\ell \circ f)(\w) \coloneqq \partial f(\w)^* \nabla^2\ell(f(\w))
      \partial f(\w).
  \end{equation*}
\end{boxdef}
As studied in \cref{optim:sec:gauss_newton}, the Gauss-Newton matrix is a key
ingredient of the Gauss-Newton method.
An advantage of the Gauss-Newton matrix is its positive semi-definiteness
provided that $\ell$ is convex.
\begin{boxprop}{Positive semi-definiteness of the GN matrix}
If $\ell$ is convex, then $\gn(\ell \circ f)(\w)$ is positive
semi-definite for all $f$.
\end{boxprop}
This means that the approximation
\begin{equation*}
L(\w + \v) \approx L(\w) + \langle \nabla L(\w), \v \rangle
+ \frac{1}{2} \langle \v, \gn L(\w)[\v] \rangle
\end{equation*}
is a \textbf{convex quadratic} \wrt $\v$.

Using the chain rule, we find that the Hessian of $L = \ell \circ f$ decomposes
into the sum of two terms
(see also \cref{higher:thm:hessian_chain_rule}). 
\begin{boxprop}{Approximation of the Hessian}
For $f$ and $\ell$ twice differentiable, we have
\begin{align*}
  \nabla^2 (\ell \circ f)(\w)
  & = \partial f(\w)^* \nabla^2 \ell(f(\w)) \partial f(\w) 
  + \partial^2 f(\w)^* [\nabla \ell(f(\w))] \\
  & = \gn(\ell \circ f)(\w) + \sum_{j=1}^{M} \nabla_j \ell(f(\w)) \nabla^2
  f_j(\w).
\end{align*}
If $f$ is linear, then the Hessian and Gauss-Newton matrices coincide,
\begin{equation*}
\nabla^2 (\ell \circ f)(\w) = \gn(\ell \circ f)(\w).
\end{equation*}
\end{boxprop}
The Gauss-Newton operator $\gn(\ell \circ f)$ can therefore be seen as an
approximation of the Hessian $\nabla^2 (\ell \circ f)$, with equality if $f$
is linear.

\subsection{Gauss-Newton chain rule}

A chain rule for computing the Hessian of a composition of two functions is
presented in \cref{higher:thm:hessian_chain_rule}, but the formula is relatively
complicated, due to the cross-terms. In contrast, a Gauss-Newton chain rule is
straightforward.
\begin{boxprop}{Gauss-Newton chain rule}
\begin{align*}
\gn(\ell \circ f \circ g)(\w) 
= \partial g(\w)^* \gn(\ell \circ f)(g(\w)) \partial g(\w).
\end{align*}
\end{boxprop}

\subsection{Gauss-Newton vector product}

As for the Hessian, we rarely need to materialize the full Gauss-Newton matrix
in memory. Indeed, we can define the Gauss-Newton vector product (GNVP),
a linear map for a direction $\v \in \cW$, as
\begin{equation}
\gn (\ell \circ f)(\w)[\v] \coloneqq \partial f(\w)^* [\nabla^2\ell(f(\w))
[\partial f(\w)[\v]]],
\label{higher:eq:gnvp}
\end{equation}
where $\nabla^2\ell(f(\w))[\u]$ is the HVP of $\ell$,
a linear map from $\cM$ to $\cM$.
The GNVP can be computed using the JVP of $f$, the HVP of $\ell$ and the VJP of
$f$. Instantiating the VJP requires $1$ forward pass through $f$, from which
we get both the value $f(\w)$ and the adjoint linear map $\u \mapsto \partial f(\w)^*[\u]$. Evaluating the VJP requires $1$ backward pass through $f$.
Evaluating the JVP requires $1$ forward pass through $f$. In total, evaluating
$\v \mapsto \gn(\ell \circ f)(\w)[\v]$ therefore requires $2$ forward passes and
$1$ backward pass through $f$.

\subsection{Gauss-Newton matrix factorization}

In this section, we assume $\cW \subseteq \RR^P$ and $\cM \subseteq \RR^M$.
When $\ell$ is convex, we know that the Gauss-Newton matrix is positive
semi-definite and therefore it can be factorized into $\gn(\ell\circ f)(\w) =
VV^\top$ for some $V \in \RR^{P \times R}$, where $R \le \min\{P,M\}$ is the
rank of the matrix. 
Such a decomposition can
actually be computed easily from a factorization of the Hessian of $\ell$.
For instance, suppose we know the eigendecomposition of the Hessian of $\ell$,
$\nabla^2 \ell(f(\w)) = \sum_{i=1}^M \lambda_i \u_i \u_i^\top$,
where the $\u_i$ are the eigenvectors
and the $\lambda_i \geq 0$ are the eigenvalues (which we know are non-negative
due to positive semidefiniteness).
Then, the Gauss-Newton matrix can be decomposed as
\begin{align*}
  \gn(\ell \circ f)(\w) 
  & = \sum_{i=1}^M \lambda_i \partial f(\w)^*[\u_i] \u_i^\top\partial f(\w) \\
  & = \sum_{i=1}^M \left(\sqrt{\lambda_i} \partial f(\w)^*[\u_i]\right)
  \left(\sqrt{\lambda_i} \partial f(\w)^*[\u_i]\right)^\top \\
  & = \sum_{i=1}^M \v_i\v_i^\top 
  \quad \mbox{where} \ \v_i \coloneqq \sqrt{\lambda_i}\partial f(\w)^*[\u_i].
\end{align*}
Stacking the vectors $\v_i$ into a matrix $V = (\v_1, \ldots, \v_M)$, we
recover the factorization $\gn(\ell\circ f)(\w) = VV^\top$. 
To form this decomposition, we need to perform the eigendecomposition of 
$\nabla^2 \ell(f(\w)) \in \RR^{M \times M}$, which takes $O(M^3)$ time.
We also need $M$ calls to
the VJP of $f$ at $\w$. Compared to the direct implementation in
\cref{higher:eq:gnvp}, the factorization, once computed, 
allows us to compute the Gauss-Newton vector product (GNVP)
as $\gn (\ell \circ f)(\w)[\v] = V V^\top \v$.
The factorization only requires $P \times M$ memory, while 
the direct implementation in \cref{higher:eq:gnvp} 
requires us to maintain the intermediate computations
of $f$. The computation-memory trade-offs therefore depend on the function
considered.

\subsection{Stochastic setting}

Suppose the objective function is of the form 
\begin{equation*}
L(\w; \x, \y) \coloneqq \ell(f(\w; \x); \y).
\end{equation*}
With some slight abuse of notation, we then have that the Gauss-Newton
matrix associated with a pair $(\x,\y)$ is
\begin{equation*}
\gn L(\w; \x, \y) \coloneqq 
\partial f(\w; \x)^* \nabla^2 \ell(\thetav; \y) \partial f(\w; \x),
\end{equation*}
where $\thetav \coloneqq f(\w; \x)$.
Given a distribution $\rho$ over $(\x,\y)$ pairs, 
the Gauss-Newton matrix associated with the averaged loss
\begin{equation*}
L(\w) \coloneqq \EE_{X,Y \sim \rho} \left[L(\w; X, Y)\right]
\end{equation*}
is then
\begin{equation*}
\gn L(\w) = \EE_{X,Y \sim \rho}\left[\gn L(\w; X, Y)\right].
\end{equation*}

\section{Fisher information matrix}
\label{higher:sec:fisher}

\subsection{Definition using the score function}

The Fisher information is a way to measure the amount of information in a random
variable $S$.
\begin{boxdef}{Fisher information matrix}
  \label{optim:def:fisher}
  The \textbf{Fisher information matrix}, or Fisher for short, associated with
  the negative log-likelihood $L(\w; S) = -\log q_{\w}(S)$ of a probability
  distribution $q_\w$ with parameters $\w$ 
  is the covariance of the gradients of $L$ at $\w$ for $S$
  distributed according to $q_\w$,
  \begin{align*}
    \fisher L (\w) 
    &\coloneqq 
    \mathbb{E}_{S \sim q_{\w}}[\nabla L(\w; S) \otimes \nabla L(\w; S)] \\
    &= \mathbb{E}_{S \sim q_{\w}}[\nabla_\w \log q_{\w}(S) \otimes \nabla_\w \log
    q_{\w}(S)].
  \end{align*}
  The gradient $\nabla_\w \log q_{\w}(S)$ is known as the \textbf{score
  function}.
\end{boxdef}
As studied in \cref{optim:sec:nat_grad}, the Fisher information matrix is a key
ingredient of the natural gradient descent method.

\subsection{Link with the Hessian}

Provided that the probability distribution is twice differentiable
\wrt $\w$ with integrable second derivatives, the Fisher information matrix
can also be expressed as the Hessian of the negative log-likelihood
~\citep{amari1998natural, martens2020new}.
\begin{boxprop}{Connection with the Hessian}
  \label{optim:lem:fisher_hess}
  The Fisher information matrix of the negative log-likelihood $L(\w; S) = -\log
  q_{\w}(S)$ satisfies
  \begin{align*}
    \fisher L (\w) 
    &= \mathbb{E}_{S \sim q_{\w}}[\nabla^2 L(\w; S)] 
    = \mathbb{E}_{S \sim q_{\w}}[-\nabla_{\w}^2\log q_{\w}(S)].
  \end{align*}
\end{boxprop}

\begin{boxrem}{Empirical Fisher}
We emphasize that in the above definitions, $S$ is sampled from the model
distribution $q_\w$, not from the data distribution $\rho$.
That is, we have
\begin{align*}
\fisher L (\w)
&=\mathbb{E}_{S \sim q_{\w}}[\nabla_\w \log q_{\w}(S)\nabla_\w \log
q_{\w}(S)^\top] \\
&\neq
\mathbb{E}_{S \sim \rho}[\nabla_\w \log q_{\w}(S)\nabla_\w \log
q_{\w}(S)^\top]
\end{align*}
The latter is sometimes called ambiguously the ``empirical'' Fisher, though this
name has generated confusion~\citep{kunstner2019limitations}. 
\end{boxrem}

\subsection{Equivalence with the Gauss-Newton matrix}

So far, we discussed the Fisher information for a generic random variable 
$S \sim q_\w$.
We now discuss the supervised probabilistic learning setting where $S = (X, Y)$
and where, using the product rule of probability, we define the PDF
$q_\w(X, Y) \coloneqq \rho_X(X) p_\thetav(Y)$, 
with the shorthand $\thetav \coloneqq f(\w; X)$.
\begin{boxprop}{Fisher matrix in supervised setting}
  \label{optim:exm:fisher_supervised}
Suppose $(X, Y) \sim q_\w$ where the PDF of $q_\w$ is
$q_\w(X, Y) \coloneqq \rho_X(X) p_\thetav(Y)$.
In that case, the Fisher information matrix of
the negative log-likelihood $L(\w; \x, \y)= -\log q_{\w}(\x, \y)$ decomposes as:
  \begin{align*}
    \fisher L (\w) & = 
    \mathbb{E}_{(X, Y) \sim q_\w}[
      \nabla_\w \log q_\w(X, Y) \otimes \nabla_\w \log q_\w(X, Y)
      ] \\
    & = 
    \mathbb{E}_{X \sim \rho_X} \left[
      \mathbb{E}_{Y\sim p_{\thetav}}\left[
        \nabla_\w \log p_{\thetav}(Y) \otimes \nabla_\w \log p_{\thetav}(Y)\right]
      \right] \\
    & = 
    \mathbb{E}_{X \sim \rho_X} \left[
      \partial f(\w; X)^* \fisher \ell(\thetav)
        \partial f(\w; X) \right],
  \end{align*}
  where we defined the shorthand $\thetav \coloneqq f(\w; X)$
  and where we defined the negative log-likelihood loss
  $\ell(\thetav; Y) \coloneqq -\log p_\thetav(Y)$.
\end{boxprop}
When $p_\thetav$ is an exponential family distribution,
we can show that the Fisher information matrix and the Gauss-Newton matrix are
equivalent.
\begin{boxprop}{Equivalence between Fisher and Gauss-Newton}
If $p_\thetav$ is an exponential family distribution, then
\begin{align*}
\fisher L(\w)
&= \EE_{X \sim \rho_X} \EE_{Y \sim p_\thetav}
[\nabla L(\w; X, Y) \otimes \nabla L(\w; X, Y)] \\
&= \EE_{X \sim \rho_X} \EE_{Y \sim p_\thetav}
[\partial f(\w; X)^*
(\nabla \ell(\thetav; Y) \otimes \nabla \ell(\thetav; Y))
\partial f(\w; X)] \\
&= \EE_{X \sim \rho_X} \EE_{Y \sim p_\thetav}
[\partial f(\w; X)^* \nabla^2 \ell(\thetav; Y) \partial f(\w; X)] \\
&= \EE_{X,Y \sim \rho}\left[\gn L(\w; X, Y)\right],
\end{align*}
where $\rho_X(\x) \coloneqq \int \rho(\x, \y)d\y$.
\label{higher:prop:equiv_fisher_gn}
\end{boxprop}
\begin{proof}
From \cref{proba_learn:prop:negative_ll_exp_family},
if $p_\thetav$ is an exponential family distribution,
$\nabla^2 \ell(\thetav; \y)$ is actually independent of $\y$.
Using Bartlett's second identity \cref{grad_est:eq:bartlett_second},
we then obtain
\begin{align*}
\nabla^2 \ell(\thetav; \cdot) 
&= \EE_{Y \sim p_\thetav} 
[\nabla^2 \ell(\thetav; Y)] \\
&= \EE_{Y \sim p_\thetav} 
[-\nabla^2_\thetav \log p_\thetav(Y)] \\
&= \EE_{Y \sim p_\thetav} 
[\nabla_\thetav \log p_\thetav(Y) \otimes \nabla_\thetav \log p_\thetav(Y)] \\
&= \EE_{Y \sim p_\thetav} 
[\nabla \ell(\thetav; Y) \otimes \nabla \ell(\thetav; Y)],
\end{align*}
where we used $\cdot$ to indicate that the result holds for all $\y$.
Plugging the result back in the Fisher information matrix concludes the proof.
\end{proof}

\section{Inverse-Hessian vector product}
\label{higher:sec:ihvp}

\subsection{Definition as a linear map}

We will see in \cref{optim:sec:newton} that Newton's method uses iterations as
\begin{align*}
  \w^{t+1} 
  & = \w^t - \nabla^2 L(\w^t)^{-1}\nabla L(\w^t).
\end{align*}
The inverse is well-defined if for example $L$ is strictly convex. 
Otherwise, some additional regularization can be added.
Newton's method therefore requires access to \textbf{inverse-Hessian vector
products} (IHVPs), as defined below.
\begin{boxdef}{Inverse-Hessian vector product}
For a twice differentiable function $L:\RR^P \rightarrow \RR$, we define the
\textbf{inverse-Hessian Vector Product} (IHVP) of $L$ at $\w \in \RR^P$ as the
linear map
\[
\u \mapsto \nabla^2 L(\w)^{-1} \u,
\]
provided that it exists. In other words, it is the linear map
which to $\u$ associates $\v$ such that $\nabla^2 L(\w) \v = \u$.
\label{higher:def:ihvp}
\end{boxdef}

\subsection{Implementation with matrix-free linear solvers}

Numerous direct methods exist to compute the inverse of a matrix, such as the
Cholesky decomposition, QR decomposition and Gaussian elimination.
However, these algorithms require accessing elementary entries of the matrix,
while an autodiff framework gives access to the Hessian through HVPs.
Fortunately, there exist so-called {\bf matrix-free} algorithms, that can solve
a linear system of equations 
\begin{equation*}
H[\v] = \u 
\end{equation*}
by only accessing the linear map $\v \mapsto H[\v]$ for any $\v$. Among such
algorithms, we have the {\bf conjugate gradient} (CG) method, that applies for
$H$ positive-definite, i.e., such that $\langle \v, H[\v]\rangle >0$ for all $\v
\neq 0$, or the {\bf generalized minimal residual} (GMRES) method, that applies
for any invertible $H$. A longer list of solvers can be found in public software
such as SciPy~\citep{scipy2020}. The IHVP of a strictly convex function
(ensuring that the Hessian is positive definite) can
therefore be computed by instantiating CG on the HVP,
\[
  \nabla^2 L(\w)^{-1} \u \approx \mathrm{CG}(\nabla^2 L(\w)[\cdot], \u).
\]
Positive-definiteness of the Hessian is indeed guaranteed for strictly convex
functions for example, while for generic nonconvex functions, such a property may
be verified around a minimizer but not in general. 
The conjugate gradient method is recalled in
\cref{higher:algo:cg} in its simplest form. 
In theory, the exact solution of the linear system is found after at most $T= P$
iterations of CG, though in practice numerical errors may prevent one from
getting an exact solution.

\begin{algorithm}
  \caption{Conjugate gradient method \label{higher:algo:cg}}
  \begin{algorithmic}[1]
    \Statex {\bf Inputs:} linear map $H[\cdot]: \RR^P \rightarrow \RR^P$, target
    $\u \in \RR^P$, initialization $\v_0$ (default $\zeros$), number of
    iterations $T$ (default $P$), target accuracy $\varepsilon$ (default machine
    precision)

    \State $\r_0 = \u - H[\v_0]$ 

    \State $\p_0 = \r_0$

    \For{$t=0, \ldots, T-1$}

      \State $\alpha_t = \frac{\langle \r_t, \r_t \rangle}{\langle \p_t, H
      [\p_t]\rangle}$

      \State $\v_{t+1} = \v_t + \alpha_t \p_t$

      \State $\r_{t+1} = \r_t - \alpha_t H[\p_t]$

      \State \algorithmicif \ $\langle \r_{t+1},
      \r_{t+1} \rangle \leq \varepsilon$ \algorithmicthen \ {\bf break}

      \State $\beta_t = \frac{\langle \r_{t+1}, \r_{t+1} \rangle}{\langle \r_t,
      \r_t \rangle}$

      \State $\p_{t+1} = \r_{t+1} + \beta_t \p_t$

    \EndFor 

    \Statex {\bf Output:} $\v_T$, such that $H[\v_T] \approx \u$
  \end{algorithmic}
\end{algorithm}

\subsection{Complexity}

  For a given matrix $H \in \RR^{P \times P}$, solving $H \v = \u$ can be done
with decomposition methods (LU, QR, Cholesky) in $O(P^3)$ time. 
For matrix-free methods such as CG or GMRES, the cost per iteration is $O(P^2)$.
Since they theoretically solve the linear system in $O(P)$
iterations, the cost to obtain an exact solution is theoretically the same,
$O(P^3)$.

However, CG or GMRES differ from decomposition methods in that they are
iterative methods, meaning that, at each iteration, they get closer to a
solution.  Unlike decomposition methods, this means that we can stop them before
an exact solution is found. In practice, the number of iterations required to
find a good approximate solution depends on the matrix. Well conditioned
matrices require only few iterations.
Badly conditioned matrices lead to some numerical instabilities for CG,
so that more than $P$ iterations may be needed to get a good solution.
In contrast, decomposition methods proceed in two steps: first they build a
decomposition of $H$ at a cost of $O(P^3)$, and second they solve 
a linear system at a cost of $O(P^2)$, by leveraging the structure. 
LU and QR decompositions are known to be
generally more stable and are therefore often preferred in practice, when we can
access entries of $H$ at no cost.

If we do not have access to the Hessian $H$, but only to its HVP,
accessing entries of $H$ comes at a prohibitive cost. 
Indeed, entries of $H$ can still be recovered from HVPs, since $\e_i^\top H
\e_j = H_{i,j}$, but accessing
each row or column of $H$ costs one HVP (matrix-vector product). To access the
information necessary to use a decomposition method, we therefore need $P$
calls to HVPs before being able to actually compute the
solution. For the same number of calls, CG or GMRES will already have
found an approximate solution. In addition, a CG method does not
require storing the matrix in memory.

\section{Second-order backpropagation}\label{higher:sec:hbp}

\subsection{Second-order Jacobian chain rule}

The essential ingredient to develop forward-mode and reverse-mode autodiff
hinged upon the chain rule for composed functions, $h = g \circ f$. For second
derivatives, a similar rule can be obtained. To do so, we slightly abuse
notation and denote
\[
  \partial^2 h(\w)^*[\u] 
  \coloneqq \nabla^2 \langle h, \u \rangle(\w) \in \RR^{P \times P},
\]
where
$h: \RR^P \rightarrow \RR^Q$, $\w \in \RR^P$, $\u \in \RR^Q$, 
and where we recall the shorthand notation
$\langle \u, h\rangle(\w) \coloneqq \langle \u, h(\w)\rangle$.
Moreover, we view the above quantity as a linear map.
Strictly speaking, the superscript ${}^*$ is not a linear adjoint anymore,
since $\v_1, \v_2 \mapsto \partial^2 h(\w)[\v_1, \v_2]$ is no
longer linear but bilinear. However, this superscript plays the same role as
the VJP, since it takes an output vector and returns the input derivatives that
correspond to infinitesimal variations along that output vector.

\begin{boxprop}{Hessian chain rule}\label{higher:thm:hessian_chain_rule} For two
  twice differentiable functions $f: \RR^P \rightarrow \RR^M$ and $g: \RR^M
  \rightarrow \RR^Q$, the second directional derivative of the composition
  $g \circ f$ is a bilinear map from $\RR^P\times \RR^P$ to $\RR^Q$
  along input directions $\v_1, \v_2 \in \RR^P$ of the form
  \begin{align*}
    \partial^2 (g\circ f)(\w)[\v_1, \v_2]
    & = \partial g(f(\w))[\partial^2 f(\w)[\v_1, \v_2]] \\
    & \quad + \partial^2 g(f(\w))[\partial f(\w)[\v_1], \partial f(\w)[\v_2]].
  \end{align*}
  The Hessian of the composition $g\circ f$ along an output direction $\u \in
  \RR^Q$ is, seen as a linear map, 
  \begin{align}
    \partial^2 (g\circ f)(\w)^*[\u] 
    & = \partial^2 f(\w)^* [\partial g(f(\w))^*[\u]] \label{higher:eq:hbp} \\
    & \quad + \partial f(\w)^* \partial^2 g(f(\w))^*[\u] \partial f(\w). \nonumber
  \end{align}
  For the composition of $f \colon \RR^P \to \RR^M$ with a scalar-valued
  function $\ell \colon \RR^M \to \RR$, 
  we have in matrix form
  \begin{align*}
    \nabla^2 (\ell \circ f)(\w)
    & = \sum_{j=1}^M (\nabla \ell(f(\w)))_j \nabla^2 f_j(\w)\\
    & \quad + \jac f(\w)^\top \nabla^2 \ell(f(\w)) \jac f(\w).
  \end{align*}
\end{boxprop}
Note that, while the Hessian is usually defined for scalar-valued functions
$h \colon \RR^P \to \RR$,
the above definition is for a generalized notion of Hessian
that works for any function $h \colon \RR^P \to \RR^Q$.

The Hessian backpropagation rule in~\cref{higher:eq:hbp} reveals two terms. The
first one $\partial^2 f(\w)^* [\partial g(f(\w))^*[\u]]$ simply computes the
Hessian of the intermediate function along the output direction normally
backpropagated by a VJP. The second term $\partial f(\w)^* \partial^2
g(f(\w))^*[\u] \partial f(\w)$ shows how intermediate first-order
variations influence second-order derivatives of the output. 

\begin{boxexm}{Composition with an elementwise nonlinear function}
  \label{higher:exm:elementwise_hbp}
  Consider the element-wise application of a twice differentiable scalar-valued
  function $f(\x) = (f(x_i))_{i=1}^M$ followed by some twice differentiable
  function $\ell$. Note that $\nabla^2 f_i(\x) = f''(x_i) \e_i\e_i^\top$.
  Hence, the Hessian of the composition reads 
  \begin{align*}
    \nabla^2 (\ell\circ f)(\x)
    & = \sum_{i=1}^M (\nabla \ell(f(\x)))_i f''(x_i) \e_i\e_i^\top \\
    & \quad + \Diag(f'(\x)) \nabla^2 \ell(f(\x)) \Diag(f'(\x)) \\
    & = \Diag(\nabla \ell(f(\x)) \odot f''(\x))\\
    & \quad + \nabla^2 \ell(f(\x)) \odot (f'(\x) f'(\x)^\top),
  \end{align*}
  where 
  $f'(\x) \coloneqq (f'(x_i))_{i=1}^M$ 
  and 
  $f''(\x) \coloneqq (f''(x_i))_{i=1}^M$.
\end{boxexm}

\begin{boxexm}{Hessian of the composition with a linear function}
  \label{higher:exm:linear_hbp}
  Consider a linear function $f(\W) = \W\x$, for $\W\in \RR^{M\times D}$,
  composed with some twice differentiable function 
  $\ell: \RR^M \rightarrow \RR$. 
  From~\cref{higher:thm:hessian_chain_rule}, we get, in terms of linear maps,
  \begin{align*}
    \nabla^2 (\ell\circ f)(\W) = \partial f(\W)^* 
    \nabla^2 \ell(f(\W)) \partial f(\W).
  \end{align*}
  As already noted in \cref{diff:sec:jvp_vjp}, we have that $\partial f(\W)[\V] =
  \V \x$ and $\partial f(\W)^*[\u] = \u \x^\top$. Hence, the 
  Hessian seen as a linear map reads
  \begin{align*}
    \nabla^2 (\ell\circ f)(\W)[\V] 
    & = \partial f(\W)^*[
      \nabla^2 \ell(f(\W)) [\partial f(\W)[\V]]] 
    & = \H \V \x\x^\top,
  \end{align*}
  where $\H \coloneqq \nabla^2 \ell(f(\W))$. 
\end{boxexm}

\subsection{Computation chains}

For a simple computation chain $f = f_K\circ \ldots \circ f_1$ as
in~\cref{auto_diff:sec:chains}, the formula derived
in~\cref{higher:thm:hessian_chain_rule} suffices to develop an algorithm that
backpropagates the Hessian, as shown in \cref{higher:algo:hessian_backprop}.
Compared to \cref{auto_diff:algo:reverse_chain}, we simply backpropagate both
the vectors $\r_k$ and the matrices $\bm{R}_k$ using intermediate first and second
derivatives.

\begin{algorithm}\caption{Hessian backprop for computation chains
\label{higher:algo:hessian_backprop}}
\begin{algorithmic}[1]
    \Statex{\bf Functions:} $\ell \circ f$ with $f \coloneqq f_K \circ \ldots \circ f_1$, 
	\Statex{\bf Inputs:} input $\x$
		\State Initialize and store $\s_0 \coloneqq \x$ \Comment{Forward pass}
        \For {$k\coloneqq 1, \ldots, K$} 
		    \State Compute and store $\s_k \coloneqq f_k(\s_{k-1})$
		\EndFor 
	\State Initialize
    $\r_K \coloneqq \nabla \ell(\s_K), \ \bm{R}_K \coloneqq \nabla^2 \ell(\s_K)$ \Comment{Backward pass}
    \For{$k\coloneqq K, \ldots, 1$} 
		\State Compute $\r_{k-1} \coloneqq \partial f_k(\s_{k-1})^*[\r_k]$
        \State Compute $\bm{R}_{k-1} \coloneqq \partial^2 f_k(\s_{k-1})^*[\r_k] 
        + \partial f_k(\s_{k-1})^* \bm{R}_k \partial f_k(\s_{k-1})$ 
    \State Release $\s_{k-1}$ from memory
	\EndFor
    \Statex{\bf Outputs:} $\ell(f(\x)) = \ell(\s_K)$, 
    $\nabla(\ell \circ f)(\x) = \r_0$,
    $\nabla^2(\ell \circ f)(\x) = \bm{R}_0$
\end{algorithmic}
\end{algorithm}

\subsection{Fan-in and fan-out}

For generic computation graphs (see \cref{auto_diff:sec:graphs}), we
saw that multi-input functions (fan-in) were crucial.
For Hessian backpropagation in computation graphs, we therefore need to develop
a similar formula.
\begin{boxprop}{Hessian chain rule for fan-in} \label{higher:thm:fan_in}
  Consider $n+1$ twice differentiable functions $f_1, \ldots, f_n$ and $g$ with
  $f_i: \RR^P \rightarrow \RR^{M_i}$ and $g: \RR^{M_1} \times \ldots \times
  \RR^{M_n} \rightarrow \RR^Q$. The Hessian of $g \circ f$ for
  $f(\w)= (f_1(\w), \ldots, f_n(\w))$ along an output direction 
  $\u \in \RR^Q$ is given by 
  \begin{align*}
    \partial^2 (g \circ f)(\w)^*[\u] 
    & = \sum_{i=1}^n \partial^2 f_i(\w)^*[ \partial_i g(f(\w))^*[\u]] \\
    & \quad + \sum_{i, j=1}^n 
    \partial f_i(\w)^* 
    \partial^2_{i, j} g(f(\w))^*[\u]
    \partial f_j(\w).
  \end{align*}
\end{boxprop}
The gradient backpropagation expression for fan-in is simple
because the functions $f_i$ are not linked by any path.
In contrast, the Hessian backpropagation involves
cross-product terms \\
$\partial f_i(\w)^* \partial^2_{i, j} g(f(\w))^*[\u] \partial f_j(\w)$ 
for $i \neq j$. 
The nodes associated with the $f_i$ computations cannot be treated independently
anymore.

On the other hand, developing a backpropagation rule for fan-out does not pose
any issue, since each output function can be treated independently.
\begin{boxprop}{Hessian chain rule for fan-out}
  Consider $n+1$ twice differentiable functions $g_1, \ldots, g_n$ and $f$ with
  $g_i: \RR^M \rightarrow \RR^{Q_i}$ and $f: \RR^{P} \rightarrow \RR^M$. The
  Hessian of $g \circ f$ for $g(\y)= (g_1(\y), \ldots, g_n(\y))$ along a
  direction $\u = (\u_1, \ldots, \u_n) \in \RR^{Q_1} \times \ldots \times
  \RR^{Q_n}$ is given by 
  \begin{align*}
    \partial^2 (g \circ f)(\w)^*[\u] 
    & = \sum_{i=1}^n \partial^2 f(\w)^*[ \partial g_i(f(\w))^*[\u_i]] \\
    & \quad + \sum_{i=1}^n 
    \partial f(\w)^* 
    \partial^2 g_i(f(\w))^*[\u_i]
    \partial f(\w).
  \end{align*}
\end{boxprop}

\section{Block diagonal approximations}\label{higher:sec:block_diagonal}

Rather than computing the whole Hessian or Gauss-Newton matrices, we
can consider computing block-diagonal or diagonal approximations, which are
easier to invert. The approximation rules we present in this section build
upon the Hessian chain rule studied in \cref{higher:sec:hbp}.

\subsection{Feedforward networks}

Recall the definition of a feedforward network:
\begin{align*}
    \s_0 &\coloneqq \x \\
    \s_k &\coloneqq f_k(\s_{k-1}, \w_k) ~ \forall k \in \{1, \ldots, K\} \\
    f(\x, \w) &\coloneqq \s_K,
\end{align*}
where $\w \coloneqq (\w_1, \dots, \w_K)$.
Rather than computing the entire Hessian of $\ell \circ f$ \wrt $\w$,
we can compute the Hessians \wrt each set of parameters $\w_k$.
For the case of computation chains,
the Hessian backpropagation recursion we used in 
\cref{higher:algo:hessian_backprop} was
\[
    \bm{R}_{k-1} \coloneqq \partial^2 f_k(\s_{k-1})^*[\r_k] 
    + \partial f_k(\s_{k-1})^* \bm{R}_k \partial f_k(\s_{k-1}).
\]
Extending this recursion to the feedforward network case,
we obtain,
starting from 
$\r_K \coloneqq \nabla \ell(\s_K)$ 
and 
$\bm{R}_K \coloneqq \nabla^2 \ell(\s_K)$,
\begin{align*}
  \r_{k-1} &\coloneqq \partial f_k(\s_{k-1}, \w_k)^*[\r_k] \\
  \begin{pmatrix}
      \bm{R}_{k-1} & \sim \\
      \sim & \bm{H}_k
  \end{pmatrix} & \coloneqq \partial^2 f_k(\s_{k-1}, \w_k)^*[\r_k] \nonumber \\
& \quad + \partial f_k(\s_{k-1}, \w_k)^* \bm{R}_k \partial f_k(\s_{k-1}, \w_k), \nonumber
\end{align*}
where we used $\sim$ to indicate that these blocks are not used.
The Hessians \wrt each set of parameters are then
\begin{align*}
    \bm{R}_0 &= \nabla_{\x\x}^2 (\ell\circ f)(\x, \w) \\
    \H_1 &= \nabla_{\w_1\w_1}^2(\ell\circ f)(\x, \w) \\
        &\vdots \\
    \H_K &= \nabla_{\w_K\w_K}^2(\ell\circ f)(\x, \w).
\end{align*}
The validity of this result stems from the fact that we can view
the Hessian \wrt $\w_k$ as computing the Hessian \wrt $\w_k$ of
\begin{equation*}
\tilde{f}_K \circ \ldots \circ \tilde{f}_{k+1} \circ f_k(\s_{k-1}, \w_k)
\end{equation*}
where $\tilde{f}_i \coloneqq f_i(\cdot, \w_i)$,
for $i \in \{k+1,\dots,K\}$.
As the computations of the
block-wise Hessians share most of the computations, they can be
evaluated in a single backward pass just as the gradients. 
\begin{boxexm}{Block-wise computation of the Gauss-Newton matrix}
  Our blockwise backpropagation scheme can readily be adapted for the
  Gauss-Newton matrix as 
  \begin{align*}
    \begin{pmatrix}
        \bm{R}_{k-1} & \sim \\
      \sim & \G_k
    \end{pmatrix} & 
    \coloneqq \partial f_k(\s_{k-1}, \w_k)^* \bm{R}_k \partial f_k(\s_{k-1}, \w_k), \nonumber
  \end{align*}
  starting from $\bm{R}_K \coloneqq \nabla^2 \ell(\s_K)$. The
  outputs $\bm{R}_0, \G_1, \ldots, \G_K$ give a block-wise approximation of the
  Gauss-Newton matrix. 
  
  Now, consider a simple multilayer perceptron such that 
  \begin{equation*}
  f_k(\s_{k-1}, \w_k) \coloneqq \a(\W_k \s_{k-1})
  \quad \text{with} \quad \w_k \coloneqq \Vec(\W_k)
  \end{equation*}
  Using~\cref{higher:exm:linear_hbp}
  and~\cref{higher:exm:elementwise_hbp} adapted to the Gauss-Newton matrix, we
  can compute the block-wise decomposition of the Gauss-Newton matrix as, for
  $k=K, \ldots, 1$,
  \begin{align*}
      \bm{R}_{k-1} &\coloneqq \W_k^\top \J_k \W_k \\
      \J_k &\coloneqq \bm{R}_k \odot (\a'(\W_k \s_{k-1})\a'(\W_k\s_{k-1})^\top ) \\
    \G_k &\coloneqq \J_k \otimes \s_{k-1} \s_{k-1}^\top
  \end{align*}
  starting from $\bm{R}_K \coloneqq \nabla^2 \ell(\s_K)$. The outputs $\G_1,
  \ldots, \G_K$
  correspond to the block-wise elements of the Gauss-Newton matrix of $f$
  for the vectorized weights $\w_1, \ldots, \w_K$.
  Similar computations were done in KFRA~\citep{botev2017practical} and
  BackPack~\citep{dangel2019backpack}.
\end{boxexm}

\subsection{Computation graphs}

For generic computation graphs, 
consider a function $f(\x, \w)$ defined by,
denoting $i_1, \ldots, i_{p_k} \coloneqq \parent(k)$,
\[
  \s_k \coloneqq f_k(\s_{i_1}, \ldots, \s_{i_{p_k}}) \quad
  \forall k \in \{1, \ldots, K\}
\]
such that $f(\x, \w) = \s_K$, and $k$ is following a topological ordering of
the graph (see~\cref{auto_diff:sec:graphs}). We can consider the following
backpropagation scheme, for $k = K, \ldots, 1$ and $j \in \parent(k)$
\begin{align}\label{higher:eq:hbp_graphs}
  \r_{i_j} & \leftarrow \r_{i_j} + \partial_j f_k(\s_{i_1}, \ldots,
  \s_{i_{p_k}})^*[\r_k] \\
  \bm{R}_{i_j} & \leftarrow \bm{R}_{i_j} + \partial^2_{jj} f_k(\s_{i_1}, \ldots, \s_{i_{p_k}})^*[\r_k] \nonumber\\
  & \quad 
  + \partial_j f_k(\s_{i_1}, \ldots, \s_{i_{p_k}})^* \bm{R}_k \partial_j
  f_k(\s_{i_1}, \ldots, \s_{i_{p_k}}),
\end{align}
starting from $\bm{R}_K \coloneqq \nabla^2 \ell(\s_K)$ and $\r_K \coloneqq \nabla
\ell(\s_K)$. 
Recall that for multiple inputs, the chain rule presented
in~\cref{higher:thm:fan_in} involves the cross-derivatives. For this
reason the backpropagation scheme in~\cref{higher:eq:hbp_graphs} only computes
an approximation. For example, one can verify that
using~\cref{higher:eq:hbp_graphs} to compute the Hessian of $\ell(f_1(\w),
f_2(\w))$
does not provide an exact expression for the Hessian of $f$. 
This scheme is easy to implement and may provide a relevant proxy for the
Hessian.

\section{Diagonal approximations}
\label{higher:sec:diag_approx}

Similarly to the idea of designing a backpropagation scheme that approximates
blocks of the Hessian, we can design a backpropagation scheme that approximates
the diagonal of the Hessian. The approach was originally proposed by
\citet{becker_1988} for feedforward networks, but our exposition, new to our
knowledge, has the benefit that it naturally extends to computational graphs, as
we shall see. 

\subsection{Computation chains}

The idea stems from modifying the Hessian backpropagation rule
in~\cref{higher:thm:hessian_chain_rule} to only keep the diagonal of
the Hessian. Formally, given a matrix $\M \in \RR^{D \times D}$, we denote by
$\diag(\M) = (\M_{ii})_{i=1}^D \in \RR^{D}$ the vector of diagonal entries of
$\M$,
and for a vector $\m\in \RR^D$, we denote $\Diag(\m) = \sum_{i=1}^D
m_i\e_i\e_i^\top$ the diagonal matrix with entries $m_i$. For the 
backpropagation of the Hessian of 
$\ell \circ f_K \circ \ldots \circ f_1$,
we see from~\cref{higher:algo:hessian_backprop}
that $\diag(\H_{k-1})$ can be expressed in terms of $\H_k$ as
\begin{align*}
\diag(\H_{k-1}) = & \diag(\partial^2 f_k(\s_{k-1})^* \r_k) \\
& + \diag(\jac f_k(\s_{k-1})^* \H_k \jac f_k(\s_{k-1})).
\end{align*}
Unfortunately, that recursion needs access to the whole Hessian $\H_k$,
and would therefore be too expensive.
A natural idea is to modify the recursion to approximate $\diag(\H_k)$ by
backpropagating \textbf{vectors}:
\begin{align*}
\d_{k-1}
\coloneqq &\diag(\partial^2 f_k(\s_{k-1})^* \r_k) \\
& + \diag(\jac f_k(\s_{k-1})^* \Diag(\d_k) \jac f_k(\s_{k-1})).
\end{align*}
The diagonal matrix $\Diag(\d_k)$ serves as a surrogate for $\H_k$.
Each iteration of this recursion can be computed in linear time in the output
dimension $D_k$ since
\begin{align*}
d_{k-1,i} =
&\sum_{j=1}^{D_k} r_{k,j} \cdot \partial_{i,i}^2 f_{k,j}(\s_{k-1}) +
\sum_{j=1}^{D_k} d_{k,j} (\partial_i f_{k,j}(\s_{k-1}))^2.
\end{align*}
To initialize the recursion, we can set $\d_K \coloneqq \diag(\nabla^2
\ell(\s_K))$. As an alternative, as proposed by \citet{elsayed_2022}, if $\H_K$
has a simple form, we can use $\nabla^2\ell(\s_K)$ instead of $\Diag(\d_K)$ at
the first iteration. This is the case for instance if $f_K$ is a cross-entropy
loss. The recursion is repeated until we obtain the approximate diagonal Hessian
$\d_0 \approx \diag(\nabla^2 (\ell\circ f)(\x))$. The gradients $\r_k$, needed to
compute $\d_k$, are computed along the way and the algorithm can
therefore also return $\r_0 = \nabla (\ell\circ f)(\x)$.

\subsection{Computation graphs}

Although this diagonal approximation was originally derived for feedforward
networks by \citet{becker_1988}, 
it is straightforward to generalize it to computation graphs.
Namely, for a function $f(\x, \w)$ decomposed along a computation graph, 
we can backpropagate a diagonal
approximation in reverse topological order as
\begin{align}
  \r_{i_j} & \leftarrow \r_{i_j} + \partial_j f_k(\s_{i_1}, \ldots, \s_{i_{p_k}})^*[\r_k]
  \nonumber
  \\
  \d_{i_j} & \leftarrow \d_{i_j} 
  + \diag(\partial^2_{jj} f_k(\s_{i_1}, \ldots, \s_{i_{p_k}})^*[\r_k]) \nonumber\\
  & \quad 
  + \diag(
  \partial_j f_k(\s_{i_1}, \ldots, \s_{i_{p_k}})^* 
  \Diag(\d_k) 
  \partial_j f_k(\s_{i_1}, \ldots, \s_{i_{p_k}})
  ),
    \label{higher:eq:diag_approx_graphs}
\end{align}
for $j \in \parent(k)$, starting from $\r_K = \nabla \ell(\s_K)$ and $\d_K =
\diag(\nabla^2 \ell(\s_K))$ or $\Diag(\d_K) = \nabla^2 \ell(\s_K)$. 
To implement such an algorithm, 
each elementary function in the computational graph
needs to be augmented with an oracle that computes the Hessian diagonal
approximation of the current function, given the previous ones.
An example with MLPs is presented in~\cref{higher:exm:diag_mlp}.

\begin{boxexm}{Hessian diagonal approximation for MLPs
  \label{higher:exm:diag_mlp}}
Consider a multilayer perceptron
\begin{align*}
    \s_{k} &\coloneqq \a_k(\W_k \s_{k-1}) \quad \forall k \in \{1,\dots,K\}\\
    f(\w, \x) &\coloneqq \s_K
\end{align*}
starting from $\s_0 = \x$. Here $\a_k$ is the element-wise activation function
(potentially the identity) and $\w$ encapsulates the weight matrices $\W_1,
\ldots, \W_K$. We consider the derivatives \wrt the flattened matrices, so that
gradients and diagonal approximations \wrt these flattened quantities are
vectors. The backpropagation scheme~\eqref{higher:eq:diag_approx_graphs} then
reduces to, denoting $\t_k = \W_k \s_{k-1}$,
\begin{align*}
  \r_{k-1} &\coloneqq \W_k^\top (\a'(\t_k) \odot \r_k) \\
  \g_k &\coloneqq \Vec((\a'(\t_k) \odot \r_k) \s_{k-1}^\top) \\ 
  \deltav_k &\coloneqq \r_k \odot \a''(\t_k) + \d_k\odot \a'(\t_k)^2 \\
  \d_{k-1} &\coloneqq \left(\sum_{j=1}^{D_k} \W_{k, j i}^2 \delta_{k, j}\right)_{i=1}^{D_{k-1}}  \\
  \h_k &\coloneqq \Vec(\deltav_k (\s_{k-1}^2)^\top)
\end{align*}
starting from $\r_K = \nabla \ell(\s_K)$ and, e.g., $\d_K = \diag(\nabla^2
\ell(\s_K))$. The algorithm returns $\g_1, \ldots, \g_K$ as the gradients of
$f$ \wrt $\w_1, \ldots, \w_K$, with $\w_i = \Vec(\W_i)$, and $\h_1, \ldots,
\h_K$ as the diagonal approximations of the Hessian \wrt $\w_1, \ldots,
\w_K$.
\end{boxexm}
\section{Randomized estimators}

In this section, we describe randomized estimators of the
diagonal of the Hessian or Gauss-Newton matrices. 

\subsection{Girard-Hutchinson estimator}

We begin with a generic estimator, originally proposed
for trace estimation by \citet{girard_1989} and extended by
\citet{hutchinson_1989}. Let $\A \in \RR^{P \times P}$ be an arbitrary square
matrix, whose matrix-vector product (matvec) is available. 
Suppose $\omegav \in \RR^P$ is an isotropic
random vector, i.e., such that $\EE_{\omegav \sim p}[\omegav \omegav^\top] = I$.
For example, two common choices are the Rademacher distribution 
$p = \mathrm{Uniform}(\{-1, 1\}^P)$ 
and the standard normal distribution
$p = \mathrm{Normal}(\zeros, I_P)$.
Then, we have
\begin{equation*}
\EE_{\omegav \sim p}[\langle \omegav, \A \omegav \rangle] = \Tr(\A).
\end{equation*}
Applications include generalized cross-validation,
computing the Kullback-Leibler divergence between two Gaussians,
and computing the derivatives of the log-determinant.

The approach can be extended \citep{bekas_2007,baston_2022,hallman_2023} to
obtain an estimator of the diagonal of $\A$,
\begin{equation*}
\EE_{\omegav \sim p}[\omegav \odot \A \omegav] 
= \diag(\A),
\end{equation*}
where $\odot$ denotes the Hadamard product (element-wise multiplication).
This suggests that we can use the Monte-Carlo method to estimate the diagonal of
$\A$,
\begin{equation*}
\diag(\A) \approx
\frac{1}{S}
\sum_{i=1}^S \omegav_i \odot \A \omegav_i,
\end{equation*}
with equality as $S \to \infty$, since the estimator is unbiased.
Since, as reviewed in \cref{higher:sec:hvp} and \cref{higher:sec:gnvp}, 
we know how to multiply efficiently with the Hessian and the Gauss-Newton
matrices, we can apply the technique with these matrices. The variance is
determined by the number $S$ of samples drawn and therefore by the number of
matvecs performed. More elaborate approaches have been proposed to further
reduce the variance \citep{meyer_2021,xtrace}.

\subsection{Bartlett estimator for the factorization}

Suppose the objective function is of the form 
$L(\w; \x, \y) \coloneqq \ell(f(\w; \x); \y)$
where $\ell$ is the negative log-likelihood
$\ell(\thetav; \y) \coloneqq -\log p_\thetav(\y)$
of an exponential family distribution,
and $\thetav \coloneqq f(\w; \x)$, for some network $f$.
We saw from the equivalence between the Fisher and Gauss-Newton matrices in
\cref{higher:prop:equiv_fisher_gn} (which follows from the Bartlett identity)
that
\begin{align*}
\gn L(\w; \x, \cdot) 
&= \EE_{Y \sim p_\thetav}
[\partial f(\w; \x)^*
\nabla \ell(\thetav; Y) \otimes \nabla \ell(\thetav; Y)
\partial f(\w; \x)] \\
&= \EE_{Y \sim p_\thetav}
[\nabla L(\w; \x, Y) \otimes \nabla L(\w; \x, Y)],
\end{align*}
where $\cdot$ indicates that the result holds for any value of the second
argument. This suggests a Monte-Carlo scheme
\begin{align*}
\gn L(\w; \x, \cdot) 
\approx 
\frac{1}{S} \sum_{j=1}^S [\nabla L(\w; \x, \y_{i_j}) \otimes \nabla L(\w; \x,
\y_{i_j})]
\end{align*}
where $\y_{i_1},\dots,\y_{i_S} \sim p_\thetav$ and $\thetav = f(\w, \x)$.
In words, we can approximate the Gauss-Newton matrix with $S$ gradient
computations. This factorization can also be used to approximate the GNVP in
\cref{higher:eq:gnvp}.

\subsection{Bartlett estimator for the diagonal}

Following a similar approach, we obtain
\begin{equation*}
\diag(\gn L(\w; \x, \cdot))
= \EE_{Y \sim p_\thetav}
[\nabla L(\w; \x, Y) \odot \nabla L(\w; \x, Y)],
\end{equation*}
where $\odot$ indicates the element-wise (Hadamard) product.
Using a Monte-Carlo scheme, 
sampling $\y_{i_1}, \dots, \y_{i_S}$ from $p_\thetav$,
we therefore obtain
\begin{equation*}
\diag(\gn L(\w; \x, \cdot))
\approx
\frac{1}{S}
\sum_{j=1}^S \nabla L(\w; \x, \y_{i_j}) \odot \nabla L(\w; \x, \y_{i_j}),
\end{equation*}
with equality when all labels in the support of $p_\thetav$ have been sampled.
That estimator, used for instance in \citep[Appendix C.1.]{wei_2020},
requires access to \textbf{individual gradients} evaluated at the sampled
labels.
Another possible estimator of the diagonal is given by
\begin{align*}
&\frac{1}{S} \diag(\gn L(\w; \x, \cdot)) \\
=& \EE_{Y_1, \dots, Y_S \sim p_\thetav}
\left[\nabla \frac{1}{S} \sum_{i=1}^S L(\w; \x, Y_i) 
\odot 
\nabla \frac{1}{S} \sum_{i=1}^S L(\w; \x, Y_i)\right].
\end{align*}
Letting $\gamma_i \coloneqq \nabla L(\w; \x, Y_i)$, this follows from
\begin{align*}
\EE\left[\left(\sum_i \gamma_i\right) \odot \left(\sum_j \gamma_j\right)\right]
&= \EE\left[\sum_i \gamma_i \odot \gamma_i 
+ \sum_{i \neq j} \gamma_i \odot \gamma_j\right] \\
&= \EE\left[\sum_i \gamma_i \odot \gamma_i\right]
\end{align*}
where we used that
$\EE[\gamma_i \odot \gamma_j] = \EE[\gamma_i] \odot \EE[\gamma_j] = \zeros$
since $\gamma_i$ and $\gamma_j$ are independent variables for $i \neq j$ 
and have zero mean, from Bartlett's first identity
\cref{grad_est:eq:bartlett_first}.  We can then use the Monte-Carlo method to
obtain
\begin{align*}
\frac{1}{S} \diag(\gn L(\w; \x, \cdot))
\approx
\left(\nabla \frac{1}{S} \sum_{j=1}^S L(\w; \x, \y_{i_j})\right)
\odot
\left(\nabla \frac{1}{S} \sum_{j=1}^S L(\w; \x, \y_{i_j})\right),
\end{align*}
with equality when all labels in the support of $p_\thetav$ have been sampled.
This estimator can be more convenient to implement, 
since it only needs access to
the gradient of the \textbf{averaged} losses.
However, it may suffer from higher variance.
A special case of this estimator is used by \citet{liu_2023}, where they draw
only one $\y$ for each $\x$.

\section{Summary}

\begin{itemize}

\item By using a Hessian chain rule, we can develop a ``Hessian backpropagation''.
While it is reasonably simple for computation chains,
it becomes computationally prohibitive for computation graphs, due to the
cross-product terms occurring with fan-in.

\item A better approach is to use Hessian-vector products (HVPs).
We saw that there are four possible methods to compute HVPs, but the
forward-on-reverse method is a priori the most efficient.
Similarly to computing gradients, computing HVPs is only a constant times
more expensive than evaluating the function itself.

\item The Gauss-Newton matrix associated with the composition $\ell \circ f$
can be seen as an approximation of the Hessian. It is a positive semidefinite
matrix if $\ell$ is convex, and can be used to build a principled quadratic
approximation of a function. It is equivalent to the Fisher information matrix
in the case of exponential families.
Gauss-Newton-vector products can be computed efficiently, like HVPs.

\item We also described other approximations, such as (block) diagonal approximations,
and randomized estimators.

\end{itemize}

%% file: chapters/graphical_models/graphical_models.tex
\chapter{Inference in graphical models as differentiation}\label{chap:gm}

A graphical model specifies how random variables depend on each other
and therefore determines how their joint probability distribution factorizes.
In this chapter, we review key concepts in graphical models and how they
relate to differentiation, drawing in the process analogies with 
computation chains and computation graphs.

\section{Chain rule of probability}
\label{gm:sec:chain_rule_proba}

The chain rule of probability is a fundamental law in probability theory 
for computing the \textbf{joint probability} of events.
In the case of only two events $A_1$ and $A_2$,
it reduces to the \textbf{product rule} 
\begin{equation*}
\PP(A_1 \cap A_2) = \PP(A_2 | A_1) \PP(A_1).
\end{equation*}
For two discrete random variables $S_1$ and $S_2$,
using the events 
$A_1 \coloneqq \{S_1 = \s_1\}$
and
$A_2 \coloneqq \{S_2 = \s_2\}$,
the product rule becomes
\begin{equation*}
\PP(S_1= \s_1, S_2= \s_2) = \PP(S_2=\s_2 | S_1=\s_1) \PP(S_1=\s_1).
\end{equation*}
More generally, using the product rule, we have for $K$ events
\begin{equation*}
\PP\left(A_1 \cap \ldots \cap A_K\right) 
= \PP\left(A_K \mid A_1 \cap \ldots \cap A_{K-1}\right)
\PP\left(A_1 \cap \ldots \cap A_{K-1}\right).
\end{equation*}
Applying the product rule one more time, we have
\begin{equation*}
\PP\left(A_1 \cap \ldots \cap A_{K-1}\right)
=
\PP\left(A_{K-1} \mid A_1 \cap \ldots \cap A_{K-2}\right) 
\PP\left(A_1 \cap \ldots \cap A_{K-2}\right).
\end{equation*}
Repeating the process recursively, we arrive at
the \textbf{chain rule of probability}
\begin{align*}
\PP\left(A_1 \cap \ldots \cap A_K\right) 
&= \prod_{j=1}^K \PP(A_j \mid A_1 \cap \dots \cap A_{j-1})\\
&= \prod_{j=1}^K \PP\left(A_j \,\Bigg|\, \bigcap_{i=1}^{j-1} A_i\right).
\end{align*}
For $K$ discrete random variables $S_j$,
using the events
$A_j \coloneqq \{S_j = \s_j \}$,
the chain rule of probability becomes
\begin{equation*}
\PP\left(S_1 = \s_1, \ldots,S_K = \s_K\right) 
= \prod_{j=1}^K \mathbb P(S_j = \s_j \mid S_1 = \s_1, \dots, S_{j-1}=\s_{j-1}).
\end{equation*}
Importantly, this factorization holds \textbf{without} any independence
assumption on the variables $S_1,\dots,S_K$. 
In other words, the space of probability distributions specified by the joint
probability on the left-hand side, and the space of probability distributions
specified by the product of conditional probabilities on the right-hand side,
are the same.
We can further simplify the
factorization if we make additional conditional independence assumptions.

\section{Conditional independence}

We know that if two events $A$ and $B$ are independent, then
\begin{equation*}
\PP(A|B) = \PP(A).
\end{equation*}
Similarly, if two random variables $S_1$ and $S_2$ are independent, then
\begin{equation*}
\PP(S_2=\s_2|S_1=\s_1) = \PP(S_2=\s_2).
\end{equation*}
More generally, if we work with $K$ variables $S_1, \dots, S_K$,
some variables may depend on each other, while others may not.
To simplify the notation, given a set $\cC$, we define the shorthands
\begin{align*}
    S_\cC &\coloneqq (S_i \colon i \in \cC) \\
    \s_\cC &\coloneqq (\s_i \colon i \in \cC).
\end{align*}
We say that 
a variable $S_j$ is independent of $S_\cD$ \textbf{conditioned} on $S_\cC$, 
with $\cC \cap \cD = \emptyset$, if for any $\s_j, \s_{\cC}, \s_{\cD}$
\begin{equation*}
\PP(S_j = \s_j \mid S_\cC = \s_\cC, S_\cD = \s_\cD) 
= \PP(S_j = \s_j \mid S_\cC = \s_\cC).
\end{equation*}

\section{Inference problems}

\subsection{Joint probability distributions}

We consider a collection of $K$ variables $\s \coloneqq (\s_1, \dots, \s_K)$,
potentially \textbf{ordered} or \textbf{unordered}.
Each $\s$ belongs to the \textbf{Cartesian product}
$\cS \coloneqq \cS_1 \times \dots \times \cS_K$.
Throughout this chapter, we assume that the sets $\cS_k$ are discrete for
concreteness, with $\cS_k \coloneqq \{\v_1, \dots, \v_{M_k}\}$.
Note that because $\cS_k$ is discrete, we can always identify it with
$\{1,\dots,M_k\}$.
A graphical model specifies a \textbf{joint probability distribution} 
\begin{align*}
\PP(S = \s) 
&= \PP(S_1=\s_1,\dots,S_K=\s_K) \\
&= p(\s) \\
&= p(\s_1,\dots,\s_K),
\end{align*}
where $p$ is the probability mass function of the joint probability
distribution.  Summing over the Cartesian product of all possible
configurations, we obtain
\begin{equation*}
\sum_{\s \in \cS} p(\s)
= \sum_{\s_1,\dots,\s_K \in \cS} p(\s_1,\dots,\s_K)
= 1.
\end{equation*}
As we shall see, the graph of a graphical model encodes the \textbf{dependencies}
between the variables $(S_1,\dots,S_K)$ and therefore how their joint
distribution \textbf{factorizes}. 
Given access to a joint probability distribution, there are several
\textbf{inference} problems one typically needs to solve.

\subsection{Likelihood}

A simple task is to compute the \textbf{likelihood} of some observations $\s =
(\s_1,\dots,\s_K)$,
\begin{equation*}
\PP(S_1 = \s_1, \dots, S_K = \s_K)
=
p(\s_1, \dots, \s_K).
\end{equation*}
It is also common to compute the \textbf{log-likelihood},
\begin{equation*}
\log \PP(S_1 = \s_1, \dots, S_K = \s_K)
=
\log p(\s_1, \dots, \s_K).
\end{equation*}

\subsection{Maximum a-posteriori inference} 

Another common task is to compute the most likely configuration,
\begin{equation*}
\argmax_{\s_1 \in \cS_1, \dots, \s_K \in \cS_K}
p(\s_1,\dots,\s_K).
\end{equation*}
This is the \textbf{mode} of the joint probability distribution.
This is also known as maximum a-posteriori (MAP) inference
in the literature \citep{wainwright_2008}.

\subsection{Marginal inference} 

The operation of \textbf{marginalization} consists in summing (or integrating)
over all possible values of a given variable in a joint probability
distribution. This allows us to compute the \textbf{marginal probability} of the
remaining variables. For
instance, we may want to marginalize all variables but $S_k=\s_k$.  To do so, we
define the Cartesian product
\begin{align}
\cC_k(\s_k)
\coloneqq 
\underbrace{\cS_1 \times \dots \times \cS_{k-1}}_{\cA_{k-1}}
\times \{\s_k\} \times 
\underbrace{\cS_{k+1} \times \dots \times \cS_K}_{\cB_{k+1}}.
\label{gm:eq:marginal_set_ck}
\end{align}
Summing over all variables but $S_k$,
we obtain the marginal probability of $S_k=\s_k$ as
\begin{align*}
\PP(S_k = \s_k) 
&= \sum_{\s_1,\dots,\s_K \in \cC_k(\s_k)} p(\s_1,\dots,\s_K) \\
&= 
\sum_{\s_1,\dots,\s_{k-1} \in \cA_{k-1}} 
\sum_{\s_{k+1},\dots,\s_K \in \cB_{k+1}} 
p(\s_1,\dots,\s_K).
\end{align*}
Defining similarly
\begin{align*}
\cC_{k,l}(\s_k,\s_l)
\coloneqq 
\cS_1 \times \dots \times \{\s_k\} \times \dots \times \{\s_l\} \times \dots
\times \cS_K,
\end{align*}
we obtain
\begin{align*}
\PP(S_k = \s_k, S_l=\s_l) 
= \sum_{\s_1,\dots,\s_K \in \cC_{k,l}(\s_k,\s_l)} p(\s_1,\dots,\s_K).
\end{align*}
In particular, we may want to compute the marginal probability of two
consecutive variables, $\PP(S_{k-1} = \s_{k-1}, S_k=\s_k)$.

\subsection{Expectation, convex hull, marginal polytope} 

Another common operation is to compute the expectation
of $\phi(S)$ under a distribution $p$. It is defined by
\begin{align*}
\muv \coloneqq \EE_{S \sim p}[\phi(S)] 
= \sum_{\s \in \cS} p(\s) \phi(\s) \in \cM.
\end{align*}
For the expectation under $p_\thetav$, we write
\begin{align*}
\mu(\thetav) \coloneqq \EE_{S \sim p_\thetav}[\phi(S)] 
= \sum_{\s \in \cS} p_\thetav(\s) \phi(\s) \in \cM.
\end{align*}
In exponential family distributions (\cref{proba_learn:sec:exp_family}), the
function $\phi$ is called a \textbf{statistic}.
It decomposes as
\begin{align*}
\phi(\s) \coloneqq (\phi_\cC(\s_\cC))_{\cC \in C},
\end{align*}
where $\cC \subseteq [K]$.
Intuitively, $\phi(\s)$ can be thought of as an \textbf{encoding} or 
\textbf{embedding} of $\s$ (a
potentially discrete object such as a sequence of integers) in a vector space.
Under this decomposition, we can also compute
\begin{equation*}
\muv_\cC 
\coloneqq \EE_S[\phi_\cC(S_\cC)]
= \sum_{\s \in \cS} p(\s) \phi_\cC(\s_\cC).
\end{equation*}

\subsubsection*{Convex hull}

The mean $\muv$ belongs to the \textbf{convex hull} of 
$\phi(\cS) \coloneqq \{\phi(\s) \colon \s \in \cS\}$,
\begin{equation*}
\cM \coloneqq \conv(\phi(\cS))
\coloneqq 
\left\{ \sum_{\s \in \cS} p(\s) \phi(\s) \colon p \in \cP(\cS) \right\},
\end{equation*}
where $\cP(\cS)$ is the set of all possible probability distributions over
$\cS$.
In other words, $\cM$ is the set of all possible convex combinations of
$\phi(\s)$ for $\s \in \cS$.  The vertices of $\cM$ are all the $\s \in \cS$.

\subsubsection*{Case of binary encodings: the marginal polytope}

In the special case of a discrete set
$\cS_k = \{\v_1,\dots,\v_M\}$ and
of a \textbf{binary encoding} (indicator function) $\phi(\s)$,
the set $\cM$ is called the \textbf{marginal polytope} \citep{wainwright_2008},
because each point $\muv \in \cM$ contains marginal probabilities.
To see why, consider the \textbf{unary} potential
\begin{equation}
[\phi(\s)]_{k,i}
= [\phi_k(\s_k)]_{i}
= \II(\s_k = \v_i) \label{gm:eq:unary_encoding}
\end{equation}
where $\II(p) \coloneqq 1$ if $p$ is true, $0$ otherwise.
We then obtain the marginal probability of $S_k=\v_i$,
\begin{align*}
[\muv]_{k,i} 
&= \EE_S[\phi(S)_{k,i}] \\
&= \EE_{S_k}[\phi_k(S_k)_i] \\
&= \EE_{S_k}[\II(S_k=\v_i)] \\
&= \sum_{\s_k \in \cS_k} \PP(S_k=\s_k)\II(\s_k=\v_i) \\
&= \PP(S_k = \v_i).
\end{align*}
Likewise, consider the \textbf{pairwise} potential
\begin{equation}
[\phi(\s)]_{k,l,i,j}
= [\phi_{k,l}(\s_k, \s_l)]_{i,j}
= \II(\s_k = \v_i, \s_l=\v_j) \label{gm:eq:pairwise_encoding}.
\end{equation}
We then obtain the marginal probability of $S_k=\v_i$ and $S_l=\v_j$,
\begin{align*}
[\muv]_{k,l,i,j} 
&= \EE_S[\phi(S)_{k,l,i,j}] \\
&= \EE_{S_k,S_l}[\phi_{k,l}(S_k,S_l)_{i,j}] \\
&= \EE_{S_k,S_l}[\II(S_k=\v_i,S_l=\v_j)] \\
&= \sum_{\s_k \in \cS_k} \sum_{\s_l \in \cS_l} \PP(S_k=\s_k,S_l=\s_l)
\II(\s_k=\v_i,\s_l=\v_j) \\
&= \PP(S_k=\v_i, S_l = \v_j).
\end{align*}
We can do the same with higher-order potential functions.

\subsection{Complexity of brute force}

Apart from computing the likelihood, which is trivial, computing the
marginal, mode and expectation 
by brute force takes $O(\prod_{k=1}^K |\cS_k|)$ time.
In particular, if $|\cS_k| = M ~\forall k \in [K]$, brute force takes
$O(M^K)$ time.

\section{Markov chains}

In this section, we briefly review Markov chains. Our notation is chosen to
emphasize the analogies with computation chains.

\subsection{The Markov property}

\begin{figure}
  \centering
  \begin{subfigure}[c]{0.35\textwidth}
    \centering
    \includegraphics[width=\linewidth]{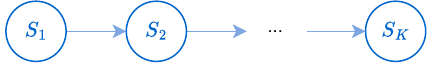}
  \end{subfigure}
  ~
  \begin{subfigure}[c]{0.6\textwidth}
    \centering
    \includegraphics[width=\linewidth]{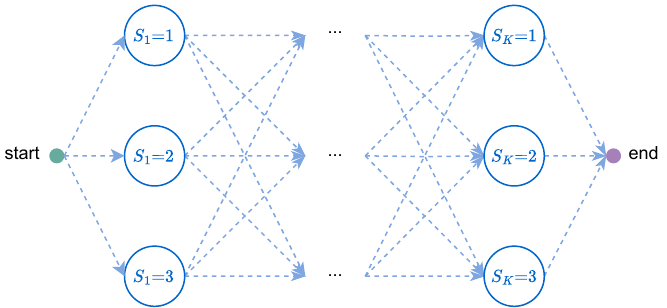}
  \end{subfigure}
  \caption{\textbf{Left:} Markov chain. \textbf{Right:} Computation graph of the
  forward-backward and the Viterbi algorithms: a lattice.
  \label{gm:fig:markov_viterbi}}
\end{figure}

When random variables are organized \textbf{sequentially} as $S_1,\dots,S_K$,
a simple example of conditional independence is when
each variable $S_k \in \cS_k$ only depends on the 
previous variable $S_{k-1} \in \cS_{k-1}$, that is,
\begin{align*}
\PP(S_k=\s_k \mid S_{k-1}=\s_{k-1}, \dots, S_1=\s_1)
&= \PP(S_k=\s_k \mid S_{k-1}=\s_{k-1}) \\
&\coloneqq p_k(\s_k \mid \s_{k-1}).
\end{align*}
A probability distribution satisfying the above
is said to satisfy the \textbf{Markov property},
and is called a \textbf{Markov chain}. 
A computation chain is specified by the functions $f_k$,
that take $\s_{k-1}$ as input and output $\s_k$.
In analogy, a Markov chain is specified by
the \textbf{conditional} probability distributions $p_k$ of
$S_k$ given $S_{k-1}$.
We can then define the \textbf{generative process} 
\begin{align*}
    S_0 &\coloneqq \s_0 \\
    S_1 &\sim p_1(\cdot \mid S_0) \\
    S_2 &\sim p_2(\cdot \mid S_1) \\
                 &\vdots \\
    S_K &\sim p_K(\cdot \mid S_{K-1}).
\end{align*}
Strictly speaking, we should write
$S_k \mid S_{k-1} \sim p_k(\cdot \mid S_{k-1})$.
We choose our notation both for conciseness and for analogy with computation
chains.
Furthermore, to simplify the notation, we assume without loss of generality
that $S_0$ is deterministic (if this is not the case, we can always move $S_0$
to $S_1$ and add a dummy variable as $S_0$). That is,
$\PP(S_0 = \s_0) = p_0(\s_0) \coloneqq 1$ and $\cS_0 \coloneqq \{\s_0\}$.
This amounts to setting the \textbf{initial distribution} of $S_1$ as
\begin{equation*}
\PP(S_1 = \s_1) 
\coloneqq \PP(S_0 = \s_0) \PP(S_1 = \s_1 | S_0 = \s_0)
= \PP(S_1 = \s_1 | S_0 = \s_0).
\end{equation*}
We can then compute the joint probability of the Markov chain by
\begin{align*}
\PP(S_1=\s_1,\dots,S_K=\s_K)
&= p(\s_1,\dots,\s_K) \\
&= \prod_{k=1}^K \PP(S_k = \s_k \mid S_{k-1} = \s_{k-1}) \\
&= \prod_{k=1}^K p_k(\s_k \mid \s_{k-1}),
\end{align*}
where we left the dependence on $\s_0$ implicit, since $p_0(\s_0) = 1$.
A Markov chain with $\cS_k = \{1,2,3\}$ is illustrated
in~\cref{gm:fig:markov_viterbi}.
A chain defines a \textbf{totally ordered set} $\{1, \dots, K\}$,
since two nodes in the graph are necessarily linked to each other by a path.

\begin{boxexm}{Chain of categorical distributions}
Suppose our goal is to predict, from $\x \in \cX$, a sequence of length $K$, 
where each $S_k$ belongs to $\cS_k = \{1, \dots, M\}$.
In natural language processing, this task is called sequence tagging.
We can define
\begin{equation*}
S_k \sim \mathrm{Categorical}(\piv_{k-1,k,S_{k-1}})
\end{equation*}
where 
\begin{align*}
\piv_{k-1,k,i} &\coloneqq \mathrm{softargmax}(\thetav_{k-1,k,i}) 
\in \triangle^M \\
               &= (\pi_{k-1,k,i,j})_{j=1}^M \\
\thetav_{k-1,k,i} &\coloneqq (\theta_{k-1,k,i,j})_{j=1}^M \in \RR^M \\
\theta_{k-1,k,i,j} &\coloneqq f_{k-1,k}(\x, i, j, \w_k) \in \RR.
\end{align*}
We therefore have
\begin{align*}
\PP(S_k = j \mid S_{k-1} = i) 
&= p_k(j \mid i) \\
&= \pi_{k-1,k,i,j} \\
&= \left[\mathrm{softargmax}(\thetav_{k-1,k,i})\right]_j \\
&= \frac{\exp(\theta_{k-1,k,i,j})}{\sum_{j'} \exp(\theta_{k-1,k,i,j'})}
\end{align*}
and
\begin{align*}
\log \PP(S_k = j \mid S_{k-1} = i) 
&= \log p_k(j \mid i) \\
&= \theta_{k-1,k,i,j} - \mathrm{logsumexp}(\thetav_{k-1,k,i}) \\
&= \theta_{k-1,k,i,j} - \log \sum_{j'} \exp(\theta_{k-1,k,i,j'}).
\end{align*}
We emphasize that because $k-1$ and $k$ are always consecutive,
the representation $\theta_{k-1,k,i,j}$ is inefficient; we could use
$\theta_{k,i,j}$ instead. Our notation is designed for consistency with 
Markov random fields.
\end{boxexm}

\subsection{Time-homogeneous Markov chains}

A \textbf{time-homogeneous} discrete-time Markov chain corresponds to the case
when the distribution of $S_k$ given $S_{k-1}$ is the same regardless of $k$:
\begin{equation*}
p_1 = \dots = p_K = p.
\end{equation*}
The \textbf{finite-space} case corresponds to when
each $S_k \in \cS$ can take a finite set of values $\cS = \{\v_1,\dots,\v_M\}$ 
and
\begin{equation*}
    \PP(S_k=\v_j \mid S_{k-1} = \v_i) = p(\v_j|\v_i) = \pi_{i,j},
\end{equation*}
where $\pi_{i,j} \in [0,1]$ is the \textbf{transition probability} 
from $\v_i$ to $\v_j$.
Because the set $\cS = \{\v_1,\dots,\v_M\}$ is discrete,
we can always identify it with $\{1,\dots,M\}$.
That is, we can instead write
\begin{equation*}
    \PP(S_k=j \mid S_{k-1} = i) = p(j|i) = \pi_{i,j}.
\end{equation*}

\subsection{Higher-order Markov chains}
\label{gm:sec:higher_order_Markov_chain}

More generally, a $n^{\text{th}}$-order Markov chain may depend,
not only on the last variable, but on the last $n$ variables,
\begin{align*}
&\PP(S_k=\s_k \mid S_{k-1}=\s_{k-1}, \dots, S_1=\s_1) \\
= &\PP(S_k=\s_k \mid S_{k-1}=\s_{k-1}, \dots, S_{k-n}=\s_{k-n}) \\
= &p_k(\s_k|\s_{k-1},\dots,\s_{k-n}).
\end{align*}
Autoregressive models such as Transformers
(\cref{neural_nets:sec:transformers})
can be seen as specifying a
higher-order Markov chain, with a context window of size $n$.
The larger context makes exact inference using dynamic programming
computationally intractable.
This is why practitioners use \textbf{beam search} or \textbf{ancestral
sampling} (\cref{gm:sec:ancestral_sampling}) instead.

\section{Bayesian networks}
\label{gm:sec:bayesian_net}

In this section, we briefly review Bayesian networks. Our notation is chosen to
emphasize the analogies with computation graphs.

\subsection{Expressing variable dependencies using DAGs}

Markov chains and more generally higher-order Markov chains are a special case
of Bayesian network. Similarly to computation graphs reviewed in
\cref{auto_diff:sec:graphs}, variable dependencies can be expressed 
using a directed acyclic graph (DAG)
$\cG = (\cV, \cE)$, where the vertices $\cV = \{1, \dots, K\}$ represent
variables and edges $\cE$ represent variable dependencies. The set 
$\{i_1, \dots, i_{n_k}\} = \parent(k) \subseteq \cV$, where $n_k \coloneqq
|\parent(k)|$,
indicates the variables $S_{i_1},\dots,S_{i_{n_k}}$ that $S_k$ depends on.
This defines a \textbf{partially ordered set} (poset).
For notational simplicity, we again assume without loss of generality 
that $S_0$ is deterministic.
A computation graph is specified by functions $f_1,\dots,f_K$ in topological
order. In analogy, a \textbf{Bayesian network} is specified by
\textbf{conditional} probability distributions $p_k$ of $S_k$ given
$S_{\parent(k)}$.  We can then define the \textbf{generative process} 
\begin{align*}
    S_0 &\coloneqq \s_0 \\
    S_1 &\sim p_1(\cdot \mid S_0) \\
    S_2 &\sim p_2(\cdot \mid S_{\parent(2)}) \\
        &\vdots \\
    S_K &\sim p_K(\cdot \mid S_{\parent(K)}).
\end{align*}
Using the chain rule of probability and variable independencies expressed by the
DAG, the \textbf{joint probability distribution} is then (assuming a topological
order for $S_0,S_1,\dots, S_K$)
\begin{align*}
\PP(S = \s)
&\coloneqq \PP(S_1=\s_1, \dots, S_K=\s_K) \\
&= \prod_{k=1}^K \PP(S_k = \s_k| S_{\parent(k)} = \s_{\parent(k)}) \\
&\coloneqq \prod_{k=1}^K p_k(\s_k|\s_{\parent(k)}).
\end{align*}
This representation is well suited to express \textbf{causal} relationships
between random variables. 

\subsection{Parameterizing Bayesian networks}

In a Bayesian framework, observed data, latent variables, parameters and noise
variables are all treated as random variables.
If the conditional distribution $p_k$ associated to node $k$
depends on some parameters, they can be provided to $p_k$ as
conditioning, using parent nodes.

A Bayesian network is specified by the conditional distributions
$p_k$. 
Therefore, unlike computation graphs,
there is no notion of function $f_k$ in a Bayesian network. 
However, the root nodes of the Bayesian
network can be the output of a neural network.
For instance, autoregressive models, such as RNNs or Transformers,
specify the conditional probability distribution of a token given past tokens,
and the chain rule of probability is used to obtain a probability distribution
over entire sequences.

\subsection{Ancestral sampling}
\label{gm:sec:ancestral_sampling}

A major advantage of Bayesian networks is that,
provided that each conditional distribution $p_k$ is normalized, 
the joint distribution of $S = (S_1, \dots, S_K)$ 
is automatically normalized.
This means that we can very easily draw \iid samples from the joint
distribution, by following the generative
process: we follow the topological order $k= 1, \dots, K$
and on iteration $k$ we draw a value
$\s_k \sim p_k(\cdot|\s_{\parent(k)})$
conditioned on the previous values $\s_{\parent(k)}$.
This is known as \textbf{ancestral sampling}.

\section{Markov random fields}
\label{gm:sec:mrf}

\subsection{Expressing factors using undirected graphs}

A Markov random field (MRF), \aka undirected graphical model,
specifies a distribution that factorizes as
\begin{equation*}
\PP(S=\s) 
= p(\s)
\coloneqq \frac{1}{Z} \prod_{\cC \in C} \psi_\cC(\s_\cC),
\end{equation*}
where $C$ is the set of maximal \textbf{cliques} of $\cG$, 
that is, subsets of $\cV$ that are fully
connected, $Z$ is a normalization constant defined by
\begin{equation*}
Z \coloneqq \sum_{\s \in \cS} \prod_{\cC \in C} \psi_\cC(\s_\cC),
\end{equation*}
and $\psi_\cC \colon \cS_\cC \to \RR_+$ is a \textbf{potential
function} (\aka compatibility function), with
$\cS_\cC \coloneqq (\cS_j)_{j \in \cC}$. 
According to the Hammersley-Clifford theorem,
an MRF can be equivalently defined in terms of Markov properties;
we refer the interested reader to \citet{wainwright_2008}.
For the sake of this chapter, the definition above is sufficient for our
purposes.

\begin{boxexm}{Markov chains as Markov random fields}
For a chain, letting 
$S = (S_1, \dots, S_K)$ and $\s = (\s_1, \dots, \s_K)$,
recall that
\begin{equation*}
\PP(S=\s) = \prod_{k=1}^K p_k(\s_k \mid \s_{k-1}).
\end{equation*}
This is equivalent to an MRF with
$Z=1$ (since a chain is automatically normalized),
\begin{equation*}
C \coloneqq \{\{0,1\}, \{1,2\}, \dots, \{K-1,K\}\}
\end{equation*}
and with potential function
\begin{equation*}
\psi_{\{k-1,k\}}(\s_{k-1},\s_k) \coloneqq p_k(\s_k|\s_{k-1}).
\end{equation*}
More generally, a Bayesian network can be similarly written as an MRF by
creating appropriate potential functions corresponding to the parents of each
node.
\label{gm:exm:mrf_chain}
\end{boxexm}

\subsection{MRFs as exponential family distributions}

Let us define the potential functions
\begin{equation*}
\psi_\cC(\s_\cC; \thetav_\cC)
\coloneqq
\exp(\langle \thetav_\cC, \phi_\cC(\s_\cC) \rangle)
\end{equation*}
for some sufficient statistic function 
$\phi_\cC \colon \cS_\cC \to \Theta_\cC$
and 
parameters $\thetav_\cC \in \Theta_\cC$.
Then,
\begin{align*}
p_\thetav(\s) 
&\coloneqq \frac{1}{Z(\thetav)} \prod_{\cC \in C} \psi_\cC(\s_\cC; \thetav_\cC) \\
&= \frac{1}{Z(\thetav)} \prod_{\cC \in C} \exp(\langle \thetav_\cC, \phi_\cC(\s_\cC)
\rangle) \\
&= \frac{1}{Z(\thetav)} \exp\left(\sum_{\cC \in C} \langle \thetav_\cC,
    \phi_\cC(\s_\cC)
    \rangle\right) \\
&= \frac{1}{Z(\thetav)} \exp\left(\langle \thetav, \phi(\s) \rangle\right) \\
&= \exp\left(\langle \thetav, \phi(\s) \rangle - A(\thetav)\right)
\end{align*}
where 
\begin{align*}
    \phi(\s) &\coloneqq (\phi_\cC(\s_\cC))_{\cC \in C} \\
    \thetav &\coloneqq (\thetav_\cC)_{\cC \in C} \\
    Z(\thetav) &\coloneqq 
    \sum_{\s \in \cS} \prod_{\cC \in C} \psi_\cC(\s_\cC; \thetav_\cC) \\
               &= \sum_{\s \in \cS}
               \exp\left(\langle \thetav, \phi(\s) \rangle\right) \\
A(\thetav) &\coloneqq \log Z(\thetav).
\end{align*}
Therefore, for this choice of potential functions,
we can view an MRF as an exponential family distribution
(\cref{proba_learn:sec:exp_family})
with \textbf{natural parameters} $\thetav$,
\textbf{sufficient statistic} $\phi$ and
\textbf{log-partition} function $A(\thetav)$.

\begin{boxexm}{Ising model}
The Ising model is a classical example of MRF.
Let $Y = (Y_1, \dots, Y_M) \in \{0,1\}^M$ be an unordered collection of binary
variables $Y_i \in \{0,1\}$.
This forms a graph $\cG = (\cV, \cE)$, where 
$\cV = [M]$ and $\cE \subseteq V^2$, such that $(i,j) \in \cE$ means that $Y_i$
interacts with $Y_j$.
In statistical physics, $Y_i$ may indicate the presence or absence of
particles, or the orientation of magnets. In image processing, $Y_i$ may
represent a black and white pixel. In multi-label classification, $Y_i$ may
indicate the presence or absence of a label.
The probability of $\y = (y_1, \dots, y_M) \in \{0,1\}^M$ is then
\begin{align*}
\PP(Y = \y) 
&= p_\thetav(\y) \\
&= \exp\left(\sum_{i \in \cV} \theta_i y_i 
+ \sum_{(i,j) \in \cE} \theta_{i,j} y_i y_j - A(\thetav) \right) \\
&= \exp\left(\sum_{\cC \in C} \langle \thetav_\cC, \phi_\cC(\y) \rangle - A(\thetav)
\right),
\end{align*}
where 
$C \coloneqq \cV \cup \cE$
and
$\thetav \in \RR^{|\cV| + |\cE|}$ is the concatenation of
$(\theta_i)_{i \in \cV}$
and
$(\theta_{i,j})_{(i,j) \in \cE}$.
These models are also known as \textbf{Boltzmann machines} in a neural network
context.
MAP inference in general Ising models is known to be NP-hard, but when the
interaction weights $\theta_{i,j}$ are non-negative, MAP inference can be
reduced to graph cut algorithms \citep{greig_1989}.
There are two ways the above equation can be extended.
First, we can use higher-order interactions, such as $y_i y_j y_k$ for 
$(i,j,k) \in \cV^3$. Second, we may want to use categorical variables, which
leads to the \textbf{Potts model}.
\end{boxexm}

\subsection{Conditional random fields}

Conditional random fields \citep{lafferty_2001,sutton_2012}
are a special case of Markov random field,
in which a conditioning variable is \textbf{explicitly} incorporated.
For example, when the goal is to predict a variable $\y$ conditioned on a
variable $\x$, CRFs are defined as
\begin{equation*}
\PP(Y=\y \mid X = \x) 
= p(\y \mid \x)
= \frac{1}{Z(\x)} \prod_{\cC \in C} \psi_\cC(\y_\cC, \x).
\end{equation*}
Note that the potential functions $\psi_\cC$
are allowed to depend on the whole $\x$, as
$\x$ is just a conditioning variable.

\subsection{Sampling}

Contrary to Bayesian networks, MRFs require an explicit normalization
constant $Z$. As a result, sampling from a distribution represented by a general
MRF is usually more involved than for Bayesian networks. A commonly-used
technique is \textbf{Gibbs sampling}.

\section{Inference on chains}

In this section, we review how to perform marginal inference and maximum
a-posteriori inference on joint distributions of the form
\begin{equation*}
p(\s_1,\dots,\s_K) = \frac{1}{Z} \prod_{k=1}^K \psi_k(\s_{k-1},\s_k),
\end{equation*}
where
\begin{equation*}
Z \coloneqq \sum_{\s \in \cS} \prod_{k=1}^K \psi_k(\s_{k-1},\s_k)
\end{equation*}
and where we used $\psi_k$ as a shorthand for $\psi_{k-1,k}$, 
since $k-1$ and $k$ are consecutive.
As explained in \cref{gm:exm:mrf_chain}, this also 
includes Markov chains by setting 
\begin{equation*}
\psi_k(\s_{k-1},\s_k) \coloneqq p_k(\s_k \mid \s_{k-1}),
\end{equation*}
in which case $Z=1$.

\subsection{The forward-backward algorithm}
\label{gm:sec:forward_backward_chain}

The key idea of the forward-backward algorithm is to use
the \textbf{distributivity} of multiplication over addition to write
\begin{equation*}
Z = \sum_{\s_1 \in \cS_1} \psi_1(\s_0, \s_1) \sum_{\s_2 \in \cS_2}
\psi_2(\s_1, \s_2) \dots \sum_{\s_K \in \cS_K} \psi_K(\s_{K-1}, \s_K).
\end{equation*}
We can compute these sums recursively, either \textbf{forward} or
\textbf{backward}.
Recalling the definitions of $\cA_{k-1}$ and $\cB_{k+1}$ in
\cref{gm:eq:marginal_set_ck}, we define
the summations \textbf{up to} and \textbf{down to} $k$,
\begin{align*}
\alpha_k(\s_k)
&\coloneqq \sum_{\s_1,\dots,\s_{k-1} \in \cA_{k-1}} 
\prod_{j=1}^k \psi_j(\s_{j-1}, \s_j) \\
&= \sum_{\s_{k-1} \in \cS_{k-1}} \psi_k(\s_{k-1}, \s_k)
\dots
\sum_{\s_1 \in \cS_1} \psi_2(\s_1,\s_2) \psi_1(\s_0,\s_1) \\
\beta_k(\s_k) 
&\coloneqq \sum_{\s_{k+1},\dots,\s_K \in \cB_{k+1}} 
\prod_{j=k+1}^K \psi_j(\s_{j-1}, \s_j) \\
&=
\sum_{\s_{k+1} \in \cS_{k+1}} \psi_{k+1}(\s_k,\s_{k+1})
\dots
\sum_{\s_K \in \cS_K} \psi_K(\s_{K-1},\s_K).
\end{align*}
We can compute the two quantities by recursing forward and backward
\begin{align*}
\alpha_k(\s_k) &= 
\sum_{\s_{k-1} \in \cS_{k-1}} 
\psi_k(\s_{k-1},\s_k) \alpha_{k-1}(\s_{k-1}) \\
\beta_k(\s_k) &= 
\sum_{\s_{k+1} \in \cS_{k+1}} \psi_{k+1}(\s_k,\s_{k+1}) \beta_{k+1}(\s_{k+1})
\end{align*}
where we defined the initializations
\begin{align*}
    \alpha_1(\s_1) &\coloneqq \psi_1(\s_0, \s_1) \quad \forall \s_1 \in \cS_1 \\
    \beta_K(\s_K) &\coloneqq 1 \quad \forall \s_K \in \cS_K.
\end{align*}
The \textbf{normalization} term can then be computed by
\begin{align*}
Z 
= \sum_{\s_K \in \cS_K} \alpha_K(\s_K) \beta_K(\s_K)
= \sum_{\s_1 \in \cS_1} \alpha_1(\s_1) \beta_1(\s_1)
\end{align*}
and the \textbf{marginal probabilities} by
\begin{align*}
\PP(S_k = \s_k) 
&= \frac{1}{Z} \alpha_k(\s_k) \beta_k(\s_k)\\
\PP(S_{k-1}=\s_{k-1}, S_k=\s_k) &= 
\frac{1}{Z}
\alpha_{k-1}(\s_{k-1})
\psi_k(\s_{k-1}, \s_k) 
\beta_k(\s_k).
\end{align*}
We can also compute the conditional probabilities by
\begin{align*}
\PP(S_k=\s_k \mid S_{k-1}=\s_{k-1}) 
&= \frac{\PP(S_{k-1}=\s_{k-1}, S_k=\s_k)}{\PP(S_{k-1}=\s_{k-1})} \\
&= \frac{\alpha_{k-1}(\s_{k-1}) \psi_k(\s_{k-1}, \s_k) \beta_k(\s_k)}{
\alpha_{k-1}(\s_{k-1}) \beta_{k-1}(\s_{k-1})} \\
&= \frac{\psi_k(\s_{k-1}, \s_k) \beta_k(\s_k)}{
\beta_{k-1}(\s_{k-1})}.
\end{align*}
In practice, the two recursions are often implemented in the 
\textbf{log-domain} for numerical stability,
\begin{align*}
\log \alpha_k(\s_k) &= 
\log \sum_{\s_{k-1} \in \cS_{k-1}} 
\exp(\log \psi_k(\s_{k-1},\s_k) + \log \alpha_{k-1}(\s_{k-1})) \\
\log \beta_k(\s_k) &= 
\log \sum_{\s_{k+1} \in \cS_{k+1}}
\exp(\log \psi_{k+1}(\s_k,\s_{k+1}) + \log \beta_{k+1}(\s_{k+1})).
\end{align*}
We recognize the \textbf{log-sum-exp} operator, which can be implemented in a
numerically stable way (\cref{neural_nets:sec:vector_to_scalar}).
The overall \textbf{dynamic programming} procedure, 
\aka \textbf{forward-backward} algorithm
\citep{baum1966statistical,rabiner_1989}, is summarized in
\cref{gm:algo:marginal_chain}.
We notice that the forward and backward passes are actually independent of each
other, and can therefore be performed in parallel.

\begin{algorithm}[t]
\caption{Marginal inference on a chain}
\label{gm:algo:marginal_chain}
\begin{algorithmic}[1]
	\Statex{\bf Potential functions:} $\psi_1, \dots, \psi_K$
	\Statex{\bf Input:} $\s_0$
    \State Initialize $\alpha_1(\s_1) \coloneqq \psi_1(\s_0, \s_1) ~ \forall
    \s_1 \in \cS_1$
    \For {$k\coloneqq 2, \ldots, K$} \Comment{Forward pass}
    \For {$\s_k \in \cS_k$} 
		    \State
$\alpha_k(\s_k) \coloneqq
\displaystyle{\sum_{\s_{k-1} \in \cS_{k-1}} \psi_k(\s_{k-1},\s_k)
\alpha_{k-1}(\s_{k-1})}$
		\EndFor
		\EndFor
    \State Initialize $\beta_{K}(\s_K) \coloneqq 1 ~ \forall \s_K \in \cS_K$
    \For{$k\coloneqq K-1, \ldots, 1$} \Comment{Backward pass}
    \For {$\s_k \in \cS_k$} 
		\State
$\beta_k(\s_k) \coloneqq 
\displaystyle{\sum_{\s_{k+1} \in \cS_{k+1}} \psi_{k+1}(\s_k,\s_{k+1})
\beta_{k+1}(\s_{k+1})}$
	\EndFor
	\EndFor
    \State Compute
$Z 
= \displaystyle{\sum_{\s_K \in \cS_K} \alpha_K(\s_K) \beta_K(\s_K)}
= \displaystyle{\sum_{\s_K \in \cS_K} \alpha_K(\s_K)}$
\Statex{\bf Outputs:} $\forall k \in [K]$:
    \Statex $\PP(S_k = \s_k) = 
    \frac{1}{Z} \alpha_k(\s_k) \beta_k(\s_k)$
    \Statex $\PP(S_{k-1}=\s_{k-1}, S_k=\s_k) = 
\frac{1}{Z} \alpha_{k-1}(\s_{k-1}) \psi_k(\s_{k-1}, \s_k) \beta_k(\s_k)$
\end{algorithmic}
\end{algorithm}

\clearpage

\subsection{The Viterbi algorithm}

Similarly, using the distributivity of multiplication over maximization,
\begin{align*}
 &\max_{\s_1 \in \cS_1, \dots, \s_K \in \cS_K}
\prod_{k=1}^K \psi_k(\s_{k-1},\s_k) \\
=& \max_{\s_K \in \cS_K} 
\max_{\s_{K-1} \in \cS_{K-1}}
\psi_K(\s_{K-1},\s_K)
\dots
\max_{\s_1 \in \cS_1} 
\psi_2(\s_1,\s_2)
\psi_1(\s_0,\s_1).
\end{align*}
Let us define for $k \in [K]$
\begin{equation*}
\delta_k(\s_k) \coloneqq
\max_{\s_{k-1} \in \cS_{k-1}}
\psi_k(\s_{k-1},\s_k)
\dots
\max_{\s_1 \in \cS_1} 
\psi_2(\s_1,\s_2)
\psi_1(\s_0,\s_1).
\end{equation*}
We can compute these quantities recursively, since for $k \in [K]$
\begin{equation*}
\delta_k(\s_k) = 
\max_{\s_{k-1} \in \cS_{k-1}} \psi_k(\s_{k-1},\s_k) \delta_{k-1}(\s_{k-1}),
\end{equation*}
with $\delta_1(\s_1) \coloneqq \psi_1(\s_0, \s_1)$.
We finally have
\begin{equation*}
\max_{\s_1 \in \cS_1, \dots, \s_K \in \cS_K} p(\s_1, \dots, \s_K)
= \frac{1}{Z} \max_{\s_K \in \cS_K} \delta_K(\s_K).
\end{equation*}
In practice, for numerical stability,
we often implement the forward recursion in the \textbf{log-domain}.
Using the fact that the logarithm is monotonic, we indeed have
for all $k \in [K]$
\begin{equation*}
\log \delta_k(\s_k) = 
\max_{\s_{k-1} \in \cS_{k-1}} 
\log \psi_k(\s_{k-1},\s_k) 
+ \log \delta_{k-1}(\s_{k-1}).
\end{equation*}
To enable efficient \textbf{backtracking},
during the forward pass, we compute
\begin{equation*}
q_k(\s_k) \coloneqq
\argmax_{\s_{k-1} \in \cS_{k-1}} \psi_k(\s_{k-1},\s_k) \delta_{k-1}(\s_{k-1})
\end{equation*}
which can be thought of as \textbf{backpointers} 
from $\s_k^\star$ to $\s_{k-1}^\star$.

The resulting dynamic programming procedure, 
\aka \textbf{Viterbi algorithm} \citep{viterbi1967error,forney1973viterbi}, 
is summarized in \cref{gm:algo:map_chain}.

\begin{algorithm}[ht]
\caption{MAP inference on a chain}
\begin{algorithmic}[1]
	\Statex{\bf Potential functions:} $\psi_1, \dots, \psi_K$
	\Statex{\bf Input:} $\s_0$
    \State Initialize $\delta_1(\s_1) \coloneqq \psi_1(\s_0, \s_1) ~ \forall \s_1
    \in \cS_1$
    \For {$k\coloneqq 2, \ldots, K$} \Comment{Forward pass}
        \For {$\s_k \in \cS_k$}
        \State
$\delta_k(\s_k) \coloneqq
\displaystyle{\max_{\s_{k-1} \in \cS_{k-1}} \psi_k(\s_{k-1},\s_k)
\delta_{k-1}(\s_{k-1})}$
        \State
$q_k(\s_k) \coloneqq
\displaystyle{\argmax_{\s_{k-1} \in \cS_{k-1}} \psi_k(\s_{k-1},\s_k)
\delta_{k-1}(\s_{k-1})}$
		\EndFor
		\EndFor
        \State $\delta^\star \coloneqq \displaystyle{\max_{\s_K \in \cS_K}
        \delta_K(\s_K)}$
        \State $\s_K^\star \coloneqq \displaystyle{\argmax_{\s_K \in \cS_K}
        \delta_K(\s_K)}$
    \For{$k\coloneqq K-1, \ldots, 1$} \Comment{Backtracking}
    \State $\s_{k}^\star \coloneqq q_{k+1}(\s_{k+1}^\star)$
	\EndFor
    \Statex{\bf Outputs:} 
    $\displaystyle{\max_{\s_1 \in \cS_1, \dots, \s_K \in \cS_K} p(\s_1, \dots, \s_K) \propto
    \delta^\star}$
\Statex \hspace{4.6em}
$\displaystyle{\argmax_{\s_1 \in \cS_1, \dots, \s_K \in \cS_K} p(\s_1, \dots, \s_K) =
(\s_1^\star, \dots, \s_K^\star)}$
\end{algorithmic}
\label{gm:algo:map_chain}
\end{algorithm}

\clearpage

\section{Inference on trees}

More generally, efficient inference based on dynamic programming
can be performed when dependencies between variables are expressed using
a \textbf{tree} or \textbf{polytree}.  The
resulting marginal inference and MAP inference algorithms are often
referred to as the \textbf{sum-product} and \textbf{max-sum} algorithms.
The sum-product algorithm is also known as \textbf{belief propagation}
or \textbf{message passing}, since it can be interpreted as propagating
``local messages'' through the graph.
See for instance \citep[Section 2.5.1]{wainwright_2008} for more details.

\section{Inference as differentiation}

In this section, we review the profound connections between
differentiating the log-partition function of an exponential family distribution
on one hand, and performing marginal inference (as well as maximum a-posteriori
inference in the zero-temperature limit) on the other hand.

\subsection{Inference as gradient of the log-partition}

We first discuss a well-known fact in the graphical model literature: 
when using a binary encoding as the
sufficient statistic $\phi$ in an exponential family distribution,
the gradient $\nabla A(\thetav)$ of the log-partition 
$A(\thetav)$ gathers all the marginals \citep{wainwright_2008}.

To see why, recall from \cref{proba_learn:sec:exp_family}
the definition of an exponential family distribution
\begin{align*}
p_\thetav(\s) 
&= h(\s) \exp\left[\langle \thetav, \phi(\s) \rangle - 
A(\thetav)\right]
\end{align*}
and of its log-partition
\begin{equation*}
A(\thetav) 
\coloneqq
\log \sum_{\s \in \cS} h(\s) \exp\left[\langle \thetav, \phi(\s) \rangle\right].
\end{equation*}
From \cref{proba_learn:prop:grad_A},
\begin{align*}
\mu(\thetav) 
\coloneqq \nabla A(\thetav)
= \EE_{Y \sim p_\thetav}[\phi(Y)] \in \cM.
\end{align*}
Therefore, with the \textbf{binary encodings} in \cref{gm:eq:unary_encoding}
and \cref{gm:eq:pairwise_encoding},
\begin{align*}
\PP(S_k = \v_i)
&= [\nabla A(\thetav)]_{k,i} \\
\PP(S_k = \v_i, S_l = \v_j)
&= [\nabla A(\thetav)]_{k,l,i,j}.
\end{align*}
Put differently, if we have an efficient algorithm for computing $A(\thetav)$,
we can perform \textbf{reverse-mode autodiff} on $A(\thetav)$ to obtain $\nabla
A(\thetav)$, and therefore obtain the marginal probabilities.  
Following \cref{auto_diff:sec:complexity_dag}, the
complexity of computing all marginal probabilities is therefore roughly the same
as that of computing $A(\thetav)$.

In the special case of chains, we obtain
\begin{align*}
\PP(S_k = \v_i)
&= [\nabla A(\thetav)]_{k,i} 
= \frac{1}{Z} \alpha_k(\v_i) \beta_k(\v_i) \\
\PP(S_{k-1} = \v_i, S_k = \v_j)
&= [\nabla A(\thetav)]_{k-1,k,i,j} 
= \frac{1}{Z} \alpha_{k-1}(\v_i) \psi_k(\v_i, \v_j) \beta_k(\v_j),
\end{align*}
where we left the dependence of $Z$, $\alpha$ and $\beta$ on $\thetav$ implicit.

If we define 
$A_\varepsilon(\thetav) \coloneqq \varepsilon A(\thetav / \varepsilon)$,
in the zero-temperature limit $\varepsilon \to 0$, 
we obtain that $\muv(\thetav)$ is a binary
encoding of the mode, i.e., of the maximum a-posteriori inference solution.

We now show i) how to unify the forward pass of the forward-backward and
Viterbi algorithms using semirings and softmax operators
ii) how to compute the gradient of the log-partition using 
backpropagation.

\subsection{Semirings and softmax operators}

The forward passes in the forward-backward and Viterbi algorithms are clearly
similar.  In fact, they can be formally linked to each other using
\textbf{semirings}.
\begin{boxdef}{Semiring}
A semiring is a set $\KK$ equipped with two binary operations
$(\oplus, \otimes)$ such that
\begin{itemize}
    \item $\oplus$ is associative and commutative,
    \item $\otimes$ is associative and distributive over $\oplus$,
    \item $\oplus$ and $\otimes$ have identity elements $\bar{0}$ and
    $\bar{1}$, respectively.
\end{itemize}
\end{boxdef}
We use the notations $\oplus$, $\otimes$, $\bar{0}$ and $\bar{1}$ to clearly
distinguish them from the classical addition, multiplication, $0$ and $1$.

We recall the following laws for binary operations:
\begin{itemize}
    \item \textbf{Commutativity} of $\oplus$: 
        $a \oplus b = b \oplus a$,
    \item \textbf{Associativity} of $\oplus$: 
        $a \oplus (b \oplus c) = (a \oplus b) \oplus c$,
    \item \textbf{Distributivity} of $\otimes$ over $\oplus$:
        $a \otimes (b \oplus c) = (a \otimes b) \oplus (a \otimes c)$.
\end{itemize}
A set equipped with a binary operation supporting associativity and an identity
element is called a \textbf{monoid}. A monoid such that every element has an
inverse element is called a \textbf{group}. The difference between a ring and
a semiring is that the latter only requires $(\KK, \oplus)$ and $(\KK, \otimes)$
to be monoids, not groups.

Equipped with these definitions, we can interpret the
forward passes in the Viterbi and forward-backward algorithms as follows:
\begin{itemize}
    \item the forward-backward algorithm in the exponential domain uses the
        semiring $\RR_+$ equipped with $(+, \times)$ and identity elements 
        $(0,1)$;
    \item the Viterbi algorithm in the log domain uses the semiring $\RR$
        equipped with $(\max, +)$ and identity elements $(-\infty, 0)$;
    \item the forward-backward algorithm in the log domain uses the semiring 
        $\RR$ equipped with $(\max_\varepsilon, +)$ and identity elements
        $(-\infty, 0)$, where we defined the soft max operator (log-add-exp)
\begin{equation*}
    \text{max}_\varepsilon(a,b) 
\coloneqq \varepsilon \log(\exp(a / \varepsilon) + \exp(b / \varepsilon)),
\end{equation*}
with $\varepsilon \coloneqq 1$ by default.
\end{itemize}
It can be checked that indeed $\text{max}_\varepsilon$ is commutative,
associative, and addition is distributive over $\text{max}_\varepsilon$. Its
identity element is $-\infty$.
By associativity,
\begin{align*}
\text{max}_\varepsilon(a_1,
\text{max}_\varepsilon(a_2, a_3))
&= \mathrm{logsumexp}_\varepsilon(a_1,a_2,a_3) \\
&= \varepsilon \log \sum_i \exp(a_i / \varepsilon).
\end{align*}
In contrast, note that the sparsemax in
\cref{smoothing:sec:max_argmax} is not associative.

Thanks to associativity, we can introduce the shorthand notations
\begin{equation*}
\underset{\v \in \cV}{\text{max}_\varepsilon} ~ f(\v)
\coloneqq \varepsilon \log \sum_{\v \in \cV} \exp(f(\v) / \varepsilon) \in \RR.
\end{equation*}
and
\begin{equation*}
\underset{\v \in \cV}{\text{argmax}_\varepsilon} ~ f(\v)
\coloneqq \left(\exp(f(\v') / \varepsilon) / 
    \sum_{\v \in \cV} \exp(f(\v) / \varepsilon) \right)_{\v' \in \cV}
    \in \cP(\cV).
\end{equation*}

Many algorithms can be generalized thanks to the use of semirings;
see among others \citet{aji2000generalized,mohri2008speech}.
The distributive and associative properties play a key role in breaking down
large problems into smaller ones \citep{verdu1987abstract}.

\subsection{Inference as backpropagation}

In this section, we show that, algorithmically,
backtracking is recovered as a special case of
backpropagation. See also \citep{eisner2016inside,mensch_2018}.

For notation simplicity, we assume 
$\cS_0 = \{1\}$
and
$\cS_k = \{1, \dots, M\}$ for all $k \in [K]$.
We focus on the case
\begin{equation*}
\log \psi_k(i,j) 
= \langle \thetav_k,\phi_k(i,j) \rangle
= \theta_{k,i,j}.
\end{equation*}
We also introduce the shorthands
\begin{align*}
a_{1,j} 
&\coloneqq
\log \alpha_1(j) = \theta_{1,1,j} \\
a_{k,j}
&\coloneqq \log \alpha_k(j) 
= \underset{i \in [M]}{\text{max}_\varepsilon} ~
\theta_{k,i,j} + a_{k-1,i}
\end{align*}
and
\begin{align*}
q_{k,j}
\coloneqq \underset{i \in [M]}{\text{argmax}_\varepsilon} ~
\theta_{k,i,j} + a_{k-1,i}.
\end{align*}

Our goal is to compute the gradient 
\wrt $\thetav \in \RR^{K \times M \times M}$ of
\begin{equation*}
\log Z = A = \underset{j \in [M]}{\text{max}_\varepsilon} ~ a_{K,j}.
\end{equation*}
The soft argmax counterpart of this quantity is
\begin{equation*}
Q \coloneqq
\underset{j \in [M]}{\text{argmax}_\varepsilon} ~ a_{K,j}
\in \triangle^M,
\end{equation*}
where we used $\cP([M]) = \triangle^M$.

Computing the gradient of $A$ is similar to
computing the gradient of a feedforward
network, in the sense that $\thetav_k$ influences not only $a_k$
but also $a_{k+1},\dots,a_K$.
Let us introduce the adjoint variable
\begin{equation*}
r_{k,i} \coloneqq \partialfrac{A}{a_{k,i}},
\end{equation*}
which we initialize as
\begin{equation*}
r_{K,i}
= \partialfrac{A}{a_{K,i}}
= Q_i.
\end{equation*}
Since $\theta_{k,i,j}$ directly influences $a_{k,j}$, 
we have for $k \in [K]$, $i \in [M]$ and $j \in [M]$
\begin{align*}
\mu_{k,i,j}
&\coloneqq \partialfrac{A}{\theta_{k,i,j}} \\
&= \partialfrac{A}{a_{k,j}} \cdot \partialfrac{a_{k,j}}{\theta_{k,i,j}} \\
&= r_{k,j} \cdot q_{k,j,i}.
\end{align*}
Since $a_{k,i}$ directly influences $a_{k+1,j}$ for $j \in [M]$, we have
for $k \in \{1, \dots, K-1\}$ and $i \in [M]$
\begin{align*}
r_{k,i}
= \partialfrac{A}{a_{k,i}}    
&=
\sum_{j \in [M]}
\partialfrac{A}{a_{k+1,j}}    
\partialfrac{a_{k+1,j}}{a_{k,i}} \\
&=
\sum_{j \in [M]} r_{k+1,j} q_{k+1,j,i} \\
&=
\sum_{j \in [M]} \mu_{k+1,i,j}.
\end{align*}

We summarize the procedure in \cref{gm:algo:chain_maxop}.
The forward pass uses the softmax operator $\text{max}_\varepsilon$ and the
softargmax operator $\text{argmax}_\varepsilon$.
In the hard max case, in \cref{gm:algo:map_chain},
we used $q$ to store backpointers from integer to integer.
In the soft max case, in \cref{gm:algo:chain_maxop},
we used $\q$ to store \textbf{soft backpointers}, that is,
discrete probability distributions.
In the zero-temperature limit, backpropagation outputs a binary encoding of the
solution of backtracking.

\begin{algorithm}[t]
\caption{Inference on a chain as backprop with max operators}
\begin{algorithmic}[1]
    \Statex{\bf Input:} $\thetav \in \RR^{K \times M \times M}$
	\Statex{\bf Max operator:} $\max_\varepsilon$
    \State Initialize $a_{1,j} \coloneqq \theta_{1,1,j} ~ \forall j 
    \in [M]$
    \For {$k\coloneqq 2, \ldots, K$} \Comment{Forward pass}
    \For {$j \in [M]$}
        \State
        $a_{k,j} \coloneqq
        \underset{i \in [M]}{\text{max}_\varepsilon} ~ \theta_{k,i,j} +
a_{k-1,i} \in \RR$
        \State
        $\q_{k,j} \coloneqq
        \underset{i \in [M]}{\text{argmax}_\varepsilon} ~
        \theta_{k,i,j} + a_{k-1,i} \in \triangle^M$
		\EndFor
		\EndFor
        \State $A \coloneqq \underset{i \in
        [M]}{\text{max}_\varepsilon} ~ a_{K,i} \in \RR$
        \State $\bm{Q} \coloneqq \underset{i \in
        [M]}{\text{argmax}_\varepsilon} ~ a_{K,i} \in \triangle^M$
        \State Initialize $r_{K,j} = Q_j$ ~ $\forall j \in [K]$
    \For{$k\coloneqq K-1, \ldots, 1$} \Comment{Backward pass}
    \For{$i \in [M]$} 
    \For{$j \in [M]$} 
        \State $\mu_{k+1,i,j} = r_{k+1,j} \cdot q_{k+1,j,i}$
        \State $r_{k,i} \leftarrow r_{k,i} + \mu_{k+1,i,j}$
        \EndFor
    \EndFor
    \EndFor
    \Statex{\bf Outputs:} 
    $\underset{i_1,\dots,i_K \in [M]^K}{\max_\varepsilon} 
    \theta_{1,1,i_1} + \sum_{k=2}^K \theta_{k,i_{k-1},i_k} = A$,
$\nabla A(\thetav) = \mu$
\end{algorithmic}
\label{gm:algo:chain_maxop}
\end{algorithm}

\newpage
\section{Summary}

\begin{itemize}

\item Graphical models represent the conditional dependencies between variables
    and therefore specify how their joint distribution factorizes.

\item There are clear analogies between the worlds of functions and of distributions:
the counterparts of computation chains and computation graphs are Markov chains
and Bayesian networks.

\item Inference on chains and more generally on trees, for exponential family
distributions, is equivalent, both
statistically and algorithmically, to differentiating the log-partition
function.

\item The forward-backward algorithm can be seen as using a sum-product algebra,
while the Viterbi algorithm can be seen as using a max-plus algebra.
Equivalently, in the log domain, we can see the former as using a soft max, and
the latter as using a hard max.

\end{itemize}

%% file: chapters/diff_thru_opt/implicit_diff_main.tex
\chapter{Differentiating through optimization}
\label{chap:imp_diff}

In this chapter, we study how to differentiate through optimization problems,
and more generally through nonlinear systems of equations.

\section{Implicit functions}

Implicit functions are functions
that do not enjoy an explicit decomposition into elementary functions, for which
automatic differentiation, as studied in \cref{chap:auto_diff}, can therefore
not be directly applied. We describe in this chapter techniques to differentiate
through such functions and how to integrate them into an autodiff framework.

Formally, we will denote an implicit function by $\wv^\star(\lambdav)$, 
where $\wv^\star \colon \Lambda \to \cW$. 
One question is then how to compute the Jacobian $\partial \wv^\star(\lambdav)$.
As a first application one can consider \textbf{sensitivity analysis} of a system. 
For example, $\wv^\star(\lambdav)$ could correspond to the equilibrium state of a
physical system and in this case, $\partial \wv^\star(\lambdav)$ would tell us
about the sensitivity of the system to some parameters $\lambdav \in \Lambda$.

\subsection{Optimization problems}

Another example is a function
implicitly defined as the solution (assumed unique) of an optimization problem
\begin{equation*}
\wv^\star(\lambdav) \coloneqq \argmax_{\wv \in \cW} f(\wv, \lambdav),
\end{equation*}
where 
$f \colon \cW \times \Lambda \to \RR$
and
$\cW$ denotes a constraint set. Note that we use an $\argmax$ for
convenience, but the same applies when using an $\argmin$. 

\subsection{Nonlinear equations}

More generally,
$\wv^\star(\lambdav)$ can be defined as the root of some function $F \colon \cW
\times \Lambda \to \cW$, i.e., $\wv^\star(\lambdav)$ is implicitly defined as the
value satisfying the (potentially nonlinear) system of equations
\begin{equation*}
F(\w, \lambdav) = \zeros
\end{equation*}
for all $\lambdav \in \Lambda$.

\subsection{Application to bilevel optimization}

\begin{figure}[t]
  \centering
  \includegraphics[width=0.8\linewidth]{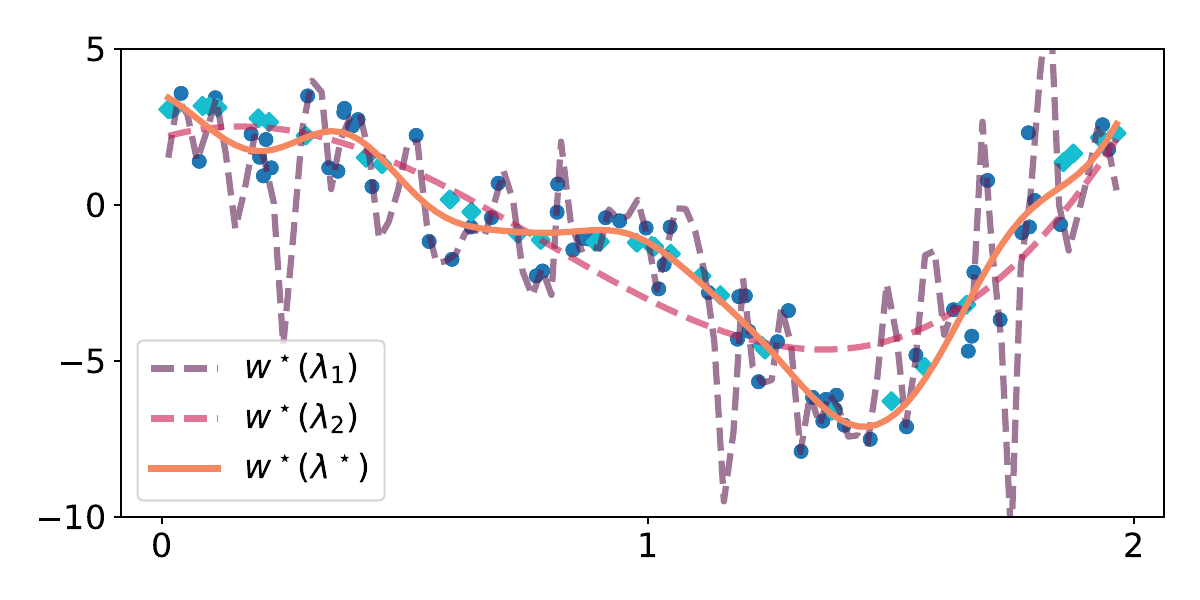}
  \caption{Hyperparameter optimization in nonlinear regression can be cast as a
      bi-level optimization problem. Each line
    corresponds to the estimator obtained by fitting some training data (in blue
    circles) using a different hyperparameter $\lambda$.
    Formally, denoting $f$ the training objective, the estimators are 
    $w^\star(\lambda) \coloneqq \argmin_w f(w; \lambda)$. 
    The goal is to find the best hyperparameter that
    fits some validation data (here in cyan diamonds), that is, minimizing
    $h(\lambda) \coloneqq g(w^\star(\lambda), \lambda)$, where $g$ is the
    validation objective.  A too small $\lambda_1$ leads to
    overfitting the training objective and performs badly on validation
    objective.
    Conversely, a larger $\lambda_2$ underfits both training and
    validation objectives. The optimal parameter $\lambda^\star$ minimizes the validation
    objective and may be obtained by iterating gradient
    descent \wrt $\lambda$. This requires gradients of 
    $h(\lambda) = g(w^\star(\lambda), \lambda)$ w.r.t. $\lambda$.
  \label{imp_diff:fig:bilevel}
  }
\end{figure}

Besides sensitivity analysis,
another example of application is \textbf{bilevel optimization}.
Many times, we want to minimize a function defined as the composition
of a fixed function and the solution of an optimization problem.
Formally, let $f, g \colon \cW \times \Lambda \to \RR$.
We consider the composition $h(\lambdav)$ defined as
\begin{equation}\label{implicit:eq:bilevel}
h(\lambdav) \coloneqq g(\wv^\star(\lambdav), \lambdav),
\quad \text{where} \quad
\wv^\star(\lambdav) \coloneqq \argmax_{\wv \in \cW} f(\wv, \lambdav).
\end{equation}
This includes for instance hyperparameter optimization, where $f$ is an inner
log-likelihood objective, $g$ is an outer validation loss, $\wv \in \cW$ are
model parameters and $\lambdav \in \Lambda$ are model hyperparameters, such as
regularization strength, as illustrated in \cref{imp_diff:fig:bilevel}. To
minimize $h(\lambdav)$, one generally resorts to a gradient descent scheme \wrt
$\lambdav$, which requires computing $\nabla h(\lambdav)$.  Assuming that
$\wv^\star(\lambdav)$ is differentiable at $\lambdav$, by the chain rule, we
obtain the Jacobian
\begin{equation*}
\partial h(\lambdav) = 
\partial_1 g(\wv^\star(\lambdav), \lambdav) \partial \wv^\star(\lambdav)
+ \partial_2 g(\wv^\star(\lambdav), \lambdav).
\end{equation*}
Using $\partial h(\lambdav)^\top = \nabla h(\lambdav)$ (see
\cref{diff:remark:jac_scalar_case}), we obtain the gradient
\begin{equation*}
\nabla h(\lambdav) = 
\partial \wv^\star(\lambdav)^\top \nabla_1 g(\wv^\star(\lambdav), \lambdav) 
+ \nabla_2 g(\wv^\star(\lambdav), \lambdav).
\end{equation*}
The only problematic term is $\partial \wv^\star(\lambdav)$, 
as it requires \textbf{argmax} differentiation.
Indeed, most of the time, there is no explicit formula for $\wv^\star(\lambdav)$
and it does not decompose into elementary functions. 

\section{Envelope theorems}
\label{implicit:sec:danskin}

\begin{figure}
  \centering
  \includegraphics[width=0.7\linewidth]{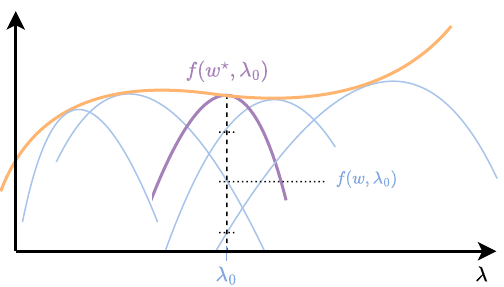}
  \includegraphics[width=0.7\linewidth]{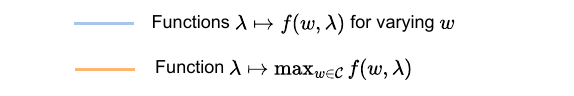}
  \caption{The graph of $h(\lambdav)= \max_{\wv \in \cW} f(\wv, \lambdav)$ is
  the upper-envelope of the graphs of the functions $\lambdav \mapsto f(\wv,
  \lambdav)$ for all $\wv \in \cW$. \label{imp_diff:fig:envelope_thm} }
\end{figure}

In the special case $g = f$, the composition $h$ defined in \cref{implicit:eq:bilevel} 
is simply given by
\begin{equation*}
h(\lambdav) 
= f(\wv^\star(\lambdav), \lambdav)
= \max_{\wv \in \cW} f(\wv, \lambdav).
\end{equation*}
That is, we no longer need \textbf{argmax} differentiation, but only
\textbf{max} differentiation, which, as we shall now see is much easier. The
function $h$ is often called a \textbf{value
function}~\citep{fleming2012deterministic}.
The reason for the name ``envelope'' is illustrated in 
\cref{imp_diff:fig:envelope_thm}.
We emphasize that there is not one, but several envelope theorems, depending
on the assumptions on $f$.

\subsection{Bertsekas' theorem}

The following theorem is due to \citet[Proposition B.25]{bertsekas1997nonlinear}.
\begin{boxthm}{Bertsekas' theorem}
\label{thm:bertsekas_env_theorem}
Let
\begin{equation*}
h(\lambdav) 
\coloneqq \max_{\wv \in \cW} f(\wv, \lambdav)
\end{equation*}
and
\begin{equation*}
\wv^\star(\lambdav)
\coloneqq \argmax_{\wv \in \cW} f(\wv, \lambdav),
\end{equation*}
where  $f \colon \cW \times \Lambda \to \R$ is \textbf{concave} in $\w$ and
$\cW$ is a compact convex set. 
If the maximum $\wv^\star(\lambdav)$ is unique, then the function $h$ is
differentiable with gradient
\begin{equation*}
\nabla h(\lambdav) = \nabla_2 f(\wv^\star(\lambdav), \lambdav).
\end{equation*}
If furthermore $f$ is convex in $\lambdav$, then $h(\lambdav)$ is convex.
\end{boxthm}
A variant of the theorem can be stated when $\cW = \RR^P$.
\begin{boxthm}{Unbounded case}
\label{thm:unbounded_env_theorem}
Suppose $\cW = \RR^P$.
If $f$ is \textbf{strongly concave} in $\wv$,
then the function $h$ is differentiable
with gradient
\begin{equation*}
\nabla h(\lambdav) = \nabla_2 f(\wv^\star(\lambdav), \lambdav).
\end{equation*}
If furthermore $f$ is convex in $\lambdav$, then $h(\lambdav)$ is convex.
\end{boxthm}
Informally, \cref{thm:bertsekas_env_theorem} and
\cref{thm:unbounded_env_theorem} mean that we can treat $\wv^\star(\lambdav)$ as
if it were a constant of $\lambdav$. In other words, we do not need to
differentiate through $\wv^\star(\lambdav)$, even though it depends on
$\lambdav$. 

\begin{boxrem}{Historical note}
Bertsekas' theorem or its unbounded variant are often incorrectly cited as
Danskin's theorem in the machine learning literature, 
because Bertsekas himself called it that way 
\citep[Proposition B.25]{bertsekas1997nonlinear}.  However, because
the assumptions differ, they should rather be thought as variants or corollaries
of Danskin's original theorem.
See \cref{imp_diff:sec:danskin} for more details.
\end{boxrem}

We now illustrate \cref{thm:unbounded_env_theorem}, 
to differentiate through a minimum, 
$h(\lambdav) = \min_{\wv \in \cW} f(\wv, \lambdav)$,
assuming $f(\wv, \lambdav)$ is strongly convex in $\wv$.
\begin{boxexm}{Differentiating the minimum of a convex quadratic}
Let us define $h(\lambda) \coloneqq \min_{w \in \R} f(w, \lambda)$,
where $f(w, \lambda) \coloneqq \frac{\lambda}{2} w^2 + b w + c$ and $\lambda > 0$. 
The derivative of $f$ \wrt $w$ is 
$\nabla_1 f(w, \lambda) = \lambda w + b$. 
The second derivative of $f$ \wrt $w$ is
$\nabla_1^2 f(w, \lambda) = \lambda$. 
Therefore $f$ is strongly convex in $w$.
Let $w^\star(\lambda)$ be the minimum of $f$ in $w$.
Setting $\nabla_1 f(w, \lambda)$ to zero,
we get $w^\star(\lambda) = -\frac{b}{\lambda}$.
The derivative of $f$ w.r.t. $\lambda$ is 
$\nabla_2 f(w, \lambda) = \frac{1}{2} w^2$.
From \cref{thm:unbounded_env_theorem}, we have
$h'(\lambda) 
= \nabla_2 f(w^\star(\lambda), \lambda)
= \frac{1}{2} (w^\star(\lambda))^2
= \frac{1}{2} \frac{b^2}{\lambda^2}$.
Let us check that the derivative is indeed correct.
Plugging $w^\star(\lambda)$ back into $f(w, \lambda)$, we get the analytical
formula 
$h(\lambda) = -\frac{1}{2} \frac{b^2}{\lambda} + c$. 
Using $(\frac{1}{\lambda})' = -\frac{1}{\lambda^2}$, 
we indeed obtain the same expression for $h'(\lambda)$.
\end{boxexm}
\cref{thm:bertsekas_env_theorem} has a simple interpretation for functions that
are affine in $\lambdav$ as shown below.
\begin{boxexm}{Convex conjugate}
\label{diff_opt:exm:conjugate_danskin}
Let $f(\wv, \lambdav) \coloneqq \langle \wv, \lambdav \rangle - \Omega(\wv)$ with $\Omega$ convex.
We then have $h(\lambdav) = \max_{\wv \in \cW} ~ \langle \wv, \lambdav \rangle
- \Omega(\wv) \eqqcolon \Omega^*(\lambdav)$, where $\Omega^*$ denotes the convex
conjugate of $\Omega$. Since $f$ is convex and assuming $\cW$ is convex compact,
the conditions of \cref{thm:bertsekas_env_theorem} are met.
Since $\nabla_2 f(\wv, \lambdav) = \wv$, 
we obtain $\nabla h(\lambdav) = \nabla \Omega^*(\lambdav) =
\wv^\star(\lambdav)$. In other words, in this special case, 
the gradient of the max is equal to the argmax. This is due to the fact that
$f(\wv, \lambdav)$ is affine in $\lambdav$.
\end{boxexm}
Another application is saddle point optimization.
\begin{boxexm}{Saddle point problem}
Consider the saddle point problem 
$\min_{\lambdav \in \Lambda} \max_{\wv \in \cW} f(\wv, \lambdav)$.
If it is difficult to minimize w.r.t. $\lambdav$ but easy to maximize w.r.t.
$\wv$, we can rewrite the problem as
$\min_{\lambdav \in \Lambda} h(\lambdav)$,
where
$h(\lambdav) \coloneqq \max_{\wv \in \cW} f(\wv, \lambdav)$,
and use $\nabla h(\lambdav)$ to perform
(projected) gradient descent w.r.t. $\lambdav$.
\end{boxexm}

\subsection{Rockafellar's theorem}

A related theorem \citep[Theorem 10.31]{rockafellar_2009} 
can be proved under different assumptions on $f$.
\begin{boxthm}{Rockafellar's envelope theorem}
\label{imp_diff:thm:envelope}
Let $f \colon \cW \times \Lambda \to \R$ and $\cW$ be a compact convex set. 
Let
\begin{equation*}
h(\lambdav) 
\coloneqq \max_{\wv \in \cW} f(\wv, \lambdav)
\end{equation*}
and
\begin{equation*}
\wv^\star(\lambdav)
\coloneqq \argmax_{\wv \in \cW} f(\wv, \lambdav).
\end{equation*}
If $f$ is continuously differentiable in $\lambdav$ for all 
$\wv \in \cW$, $\nabla_1 f$ is continuous and 
the maximum $\wv^\star(\lambdav)$ is unique,
then the function $h$ is differentiable with gradient 
\begin{equation*}
\nabla h(\lambdav) = \nabla_2 f(\wv^\star(\lambdav), \lambdav).
\end{equation*}
\end{boxthm}
Compared to Bertsekas' theorem, Rockafellar's theorem does not require $f$ to be
concave in $\w$, but it requires stronger assumptions on the differentiability
of $f$.

\subsection{Danskin's theorem}
\label{imp_diff:sec:danskin}

Finally, we state Danskin's original theorem \citep{danskin1966theory}.
\begin{boxthm}{Danskin's theorem}
\label{thm:danskin_original}
Let $\cW$ be a compact topological space. Let $f \colon \cW \times \Lambda \to
\R$ be continuous and continuously differentiable with respect to $\lambdav \in
\Lambda$.
Let the value function and the set of maximizers be defined as
\begin{equation*}
h(\lambdav) \coloneqq \max_{\wv \in \cW} f(\wv, \lambdav)
\quad \text{and} \quad
\cW^\star(\lambdav) \coloneqq \argmax_{\wv \in \cW} f(\wv, \lambdav).
\end{equation*}
Then, the directional derivative of $h$ at $\lambdav \in \Lambda$ in direction
$\dv \in \Lambda$,
denoted $h'(\lambdav; \dv)$, exists and is given by
\begin{equation*}
h'(\lambdav; \dv) 
= \max_{\wv \in \cW^\star(\lambdav)} \langle \nabla_2 f(\wv, \lambdav), \dv \rangle.
\end{equation*}
\end{boxthm}
We sum up envelope theorems and their assumptions in
\cref{tab:danskin_comparison}.

\begin{table}[t]
\centering
\caption{Comparison of different envelope theorems.}
\label{tab:danskin_comparison}
\begin{small}
\begin{tabular}{lcccc}
\toprule
Theorem & Unique $\wv^\star$? & Concave in $\w$? & Smooth in $\wv$? & Result \\
\midrule
Danskin & No & No & No & Dir. Deriv. \\
Bertsekas & Yes & Yes & No & Gradient \\
Rockafellar & Yes & No & Yes & Gradient \\
\bottomrule
\end{tabular}
\end{small}
\end{table}

\section{Implicit function theorem}
\label{imp_diff:sec:ift}

\subsection{Univariate functions}

The implicit function theorem (IFT) provides conditions under which
an implicit relationship of the form $F(w, \lambda) = 0$ can be rewritten
as a function $w = w^\star(\lambda)$ locally, and provides a way to compute its
derivative w.r.t. $\lambda$.

\begin{boxthm}{Implicit function theorem, univariate case}

Let\\$F \colon \R \times \R \to \R$.
Assume $F(w, \lambda)$ is a continuously differentiable function in a 
neighborhood $\cU$ of $(w_0, \lambda_0)$ such that
$F(w_0, \lambda_0) = 0$
and
$\partial_1 F(w_0, \lambda_0) \neq 0$.
Then there exists a neighborhood $\cV \subseteq \cU$ of $(w_0, \lambda_0)$ in
which there is a function $w^\star(\lambda)$ such that
\begin{itemize}
\item $w^\star(\lambda_0) = w_0$,

\item $F(w^\star(\lambda), \lambda) = 0$ for all $\lambda$ in the neighborhood
    $\cV$,

\item $\partial w^\star(\lambda) = -\frac{\partial_2 F(w^\star(\lambda), \lambda)}{
\partial_1 F(w^\star(\lambda), \lambda)}$.

\end{itemize}
\end{boxthm}
We postpone the proof to the multivariate case and begin with a classical
example of application of the theorem.
\begin{figure}[t]
  \centering
  \includegraphics[width=0.5\linewidth]{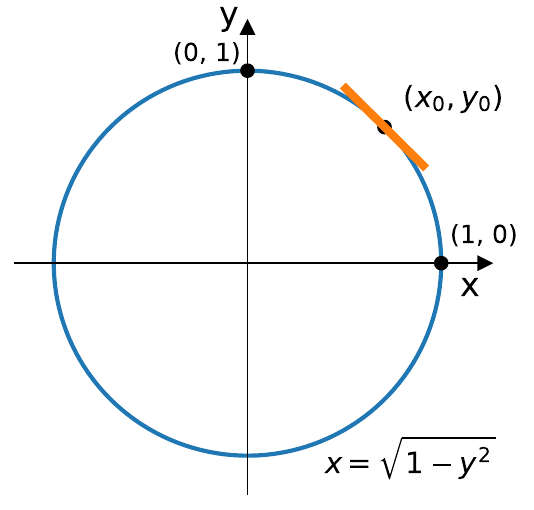}
  \caption{
The circle equation $F(x, y) \coloneqq x^2 + y^2 - 1 = 0$ is not a function from
$y \in [-1, 1]$ to $x \in [-1, 1]$, as there are always two possible $x$ values,
$x = \sqrt{1 - y^2}$ or $x = -\sqrt{1 - y^2}$.
However, locally around some
point $(x_0, y_0)$, e.g., such that $x_0 > 0$ and $y_0 > 0$ (upper-right
quadrant), the function $x = x^\star(y) = \sqrt{1 - y^2}$ is well-defined.
The implicit function theorem gives conditions for such a function to exist
locally and provides a way to compute its derivative.
  \label{imp_diff:fig:ift_circle}
  }
\end{figure}
\begin{boxexm}{Equation of the unit circle}
We use $w \equiv x$ and $\lambda \equiv y$ for clarity.
Let $F(x, y) \coloneqq x^2 + y^2 - 1$. In general, we cannot rewrite the unit
circle equation $F(x, y) = 0$ as a function from $y$ to $x$, because for every
$y \in [-1, 1]$, there are always two possible $x$ values, namely,
$x = \sqrt{1 - y^2}$ or $x = -\sqrt{1 - y^2}$. 
However, locally around some
point $(x_0, y_0)$, e.g., such that $x_0 > 0$ and $y_0 > 0$ (upper-right
quadrant), the function $x = x^\star(y) = \sqrt{1 - y^2}$ is well-defined.
Using $\partial_1 F(x, y) = 2x$ and $\partial_2 F(x, y) = 2y$,
we get 
$\partial x^\star(y) 
= -\frac{\partial_2 F(x^\star(y), y)}{\partial_1 F(x^\star(y), y)}
= - \frac{2y}{2 x^\star(y)} 
= -\frac{y}{\sqrt{1 - y^2}}$
in that neighborhood (the upper right quadrant in this case).
This is indeed the same derivative expression as if we used the chain rule on 
$\sqrt{1 - y^2}$ and is well-defined on $y \in [0, 1)$.
\end{boxexm}
In the above simple example, we can easily derive an explicit function relating
$y$ to $x$ in a given neighborhood, but this is not always the case. 
The IFT gives us conditions guaranteeing that such function \textbf{exists}
and a way to \textbf{differentiate} it, but not a way to \textbf{construct} such
a function.  In fact, finding $w^\star(\lambda)$ such that $F(w^\star(\lambda),
\lambda) = 0$ typically involves a root finding algorithm, an optimization
algorithm, a nonlinear system solver, etc.
\begin{boxexm}{Polynomial}
Let $F(w, \lambda) = w^5 + w^3 + w - \lambda$.
According to the Abel-Ruffini theorem~\citep{tignol2015galois}, quintics
(polynomials of degree $5$) do not enjoy roots in terms of radicals and one must
resort to numerical root finding.
In addition, odd-degree polynomials have real roots. Moreover, $\partial_1 F(w,
\lambda) =
5w^4 + 3w^2 + 1$ is strictly positive. Therefore, by the intermediate value
theorem, there must be only one root $w^\star(\lambda)$ such that
$F(w^\star(\lambda), \lambda) = 0$. This unique root can for example be found by
bisection. Using the IFT, its derivative is found to be
$\partial w^\star(\lambda) = (5w^\star(\lambda)^4 + 3 w^\star(\lambda)^2 +
1)^{-1}$.
\end{boxexm}
While an implicit function is differentiable at a point if the assumptions of
the IFT hold in a neighborhood of that point, the reciprocal is not true:
failure of the IFT assumptions does not necessarily mean that the implicit
function is not differentiable, as we now illustrate.
\begin{boxexm}{IFT conditions are not necessary for differentiability}
Consider $F(w, \lambda) = (w - \lambda)^2$. 
We clearly have that $F(w^\star(\lambda), \lambda) = 0$ if
we define $w^\star(\lambda) = \lambda$, the identity function. 
It is clearly differentiable for all $\lambda$, yet the assumptions of the IFT
fail, since we have $\partial_1 F(w, \lambda) = 2(w-\lambda)$ and 
therefore $\partial_1 F(0, 0) = 0$.
\end{boxexm}

\subsection{Multivariate functions}

We now present the IFT in the general multivariate setting.
Informally, if $F(\w^\star(\lambdav), \lambdav) = \zeros$,
then by the chain rule, we have
\begin{equation*}
\partial_1 F(\w^\star(\lambdav), \lambdav) \partial \w^\star(\lambdav) +
\partial_2 F(\w^\star(\lambdav), \lambdav) = \zeros,
\end{equation*}
meaning that the Jacobian $\partial \w^\star(\lambdav)$, assuming that it
exists, satisfies
\begin{equation*}
-\partial_1 F(\w^\star(\lambdav), \lambdav) \partial \w^\star(\lambdav) =
\partial_2 F(\w^\star(\lambdav), \lambdav).
\end{equation*}
The IFT gives us conditions for the existence of $\partial \w^\star(\lambdav)$.
\begin{boxthm}{Implicit function theorem, multivariate case}\label{implicit:thm:imp_diff}

Let us define $F \colon \cW \times \Lambda \to \cW$.
Assume $F(\wv, \lambdav)$ is a continuously differentiable function in a 
neighborhood of $(\wv_0, \lambdav_0)$ such that
$F(\wv_0, \lambdav_0) = \zeros$
and
$\partial_1 F(\wv_0, \lambdav_0)$ is invertible, i.e.,
its determinant is nonzero.
Then there exists a neighborhood of $\lambdav_0$ in which
there is a function $\wv^\star(\lambdav)$ 
such that
\begin{itemize}
\item $\wv^\star(\lambdav_0) = \wv_0$,

\item $F(\wv^\star(\lambdav), \lambdav) = \zeros$ for all $\lambdav$ in the
    neighborhood,

\item $-\partial_1 F(\wv^\star(\lambdav), \lambdav)  \partial \wv^\star(\lambdav) =  \partial_2 F(\wv^\star(\lambdav),\lambdav)$

$ \iff \partial \wv^\star(\lambdav) 
= -\partial_1 F(\wv^\star(\lambdav), \lambdav)^{-1} \partial_2 F(\wv^\star(\lambdav),
\lambdav). $
\end{itemize}
\label{implicit:thm:ift_multivariate}
\end{boxthm}

We begin with a simple unconstrained optimization algorithm.
\begin{boxexm}{Unconstrained optimization}
Assume we want to differentiate through 
$\wv^\star(\lambdav) = \argmin_{\wv \in \R^P} f(\wv, \lambdav)$,
where $f$ is strictly convex in $\wv$, which ensures that the solution is unique.
From the stationary conditions,
if we define $F(\wv, \lambdav) \coloneqq \nabla_1 f(\wv, \lambdav)$,
then $\wv^\star(\lambdav)$ is uniquely characterized as the root of $F$ in the
first argument, i.e., $F(\wv^\star(\lambdav), \lambdav) = \zeros$.
We have $\partial_1 F(\wv, \lambdav) = \nabla^2_1 f(\wv, \lambdav)$, the Hessian
of $f$ in $\wv$, and $\partial_2 F(\wv, \lambdav) = \partial_2 \nabla_1 f(\wv,
\lambdav)$, 
the cross derivatives of $f$ in $\wv$ and $\lambdav$.
Therefore, assuming that the Hessian is well-defined and
invertible at $(\wv^\star(\lambdav), \lambdav)$, we can use the IFT to differentiate
through $\wv^\star(\lambdav)$ and obtain 
$\partial \wv^\star(\lambdav) = -(\nabla^2_1 f(\wv^\star(\lambdav),
\lambdav))^{-1}
 \partial_2 \nabla_1 f(\wv^\star(\lambdav), \lambdav)$.
\end{boxexm}
Next, we generalize the previous example, by allowing constraints in the
optimization problem.
\begin{boxexm}{Constrained optimization}
Now, assume we want to differentiate through $\wv^\star(\lambdav) = \argmin_{\wv
\in \cC} f(\wv, \lambdav)$, where $f$ is strictly convex in $\wv$ and $\cC
\subseteq \cW$ is a
convex set. A solution is characterized by the fixed point equation
$\wv^\star(\lambdav) = P_\cC(\wv^\star(\lambdav) - \eta \nabla_1
f(\wv^\star(\lambdav), \lambdav))$, for any $\eta > 0$, where $P(\yv) \coloneqq
\argmin_{\xv \in \cC} \|\xv - \yv\|_2^2$ is the Euclidean projection of $\yv$
onto $\cC$. Therefore, $\wv^\star(\lambdav)$ is the root of $F(\wv, \lambdav) =
\wv - P_\cC(\wv - \eta \nabla_1 f(\wv, \lambdav))$ (see \cref{chap:optim}). We
can differentiate through $\wv^\star(\lambdav)$ using the IFT, assuming that the
conditions of the theorem apply. Note that $\partial_1 F(\wv, \lambdav)$
requires the expression of the Jacobian $\partial P_\cC(\yv)$. Fortunately,
$P_\cC(\yv)$ and its Jacobian are easy to compute for many sets $\cC$
\citep{blondel_implicit_diff}.
\end{boxexm}

\subsection{JVP and VJP of implicit functions}

To integrate an implicit function $\w^\star(\lambdav)$ in an autodiff framework,
we need to be able to compute its JVP or VJP. This is the purpose of the next
proposition.
\begin{boxprop}{JVP and VJP of implicit functions}
\label{imp_diff:prop:jvp_vjp_implicit_function}
Let
$\wv^\star \colon \Lambda \to \cW$
be a function implicitly defined as the solution of
$F(\wv^\star(\lambdav), \lambdav) = \zeros$,
for some function $F \colon \cW \times \Lambda \to \cW$.
Define
\begin{align*}
  A & \coloneqq -\partial_1 F(\w^\star(\lambdav), \lambdav) \\
  B & \coloneqq \partial_2 F(\w^\star(\lambdav),\lambdav).
\end{align*}
Assume the assumptions of the IFT hold.
The JVP $\t \coloneqq \partial \wv^\star(\lambdav) \v$ in the input direction
$\v \in \Lambda$ is obtained by solving the linear system
\begin{equation*}
  A \t = B \v.
\end{equation*}
The VJP $\partial \wv^\star(\lambdav)^* \u$ in the output direction $\u \in \cW$
is obtained by solving the linear system
\begin{equation*}
  A^* \r = \u.
\end{equation*}
Using the solution $\r$, we get
\[
\partial \wv^\star(\lambdav)^* \u 
= \partial \wv^\star(\lambdav)^* A^* \r
= B^*\r.
\]
\end{boxprop}
Note that in the above linear systems,
we can access $A$ and $B$ as linear maps, the JVPs of $F$.
Their adjoints, $A^*$ and $B^*$, correspond to the VJPs of $F$.
To solve these systems, we can therefore use \textbf{matrix-free} solvers as
detailed in \cref{higher:sec:ihvp}. For
example, when $A$ is symmetric positive semi-definite, we can use the
conjugate gradient method \citep{conjugate_gradient}. When $A$ is not symmetric
positive definite, we can use GMRES \citep{saad_1986} or BiCGSTAB
\citep{Vorst1992-bicgstab}.

\subsection{Proof of the implicit function theorem}

  We prove the theorem using the inverse function theorem presented
  in~\cref{implicit:thm:inv_diff}. Define 
  \[
    f(\lambdav, \w) = (\lambdav, F(\w, \lambdav)) 
  \]
  which goes from $\RR^Q \times \RR^P$ onto  $\RR^Q \times \RR^P$. The Jacobian of $f$ is 
  \[
  \jac f(\lambdav, \w)  =
  \begin{pmatrix}
    \idm & 0 \\
    \jac_2  F(\w, \lambdav) & \jac_1 F(\w, \lambdav)
  \end{pmatrix}.
  \]
  So at $\w_0, \lambdav_0$, we have $\det(\jac f(\lambdav_0, \w_0)) =
  \det(\idm)\det( \jac_1 F(\w_0, \lambdav_0)) \neq 0$ since we assumed $\partial_1
  F(\wv_0, \lambdav_0)$ invertible. By the inverse function theorem, the
  function $f$ is then invertible in a neighborhood $N$ of $f(\lambdav_0,
  \w_0) = (\lambdav_0, \zeros)$. In particular, it is invertible in $N \cap
  \{(\lambdav, \zeros), \lambdav \in \RR^Q\}$. The solution of the implicit
  equation in a neighborhood of $\lambdav_0$ is then $( \lambdav,
  \w^*(\lambdav)) = f^{-1}(\lambdav, \zeros)$. By the inverse function theorem,
  $f^{-1}$ is continuously differentiable inverse and so is $\w^*(\lambdav)$.
  The derivative $\partial \w^*(\lambdav)$ is obtained
  from the differential of the inverse as
  \[
    \begin{pmatrix}
      \sim  & \sim \\
      \partial \w^*(\lambdav) & \sim
    \end{pmatrix} = \partial f^{-1}(\lambdav, \zeros),
  \]
  and by the inverse function theorem, we have
  \begin{equation*}
  \partial f^{-1}(\lambdav, \zeros) = 
  (\partial f(\lambdav, \w^*(\lambdav)))^{-1}.
  \end{equation*}
  Using the block matrix inversion formula, we therefore obtain
  \[
    \begin{pmatrix}
      \A & \B \\
      \C & \D
    \end{pmatrix}^{-1} 
    = 
    \begin{pmatrix}
      \sim & \sim \\
      -(\D - \C\A^{-1}\B)^{-1}\C\A^{-1} & \sim
    \end{pmatrix},
  \]
  we get the claimed expression. Though we expressed the proof in terms of
  Jacobians and matrices, the result naturally holds for the corresponding
  linear operators, JVPs, VJPs, and their inverses.

\section{Adjoint state method}\label{imp_diff:sec:adjoint_method}

\subsection{Differentiating nonlinear equations}

We describe in this section the adjoint state method (\aka adjoint method,
method of adjoints, adjoint sensitivity method). 
The method can be used to compute the gradient of the composition of an explicit
function and an \textbf{implicit function}, defined through an \textbf{equality
constraint} (e.g., a \textbf{nonlinear equation}).
The method dates back to \citet{cea_1986}.

Suppose a variable $\s \in \cS$ (which corresponds to a \textbf{state} in
optimal control) is implicitly defined given some parameters $\w \in \cW$
through the (potentially nonlinear) equation
$c(\s, \w) = \zeros$, where $c \colon \cS \times \cW \to \cS$.
Assuming $\s$ is \textbf{uniquely} determined for all $\w \in \cW$, this defines
an \textbf{implicit function} $\s^\star(\w)$ from $\cW$ to $\cS$ such that
$c(\s^\star(\w), \w) = \zeros$.
Given an objective function $L \colon \cS \times \cW \to \RR$,
the goal of the adjoint state method is then to compute the gradient of
\begin{equation*}
L(\w) \coloneqq L(\s^\star(\w), \w).
\end{equation*}
However, this is not trivial as $\s^\star(\w)$
is an implicit function.
For instance, this can be used to convert the \textbf{equality-constrained}
problem
\begin{align*}
\min_{\w \in \cW} L(\s, \w) 
\quad \text{s.t.} \quad c(\s, \w) = \zeros.
\end{align*}
into the \textbf{unconstrained} problem
\begin{align*}
\min_{\w \in \cW} L(\s^\star(\w), \w).
\end{align*}
Access to $\nabla L(\w)$ allows us to solve this problem by gradient
descent.
\begin{boxprop}{Adjoint state method}
Let $c:\cS \times \cW \rightarrow \cS$ be a mapping defining constraints of
the form $c(\s, \w)$. Assume that for each $\w \in \cW$, there exists a unique
$\s^\star(\w)$ satisfying $c(\s^\star(\w), \w)= \zeros$ and that $\s^\star(\w)$
is differentiable.  The gradient of 
\[
  L(\w) \coloneqq L(\s^\star(\w), \w),
\]
for some differentiable function $L:\cS \times \cW \rightarrow \RR$, is given by 
\begin{equation*}
  \nabla L(\w) = 
  \nabla_2 L(\s^\star(\w), \w) + \partial_2 c(\s^\star(\w), \w)^* \r^\star(\w),
\end{equation*}
where $\r^\star(\w)$ is the solution of the linear system
\begin{equation*}
\partial_1 c(\s^\star(\w), \w)^* \r = -\nabla_1 L(\s^\star(\w), \w).
\end{equation*}

\end{boxprop}
As shown in the proof below, $\r^\star(\w)$ corresponds to a \textbf{Lagrange
multiplier}. The linear system can be solved using matrix-free solvers.

\subsection{Relation with envelope theorems}

Because $\s$ is uniquely determined for any $\w \in \cW$ by $c(\s, \w) =
\zeros$, we can alternatively rewrite $L(\w)$ as
the trivial minimization or maximization,
\begin{align*}
L(\w) 
&= \min_{\s \in \cS} L(\s, \w) \quad \text{s.t.} \quad c(\s, \w) = \zeros\\
&= \max_{\s \in \cS} L(\s, \w) \quad \text{s.t.} \quad c(\s, \w) = \zeros.
\end{align*}
Therefore, the adjoint state method can be seen as an envelope theorem for
computing $\nabla L(\w)$, for the case when $\w$ is involved in \textbf{both}
the objective function and in the \textbf{equality constraint}.

\subsection{Proof using the method of Lagrange multipliers}
\label{implicit:sec:proof_adjoint_lagrange}

Classically, the adjoint state method is derived using the method of Lagrange
multipliers.
Let us introduce the \textbf{Lagrangian} associated with $L$ and $c$,
\begin{equation*}
\cL(\s, \w, \r) \coloneqq L(\s, \w) + \langle \r, c(\s, \w) \rangle,
\end{equation*}
where $\r \in \cS$ is the \textbf{Lagrange multiplier} associated with the
equality constraint $c(\s, \w) = \zeros$. In the optimal control literature,
$\r$ is often called the \textbf{adjoint variable} or \textbf{adjoint state}.
The gradients of the Lagrangian are
\begin{align*}
\nabla_\s \cL(\s, \w, \r) &= \nabla_1 L(\s, \w) + \partial_1 c(\s, \w)^* \r \\
\nabla_\w \cL(\s, \w, \r) &= \nabla_2 L(\s, \w) + \partial_2 c(\s, \w)^* \r \\
\nabla_\r \cL(\s, \w, \r) &= c(\s, \w),
\end{align*}
where $\partial_i c(\s, \w)^*$ are the \textbf{adjoint operators}.
Setting $\nabla_\r \cL(\s, \w, \r)$ to zero gives the constraint
$c(\s, \w) = \zeros$.
Setting $\nabla_\s \cL(\s, \w, \r)$ to zero gives the so-called 
\textbf{adjoint state equation}
\begin{equation*}
\partial_1 c(\s, \w)^* \r = -\nabla_1 L(\s, \w).
\end{equation*}
Solving this linear system \wrt $\r$ at $\s = \s^\star(\w)$ gives the adjoint
variable $\r^\star(\w)$. We then get 
\begin{align*}
\nabla L(\w)
&= \nabla_2 \cL(\s^\star(\w), \w, \r^\star(\w)) \\
&=
\nabla_2 L(\s^\star(\w), \w) + \partial_2 c(\s^\star(\w), \w)^* \r^\star(\w),
\end{align*}
which concludes the proof.

\subsection{Proof using the implicit function theorem}
\label{implicit:sec:proof_adjoint_ift}

A more direct proof is possible thanks to the implicit function theorem
(\cref{imp_diff:sec:ift}).
Using the chain rule, we get
\begin{equation*}
\nabla L(\w) = \nabla_2 L(\s^\star(\w), \w) 
+ \partial \s^\star(\w)^* \nabla_1 L(\s^\star(\w), \w),
\end{equation*}
where $\partial \s^\star(\w)^*$ is the VJP of $\s^\star$,
a linear map from $\cS$ to $\cW$.

Computationally, the main difficulty is to apply $\partial \s^\star(\w)^*$ to
the vector $\u = \nabla_1 L(\s^\star(\w), \w) \in \cS$. 
Using the implicit function theorem (\cref{imp_diff:sec:ift}) on the implicit
function $c(\s^\star(\w), \w) = \zeros$,
and \cref{imp_diff:prop:jvp_vjp_implicit_function},
we get the linear system
$A^* \r = \u$,
where $A^* \coloneqq \partial_1 c(\s^\star(\w), \w)^*$
is a linear map from $\cS$ to $\cS$. After solving for $\r$,
we get
$\partial \s^\star(\w)^* \u = B^* \r$,
where
$B^* \coloneqq \partial_2 c(\s^\star(\w), \w)^*$ is a linear map
from $\cS$ to $\cW$.
Putting everything together, we get
\begin{equation*}
\nabla L(\w) = \nabla_2 L(\s^\star(\w), \w) 
+ \partial_2 c(\s^\star(\w), \w)^* \r.
\end{equation*}

\subsection{Reverse mode as adjoint method with backsubstitution}

In this section, we revisit reverse-mode autodiff from the perspective of the
adjoint state method. For clarity, we focus our exposition on feedforward
networks with input $\x \in \cX$ and network weights $\w = (\w_1,
\ldots, \w_k) \in \cW_1 \times \ldots \times \cW_K$, defined by
\begin{align}
	\s_0 &\coloneqq \x \in \cX \nonumber \\
	\s_1 &\coloneqq f_1(\s_0, \w_1) \in \cS_1 \nonumber \\ 
		 & \hspace{6pt} \vdots \nonumber\\
	\s_K &\coloneqq f_K(\s_{K-1}, \w_K) \in \cS_K \nonumber \\
	f(\w) &\coloneqq \s_K. \label{auto_diff:eq:feedforward_autodiff_viewpoints}
\end{align}
Here we focus on gradients with respect to the parameters $\w$, hence the
notation $f(\w)$. We can use the adjoint state method 
to recover reverse-mode autodiff, and prove its \textbf{correctness}
in the process.
While we focus for simplicity on feedforward networks, our exposition can be
generalized to computation graphs.

\subsubsection*{Feedforward networks as the solution of a nonlinear equation}

While we defined the set of intermediate computations $\s = (\s_1, \ldots, \s_K)
\in \cS_1 \times  \ldots \times \cS_K$ as a sequence of operations, they can
also be defined as the unique solution of the
\textbf{nonlinear equation}
$c(\s, \w) = \zeros$, 
where
\begin{equation*}
  c(\s, \w) \coloneqq 
  \begin{pmatrix}
    \s_1 - f_1(\x, \w_1) \\
    \s_2 - f_2(\s_1, \w_2) \\
    \vdots \\
    \s_K - f_K(\s_{K-1}, \w_K)
  \end{pmatrix}.
\end{equation*}
This defines an \textbf{implicit function} 
$\s^\star(\w) = (\s_1^\star(\w), \ldots, \s_K^\star(\w))$,
the solution of this nonlinear system, which is given by the variables $\s_1,
\ldots, \s_K$ defined in~\cref{auto_diff:eq:feedforward_autodiff_viewpoints}.
The output of the feedforward network is then $f(\w) = \s_K^\star(\w)$.

In machine learning, the final layer $\s^\star_K(\w)$ is typically fed into a
loss $\ell$, to define
\begin{equation*}
L(\w) \coloneqq \ell(\s_K^\star(\w); \y).
\end{equation*}
Note that an alternative is to write $L(\w)$ as
\begin{equation*}
L(\w) = \min_{\s \in \cS} \ell(\s; \y)
\quad \text{s.t.} \quad
c(\s, \w) = \zeros.
\end{equation*}

More generally, if we just want to compute the VJP of $\s_K^\star(\w)$ in
some direction $\u_K \in \cS_K$, we can define the scalar-valued function
\begin{equation*}
L(\w) 
\coloneqq \ell(\s_K^\star(\w); \u_K)
\coloneqq \langle \s_K^\star(\w), \u_K \rangle
\end{equation*}
so that
\begin{equation*}
\partial f(\w)^* \u_K = \nabla L(\w).
\end{equation*}
Let us define 
$\u \in \cS_1 \times \dots \times \cS_{K-1} \times \cS_K$
as
$\u \coloneqq (\zeros, \ldots, \zeros, \nabla_1 \ell(f(\w); \y))$ 
(gradient of the loss $\ell$ case)
or
$\u \coloneqq (\zeros, \ldots, \zeros, \u_K)$ 
(VJP of $f$ in the direction $\u_K$ case).
Using the adjoint state method, we know that
the gradient of this objective is obtained as 
\begin{equation*}
\nabla  L(\w) = \partial_2 c(\s(\w), \w)^* \r^\star(\w),
\end{equation*}
for $\r^\star(\w)$ the solution of the linear system
\begin{equation*}
\partial_1 c(\s(\w), \w)^* \r =  -\u.
\end{equation*}

\subsubsection*{Solving the linear system using backsubtitution}

The JVP of the constraint function $c$ at $\s^\star(\w)$, materialized as
a matrix, takes the form of a \textbf{block lower-triangular matrix}
\[
  \jac_1 c(\s^\star(\w), \w) 
  =
  \begin{pmatrix}
    \idm & 0 & \dots & \dots & 0 \\
    - A_1 & \idm & \ddots &  & \vdots\\
    0 & -A_2 & \idm & \ddots & \vdots \\
    \vdots & \ddots & \ddots & \ddots & 0 \\
    0 & \dots & 0 & -A_K & \idm
  \end{pmatrix},
\]
where $A_k \coloneqq \jac_1 f_k(\s_{k-1}, \w_k)$. 
Crucially the triangular structure of
the JVP stems from the fact that each intermediate activation only depends on
the past intermediate activations. Therefore, the constraints, corresponding to
the lines of the Jacobian, cannot introduce non-zero values beyond its diagonal.
The VJP takes the form of a \textbf{block upper-triangular matrix}
\[
  \jac_1 c(\s^\star(\w), \w)^*
  =
  \begin{pmatrix}
    \idm & - A_1^* & 0 & \dots & 0 \\
    0 & \idm & -A_2^* & \ddots & \vdots\\
    \vdots & \ddots & \idm & \ddots & 0 \\
    \vdots &  & \ddots & \ddots & -A_K^* \\
    0 & \dots & \dots & 0 & \idm
  \end{pmatrix}.
\]
Solving an upper triangular system like $\partial_1 c(\s(\w), \w)^*\r = \u$ can
then be done efficiently by \textbf{backsubstitution}. Starting from the last
adjoint state $\r_K = \u$, we can compute each adjoint state $\r_k$ from that
computed at $k+1$. Namely, for $k \in (K-1,\dots,1)$, we have 
\begin{align*}
\r_k -  A_{k+1}^*\r_{k+1} = \zeros
\iff 
\r_k = \partial f_{k+1}(\s_k, \w_{k+1})^*\r_{k+1}.
\end{align*}
The VJPs with respect to the parameters are then obtained by
\begin{equation*}
  \partial_2 c(\s(\w), \w)^* \r
  = \begin{pmatrix}
    \partial_2 f_1(\x, \w_1)^* \r_1 \\
    \partial_2 f_2(\s_1(\w), \w_2)^* \r_2 \\
    \vdots \\
    \partial_2 f_K(\s_{K-1}(\w), \w_K)^* \r_K
  \end{pmatrix},
\end{equation*}
recovering reverse-mode autodiff.

The Lagrangian perspective of backpropagation for networks with
separate parameters $\w = (\w_1, \dots, \w_K)$ is well-known;
see for instance \citet{lecun_1988_lagrangian} or \citet{mates_of_costate}.
The Lagrangian perspective of backpropagation through time \citep{werbos_1990}
for networks with shared parameter $\w$ is discussed for instance by
\citet{franceschi_2017}. 
Our exposition uses the adjoint state method,
which can itself be proved either using the method of Lagrange multipliers
(\cref{implicit:sec:proof_adjoint_lagrange}) or by
the implicit function theorem (\cref{implicit:sec:proof_adjoint_ift}), 
combined with backsubtitution for solving the
upper-triangular linear system.
Past works often minimize over $\w$ but we do not require this, as gradients are
not necessarily used for optimization.
Our exposition also supports computing 
the VJP of any vector-valued function $f$, while existing works derive the gradient of a
scalar-valued loss function.

\section{Inverse function theorem}
\label{implicit:sec:inverse_function}

\subsection{Differentiating inverse functions}

In some cases (see for instance \cref{grad_est:sec:push_forward}),
it is useful to compute the Jacobian of an inverse function $f^{-1}$.
The inverse function theorem below allows us to relate the Jacobian of $f^{-1}$
with the Jacobian of $f$.
\begin{boxthm}{Inverse function theorem}\label{implicit:thm:inv_diff}
Assume $f \colon \cW \to \cW$ is continuously differentiable with invertible
Jacobian $\partial f(\wv_0)$ at $\wv_0$. Then $f$ is bijective from a
neighborhood of $\wv_0$ to a neighborhood of $f(\w_0)$. Moreover, the inverse
$f^{-1}$ is continuously differentiable near $\omegav_0 = f(\wv_0)$ and the
Jacobian of the inverse $\partial f^{-1}(\omegav)$ is
\begin{equation*}
  \partial f(\wv) \partial f^{-1}(\omegav)= I
\Leftrightarrow
\partial f^{-1}(\omegav) = (\partial f(\wv))^{-1},
\end{equation*}
with $\wv = f^{-1}(\omegav)$.
\end{boxthm}

\subsection{Link with the implicit function theorem}

The inverse function theorem can be used to prove the implicit function theorem;
see proof of \cref{implicit:thm:ift_multivariate}.
Conversely, recall that, in order to use the implicit function theorem,
we need to choose a root objective $F \colon \cW \times \Lambda \to \cW$.
If we set $\cW=\Lambda=\RR^Q$ and $F(\wv, \omegav) = f(\wv) - \omegav$, 
with $f \colon \R^Q \to \R^Q$, then
we have that the root $\wv^\star(\omegav)$ satisfying
$F(\wv^\star(\omegav), \omegav) = \zeros$ is exactly
$\wv^\star(\omegav) = f^{-1}(\omegav)$. Moreover,
$\partial_1 F(\wv, \omegav) = \partial f(\wv)$
and
$\partial_2 F(\wv, \omegav) = -I$.
By applying the implicit function theorem with this $F$,
we indeed recover the inverse function theorem.

\subsection{Proof of inverse function theorem}

  We first give a proof of the formula assuming that $f^{-1}$ is well-defined
  and continuously differentiable in a neighborhood of $f(\w_0)$. In that case, 
  we have for any $\omegav$ in a neighborhood of $f(\w_0)$,
  \[
     f \circ f^{-1}(\omegav) = \omegav.
  \]
  Differentiating both sides w.r.t. $\omegav$, we get
  \[
    \partial f(f^{-1}(\omegav)) \partial f^{-1}(\omegav) = \idm,
  \]
  where $\idm$ is the identity function in $\RR^Q$. In particular, for $\w =
  f^{-1}(\omegav)$ we recover the formula presented in the statement.

  Now, it remains to show that invertibility of the JVP ensures that the
  function is invertible in a neighborhood of $f(\w_0)$ and that the inverse is
  continuously differentiable. For that, denote $\l=\partial f(\w_0)$ such that
  $\l^{-1}$ is well-defined by definition. $f$ is invertible with continuously
  differentiable inverse, if and only if $\l^{-1}(f(\w)) - f(\w_0)$ is
  invertible with continuously differentiable inverse. So without loss of
  generality, we consider $\partial f(\w_0) = \idm$, $f(\w_0) = \zeros$, 
  $\w_0=\zeros$.
  
  As $f$ is continuously differentiable, there exists a neighborhood 
  $\cN \coloneqq \{\w: \|\w-\w_0\|_2\leq \delta\}$ 
  on which we have $\|\partial f(\w) - \idm\|_2\leq
  1/2$. In this neighborhood, the function $g(\w) = f(\w) - \w$ is contractive
  by the mean value theorem with contraction factor $1/2$. For any $\omegav$
  such that $\|\omegav-f(\w_0)\|_2 \leq \delta/2$, the sequence $\w_{k+1} =
  \w_k - f(\w_k) - \omegav'$ remains in $\mathcal{N}$ and converges (since it
  is a Cauchy sequence by the contraction of $g$) to a unique fixed point
  $\w_\infty$ satisfying $\w_\infty = \w_\infty - f(\w_\infty) - \omegav \iff
  f(\w_\infty) = \omegav$. This shows the existence of the inverse in the
  neighborhood $\cM \coloneqq \{\omegav: \|\omegav-\omegav_0\|_2\leq \delta/2\}$
  of $\omegav_0 = f(\w_0)$ onto $\cN$.

  We tackle now the differentiability (hence the continuity) of $f^{-1}$. For
  any $\omegav$ in the neighborhood of $\omegav_0$ with inverse $\w \coloneqq
  f^{-1}(\omegav) \in N$, the JVP of $f$ at $\w$ satisfies by assumption
  $\|\partial f(\w) - \idm\|_2 \leq 1/2$. Hence, $\a = \partial f(\w) -\idm$
  defines a convergent series $\b = \sum_{k=0}^{+\infty} \a^k$ and one verifies
  easily that $\b = \partial f(\w)^{-1}$, that is $\partial f(\w)$ is
  invertible and $\|\partial f(\w)^{-1}\|\leq 2$. To compute the JVP of the
  inverse, we consider then $ \partial f(\w)^{-1}$ as the candidate JVP and
  examine
  \[
    \frac{\|f^{-1}(\omegav + \etav) - f(\omegav) - (\partial f(\w))^{-1}\etav\|_2}
    {\|\etav\|_2}.
  \]
  Denote then $\v$ such that $f^{-1}(\omegav + \etav) = \w + \v$. As $g(\w) =
  f(\w) - \w$ is $1/2$-contractive in $\cN$, we have $\|\v -\etav\|_2 = \|g(\w +
  \v) - g(\w)\|_2 \leq 1/2\|\v\|_2$. So $\|\v\|_2 \geq \|\etav\|/2$. We then get
  \begin{align*}
    & \frac{\|f^{-1}(\omegav + \etav) - f(\omegav) - (\partial f(\w))^{-1}\etav\|_2}
    {\|\etav\|_2} \\
    & = \frac{\|\v - (\partial f(\w))^{-1}(f(\w + \v) - f(\w))\|_2}
    {\|\etav\|_2} \\
    & \leq 4 \frac{\|f(\w+\v) - f(\w) - \partial f(\w)\v\|_2}{\|\v\|_2}
  \end{align*}
  As $\|\etav\|_2 \rightarrow 0$, we have $\|\v\|_2 \rightarrow 0$ and so
  $\|f(\w+\v) - f(\w) - \partial f(\w)\v\|_2/\|\v\|_2 \rightarrow 0$. Hence,
  $f^{-1}$ is differentiable with JVP $\partial f^{-1}(\omegav) =  (\partial
  f(\w))^{-1} =  (\partial f(f^{-1}(\omegav)))^{-1}$. This shows that $f^{-1}$
  is continuous and so $\partial f^{-1}(\omegav)$ is continuous as a composition
  of continuous functions.

\section{Summary}

\begin{itemize}

\item Implicit functions are functions that cannot be decomposed into elementary
operations and for which autodiff can therefore not be directly applied.
Examples are optimization problems and nonlinear equations.

\item Envelope theorems can be used for differentiating through the min or max value
(not solution) of a function.

\item More generally, the implicit function theorem allows us to differentiate through
implicit functions. It gives conditions for the existence of
derivatives and how to obtain them. 

\item The adjoint state method can be used to
obtain the gradient of the composition of an explicit function and of an
implicit function, specified by equality constraints.
It can be used to prove the correctness of reverse-mode autodiff.

\item The inverse function theorem can be used to differentiate function inverses.

\item In a sense, the implicit function theorem can be thought of as the mother
    theorem, as it can be used to prove envelope theorems, the adjoint state
    method and the inverse function theorem.

\end{itemize}

%% file: chapters/diff_thru_int/grad_est_main.tex
\chapter{Differentiating through integration}
\label{chap:grad_est}

In this chapter, we study how to differentiate through integrals,
with a focus on expectations and solutions of
ordinary differential equations.

\section{Differentiation under the integral sign}
\label{grad_est:rem:swap_derivative_integration} 

Given two Euclidean spaces $\Theta$ and $\cY$, and a function $f: \Theta
\times\cY \rightarrow \RR$, 
we often want to differentiate an integral of the form
\begin{equation*}
    F(\thetav) \coloneqq \int_\cY f(\thetav, \y)d\y.
\end{equation*}
Provided that we can swap integration and differentiation,
we have
\begin{equation*}
\nabla F(\thetav) = \int_\cY \nabla_\thetav f(\thetav, \y)d\y.
\end{equation*}
The conditions enabling us
to do so are best examined in the context of measure theory. We refer
the reader to e.g.~\citep{cohn2013measure} for a course on measure theory and
\citet{flanders1973differentiation} for an in-depth study of the differentiation
under the integral sign. 
Briefly, if $\Theta = \cY = \RR$, the following
conditions are sufficient.
  \begin{enumerate}
    \item $f$ is measurable in both its arguments, and $f(\theta,
    \cdot)$ is integrable for almost all $\theta \in \Theta$ fixed,

    \item $f(\cdot, y)$ is absolutely continuous for almost all $y \in \cY$,
    that is, there exists an integrable function $g(\cdot, y)$ such that
    $f(\theta, y) = f(\theta_0, y) + \int_{\theta_0}^\theta g(\tau, y)d\tau$,

    \item $\partial_1 f(\theta, y)$ (which exists almost everywhere if
    $f(\cdot, y)$ is absolutely continuous), is locally integrable, that is, for
    any closed interval $[\theta_0, \theta_1]$, the integral
    $\int_{\theta_0}^{\theta_1} \int |\partial_1 f(\theta, y)| dy d\theta$ is
    finite.
  \end{enumerate} 
Any differentiable function $f:\Theta \times \cY \rightarrow \RR$ is absolutely
continuous. However, the conditions also hold if $f$ is just absolutely
continuous, that is, if $f(\cdot, y)$ is differentiable for almost all $y$.  
This weaker assumption can be used to smooth out differentiable
almost-everywhere functions, such as the ReLU, as we study in
\cref{perturb:sec:blackbox}.

\section{Differentiating through expectations}\label{grad_est:sec:diff_thru_expect} 

A special case of differentiating through integrals is differentiating through
expectations. We can distinguish between two cases, depending on whether the
parameters $\thetav$ we wish to differentiate
are involved in the distribution or in the function, whose expectation we
compute.

\subsection{Parameter-independent distributions}\label{grad_est:sec:param_indepdent_expect}

We first consider expectations of the form
\begin{equation*}
F(\thetav) 
\coloneqq \EE_{Y \sim p}[g(Y, \thetav)]
= \int_{\cY} g(\y, \thetav) p(\y) d\y,
\end{equation*}
for a random variable $Y \in \cY \subseteq \RR^M$, distributed according to a
distribution $p$, and a function $g \colon \cY \times \Theta \to \R$.
Importantly,
the distribution is independent of the parameters $\thetav$.
Under mild conditions recalled
in~\cref{grad_est:rem:swap_derivative_integration}, 
we can swap differentiation and integration to obtain
\begin{align*}
\nabla F(\thetav)
&= \nabla_\thetav \int_{\cY} g(\y, \thetav) p(\y) d\y \\
&= \int_{\cY} \nabla_\thetav g(\y, \thetav) p(\y) d\y \\
&= \EE_{Y \sim p}[\nabla_\thetav g(Y, \thetav)].
\end{align*}
Generally, the expectation cannot be computed in closed form. 
However, provided that we can sample from $p$, 
we can define a Monte-Carlo estimator of the value
\begin{equation*}
\widehat{F}_N(\thetav) \coloneqq \frac{1}{N}\sum_{i=1}^{N} g(Y_i, \thetav)
\end{equation*}
and of the gradient

\[
\nabla \widehat{F}_N(\thetav) 
= \frac{1}{N}\sum_{i=1}^{N} \nabla_\thetav g(Y_i, \thetav),
\]
for $N$ i.i.d. samples $Y_1, \ldots, Y_N$ from $p$. 
These estimators are unbiased, meaning that 
$\EE[\widehat{F}_N(\thetav)] = F(\thetav)$
and
$\EE[\nabla \widehat{F}_N(\thetav)] = \nabla F(\thetav)$, 
and they converge
to the true quantity as $N \rightarrow +\infty$.
This suggests a simple implementation in an autodiff framework
of the approximation of $\nabla F(\thetav)$:
\begin{enumerate}
  \item Sample $y_1, \ldots, y_N$ from $p$.
  \item Compute $\widehat{F}_N(\thetav) = \frac{1}{N}\sum_{i=1}^{N} g(y_i,
      \thetav)$.
  \item Compute the gradient $\nabla \widehat{F}_N(\thetav)$ by automatic
      differentiation.
\end{enumerate}
Computing higher-order derivatives follows the same principle: to get an
approximation of $\nabla^2 F(\thetav)$, we can simply compute $\nabla^2
\widehat{F}_N(\thetav)$ by autodiff. As such, the implementation delineated above is akin to
the ``discretize-then-optimize'' approach used to differentiate through the
solution of an ODE (\cref{chap:ode}): we implement an approximation of the
objective and simply call autodiff on it. 

\subsection{Parameter-dependent distributions}

A more challenging case arises when the distribution depends on the parameters
$\thetav$:
\begin{equation*}
E(\thetav) \coloneqq \EE_{Y \sim p_\thetav}[g(Y)]
= \int_{\cY} g(\y) p_\thetav(\y) d\y,
\end{equation*}
where $Y \in \cY \subseteq \RR^M$ is a random variable, distributed according to
a distribution $p_\thetav$ parameterized by $\thetav \in \Theta$ and where $g
\colon \cY \to \R$ is, depending on the setting, potentially a blackbox
function (i.e., we do not have access to its gradients). 
Typically, $\thetav \in
\Theta$ could be parameters we wish to estimate, or it could indirectly be
generated by $\thetav = f(\x, \w) \in \Theta$, where $f$ is a neural network
with parameters $\w \in \cW$ we wish to estimate. 
The main difficulty in computing $\nabla E(\thetav)$ stems from the fact that
$\thetav$ are the parameters of the distribution $p_\thetav$.  Estimating an
expectation $E(\thetav) = \EE_{Y \sim p_\thetav}[g(Y)]$ using Monte-Carlo
estimation requires us to sample from $p_\thetav$. However, it is not clear how
to differentiate $E$ w.r.t. $\thetav$ if $\thetav$ is involved in the sampling
process. 

\subsubsection{Continuous case}

When $\cY$ is a continuous set (that is, $p_\thetav(\y)$ is a probability
density function), we can rewrite $E(\thetav)$ as
\begin{equation*}
E(\thetav) = \int_{\cY} p_\thetav(\y) g(\y) d\y.
\end{equation*}
Provided that we can swap integration and differentiation
(see \cref{grad_est:rem:swap_derivative_integration}), we then have
\begin{align*}
\nabla E(\thetav) 
&= \nabla_\thetav \int_{\cY} p_\thetav(\y) g(\y) d\y \\
&= \int_{\cY} \nabla_\thetav p_\thetav(\y) g(\y) d\y.
\end{align*}
Unfortunately, this integral is not an expectation and it could be
intractable in general.

\subsubsection{Discrete case}

When $\cY$ is a discrete set (that is, $p_\thetav(\y)$ is a probability
mass function), we can rewrite $E(\thetav)$ as
\begin{equation*}
E(\thetav) = \sum_{\y \in \cY} p_\thetav(\y) g(\y).
\end{equation*}
We then obtain
\begin{equation*}
\nabla E(\thetav) 
= \sum_{\y \in \cY}  g(\y) \nabla_{\thetav} p_\thetav(\y).
\end{equation*}
Again,
$\nabla E(\thetav)$ 
is not an expectation.
We therefore cannot use Monte-Carlo estimation to 
estimate the gradient.
Instead, we can compute it by brute force, i.e., 
by summing over all possible $\y \in \cY$.
However, this is clearly only computationally tractable 
if $|\cY|$ is small 
or if $p_\thetav$ is designed to have sparse support, i.e.,
so that the set $\{\y \in \cY \colon p_\thetav(\y) \neq 0\}$
is small. Moreover, even if these conditions hold,
summing over $\y$ could be problematic if $g(\y)$ is expensive to
compute. Therefore, exact gradients are seldom used in practice.

In Sections~\ref{grad_est:sec:sfe} and~\ref{grad_est:sec:pge}, we
review the score function and pathwise gradient estimators,
to (approximately) compute $\nabla E(\thetav)$,
allowing us to optimize $\thetav$ (or $\w$ using the chain rule) by
gradient-based algorithms.

\subsection{Application to expected loss functions}

Differentiating through expectations
is particularly useful when working with expected loss
functions of the form
\begin{equation*}
L(\thetav; \y) \coloneqq \EE_{\hat Y \sim p_\thetav}[\ell(\hat Y, \y)],
\end{equation*}
where $\y$ is some ground truth.
Equivalently, we can set $\ell = -r$, where $r$ is a \textbf{reward function}.
As we shall see, the score function estimator will support a discrete loss
function
$\ell \colon \cY \times \cY \to \R$, while the pathwise gradient estimator
will require
a differentiable loss function $\ell \colon \RR^M \times \cY \to \R$.
Intuitively, $L(\thetav; \y)$ will be low if 
$p_\thetav$ assigns high probability to predictions $\widehat \y$ with low loss
value $\ell(\widehat \y, \y)$.

In the classification setting, where $\cY = [M]$,
$p_\thetav$ is often chosen to be the \textbf{Gibbs distribution}, which is
a categorical distribution induced by a softargmax
\begin{equation*}
p_\thetav(y) 
\coloneqq \frac{\exp(\theta_{y})}{\sum_{i \in [M]} \exp(\theta_i)} 
= \left[\mathrm{softmax}(\thetav)\right]_{y} \in (0, 1),
\end{equation*}
where $\theta_{y} \coloneqq f(\x, y, \w) \in \RR$ are logits produced by a
neural network $f$. More generally, in the structured prediction setting, 
where $\cY \subseteq \R^M$ but $|\cY| \gg M$, 
we often use the distribution
\begin{equation*}
p_\thetav(\y) \coloneqq
\frac{\exp(\langle \phi(\y), \thetav \rangle)}{\sum_{\y' \in \cY} \exp(\langle
    \phi(\y'), \thetav \rangle)},
\end{equation*}
where $\thetav = f(\x, \w) \in \RR^M$.

Given a distribution $\rho$ over $\cX \times \cY$, 
we then want to minimize
the expected loss function, also known as \textbf{risk},
\begin{equation*}
R(\w) \coloneqq \EE_{(X,Y) \sim \rho}[L(f(X, \w); Y)].
\end{equation*}
Typically, minimizing $R(\w)$ is done through some form of gradient descent,
which requires us to be able to compute 
\begin{equation*}
\begin{aligned}
\nabla R(\w) 
&= \EE_{(X, Y) \sim \rho}[\nabla_\w L(f(X, \w); Y)] \\
&= \EE_{(X, Y) \sim \rho}[\partial_2 f(X, \w)^* \nabla L(f(X, \w); Y)].
\end{aligned}
\end{equation*}
Computing $\nabla R(\w)$ therefore boils down to computing the gradient
of $L(\thetav; \y)$, which is the gradient of an expectation.

\subsection{Application to experimental design}

In experimental design, we wish to minimize a function $g(\lambdav)$, which we
assume costly to evaluate. As an example, evaluating $g(\lambdav)$ could
require us to run a scientific experiment with parameters $\lambdav \in \RR^Q$.
As another example, in
hyperparameter optimization, evaluating $g(\lambdav)$ would require us to run a
learning algorithm with hyperparameters $\lambdav \in \RR^Q$. Instead of solving
the problem $\argmin_{\lambdav \in \RR^Q} g(\lambdav)$, we can lift the problem
to probability distributions and solve $\argmin_{\thetav \in \RR^M} E(\thetav)$,
where $E(\thetav) = \EE_{\lambdav \sim p_\thetav}[g(\lambdav)]$.  
This requires the probability distribution $p_\thetav$ to assign high
probability to $\lambdav$ values that achieve small $g(\lambdav)$ value.
Solving this
problem by stochastic gradient descent requires us to be able to compute
estimates of $\nabla E(\thetav)$. This can be done for instance with SFE
explained in \cref{grad_est:sec:sfe}, which does not require gradients of $g$,
unlike implicit differentiation explained in \cref{chap:imp_diff}.  This
approach also requires us to choose a distribution $p_\thetav$ over $\lambdav$.
For continuous hyperparameters, a natural choice would be the normal
distribution $\lambdav \sim \mathrm{Normal}(\muv, \Sigmav)$, setting $\thetav =
(\muv, \Sigmav)$. Once we have obtained $\thetav$ by minimizing $E(\thetav)$, 
we need a way to recover $\lambdav$.
This can be done for example by choosing the mode of the distribution, i.e.,
$\argmax_{\lambdav \in \RR^Q} p_\thetav(\lambdav)$, or the mean of the
distribution
$\EE_{\lambdav \sim p_\thetav(\lambdav)}[\lambdav]$. Of course,
in the case of the normal distribution, they coincide.

\section{Score function estimators, REINFORCE}
\label{grad_est:sec:sfe}

\subsection{Scalar-valued functions}

The key idea of the \textbf{score function estimator} (SFE),
also known as REINFORCE, is to rewrite $\nabla E(\thetav)$
as an expectation.
The estimator is based on the \textbf{logarithmic derivative identity}
\begin{equation*}
\nabla_\thetav \log p_\thetav(\y) 
= \frac{\nabla_\thetav p_\thetav(\y)}{p_\thetav(\y)}
\Longleftrightarrow
\nabla_\thetav p_\thetav(\y)
= 
p_\thetav(\y)
\nabla_\thetav \log p_\thetav(\y).
\end{equation*}
Using this identity, we obtain
the following gradient estimator.
\begin{boxprop}{SFE for scalar-valued functions}
\label{grad_est:prop:sfe}
Given a family of distributions $p_\thetav$ on $\cY$, for $\theta \in \Theta$,
define
\begin{equation*}
E(\thetav) 
\coloneqq \EE_{Y \sim p_\thetav}[g(Y)]
= \int_\cY p_\thetav(\y) g(\y) d\y,
\end{equation*}
where $Y \in \cY \subseteq \R^M$ and $g \colon \cY \to \R$.
Then,
\begin{equation*}
\nabla E(\thetav)
= \EE_{Y \sim p_\thetav}[g(Y) \nabla_\thetav \log p_\thetav(Y)].
\end{equation*}
\end{boxprop}
\begin{proof}
\begin{equation*}
\begin{aligned}
\nabla E(\thetav)
&= \int_{\cY} \nabla_\thetav p_\thetav(\y) g(\y) d\y \\
&= \int_{\cY} p_\thetav(\y) g(\y) \nabla_\thetav \log p_\thetav(\y) d\y \\
&= \EE_{Y \sim p_\thetav}[g(Y) \nabla_\thetav \log p_\thetav(Y)].
\end{aligned}
\end{equation*}
\end{proof}
The gradient of the log-PDF \wrt $\thetav$, $\nabla_\thetav \log p_\thetav(\y)$,
is known as the \textbf{score function}, hence the estimator name.
SFE is suitable when two requirements are met: it is easy to sample from
$p_\thetav$ and the score function is available in closed form.
Since the SFE gradient is an expectation,
we can use Monte-Carlo estimation to compute an unbiased
estimator of $\nabla E(\thetav)$:
\begin{equation}
\nabla E(\thetav)
\approx
\widehat{\gammav}_N(\thetav)
\coloneqq
\frac{1}{N} \sum_{i=1}^N g(Y_i) \nabla_\thetav \log p_\thetav(Y_i),
\label{grad_est:eq:gamma_n_def}
\end{equation}
where $Y_1, \dots, Y_N$ are sampled from $p_\thetav$.

Interestingly, the gradient of $g$ is not needed in this estimator. Therefore,
there is no differentiability assumption about $g$. This is why SFE is useful
when $g$ is a discrete loss function or more generally a blackbox function.

\begin{boxexm}{SFE with a language model}
In a language model, the probability of a sentence 
$\y = (y_1, \dots, y_L)$
is typically factored using the chain rule of probability
(see \cref{gm:sec:chain_rule_proba})
\begin{equation*}
p_\thetav(\y) 
\coloneqq
p_\thetav(y_1) 
p_\thetav(y_2| y_1) 
\dots 
p_\thetav(y_L | y_1, \dots, y_{L-1}),
\end{equation*}
where $p_\thetav$ is modeled using a transformer or RNN.
Note that the probabilities are normalized by construction, so there is no need
for an explicit normalization constant.
Thanks to this factorization, it is easy to sample from $p_\thetav$ 
using ancestral sampling (see \cref{gm:sec:ancestral_sampling})
and 
the log-probability enjoys the simple expression
\begin{align*}
\nabla_\thetav \log p_\thetav(\y) 
& = 
\nabla_\thetav \log p_\thetav(y_1) +
\nabla_\thetav \log p_\thetav(y_2| y_1) + \dots \\
& \quad +
\nabla_\thetav \log p_\thetav(y_L | y_1, \dots, y_{L-1}).
\end{align*}
This gradient is easy to compute, since the token-wise distributions
$p_\thetav(y_j|y_1,\dots,y_{j-1})$
are typically defined using a softargmax.
We can therefore easily compute $\nabla E(\thetav)$ under $p_\thetav$ using
SFE.
This is for instance useful to optimize an expected reward, 
in order to finetune or align a language model
\citep{ziegler_2019}.
\end{boxexm}
Another example when $\nabla_\thetav p_\thetav(\y)$ is available in closed form
is in the context of reinforcement learning, where $p_\thetav(\y)$ is a Markov
Decision Process (MDP) and is called the policy. Applying the SFE
leads to the (vanilla) policy gradient method
\citep{sutton_1999} and can then be used to compute the gradient of an
expected cumulative reward.
However, SFE is more problematic when used with the Gibbs distribution,
due to the explicit normalization constant.
\begin{boxexm}{SFE with a Gibbs distribution}
The Gibbs distribution is parameterized, for $\thetav \in \RR^\mathcal{Y}$,
\begin{align*}
p_\thetav(y) 
\coloneqq \exp(\theta_y /\gamma - A(\thetav))
= \exp(\theta_y / \gamma) / \exp(A(\thetav))
\end{align*}
where we defined the log-partition function
\begin{equation*}
A(\thetav) \coloneqq \log \sum_{y \in \cY} \exp(\theta_y / \gamma).
\end{equation*}
A typical parametrization is $\theta_y = f(\x, y, \w)$ with $f$ the output of
a network on a sample $\x$ with parameters $\w$. We then have 
\begin{equation*}
\log p_\thetav(y)   
= \theta_y / \gamma - A(\thetav),
\end{equation*}
so that
\begin{equation*}
\nabla_\thetav \log p_\thetav(y)   
= \e_y / \gamma - \nabla A(\thetav).
\end{equation*}
We therefore see that $\nabla_\thetav \log p_\thetav(y)$ crucially depends on
$\nabla A(\thetav)$, the gradient of the log-partition. This gradient is
available for some structured sets $\cY$, see e.g.~\citep{mensch_2018}, but not
in general. 
\end{boxexm}
As another example, we apply SFE in \cref{perturb:sec:blackbox}
to derive the gradient of perturbed functions.

\subsubsection*{Differentiating through both the distribution and the function}

Suppose both the distribution and the function now depend on $\thetav$.
When $g$ is scalar-valued and differentiable w.r.t. $\thetav$,
we want to differentiate
\begin{equation*}
E(\thetav) \coloneqq
\EE_{Y \sim p_\thetav} [g(Y, \thetav)].
\end{equation*}
Using the product rule, we obtain
\begin{equation*}
\nabla E(\thetav) =
\EE_{Y \sim p_\thetav} [g(Y, \thetav) \nabla_\thetav \log p_\thetav(Y)]
+
\EE_{Y \sim p_\thetav} [\nabla_\thetav g(Y, \thetav)].
\end{equation*}

\subsubsection*{Differentiating through joint distributions}

Suppose we now want to differentiate through
\begin{equation*}
E(\thetav) \coloneqq \EE_{Y_1 \sim p_\thetav, Y_2 \sim q_\thetav}[g(Y_1,Y_2)].
\end{equation*}
The gradient is then given by
\begin{equation*}
\nabla E(\thetav) 
= \EE_{Y_1 \sim p_\thetav, Y_2 \sim q_\thetav}[(\nabla_\thetav \log p_\thetav(Y_1) +
\nabla_\thetav \log q_\thetav(Y_2)) g(Y_1,Y_2)],
\end{equation*}
which is easily seen by applying \cref{grad_est:prop:sfe} on the joint
distribution $\rho_\thetav \coloneqq p_\thetav \cdot q_\thetav$.
The extension to more than two variables is straightforward.

\subsection{Variance reduction}

We discuss in this section various techniques for reducing the variance of
estimators of $\nabla E(\thetav)$, where
$E(\thetav) \coloneqq \EE_{Y \sim p_\thetav}[g(Y)]$.

\subsubsection*{Bias and variance}

Recall the definition of $\widehat{\gammav}_N$ 
in \cref{grad_est:eq:gamma_n_def}.
SFE is an \textbf{unbiased} estimator, meaning that
\begin{equation*}
\nabla E(\thetav)
= 
\EE[\widehat{\gammav}_N(\thetav)],
\end{equation*}
where the expectation is taken with respect to the $N$ samples drawn.  
Since the gradient is vector-valued, 
we need to define a scalar-valued notion of variance.  
We do so by using the squared Euclidean distance in the usual
variance definition to define
\begin{align*}
\VV[\widehat{\gammav}_N(\thetav)]
&\coloneqq \EE [\|\widehat{\gammav}_N(\thetav) - \nabla E(\thetav)\|_2^2] \\
&= \EE[\|\widehat{\gammav}_N(\thetav)\|_2^2] - \|\nabla E(\thetav)\|^2_2.
\end{align*}
The variance naturally goes to zero as $N \to \infty$.

\subsubsection*{Baseline}

SFE is known to suffer from 
\textbf{high variance} \citep{mohamed_2020}.
This means that this estimator may require us to draw many samples from the
distribution $p_\thetav$ to work well in practice.
One of the simplest variance reduction technique 
consists in shifting the function $g$ with a constant $\beta$, called a
\textbf{baseline}, to obtain
\begin{equation*}
\nabla E(\thetav)
= \EE_{Y \sim p_\thetav}[(g(Y) - \beta) \nabla_\thetav \log p_\thetav(Y)].
\end{equation*}
The reason this is still a valid estimator of $\nabla E(\thetav)$
stems from
\begin{align*}
\EE_{Y \sim p_\thetav}[
\nabla_\thetav \log p_\thetav(Y)
] 
&= \EE_{Y \sim p_\thetav}\left[
\frac{\nabla_\thetav p_\thetav(Y)}{p_\thetav(Y)}
\right] \\
&= \nabla_\thetav \EE_{Y \sim p_\thetav}[\ones] \\
&= \nabla_\thetav \ones \\
&= \zeros,
\end{align*}
for any valid distribution $p_\thetav$.
The baseline $\beta$ is often set to the running average of past values of the
function $g$, though it is neither optimal nor does it guarantee to lower the
variance \citep{mohamed_2020}. 

\subsubsection*{REINFORCE leave-one-out (RLOO)}

A simple way to reduce the variance is REINFORCE leave-one-out (RLOO)
\citep{kool2019buy,ahmadian2024back}. The key idea is to draw multiple
independent samples $Y_1, \dots, Y_N$ from $p_\thetav$ and for each sample
$Y_i$, to use the remaining samples $(Y_j)_{j \neq i}$ as a baseline. The RLOO
gradient estimator is defined as
\begin{equation*}
\hat{\Gamma}_N(\thetav) \coloneqq
\frac{1}{N} \sum_{i=1}^N 
\big(g(Y_i) - \frac{1}{N-1} \sum_{j \neq i} g(Y_j)\big)
\nabla_\thetav \log p_\thetav(Y_i).
\end{equation*}
This is an unbiased estimator, meaning that
\begin{equation*}
\nabla E(\thetav) = \EE[\hat{\Gamma}_N(\thetav)],
\end{equation*}
where the expectation is taken over $Y_1, \dots, Y_N$. This comes from the
independence of $Y_i, Y_j$ for $i\neq j$, such that $\EE[g(Y_j)
\nabla_\thetav\log p_\thetav(Y_i)] = \EE[g(Y_j)]\EE[\nabla_\thetav\log
p_\thetav(Y_i)]= 0$ since $\EE[\nabla_\thetav\log p_\thetav(Y_i)]= 0$ by
Bartlett's first identity~\eqref{grad_est:eq:bartlett_first}. 

It can be checked that the estimator can equivalently be rewritten as
\begin{equation*}
\hat{\Gamma}_N(\thetav) =
\frac{1}{N-1} \sum_{i=1}^N 
\big(g(Y_i) - \frac{1}{N} \sum_{j=1}^N g(Y_j)\big)
\nabla_\thetav \log p_\thetav(Y_i).
\end{equation*}
This is slightly more convenient to implement, as we can use
$\beta \coloneqq \frac{1}{N} \sum_{j=1}^N g(Y_j)$ as a shared baseline across
samples.

\subsubsection*{Control variates}

Another general technique are \textbf{control variates}.
Let us denote the expectation of a function
$h \colon \RR^M \to \RR$ under the distribution $p_\thetav$ as
\begin{equation*}
H(\thetav) \coloneqq \EE_{Y \sim p_\thetav}[h(Y)].
\end{equation*}
Suppose that $H(\thetav)$ and its gradient $\nabla H(\thetav)$ are known in
closed form.
Then, for any $\gamma \ge 0$, we clearly have
\begin{align*}
E(\thetav)
&= \EE_{Y \sim p_\thetav}[g(Y)] \\
&= \EE_{Y \sim p_\thetav}[g(Y) - \gamma (h(Y) - H(\thetav))] \\
&= \EE_{Y \sim p_\thetav}[g(Y) - \gamma h(Y)] + \gamma H(\thetav)
\end{align*}
and therefore
\begin{align*}
\nabla E(\thetav)
&= \nabla_\thetav \EE_{Y \sim p_\thetav}[g(Y) - \gamma h(Y)]
+ \gamma \nabla H(\thetav).
\end{align*}
Applying SFE, we then obtain
\begin{equation*}
\nabla E(\thetav)
= \EE_{Y \sim p_\thetav}[(g(Y) - \gamma h(Y)) \nabla_\thetav \log p_\thetav(Y)]
+ \gamma \nabla H(\thetav).
\end{equation*}
Examples of function $h$ include a bound on $f$ or a second-order Taylor
expansion of $f$, assuming that these approximations are easier to integrate
than $f$ \citep{mohamed_2020}.

\subsection{Vector-valued functions}

It is straightforward to extend the SFE to vector-valued functions.
\begin{boxprop}{SFE for vector-valued functions}\label{grad_est:prop:jvp_sfe}
Given a family of distributions $p_\thetav$ on $\cY$, for $\theta \in \Theta$,
define
\begin{equation*}
E(\thetav) 
\coloneqq \EE_{Y \sim p_\thetav}[g(Y)]
= \int_\cY p_\thetav(\y) g(\y) d\y,
\end{equation*}
where 
$Y \in \cY$, 
$g \colon \cY \to \cG$.
The JVP of $E$ at $\thetav\in \Theta$ along $\v\in \Theta$ is 
\begin{equation*}
\partial E(\thetav) \v
= 
\EE_{Y \sim p_\thetav}[\langle \nabla_\thetav \log p_\thetav(Y), \v \rangle
g(Y)] \in \cG
\end{equation*}
and the VJP of $E$ at $\thetav\in \Theta$ along $\u \in \cG$ is 
\begin{equation*}
  \partial E(\thetav)^* \u
  = 
  \EE_{Y \sim p_\thetav}[\nabla_\thetav \log p_\thetav(Y) \langle\u, g(Y)
  \rangle] \in \Theta.
\end{equation*}
The Jacobian of $E$ at $\thetav \in \Theta$ can then be written as
\begin{equation*}
  \jac E(\thetav)
  = 
  \EE_{Y \sim p_\thetav}[g(Y) \otimes \nabla_\thetav \log p_\thetav(Y)],
\end{equation*}
where $\otimes$ denotes the outer product.
\end{boxprop}
\begin{proof}
The VJP of $E$ at $\thetav \in \Theta$ along $\u \in \cG$ amounts to
computing the gradient of the scalar function 
\[
  \langle E(\thetav), \u\rangle
  = \EE_{Y \sim p_\thetav}[\langle g(Y), \u\rangle]
\]
The expression of the VJP follows by using the SFE on the scalar valued
integrand $\langle g(Y), \u\rangle$. The JVP is obtained as the adjoint operator
of the VJP and the Jacobian follows.
\end{proof}

\subsubsection*{Differentiating through both the distribution and the function}

If $\thetav$ now influences both the distribution and the function,
\begin{equation*}
E(\thetav) \coloneqq
\EE_{Y \sim p_\thetav} [g(Y, \thetav)],
\end{equation*}
then, we obtain
\begin{equation*}
\partial E(\thetav) =
\EE_{Y \sim p_\thetav} 
[g(Y, \thetav) \otimes \nabla_\thetav \log p_\thetav(Y)]
+
\EE_{Y \sim p_\thetav} [\partial_\thetav g(Y, \thetav)].
\end{equation*}

\subsection{Second derivatives}

Using the previous subsection with
$g(\y, \thetav) = g(\y) \nabla_\thetav \log p_\thetav(\thetav)$,
we easily obtain an estimator of the Hessian.
\begin{boxprop}{SFE for the Hessian}
\label{grad_est:prop:sfe_hessian}
Let us define the scalar-valued function
$E(\thetav) \coloneqq \EE_{Y \sim p_\thetav}[g(Y)]$.
Then,
\begin{align*}
\nabla^2 E(\thetav)
=&
\EE_{Y \sim p_\thetav} 
[g(Y) \nabla_\thetav \log p_\thetav(Y) \otimes \nabla_\thetav \log p_\thetav(Y)]
+ \\
 &\EE_{Y \sim p_\thetav} 
[g(Y) \nabla^2_\thetav \log p_\thetav(Y)].
\end{align*}
\end{boxprop}
This can also be derived using the second-order log-derivative
\begin{equation*}
\nabla^2_\thetav \log p_\thetav(\y)
= \frac{1}{p_\thetav(\y)} \nabla^2_\thetav p_\thetav(\y)
- \frac{1}{p_\thetav(\y)^2} \nabla_\thetav p_\thetav(\y) \otimes \nabla_\thetav
p_\thetav(\y)
\end{equation*}
so that
\begin{equation*}
\nabla^2_\thetav p_\thetav(\y)
=
p_\thetav(\y)
\left[
\nabla^2_\thetav \log p_\thetav(\y)
+ \nabla_\thetav \log p_\thetav(\y) \otimes \nabla_\thetav \log p_\thetav(\y)
\right].
\end{equation*}

\subsubsection*{Link with the Bartlett identities}

The Bartlett identities are expressions relating the moments of the score
function (gradient of the log-likelihood function). Using 
\cref{grad_est:prop:sfe} with $g(\y) = 1$ 
and $\int_\cY p_\thetav(\y) d\y = 1$, we obtain
\begin{equation}
\EE_{Y \sim p_\thetav}[\nabla_\thetav \log p_\thetav(Y)]
=
\zeros,
\label{grad_est:eq:bartlett_first}
\end{equation}
which is known as Bartlett's first identity.
Similarly, using \cref{grad_est:prop:sfe_hessian}, we obtain
\begin{equation}
\label{grad_est:eq:bartlett_second}
\begin{aligned}
&\EE_{Y \sim p_\thetav} 
[\nabla^2_\thetav \log p_\thetav(Y)]
+
\EE_{Y \sim p_\thetav} 
[\nabla_\thetav \log p_\thetav(Y) \otimes \nabla_\thetav \log p_\thetav(Y)] \\
=& \EE_{Y \sim p_\thetav} 
[\nabla^2_\thetav \log p_\thetav(Y)]
+
\mathrm{cov}[\log p_\thetav(Y)] \\
=& \zeros,
\end{aligned}
\end{equation}
which is known as Bartlett's second identity.

\subsection{Autodiff-friendly implementation}

So far, we derived expressions of score function estimators
for first and second derivatives.
From a user perspective, instead of manually implementing these expressions,
it would be more convenient if we could just call
autodiff on some function and recover the correct expressions of the score
function estimators.
This can be achieved using the so-called \textbf{magicbox operator}
\citep{foerster2018dice}. This operator is defined as
\begin{equation*}
\cM(\tau) \coloneqq \exp(\tau - \cS(\tau)), 
\end{equation*}
where $\cS$ is the \textbf{stop gradient} operator.

During the evaluation of a program, the stop gradient operator acts as the
identity. During automatic differentiation, that is, during forward or reverse
mode autodiff, this operator stops any flow of derivatives. In other words, its
Jacobian is zero on any inputs despite evaluating as the identity. Such an
operator does not exist mathematically, but it can be defined in an autodiff
framework using \textbf{custom derivation rules}. Evaluations and derivatives of
programs making $\cS$ intervene are then to be understood first in terms of the
autodiff framework and its custom rules.

Coming back to the magicbox operator, its evaluation and its derivatives in an
autodiff framework are then for any $\tau \in \cR$
\begin{align*}
\cM(\tau) & = \exp(\tau - \tau) = 1\\
\cM'(\tau) & = (\tau - \cS(\tau))' \cM(\tau) = 1 \cdot 1 = 1\\ 
\cM''(\tau) & = \dots = 1
\end{align*}
The ``magic'' happens when we compose the magicbox operator with another
function.  Consider a (twice) differentiable function $f \colon \RR^M \to \RR$.
We then have
\begin{align*}
(\cM \circ f)(\thetav) &= 1 \\
\nabla (\cM \circ f)(\thetav) 
&= \cM'(f(\thetav))\nabla f(\thetav)
= \nabla f(\thetav) \\
\nabla^2 (\cM \circ f)(\thetav) 
&= \cM''(f(\thetav)) \nabla f(\thetav) \otimes \nabla f(\thetav) +
\cM'(f(\thetav)) \nabla^2 f(\thetav) \\
&= \nabla f(\thetav) \otimes \nabla f(\thetav) + \nabla^2 f(\thetav).
\end{align*}
Let $Y$ be a sample from $p_\thetav$ and let us define $f(\thetav) \coloneqq
\log p_\thetav(Y)$. Then,
\begin{align*}
(\cM \circ f)(\thetav) &= 1 \\
\nabla (\cM \circ f)(\thetav) 
&= \nabla_\thetav \log p_\thetav(Y) \\
\nabla^2 (\cM \circ f)(\thetav) 
&= \nabla_\thetav \log p_\thetav(Y) \otimes \nabla_\thetav \log p_\thetav(Y) 
+ \nabla^2_\thetav \log p_\thetav(Y),
\end{align*}
correctly recovering the expressions for the first derivative in
\cref{grad_est:prop:sfe} and for the second derivative in
\cref{grad_est:prop:sfe_hessian}.  Let us define the functions
\begin{align*}
E(\thetav) &\coloneqq \EE_{Y \sim p_\thetav}[g(Y)] \\
\bar{E}(\thetav) &\coloneqq \EE_{Y \sim \cS(p_\thetav)}[g(Y) \cM(\log p_\thetav(Y))].
\end{align*}
We then correctly obtain for all $\thetav \in \RR^M$
\begin{align*}
    E(\thetav) &= \bar{E}(\thetav) \\
    \nabla E(\thetav) &= \nabla \bar{E}(\thetav) \\
    \nabla^2 E(\thetav) &= \nabla^2 \bar{E}(\thetav).
\end{align*}
The key benefit is that $\bar{E}$ is compatible with autodiff frameworks, while
$E$ is not.
In practice, we can draw $N$ samples $Y_1, \dots, Y_N$ (without
differentiating through the sampling) and define
\begin{equation*}
\bar{E}_N(\thetav) \coloneqq \frac{1}{N} \sum_{i=1}^N g(Y_i) \cM(\log
p_\thetav(Y_i)).
\end{equation*}
By Monte-Carlo estimation, we then have
\begin{align*}
    E(\thetav) &= \EE[\bar{E}_N(\thetav)] \\
    \nabla E(\thetav) &= \EE[\nabla \bar{E}_N(\thetav)] \\
    \nabla^2 E(\thetav) &= \EE[\nabla^2 \bar{E}_N(\thetav)],
\end{align*}
where the expectation is taken over $Y_1, \dots, Y_N$.

\section{Path gradient estimators, reparametrization trick}
\label{grad_est:sec:pge}

As we saw previously, the main difficulty in computing gradients of expectations
arises when the parameters $\thetav$ play a role in the distribution $p_\thetav$
being sampled.
The key idea of path gradient estimators (PGE), 
also known as reparametrization trick, 
is to rewrite the expectation in such a way that the parameters are 
moved from the distribution to the function, using a \textbf{change of
variable}.

\subsection{Location-scale transforms}
\label{grad_est:sec:location_scale_transform}

The canonical example of path gradient estimator is 
differentiating through the expectation
\begin{equation*}
E(\mu, \sigma)
\coloneqq \EE_{U \sim \mathrm{Normal}(\mu, \sigma^2)}[g(U)],
\end{equation*}
where 
$g \colon \RR \to \RR$ 
is a differentiable function.
If we let $Z \sim \mathrm{Normal}(0, 1)$,
it is easy to check that $U = \mu + \sigma Z$.
We can therefore write
\begin{equation*}
E(\mu, \sigma) 
= \EE_{Z \sim \mathrm{Normal}(0, 1)}[g(\mu + \sigma Z)].
\end{equation*}
The key advantage is that we can now easily compute the derivatives by mere
application of the chain rule,
since the parameters $\mu$ and $\sigma$ are moved from the distribution to the
function:
\begin{align*}
\frac{\partial}{\partial \mu} E(\mu, \sigma) 
&= \EE_{Z \sim \mathrm{Normal}(0, 1)}[g'(\mu + \sigma Z)] \\
\frac{\partial}{\partial \sigma} E(\mu, \sigma) 
&= \EE_{Z \sim \mathrm{Normal}(0, 1)}[Z \cdot g'(\mu + \sigma Z)].
\end{align*}
The change of variable
\begin{equation}
U \coloneqq \mu + \sigma Z
\label{grad_est:eq:loc_scale_transform}
\end{equation}
is called a \textbf{location-scale transform}.
Such a transformation exists, not only for the normal distribution,
but for \textbf{location-scale family} distributions,
i.e., distributions parametrized by a location parameter $\mu$ and a scale
parameter $\sigma > 0$, such that $U$ is distributed according to a 
distribution in the same family as $Z$ is distributed.
Besides the normal distribution, examples of location-scale family 
distributions include
the Cauchy distribution,
the uniform distribution,
the logistic distribution,
the Laplace distribution,
and Student's $t$-distribution.

We can easily relate the cumulative distribution function (CDF) and the
probability density function (PDF) of $Z$ to that of $U$, and vice-versa.
\begin{boxprop}{CDF and PDF of location-scale family distributions}
\label{grad_est:prop:cdf_pdf_location_scale}

Let $F_Z(z) \coloneqq \PP(Z \le z)$ and $f_Z(z) \coloneqq F_Z'(z)$.
If $U \coloneqq \mu + \sigma Z$, then
\begin{align*}
    F_Z(z) &= F_U(\mu + \sigma z) 
    \iff
    F_U(u) = F_Z\left(\frac{u - \mu}{\sigma}\right) \\
    f_Z(z) &= \sigma f_U(\mu + \sigma z) 
    \iff
    f_U(u) = \frac{1}{\sigma} f_Z\left(\frac{u - \mu}{\sigma}\right).
\end{align*}
\end{boxprop}
\begin{proof}
We have
\begin{align*}
F_Z(z) 
&= \PP(Z \le z) \\
&= \PP\left(\frac{U-\mu}{\sigma} \le z\right) \\
&= \PP(U \le \mu + \sigma z) \\
&= F_U(\mu + \sigma z)
\end{align*}
and we obtain $f_Z(z)$ by differentiating $F_Z(z)$.
\end{proof}

\subsection{Differentiable transforms}

We can generalize the idea of path gradient estimator (PGE)
to any change of variable
\begin{equation*}
U \coloneqq T(Z, \thetav),
\end{equation*}
where $T \colon \RR^M \times \RR^Q \to \RR^M$ is a differentiable
transformation.
For example, if we gather $\mu$ and $\sigma$ as 
$\thetav \coloneqq (\mu, \sigma)$,
we can write the location-scale transform as
\begin{equation*}
U 
= T(Z, \thetav) 
= \mu + \sigma Z.
\end{equation*}
We can derive the path gradient estimator for any such differentiable
transformation $T$.
\begin{boxprop}{Path gradient estimator}
\label{grad_est:prop:pge}
Let us define
\begin{equation*}
E(\thetav) \coloneqq \EE_{U \sim p_\thetav}[g(U)],
\end{equation*}
where $U \in \cU \subseteq \RR^M$ and $g \colon \RR^M \to \RR$ is
differentiable.
Suppose there is a differentiable transformation 
$T \colon \R^M \times \R^Q \to \R^M$ such that
if $Z \sim p$ (where $p$ does not depend on $\thetav$) 
and 
$U \coloneqq T(Z, \thetav)$,
then $U \sim p_\thetav$.
Then, we have
\begin{equation*}
E(\thetav) 
= \EE_{Z \sim p}[h(Z, \thetav)]
= \EE_{Z \sim p}[g(T(Z, \thetav))],
\end{equation*}
where $h(\z, \thetav) \coloneqq g(T(\z, \thetav))$.
This implies
\begin{equation*}
\begin{aligned}
\nabla E(\thetav) 
&= \EE_{Z \sim p}[\nabla_2 h(Z, \thetav)] \\
&= \EE_{Z \sim p}[\partial_2 T(Z, \thetav)^* \nabla g(T(Z, \thetav))].
\end{aligned}
\end{equation*}
\end{boxprop}
The path gradient estimator (\aka reparametrization trick) 
gives an \textbf{unbiased estimator} of $\nabla E(\thetav)$.
It has however two key disadvantages. First, it assumes that $g$ is
\textbf{differentiable} (almost everywhere), 
which may not always be the case. Second, it assumes that $g$
is well-defined on $\R^M$, not on $\cU$, which could be problematic for some
discrete loss functions, such as the zero-one loss function or 
ranking loss functions.

As an example of differentiable transform,
in machine learning, we can sample Gaussian noise $Z$ and make it go through a
neural network with parameters $\w$ to generate an image
$X \coloneqq T(Z, \w)$. In statistics, many distributions are related to each
other through differentiable transforms, as we recall below.
\begin{boxexm}{Some differentiable transforms in statistics}

We give below a non-exhaustive list of differentiable transform examples.
\begin{itemize}

\item If $X \sim \mathrm{Normal}(\mu, \sigma^2)$,
then $\exp(X) \sim \mathrm{Lognormal}(\mu, \sigma^2)$.

\item If $U \sim \mathrm{Uniform}(0,1)$,
then $-\log(U) / \lambda \sim \mathrm{Exponential}(\lambda)$.

\item If $X_1,\dots,X_N \sim \mathrm{Exponential}(\lambda)$ (i.i.d.),
then $\sum_{i=1}^N X_i \sim \mathrm{Gamma}(N, \lambda)$.

\item If $X_i \sim \mathrm{Gamma}(\alpha_i,\theta)$ for $i \in [K]$,
then 
$\left(
\frac{X_1}{\sum_{i=1}^K X_i}, 
\dots, 
\frac{X_K}{\sum_{i=1}^K X_i}
\right) 
\sim 
\mathrm{Dirichlet}(\alpha_1,\dots,\alpha_K)$.

\end{itemize}
\end{boxexm}

\subsection{Inverse transforms}
\label{grad_est:sec:its}

The inverse transform method can be used for sampling from a probability
distribution, given access to its associated \textbf{quantile function}. Recall
that the cumulative distribution function (CDF) associated with a random
variable $Y$ is the function $F_Y \colon \RR \to [0,1]$ defined by
\begin{equation*}
F_Y(y) \coloneqq \PP(Y \le y). 
\end{equation*}
The quantile function is then a function $Q_Y \colon [0,1] \to \RR$ such
that $Q_Y(\pi) = y$ for $\pi = F_Y(y)$. Assuming $F_Y$ is continuous and strictly
increasing, we have that $Q_Y$ is the \textbf{inverse CDF},
\begin{equation*}
Q_Y(\pi) = F^{-1}_Y(\pi). 
\end{equation*}
In the general case of CDF functions that are not strictly increasing,
the quantile function is usually defined as
\begin{equation*}
Q_Y(\pi) \coloneqq \inf \{y \in \RR \colon \pi \le F_Y(y) \}.
\end{equation*}
Given access to the quantile function $Q_Y(\pi)$ 
associated with a distribution $p$, 
inverse transform sampling allows us to sample from $p$ by first
drawing a sample from the \textbf{uniform distribution} and 
then making this sample go through the quantile function.
\begin{boxprop}{Inverse transform sampling}
Suppose $Y \sim p$, \\ where $p$ is a distribution with quantile function 
$Q_Y$. If $U \sim \text{Uniform}(0,1)$, then $Q_Y(U) \sim p$.
\end{boxprop}
\begin{proof}
  If $\pi \leq F_Y(t)$, then by definition of $Q_Y$, $Q_Y(\pi) \leq t$. If $\pi
  \geq F_Y(t)$, then by definition of $Q_Y$, $F_Y(Q_Y(\pi)) \geq \pi$, so
  $F_Y(Q_Y(\pi)) \geq F_Y(t)$ and since a CDF is always non-decreasing,
  $Q_Y(\pi) \geq t$. Hence, we have, $Q_Y(\pi) \leq t \iff \pi \leq F_Y(t)$, so
  \begin{align*}
    \PP(Q_Y(U)\leq t) & = \PP(U \leq F_Y(t)) \\
    & = F_Y(t).
  \end{align*}
  The CDFs of $Q_Y(U)$ and $Y$ coincide, hence they have the same distribution.
\end{proof}

If the quantile function is differentiable, we can therefore use it as a
\textbf{transformation} within the \textbf{reparametrization trick}.
Indeed, if $Y \sim p_\thetav$, where $p_\thetav$ is a distribution with
parameter $\thetav$ and quantile function $Q_Y(\pi, \thetav)$,
then we have
\begin{equation*}
E(\thetav)
= \EE_{Y \sim p_\thetav}[g(Y)]
= \EE_{\pi \sim \text{Uniform}(0,1)}[g(Q_Y(\pi, \thetav))]
\end{equation*}
and therefore,
by the reparametrization trick (\cref{grad_est:prop:pge}),
\begin{equation*}
\nabla E(\thetav)
= \EE_{\pi \sim \text{Uniform}(0,1)}[
  \partial_2 Q_Y(\pi, \thetav)^* \nabla g(Q_Y(\pi, \thetav))
  ].
\end{equation*}
\begin{boxexm}{Examples of quantile functions}
If \\ $Y \sim \text{Exponential}(\lambda)$,
the CDF of $Y$ is $\pi = F_Y(y) = 1 - \exp(-\lambda y)$ for $y \ge 0$
and therefore the quantile function is
$Q_Y(\pi, \lambda) = -\frac{\log(1-\pi)}{\lambda}$.
If $Y \sim \text{Normal}(\mu, \sigma^2)$, 
the CDF is $F_Y(y) = \frac{1}{2}\left[1 + \text{erf}\left(\frac{y -
\mu}{\sigma\sqrt{2}}\right)\right]$ and the quantile function is
$Q_Y(\pi, \thetav) = \mu + \sigma \sqrt{2} \cdot \mathrm{erf}^{-1}(2\pi - 1)$,
where $\thetav = (\mu, \sigma)$. This therefore defines an alternative
transformation to the location-scale
transformation in \cref{grad_est:eq:loc_scale_transform}.
\end{boxexm}
Note that, in the above example, the error function $\mathrm{erf}$ and its
inverse do not enjoy analytical expressions but autodiff packages usually
provide numerical routines to compute them and differentiate through them.
Nonetheless, one caveat of the inverse transform is that it indeed requires
access to (approximations of) the quantile function and its derivatives, which
may be difficult for complicated distributions.

\subsection{Pushforward operators}
\label{grad_est:sec:push_forward}

\subsubsection*{Pushforward distributions}

We saw so far that the reparametrization trick
is based on using a change of variables in order to differentiate 
an expectation \wrt the parameters of the distribution.
In this section,
we further formalize that approach using pushforward distributions.
\begin{boxdef}{Pushforward distribution}
Suppose $Z \sim p$, where $p$ is a distribution over $\cZ$. 
Given a continuous map $T \colon \cZ \to \cU$, 
the pushforward distribution of $p$ through $T$
is the distribution $q$ according to which 
$U \coloneqq T(Z) \in \cU$ is distributed, 
i.e., $U \sim q$.
\end{boxdef}
Although not explicit in the above, the transformation $T$ can depend on some
learnable parameters, for example if $T$ is a neural network.
Intuitively, the pushforward distribution is obtained by moving the position of
all the points in the support of $p$.
We give a few examples below.
\begin{itemize}
\item Inverse transform sampling studied in \cref{grad_est:sec:its} can be
seen as performing the pushforward of the uniform distribution through $T = Q$,
where $Q$ is the quantile function.

\item The Gumbel trick studied in \cref{pert:sec:gumbel_trick}
can be seen as the pushforward of Gumbel noise through $T =
\mathrm{argmax}$ (a discontinuous function).

\item Gumbel noise can itself be obtained by pushing forward 
the uniform distribution through
$T = -\log(-\log(\cdot))$ (\cref{pert:rem:gumbel_noise}).

\item In a generative modeling setting, as we mentioned previously,
we use the pushforward of Gaussian noise through a parametrized transformation
$X = T(Z, \w)$ called a generator, typically a neural network. 

\item It is possible to define distributions over (sparse) probability vectors
    by sampling then projecting \citep{farinhas2021sparse}.
\end{itemize}
A crucial aspect of the pushforward distribution $q$ is that it can be
\textbf{implicitly} defined, meaning that we do not necessarily need to know the
explicit form of the associated PDF. In fact, it is easy to
\textbf{sample} from $q$, provided that it is easy to sample from $p$:
\begin{align*}
U \sim q
\iff
Z \sim p, U &\coloneqq T(Z).
\end{align*}
Hence the usefulness of the pushforward distribution in \textbf{generative
modeling}. Furthermore, if $p$ has associated PDF $p_Z$, we can compute the
expectation of a function $f$ according to $q$ as
\begin{align*}
\EE_{U \sim q}[f(U)] 
= \EE_{Z \sim p}[f(T(Z))]
= \int_{\cZ} f(T(\z)) p_Z(\z) d\z,
\end{align*}
even though we do not know the explicit form of the PDF of $q$.

\subsubsection*{Pushforward measures}

More generally, we can define the notion of pushforward, in the language of
measures. Denote $\cM(\cZ)$ the set of measures on a set $\cZ$. A
\textbf{measure} $\alpha \in \cM(\cZ)$, that has a density $d\alpha(\z)
\coloneqq p_Z(\z) d\z$, can be integrated against a function $f$ as
\begin{equation*}
\int_\cZ f(\z) d\alpha(\z) = \int_\cZ f(\z) p_Z(\z) d\z.
\end{equation*}
A measure $\alpha$ is called a probability measure if it is positive and
satisfies $\alpha(\cZ) = \int_\cZ d\alpha(\z) = \int_\cZ p_Z(\z) d\z = 1$.
See \citet[Chapter 2]{peyre_2019} for a concise introduction.
\begin{boxdef}{Pushforward operator and measure}
Given a continuous map $T \colon \cZ \to \cU$ and some measure 
$\alpha \in \cM(\cZ)$,
the pushforward measure $\beta = T_\sharp \alpha \in \cM(\cU)$ 
is such that for all continuous functions $f \in \cC(\cU)$
\begin{equation*}
\int_\cU f(\u) d\beta(\u) = \int_\cZ f(T(\z)) d\alpha(\z). 
\end{equation*}
Equivalently, for any measurable set $\cA \subset \cU$, we have
\begin{equation*}
\beta(\cA) = \alpha(\{\z \in \cZ \colon T(\z) \in \cA\}) 
= \alpha(T^{-1}(\cA)),
\end{equation*}
where $T^{-1}(\cA) = \{\z \in \cZ: T(\z)\in \cA\}$.
\end{boxdef}
Importantly, the pushforward operator preserves positivity and mass, therefore
if $\alpha$ is a probability measure, then so is $T_\sharp \alpha$.
The pushforward of a probability measure therefore defines a
pushforward distribution (since a distribution can be parametrized by 
a probability measure).

\subsection{Change-of-variables theorem}
\label{grad_est:sec:change_of_var_theorem}

We saw that a pushforward distribution associated with a variable $U$
is implicitly defined through a
transform $U \coloneqq T(Z)$ and can be easily sampled from as long as it is 
easy to sample $Z$. However, in some applications (e.g., density estimation), 
we may want to know the PDF associated with $U$.
Assuming the transform $T$ is invertible, 
we have $Z = T^{-1}(U)$ and therefore
for $\cA \subseteq \cU$, we have
\begin{align*}
\PP(U \in \cA) 
= \PP(Z \in T^{-1}(\cA)) 
= \int_{T^{-1}(\cA)} p_Z(\z) d\z.
\end{align*}
Using the \textbf{change-of-variables theorem} from multivariate calculus,
assuming $T^{-1}$ is available, we can give an explicit formula for the PDF of
the pushforward distribution, see e.g.~\citep{schwartz1954formula,
taylor2002differential}.
\begin{boxprop}{PDF of the pushforward distribution}
Suppose \\
$Z \sim p$, where $p$ is a distribution over $\cZ$, with PDF $p_Z$. 
Given a \textbf{diffeomorphism} $T \colon \cZ \to \cU$ (i.e., an invertible and
differentiable map), 
the pushforward distribution of $p$ through $T$ is the distribution $q$ 
such that $U \coloneqq T(Z) \sim q$ and its PDF is
\begin{equation*}
q_U(\u) = |\det(\partial T^{-1}(\u))| p_Z(T^{-1}(\u)),
\end{equation*}
where $\partial T^{-1}(\u)$ is the Jacobian of $T^{-1} \colon \cU \to \cZ$.
\end{boxprop}
Using this formula, we obtain
\begin{align*}
\PP(U \in \cA)  
&= \int_{\cA} p_U(\u) d\u \\
&= \int_{\cA} |\det(\partial T^{-1}(\u))| p_Z(T^{-1}(\u)) d\u.
\end{align*}
Using the inverse function theorem (\cref{implicit:thm:inv_diff}),
we then have
\begin{equation*}
\partial T^{-1}(\u)
= (\partial T(T^{-1}(\u)))^{-1},
\end{equation*}
under the assumption that $T(\z)$ is continuously differentiable 
and has invertible Jacobian $\partial T(\z)$.
\textbf{Normalizing flows} are parametrized transformations $T$ designed such that
$T^{-1}$ and its Jacobian $\partial T^{-1}$ are easy to compute; 
see e.g. \citet{kobyzev2019normalizing,papamakarios_2021} for a review.

\section{Stochastic programs}

A stochastic program is a program that involves some form of randomness. In a
stochastic program, the final output, as well as intermediate variables, may
therefore be random variables. In other words, a stochastic program induces a
probability distribution over program outputs, as well as over execution
trajectories.

\subsection{Stochastic computation graphs}

A stochastic program can be represented by a stochastic computation graph
as originally introduced by \citet{schulman2015gradient}. 
Departing from that work, our exposition explicitly supports two types of
intermediate operations: sampling from a \textbf{conditional distribution} or
evaluating a \textbf{function}.  These operations can produce either
\textbf{deterministic} variables or \textbf{random} variables.

\subsubsection*{Function and distribution nodes}

Formally, we define a stochastic computation graph as a directed acyclic graph
$\cG = (\cV, \cE)$, where $\cV = \cV_f \cup \cV_p$,
$\cV_f$ is the set of function nodes and $\cV_p$ is the set of distribution
nodes.  Similarly to computation graphs reviewed in 
\cref{neural_nets:sec:comput_graphs}, 
we number the nodes as $\cV =\{0, 1, \ldots, K\}$.
Node $0$ corresponds to the input $\s_0 \in \cS_0$, which we assume to be
deterministic.
It is the variable with respect to which we wish to differentiate.
Node $K$ corresponds to the program output $S_K \in \cS_K$, which we assume to
be a random variable.
A node $k \in \{1,\dots,K\}$
can either be a \textbf{function node} $k \in \cV_f$ with an
associated function $f_k$
or a \textbf{distribution node} 
$k \in \cV_p$, with associated conditional distribution $p_k$.
A stochastic program has at least one distribution node, the source of
randomness. Otherwise, it is a deterministic program.
As for computation graphs, the set of edges $\cE$ is used to
represent dependencies between nodes.
We denote the parents of node $k$ by $\parent(k)$.

\subsubsection*{Deterministic and random variables}

We distinguish between two types of intermediate variables:
\textbf{deterministic} variables $\s_k$
and \textbf{random} variables $S_k$.
Therefore, a distribution $p_k$ or a function $f_k$ may receive both types of
variables as \textbf{conditioning} or \textbf{input}.
It is then convenient to split $\parent(k)$ as
$\parent(k)= \dparent(k) \cup \rparent(k)$,
where we defined the \textbf{deterministic parents}
$\dparent(k) \coloneqq \{i_1, \ldots, i_{p_k}\}$
and the \textbf{random parents}
$\rparent(k) \coloneqq \{j_1, \ldots, j_{q_k}\}$.
Therefore,
$\s_{i_1}, \ldots, \s_{i_{p_k}}$
are the deterministic parent variables and
$S_{j_1}, \ldots, S_{j_{q_k}}$ 
are the random parent variables, of node $k$.

\subsubsection*{Executing a stochastic program}

We assume that nodes $0, 1, \dots, K$ are in topological order (if this is not
the case, we need to perform a topological sort).
Given parent variables
$\s_{i_1}, \ldots, \s_{i_{p_k}}$
and
$S_{j_1}, \ldots, S_{j_{q_k}}$,
a node $k \in \{1, \dots, K\}$
produces an output as follows.
\begin{itemize}
\item If $k \in \cV_p$ (distribution node), the output is 
\begin{align*}
&S_k
\sim p_k(\cdot \mid \s_{\dparent(k)}, S_{\rparent(k)}) \\
\iff
&S_k
\sim p_k(\cdot \mid \s_{i_1}, \ldots, \s_{i_{p_k}}, S_{j_1}, \ldots, S_{j_{q_k}})
\end{align*}
Note that technically $p_k$ is the distribution of $S_k$ conditioned on its
parents, not the distribution of $S_k$.
Therefore, we should in principle write
$S_k \mid \s_{\dparent(k)}, S_{\rparent(k)}
\sim p_k(\cdot \mid \s_{\dparent(k)}, S_{\rparent(k)} )$.
We avoid this notation for conciseness and for symmetry with function nodes.

Contrary to a function node, a distribution node can have no parents.
That is, if $k \in \cV_p$, it is possible that $\parent(k) = \emptyset$.
A good example would be a parameter-free noise distribution.

\item If $k \in \cV_f$ (function node), the output is in general
\begin{align*}
S_k 
&\coloneqq f_k(\s_{\dparent(k)}, S_{\rparent(k)}) \\
&\coloneqq f_k(\s_{i_1}, \ldots, \s_{i_{p_k}}, S_{j_1}, \ldots, S_{j_{q_k}})
\end{align*}
and in the special case $q_k = |\rparent(k)| = 0$, the output is
\begin{align*}
\s_k 
&\coloneqq f_k(\s_{\dparent(k)}) \\
&\coloneqq f_k(\s_{i_1}, \ldots, \s_{i_{p_k}}).
\end{align*}
\end{itemize}
Unless the associated conditional distribution $p_k$ is a delta distribution,
that puts all the probability mass on a single point, the output of a
distribution node $k \in \cV_p$ is necessarily a random variable 
$S_k \in \cS_k$.
For function nodes $k \in \cV_f$, the output of the function $f_k$ is a random
variable $S_k \in \cS_k$ if at least one of the parents of $k$
produces a random variable. Otherwise, if all
parents of $k$ produce deterministic variables, the output
of $f_k$ is a deterministic variable $\s_k \in \cS_k$. 

The entire procedure is summarized in \cref{grad_est:algo:execute_sp}.
We emphasize that $S_K = f(\s_0) \in \cS_K$ is a random variable.
Therefore, a stochastic program (implicitly) induces
a distribution over $S_K$, and also over intermediate random variables $S_k$.
Executing the stochastic program allows us to draw samples from that
distribution.

\begin{algorithm}[t]\caption{Executing a stochastic program}
\label{grad_est:algo:execute_sp}
\begin{algorithmic}[1]
  \Statex \textbf{Nodes:} $1,\dots,K$ in topological order,
  where node $k$ is either a function
  $f_k$ or a conditional distribution $p_k$
  \Statex {\bf Input:} input $\s_0 \in \cS_0$
  \For{$k \coloneqq 1, \ldots, K$}
    \State Retrieve 
    $\parent(k) = \dparent(k) \cup \rparent(k)$
    \If{$k \in \cV_p$} \Comment{Distribution node}
      \State $S_k \sim p_k(\cdot|\s_{\dparent(k)},S_{\rparent(k)})$
    \ElsIf{$k \in \cV_f$} \Comment{Function node}
      \If{$|\rparent(k)| \neq 0$} 
        \State 
          $S_k \coloneqq 
          f_k(\s_{\dparent(k)}, S_{\rparent(k)})$
          \Comment{Output is a R.V.}
        \ElsIf{$|\rparent(k)| = 0$} 
          \State $\s_k \coloneqq f_k(\s_{\dparent(k)})$ 
          \Comment{Output is deterministic} 
      \EndIf
    \EndIf
  \EndFor
  \State {\bf Output:} $f(\s_0) \coloneqq S_K \in \cS_K$
  \end{algorithmic}
\end{algorithm}

\subsubsection*{Special cases}

If all nodes are function nodes, we recover computation graphs,
reviewed in \cref{neural_nets:sec:comput_graphs}.
If all nodes are distribution nodes, we recover 
Bayesian networks, reviewed in \cref{gm:sec:bayesian_net}.

\subsection{Examples}

We now present several examples that illustrate our formalism.
We use the legend below in the following illustrations.

\begin{center}
  \vspace*{1em}
  \includegraphics[width=0.7\linewidth]{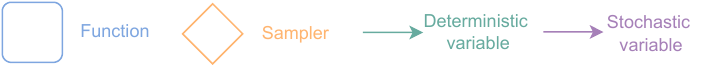}
\end{center}

\begin{itemize}
  \item Example 1 (SFE estimator):
  
  \begin{center}
    \includegraphics[width=0.45\linewidth]{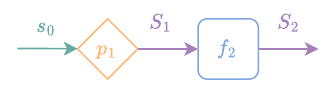}
    \vspace*{-3em}
  \end{center}

  \begin{align*}
      S_1 &\sim p_1(\cdot \mid \s_0) \\
      S_2 &\coloneqq f_2(S_1) \\
  E(\s_0) &\coloneqq \EE[S_2] \\
  \nabla E(\s_0) &= \EE_{S_1}[f_2(S_1) \nabla_{\s_0} \log p_1(S_1 \mid \s_0)]
  \end{align*}
  
  \item Example 2 (Pathwise estimator):
  
  \begin{center}
    \includegraphics[width=0.3\linewidth]{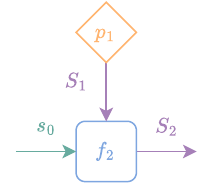}
    \vspace*{-3em}
  \end{center}

  \begin{align*}
      S_1 & \sim p_1 \\
      S_2 &\coloneqq f_2(S_1, \s_0) \\
      E(\s_0) &\coloneqq \EE[S_2] \\
      \nabla E(\s_0) &= \EE_{S_1}\left[ \nabla_{\s_0} f_2(S_1, \s_0) \right]
  \end{align*}
  
  \item Example 3 (SFE estimator + chain rule):
  
  \begin{center}
    \includegraphics[width=0.7\linewidth]{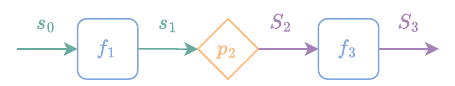}
    \vspace*{-3em}
  \end{center}

  \begin{align*}
      \s_1 &\coloneqq f_1(\s_0) \\
      S_2 & \sim p_2(\cdot \mid \s_1) \\
      S_3 &\coloneqq f_3(S_2) \\
      E(\s_0) &\coloneqq \EE[S_3] \\
      \nabla E(\s_0)
        &= \partial f(\s_0)^* \EE_{S_2}[f_3(S_2) \nabla_{\s_1} \log p_2(S_2
        \mid \s_1)]
  \end{align*}
  
  \item Example 4:
  
  \begin{center}
    \includegraphics[width=0.7\linewidth]{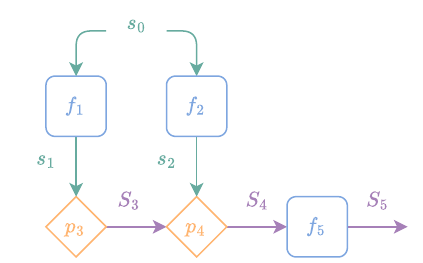}
    \vspace*{-3em}
  \end{center}

  \begin{align*}
      \s_1 &\coloneqq f_1(\s_0) \\
      \s_2 &\coloneqq f_2(\s_0) \\
      S_3 & \sim p_3(\cdot \mid \s_1) \\
      S_4 &\sim p_4(\cdot \mid \s_2, S_3) \\
      S_5 &\coloneqq f_5(S_4) \\
      E(\s_0) &\coloneqq \EE[S_5]
      = \EE_{S_3} \left[ \EE_{S_4}[f_5(S_4)]\right] \\ 
      \nabla E(\s_0) &=
      \EE_{S_3} \left[\partial f_1(\s_0)^*\nabla_{\s_1} \log p(S_3 \mid \s_1)
      \EE_{S_4}[f_5(S_4)]\right] \\
      &\quad  + \EE_{S_3}
      \left[\EE_{S_4}\left[\partial f_2(\s_0)^*\nabla_{\s_2} \log
      p_4(S_4|\s_2,S_3)f_5(S_4) \right]\right]
  \end{align*}

  \end{itemize}

As can be seen, the gradient expressions can quickly become quite complicated,
demonstrating the merits of automatic differentiation in stochastic computation
graphs.

\subsection{Unbiased gradient estimators}

The output of a stochastic program is a random variable
\begin{equation*}
S_K \coloneqq f(\s_0). 
\end{equation*}
It implicitly defines a probability distribution
$p(\cdot|\s_0)$ such that
$S_K \sim p(\cdot|\s_0)$.
Executing the stochastic program once gives us an \iid sample 
from $p(\cdot|\s_0)$.

Since derivatives are defined for deterministic variables,
we need a way to convert a random variable to a deterministic variable.
One way to do so is to consider the expected value (another way would be the
mode)
\begin{equation*}
E(\s_0) \coloneqq \EE[S_K] = \EE[f(\s_0)] \in \conv(\cS_K), 
\end{equation*}
where the expectation is over 
$S_K \sim p(\cdot|\s_0)$
or equivalently over the intermediate random variables $S_k$
\[
  S_k
  \sim p_k(\cdot|\s_{\dparent(k)}, S_{\rparent(k)}),
\]
for $k \in \cV_p$ (the distribution nodes).
We then wish to compute the gradient or more generally the Jacobian of
$E(\s_0)$.

If all nodes in the stochastic computation graph are function nodes, we can
estimate the gradient of $E(\s_0)$ using the pathwise estimator \aka
reparametrization trick (\cref{grad_est:sec:pge}). This is the approach taken
by \citet{kingma_2013} and \citet{rezende2014stochastic}.

If all nodes in the stochastic computation graph are distribution nodes, 
we can use the SFE estimator (\cref{grad_est:sec:sfe}). 
\citet{schulman2015gradient} propose a
surrogate loss so that using autodiff on that loss produces an unbiased gradient
of the expectation, using the SFE estimator. \citet{foerster2018dice} extend the
approach to support high-order differentiation. \citet{krieken2021storchastic}
further extend the approach by supporting different estimators per node,
as well as control variates.

\subsubsection*{Converting distribution nodes into function nodes and
vice-versa}

Our formalism uses two types of nodes: distribution nodes with associated
conditional distribution $p_k$ and function nodes with associated function
$f_k$. It is often possible to convert between node types.

Converting a distribution node into a function node is exactly the
reparametrization trick studied in \cref{grad_est:sec:pge}.
We can use transformations such as the location-scale transform or the inverse
transform.

Converting a function node into a distribution node can be done using the
change-of-variables theorem, studied in
\cref{grad_est:sec:change_of_var_theorem},
on a pushforward distribution.

\begin{figure}
  \centering
  \includegraphics[width=\linewidth]{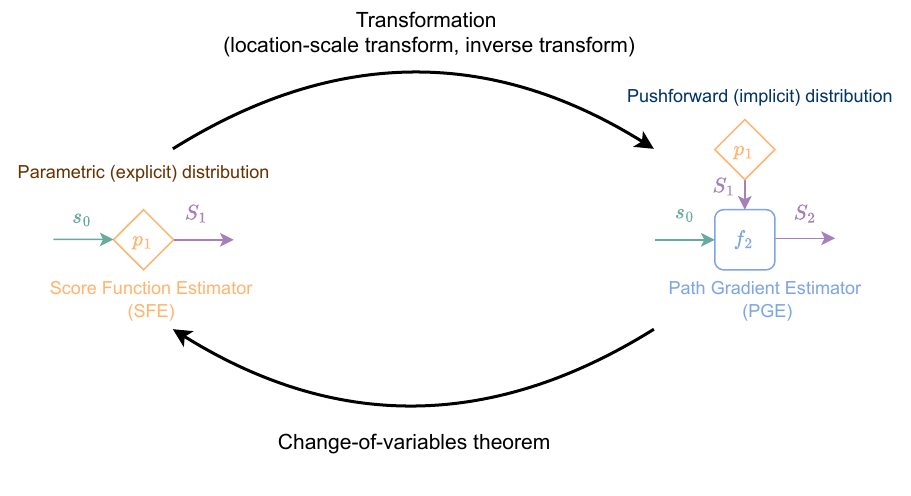}
  \caption{
    It is sometimes possible to convert a distribution node to a function node
    and vice-versa using a suitable transformation. \label{gm:fig:dist_vs_fun}
  }
\end{figure}

Because the pathwise estimator has lower variance than SFE, this is the method
of choice when the $f_k$ functions are available. The conversion from
distribution node to function node and vice-versa is illustrated in
\cref{gm:fig:dist_vs_fun}.

\subsection{Local vs. global expectations}

A stochastic computation graph can be seen as a \textbf{stochastic process}, a
collection of random variables $S_k$, indexed by $k$, the position
in the topological order. However,
random variables are incompatible with autodiff. Replacing random variables by
their expectation can be seen as a way to make them compatible with autodiff.
Two strategies are then possible.

As we saw in the previous section,
a strategy is to consider the expectation of the last output $S_K$.
This strategy corresponds to a \textbf{global smoothing}.
The two major advantages are that i) we do not need to assume that $f_{k+1}$ is
well-defined on $\conv(\cS_k)$ and ii) this induces a probability distribution
over program executions. This is
for instance useful to compute the variance of the program.
The gradient of the program's expected value can be estimated by the
reparametrization trick or by the SFE, depending on the type of nodes used.

A second strategy is to replace an intermediate random variable $S_k \in
\cS_k$, for $k \in \{1, \dots, K\}$, by its expectation $\EE[S_k] \in
\conv(\cS_k)$.  This
strategy corresponds to a \textbf{local smoothing}.  A potential drawback of
this approach is that $\EE[S_k]$ belongs to $\conv(\cS_k)$, the convex hull of
$\cS_k$. Therefore, the function $f_{k+1}$ in which $\EE[S_k]$ is fed must be
well-defined on $\conv(\cS_k)$, which may not always be the case.
In the case of control flows, another disadvantage is computational.
We saw in \cref{cf:sec:if_else} and \cref{cf:sec:else_if} that
using a soft comparison operator within a conditional statement
induces a distribution on a binary or
categorical random variable, corresponding to the branch to be selected.
A conditional statement can then be locally smoothed out by replacing the random
variable by its expectation i.e., a \textbf{convex combination} of all the
branches.  This means that, unless the distribution has sparse support, all
branches must be evaluated.

%% file: chapters/diff_thru_int/ode.tex
\section{Differential equations}\label{chap:ode}

\subsection{Parameterized differential equations}

\subsubsection*{From residual networks to neural ODEs}

Starting from $\s_0 \coloneqq \x$,
residual networks, reviewed in \cref{neural_nets:sec:residual},
iterate for $k \in \{1, \dots,K\}$
\[
  \s_k \coloneqq \s_{k-1} + h_k(\s_{k-1}, \w_k).
\]
A residual network can be seen as parameterizing incremental discrete-time
input changes (hence the name ``residual'')
\begin{equation*}
\s_k - \s_{k-1} = h_k(\s_{k-1}, \w_k).
\end{equation*}
\citet{chen2018neural} proposed to parameterize continuous-time (instantaneous)
changes instead. They considered the evolution $\s(t)$ of the inputs in continuous
time driven by a function $h(t, \s, \w)$ parameterized by $\w$, starting
from $\x$. Formally, the evolution $\s(t)$ is the solution of the
\textbf{ordinary differential equation} (ODE)
\begin{align}
  \s(0) &= \x \nonumber \\
  \s'(t) &= h(t, \s(t), \w) \quad t \in [0, T] \label{ode:eq:ode}
\end{align}
Here, $\s'(t)$ is the vector of derivatives of $\s$ as defined
in~\cref{diff:remark:jac_scalar_case}, and $T$ denotes a final time for the
trajectory. The output of such a \textbf{neural ODE}~\citep{chen2018neural} is 
then $f(\x, \w) \coloneqq \s(T)$.
Alternatively, the output can be seen as the solution of an \textbf{integration}
problem
\begin{equation}\label{ode:eq:ode_integral_form}
  f(\x, \w) = \s(T) = \x + \int_0^T h(t, \s(t), \w)dt.
\end{equation}
Differential equations like~\cref{ode:eq:ode} arise in many contexts beyond
neural ODEs, ranging from modeling physical systems to
pandemics~\citep{braun1983differential}. Moreover, the differential equation
presented in~\cref{ode:eq:ode} is just an example of an ordinary differential
equation, while controlled differential equations or stochastic differential
equations can also be considered.

\subsubsection*{Existence of a solution}

First and foremost, the question is whether $\s(t)$ is well-defined. Fortunately,
the answer is positive under mild conditions, as shown by Picard-Lindel\"of's
theorem recalled below~\citep[Theorem 16]{butcher2016numerical}.
\begin{boxthm}{Existence and uniqueness of ODE solutions}
  If $h: [0, T] \times \cS \rightarrow \cS$ is continuous in its first
  variable and Lipschitz-continuous in its second variable, then there exists a
  unique differentiable map $\s: [0, T] \rightarrow \cS$ satisfying
  \begin{align*}
      \s(0) &= \s_0 \\ 
      \s'(t) &= h(t, \s(t)) \quad t \in [0, T],
  \end{align*}
  for some given $\s_0 \in \cS$.
\end{boxthm}
For time-independent linear functions $h(t, \s) =\A\s$, 
the integral in~\cref{ode:eq:ode_integral_form} can be computed in closed form
as
\[
\s_t = \exp(t\A)(\s_0),
\]
where $\exp(\A)$ is the matrix exponential. Hence, the output $\s(T)$ can be
expressed as a simple function of the parameters ($\A$ in this case). 
However, generally,
we do not have access to such analytical solutions, and, just as for solving 
optimization problems in~\cref{chap:imp_diff}, we need to resort to some
iterative algorithms. 

\subsubsection*{Integration methods}

To numerically solve an ODE, 
we can use \textbf{integration methods}, whose goal is to build a
sequence $\s_k$ that approximates the solution $\s(t)$ at times $t_k$. The
simplest integration method is the \textbf{explicit Euler method}, which
approximates the solutions between times $t_{k-1}$ and $t_k$ as
\begin{align*}
  \s(t_{k-1}) - \s(t_k) 
  & = \int_{t_{k-1}}^{t_k} h(t, \s(t), \w)dt \\
  & \approx \delta_k h(t_{k-1}, \s(t_{k-1}), \w),
\end{align*}
for a time-step 
\[
  \delta_k \coloneqq t_k - t_{k-1}.
\]
The resulting integration scheme consists in computing
starting from $\s_0 = \x$,
for $k \in \{1, \ldots, K\}$,
\begin{equation*}
\s_k \coloneqq \s_{k-1} + \delta_k h(t_{k-1}, \s_{k-1}, \w).
\end{equation*}
Assimilating $\delta_k h(t_{k-1}, \s_{k-1}, \w)$ with $h_k(\s_{k-1}, \w_k)$, we
find that residual networks are essentially the
discretization of a neural ODE by an explicit Euler method; more precisely, a
non-autonomous neural ODE, see e.g.~\citep{davis2020time}.

Euler's forward method is only one integration method among many. To cite a
few, there are implicit Euler methods, semi-implicit methods,
Runge-Kutta methods, linear multistep methods, etc. See,
e.g.,~\citet{gautschi2011numerical} for a detailed review. The quality of
an integration method is measured by its consistency and its
stability~\citep{gautschi2011numerical}. These concepts naturally influence the
development of evaluation and differentiation techniques for ODEs. We briefly
summarize them below. 

Given a fixed time interval $\delta_k = \delta$ and $K = \lceil
T/\delta\rceil$ points, an integration method is \textbf{consistent of order
$k$} if $\|\s_k - \s(k \delta)\| = O(\delta^k)$ as $\delta\rightarrow 0$ and
therefore $k\rightarrow +\infty$. The higher the order $k$, the fewer points we
need to reach an approximation error $\varepsilon$ on the points considered. The
term $\|\s_k - \s(k \delta)\| = O(\delta^k)$ is reminiscent of the error
encountered in finite differences (\cref{chap:finite_diff}) and is called
the \textbf{truncation error}.

The (absolute) \textbf{stability} of a method is defined by the set of
time-steps such that the integration method can integrate $s'(t) =
\lambda s(t)$ for some $\lambda \in \CC$ without blowing up as $t\rightarrow
+\infty$.

\subsection{Continuous adjoint method}

Since different parameters $\w$ induce different trajectories associated with
$h(t, \s, \w)$ in~\cref{ode:eq:ode}, we may want to select one of these
trajectories by minimizing some criterion. For example, we may consider
selecting $\w \in \cW$ by minimizing a loss $L$ on the final point of the
trajectory,
\begin{equation}\label{ode:eq:optim_through_ode}
  \min_{\w\in \cW} L(f(\x, \w), \y), 
\end{equation}
where
\begin{equation*}
f(\x, \w) \coloneqq \s(T) = \x + \int_0^T h(t, \s(t), \w)dt.
\end{equation*}
To solve such problems, we need to access gradients of $L$ composed
with $f$ through VJPs of the solution of the ODE. The VJPs can actually be
characterized as solutions of an ODE themselves thanks to the \textbf{continuous
time adjoint method}~\citep{pontryagin1985mathematical}, presented below, and
whose proof is postponed to \cref{ode:sec:proofs}.
\begin{boxprop}{Continuous-time adjoint method}
    \label{ode:thm:adjoint_method}
  Consider a function $h: [0, T] \times \cS \times \cW \rightarrow \cS$,
  continuous in its first variable, Lipschitz-continuous and continuously
  differentiable in its second variable. Assume that $\partial_3 h(t, \s, \w)$
  exists for any $t, \s, \w$, and is also continuous in its first variable,
  Lipschitz-continuous in its second variable.
  Denote $\s:[0,T] \rightarrow \cS$ the solution of the ODE
  \begin{align*}
      \s(0) &= \x \\
      \s'(t) &= h(t, \s(t), \w) \quad  t \in [0, T],
  \end{align*}
  and $f(\x, \w) = \s(T)$ the final state of the ODE at time $T$.
  
  Then, the function $f$ is differentiable, and for an output direction $\u \in
  \cS$, its VJP along $\u$ is given by 
  \[
    \partial f(\x, \w)^*\u = (\r(0), \g)
  \] 
  for
  \begin{equation*}
    \g = \int_{0}^{T} \partial_3 h(t, \s(t), \w)^* \r(t) dt
  \end{equation*}
  and for $\r$ solving the \textbf{adjoint} (backward) ODE
  \begin{align*}
    \r'(t) & = - \partial_2 h(t, \s(t), \w)^* \r(t) \\
    \r(T) &= \u \nonumber.
  \end{align*}
  In particular, the gradient 
    $\nabla (L \circ f)(\x, \w)$
  for $L: \cS \rightarrow \RR$ a differentiable loss is obtained 
  by solving the adjoint ODE with $\r(T) = \nabla L(\s(T))$.
\end{boxprop}

\begin{figure}[t]
  \centering
  \includegraphics[width=0.6\linewidth]{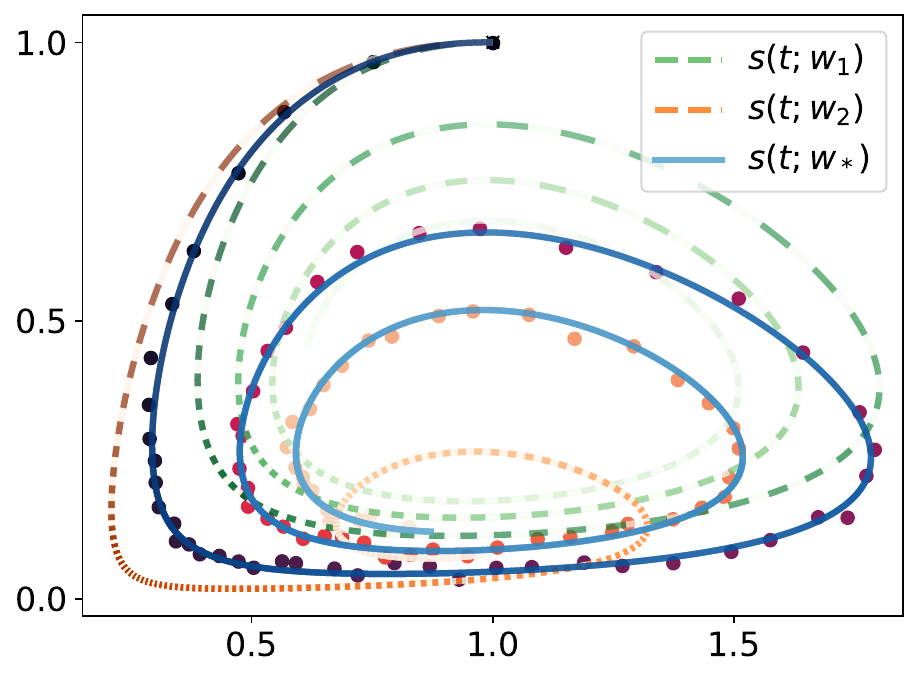}
  \caption{Finding the optimal parameters of an ODE to fit some observed data.
  The dots represent the trajectories of a dynamical system observed at regular
  times (time is represented here by a gradient color, the lighter the color,
  the larger the time). Each line represents the solution of an ODE given by
  some hyperparameters $\w$. The objective is to find the hyperparameters of the
  ODE such that its solution fits the data points. Green and orange lines fail
  to do so while the blue line fits the data. To compute such parameters $\w$,
  we need to backpropagate through the solution of the ODE.
  \label{ode:fig:regression_ode}
  }
  
\end{figure}

\begin{boxexm}{Fitting data through the solution of an ODE}
  As an illustrative example, we can consider optimizing the parameters of an
  ODE to fit some data points. Namely, we may seek a continuous time
  solution $\z(t; \w)$ of a modified Lotka Volterra ODE
  \[
    \z'(t; \w) = \begin{pmatrix}
      \alpha z_1(t;\w) - \beta z_1(t;\w) z_2(t;\w) \\
      -\gamma z_2(t;\w) + \delta z_1(t;\w) z_2(t;\w)
    \end{pmatrix} + \c,
  \]
  for $\w = (\alpha, \beta, \gamma, \delta, \c)$, that fits some observations
  $\z_1, \ldots, \z_T$. The optimization problem then consists of
  \[
  \min_{\w}\sum_{j=1}^{T} \|\z(t_j;\w) - \z_j\|^2,
  \]
  and requires backpropagating through the solution $\z(\cdot;\w)$ of the ODE
  \wrt its candidate parameters $\w$. \cref{ode:fig:regression_ode}
  illustrates such a problem with varying candidate parameters.
\end{boxexm}

\subsection{Gradients via the continuous adjoint method}

\cref{ode:thm:adjoint_method} gives a formal definition of the gradient.
However, just as computing the mapping $f(\x, \w)$ itself, computing its VJP or the
gradient of $L\circ f$ requires solving an integration problem. Note that the
integration of $\r(t)$ in~\cref{ode:thm:adjoint_method} requires also values of
$\s(t)$. Therefore, we need to integrate both $\r(t)$ and $\s(t)$. Such an approach is
generally referred to as \textbf{optimize-then-discretize} because we first
formulate the gradient in continuous time (the ``optimize part'') and then
discretize the resulting ODE.

\subsubsection*{Simple discretization scheme}

A first approach consists in defining a backward discretization scheme that can
approximate $\s(t)$ backward in time.  Namely, by defining $\sigmav(t)= \s(T-t),
\rhov(t) = \r(T-t)$, and $\gammav(t) = \int_t^T \partial_3 h(\tau, \s(\tau),
\w)^*\r(\tau) d\tau$, the derivative of $L\circ f$
is given by $(\rhov(T), \gammav(T))$. The
functions $\sigmav, \rhov, \gammav$ are solutions of a standard ODE
\begin{align*}
  \sigmav(0) & = \s(T),
  \hspace*{30pt} \sigmav'(t) = -h(T-t, \sigmav(t), \w), \\
  \rhov(0) & = \nabla L(\s(T)),
  \hspace*{6pt} \rhov'(t) = \partial_2 h(T-t, \sigmav(t), \w)^* \rhov(t), \\
  \gammav(0) & = 0,
  \hspace*{48pt} \gammav'(t) = \partial_3 h(T-t, \sigmav(t), \w)^*\rhov(t).
\end{align*}
The above ODE can then be solved by any integration method. Note, however, that
it requires first computing $\s(T)$ and $\nabla L(\s(T))$ by an integration
method. The overall computation of the gradient using an explicit Euler method
to solve forward and backward ODEs is summarized in
\cref{ode:algo:optimize_then_discretize}.

\begin{algorithm}\caption{Gradient computation via continuous adjoint method
  with Euler explicit discretization\label{ode:algo:optimize_then_discretize}}
  \begin{algorithmic}[1]
    \State{\bf Functions:} $h: [0, T] \times \cS \times \cW \rightarrow \RR$,
    $L: \cS \rightarrow \RR$
    \State{\bf Inputs:} input $\x$, parameters $\w$,
    number of discretization steps $K$. 
    \State Set discretization step $\delta = T/K$, denote $h_k(\s, \w) = h(k\delta, \s, \w)$.
    \State Set $\s_0 \coloneqq \x$
    \For{$k \coloneqq 1, \ldots, K$}  \Comment{Forward discretization}
    \State Compute $\s_k \coloneqq \s_{k-1} + \delta h_{k-1}(\s_{k-1}, \w)$.
    \EndFor
    \State Compute $\u \coloneqq \nabla L(\s_K)$.
    \State Initialize $\r_K \coloneqq \u$, $\hat \s_K = \s_K$, $\g_K= \zeros$
    \For{$k \coloneqq K, \ldots, 1$}  \Comment{Backward discretization}
    \State Compute $\hat \s_{k-1}  \coloneqq \hat \s_k - \delta h_k(\hat \s_k, \w)$
    \State Compute $\r_{k-1} \coloneqq \r_k + \delta \partial_2 h_k(\hat \s_k, \w)^* \r_k$
    \State Compute $\g_{k-1} \coloneqq \g_k + \delta \partial_3 h_k(\hat \s_k, \w)^*\r_k$
    \EndFor
    \State{\bf Output:} $(\r_0, \g_0) \approx \nabla (L\circ f)(\x, \w)$
  \end{algorithmic}
\end{algorithm}

\cref{ode:algo:optimize_then_discretize} naturally looks like the reverse mode
of autodiff for a residual neural network with \textbf{shared weights}. A
striking difference is that the intermediate computations $\s_k$ are not kept in
memory and, instead, new variables $\hat \s_k$ are computed along the backward
ODE. One may believe that by switching to continuous time,
we solved the memory issues encountered in reverse-mode autodiff. 
Unfortunately, this comes at the cost of numerical stability.
As we use a discretization scheme to
recompute the intermediate states backward in time through $\hat \s_k$ in
\cref{ode:algo:optimize_then_discretize}, we accumulate some truncation errors.

\begin{figure}
  \includegraphics[width=\linewidth]{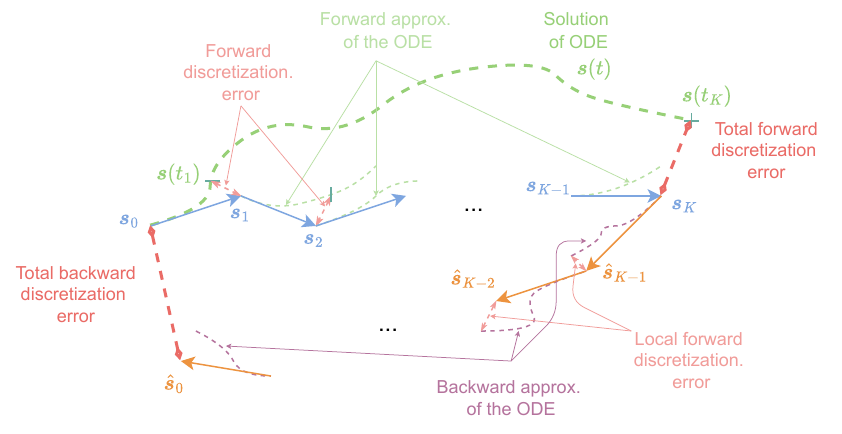}
  \caption{Forward and backward discretizations when using the continuous adjoint method.}
  \label{ode:fig:diff_ode}
\end{figure}

To understand the issue here, consider
applying~\cref{ode:algo:optimize_then_discretize} repeatedly on the same
parameters but using $\hat \s_0$ instead of $\s_0=\x$ each time. In the
continuous realm, $\sigmav(T) = \s(0)$. But after discretization, $\hat \s_0
\approx \sigmav(T)$ does not match $\s_0$. Therefore, by applying
\cref{ode:algo:optimize_then_discretize} with $\s_0= \hat \s_0$, we would not get
the same output even if in continuous time we naturally should have. 
This phenomenon is illustrated in \cref{ode:fig:diff_ode}. 
It intuitively shows why \cref{ode:algo:optimize_then_discretize}
induces some noise in the estimation of the gradient.

\subsubsection*{Multiple shooting scheme}

An alternative approach consists in integrating both the forward and backward
ODEs jointly. Namely, we may solve an ODE with boundary values
\begin{align}
  \s'(t) & = h(t, \s(t), \w), 
  \quad \s(0) = \x, \nonumber\\
  \r'(t) & = - \partial_2 h(t, \s(t), \w)^* \r(t), 
  \quad \r(T) = \nabla L(\s(T))\nonumber \\
  \g'(t) & = - \partial_3 h(t, \s(t), \w)^*\r(t), 
  \quad \g(T) = 0, \nonumber
\end{align}
by means of a multiple shooting method or a collocation
method~\citep{stoer1980introduction}. This approach still requires
$\nabla L(\s(T))$ to be approximated first.

\subsection{Gradients via reverse-mode on discretization}

A simpler approach consists in replacing the objective
in~\cref{ode:eq:optim_through_ode} by its version discretized using some
numerical method, such as an Euler forward discretization scheme.
That is, we seek to solve
\begin{equation*}
\min_{\w \in \cW} \quad L(\s_K)
\quad \text{where} \quad
\s_k = \s_{k-1} + \delta h(k\delta, \s_{k-1}, \w) ~
k \in \{1, \ldots, K\},
\end{equation*}
with $\s_0 = \zeros$ and $\delta$ some discretization step. 
Gradients of the objective
can be computed by automatic
differentiation. That approach is often referred to as
\textbf{discretize-then-optimize}. At first glance, this approach may suffer
from very high memory requirements. Indeed, to get an accurate solution of the
ODE, a numerical integration method may require $K$ to be very large. Since a
naive implementation of reverse-mode automatic differentiation has a memory that
scales linearly with $K$, computing the gradient by a discretize-then-optimize
method could be prohibitive. However, the memory requirements may easily be
amortized using checkpointing, as explained
in~\cref{auto_diff:sec:checkpointing}; see also \citep{gholaminejad2019anode}.

As for the \textbf{optimize-then-discretize} method, we still accumulate some
truncation errors in the forward discretization process. This discretization
error occurs when computing the gradient in reverse-mode too. 
The discretize-then-optimize method
can be seen as computing gradients of a surrogate objective. For that objective,
the gradients are correct and well-defined. However, they may not match the
gradients of the true ODE formulation. 

To compare the discretize-then-optimize and optimize-then-discretize approaches,
\citet{gholaminejad2019anode} compared their performance on
an ODE whose solution can be computed analytically by selecting $h$ to be
linear in $\s$. The authors observed that
discretize-then-optimize generally outperformed optimize-then-discretize. A
middle ground can actually be found by using reversible differentiation
schemes.

\subsection{Reversible discretization schemes}

Our exposition of the optimize-then-discretize or discretize-then-optimize
approaches used a simple Euler explicit discretization scheme. However, for both
approaches, we could have used other discretization schemes instead, such as
reversible discretization schemes.

A reversible discretization scheme is a discretization scheme such that we have
access to a closed-form formula for the inverse of its discretization step. Formally,
a discretization method $\cM$ builds an approximation $(\s_k)_{k=1}^{K}$ of the
solution of an ODE $\s'(t) = h(t, \s(t))$ on an interval $[0, T]$ by computing
for $k \in (1,\dots, K)$
\begin{align}\label{ode:eq:forward_reversible}
  t_k, \s_k, \c_k = \cM(t_{k-1}, \s_{k-1}, \c_{k-1}; h, \delta),
\end{align}
where $\delta>0$ is some fixed discretization step, $t_k$ is the time step
(typically $t_k = t_{k-1} + \delta$), $\s_k$ is the approximation of $\s(t_k)$,
and $\c_k$ is some additional context variable used by the discretization
method to build the iterates. An explicit Euler method does not have a context,
but just as an optimization method may update some internal states, a
discretization method can update some context variable. The discretization
scheme in \cref{ode:eq:forward_reversible} is a forward discretization scheme as
we took a positive discretization step. By taking a negative discretization
step, we obtain the corresponding backward discretization scheme, 
for $k \in (K, \dots, 1)$,
\begin{align*}
t_{k-1}, \s_{k-1}, \c_{k-1} = \cM(t_k, \s_k, \c_k; h, -\delta).
\end{align*}
A discretization method is \textbf{reversible} if we have access to $\cM^{-1}$
to recompute the inputs of the discretization step from its outputs, 
\[
  t_{k-1}, \s_{k-1}, \c_k = \cM^{-1}(t_k, \s_k, \c_k; h, \delta).
\]
A reversible discretization method is \textbf{symmetric} if the
backward discretization scheme is exactly the inverse of the forward
discretization scheme, i.e.,
\begin{align*}
  \cM(t_k, \s_k, \c_k; h, -\delta) = \cM^{-1}(t_k,
  \s_k, \c_k; h, \delta).
\end{align*}
The explicit Euler method is clearly not symmetric and a priori not reversible,
unless we can solve for $\y_{k-1}$, the equation $\y_k = \y_{k-1} + \delta
f(\y_{k-1})$. 

\subsubsection*{Leapfrog method}
The (asynchronous) \textbf{leapfrog method} \citep{zhuang2021mali,
mutze2013asynchronous} on the other hand is an example of symmetric reversible
discretization method. For a constant discretization step $\delta$, given
$t_{k-1}, \s_{k-1}, \c_{k-1}$ and a function $h$, it computes
\begin{align*}
  \bar t_{k-1} & \coloneqq t_{k-1} + \frac{\delta}{2} \\
  \bar \s_{k-1} & \coloneqq \s_{k-1} + \frac{\delta}{2} \c_{k-1} \\
  \bar \c_{k-1} & \coloneqq h(\bar t_{k-1}, \bar \s_{k-1}) \\
  t_k & \coloneqq \bar t_{k-1} + \frac{\delta}{2} \\
  \s_k & \coloneqq \bar \s_{k-1} + \frac{\delta}{2} \bar \c_{k-1} \\
  \c_k & \coloneqq  2 \bar \c_{k-1} - \c_{k-1} \\
  \mathcal{M}(t_{k-1}, \s_{k-1}, \c_{k-1}; h, \delta) & \coloneqq (t_k, \s_k, \c_k). \hspace*{60pt}
\end{align*}
One can verify that we indeed have $\mathcal{M}(t_k, \s_k, \c_k; h, -\delta) =
(t_{k-1}, \s_k, \c_k)$.

By using a reversible symmetric discretization scheme in the
optimize-then-discretize approach, we ensure that, at the end of the backward
discretization pass, we recover exactly the original input. Therefore, by
repeating forward and backward discretization schemes we always get the same
gradient, which was not the case for an Euler explicit scheme. 

By using a reversible discretization scheme in the discretize-then-optimize
method, we address the memory issues of reverse mode autodiff. As
explained in \cref{auto_diff:sec:reversible}, we can recompute intermediate
values during the backward pass rather than storing them. 

\subsubsection*{Momentum residual networks}
In the leapfrog method, the additional variables $\c_k$ may actually be
interpreted as velocities of a system whose acceleration is driven by the
given function, that is, $\s''(t) = h(t, \s(t), \w)$. Such an interpretation
suggests alternatives to the usual neural ODE paradigm. 
For instance, \textbf{momentum neural networks} \citep{sander2021momentum}, 
can be interpreted as the
discretization of a \textbf{second-order ordinary differential equation}, which
is naturally amenable to reversible differentiation schemes with a low memory
footprint.

\subsection{Proof of the continuous adjoint method}\label{ode:sec:proofs}

  In the following, we denote $\s(t, \x, \w)$ the solution of the ODE at time
  $t$ given the input $\x$ and the parameters $\w$. We focus here on the
  formulation of the VJP. The proof relies on the existence of partial
  derivatives of $\s(t, \x, \w)$, which we do not cover here and refer to, e.g.,
  \citet{pontryagin1985mathematical} for a complete proof of such facts given
  the assumptions.
  
  We use the ODE constraint to introduce adjoint variables, this time in the
  form of a continuously differentiable function $\r$. For any such function
  $\r$, we have 
  \begin{align*}
    \langle f(\x, \w), \u \rangle 
    & = \langle \s(T, \x, \w), \u \rangle  \\
    & \quad + 
      \int_0^T 
        \langle 
          \r(t),
          h(t, \s(t, \x, \w), \w) - \partial_t \s(t, \x, \w)
        \rangle
      dt,
  \end{align*}
  using Leibniz notations such as $\partial_t \s(t, \x, \w) = \partial_1 \s(t,
  \x, \w)$. The VJPs then decompose as
  \begin{align*}
    & \partial_\w f(\x, \w)^*[\u] \\
    & = \partial_\w \s(T, \x, \w)^* \u \\
    & \quad + \int_0^T 
    (\partial_\w \s(t, \x, \w)^* \partial_\s h(t, \s(t, \x, \w), \w)^*
    - \partial_{\w t}^2 \s(t, \x, \w)^*
    )\r(t) dt \\
    & \quad + \int_0^T \partial_\w h(t, \s(t, \x, \w), \w)^* \r(t) dt,  \\
    & \partial_\x f(\x, \w)^*[\u] \\
    & = \partial_\x \s(T, \x, \w)^* \u \\
    & \quad + \int_0^T 
    (\partial_\x \s(t, \x, \w)^* \partial_\s h(t, \s(t, \x, \w), \w)^*
    - \partial_{\x t}^2 \s(t, \x, \w)^*
    )\r(t) dt
  \end{align*}
  Here the second derivative terms $\partial_{\w t}^2 \s(t, \x, \w)^*\r,
  \partial_{\x t}^2 \s(t, \x, \w)^*\r$ correspond to second derivatives of
  $\langle \s(t, \x, \w), \r \rangle$. Since the Hessian is symmetric
  (Schwarz's theorem presented in~\cref{diff:thm:symmetry_hessian}), we can
  swap the derivatives in $t$ and $\w$ or $\x$. Then, to express the gradient
  uniquely in terms of first derivatives of $\s$, we use an integration by parts
  to have for example
  \begin{align*}
    \int_0^T \partial_{\w t}^2 \s(t, \x, \w)^* \r(t)dt 
    & = \int_0^T \partial_{t \w}^2 \s(t, \x, \w)^* \r(t)dt \\
    & = (
      \partial_{\w} \s(T, \x, \w)^* \r(T) 
      - \partial_{\w} \s(0, \x, \w)^* \r(0)
      ) \\
    & \quad - \int_0^T \partial_{\w} \s(t, \x, \w)^* \partial_t \r(t)dt.
  \end{align*}
  Since $\s(0) = \x$, we have $\partial_{\w} \s(0, \x, \w)^* \r(0) = 0$. The VJP
  \wrt $\w$ can then be written as
  \begin{align*}
    & \partial_\w f(\x, \w)^*[\u]  \\
    & = \partial_\w \s(T, \x, \w)^* [\u- \r(T)] \\
    & \quad + \int_0^T 
      \partial_\w \s(t, \x, \w)^*[
        \partial_\s h(t, \s(t, \x, \w), \w)^* \r(t)
        + \partial_t \r(t)
        ]dt \\
    & \quad + \int_0^T \partial_\w h(t, \s(t, \x, \w), \w)^* \r(t) dt.
  \end{align*}
  By choosing $\r(t)$ to satisfy the adjoint ODE
  \[
    \partial_t \r(t) = -\partial_\s h(t, \s(t, \x, \w), \w)^* \r(t), 
    \quad \r(T) = \u,
  \]
  the expression of the VJP simplifies as
  \[
    \partial_\w f(\x, \w)^*[\u] = \int_0^T \partial_\w h(t, \s(t, \x, \w), \w)^* \r(t) dt.
  \]
  For the VJP \wrt $\x$, we can proceed similarly. Using an integration by
  parts, we have, this time, $\partial_{\x} \s(0, \x, \w)^* \r(0) = \r(0)$ since
  $\s(0) = \x$. Choosing the same curve $\r(t)$ satisfying the adjoint ODE we
  get 
  \[
    \partial_\x f(\x, \w)^*[\u] = \r(0).
  \]
  The existence of a curve $\r$ solution of the backward ODE can easily be shown
  from Picard Lindel\"of's theorem and the assumptions.

\section{Summary}

\begin{itemize}

\item We studied how to differentiate integrals,
with a focus on expectations and solutions of a differential equation.

\item For differentiating through expectations, we studied two main methods: the score
function estimator (SFE, \aka REINFORCE) and the path gradient estimator (PGE,
\aka reparametrization trick).  

\item The SFE is suitable when it is easy to sample
from the distribution and its log-PDF is \textbf{explicitly} available. It is an
unbiased estimator, but is known to suffer from high variance. 

\item The PGE is
suitable for pushforward distributions, distributions that are
\textbf{implicitly} defined through a transformation, or a sequence of them.
These distributions can be easily sampled from, by injecting a source of
randomness (such as noise) through the transformations. An unbiased,
low-variance estimator of the gradient of their expectation is easily obtained,
provided that we can interchange integration and differentiation.

\item If we have an explicit distribution, we can sometimes convert it to an implicit
distribution, thanks to the \textbf{location-scale transformation} or the
\textbf{inverse transformation}. 

\item Conversely, if we have an implicit
distribution, we can convert it to an explicit distribution using the
\textbf{change-of-variables theorem}.  However, this formula requires computing
the determinant of an inverse Jacobian, and is computationally expensive in
general.  Normalizing flows use invertible transformations so that the inverse
Jacobian is cheap to compute, by design.  

\item \textbf{Stochastic computation graphs} can use a mix of explicit and
implicit distributions at each node.

\item For differentiating through the solution of a differential equation, two approaches
can be considered. 

\item We can express the gradient as the solution of a differential
equation thanks to the \textbf{continuous adjoint method}. We may then
discretize backwards in time the differential equation that the gradient
satisfies. This is the \textbf{optimize-then-discretize} approach. 

\item We can also
first discretize the problem in such a way that the gradient can simply be computed by
reverse mode auto-diff, applied on the discretization steps. This is the
\textbf{discretize-then-optimize} approach. The optimize-then-discretize
approach has no memory cost, but discrepancies between the forward and backward
discretization passes often lead to numerical errors. The
discretize-then-optimize approach introduces no such discrepancies but may come
at a large memory cost. 

\item \textbf{Reversible discretization schemes} can circumvent the
memory cost, as they enable the recomputation of intermediate discretization
steps backwards in time.

\end{itemize}

%% file: chapters/smoothing/smoothing.tex
\chapter{Smoothing by optimization}
\label{chap:smoothing}

When a function is non-differentiable (or worse, discontinuous), a reasonable
approach is to replace it by a differentiable approximation (or at least, by a
continuous relaxation). We refer to the process of transforming a
non-differentiable function into a differentiable one as ``smoothing'' the
original function. In this chapter, we begin by reviewing a smoothing technique
based on \textbf{infimal convolution}. We then review an equivalent dual
approach, based on the \textbf{Legendre-Fenchel transform}. We illustrate how to
apply these techniques to compute smoothed ReLUs and smoothed max operators, as
well as continuous relaxations of step functions and argmax operators.

\section{Primal approach}
\label{smoothing:sec:primal}

We first review how to smooth functions in the original, primal space of the
function, using the infimal convolution and more particularly the Moreau
envelope, \aka Moreau-Yoshida regularization. In this chapter, we consider
functions taking potentially infinite positive values, that is, functions taking
values in the half-extended real line $\R\cup \{\infty\}$. For a function
$f:\RR^M \rightarrow \RR \cup \{\infty\}$, we define its domain as 
\[
\mathrm{dom}(f) = \{\u\in \RR^M: f(\u) < \infty\}.
\]

\subsection{Infimal convolution}

Sometimes abbreviated inf-conv, the infimal convolution between two functions
$f$ and $g$ creates a new function $f \square g$. It is defined as follows.
\begin{boxdef}{Infimal convolution}
The infimal convolution between two functions $f \colon \RR^M \to \RR \cup
\{\infty\}$ and $g \colon \RR^M \to \RR \cup \{\infty\}$ is defined by
\begin{align*}
(f \square g)(\muv) 
&\coloneqq \inf_{\u \in \RR^M} f(\u) + g(\muv - \u) \\
&= \inf_{\z \in \RR^M} f(\muv + \z) + g(-\z) \\
&= \inf_{\u, \z \in \RR^M} f(\u) + g(-\z) \text{ s.t. } \u = \muv + \z.
\end{align*}
\end{boxdef}
It is easy to check that the three definitions are indeed equivalent,
by using the change of variable $\u \coloneqq \muv + \z$,
which is a location-scale transform; 
see \cref{grad_est:sec:location_scale_transform}.

The infimal convolution can be seen as a counterpart of the usual convolution,
in which integration has been replaced by minimization (hence its name).
Similarly to the classical convolution, 
it is \textbf{commutative},
meaning that for all $\muv \in \RR^M$, we have
\begin{equation*}
    (f \square g)(\muv) = (g \square f)(\muv).
\end{equation*}

Computing the infimal convolution involves the resolution of a minimization
problem, that may or may not enjoy an analytical solution.
Some examples are given in \cref{smoothing:tab:inf_conv}.

\subsubsection*{Existence}

The infimal convolution $(f \square g)(\muv)$ exists if the infimum $\inf_{\u
\in \RR^M} f(\u) + g(\muv - \u)$ is finite \citep[Proposition
12.6]{bauschke2011convex}. A sufficient condition to achieve this is that $\u
\mapsto f(\u) + g(\muv - \u)$ is convex for all $\muv \in \RR^M$. However, this
is not a necessary condition. For example, the infimum can be finite even if $f$
or $g$ are nonconvex, for example if their domain is a compact set.

\subsubsection*{Infimal convolution with a regularization function}

When a function $f$ is non-differentiable, a commonly-used technique is to
replace it by its infimal convolution $f \square R$, with some regularization
$R$. The most used regularization is the squared $2$-norm, leading to the Moreau
envelope, as we now review.

\begin{table}[t]
\caption{Examples of infimal convolutions. 
We use $\indic_\cC$ to denote the indicator function of the set $\cC$.}
\centering
\begin{tabular}{lcc}
\toprule
$f(\u)$ & $g(\z)$ & $(f \square g)(\muv)$ \\
\midrule
$f(\u)$ & $0$ & $\inf_{\u \in \RR^M} f(\u)$ \\
$f(\u)$ & $\indic_{\{\v\}}(\z)$ & $f(\muv - \v)$ \\
$\indic_\cC(\u)$ & $\indic_\cD(\z)$ & $\indic_{\cC + \cD}(\muv)$ \\
$\indic_\cC(\u)$ & $\|\z\|_2$ & 
$d_\cC(\muv) = \inf_{\u \in \cC} \|\muv - \u\|_2$ \\
$f(\u)$ & $\frac{1}{2}\|\z\|_2^2$ &  \qquad
$\env_f(\muv) = \inf_{\u \in \RR^M} \frac{1}{2}\|\muv - \u\|_2^2 + f(\u)$ \\
\bottomrule
\end{tabular}
\label{smoothing:tab:inf_conv}
\end{table}

\subsection{Moreau envelope}
\label{smoothing:sec:moreau_env}

When $R(\z) \coloneqq \frac{1}{2} \|\z\|^2_2$, 
the infimal convolution $f \square R$ 
gives the so-called \textbf{Moreau envelope} of $f$,
which is also known as Moreau-Yoshida regularization of $f$.
\begin{boxdef}{Moreau envelope}
Given a function $f \colon \RR^M \to \RR \cup \{\infty\}$, its Moreau envelope
is defined as
\begin{align*}
\env_f(\muv) 
&\coloneqq \left(f \square \frac{1}{2} \|\cdot\|^2_2\right)(\muv) \\
&= \inf_{\u \in \RR^M}
f(\u) + \frac{1}{2} \|\muv - \u\|_2^2 \\
&= \inf_{\z \in \RR^M}
f(\muv + \z) + \frac{1}{2} \|\z\|_2^2.
\end{align*}
\label{smoothing:def:moreau_env}
\end{boxdef}
Intuitively, the Moreau envelope is the minimal value over $\u \in \RR^M$ of
a trade-off between staying close to the input $\muv$ according to the
\textbf{proximity term}
$\frac{1}{2}\|\muv-\u\|_2^2$ and minimizing $f(\u)$. Provided that the minimizer
exists and is unique, we can define the associated \textbf{proximal operator} of
$f$ as
\begin{equation*}
\mathrm{prox}_{f}(\muv) 
\coloneqq \argmin_{\u \in \RR^M}
\frac{1}{2} \|\muv - \u\|_2^2 + f(\u).
\end{equation*}
In other words, we have for $\mathrm{prox}_{f}(\muv)$
well defined,
\begin{equation}
\env_f(\muv) = 
f(\mathrm{prox}_f(\muv))
+
\frac{1}{2} \|\muv - \mathrm{prox}_f(\muv)\|_2^2.
\label{smoothing:eq:env_from_prox}
\end{equation}

\subsubsection*{Properties}

A crucial property of the Moreau envelope $\env_f$ is that for any convex
function $f$, it is always a smooth function, even when $f$ itself is not
smooth. By smooth, we formally mean that the resulting function $\env_f$ is
differentiable everywhere with Lipschitz-continuous gradients. We say
$L$-smooth, if the gradients are $L$-Lipschitz continuous. Such a property can
determine the efficiency of optimization algorithms as reviewed in
\cref{optim:sec:function_classes}. We recap below useful properties of the
Moreau envelope.
\begin{boxprop}{Properties of Moreau envelope}
\label{smoothing:prop:moreau_properties}
Let $f \colon \RR^M \to \RR \cup \{\infty\}$.
\begin{enumerate}

\item \textbf{Smoothness}: If $f$ is convex, the function $\env_f$ is $1$-smooth.

\item \textbf{Gradient}: Provided that $\mathrm{prox}_f(\muv)$ is well-defined
on $\muv \in \RR^M$, the gradient of the Moreau envelope can be expressed in
terms of the proximal operator as
\begin{equation*}
\nabla \env_f(\muv) = \muv - \mathrm{prox}_f(\muv).
\end{equation*}

\item \textbf{Moreau decomposition}: If $f$ is convex, then for any $\muv \in
\RR^M$, we have the following identity
\begin{equation*}
\mathrm{prox}_f(\muv) + \mathrm{prox}_{f^*}(\muv) = \muv,
\end{equation*}
where $f^*$ is the convex conjugate of $f$, detailed in
\cref{duality:sec:conjugates}. In particular, we get 
\[
\nabla \env_{f^*}(\muv) = \mathrm{prox}_f(\muv).
\]

\item \textbf{Convexity}: $\env_f$ is convex if $f$ is convex.

\item \textbf{Infimums coincide:} $\env_f$ has the same infimum as the original
function $f$:
\begin{equation*}
\min_{\muv \in \RR^M} \env_f(\muv) = \min_{\u \in \RR^M} f(\u).
\end{equation*}
\end{enumerate}
\end{boxprop}
\begin{proof}
\begin{enumerate}

\item This is best seen using the dual approach
detailed in \cref{smoothing:sec:dual_approach}.

\item This follows from Danskin's theorem, reviewed in 
\cref{implicit:sec:danskin}.

\item See, e.g., \citet[Theorem 14.3]{bauschke2011convex}.

\item This follows from the fact that the infimum of a jointly convex function
is convex.

\item We have
\begin{align*}
\inf_{\muv \in \RR^M} \env_f(\muv)
&= \inf_{\muv \in \RR^M} \inf_{\u \in \RR^M} \frac{1}{2} \|\muv - \u\|^2_2 +
f(\u) \\
&= \inf_{\u \in \RR^M} \inf_{\muv \in \RR^M} \frac{1}{2} \|\muv - \u\|^2_2 +
f(\u) \\
&= \inf_{\u \in \RR^M} f(\u).
\end{align*}
\end{enumerate}
\end{proof}

\subsubsection*{Examples}

\begin{figure}[t]
\centering
\includegraphics[scale=0.4]{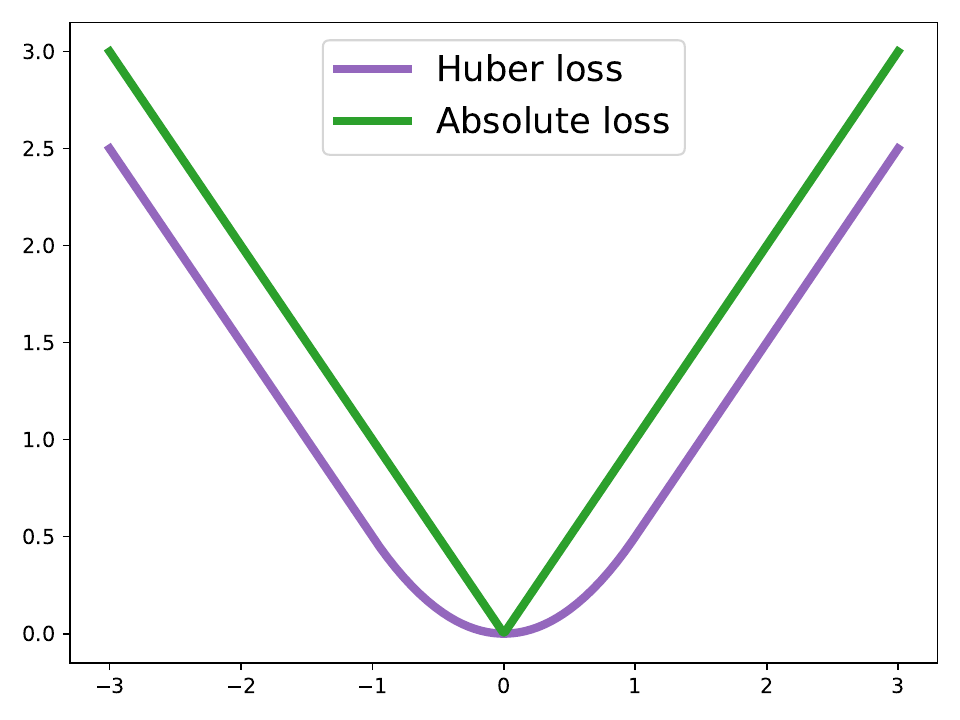}
\caption{The Huber loss is the Moreau envelope of the absolute loss.}
\label{smoothing:fig:huber}
\end{figure}

\begin{figure}[t]
\centering
\includegraphics[scale=0.35]{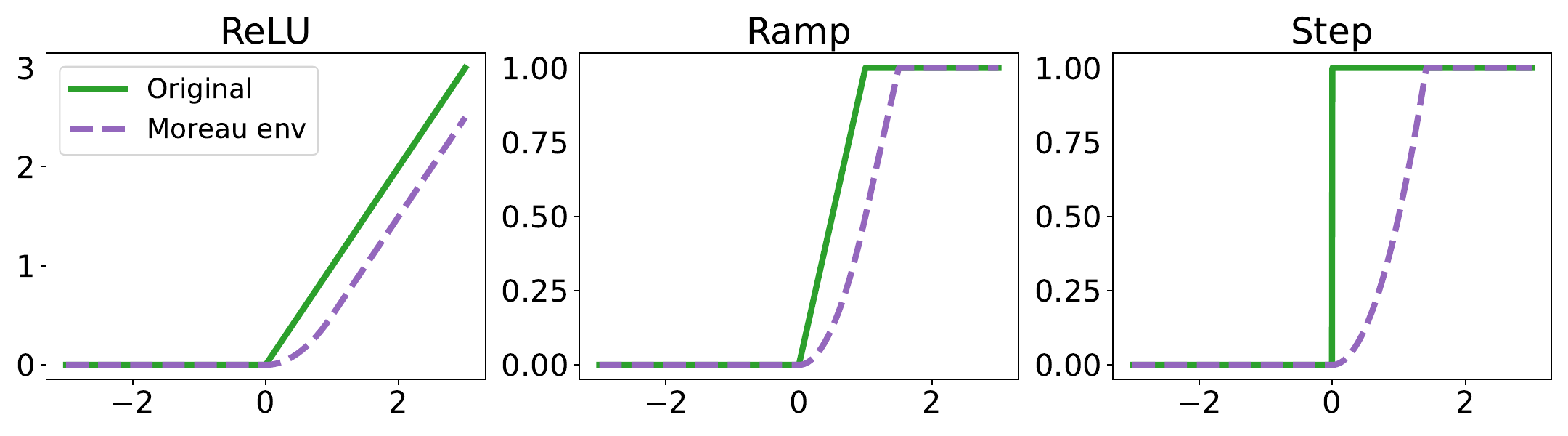}
\caption{The Moreau envelope is not limited to convex functions.
For instance, the ramp function is continuous but nonconvex,
and the step function is not only nonconvex but also discontinuous.
In this figure, we approximately computed the infimum over $u \in \RR$
in \cref{smoothing:def:moreau_env} by restricting the search on a finite grid,
in a closed interval.
}
\label{smoothing:fig:moreau}
\end{figure}

To illustrate smoothing from the Moreau envelope perspective, we 
show how to smooth the $1$-norm. In this case,
we obtain an analytical expression for the Moreau envelope.
\begin{boxexm}{Smoothing the $1$-norm via infimal convolution}
We wish to smooth $f(\u) \coloneqq \|\u\|_1 = \sum_{j=1}^M |u_j|$. 
The corresponding proximal operator is
the soft-thresholding operator (see \cref{optim:sec:prox_grad}),
\begin{align*}
\mathrm{prox}_{f}(\muv) 
&= \argmin_{\u \in \RR^M} \frac{1}{2} \|\muv - \u\|^2_2 + \|\u\|_1 \\
&= \mathrm{sign}(\muv) \cdot \max(|\muv| - 1, 0).
\end{align*}
Using \cref{smoothing:eq:env_from_prox}
and after some algebraic manipulations,
we obtain
\begin{equation*}
\env_f(\muv) 
= \sum_{j=1}^M \mathrm{huber}(\mu_j)
\approx \sum_{j=1}^M |\mu_j|,
\end{equation*}
where we defined the \textbf{Huber loss}
\begin{equation*}
\mathrm{huber}(\mu_j)
\coloneqq
\begin{cases}
\frac{\mu_j^2}{2} &\mbox{ if } |\mu_j| \le 1 \\
|\mu_j| - \frac{1}{2} &\mbox{ if } |\mu_j| > 1
\end{cases}.
\end{equation*}
This is illustrated in \cref{smoothing:fig:huber} with $M=1$.
\label{smoothing:exm:huber_primal}
\end{boxexm}
We also illustrate in \cref{smoothing:fig:moreau}
that the Moreau envelope of nonconvex functions can be approximately computed
numerically.

\subsection{Vector-valued functions}

The Moreau envelope is defined by $\env_f(\muv) \coloneqq
\inf_{\u \in \RR^M} f(\u) + \frac{1}{2} \|\muv - \u\|_2^2$.
As such, it is limited to scalar-valued functions $f \colon \RR^M \to \RR$.
To extend the Moreau envelope to vector-valued functions $f \colon \RR^M \to
\RR^T$, where $f(\u) = (f_1(\u), \dots, f_T(\u))$ and $f_i \colon \RR^M \to \RR$
for $i \in [T]$, we may choose to smooth each $f_j$ separately to define
\begin{equation*}
\envv_f(\muv) \coloneqq (\env_{f_1}(\muv), \dots, \env_{f_T}(\muv)),
\end{equation*}
where
\begin{equation*}
\env_{f_i}(\muv) = \inf_{\u_i \in \RR^M} f_i(\u_i) 
+ \frac{1}{2} \|\muv - \u_i\|_2^2.
\end{equation*}
This approach requires to solve $T$ separate minimization problems
and performs the smoothing of each output coordinate $i \in [T]$ independently.
From \cref{diff:prop:multiple_outputs}, 
we then have that the VJP of $\envv_f(\muv)$ with any direction 
$\d = (d_1, \dots, d_T) \in \RR^T$ is
\begin{align*}
\partial  \envv_f(\muv)^*[\d]
&= \sum_{i=1}^T \partial \env_{f_i}(\muv)^*[d_i] \\
&= \sum_{i=1}^T d_i \nabla \env_{f_i}(\muv).
\end{align*}
In the particular case $f(\u) = (f_1(u_1), \dots, f_T(u_T))$, we obtain
\begin{equation*}
\partial \envv_f(\muv)^*[\d] = \sum_{i=1}^T d_i \nabla \env_{f_i}(\mu_i).
\end{equation*}

An alternative was proposed by \citet{roulet_2022}.
For a differentiable function $f \colon \RR^M \to \RR^T$,
we recall that the VJP of $f$ with a direction $\d \in \RR^T$ reads
\begin{equation*}
\partial f(\u)^*[\d] = \nabla \langle f, \d \rangle(\u),
\end{equation*}
where we defined the scalar-valued function
$\langle f, \d \rangle(\u) \coloneqq \langle f(\u), \d \rangle$.
As a result, if $f$ is non-differentiable, a natural idea is to approximate
its VJP $\partial f(\u)^*[\d]$ (had it existed) 
by the gradient $\nabla \env_{\langle f, \d \rangle}(\muv)$ of the Moreau envelope
\begin{equation}
\env_{\langle f, \d \rangle}(\muv) 
= \inf_{\u \in \RR^M} \langle f(\u), \d \rangle
+ \frac{1}{2} \|\muv - \u\|^2_2.
\label{smoothing:eq:moreau_vjp}
\end{equation}
This requires solving a single optimization problem, independently of the
number of outputs $T$.
Moreover, for $\d = \e_i$, this recovers $\env_{f_i}(\muv)$ as a special case.

This approach allows in principle to perform reverse-mode autodiff (gradient
backpropagation) on a neural
network whose layers use the Moreau envelope. 
Indeed, following \cref{smoothing:prop:moreau_properties},
the approximate VJP of $f$ with a direction $\d$ is given by
\begin{equation*}
\partial f(\muv)^*[\d] 
\approx    
\nabla \env_{\langle f, \d \rangle}(\muv) 
= \muv - \u^\star,
\end{equation*}
where $\u^\star$ is the solution of the minimization problem in
\cref{smoothing:eq:moreau_vjp}.
However, we emphasize that this minimization problem could be difficult to
solve in general.  Indeed, 
when performing gradient backpropagation,
the direction $\d$ is not necessarily non-negative,
therefore the function being minimized in \cref{smoothing:eq:moreau_vjp} could
be nonconvex, even if each $f_i$ is convex.
Another potential caveat is that the direction $\d$ influences the smoothing
strength,
while in principle we should be able to smooth a function independently of
whether we compute its VJP or not.
To see that,
for example in the particular case $f(\u) = (f_1(u_1), \dots, f_T(u_T))$,
one easily checks that for $\d = (d_1, \dots, d_T)$, we get
\begin{equation*}
\env_{\langle f, \d \rangle}(\muv) = \sum_{i=1}^T \env_{d_i f_i}(\mu_i).
\end{equation*}
Smoothing vector-valued functions by Moreau envelope (or more generally, by
infimal convolution) remains an open area of research.
We will see in \cref{chap:conv} that smoothing by convolution more naturally
supports vector-valued functions.

\section{Legendre–Fenchel transforms, convex conjugates}
\label{duality:sec:conjugates}

The Legendre-Fenchel transform,
\aka convex conjugate,
is a way to turn a function $f$
into a new function, denoted $f^*$.
We now review it in detail, as it plays a major role for the dual approach to
smoothing.

\subsection{Definition}

\begin{figure}
\centering
\includegraphics[width=0.8\linewidth]{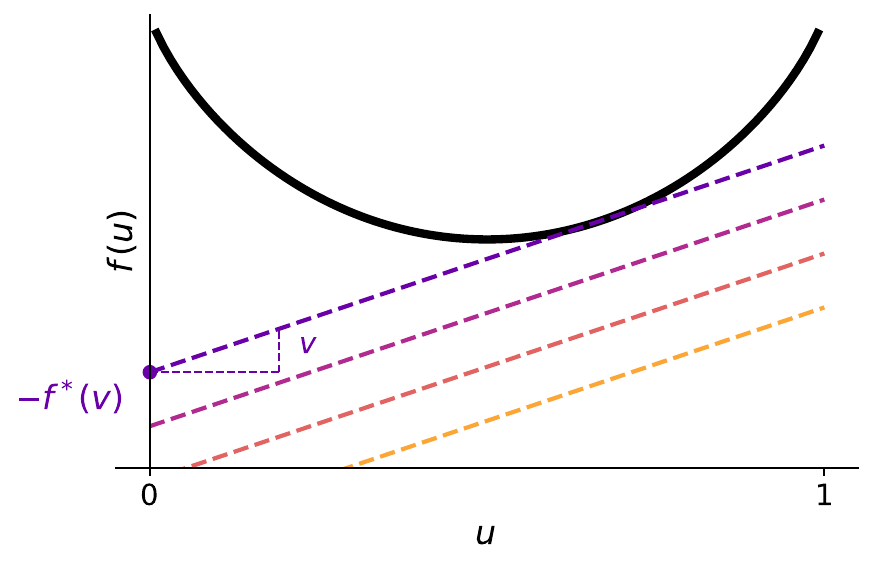}
\caption{For a fixed slope $v$, the function $u \mapsto uv - f^*(v)$
is the tightest affine lower bound of $f$ with slope $v$.}
\label{duality:fig:conjugate}
\end{figure}

Consider the class of affine functions of the form
\begin{equation*}
\u \mapsto \langle \u, \v \rangle - b.
\end{equation*}
These functions are parametrized by their slope 
$\v \in \RR^M$ 
and their intercept 
$-b \in \RR$.
Now, suppose we fix $\v$. 
Given a function $f(\u)$, affine lower bounds of $f(\u)$ are all the 
functions of $\u$ such that $b$ satisfies for all $\u \in \RR^M$,
\begin{equation*}
\langle \u, \v \rangle - b \le f(\u)
\iff
\langle \u, \v \rangle - f(\u) \le b.
\end{equation*}
The \textbf{tightest} 
lower bound is then defined by $b$ such that
\begin{equation*}
b \coloneqq \sup_{\u \in \dom(f)} \langle \u, \v \rangle - f(\u),
\end{equation*}
where we recall that the domain of $f$ is defined by
\begin{equation*}
\dom(f) \coloneqq \{\u \in \RR^M \colon f(\u) < \infty\}.
\end{equation*}
This leads to the definition of \textbf{Legendre-Fenchel transform},
\aka \textbf{convex conjugate}.
\begin{boxdef}{Legendre-Fenchel transform, convex conjugate}
Given a function $f \colon \RR^M \to \RR \cup \{\infty\}$,
its convex conjugate is defined by
\begin{equation*}
f^*(\v) \coloneqq \sup_{\u \in \dom(f)}
\langle \u, \v \rangle - f(\u).
\end{equation*}
\label{duality:def:conjugate}
\end{boxdef}
We use a $\sup$ rather than a $\max$ to indicate that $f^*(\v)$ is potentially
$\infty$.
Following the previous discussion, $-f^*(\v)$ is the intercept of the tightest
affine lower bound with slope $\v$ of $f(\u)$.
This is illustrated in \cref{duality:fig:conjugate}.

\begin{figure}
  \centering
  \includegraphics[width=\linewidth]{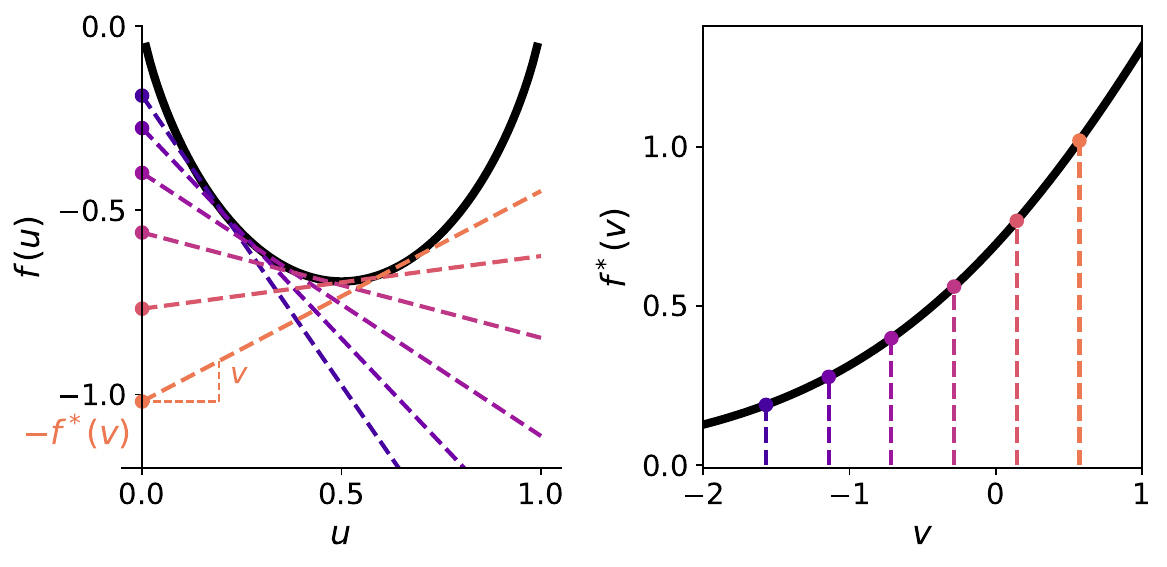}
  \caption{
\textbf{Left:}
instead of representing a convex function $f$ by its graph $(\u, f(\u))$ for $\u
\in \dom(f)$, we can represent it by the set of tangents with slope $\v$ and
intercept $-f^*(\v)$ for $\v \in \dom(f^*)$.
\textbf{Right:}
by varying the slope $v$ of all possible tangents, we
obtain a function of the slope $v$ rather than of the original input
$u$. The colors of the tangents on the left are chosen to match the colors of
the vertical lines on the right.
\label{smoothing:fig:convex_conjugate}
}
\end{figure}

The Legendre-Fenchel transform is a function transformation, as it produces a
new function $f^*$.  It can be seen as a dual representation of a function:
instead of representing a convex function $f$ by its graph $(\u, f(\u))$ for $\u
\in \dom(f)$, we can represent it by the set of tangents with slope $\v$ and
intercept $-f^*(\v)$ for $\v \in \dom(f^*)$, as illustrated
in~\cref{smoothing:fig:convex_conjugate}.  As the name ``convex conjugate''
indicates, it is convex, even if the original function is not. 

\subsection{Closed-form examples}

Computing $f^*(\v)$ involves the resolution of a maximization problem, which
could be difficult in general without assumption on $f$. 
In some cases, however, we can compute an analytical expression, as we now
illustrate.
\begin{boxexm}{Analytical conjugate examples}
When $f(\u) = \frac{1}{2} \|\u\|_2^2$, 
with $\dom(f) = \RR^M$,
the conjugate is
\begin{equation*}
f^*(\v) = \max_{\u \in \RR^M} \langle \u, \v \rangle - \frac{1}{2} \|\u\|^2_2. 
\end{equation*}
Setting the gradient 
$\u \mapsto \langle \u, \v \rangle - \frac{1}{2} \|\u\|^2_2$
to zero, we obtain $\u^\star = \v$.
Plugging $\u^\star$ back, we therefore obtain
\begin{equation*}
f^*(\v) 
= \langle \u^\star, \v \rangle - \frac{1}{2} \|\u^\star\|^2_2
= \frac{1}{2} \|\v\|^2_2.
\end{equation*}
Therefore, $f = f^*$ in this case.

When $f(\u) = \langle \u, \log \u \rangle$, with $\dom(f) = \RR_+^M$, the
minimizer of $\u \mapsto \langle \u, \v \rangle - \langle \u, \log \u \rangle$ 
is $\u^\star = \exp(\v-\ones)$ and the conjugate is
\begin{equation*}
f^*(\v) =  \sum_{j=1}^M \exp(v_j - 1).
\end{equation*}
\label{duality:ex:conjugates}
\end{boxexm}
See for instance \citet{boyd_2004} or \citet{beck_2017} for many more examples.

\subsubsection*{Constraining the domain}

We can incorporate constraints using an \textbf{indicator function} with values
in the extended real line $\RR \cup \{\infty\}$,
\begin{equation*}
\indic_\cC(\u)
\coloneqq \begin{cases}
    0 &\mbox{ if } \u \in \cC \\
    +\infty &\mbox{ otherwise }
\end{cases}.
\end{equation*}
\begin{boxexm}{Incorporating constraints}
If $f(\u) = \indic_\cC(\u)$,
where $\cC$ is a convex set, then
\begin{equation*}
f^*(\v) 
= \sup_{\u \in \dom(f)} ~ \langle \u, \v \rangle - f(\u)
= \sup_{\u \in \cC} ~ \langle \u, \v \rangle
\coloneqq \sigma_{\cC}(\v),
\end{equation*}
which is known as the \textbf{support function} of $\cC$. 
The corresponding argmax (assuming that it exists),
\begin{equation*}
\v \mapsto \argmax_{\u \in \cC} \langle \u, \v \rangle,
\end{equation*}
is known as the \textbf{linear maximization oracle} (LMO) of $\cC$. As another
example, if $f(\u) = \langle \u, \log \u \rangle + \indic_{\triangle^M}(\u)$
then
\begin{equation*}
f^*(\v) = \lse(\v) = \log \sum_{i=1}^M \exp(v_i). 
\end{equation*}
We postpone a proof to \cref{smoothing:prop:softmax_analytical_expression}.
\end{boxexm}

\subsection{Properties}

The conjugate enjoys several useful properties, that we now summarize.
\begin{boxprop}{Convex conjugate properties}
~
\begin{enumerate}
\item \textbf{Convexity:}
$f^*(\v)$ is a \textbf{convex} function for \textbf{all} 
$f \colon \RR^M \to \RR \cup \{\infty\}$ (even if $f$ is nonconvex).

\item \textbf{Fenchel-Young inequality:}
for all $\u, \v \in \RR^M$
\begin{equation*}
f(\u) + f^*(\v) - \langle \u, \v \rangle \ge 0.
\end{equation*}

\item \textbf{Gradient:}
if the supremum in \cref{duality:def:conjugate} is uniquely achieved,
then $f^*(\v)$ is differentiable at $\v$ and its gradient is
\begin{equation*}
\nabla f^*(\v) 
= \argmax_{\u \in \dom(f)} \langle \u, \v \rangle - f(\u).
\end{equation*}
Otherwise, $f^*(\v)$ is sub-differentiable at $\v$ and we get a sub-gradient
instead.

\item \textbf{Maps:} If $f$ and $f^*$ are differentiable, then
\begin{equation*}
\v = \nabla f(\u) 
\iff
\u = \nabla f^*(\v) 
\iff
f^*(\v) + f(\u) - \langle \u, \v \rangle = 0.
\end{equation*}

\item \textbf{Biconjugate:} $f = f^{**}$ if and only if $f$ is convex and closed
(i.e., its sublevel sets form a closed set), otherwise $f^{**} \le f$.
\end{enumerate}
\label{duality:prop:conjugate_properties}
\end{boxprop}
\begin{proof}
\begin{enumerate}
    \item This follows from the fact that 
$\v \mapsto \sup_{\u \in \cC} g(\u, \v)$ is convex
if $g$ is convex in $\v$. 
Note that this is true even if $g$ is nonconvex in $\u$.
Here, $g(\u, \v) = \langle \u, \v \rangle - f(\u)$,
which is affine in $\v$ and therefore convex in $\v$.

\item This follows immediately from \cref{duality:def:conjugate}.

\item This follows from Danskin's theorem, reviewed in
    \cref{implicit:sec:danskin}. Another way to see this is by observing that
\begin{align*}
f^*(\v) &= \langle \g, \v \rangle - f(\g) \\
f^*(\v') &\ge \langle \g, \v' \rangle - f(\g),
\end{align*}
where $\g \coloneqq \displaystyle{\argmax_{\u \in \dom(f)}} \langle \u, \v
\rangle - f(\u)$. Subtracting the two, we obtain
\begin{equation*}
f^*(\v') \ge f^*(\v) + \langle \g, \v' - \v \rangle.
\end{equation*}
Now, using that $f^*$ is convex and \cref{optim:def:convex_differentiable},
we obtain that $\g = \nabla f^*(\v)$.

\item See, e.g., \citet[Proposition 16.10]{bauschke2011convex}.

\item See \citet[Section 3.3]{boyd_2004}.

\end{enumerate}
\end{proof}

\subsection{Conjugate calculus}

While deriving a convex conjugate expression can be difficult in general,
in some cases, it is possible to use simple rules to derive conjugates in terms
of other conjugates.
\begin{boxprop}{Conjugate calculus rules}
~
\begin{enumerate}

\item \textbf{Separable sum of functions:}
if $f(\u) = \sum_{j=1}^M f_j(u_j)$, then
\begin{equation*}
f^*(\v) = \sum_{j=1}^M f^*_j(v_j).
\end{equation*}

\item \textbf{Scalar multiplication:}
if $f(\u) = c \cdot g(\u)$, for $c > 0$, then
\begin{equation*}
f^*(\v) = c \cdot g^*(\v / c).
\end{equation*}

\item \textbf{Addition to an affine function and translation:}
if $f(\u) = g(\u) + \langle \alphav, \u \rangle + \beta$, then
\begin{equation*}
f^*(\v) = g^*(\v - \alphav) - \beta. 
\end{equation*}

\item \textbf{Composition with an invertible linear map:}
if $f(\u) = g(M\u)$,
where $x\mapsto M\x$ is an invertible linear map,
then
\begin{equation*}
f^*(\v) = g^*(M^{-T} \v).
\end{equation*}

\item \textbf{Non-separable sum of functions:}
if $h_1$ and $h_2$ are convex functions, then
$(h_1 + h_2)^* = h_1^* \square h_2^*$, where $\square$ is the infimal
convolution operator.

\end{enumerate}
\label{duality:prop:conjugate_calculus}
\end{boxprop}

\subsection{Fast Legendre transform}

When an analytical expression is not available, 
we can resort to numerical schemes
to approximately compute the transform / conjugate.
When $f$ is convex, because $-f$ is concave, the maximization
in \cref{duality:def:conjugate} is that of a concave function. 
Therefore, the conjugate can be
computed to arbitrary precision in polynomial time using classical iterative
algorithms for constrained optimization such as
projected gradient descent (\cref{optim:sec:proj_grad})
or conditional gradient \aka Frank-Wolfe \citep{jaggi_2013}.
Without convexity assumption on $f$,
$f^*(\v)$ can be approximated by
\begin{equation*}
f^*(\v) \approx \sup_{\u \in \cU} \langle \u, \v \rangle - f(\u),
\end{equation*}
where $\cU \subseteq \dom(f)$ is a discrete grid of values.
We can then compute $f^*(\v)$ for several inputs $\v \in \cV$
using the linear-time Legendre transform algorithm
\citep{lucet_1997}, where $\cV \subseteq \dom(f^*)$
is another discrete grid.
The complexity is $O(|\cU| \cdot |\cV|)$, which is linear in the grid sizes.
However, the grid sizes are typically $|\cU| = |\cV| = O(N^M)$, for $N$
equally-distributed points in each of the $M$ dimensions. Therefore, this
approach is limited to small-dimensional settings, e.g., $M \in \{1,2,3\}$.

\section{Dual approach}
\label{smoothing:sec:dual_approach}

Previously, we presented how to smooth a function by performing its infimal
convolution with a primal-space regularization $R$. 
We now present how to smooth a
function by regularizing its Legendre-Fenchel transform (convex conjugate)
instead.
This dual, equivalent approach, is often mathematically more convenient.

\subsection{Duality between strong convexity and smoothness}

We begin by stating a well-known result that will underpin this whole section:
smoothness and strong convexity are dual to each other
\citep{hiriart_1993,kakade_2009,beck_2017,zhou_2018}.
\begin{boxprop}{Duality between strong convexity and smoothness}
$f$ is $\frac{1}{\mu}$-strongly convex \wrt the norm $\|\cdot\|$ over $\dom(f)$
if and only if $f^*$ is $\mu$-smooth \wrt the dual norm $\|\cdot\|_*$ over
$\dom(f^*)$.
\label{smoothing:prop:duality_smoothness}
\end{boxprop}
For a review of the notions of smoothness and strong convexity, see
\cref{optim:sec:function_classes}. 
We give two examples of strongly-convex and smooth conjugate pairs in 
\cref{smoothing:tab:conjugates}.

\begin{table}[t]
\caption{Examples of strongly-convex and smooth conjugate pairs.}
\begin{tabular}{lccccc}
\toprule
Function & Norm & Domain & Conjugate & Dual norm & Dual domain \\
\midrule
  $\frac{1}{2} \|\u\|_2^2$ & $\|\cdot\|_2$ & $\RR^M$
& $\frac{1}{2} \|\v\|_2^2$ & $\|\cdot\|_2$ & $\RR^M$ \\
  $\langle \u, \log \u \rangle$ & $\|\cdot\|_1$ & $\triangle^M$
& $\mathrm{logsumexp}(\v)$ & $\|\cdot\|_\infty$ & $\RR^M$ \\
\bottomrule
\end{tabular}
\label{smoothing:tab:conjugates}
\end{table}

\subsection{Smoothing by dual regularization}

The duality between smoothness and strong convexity suggests a generic approach
in order to smooth a function $f \colon \RR^M \to \RR$, by going through the
\textbf{dual} space.
\begin{enumerate}
    \item Compute the conjugate $f^*$:
\begin{equation*}
f^*(\v) \coloneqq \sup_{\u \in \dom(f)} \langle \u, \v \rangle - f(\u).
\end{equation*}
\item Add strongly-convex regularization $\Omega$ to the conjugate:
\begin{equation}
f_\Omega^*(\v) \coloneqq f^*(\v) + \Omega(\v).
\label{smoothing:eq:regularized_conjugate}
\end{equation}

\item Go back to the primal space, by computing the conjugate of $f^*_\Omega$:
\begin{equation*}
f_\Omega(\u) 
\coloneqq f^{**}_\Omega(\u) 
= \max_{\v \in \RR^M} \langle \u, \v \rangle - f_\Omega^*(\v).
\end{equation*}
\end{enumerate}
Note that $\u$ and $\v$ belong to different spaces, i.e.,
$\u \in \dom(f)$ and $\v \in \dom(f^*)$.
Following \cref{smoothing:prop:duality_smoothness},
if $\Omega$ is $\mu$-strongly convex, then $f_\Omega(\u)$ is
$\frac{1}{\mu}$-smooth.
Furthermore, following \cref{duality:prop:conjugate_properties},
$f_\Omega(\u)$ is convex, even if $f$ is nonconvex.
Therefore, $f_\Omega(\u)$ is a \textbf{smooth and convex relaxation} of
$f(\u)$. 

Steps $1$ and $3$ are the most challenging, as they both require the
derivation of a conjugate.
While an analytical solution may not exist in general,
in some simple cases, there is, as we now illustrate.
\begin{boxexm}{Smoothing the $1$-norm via dual regularization}
We revisit \cref{smoothing:exm:huber_primal},
this time from the dual perspective.
We wish to smooth out the $1$-norm 
$f(\u) \coloneqq \|\u\|_1 = \sum_{j=1}^M |u_j|$.
\begin{enumerate}
    \item \textbf{Compute the conjugate.} The conjugate of any norm $\|\cdot\|$
is the indicator function of the dual norm's unit ball $\{\v \in \RR^M \colon
\|\v\|_* \le 1\}$ (see e.g.~\citet[Example 3.26]{boyd_2004}). The dual norm of
$\|\u\|_1$ is $\|\v\|_\infty$. Moreover,
\begin{equation*}
\{ \v \in \RR^M \colon \|\v\|_\infty \le 1\}
=
[-1, 1]^M.
\end{equation*}
Recalling that $\indic_\cC$ is the indicator function of $\cC$,
we obtain
\begin{equation*}
f^*(\v) = \indic_{[-1,1]^M}(\v).
\end{equation*}

\item \textbf{Adding strongly-convex regularization.}
We add quadratic regularization 
$\Omega(\v) \coloneqq \frac{1}{2} \|\v\|_2^2$ 
to define
\begin{equation*}
f^*_\Omega(\v) \coloneqq \indic_{[-1,1]^M}(\v) + \Omega(\v).
\end{equation*}

\item \textbf{Going back to the primal.}
\begin{equation*}
f_\Omega(\u) 
= f_\Omega^{**}(\u)
= \langle \u, \v^\star \rangle - \Omega(\v^\star) 
= \sum_{i=1}^M \mathrm{huber}(u_i),
\end{equation*}
where
$
\v^\star 
= \mathrm{clip}\left(\u\right)
\coloneqq \max\left(\min\left(\u, 1\right), -1\right)
$.
\end{enumerate}
We therefore indeed recover the Huber loss from
\cref{smoothing:exm:huber_primal}.
\label{smoothing:ex:1_norm}
\end{boxexm}
ReLU functions can be smoothed out in a similar way, as we see in more details
in \cref{smoothing:sec:relu}. 

The dual approach allows us to easily bound the smoothed function in terms of
the original function.
\begin{boxprop}{Bounding the smoothed function}\label{smoothing:prop:bounds}

Let $f$ be a closed convex function.
Let $\Omega$ be a convex function.
If $\Omega$ is such that
$\cL_\Omega \le \Omega(\v) \le \cU_\Omega$ for all $\v \in \dom(\Omega)$,
then for all $\u \in \RR^M$,
\begin{equation*}
    f(\u) - \cU_\Omega \le f_\Omega(\u) \le f(\u) - \cL_\Omega.
\end{equation*}
\end{boxprop}
\begin{proof}
Let us define
\begin{align*}
\v^\star &\coloneqq \argmax_{\v \in \RR^M} \langle \u, \v \rangle - f^*(\v)\\
\v^\star_\Omega &\coloneqq \argmax_{\v \in \RR^M} \langle \u, \v \rangle 
- f^*_\Omega(\v),
\end{align*}
where we recall that $f^*_\Omega \coloneqq f^* + \Omega$.
We then have for all $\u \in \RR^M$
\begin{align*}
f_\Omega(\u)
&= \langle \u, \v^\star_\Omega \rangle - f_\Omega^*(\v^\star_\Omega) \\
&\ge \langle \u, \v^\star \rangle - f_\Omega^*(\v^\star) \\
&= \langle \u, \v^\star \rangle - f^*(\v^\star) - \Omega(\v^\star) \\
&= f^{**}(\u) - \Omega(\v^\star) \\
&= f(\u) - \Omega(\v^\star),
\end{align*}
Similarly,
\begin{align*}
f(\u) - \Omega(\v^\star_\Omega)
&= \langle \u, \v^\star \rangle - f^*(\v^\star) - \Omega(\v^\star_\Omega) \\
&\ge \langle \u, \v^\star_\Omega \rangle - f_\Omega^*(\v^\star_\Omega) \\
&= f_\Omega^{**}(\u) \\
&= f_\Omega(\u).
\end{align*}
Combining the two with
$\cL_\Omega \le \Omega(\v) \le \cU_\Omega$ for all $\v \in \dom(\Omega)$,
we obtain
\begin{align*}
f(\u) - \cU_\Omega
&\le
f(\u) - \Omega(\v^\star) \\
&\le
f_\Omega(\u) \\
&\le
f(\u) - \Omega(\v^\star_\Omega) \\
&\le f(\u) - \cL_\Omega.
\end{align*}
\end{proof}

\begin{boxrem}{The gradient is differentiable almost everywhere}
\label{smoothing:rem:grad_diff_ae}

From \cref{duality:prop:conjugate_properties},
the gradient of $f_\Omega(\u)$ equals
\begin{equation*}
\nabla f_\Omega(\u)
= \argmax_{\v \in \RR^M}
\langle \u, \v \rangle - f_\Omega^*(\v).
\end{equation*}
If $\Omega$ is strongly convex, then $f_\Omega$ is smooth,
meaning that $\nabla f_\Omega$ is Lipschitz continuous.
From Rademacher's theorem reviewed in \cref{diff:sec:rademacher},
$\nabla f_\Omega$ is then differentiable almost everywhere (that is,
$f_\Omega$ is twice differentiable almost everywhere).
We use this property in the sequel to define continuous
differentiable almost everywhere relaxations of step functions
and argmax operators.
\end{boxrem}

\subsection{Equivalence between primal and dual regularizations}

So far, we saw two approaches to obtain a smooth approximation of a function $f$.
The first approach is based on the infimal convolution
$f \square R$, where $R \colon \dom(f) \to \RR$ denotes primal regularization.
The second approach is based on regularizing the Legendre-Fenchel transform
(convex conjugate) $f^*$ of $f$ with some dual regularization $\Omega$, to
define $f_\Omega = (f^* + \Omega)^*$.
It turns out that both approaches are equivalent.
\begin{boxprop}{Equivalence between primal and dual regularizations}
Let $f \colon \RR^M \to \RR \cup \{\infty\}$
and $R \colon \RR^M \to \RR \cup \{\infty\}$, both convex and closed.
Then,
$f_\Omega = (f^* + \Omega)^* = f \square R$
with $\Omega = R^*$.
\end{boxprop}
\begin{proof}
We have 
\begin{equation*}
f_\Omega(\u) 
= (f^* + \Omega)^*(\u)
= \sup_{\v \in \dom(f^*)} \langle \u, \v \rangle - f^*(\v) - \Omega(\v).
\end{equation*}
If $h_1$ and $h_2$ are convex, we have $(h_1 + h_2)^* = h_1^* \square h_2^*$
\citep[Theorem 4.17]{beck_2017}. Using $h_1 = f^*$ and $h_2 = \Omega = R^*$
gives the desired result using that $f^{**}= f$ and $R^{**} = R$ since both are
convex and closed (see \cref{duality:prop:conjugate_properties}).
\end{proof}
In particular, with $\Omega = \frac{1}{2}\|\cdot\|^2_2 = \Omega^*$, 
this shows that the Moreau envelope can equivalently be written as
\begin{equation*}
\env_f = f_\Omega = f_{\Omega^*}.
\end{equation*}
Given the equivalence between the primal and dual approaches, 
using one approach or the other is mainly a matter of mathematical or
algorithmic convenience, depending on the case. 

In this book, we focus on applications of smoothing techniques to differentiable
programming.  For applications to non-smooth optimization,
see \citet{nesterov_2005_smooth,beck_2012_smoothing}.

\subsection{Regularization scaling}

\subsubsection*{Dual approach}

If $\Omega$ is $1$-strongly convex, then $f_\Omega$ is a $1$-smooth
approximation of the original function $f$. To control the smoothness of the
approximation, we regularize with $\varepsilon \Omega$ for
$\varepsilon>0$, defining
\begin{equation*}
f_{\varepsilon \Omega}(\u) \coloneqq (f^* + \varepsilon \Omega)^*(\u) = \sup_{\v
\in \dom(f^*)} \langle \u, \v \rangle - f^*(\v) - \varepsilon \Omega(\v).
\end{equation*}
Consequently, the gradient is given by
\begin{equation*}
\nabla f_{\varepsilon \Omega}(\u) = \argmax_{\v \in \dom(f^*)} \langle \u, \v
\rangle - f^*(\v) - \varepsilon \Omega(\v).
\end{equation*}
This formulation works for any closed convex $f$.
The approximation error induced by the smoothing can be quantified using
\cref{smoothing:prop:bounds}. Provided that
$\cL_\Omega \le \Omega(\v) \le \cU_\Omega$ for all $\v \in \dom(\Omega)$,
we have
\begin{equation*}
    f(\u) - \varepsilon \cU_\Omega
    \le f_{\varepsilon \Omega}(\u)
    \le f(\u) - \varepsilon \cL_\Omega.
\end{equation*}

\begin{boxrem}{Homogeneous case}
In the special case where $f$ is positively homogeneous of degree $1$ (e.g., a
norm), its conjugate $f^*$ is an indicator function satisfying $f^* =
\varepsilon f^*$.  In this specific setting, we recover the scaling
relationships:
\begin{align*}
f_{\varepsilon \Omega}(\u) &= \varepsilon f_\Omega(\u/\varepsilon) \\
\nabla f_{\varepsilon \Omega}(\u) &= \nabla f_\Omega(\u/\varepsilon).
\end{align*}
\end{boxrem}

\subsubsection*{Primal approach}

Following \cref{smoothing:def:moreau_env},
if we use dual regularization $\varepsilon \Omega$, where $\varepsilon
> 0$ controls the regularization strength, 
assuming $f$ is convex, the corresponding primal regularization is 
$R = \varepsilon \Omega^*(\cdot / \varepsilon)$
and we have
\begin{equation*}
f_{\varepsilon \Omega} = f \square \varepsilon \Omega^*(\cdot / \varepsilon).
\end{equation*}
In the particular case $\Omega(\v) = \frac{1}{2} \|\v\|^2_2$,
we have
\begin{equation*}
R(\u) 
= \frac{\varepsilon}{2} \|\u / \varepsilon\|^2_2
= \frac{1}{2 \varepsilon} \|\u\|^2_2
= \frac{1}{\varepsilon} \Omega(\u).
\end{equation*}
We therefore get
\begin{equation*}
f_{\varepsilon \Omega} 
= f \square \frac{1}{\varepsilon} \Omega
= \frac{1}{\varepsilon} (\varepsilon f \square \Omega)
= \frac{1}{\varepsilon} \env_{\varepsilon f}.
\end{equation*}

\subsection{Generalized entropies}
\label{smoothing:sec:entropies}

A natural choice of dual regularization $\Omega(\piv)$, 
when $\piv \in \triangle^M$ is a discrete probability distribution,
is a negative entropy function, also known as \textbf{negentropy}.
Since negentropies play a major role in smoothed max operators,
we discuss them in detail here.

\subsubsection*{Information content and entropy}

An entropy function measures the amount of ``surprise'' of a random variable or
equivalently of a distribution. 
To define an entropy, we must first define the \textbf{information content} 
$I(E)$
of an event $E$. The value returned by such a function should be $0$ if the
probability of the event is $1$, as there is no surprise. Conversely,
information content should attain its maximal value if the probability of the
event is $0$, as it is maximally surprising.
Furthermore, the more probable an event $E$ is, the less surprising it is.
Therefore, when $p(E)$ increases, $I(E)$ should decrease.
Overloading the notation, we also write the information content of the outcome
$y$ of a random variable $Y$ as the information content of the event $\{Y =
y\}$,
\begin{equation*}
I(y) \coloneqq I(\{Y = y\}).
\end{equation*}

Given an information content function, we can then define the \textbf{entropy}
$H(Y)$ of a random variable $Y \in \cY$ as the expected information content,
\begin{equation*}
H(Y) \coloneqq \EE[I(Y)].
\end{equation*}
Different definitions of information content lead to different definitions of
entropy.

\subsubsection*{Shannon's entropy}

A definition of information content satisfying the criteria above is
\begin{equation*}
I(E) 
\coloneqq \log\left(\frac{1}{p(E)}\right )
= - \log p(E).
\end{equation*}
Indeed, $-\log 1 = 0$, $-\log 0 = \infty$ and $-\log$ is a decreasing function
over $(0,1]$.
Using this information content definition
leads to \textbf{Shannon's entropy} \citep{shannon_1948}
\begin{equation*}
H(Y) 
= \EE[I(Y)]   
= -\sum_{y \in \cY} p(y) \log p(y).
\end{equation*}
We can therefore define the Shannon entropy of a discrete probability
distribution $\piv \in \triangle^M$ as
\begin{equation*}
H(\piv) 
= -\sum_{i=1}^M \pi_i \log \pi_i 
= -\langle \piv, \log \piv \rangle
\end{equation*}
and use the corresponding negentropy as regularization
\begin{equation*}
\Omega(\piv) = -H(\piv) = \langle \piv, \log \piv \rangle.
\end{equation*}
The function is strongly convex \wrt $\|\cdot\|_1$ over $\triangle^M$.
However, it is not strongly convex over $\RR_+^M$, since this is not a bounded
set; see for instance \citep[Proposition 2]{blondel_2019}.
Since $\Omega$ is added to $f^*$
in \cref{smoothing:eq:regularized_conjugate},
we can therefore use this choice of $\Omega$ to smooth out a function $f$
if $\dom(f^*) \subseteq \triangle^M$.

\subsubsection*{Gini's entropy}

As an alternative, we can define information content as
\begin{equation*}
I(E) = \frac{1}{2}(1 - p(E)).
\end{equation*}
The $\frac{1}{2}$ factor is for later mathematical convenience.
This again satisfies the criteria of an information content function.
Indeed, 
i) when $p(E) = 1$, $I(E) = 0$ 
ii) when $p(E) = 0$, $I(E)$ attains its maximum of $\frac{1}{2}$
iii) the function is decreasing \wrt $p(E)$.
Using this information content definition
leads to \textbf{Gini's entropy} \aka Gini index \citep{gini_index}
\begin{equation*}
H(Y) 
= \EE[I(Y)]   
= \frac{1}{2} \sum_{y \in \cY} p(y) (1 - p(y)).
\end{equation*}
We can use Gini's negative entropy to define
for all $\piv \in \triangle^M$
\begin{equation*}
\Omega(\piv) 
= \frac{1}{2} \langle \piv, \piv - \ones \rangle
= \frac{1}{2} (\|\piv\|^2_2 - 1).
\end{equation*}
The function is strongly convex \wrt $\|\cdot\|_2$ over $\RR^M$.
We can therefore use this choice of $\Omega$ to smooth out a function $f$
if $\dom(f^*) \subseteq \RR^M$.
This means that the set of functions that we can smooth out with Gini entropy is
larger than the set of functions we can smooth out with Shannon entropy.

\begin{figure}[t]
\centering
\includegraphics[scale=0.4]{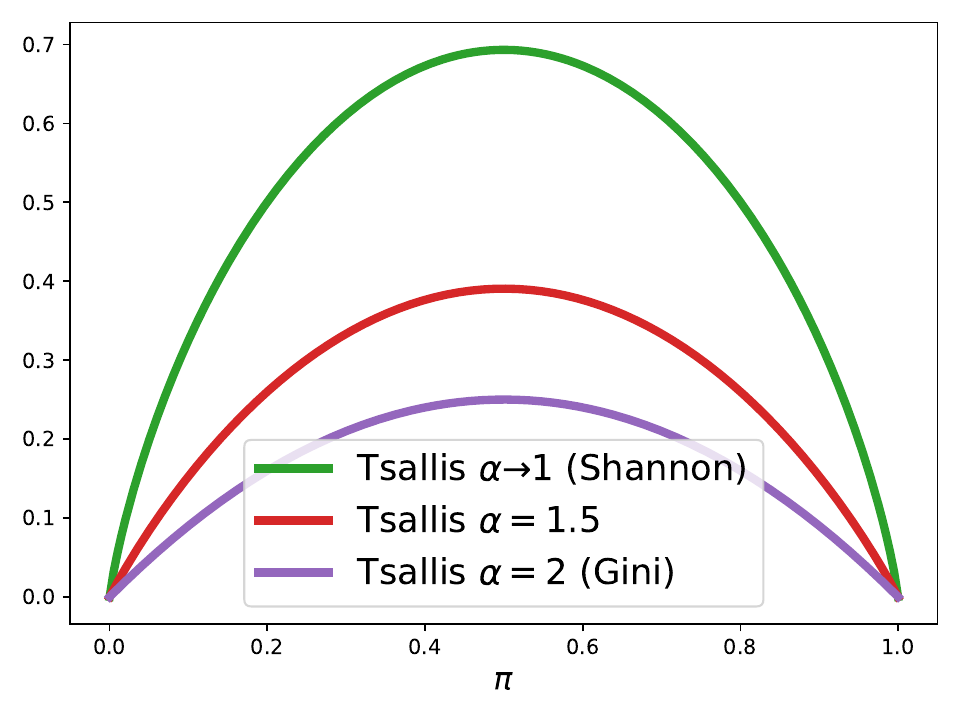}
\caption{Tsallis entropies of the distribution 
$\piv = (\pi, 1-\pi) \in \triangle^2$,
for $\pi \in [0,1]$. An entropy is a non-negative concave function
that attains its maximum at the uniform distribution, here $(0.5, 0.5)$.
A negative entropy, \aka negentropy, can be used as a dual regularization
function $\Omega$ to smooth out a function $f$ when $\dom(f^*) \subseteq
\triangle^M$.}
\label{smoothing:fig:entropies}
\end{figure}

\subsubsection*{Tsallis entropies}

\begin{figure}[t]
\centering
\includegraphics[scale=0.4]{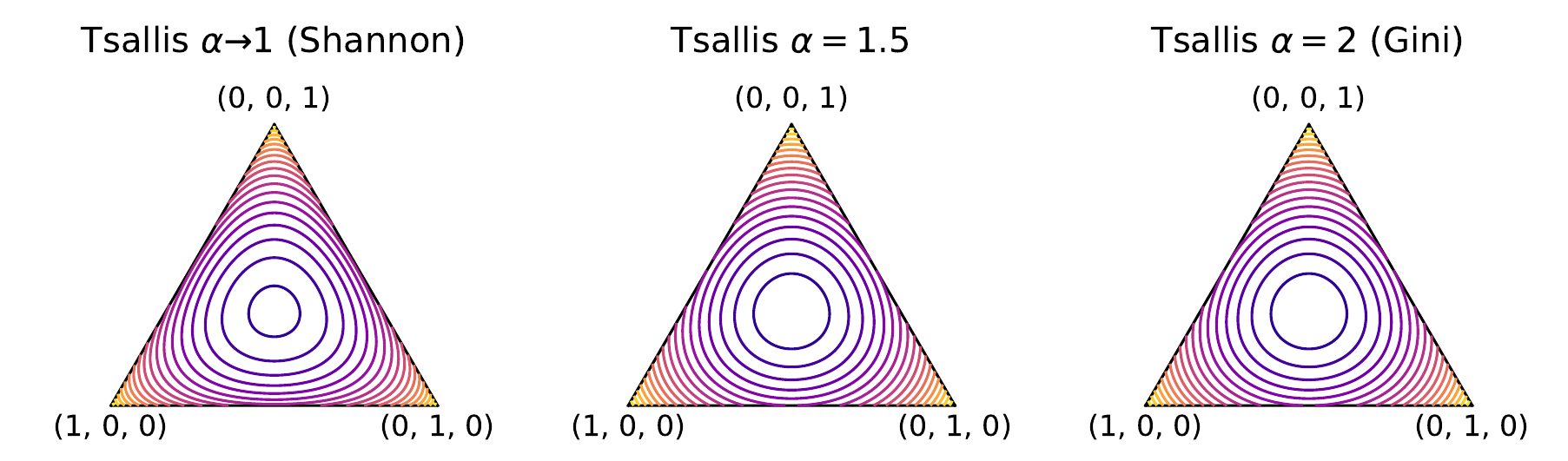}
\caption{
Contours of Tsallis entropies on the probability simplex.
}
\label{smoothing:fig:entropies_contours}
\end{figure}

Given $\alpha \ge 1$, a more general information content definition is
\begin{equation*}
I(E) = \frac{1}{\alpha (\alpha-1)}(1 - p(E)^{\alpha-1}).
\end{equation*}
Using this definition leads to the \textbf{Tsallis entropy} \citep{tsallis_1988}
\begin{equation*}
H(Y) 
= \EE[I(Y)]   
= \frac{1}{\alpha(\alpha-1)} \sum_{y \in \cY} p(y) (1 - p^{\alpha-1}(y)).
\end{equation*}
The Tsallis entropy recovers the Shannon entropy in the limit $\alpha \to 1$
and the Gini entropy when $\alpha = 2$.
We can use the Tsallis negative entropy to define
for all $\piv \in \triangle^M$
\begin{equation*}
\Omega(\piv) 
= \frac{1}{\alpha(\alpha-1)}\langle \piv, \piv^{\alpha-1} - \ones \rangle
= \frac{1}{\alpha(\alpha-1)} (\|\piv\|^\alpha_\alpha - 1),
\end{equation*}
where $\|\v\|_p$ is the $p$-norm for ($p \ge 1$)
\begin{equation*}
\|\v\|_p \coloneqq 
\left(\sum_{i=1}^M v_i^p\right)^{\frac{1}{p}},
\end{equation*}
so that
\begin{equation*}
\|\v\|_p^p = \sum_{i=1}^M v_i^p.
\end{equation*}
Tsallis entropies for $\alpha \to 1$ (Shannon entropy),
$\alpha=1.5$ and $\alpha=2$ (Gini entropy)
are illustrated in 
\cref{smoothing:fig:entropies}
and
\cref{smoothing:fig:entropies_contours}.

\subsubsection*{Definition and properties of generalized entropies}

So far, we saw how to define an entropy as the expected information content.
However, generalized entropies \citep{degroot_1962,grunwald2004game}
do not necessarily need to take this form.
We follow the definition of \citet{blondel_2020}.
\begin{boxdef}{Entropy function}

A function $H \colon \triangle^M \to \RR_+$ is an entropy if
\begin{enumerate}
    \item $H(\piv) = 0$ if $\piv \in \{\e_1,\dots,\e_M\}$,
    \item $H$ is strictly concave,
    \item $H(P \piv) = H(\piv)$ for any permutation matrix $P$.
\end{enumerate}
\end{boxdef}
This definition implies that $H$ is non-negative and is uniquely maximized by
the uniform distribution \citep[Proposition 4]{blondel_2020}.
This is indeed what we expect from an entropy function.
An example is the squared $p$-norm entropy \citep{blondel_2020}
\begin{equation*}
    H(\piv) = \frac{1}{2} - \frac{1}{2} \|\piv\|_p^2.
\end{equation*}
Since the squared $p$-norm is strongly convex for $p \in (1,2]$
\citep{ball2002sharp},
this entropy is strongly concave for $p \in (1,2]$ 
and can therefore be used to smooth out functions.

We now illustrate how to apply these techniques to compute smoothed ReLUs and
smoothed max operators, as well as continuous relaxations of 
step functions and argmax operators.

\section{Smoothed ReLU functions}
\label{smoothing:sec:relu}

To demonstrate the application of the smoothing techniques discussed in this
chapter, we begin by explaining how to smooth the ReLU function.
The ReLU function is defined by
\begin{equation*}
\mathrm{relu}(u)
\coloneqq 
\begin{cases}
    u &\mbox{ if } u \ge 0 \\
    0 &\mbox{ otherwise }
\end{cases}
= \max(u, 0).
\end{equation*}
We recall that in order to smooth a function $f$ by the dual approach,
we calculate its conjugate $f^*$, add regularization $\Omega$ to it to obtain
$f^*_\Omega \coloneqq f^* + \Omega$ 
and then obtain $f_\Omega$ by computing $f_\Omega^{**}$.

Here, we wish to smooth out $f = \mathrm{relu}$.
Its convex conjugate is 
\begin{equation*}
\mathrm{relu}^*(\pi) 
= \indic_{[0,1]}(\pi)
= \begin{cases}
    0 &\mbox{ if } \pi \in [0,1] \\
    \infty &\mbox{ if } \pi \not \in [0,1]
\end{cases}.
\end{equation*}
To notice why, we observe that 
\begin{equation}
\mathrm{relu}(u) =
\max_{\pi \in [0,1]} u \cdot \pi
= \max_{\pi \in \{0,1\}} u \cdot \pi
= \begin{cases}
    u &\mbox{ if } u \ge 0 \\
    0 &\mbox{ otherwise}
\end{cases}.
\label{smoothing:eq:relu_variational_form}
\end{equation}
Indeed, since the objective is linear in $\pi$, 
the maximum is attained at one of
the extreme points of $[0,1]$, so that we can replace the constraint
$\pi \in [0,1]$ with $\pi \in \{0,1\}$.
This shows that the ReLU is exactly the support function of $[0,1]$.
Since the conjugate of the support function is the indicator function,
we indeed obtain $\mathrm{relu}^* = \iota_{[0,1]}$.
We therefore have
\begin{equation*}
\mathrm{relu}^*_\Omega(\pi) 
= \mathrm{relu}^*(\pi) + \Omega(\pi)
= \iota_{[0,1]}(\pi) + \Omega(\pi)
\end{equation*}
and for some choice of $\Omega$, we need to be able to compute
\begin{align*}
\mathrm{relu}_\Omega(u) 
&= \max_{\pi \in \RR} u \cdot \pi - 
(\iota_{[0,1]}(\pi) + \Omega(\pi)) \\
&= \max_{\pi \in [0,1]} u \cdot \pi  
- \Omega(\pi).
\end{align*}

\subsubsection*{The softplus}

If we use the regularizer
$\Omega(\pi) = \pi \log \pi + (1-\pi) \log (1-\pi)$,
which comes from using Shannon's negentropy 
$\langle \piv, \log \piv \rangle$ with $\piv = (\pi, 1 - \pi)$,
we obtain
\begin{equation*}
\mathrm{relu}_\Omega(u) = \mathrm{softplus}(u) = \log(1 + \exp(u)).
\end{equation*}
This result is a special case of
\cref{smoothing:prop:softmax_analytical_expression}.

\subsubsection*{The sparseplus}

If we use the regularizer $\Omega(\piv) = \pi(\pi - 1)$, 
which comes from using Gini's negentropy with
$\frac{1}{2} \langle \piv, \piv - \ones \rangle$ 
with 
$\piv = (\pi, 1-\pi)$,
we obtain
\begin{equation*}
\mathrm{relu}_\Omega(u)
=
\mathrm{sparseplus}(u) 
=
\begin{cases}
  0, & u \leq -1\\
  \frac{1}{4}(u+1)^2, & -1 < u < 1 \\  
  u, & u \geq 1
\end{cases}.
\end{equation*}
See \cref{smoothing:fig:activ_sig} (left figure)
for a comparison of softplus and sparseplus.

\section{Smoothed max operators}
\label{smoothing:sec:max_argmax}

As a more elaborate application of the smoothing techniques discussed in this
chapter, we explain how to smooth max operators.
Smoothed max operators include smoothed ReLU functions as a special case.

\subsection{Definition and properties}

\begin{figure}[t]
  \centering
  \includegraphics[width=0.3\linewidth]{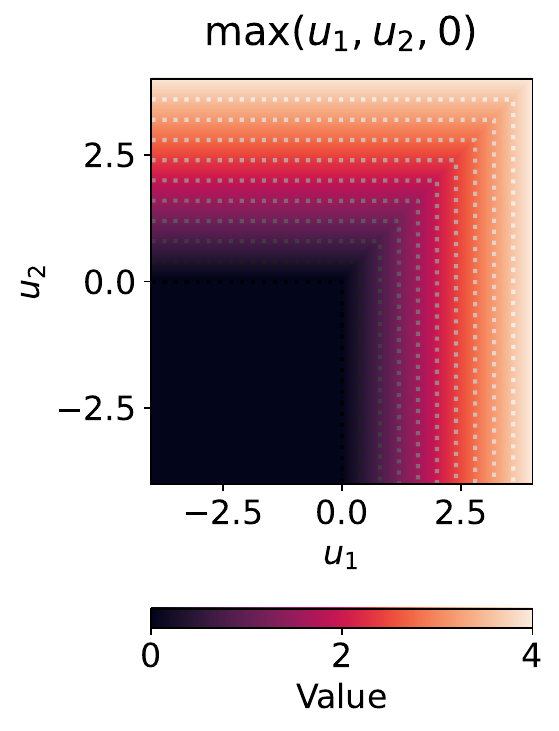}
  \includegraphics[width=0.3\linewidth]{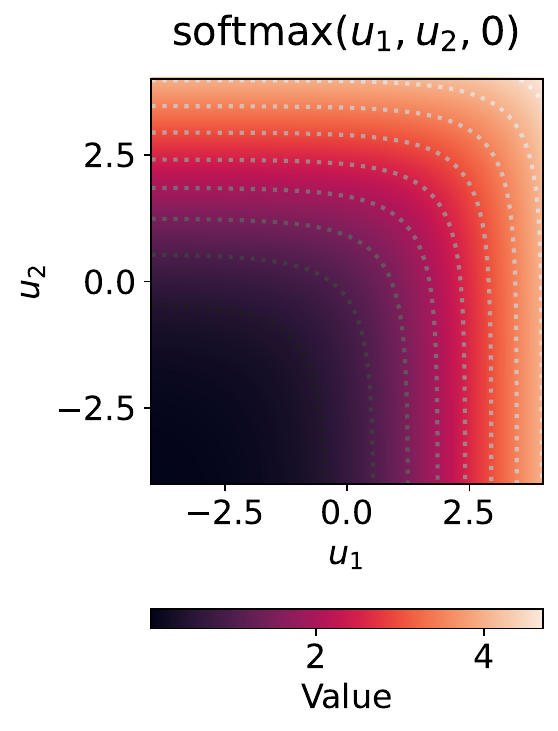}
  \includegraphics[width=0.3\linewidth]{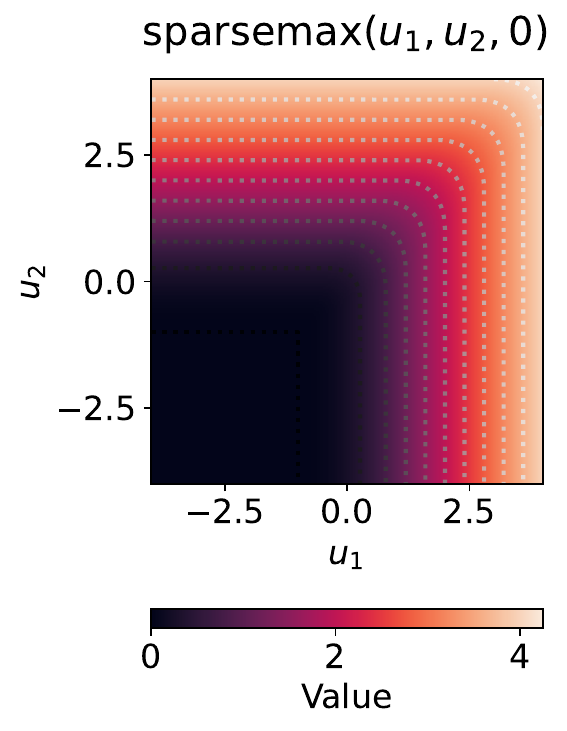}
  \caption{Max, softmax and sparsemax functions. The max function has non-smooth contour
  lines (set of points $\{\u \in \RR^3: f(\u) = c\}$ for some constant $c$
  represented by dashed gray lines). The gradient along these contour lines
  switches suddenly at the corners of the contour lines switch. This shows that
  the max function is not differentiable everywhere, namely, non-differentiable
  on the set of points $\{\u \in \RR^3: u_i = u_j\ \text{for any} \ i\neq j\}$. The
  contour lines of the softmax and sparsemax functions on the other hand are smooth
  illustrating that these functions are smooth counterparts of the max
  function. 
  \label{smoothing:fig:max}
  }
\end{figure}

With a slight notation overloading, 
given a vector $\u = (u_1, \dots, u_M) \in \RR^M$,
we define its maximum as
\begin{equation*}
\mathrm{max}(\u) \coloneqq \max_{j \in [M]} u_j.
\end{equation*}
To obtain a smooth approximation $\max_\Omega$ of $\max$,
we again apply the dual approach.
The conjugate of $\max$ is
\begin{equation*}
    \mathrm{max}^*(\piv) = \indic_{\triangle^M}(\piv).
\end{equation*}
To notice why, we observe that the vertices of the probability simplex 
$\triangle^M$ are
the standard basis vectors $\e_1, \dots, \e_M$. 
Since the objective is linear,
we then have
\begin{equation*}
\max(\u) 
=
\max_{\piv \in \triangle^M} \langle \u, \piv \rangle 
=
\max_{\piv \in \{\e_1, \dots, \e_M\}} \langle \u, \piv \rangle.
\end{equation*}
In other words, the maximum operator is exactly the support function
of $\triangle^M$. Since the conjugate of the support function is the indicator
function, we indeed obtain
$\mathrm{max}^* = \indic_{\triangle^M}$.
We can therefore write
\begin{equation*}
\text{max}_\Omega^*(\piv) 
= \text{max}^*(\piv) + \Omega(\piv) 
= \Omega(\piv) + \iota_{\triangle^M}(\piv)
\end{equation*}
and
\begin{align*}
\mathrm{max}_\Omega(\u)
&= (\Omega + \iota_{\triangle^M})^*(\u) \\ 
&= \max_{\piv \in \RR^M} \langle \u, \piv \rangle -
(\Omega(\piv) + \iota_{\triangle^M}(\piv)) \\
&= \max_{\piv \in \triangle^M} \langle \u, \piv \rangle - \Omega(\piv).
\end{align*}
The smoothed max operator $\max_\Omega$ can be useful in a neural network, for
example as a smoothed max pooling layer.
Its properties have been studied in \citep[Lemma 1]{mensch_2018},
as we recall here for convenience.
\begin{boxprop}{Properties of $\max_\Omega$}
\label{smoothing:prop:properties_max_Omega}

The following properties hold.
\begin{enumerate}
    \item \textbf{Bounds:} 
if $\cL_\Omega \le \Omega(\piv) \le \cU_\Omega$ for all $\piv \in \triangle^M$,
then $\max(\u) - \cU_\Omega \le \text{max}_\Omega(\u) \le \max(\u) - \cL_\Omega$
for all $\u \in \RR^M$.

\item \textbf{Monotonicity:} if $\u \le \v$ (element-wise),
    then $\text{max}_\Omega(\u) \le \text{max}_\Omega(\v)$.

\item \textbf{Commutativity:}
if $\Omega(P \piv) = \Omega(\piv)$ for any permutation matrix $P$ and any $\piv
\in \triangle^M$,
then $\text{max}_\Omega(P\u) = \text{max}_\Omega(\u)$ for any permutation matrix
$P$.

\item \textbf{Distributivity of $+$:}
$\text{max}_\Omega(\u + c \ones) = 
\text{max}_\Omega(\u) + c$ for all $\u \in \RR^M$ and all $c \in \RR$.
\end{enumerate}
\end{boxprop}
These properties are leveraged in \citep{mensch_2018} to create differentiable
dynamic programs. We consider in the following two possible choices of $\Omega$
leading to the softmax and sparsemax operators illustrated in
\cref{smoothing:fig:max}.

\subsubsection*{Smoothed min operators}

The minimum operator can be expressed in terms of the maximum operator,
since for all $\u \in \RR^M$,
\begin{equation*}
\min(\u) = -\max(-\u). 
\end{equation*}
Given a smoothed max operator $\mathrm{max}_\Omega$, 
we can therefore easily define a smoothed min operator as
\begin{align*}
\mathrm{min}_\Omega(\u) &\coloneqq -\mathrm{max}_\Omega(-\u).
\end{align*}

\subsection{Reduction to root finding}

Computing $\max_\Omega(\u)$ for a general strongly-convex regularization
$\Omega$ involves the resolution of a maximum over probability simplex
constraints. For convenience, let us define the notation
\begin{equation*}
\delta_\Omega(\u)
\coloneqq (\Omega + \iota_{\RR_+^M})^*(\u)
= \max_{\v \in \RR^M_+} \langle \u, \v \rangle - \Omega(\v). 
\end{equation*}
The following proposition shows that we can reduce 
computing $\max_\Omega$ to solving a root equation involving $\delta_\Omega$.
\begin{boxprop}{Computing $\max_\Omega$ as root finding}
\label{smoothing:prop:reduction_root_finding}

Suppose $\Omega$ is strongly convex.
For all $\u \in \RR^M$,
\begin{align*}
\text{max}_\Omega(\u)    
&= \min_{\tau \in \RR} \tau + \delta_\Omega(\u - \tau \ones) \\
&= \tau^\star + \delta_\Omega(\u - \tau^\star \ones)
\end{align*}
and
\begin{equation*}
\nabla \text{max}_\Omega(\u)    
= \nabla \delta_\Omega(\u - \tau^\star \ones),
\end{equation*}
where $\tau^\star$ is the solution \wrt $\tau$ of the above $\min$, 
which satisfies the root equation
\begin{equation*}
\langle \nabla \delta_\Omega(\u - \tau^\star \ones), \ones \rangle = 1.
\end{equation*}
\end{boxprop}
\begin{proof}
The idea is to keep the non-negativity constraint explicit, but to use a
Lagrange multiplier for the equality constraint of the probability simplex.  We
then have
\begin{align*}
\text{max}_\Omega(\u)
&= \max_{\v \in \triangle^M} \langle \u, \v \rangle - \Omega(\v) \\
&= \max_{\v \in \RR_+^M} \min_{\tau \in \RR} \langle \u, \v \rangle - \Omega(\v) 
- \tau(\langle \v, \ones \rangle - 1) \\
&= \min_{\tau \in \RR} \tau + \max_{\v \in \RR_+^M} 
\langle \u -  \tau \ones, \v \rangle - \Omega(\v) \\
&= \min_{\tau \in \RR} \tau + \delta_\Omega(\u -  \tau \ones),
\end{align*}
where we used that we can swap the $\min$ and the $\max$,
since $(\u,\v) \mapsto \langle \u, \v \rangle - \Omega(\v)$
is convex-concave and $\v \in \triangle^M$ is an affine constraint.
The gradient $\nabla \delta_\Omega(\u)$ follows from Danskin's theorem.
The root equation follows from computing the derivative of
$\tau \mapsto \tau + \delta_\Omega(\u - \tau \ones)$ and setting it to zero.
\end{proof}

\subsection{The softmax}

When $\Omega$ is Shannon's negentropy, we obtain
that $\max_\Omega$ is the softmax,
already briefly discussed in \cref{neural_nets:sec:vector_to_scalar}.
\begin{boxprop}{Analytical expression of the softmax}
\label{smoothing:prop:softmax_analytical_expression}
When $\Omega(\piv) = \langle \piv, \log \piv \rangle$, we get
\begin{align*}
\mathrm{softmax}(\u)
&\coloneqq \mathrm{max}_\Omega(\u) \\
&= \max_{\piv \in \triangle^M} \langle \u, \piv \rangle - \Omega(\piv) \\
&= \mathrm{logsumexp}(\u) \\
&= \log \sum_{j=1}^M e^{u_j}.
\end{align*}
\end{boxprop}
\begin{proof}
Since $\dom(\Omega) = \RR_+^M$, 
we have $\delta_\Omega = \Omega^*$ (i.e., the non-negativity constraint is
redundant).
From \cref{duality:ex:conjugates},
we therefore have 
$\delta_\Omega(\u) = \sum_{j=1}^M \exp(u_j - 1)$.
From \cref{smoothing:prop:reduction_root_finding},
$\max_\Omega(\u) = \tau^\star + \delta_\Omega(\u - \tau^\star \ones)$
where $\tau^\star$ satisfies
$\langle \nabla \delta_\Omega(\u - \tau^\star \ones), \ones \rangle = 1.$
Since $\nabla \delta_\Omega(\u) = \exp(\u - \ones)$,
we need to solve $\sum_{j=1}^M \exp(u_j - 1 - \tau) = 1$.
We therefore get
$\tau^\star + 1 = \lse(\u)$
and therefore
$\max_\Omega(\u) 
= \lse(\u) - 1 + \sum_{j=1}^M\exp(u_j - \lse(\u))
= \lse(\u)$.
\end{proof}
Since $-\log M \le \Omega(\piv) \le 0$ for all $\piv \in \triangle^M$,
following \cref{smoothing:prop:properties_max_Omega}, we get
for all $\u \in \RR^M$
\begin{equation*}
\max(\u) \le \mathrm{softmax}(\u) \le \max(\u) + \log M.
\end{equation*}
A unique property of the softmax,
which is not the case of all $\max_\Omega$ operators,
is that it supports \textbf{associativity}.
\begin{boxprop}{Associativity of the softmax}

For all $a, b, c \in \RR$,
\begin{equation*}
\text{softmax}(\text{softmax}(a, b), c) 
=
\text{softmax}(a, \text{softmax}(b, c)).
\end{equation*}
\end{boxprop}

\subsection{The sparsemax}

Alternatively, choosing $\Omega$ to be Gini's negentropy
leads to the sparsemax \citep{martins_2016,mensch_2018}.
\begin{boxprop}{Variational formulation of sparsemax}
When $\Omega(\piv) = \frac{1}{2} \langle \piv, \piv - \ones \rangle$,
we have
\begin{align*}
\mathrm{sparsemax}(\u)
&\coloneqq \text{max}_\Omega(\u) \\
&= \max_{\piv \in \triangle^M} \langle \u, \piv \rangle - \Omega(\piv) \\
&= \langle \u, \piv^\star \rangle - \Omega(\piv^\star)
\end{align*}
where
\begin{equation*}
\piv^\star 
= \mathrm{sparseargmax}(\u)
\coloneqq \argmin_{\piv \in \triangle^M} \|\u - \piv\|_2^2.
\end{equation*}
\end{boxprop}
\begin{proof}
This follows from the fact that $\Omega(\piv)$ is up to a constant equal to
$\frac{1}{2}\|\piv\|^2_2$ and completing the square.
\end{proof}
Therefore, computing the sparsemax can use the sparseargmax (the Euclidean
projection onto the probability simplex) as a building block.
We discuss how to compute it in more detail in
\cref{inf_conv:sec:relaxed_argmax}.
Applying \cref{smoothing:prop:reduction_root_finding} gives an alternative
formulation.
\begin{boxprop}{Sparsemax as root finding}
When $\Omega(\piv) = \frac{1}{2} \langle \piv, \piv - \ones \rangle$,
we have
\begin{equation*}
\mathrm{sparsemax}(\u)
= \text{max}_\Omega(\u)
= \min_{\tau \in \RR}
\tau + \frac{1}{2} \sum_{i=1}^M [u_i - \tau]_+^2
\end{equation*}
and $\tau^\star$ satisfies
\begin{equation*}
\sum_{i=1}^M [u_i - \tau]_+ = 1.
\end{equation*}
\end{boxprop}
\begin{proof}
First, we compute the expression of
$\delta_\Omega(\u) = \max_{\v \in \RR_+^M} \langle \u, \v \rangle - \Omega(\v)$.
Setting the gradient of  
$\v \mapsto \langle \u, \v \rangle - \Omega(\v)$ to zero and clipping, we
obtain $\v^\star = [\u]_+$.
Plugging $\v^\star$ back, we obtain $\delta_\Omega(\u) = \frac{1}{2}
\sum_{i=1}^M [u_i]_+^2$.
Using \cref{smoothing:prop:reduction_root_finding} proves the proposition's
first part. Setting the derivative \wrt $\tau$ to zero gives the second part.
\end{proof}
It can be shown \citep{duchi,Condat2016} that the exact solution $\tau^\star$ is
obtained by
\begin{equation}
\tau^\star 
= \frac{1}{j^\star} \left(\sum_{i=1}^{j^\star} u_{[i]} - 1\right),
\label{smoothing:eq:sparsemax_optimal_threshold}
\end{equation}
where $j^\star$ is the largest $j \in [M]$ such that
\begin{equation*}
u_j - \frac{1}{j} \left(\sum_{i=1}^j u_{[i]} - 1\right) > 0,
\end{equation*}
and where we used the notation
$u_{[1]} \ge u_{[2]} \ge \dots \ge u_{[M]}$.
As an alternative, we can also compute $\tau^\star$ approximately using a
bisection or by gradient descent \wrt $\tau$.

For $\piv \in \triangle^M$, since $\tfrac{1}{M} \leq \lVert \piv \rVert_2^2 \leq
1$, we have $-\tfrac{M-1}{2M} \leq \Omega(\piv) \leq 0.$
Following \cref{smoothing:prop:properties_max_Omega}, we therefore get
for all $\u \in \RR^M$
\begin{equation*}
\max(\u) \le \mathrm{sparsemax}(\u) \le \max(\u) + \frac{M-1}{2M}.
\end{equation*}

\subsection{Recovering smoothed ReLU functions}

Using the vector $\u = (u, 0) \in \RR^2$ as input,
the smoothed max operator recovers the smoothed ReLU:
\begin{equation*}
\mathrm{max}_\Omega((u, 0)) = \mathrm{relu}_\Psi(u),
\end{equation*}
where we defined $\Psi(\pi) \coloneqq \Omega((\pi, 1-\pi))$.
With $\Omega$ being Shannon's negentropy, 
we recover $\Psi(\pi) = \pi \log \pi + (1-\pi) \log (1-\pi)$;
with $\Omega$ being Gini's negentropy, we recover
$\Psi(\pi) = \pi(\pi-1)$, that we used to smooth the ReLU.

\section{Relaxed step functions (sigmoids)}
\label{inf_conv:sec:sigmoid}

We now turn to creating continuous relaxations of step functions.
The binary step function, a.k.a. Heaviside step function, is defined by
\begin{equation*}
\mathrm{step}(u)
\coloneqq \begin{cases}
    1 &\mbox{ if } u \ge 0 \\
    0 &\mbox{ otherwise}
\end{cases}.
\end{equation*}
From \cref{smoothing:eq:relu_variational_form}, its variational form is
\begin{equation*}
\mathrm{step}(u) = 
\argmax_{\pi \in [0,1]} u \cdot \pi.
\end{equation*}
We can therefore define the relaxation
\begin{equation*}
\mathrm{step}_\Omega(u) \coloneqq 
\argmax_{\pi \in [0,1]} u \cdot \pi - \Omega(\pi).
\end{equation*}
Notice that, unlike the case of the smoothed ReLU,
it is a regularized argmax, not a regularized max.
Following \cref{smoothing:rem:grad_diff_ae},
strongly convex regularization $\Omega$ ensures that
$\mathrm{step}_\Omega(u)$ is a Lipschitz continuous function of $u$
and is therefore, at least, differentiable almost everywhere,
unlike $\mathrm{step}(u)$.

\subsubsection*{The logistic function}

If we use the regularizer $\Omega(\pi) = \pi \log \pi + (1-\pi) \log (1-\pi)$,
we obtain the closed form
\begin{equation*}
\mathrm{step}_\Omega(u) =
\mathrm{logistic}(u) \coloneqq \frac{1}{1 + e^{-u}} = \frac{e^u}{1 + e^u}.
\end{equation*}
This function is differentiable everywhere.

\subsubsection*{The sparse sigmoid}

As an alternative, 
if we use $\Omega(\pi) = \pi(\pi - 1)$,
we obtain a piecewise linear sigmoid,
\begin{equation*}
\mathrm{step}_\Omega(u) =
\mathrm{sparsesigmoid}(u) \coloneqq
\begin{cases}
0, & u \le -1\\
\frac{1}{2}(u+1), & -1 < u < 1 \\  
1, & u \ge 1
\end{cases}.
\end{equation*}
Unlike the logistic function, it can reach the exact values $0$ or $1$.
However, the function has two kinks, where the function is non-differentiable.

\subsubsection*{Link between smoothed ReLU functions and sigmoids}

\begin{figure}[t]
  \begin{center}
    \includegraphics[width=\linewidth]{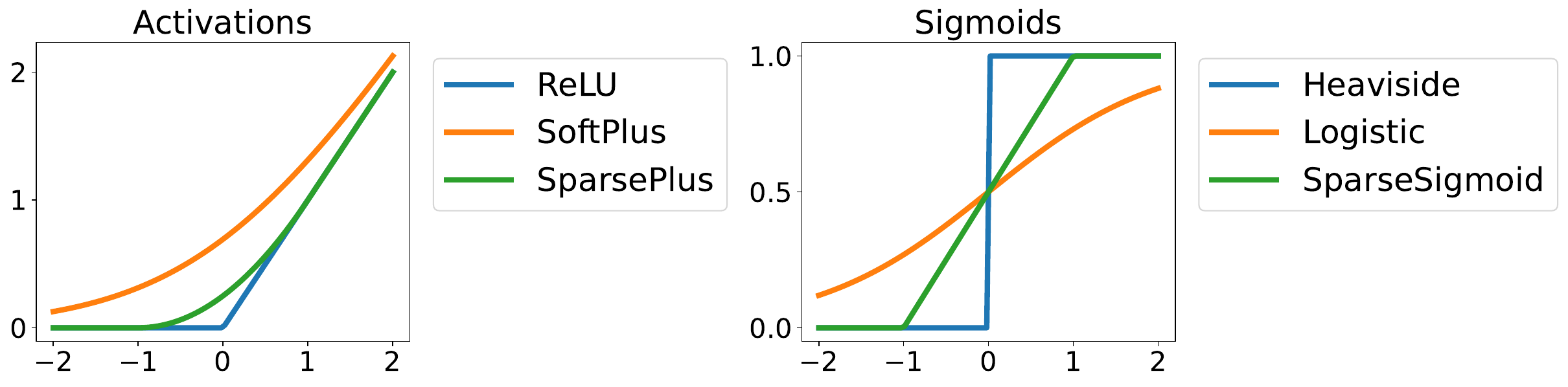}
  \end{center}
\caption{Smoothed ReLU functions and relaxed step functions (sigmoids). Differentiating the left functions
  gives the right functions.
  \label{smoothing:fig:activ_sig}}
\end{figure}

It turns out that the three sigmoids we presented above (step, logistic,
sparsesigmoid) are all equal to the derivative of their corresponding
smoothed ReLU function:
\begin{align*}
    \mathrm{step}(u) &= \mathrm{relu}'(u) \\
    \mathrm{logistic}(u) &= \mathrm{softplus}'(u) \\
    \mathrm{sparsesigmoid}(u) &= \mathrm{sparseplus}'(u)
\end{align*}
and more generally
\begin{equation*}
\mathrm{relu}_\Omega'(u) = \mathrm{step}_\Omega(u).
\end{equation*}
This is a consequence of Danskin's theorem;
see \cref{diff_opt:exm:conjugate_danskin}.
We illustrate the smoothed ReLU functions and relaxed step functions (sigmoids)
in \cref{smoothing:fig:activ_sig}.

\section{Relaxed argmax operators}
\label{inf_conv:sec:relaxed_argmax}

We now turn to argmax operators, which are a generalization of step functions.
With a slight notation overloading, let us now define
\begin{equation*}
\mathrm{argmax}(\u) 
\coloneqq \phi(\argmax_{j \in [M]} u_j),
\end{equation*}
where $\phi(j) = \mathrm{onehot}(j) = \e_j$ is used to embed any integer $j \in
[M]$ into $\RR^M$.
Following the previous discussion, we have the variational form
\begin{equation*}
\mathrm{argmax}(\u) = 
\argmax_{\piv \in \triangle^M} \langle \u, \piv \rangle
= \argmax_{\piv \in \{\e_1, \dots, \e_M\}} \langle \u, \piv \rangle,
\end{equation*}
where the second equality uses that a linear function is maximized at one of the
vertices of the simplex. This variational form suggests to define the relaxation
\begin{equation*}
\mathrm{argmax}_\Omega(\u) \coloneqq
\argmax_{\piv \in \triangle^M} \langle \u, \piv \rangle - \Omega(\piv).
\end{equation*}
Again, following \cref{smoothing:rem:grad_diff_ae},
$\mathrm{argmax}_\Omega(\u)$ is guaranteed to be, at least,
a differentiable almost everywhere function of $\u$ if $\Omega$ is strongly
convex.

Similarly to sigmoids, it turns out that these mappings are equal 
to the gradient of their corresponding smoothed max operator:
\begin{equation*}
\mathrm{argmax}_\Omega(\u) = \nabla \mathrm{max}_\Omega(\u).
\end{equation*}
This is again a consequence of Danskin's theorem.

\subsubsection*{The softargmax}

When using Shannon's negative entropy
$\Omega(\piv) = \langle \piv, \log \piv \rangle$, 
we obtain
\begin{equation*}
\mathrm{argmax}_\Omega(\u) = 
\mathrm{softargmax}(\u) =
\frac{\exp(\u)}{\sum_{j=1}^M \exp(u_j)},
\end{equation*}
which is differentiable everywhere.
\begin{proof}
We know that $\max_\Omega(\u) = \lse(\u)$ and that $\nabla \max_\Omega(\u) =
\mathrm{argmax}_\Omega(\u)$.
Differentiating $\lse(\u)$ gives $\mathrm{softargmax}(\u)$.
\end{proof}

\subsubsection*{The sparseargmax}

\begin{figure}
\begin{center}
\includegraphics[width=\linewidth]{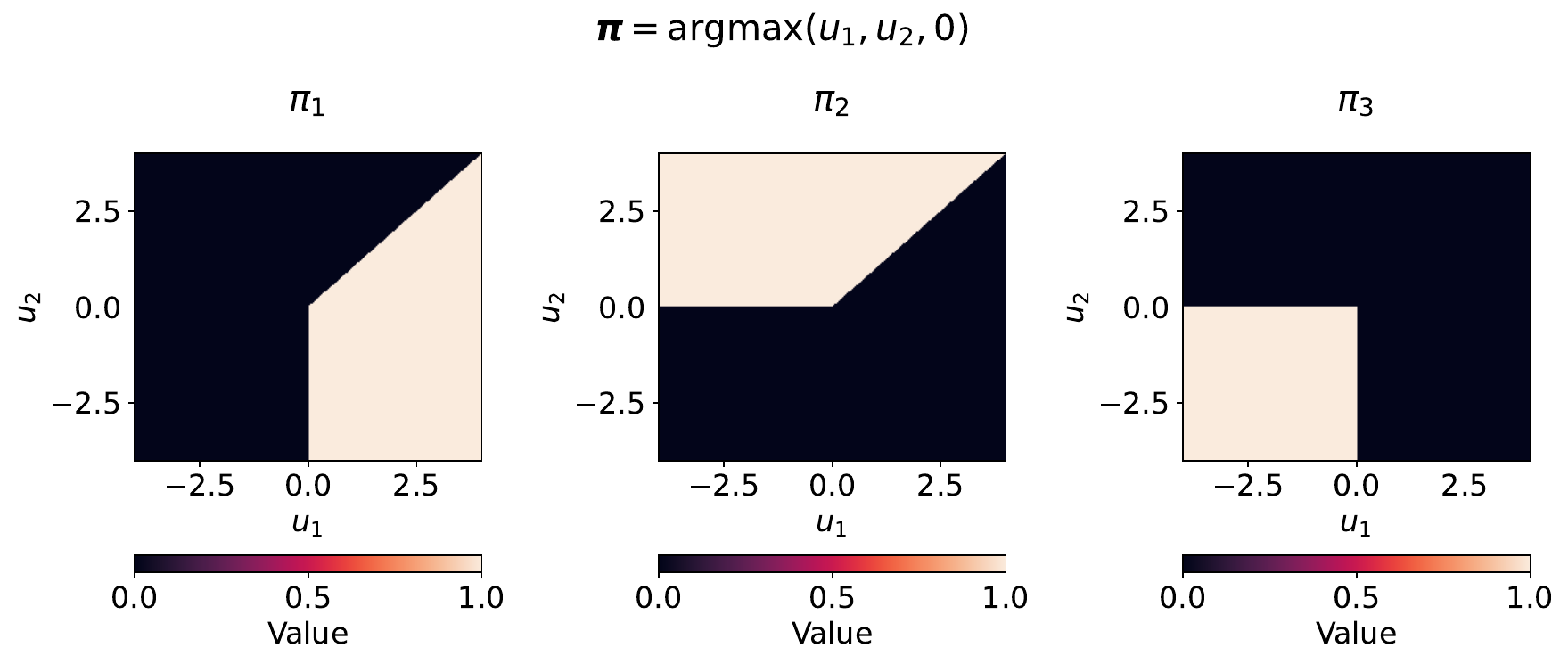}
\includegraphics[width=\linewidth]{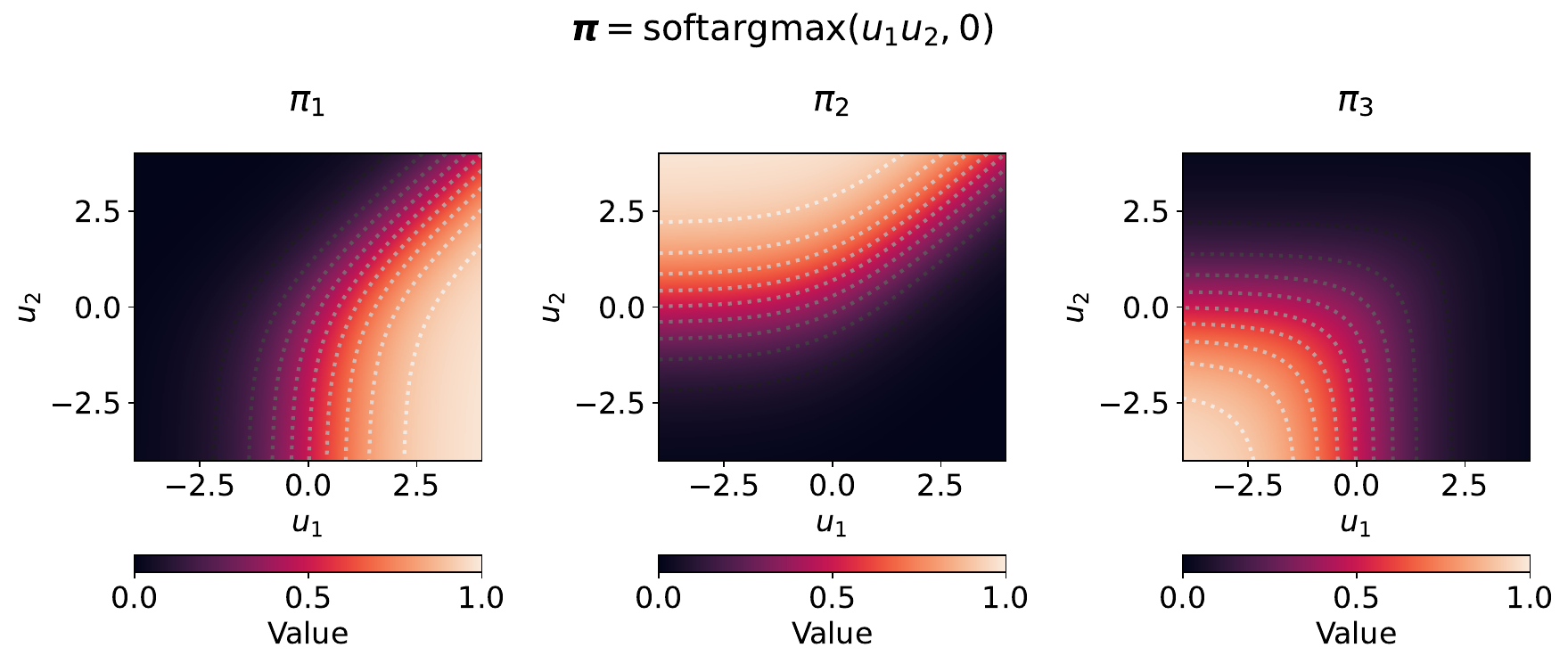}
\includegraphics[width=\linewidth]{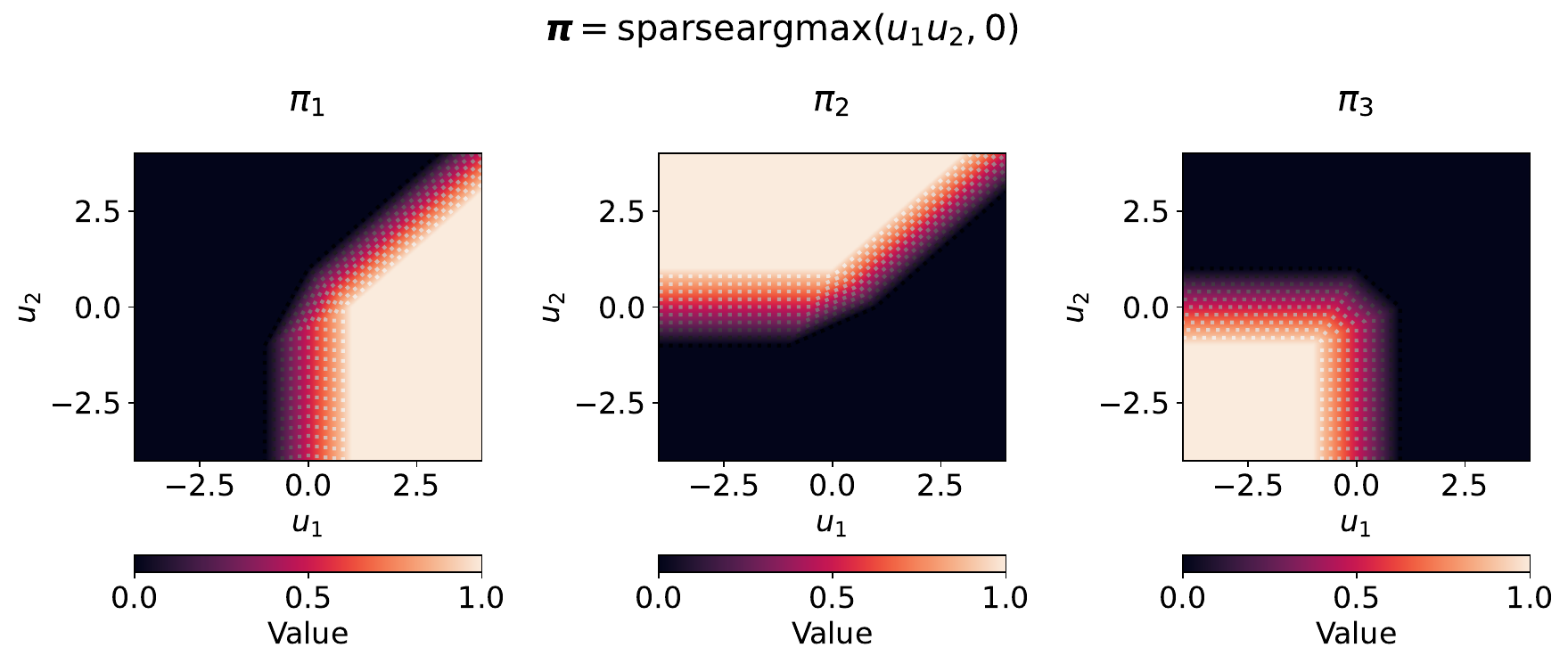}
\end{center}
\caption{
Values of $\mathrm{argmax}(\u)$, $\mathrm{softargmax}(\u)$,
and $\mathrm{sparseargmax}(\u)$ for $\u = (u_1, u_2, 0)$, when varying $u_1$
and $u_2$.
The argmax is a piecewise constant, discontinuous function.
The softargmax is a continuous and differentiable everywhere function, but it
is always strictly positive and therefore dense.
The sparseargmax is a continuous function and its output can be sparse,
but it is only a differentiable almost everywhere function.
\label{smoothing:fig:argmax_heatmap}}
\end{figure}

\begin{figure}
\begin{center}
\includegraphics[width=0.32\linewidth]{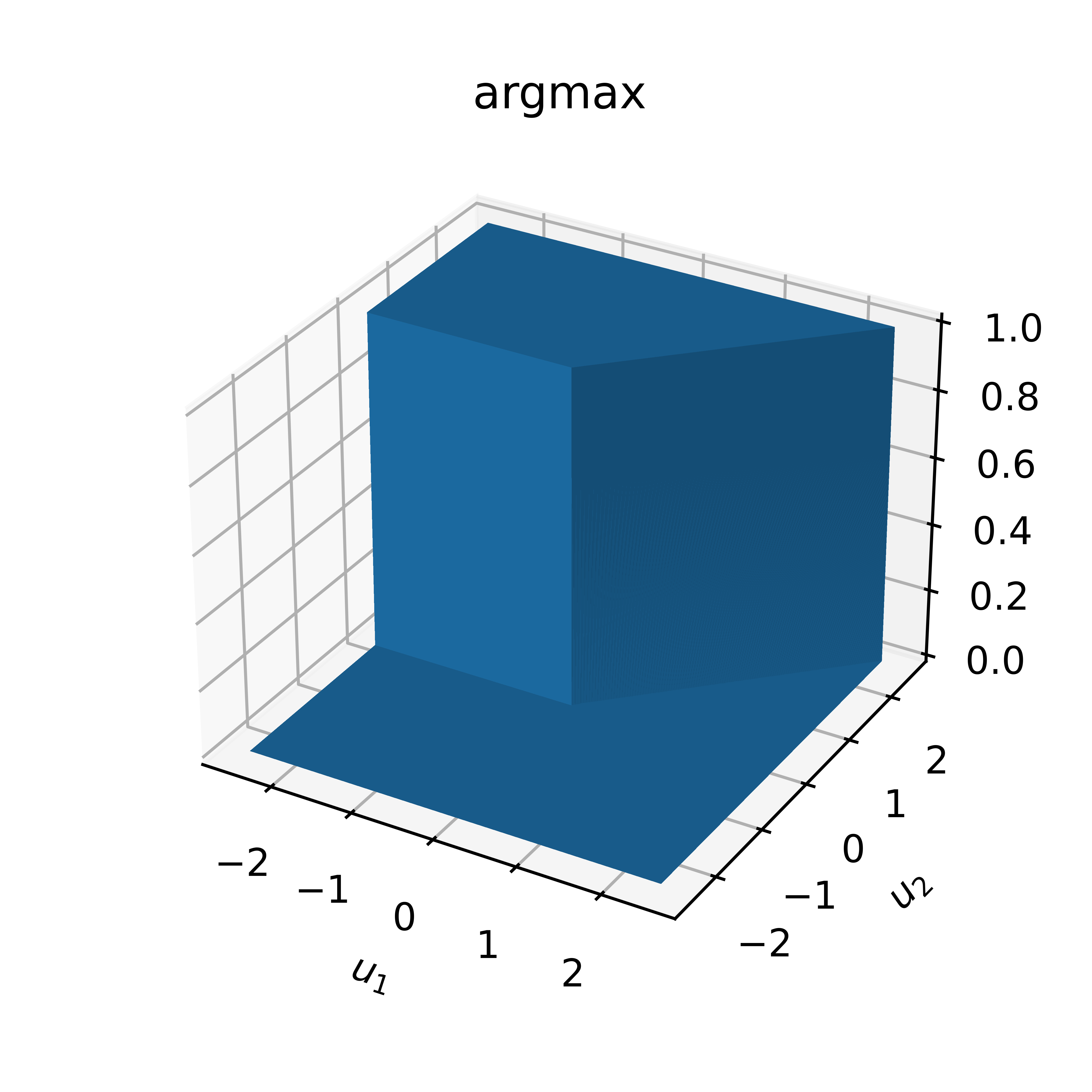}
\includegraphics[width=0.32\linewidth]{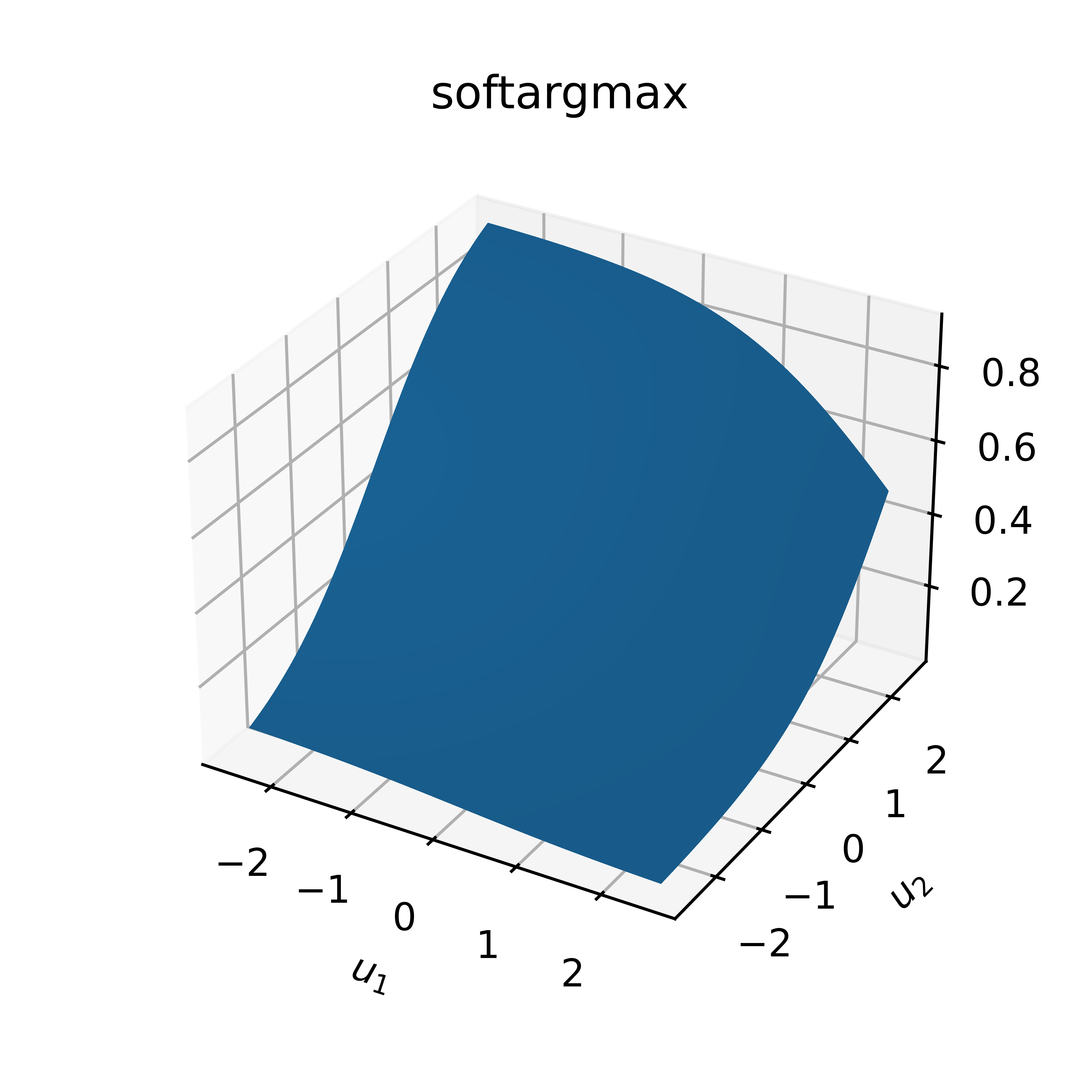}
\includegraphics[width=0.32\linewidth]{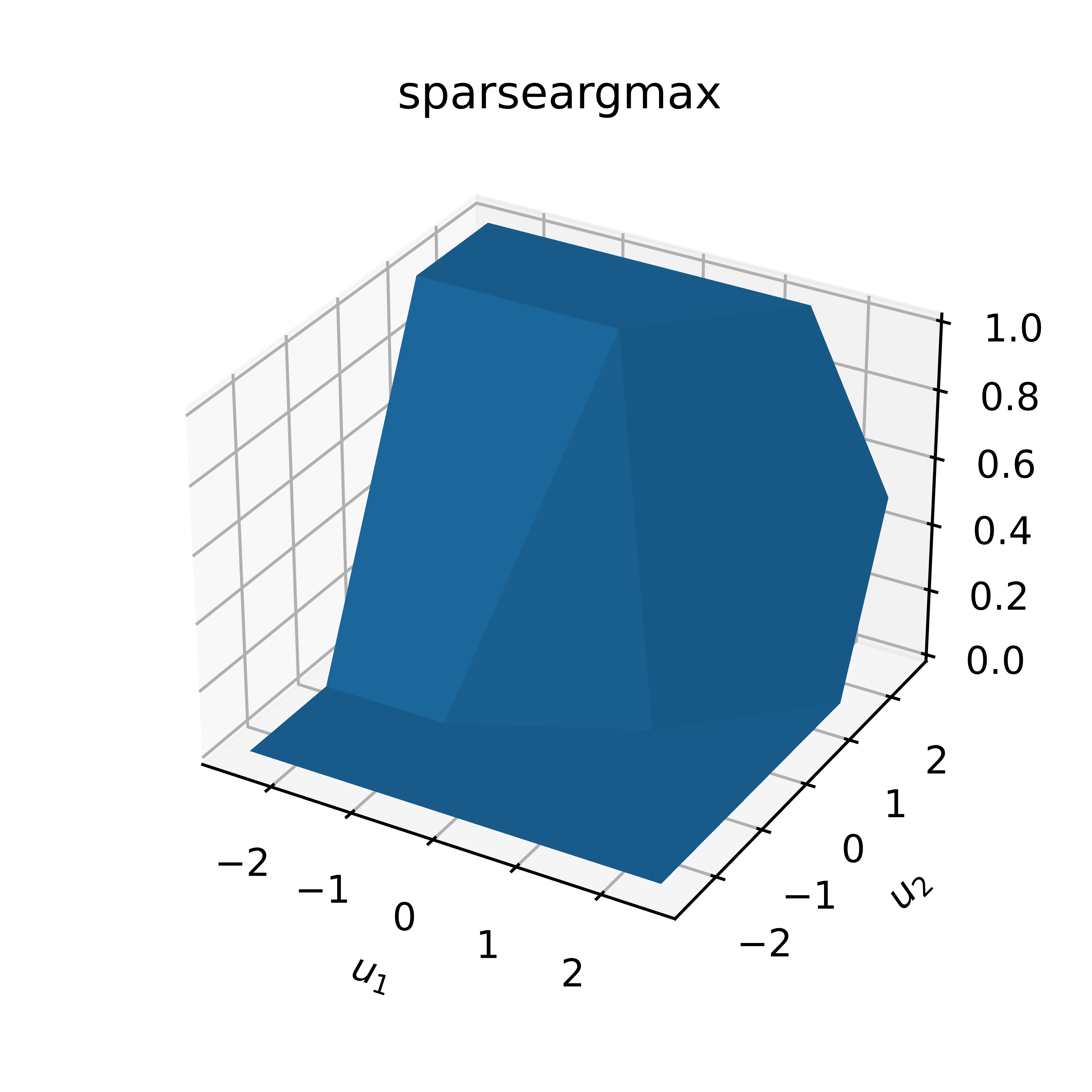}
\end{center}
\caption{
Same as \cref{smoothing:fig:argmax_heatmap} but using a 3D plot.
\label{smoothing:fig:argmax_3d}}
\end{figure}

When using Gini's entropy 
$\Omega(\piv) = \frac{1}{2} \langle \piv, \piv - \ones \rangle$,
which is up to a constant equal to $\frac{1}{2} \|\piv\|^2_2$,
we obtain
the sparseargmax \citep{martins_2016}
\begin{align*}
\mathrm{argmax}_\Omega(\u) 
&= \mathrm{sparseargmax}(\u) \\
&\coloneqq \argmax_{\piv \in \triangle^M} \langle \u, \piv \rangle - 
\frac{1}{2} \langle \piv, \piv - \ones \rangle \\
&= \argmax_{\piv \in \triangle^M} \langle \u, \piv \rangle - 
\frac{1}{2} \|\piv\|_2^2 \\
&=\argmin_{\piv \in \triangle^M} \|\u - \piv\|^2_2,
\end{align*}
which is nothing but the Euclidean projection onto the probability simplex
(see also \cref{optim:sec:proj_grad}).
The Euclidean projection onto the probability simplex $\triangle^M$
can be computed exactly using a median-finding-like algorithm.
The complexity is $O(M)$ expected time and $O(M \log M)$ worst-case time
\citep{Brucker1984,michelot,duchi,Condat2016}.
Computing the Euclidean projection onto the probability simplex
boils down to computing $\tau^\star$ given in
\cref{smoothing:eq:sparsemax_optimal_threshold}.
Once we computed it, we have
\begin{equation*}
\mathrm{sparseargmax}(\u) = [\u - \tau^\star]_+,
\end{equation*}
As its name indicates, and as the above equation shows, sparseargmax is
\textbf{sparse}, but it is only differentiable almost everywhere. Note that the
operator is originally known as sparsemax \citep{martins_2016}, but this is a
misnomer, as it is really an approximation of the argmax. Therefore, in analogy
with the softargmax, we use the name sparseargmax. We compare the argmax,
softmax and sparseargmax in 
\cref{smoothing:fig:argmax_heatmap}
and
\cref{smoothing:fig:argmax_3d}.

\subsubsection*{Relaxed argmin operators}

The argmin operator can be expressed in terms of the argmax operator,
\begin{equation*}
    \argmin(\u) = \argmax(-\u).
\end{equation*}
Given a relaxed argmax operator $\mathrm{argmax}_\Omega$, we can therefore
define a relaxed argmin by
\begin{equation*}
\mathrm{argmin}_\Omega(\u) \coloneqq 
\mathrm{argmax}_\Omega(-\u).
\end{equation*}
We then have for all $\u \in \RR^M$
\begin{equation*}
\mathrm{argmin}_\Omega(\u) = \nabla \text{min}_\Omega(\u).
\end{equation*}

\section{Summary}

\begin{itemize}

\item When a function $f$ is non-differentiable (or worse, discontinuous),
a reasonable approach is to replace it by its smooth approximation (or
continuous relaxation).

\item The first approach we reviewed is infimal convolution between $f$
and primal regularization $R$.
The Moreau envelope is a special case, obtained by using $R = \frac{1}{2}
\|\cdot\|_2^2$.

\item The second approach we reviewed is regularizing the convex conjugate $f^*$ of
$f$ with some dual regularization $\Omega$.
We saw that the primal and dual approaches are equivalent when $R = \Omega^*$.

\item The Legendre-Fenchel transformation, \aka convex conjugate, can be seen as a
dual representation of a function: instead of representing $f$ by its graph
$(\u, f(\u))$ for $\u \in \dom(f)$, we can represent it by the set of tangents
with slope $\v$ and intercept $-f^*(\v)$ for $\v \in \dom(f^*)$. As its name
indicates, it is convex, even if the original function is not. 

\item We showed how to apply smoothing techniques to create smoothed ReLU functions
and smoothed max operators. We also showed that taking their gradients allowed
us to obtain generalized sigmoid functions and argmax operators.

\end{itemize}

%% file: chapters/convolution/convolution.tex
\chapter{Smoothing by integration}
\label{chap:conv}

In this chapter, we review smoothing techniques based on \textbf{convolution}.

\section{Convolution}

\subsection{Convolution operators}

The convolution between two functions $f$ and $g$ produces another function,
denoted $f \ast g$. It is defined by
\begin{equation}
(f \ast g)(\mu) \coloneqq \int_{-\infty}^\infty f(u) g(\mu - u) \, du,
\label{conv:eq:conv_def}
\end{equation}
assuming that the integral is well-defined.
It is therefore the integral of the product of $f$ and $g$ after
$g$ is reflected about the $y$-axis and shifted. It can be seen as a
generalization of the \textbf{moving average}. Using the change of variable
$u \coloneqq \mu + z$, we can also write
\begin{equation}
\begin{aligned}
(f \ast g)(\mu) 
&= \int_{-\infty}^\infty f(\mu + z) g(-z) \, dz \\
&= \int_{\infty}^{-\infty} f(\mu - v) g(v) \, (-dv) \\
&= \int_{-\infty}^\infty g(v) f(\mu - v) \, dv \\
&= (g \ast f)(\mu).
\end{aligned}
\label{conv:eq:conv_def2}
\end{equation}
The convolution operator is therefore \textbf{commutative}.

\begin{boxrem}{Change of variables}
Readers with a background in signal processing might find the change of
variables $u \coloneqq \mu + z$ slightly unconventional for proving the
commutativity of the convolution, as $u \coloneqq \mu - z$ more directly proves
commutativity. However, we intentionally use
$u \coloneqq \mu + z$ here, as this perfectly aligns our notation with 
the \textbf{location-scale transform} $U \coloneqq \muv + \sigma Z$ 
from \cref{grad_est:sec:location_scale_transform}. 
It allows us to seamlessly interpret smoothing by convolution as the injection
of additive noise.
\end{boxrem}

\subsection{Convolution with a kernel}

The convolution is frequently used together with a {\bf kernel} $\kappa$ to create a
smooth approximation $f \ast \kappa$ of $f$.  The most frequently used kernel is the
{\bf Gaussian kernel} with width $\sigma$, defined by
\begin{equation*}
\kappa_\sigma(z) 
\coloneqq \frac{1}{\sqrt{2\pi}\sigma} e^{-\frac{1}{2}(\frac{z}{\sigma})^2}.
\end{equation*}
This is the probability density function (PDF) of the normal distribution with
zero mean and variance $\sigma^2$.
The term $\frac{1}{\sqrt{2\pi}\sigma}$ is a normalization constant, ensuring
that the kernel sums to $1$ for all $\sigma$. We therefore say that $\kappa_\sigma$
is a \textbf{normalized kernel}.

\subsubsection*{Averaging perspective}

Applying the definition of the convolution in \cref{conv:eq:conv_def}, 
we obtain
\begin{align*}
(f \ast \kappa_\sigma)(\mu)
&\coloneqq 
\frac{1}{\sqrt{2\pi}\sigma}
\int_{-\infty}^\infty f(u) e^{-\frac{1}{2} (\frac{\mu -
u}{\sigma})^2} du \\
&= \EE_{U \sim p_{\mu,\sigma}}[f(U)],
\end{align*}
where 
\begin{equation*}
p_{\mu,\sigma}(u) 
\coloneqq \kappa_\sigma(\mu - u)
= \frac{1}{\sqrt{2\pi}\sigma} e^{-\frac{1}{2} (\frac{\mu - u}{\sigma})^2}
\end{equation*}
is the PDF of the Gaussian distribution with mean
$\mu$ and variance $\sigma^2$.
Therefore, we can see $f \ast \kappa_\sigma$ as the \textbf{expectation} of
$f(u)$ over a Gaussian \textbf{centered around} $\mu$.
This property is true for all translation-invariant kernels, 
that correspond to a location-scale family distribution (e.g., the Laplace
distribution). 
The convolution therefore performs an averaging with all points, with
points nearby $\mu$ given more weight by the distribution.
The parameter $\sigma$ controls the importance we want to give to farther
points. We call this viewpoint averaging, as we replace $f(u)$ by
$\EE[f(U)]$.

\subsubsection*{Perturbation perspective}

Conversely, using the alternative definition of the convolution operator in
\cref{conv:eq:conv_def2}, which stems from the commutativity of the convolution, 
we have
\begin{align*}
(f \ast \kappa_\sigma)(\mu)
&\coloneqq \int_{-\infty}^\infty f(\mu + z)
e^{-\frac{1}{2} (\frac{-z}{\sigma})^2} dz \\
&= \int_{-\infty}^\infty f(\mu + z)
e^{-\frac{1}{2} (\frac{z}{\sigma})^2} dz \\
&= \EE_{Z \sim p_{0,\sigma}}[f(\mu + Z)],
\end{align*}
where, in the second line, we used that the Gaussian kernel $p_{0,\sigma}$ is sign invariant,
i.e., $p_{0,\sigma}(-z) = p_{0,\sigma}(z)$.
This viewpoint shows that smoothing by convolution with a Gaussian kernel can
also be seen as injecting Gaussian \textbf{noise} or \textbf{perturbations} to 
the function's input.

\subsubsection*{Limit case}

When $\sigma \to 0$, the kernel $\kappa_\sigma$ converges to a Dirac delta function,
\begin{equation*}
\lim_{\sigma \to 0} \kappa_\sigma(z) = \delta(z).
\end{equation*}
Since the Dirac delta is the multiplicative identity of the
convolution algebra (this is also known as the sifting property),
when $\sigma \to 0$, $f \ast \kappa_\sigma$ converges to $f$, i.e.,
\begin{equation*}
\lim_{\sigma \to 0} (f \ast \kappa_\sigma)(u) = f(u).
\end{equation*}

\subsection{Discrete convolution}
\label{conv:sec:discrete_conv}

Many times,
we work with functions whose convolution does not have an analytical form. In
these cases, we can use a discrete convolution on a grid of values. For two
functions $f$ and $g$ defined over $\ZZ$, the discrete convolution is defined by
\begin{equation*}
(f \ast g)[i] \coloneqq \sum_{j=-\infty}^\infty f[j] g[i - j].
\end{equation*}
As for its continuous counterpart, the discrete convolution is commutative,
namely,
\begin{equation*}
(f \ast g)[i] = \sum_{j=-\infty}^\infty f[i - j] g[j] = (g \ast f)[i].
\end{equation*}
When $g$ has finite support over the set 
$S \coloneqq \{-M, -M+1, \dots, 0, \dots, M-1, M\}$,
meaning that $g[i] = 0$ for all $i \not \in S$,
a finite summation may be used instead, i.e.,
\begin{equation*}
(f \ast g)[i] = \sum_{j=-M}^M f[i - j] g[j] = (g \ast f)[i].
\end{equation*}

\begin{figure}[t]
\centering
\includegraphics[scale=0.5]{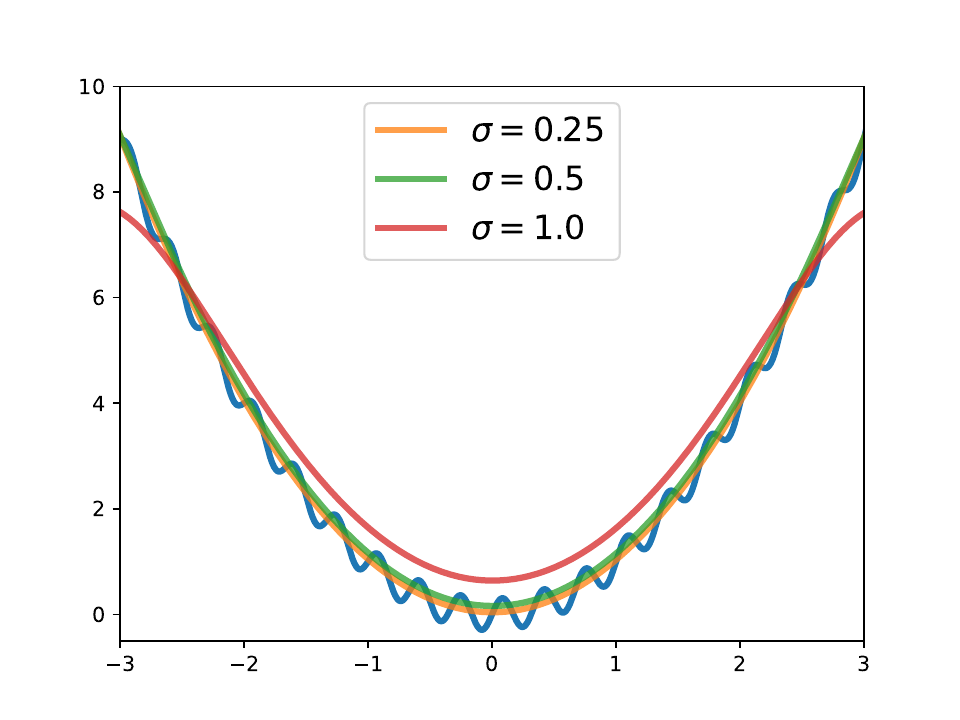}
\caption{Smoothing of the signal
$f[t] \coloneqq t^2 + 0.3 \sin(6\pi t)$
with a sampled and renormalized Gaussian kernel.}
\label{conv:fig:denoising}
\end{figure}

In practice, convolution between a discrete signal $f \colon \ZZ \to \RR$ 
and a continuous kernel $\kappa \colon \RR \to \RR$ is implemented by discretizing
the kernel. 
One of the simplest approaches consists in sampling points on an interval,
evaluating the kernel at these points and renormalizing the obtained values,
so that the sampled kernel sums to $1$. 
This is illustrated with the Gaussian kernel in \cref{conv:fig:denoising}. 
Since the Gaussian kernel decays exponentially fast,
we can choose a small interval around $0$.
For a survey of other possible discretizations of the Gaussian kernel, see
\citet{getreuer_2013}.

\subsection{Differentiation}

Remarkably, provided that the two functions are integrable with integrable
derivatives, the derivative of the convolution satisfies
\begin{equation*}
(f \ast g)' = (f' \ast g) = (f \ast g'),
\end{equation*}
which simply stems from switching derivative and integral in the definition of
the convolution. Moreover, we have the following proposition.
\begin{boxprop}{Differentiability of the convolution}
If $g$ is $n$-times differentiable with compact support over $\RR$
and $f$ is locally integrable over $\RR$, then $f \ast g$ is $n$-times
differentiable over $\RR$.
\end{boxprop}

\subsection{Multidimensional convolution}

So far, we studied the convolution of one-dimensional functions.
The definition can be naturally extended to multidimensional functions $f \colon
\RR^M \to \RR$ and $g \colon \RR^M \to \RR$ as
\begin{equation*}
(f \ast g)(\muv) \coloneqq \int_{\RR^M} f(\u) g(\muv - \u) \, d\u,
\end{equation*}
assuming again that the integral exists.
Typically, a Gaussian kernel with diagonal covariance matrix is used
\begin{align}
\kappa_\sigma(\z) 
&\coloneqq 
\prod_{j=1}^M
\frac{1}{\sqrt{2\pi}\sigma_j}
e^{-\frac{1}{2}(\frac{z_j}{\sigma_j})^2} = \frac{1}{\sqrt{2\pi}^M\sigma^M}
e^{-\frac{1}{2}\frac{\|\z\|^2_2}{\sigma^2}},
\label{conv:eq:gaussian_kernel_diag_cov}
\end{align}
where, in the second equality, we assumed $\sigma_1 = \dots = \sigma_M$.  In an
image processing context, where $M=2$, it is approximated using a discrete
convolution and it is called a \textbf{Gaussian blur}.

\subsection{Link between convolution and infimal convolution}
\label{conv:sec:link_conv_inf_conv}

The infimal convolution we studied in \cref{smoothing:sec:primal} takes the form
\begin{equation*}
(f \square g)(\muv) \coloneqq \inf_{\u \in \RR^M} f(\u) + g(\muv - \u).
\end{equation*}
In comparison, the classical convolution takes the form
\begin{equation*}
(F \ast G)(\muv) \coloneqq \int_{\RR^M} F(\u) G(\muv - \u) \, d\u.
\end{equation*}
The two forms of convolution are clearly related.
Infimal convolution performs an infimum and uses the sum of $f$ and $g$:
it uses a \textbf{min-plus algebra}.
Classical convolution performs an integral and uses the product of $F$ and $G$:
it uses a \textbf{sum-product algebra}.

\subsection{The soft infimal convolution}

The link between the infimal convolution and the classical convolution
can be further elucidated 
if we replace the infimum with a soft minimum in the definition of the
infimal convolution.
\begin{boxdef}{Soft infimal convolution}
The soft infimal convolution between $f \colon \RR^M \to \RR$ and $g \colon
\RR^M \to \RR$ is
\begin{equation*}
(f \square_\varepsilon g)(\muv) \coloneqq
\underset{\u \in \RR^M}{\mathrm{softmin}_\varepsilon} ~
f(\u) + g(\muv - \u),
\end{equation*}
where we defined the soft minimum (assuming that it exists) 
over $\cS$ of any function $h \colon \cS \to \RR$ as
\begin{equation*}
\underset{\u \in \cS}{\mathrm{softmin}_\varepsilon} ~ h(\u)
\coloneqq -\varepsilon \log \int_{\cS} 
\exp\left(-h(\u) / \varepsilon \right) d\u.
\end{equation*}
\end{boxdef}
We recover the infimal convolution as $\varepsilon \to 0$.

\subsubsection*{Computation using a convolution}

We now show that we can rewrite the soft infimal convolution using a
classical convolution.
Indeed, by using the exponential change of variable
(sometimes referred to as \textbf{Cole-Hopf transformation} 
in a partial differential equation context)
\begin{align*}
\cC_\varepsilon\{f\}(\u) &\coloneqq  \exp(-f(\u) / \varepsilon) \\
\cC_\varepsilon^{-1}\{F\}(\v) &= -\varepsilon \log F(\v),
\end{align*}
we can define each function in the exponential domain,
\begin{align*}
F_\varepsilon &\coloneqq \cC_\varepsilon\{f\} \\
G_\varepsilon &\coloneqq \cC_\varepsilon\{g\} \\
H_\varepsilon &\coloneqq \cC_\varepsilon\{h_\varepsilon\}.
\end{align*}
It is easy to check that we then have
\begin{equation*}
H_\varepsilon(\muv) = (F_\varepsilon \ast G_\varepsilon)(\muv).
\end{equation*}
Back to log domain, we obtain
\begin{equation*}
h_\varepsilon(\muv) = \cC_\varepsilon^{-1}\{H_\varepsilon\}(\muv). 
\end{equation*}
Combining the transformation and its inverse, we can write
\begin{align*}
h_\varepsilon(\muv) 
=  \cC_\varepsilon^{-1}\{\cC_\varepsilon\{f\} \ast \cC_\varepsilon\{g\}\}(\muv).
\end{align*}
What we have shown is that,
after an exponential change of variable, 
the soft infimal convolution can be reduced to the computation of a convolution.
This is useful as a discrete convolution on a grid of size $n$ can be computed
in $O(n \log n)$.

\subsection{The soft Moreau envelope}

We saw in \cref{smoothing:sec:moreau_env}
that the infimal convolution between $f$ and 
$R(z) = \frac{1}{2} z^2$ is the Moreau envelope,
\begin{equation*}
M_f(\muv) 
\coloneqq (f \square R)(\muv)
= \inf_{\u \in \RR^M} f(\u) + \frac{1}{2} \|\muv - \u\|^2_2.
\end{equation*}
Replacing the infimal convolution with a soft infimal convolution, we can
define the ``soft'' Moreau envelope,
\begin{equation*}
M_f^\varepsilon(\muv) 
\coloneqq (f \square_\varepsilon R)(\muv)
= \underset{\u \in \RR^M}{\mathrm{softmin}_\varepsilon} ~ f(\u) + \frac{1}{2}
\|\muv - \u\|^2_2.
\end{equation*}
We emphasize that this operation is \textbf{not} the same as the convolution
of $f$ with a Gaussian kernel.
Indeed, we have
\begin{align*}
M_f^\varepsilon(\muv) =
-\varepsilon \log \int_{\RR^M} 
\exp\left((-f(\u) - \frac{1}{2} \|\muv - \u\|_2^2) / \varepsilon \right) d\u.
\end{align*}
while
\begin{align*}
(f \ast \kappa_\sigma)(\muv) \coloneqq \int_{\RR^M} 
f(\u) \kappa_\sigma(\muv - \u) \, d\u,
\end{align*}
where $\kappa_\sigma$ is for instance defined in 
\cref{conv:eq:gaussian_kernel_diag_cov}.

We saw that the Moreau envelope is a smooth function. 
One may therefore ask what do we gain from using a soft Moreau envelope.
The benefit can be computational, as the latter can be approximated using a
discrete convolution.

\section{Fourier and Laplace transforms}

Let us define the \textbf{Fourier transform} of $f$ by
\begin{equation*}
F(s) 
\coloneqq \cF\{f\}(s) 
\coloneqq \int_{-\infty}^{\infty}f(t) e^{-i 2 \pi s t} \, dt, \quad s \in \RR.
\end{equation*}
Note that $\cF\{f\}$ is a function transformation: it transforms $f$ into
another function $F$. 

\subsection{Convolution theorem}

Now, consider the convolution
\begin{equation*}
h(t) \coloneqq (f \ast g)(t).
\end{equation*}
If we define the three transformations
\begin{align*}
    F &\coloneqq \cF\{f\}, \
    G \coloneqq \cF\{g\}, \
    H \coloneqq \cF\{h\},
\end{align*}
the \textbf{convolution theorem} states that
\begin{equation*}
H(s) = F(s) \cdot G(s), \quad s \in \RR.
\end{equation*}
Written differently, we have
\begin{equation*}
\cF\{f \ast g\} = \cF\{f\} \cdot \cF\{g\}.
\end{equation*}
In words, in the Fourier domain, the convolution operation becomes a
multiplication.
Conversely,
\begin{equation*}
h(t) = (f*g)(t) = \cF^{-1}\{F \cdot G\}(t), \quad t \in \RR.
\end{equation*}
The convolution theorem also holds if we replace the Fourier transform 
with the Laplace transform or with the two-sided (bilateral) Laplace transform.

\subsection{Link between Fourier and Legendre transforms}

In \cref{duality:sec:conjugates}, we studied another function transformation:
the convex conjugate, also known as
Legendre-Fenchel transform. 
We recap the analogies between these transforms
in \cref{conv:tab:analogy_fourier_legendre}.
In particular, the counterpart of
\begin{equation*}
\cF\{f \ast g\} = \cF\{f\} \cdot \cF\{g\}
\end{equation*}
for the infimal convolution is
\begin{equation*}
(f \square g)^* = f^* + g^*.
\end{equation*}
In words, the Legendre-Fenchel transform is to the infimal convolution what the
Fourier transform is to the convolution.

\begin{table}[t]
\caption{Analogy between Fourier and Legendre transforms.
See \cref{duality:prop:conjugate_calculus} for more conjugate calculus rules.}
\begin{center}
\begin{tabular}{lcc}
\toprule
& Fourier $\cF\{f\}$ & Legendre $f^*$  \\
\midrule
Semiring & $(+, \cdot)$ & $(\min, +)$ \\
Scaling $(a > 0)$ & $f(t) = g(t / a)$ & $f(t) = a g(t/a)$\\
                  & $\cF\{f\}(s) = a\cF\{g\}(as)$ & $f^*(s) = a g^*(s)$ \\
Translation & $f(t) = g(t - t_0)$ & $f(t) = g(t - t_0)$ \\
            & $\cF\{f\}(s) = e^{-i 2 \pi t_0 s} \cF\{g\}(s)$ & 
            $f^*(s) = g^*(s) + t_0$ \\
Convolution & $h = f \ast g$ & $h = f \square g$ \\
            & $\cF\{h\} = \cF\{f\} \cdot \cF\{g\}$ & $h^* = f^* + g^*$ \\
Gaussian / quadratic & $f(t) = e^{-at^2}$ & $f(t) = \frac{a}{2} t^2$ \\
             & $\cF\{f\}(s) = \sqrt{\frac{\pi}{a}} e^{-\pi^2 s^2/a}$ 
             & $f^*(s) = \frac{1}{2a} s^2$ \\
Smoothing & $f \ast \kappa_\sigma$ & 
$f \square \frac{1}{2\varepsilon} \|\cdot\|^2_2$ \\
\bottomrule
\end{tabular}
\end{center}
\label{conv:tab:analogy_fourier_legendre}
\end{table}

\subsection{The soft Legendre-Fenchel transform}

We saw in \cref{duality:sec:conjugates}
that the Legendre-Fenchel transform (convex conjugate) of a function
$f \colon \RR^M \to \RR$ is 
\begin{equation*}
f^*(\v) 
\coloneqq \max_{\u \in \RR^M} ~ \langle \u, \v \rangle - f(\u).
\end{equation*}
If necessary, we can support constraints by including an indicator function
in the definition of $f$.
The conjugate can be smoothed out using a log-sum-exp,
which plays the role of a soft maximum
(\cref{smoothing:sec:max_argmax}).
\begin{boxdef}{Soft convex conjugate}
\label{conv:def:smoothed_conj}
\begin{equation*}
f^*_\varepsilon(\v)
\coloneqq 
\underset{\u \in \RR^M}{\mathrm{softmax}_\varepsilon} ~
\langle \u, \v \rangle - f(\u),
\end{equation*}
where we defined the soft maximum (assuming that it exists) 
over $\cS$ of any function $g \colon \cS \to \RR$ as
\begin{equation*}
\underset{\u \in \cS}{\mathrm{softmax}_\varepsilon} ~ g(\u)
\coloneqq \varepsilon \log \int_{\cS} 
\exp\left(g(\u) / \varepsilon \right) d\u.
\end{equation*}
\end{boxdef}
In the limit $\varepsilon \to 0$, we recover the convex conjugate.

\subsubsection*{Computation using a convolution}

We now show that this smoothed conjugate can be rewritten
using a convolution if we apply a bijective transformation to $f$.
\begin{boxprop}{Smoothed convex conjugate as convolution}
The smoothed conjugate can be rewritten as
\begin{equation*}
f^*_\varepsilon(\v)
= \cQ_\varepsilon^{-1}\left\{\frac{1}{\cQ_\varepsilon\{f\} \ast
G_\varepsilon}\right\}\left(\v\right)
\end{equation*}
where
\begin{align*}
G_\varepsilon
&\coloneqq 
\cC_\varepsilon\left\{\frac{1}{2} \|\cdot\|^2_2\right\}
= \exp\left(-\frac{1}{2} \frac{\|\cdot\|^2_2}{\varepsilon}\right) \\
\cQ_\varepsilon\{f\}
&\coloneqq 
\cC_\varepsilon\left\{f(\cdot) - \frac{1}{2} \|\cdot\|^2_2\right\} =
\exp\left(\frac{1}{2\varepsilon} \|\cdot\|^2_2 -
\frac{1}{\varepsilon}f(\cdot)\right) \\
\cQ_\varepsilon^{-1}\{F\}
&\coloneqq \frac{1}{2} \|\cdot\|^2_2 - \varepsilon \log(F(\cdot)).
\end{align*}
\end{boxprop}
This insight was
\href{https://twitter.com/gabrielpeyre/status/1253186839776768000}{tweeted}
by Gabriel Peyr\'{e} in April 2020.
\begin{proof}
\begin{equation*}
\begin{small}
\begin{aligned}
f^\ast_\varepsilon(\v) &\coloneqq 
\varepsilon \log \int \exp\left(\frac{1}{\varepsilon} \langle \u, \v \rangle -
\frac{1}{\varepsilon} f(\u)\right) d\u \\
&= \varepsilon \log \int \exp\left(-\frac{1}{2\varepsilon} \|\u - \v\|^2_2 +
    \frac{1}{2\varepsilon} \|\u\|^2_2 + \frac{1}{2\varepsilon}\|\v\|^2_2 
    - \frac{1}{\varepsilon} f(\u)\right) d\u \\
&= \varepsilon \log \int \exp\left(-\frac{1}{2\varepsilon} \|\u - \v\|^2_2 +
    \frac{1}{2\varepsilon} \|\u\|^2_2 - \frac{1}{\varepsilon} f(\u)\right) d\u
    + \frac{1}{2} \|\v\|^2_2 \\
&= \varepsilon \log \int G_\varepsilon(\v - \u) \cQ_\varepsilon\{f\}(\u) d\u 
+ \frac{1}{2} \|\v\|^2_2 \\
&= \varepsilon \log (\cQ_\varepsilon\{f\} \ast G_\varepsilon)(\v) 
+ \frac{1}{2} \|\v\|^2_2 \\
&=  \frac{1}{2} \|\v\|^2_2 -
\varepsilon \log 
\left(\frac{1}{\cQ_\varepsilon\{f\} \ast G_\varepsilon}\right)(\v) \\
&= \cQ_\varepsilon^{-1}\left\{\frac{1}{\cQ_\varepsilon\{f\} \ast
G_\varepsilon}\right\}\left(\v\right)
\end{aligned}
\end{small}
\end{equation*}
\end{proof}
\begin{figure}[t]
\centering
\includegraphics[scale=0.5]{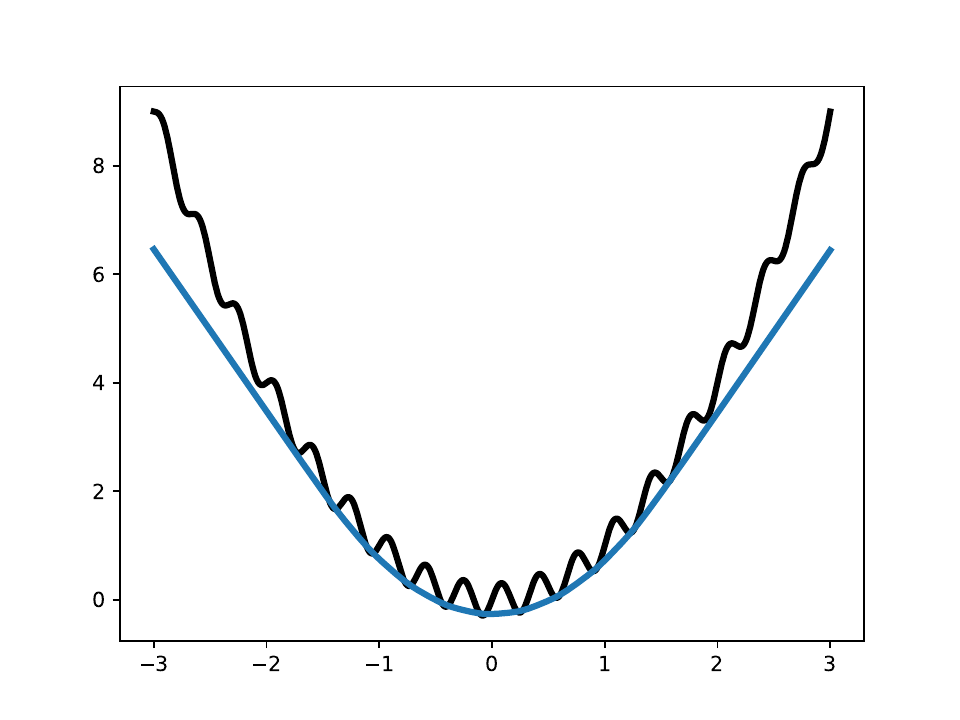}
\caption{Applying the smoothed conjugate twice
gives a smoothed biconjugate (convex envelope) of the function.}
\label{conv:fig:smoothed_biconj}
\end{figure}
What did we gain from this viewpoint?
The convex conjugate can often be difficult to compute in closed form.
If we replace $\RR^M$ with a discrete set $\cS$ (i.e., a grid), we can then
approximate the smoothed convex conjugate in $O(n \log n)$, where $n = |\cS|$,
using a discrete convolution,
\begin{align*}
(\cQ_\varepsilon\{f\} \ast G_\varepsilon)(\v)
&\approx \sum_{\u \in \cS} G_\varepsilon(\v - \u) \cQ_\varepsilon\{f\}(\u) \\
&= \K \q,
\end{align*}
where $\K$ is the $n \times n$ Gaussian kernel matrix whose entries correspond
to $\exp(-\frac{1}{2\varepsilon} \|\u - \u'\|^2_2)$ for $\u,\u' \in \cS$
and $\q$ is the $n$-dimensional vector whose entries correspond to
$\exp(\frac{1}{\varepsilon}(\frac{1}{2} \|\v\|^2_2 - f(\u))$ for $\u \in \cS$.
This provides a GPU-friendly alternative to the fast Legendre transform
algorithm, discussed in \cref{duality:sec:conjugates}. Of course, due to the
curse of dimensionality, the technique is limited to functions defined on
low-dimensional sets. We illustrate in \cref{conv:fig:smoothed_biconj}
the application of the technique to computing an approximate biconjugate (convex
envelope) of a function.

\begin{boxrem}{Link with the two-sided Laplace transform}
For \\
one-dimensional functions, instead of using a convolution,
we can also write the soft convex conjugate as
\begin{align*}
f^*_\varepsilon(v) 
&= \varepsilon \log \int_{-\infty}^\infty \exp\left(\frac{1}{\varepsilon}\left[
uv - f(u)\right])\right) du \\
&= \varepsilon \log
\cB\left\{e^{-\frac{f}{\varepsilon}}\right\}\left(-\frac{v}{\varepsilon}\right)
\\
&= 
-\cC^{-1}_\varepsilon\left\{\cB\left\{\cC_\varepsilon\{f\}\right\}\right\}
\left(-\frac{v}{\varepsilon}\right)
\end{align*}
where we defined the two-sided (bilateral) Laplace transform
\begin{equation*}
\cB\{g\}(v) \coloneqq \int_{-\infty}^\infty e^{-uv} g(u) du
\end{equation*}
and where we assumed that the integral exists.
\end{boxrem}

\section{Examples}

In this section, we review practical examples for which the convolution with a
Gaussian kernel enjoys an analytical solution.

\subsection{Smoothed step function}

\begin{boxexm}{Smoothed Heaviside}
The Heaviside step function is defined by
\begin{equation*}
\mathrm{step}(u) \coloneqq h(u)
\coloneqq \begin{cases}
    1 &\mbox{ if } u \ge 0 \\
    0 &\mbox{ otherwise}
\end{cases}.
\end{equation*}
With the Gaussian kernel, we therefore obtain
\begin{align*}
(h \ast \kappa_\sigma)(\mu)
&= \int_{-\infty}^\mu \kappa_\sigma(z) h(\mu - z) dz
+ \int_{\mu}^\infty \kappa_\sigma(z) h(\mu - z) dz \\
&= \int_{-\infty}^\mu \kappa_\sigma(z) dz \\
&= \Phi_\sigma(\mu) \\
&= \frac{1}{2} \left[1 +
\mathrm{erf}\left(\frac{\mu}{\sqrt{2}\sigma}\right)\right],
\end{align*}
where $\Phi_\sigma(\mu)$ is the CDF of the Gaussian distribution with zero mean
and variance $\sigma^2$, and  where we used the error function
\begin{equation*}
\operatorname{erf}(z) 
\coloneqq \frac{2}{\sqrt\pi}\int_0^z e^{-t^2}\,\mathrm dt,
\end{equation*}
that we both already encountered in \cref{chap:proba_learn}.
Although there is no closed form for the error function, it is commonly
available in numerical analysis software, such as SciPy.
\label{conv:ex:smoothed_heaviside}
\end{boxexm}

\subsection{Smoothed ReLU function}

\begin{boxexm}{Smoothed ReLU}
The ReLU is defined by
\begin{equation*}
r(u)
\coloneqq \begin{cases}
    u &\mbox{ if } u \ge 0 \\
    0 &\mbox{ otherwise}
\end{cases}
= u \cdot h(u).
\end{equation*}
Similarly to the previous example, we obtain
\begin{align*}
(r \ast \kappa_\sigma)(\mu)
&= \int_{-\infty}^\mu \kappa_\sigma(z) r(\mu - z) dz \\
&= \int_{-\infty}^\mu \kappa_\sigma(z) (\mu - z) dz \\
&= \mu \int_{-\infty}^\mu \kappa_\sigma(z) dz - 
\int_{-\infty}^\mu \kappa_\sigma(z) z dz \\
&= \mu \Phi_\sigma(\mu) + \sigma^2 \kappa_\sigma(\mu).
\end{align*}
In the second integral, setting $a \coloneqq \frac{1}{2\sigma^2}$, we used
\begin{equation*}
\int z e^{-a z^2} dz
= -\frac{1}{2a} \int e^t d t
= -\frac{1}{2a} e^t + C 
= -\frac{1}{2a} e^{-a z^2} + C
\end{equation*}
and $t \coloneqq -a z^2 \Rightarrow z dz = -\frac{1}{2a} d t$.
\label{conv:ex:smoothed_relu}
\end{boxexm}

\begin{figure}[t]
\centering
\includegraphics[scale=0.5]{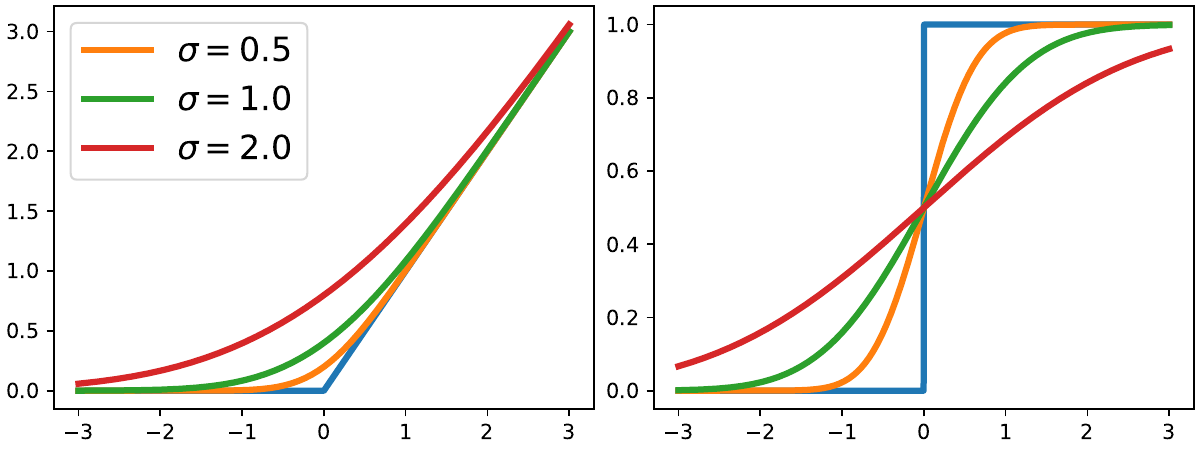}
\caption{Smoothing of the ReLU and Heaviside functions by convolution with a
Gaussian kernel, for three values of the width $\sigma$.}
\label{conv:fig:gaussian_cdf}
\end{figure}

To illustrate differentiation of the convolution, 
we show how to differentiate the smoothed ReLU.
\begin{boxexm}{Differentiating the smoothed ReLU}
Differentiating the smoothed ReLU from \cref{conv:ex:smoothed_relu}, we obtain
\begin{equation*}
(r \ast \kappa_\sigma)' = (r' \ast \kappa_\sigma) = h \ast \kappa_\sigma = \Phi_\sigma.
\end{equation*}
Therefore, unsurprisingly, the derivative of the smoothed ReLU is the smoothed
Heaviside step function. Differentiating once again, we obtain,
\begin{equation*}
(r \ast \kappa_\sigma)'' 
= (h \ast \kappa_\sigma)' 
= (h' \ast \kappa_\sigma) 
= \delta \ast \kappa_\sigma 
= \kappa_\sigma,
\end{equation*}
where the derivative $h'$ is well-defined almost everywhere.
We can arrive at the same result by using
that $h \ast \kappa_\sigma = \Phi_\sigma$ and $\Phi_\sigma' = \kappa_\sigma$, since
$\Phi_\sigma$ and $\kappa_\sigma$ are the CDF and PDF of the Gaussian with zero mean
and $\sigma^2$ variance.
\end{boxexm}

\section{Perturbation of blackbox functions}
\label{perturb:sec:blackbox}

In this section, we review how to approximately compute a convolution with a
kernel and its gradient using Monte-Carlo estimation.

\subsection{Expectation in a location-scale family}

A rather intuitive approach to smooth a function $f: \RR^M\rightarrow \RR$ is
to average its values on an input $\muv$, perturbed by some additive
noise $Z \sim p$, for some noise distribution $p$.
This defines the surrogate
\[
f_\sigma(\muv) \coloneqq \EE_{Z\sim p}[f(\muv + \sigma Z)].
\]
The parameter $\sigma$ controls the perturbation strength:
as $\sigma \rightarrow 0$, we naturally recover $f$. 
An equivalent viewpoint is obtained by defining the transformation 
(change of variables)
\begin{equation*}
U \coloneqq \muv + \sigma Z.
\end{equation*}
We then have
\begin{equation*}
U \sim p_{\muv, \sigma},
\end{equation*}
where 
$p_{\muv, \sigma}$ is the \textbf{location-family distribution} generated
by the noise distribution $p$.
It is the pushforward distribution of $Z$ through the transformation (see
\cref{grad_est:sec:push_forward}). 
In this notation, the initial noise distribution $p$ is then simply 
$p = p_{\zeros,1}$. 
The perturbed function can then be expressed from these two perspectives as
\begin{align}
  f_\sigma(\muv)
  &= \EE_{Z \sim p_{\zeros,1}}[f(\muv + \sigma \cdot Z)] 
  \nonumber
  \\
  &= \EE_{U \sim p_{\muv,\sigma}}[f(U)].
  \label{perturb:eq:smooth_perturb2}
\end{align}
Writing the expectation as the integral of a p.d.f, we naturally recover
the smoothing by convolution presented earlier,
\begin{align*}
  f_\sigma(\muv) & = \int f(\muv + \sigma \z) p_{\zeros,1}(\z) d\z \\
  & = f \ast \kappa_\sigma(\muv),
\end{align*}
where we defined the kernel
\begin{equation*}
\kappa_\sigma(\z) 
\coloneqq 
p_{\zeros, \sigma}(-\z).
\end{equation*}
In the sequel, we assume that the noise distribution decomposes as
\begin{equation*}
  p_{\zeros,1}(\z) \coloneqq \exp(-\nu(\z)) / C,
\end{equation*}
where $\nu(\z)$ is the \textbf{log-density} of the noise distribution and $C$ is
a normalization constant.
For instance, the Gaussian distribution with diagonal covariance matrix 
and the corresponding \textbf{Gaussian kernel} are obtained with
$\nu(\z) = \frac{1}{2} \|\z\|^2_2$ and $C=\sqrt{2\pi}^M$.

\subsubsection*{Approximation by Monte-Carlo estimation}

Instead of approximating the integral above (continuous convolution)
with a discrete convolution
on a grid, as we did in \cref{conv:sec:discrete_conv},
the expectation perspective suggests that we can estimate
$f_\sigma(\muv)$ by Monte-Carlo estimation: we simply draw samples from the
distribution, evaluate the function at these samples and average.
Beyond mere Monte-Carlo estimation, more elaborate approximation schemes are
studied in \citep{chaudhuri2010smooth}.

\subsection{Gradient estimation by reparametrization}

Provided that the conditions for swapping differentiation and integration hold 
(see \cref{grad_est:rem:swap_derivative_integration}), we have
\begin{equation}\label{perturb:eq:expected_grad}
  \nabla f_\sigma(\muv) = 
  \EE_{Z \sim p_{\zeros,1}}[\nabla f(\muv + \sigma \cdot Z)].
\end{equation}
Note that if $f$ is only differentiable almost everywhere, the formula may still hold. 
For example, if $f$ is the
ReLU, then $\nabla f$ is the Heaviside step function, and we obtain the correct
gradient of $f_\sigma$ using the formula above; see
\cref{conv:ex:smoothed_heaviside}. However, if $f$ is not absolutely continuous,
the formula may not hold. For example, if $f$ is the Heaviside function, the
right-hand side of~\eqref{perturb:eq:expected_grad} is 0 which does not match
the gradient of $f_\sigma$; see again \cref{conv:ex:smoothed_heaviside}.

From the second expression of $f_{\sigma}$
in~\eqref{perturb:eq:smooth_perturb2}, we can see the formula of the gradient
in~\eqref{perturb:eq:expected_grad} as a reparametrization trick $U =
\muv+\sigma Z$; see \cref{grad_est:sec:pge}. Namely, we have
\begin{align}
\nabla f_\sigma(\muv) \nonumber
&= \nabla_\muv \EE_{U \sim p_{\muv,\sigma}}[f(U)] \nonumber \\
&= \nabla_\muv \EE_{Z \sim p_{\zeros,1}}[f(\muv + \sigma \cdot Z)] \nonumber \\
&= \EE_{Z \sim p_{\zeros,1}}[\nabla_\muv f(\muv + \sigma \cdot Z)] \nonumber \\
&= \EE_{Z \sim p_{\zeros,1}}[\nabla f(\muv + \sigma \cdot Z)].
\label{perturb:eq:grad_pge}
\end{align}

\subsection{Gradient estimation by SFE, Stein's lemma}

In some cases, we may not have access to $\nabla f$ or $f$ may not be absolutely
continuous and therefore the formula in~\eqref{perturb:eq:expected_grad} cannot
apply. For these cases, we can use the score function estimator (SFE)
from \cref{grad_est:sec:sfe}. Here,
for $f_\sigma(\muv)  = \EE_{U \sim p_{\muv,\sigma}}[f(U)]$, we obtain
\begin{equation*}
\nabla f_\sigma(\muv) 
= \EE_{U \sim p_{\muv, \sigma}}[f(U) \nabla_{\muv} \log p_{\muv,\sigma}(U)].
\end{equation*}
Since the PDF can be written as
\begin{equation*}
p_{\muv,\sigma}(\u) 
= \frac{1}{\sigma} p_{\zeros,1}((\u - \muv) / \sigma),
\end{equation*}
where
\begin{equation*}
p_{\zeros,1}(\z) \coloneqq \exp(-\nu(\z)) / C,
\end{equation*}
we obtain
\begin{equation*}
\nabla_\muv \log p_{\muv,\sigma}(\u)
= \nabla \nu((\u - \muv) / \sigma) / \sigma.
\end{equation*}
To summarize, we have shown that
\begin{align}
\nabla f_\sigma(\muv) 
&= \EE_{U \sim p_{\muv, \sigma}}[f(U) \nabla \nu((U - \muv) / \sigma) /
\sigma] \nonumber \\
&= \EE_{Z \sim p_{\zeros, 1}}[f(\muv + \sigma \cdot Z)
\nabla \nu(Z) / \sigma],
\label{perturb:eq:gradient_sfe}
\end{align}
where we used the change of variable $Z = (U - \muv) / \sigma$.
The same technique can also be used if we want to estimate the gradient w.r.t.
$\thetav = (\muv, \sigma)$ or if we want to estimate the Jacobian of the
expectation of a vector-valued function.

In the particular case of Gaussian noise, 
since $\nabla \nu(\z) = \z$,
we obtain
\begin{align*}
\nabla f_\sigma(\muv) 
&= \EE_{Z \sim p_{\zeros, 1}}[f(\muv + \sigma \cdot Z) Z / \sigma].
\end{align*}
This is known as \textbf{Stein's lemma}.
It should be noted that the above is an unbiased estimator of
the gradient of the smoothed function $f_\sigma$, but a biased estimator of the
gradient of the original function $f$ (assuming that it exists).
However, smoothing is usually a good thing, as it can accelerate the convergence
of gradient-based algorithms.
Computing the gradient of perturbed general programs is studied in detail in  
\citep{kreikemeyer2023smoothing}.

\subsection{Link between reparametrization and SFE}

Using the log-derivative identity, we have for any distribution with
differentiable density $p$
\begin{align*}
\EE_{Z \sim p} [h(Z) \nabla \log p(Z)]
&= \int_{\RR^M} h(\z) \left(\frac{\nabla p(\z)}{p(\z)}\right) p(\z) d\z \\
&= \int_{\RR^M} h(\z) \nabla p(\z) d\z.
\end{align*}
Using \textbf{integration by parts} and assuming that
$h(\z)p(\z)$ goes to zero when $\|\z\| \to \infty$,
we have
\begin{equation*}
\int_{\RR^M} h(\z) \nabla p(\z) d\z 
=
-\int_{\RR^M} p(\z) \nabla h(\z) d\z.
\end{equation*}
We have therefore the identity
\begin{equation*}
\EE_{Z \sim p} [h(Z) \nabla \log p(Z)]
= -\EE_{Z \sim p}[\nabla h(Z)].
\end{equation*}
Importantly, contrary to the SFE estimator from \cref{grad_est:sec:sfe},
this identity uses gradients with respect to $\z$, not with respect to the
parameters of the distribution. Nevertheless, 
using the reparametrization
$h(\z) \coloneqq f(\muv + \sigma \cdot \z)$,
we have
$\nabla h(\z) = \nabla f(\muv + \sigma \cdot \z) \cdot \sigma$
so that
\begin{align*}
\nabla f_\sigma(\muv)
&= \nabla_\muv \EE_{U \sim p_{\muv,\sigma}}[f(U)] \\
&= \EE_{Z \sim p_{\zeros,1}}[\nabla f(\muv + \sigma \cdot Z)]
\quad \text{(reparametrization trick)}\\
&= \EE_{Z \sim p_{\zeros,1}}[\nabla h(Z) / \sigma] \\
&= -\EE_{Z \sim p_{\zeros,1}}[h(Z) \nabla \log p(Z) / \sigma] \\
&= \EE_{Z \sim p_{\zeros,1}}[h(Z) \nabla \nu(Z) / \sigma] \\
&= \EE_{Z \sim p_{\zeros,1}}[f(\muv + \sigma \cdot Z) \nabla \nu(Z) / \sigma]
\quad \text{(score function estimator)}
\end{align*}
Essentially, integration by parts allowed us to convert the 
reparametrization trick estimator into the SFE estimator.
For more applications of integration by parts in machine learning, see Francis
Bach's excellent
\href{https://francisbach.com/integration-by-parts-randomized-smoothing-score-functions/}{blog post}.

\subsection{Variance reduction and evolution strategies}

\begin{figure}[t]
\centering
\includegraphics[scale=0.4]{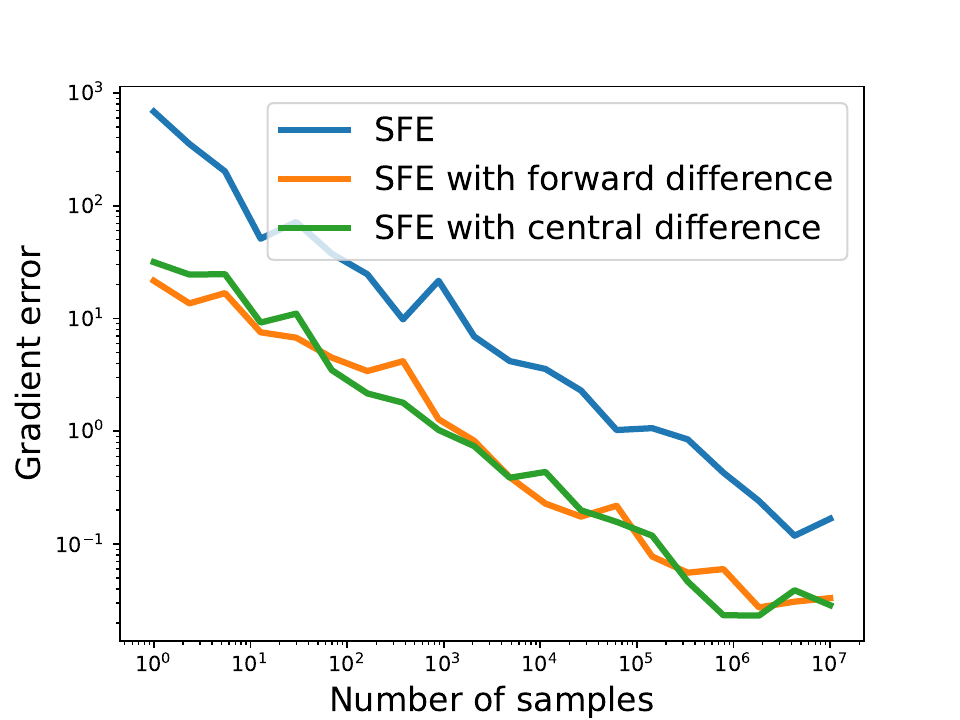}
\caption{
Comparison of the score function estimator (SFE) with or without variance
reduction for blackbox gradient estimation.
We show the
error $|\nabla f(\mu) - \nabla f_\sigma(\mu)|$ for
$f(u) \coloneqq u^3$ and $f_\sigma(\mu) \coloneqq \EE[f(\mu + \sigma
Z)]$, where $Z \sim \mathrm{Normal}(0,1)$ and $\sigma \coloneqq 0.1$.
To estimate $\nabla f_\sigma(\mu)$,
we compare three estimators: 
the vanilla SFE \cref{perturb:eq:gradient_sfe},
the SFE estimator with forward difference (variance reduced)
\cref{perturb:eq:finite_diff},
and the SFE estimator with central difference (variance reduced) 
\cref{perturb:eq:central_diff}.
In all three cases, we approximate the expectation by Monte-Carlo estimation
using some number of samples.
The variance-reduced estimators not only achieve smaller error, they are also
more numerically stable as $\sigma$ gets smaller.
}
\label{perturb:fig:sfe_error}
\end{figure}

As discussed in \cref{chap:grad_est}, the SFE suffers from high variance. We
now apply variance reduction techniques to it. To do so, we assume that $\nabla
\nu(Z)$ has zero mean for $Z \sim p_{\zeros, 1}$. This assumption for
example holds for Gaussian noise. This assumption implies that
\begin{equation*}
\EE_{Z \sim p_{\zeros, 1}}[f(\muv) \nabla \nu(Z) / \sigma] 
= f(\muv) \EE_{Z \sim p_{\zeros, 1}}[\nabla \nu(Z) / \sigma] 
= \zeros
\end{equation*}
and therefore
\begin{equation}\label{perturb:eq:finite_diff}
\nabla f_\sigma(\muv) 
= \EE_{Z \sim p_{\zeros, 1}}\left[(f(\muv + \sigma \cdot Z)
- f(\muv)) \nabla \nu(Z) / \sigma\right].
\end{equation}
This is an example of \textbf{control variate} discussed in
\cref{grad_est:sec:sfe}.
This can be interpreted as using a \textbf{finite difference}
for computing a directional derivative in the \textbf{random direction} $Z$ 
(see ``limit case'' below).
Inspired by a \textbf{central finite difference}, we can also use
\begin{equation}
\nabla f_\sigma(\muv) 
= \EE_{Z \sim p_{\zeros, 1}}\left[(f(\muv + \sigma \cdot Z)
- f(\muv - \sigma \cdot Z)) \nabla \nu(Z) / (2 \sigma)\right].
\label{perturb:eq:central_diff}
\end{equation}
These estimators have been used as part of blackbox (zero-order)
optimization algorithms,
such as \textbf{evolution strategies} \citep{salimans_2017}
or \textbf{random gradient-free optimization} \citep{nesterov2017random}.
For quadratic functions, it is easy to show that the second estimator achieves
lower variance \citep{recht_evolution}.
The idea of sampling both $Z$ and $-Z$ simultaneously is called 
antithetic \citep{geweke_1988} or mirrored sampling \citep{brockhoff_2010}.
Evolution strategies have also been used to obtain unbiased gradient estimators 
of partially unrolled computational graphs \citep{vicol_2021}.
We empirically compare the SFE with or without variance reduction for blackbox
gradient estimation in \cref{perturb:fig:sfe_error}.

\subsection{Zero-temperature limit}
\label{pert:sec:zero_temp_limit}

We now discuss the limit case $\sigma \to 0$.
That is, we assume that we do \textbf{not} want to perform smoothing
and that $\nabla f$ exists.
We recall that the directional derivative of $f$ at $\muv$ 
in the direction $\z$ is
\begin{align*}
\partial f(\muv)[\z]
& = \langle \nabla f(\muv), \z \rangle \\
& = \lim_{\sigma \to 0} 
\left[f(\muv + \sigma \cdot \z) - f(\muv)\right] / \sigma.
\end{align*}
When $\sigma \to 0$ and $Z$ follows the standard Gaussian distribution,
meaning that $\nabla \nu(\z) = \z$, 
\cref{perturb:eq:finite_diff} therefore becomes
\begin{align*}
\nabla f_\sigma(\muv) 
&= \EE_{Z \sim p_{\zeros, 1}} 
\left[\partial f(\muv)[Z] \nabla \nu(Z) \right] \\
&= \EE_{Z \sim p_{\zeros, 1}} 
\left[\partial f(\muv)[Z] Z \right] \\
&= \EE_{Z \sim p_{\zeros, 1}} 
\left[\langle \nabla f(\muv), Z \rangle Z\right] \\
&= \EE_{Z \sim p_{\zeros, 1}} 
\left[Z Z^\top \nabla f(\muv) \right] \\
&= \nabla f(\muv).
\end{align*}
This should not be too surprising, as we already know from the convolution
perspective that $f_\sigma(\muv) = (f \ast \kappa_\sigma)(\muv) \to f(\muv)$
when $\sigma \to 0$.
This recovers the randomized forward-mode estimator already presented in
\cref{auto_diff:sec:randomized_forward}.

\section{Gumbel tricks}
\label{pert:sec:gumbel_trick}

\subsection{The Gumbel distribution}

\begin{figure}[t]
\centering
\includegraphics[scale=0.4]{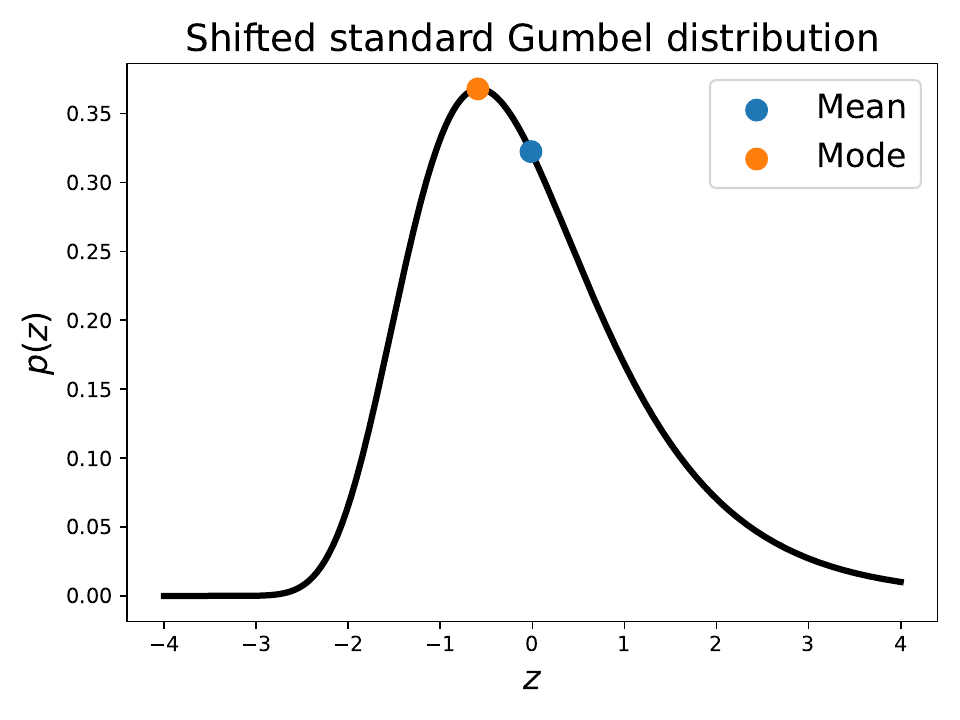}
\caption{We use a shifted definition of the standard Gumbel distribution so that
the mean is achieved at $z = 0$, and the mode at $z = -\gamma$.
With an unshifted definition, the mean and the mode would be achieved
at $\gamma$ and $0$, respectively.}
\label{conv:fig:gumbel_pdf}
\end{figure}

The Gumbel distribution is a distribution frequently used in extreme value
theory. 
As illustrated in \cref{conv:fig:gumbel_pdf},
we consider the shifted standard Gumbel distribution, 
whose PDF is defined by
\[
  p(z) \coloneqq \exp(-\nu(z)),
\]
where
\[
  \nu(z) \coloneqq z + \gamma + \exp(-(z+\gamma)),
\]
and where $\gamma \approx 0.577 $ is Euler's constant. 
Note that in some formulations, the distribution is not shifted, i.e.,
$\gamma$ is not added.
If $Z$ is distributed according to the shifted standard Gumbel distribution, 
we write $Z \sim \mathrm{Gumbel}(0, 1)$.

To obtain a multivariate extension with location-scale parameters $\muv$ and
$\sigma$, we take $M$ independent random variables $Z \coloneqq (Z_1, \ldots,
Z_M)$ and apply the location-scale transform
(\cref{grad_est:sec:location_scale_transform}). That is,
\begin{equation*}
U \sim \mathrm{Gumbel}(\muv, \sigma) 
\iff 
U = \muv + \sigma Z, ~ Z_i \sim \mathrm{Gumbel}(0, 1).
\end{equation*}
As we used shifted standard Gumbel distributions, we naturally get that $\EE[U]
= \muv$ and $\Var(U) = \sigma^2$. 
We can use Gumbel noise as an alternative to the Gaussian noise used in
\cref{perturb:sec:blackbox}. Thankfully, in particular cases, we can compute
closed-form expressions of the expectation of perturbed functions.

\begin{boxrem}{Link between Gumbel and exponential distribution}
  \label{perturb:rem:gumbel_exp_link}
  A random variable $Z$ is distributed as $\Gumbel(\mu, 1)$ if
  and only if $\exp(-Z)$ is distributed as an exponential distribution
  $\Exp(\exp(\mu - \gamma))$.
  To see this, one can simply compute the CDF of $\exp(-Z)$ and 
  recognize the CDF of $\Exp(\exp(\mu-\gamma))$. 
  Therefore, when comparing Gumbel distributions, we can use standard
  properties of the exponential distribution. 
\end{boxrem}

\begin{boxrem}{Sampling Gumbel noise}
If $U \sim \mathrm{Uniform}(0,1)$,
then $Z \coloneqq -\log(-\log(U)) - \gamma$ satisfies $Z \sim \mathrm{Gumbel}(0,
1)$, where we recall that we use $\mathrm{Gumbel}(0,1)$ to denote the shifted standard Gumbel distribution. To see this, note that 
$\PP(Z \leq t) = \PP(U \leq \exp(\exp(-(\gamma +t)))) = \exp(\exp(-(\gamma +t)))$, where the last expression matches the CDF of $\mathrm{Gumbel}(0,1)$.
\label{pert:rem:gumbel_noise}
\end{boxrem}

\subsection{Perturbed comparison}

To start with, the Gumbel distribution can be used to smooth a binary comparison
like the greater than or equal operators. Recall that the latter is defined for
any $\mu_1, \mu_2 \in \RR$ as
\[
  \mathrm{gt}(\mu_1, \mu_2) \coloneqq \begin{cases}
    1 & \mbox{if} \ \mu_1 \geq  \mu_2 \\
    0 & \mbox{if} \ \mu_1 < \mu_2
  \end{cases} = \heavistep(\mu_1-\mu_2),
\]
where $\heavistep$ is the Heaviside function. 
As shown below, by perturbing each variable with Gumbel noise,
we recover $\logistic(a-b)= 1/(1+ e^{-(a-b)})$ as
an approximation of $\heavistep(a-b)$.
\begin{boxprop}{Gumbel trick for binary variables}\label{perturb:prop:gumbel_binary}
   Let\\ $Z_1, Z_2 \sim \mathrm{Gumbel}(0,1)$ be
   two independent random variables.
   The difference of their location-scale transform
   (\cref{grad_est:sec:location_scale_transform}) is distributed according to a
   logistic distribution (\cref{proba_learn:rem:logistic}), i.e.,
  \[
      \mu_1 + \sigma Z_1 - (\mu_2 + \sigma Z_2)  \sim
      \mathrm{Logistic}(\mu_1-\mu_2,\sigma),
  \]
  for $\mu_1, \mu_2 \in \RR$ and $\sigma >0$. In particular, we have 
  \[
      \EE_{Z_1,Z_2}[\mathrm{gt}(\mu_1 + \sigma Z_1, \mu_2 + \sigma Z_2)] 
    = \frac{1}{1+ e^{-(\mu_1-\mu_2)/\sigma}}.
  \]
\end{boxprop}
\begin{proof}
    We first derive the CDF of $\mu_1 + \sigma Z_1 - (\mu_2 + \sigma Z_2)$ as 
  \begin{align*}
    \PP(\mu_1 + \sigma Z_1 - (\mu_2 + \sigma Z_2) \leq t) 
    & = \PP\left(\mu_1/\sigma + Z_1 \leq (\mu_2 + t)/\sigma + Z_2\right) \\
    & = \PP\left(e^{-(\mu_1/\sigma + Z_1)} \geq e^{-((\mu_2 + t)/\sigma + Z_2)}\right).
  \end{align*}
  By~\cref{perturb:rem:gumbel_exp_link}, $e^{-(\mu_1/\sigma + Z_1)}\sim
  \Exp(\exp(\mu_1/\sigma - \gamma))$, and similarly for $e^{-(\mu_2 + t)/\sigma + Z_2}$. Now one easily
  shows that if $U \sim \Exp(u), V \sim \Exp(v)$ independent, then  $\PP(U\leq
  V) = u/(u+v)$. Hence, we get 
  \begin{align*}
    \PP(\mu_1 + \sigma Z_1 - (\mu_2 + \sigma Z_2) \leq t) 
    & = \frac{e^{(\mu_2 + t)/\sigma - \gamma }}{e^{(\mu_2 + t)/\sigma - \gamma }
    + e^{\mu_1/\sigma - \gamma}} \\
    & = \frac{1}{1+ e^{-(t- (\mu_1-\mu_2))/\sigma}}.
  \end{align*}
  We recognize the CDF of the logistic distribution with mean $\mu_1-\mu_2$ and scale
  $\sigma$, denoted $\mathrm{Logistic}(\mu_1 - \mu_2, \sigma)$.
  For the last claim, we simply have that
  \begin{align*}
    \EE[\mathrm{gt}(\mu_1 + \sigma Z_1, \mu_2 + \sigma Z_2)] 
    & = \EE\left[\heavistep(\mu_1+ \sigma Z_1 - (\mu_2+ \sigma Z_2))\right] \\
    & = \PP(\mu_1 + \sigma Z_1 - (\mu_2 + \sigma Z_2)\geq 0) \\
    & = \frac{1}{1+e^{-(\mu_1-\mu_2)/\sigma}}.
  \end{align*}
\end{proof}

\subsection{Perturbed argmax}
\label{perturb:sec:argmax}

Suppose we want to smooth
\begin{equation*}
\y(\u) 
\coloneqq 
\argmax_{\y \in \{\e_1, \dots, \e_M\}} \langle \y, \u \rangle
= \phi(i(\u)),
\end{equation*}
where
\begin{align*}
i(\u) & \coloneqq \argmax_{i \in [M]} u_i  \\
\phi(i) & \coloneqq \e_i
\end{align*}
where $\phi(i)$ is the one-hot encoding of $i \in [M]$. It turns out that the
function $\y(\u)$ perturbed using Gumbel noise enjoys a closed form expectation,
which is nothing else than the softargmax.
\begin{boxprop}{Gumbel trick for categorical variables}\label{perturb:prop:gumbel_trick}
Let us define $M$ independent random variables 
$Z \sim \mathrm{Gumbel}(\zeros, 1) \in \RR^M$.
Then, 
for $\muv \in \RR^M$ and $\sigma >0$,
\begin{equation*}
Y \coloneqq i(\muv + \sigma \cdot Z)
\implies
Y \sim
\mathrm{Categorical}(\mathrm{softargmax}(\muv / \sigma)).
\end{equation*}
Moreover, we have
\begin{align*}
\y_\sigma(\muv) 
& = \EE_Z[\y(\muv + \sigma \cdot Z)] \\
& = \EE_Y[\phi(Y)] \\
& = \mathrm{softargmax}(\muv / \sigma).
\end{align*}
\end{boxprop}
\begin{proof}
  For $k \in [M]$, we have that  
  \begin{align*}
    \PP(Y = k) & = \PP\left(\argmax_{i \in [M]} \{\mu_i + \sigma Z_i\} = k\right) \\
    & = \PP\left(\argmin_{i \in [M]} \{e^{-\mu_i/\sigma - Z_i} \} = k\right)
  \end{align*}
  By~\cref{perturb:rem:gumbel_exp_link}, we have that $e^{-\mu_i/\sigma - Z_i}
  \sim \Exp(\exp(\mu_i/\sigma - \gamma))$. One easily verifies as an exercise, that, for
  $U_1, \ldots, U_M$ independent exponential variables with parameters $u_1,
  \ldots, u_M$, we have $\PP(\argmin_{i \in [M]} \{U_i\} = k) =
  u_k/\sum_{i=1}^{M} u_i$. Hence, we get 
  \[
    \PP(Y = k) = \frac{\exp(\mu_k/\sigma)}{\sum_{i=1}^M \exp(\mu_i/\sigma)},
  \]
  that is, 
  \[
    Y \sim \mathrm{Categorical}(\mathrm{softargmax}(\muv / \sigma)).
  \]
  The last claim follows from the distribution of $Y$ and the definition of
  $\phi$.
\end{proof}

\subsection{Perturbed max}

A similar result holds if we now wish to perturb the max instead of the argmax.
\begin{boxprop}{Link to log-sum-exp}
Let us define $M$ independent random variables $Z \sim \mathrm{Gumbel}(\zeros,
1) \in \RR^M$ and
\begin{equation*}
  f(\u) \coloneqq \max_{i \in [M]} u_i.
\end{equation*}
Then,
for $\muv \in \RR^M$ and $\sigma >0$,
\begin{equation*}
  V \coloneqq f(\muv + \sigma \cdot Z)
\implies
V \sim
\mathrm{Gumbel}(\sigma \mathrm{LSE}(\muv/\sigma), \sigma).
\end{equation*}
Moreover, we have
\begin{equation*}
f_\sigma(\muv)
\coloneqq \EE_Z[f(\muv + \sigma \cdot Z)]
= \EE_V[V]
= \sigma \cdot \mathrm{LSE}(\muv / \sigma).
\end{equation*}
\end{boxprop}
\begin{proof}
  We derive the CDF of $f(\muv + \sigma \cdot Z)$ as 
  \begin{align*}
    \PP(\max_{i\in [M]} \{\mu_i + \sigma Z_i\} \leq t) 
    & = \PP\left(\min_{i \in [M]} \{ e^{-(\mu_i/\sigma -Z_i)} \}\geq e^{-t/\sigma}\right)
  \end{align*}
  We have $ e^{-(\mu_i/\sigma -Z_i)} \sim \Exp(\exp(\mu_i/\sigma) - \gamma)$ and for
  $U_1, \ldots, U_M$ independent exponential random variables with parameters
  $u_i$, we have $\min_{i \in [M]} U_i \sim \Exp(\sum_{i=1}^M u_i)$. Hence, 
  \begin{align*}
    \PP\left(\max_{i\in [M]} \{\mu_i + \sigma Z_i\} \leq t\right) 
    & = \exp\left(-\sum_{i=1}^M (\exp(\mu_i/\sigma -\gamma) ) \exp(-t/\sigma)\right) \\
    & = \exp(-\exp(-(t - \sigma\mathrm{LSE}(\muv/\sigma))/\sigma - \gamma)). \\
  \end{align*}
  We recognize the CDF of the shifted Gumbel distribution with location-scale parameters
  $\sigma \mathrm{LSE}(\muv/\sigma)$ and $\sigma$.
\end{proof}
For further reading on the Gumbel trick, see
Tim Vieira's great \href{https://timvieira.github.io/blog/}{blog}.

\subsection{Gumbel trick for sampling}

The Gumbel trick is also useful in its own right for \textbf{sampling}
without computing the normalization constant of the softargmax.
Indeed, \cref{perturb:prop:gumbel_trick} ensures that if
$Z$ is Gumbel noise, then $Y$ is distributed according to
$\mathrm{Categorical}(\mathrm{softargmax}(\muv / \sigma))$.
Computing the arg-maximum, as required to compute $Y$, can be done in one pass.
Therefore, we obtain a one-pass algorithm to sample directly from the logits
$\muv$, without explicitly computing the probabilities
$\mathrm{softargmax}(\muv/\sigma)$. 

One may wonder whether such trick could also be used with the normal
distribution. Unfortunately, there is no closed form in this case because it
would require integrating the CDF of the maximum of $M-1$ Gaussian
distributions. However, other tricks can be defined such as using Weibull
distributions, see~\citet{balog2017lost}. 

\subsection{Perturb-and-MAP}

Previously, we discussed the Gumbel trick in
the classification setting, where $\cY = [M]$. 
In the structured prediction setting, outputs are typically embedded in $\RR^M$ 
but the output space is very large. That is,
$\cY \subseteq \RR^M$ but $|\cY| \gg M$.
Structured outputs are then decoded using a 
maximum a-posteriori (MAP) oracle
\begin{align*}
f(\u)
&\coloneqq \max_{\y \in \cY} \langle \y, \u \rangle \\
\y(\u)
&\coloneqq \argmax_{\y \in \cY} \langle \y, \u \rangle.
\end{align*}
For this setting, the perturbed versions of $f$ and $\y$,
\begin{align*}
f_\sigma(\muv) 
&\coloneqq \EE_Z
[f(\muv + \sigma \cdot Z)] \\
\y_\sigma(\muv) 
&\coloneqq \EE_Z
[\y(\muv + \sigma \cdot Z)],
\end{align*}
no longer enjoy a closed form in general.  
However, we can approximate them using Monte-carlo estimation.
For the gradient of $\nabla f_\sigma(\muv)$, two estimators exist
\citep{abernethy_2016,berthet_2020}.
\begin{boxprop}{Gradient of perturbed max}
Let $\cY \subseteq \RR^M$ and $Z$ be noise with density
\begin{equation*}
p_{\zeros,1}(\z) \coloneqq \exp(-\nu(\z)) / C.
\end{equation*}
Then, $f_\sigma(\muv)$ is smooth, and its gradient is given by
\begin{align*}
  \nabla f_\sigma(\muv) 
&= \EE_Z [\y(\muv + \sigma \cdot Z)] \\
&= \EE_Z [f(\muv + \sigma \cdot Z) \nabla \nu(Z) /
\sigma] \\
& \in \conv(\cY).
\end{align*}
We therefore have $\nabla f_\sigma(\muv) = \y_\sigma(\muv)$.
\label{perturb:prop:gradient_perturbed_max}
\end{boxprop}
The first estimator is simply a consequence of the
reparametrization trick seen in \cref{perturb:eq:grad_pge} and of
$\y = \nabla f$, which follows from Danskin's theorem
(see \cref{implicit:sec:danskin}).
The second estimator is just SFE seen in
\cref{perturb:eq:gradient_sfe}. The first estimator usually has lower variance,
as it uses more information, namely that $\y = \nabla f$.

The Jacobian of $\y_\sigma(\muv)$ also has two estimators
\citep{abernethy_2016,berthet_2020}.
\begin{boxprop}{Jacobian of perturbed argmax}
Under the same notation as in \cref{perturb:prop:gradient_perturbed_max}, 
we have
\begin{align*}
\partial \y_\sigma(\muv) 
&= \EE_Z\left[\y(\muv + \sigma Z)\nabla \nu(Z)^\top /
\sigma\right] \\
&= \EE_Z\left[f(\muv + \sigma Z)
    \left(\nabla \nu(Z)\nabla \nu(Z)^\top - \nabla^2 \nu(Z)\right) /
\sigma^2\right].
\end{align*}
\end{boxprop}
The first estimator uses SFE.
The second estimator is obtained by differentiating through 
\begin{equation*}
\y_\sigma(\muv)
= \nabla f_\sigma(\muv)
= \EE_Z [f(\muv + \sigma \cdot Z) \nabla \nu(Z) /
\sigma].
\end{equation*}
The first estimator usually has lower variance.
Note that we cannot use the reparametrization trick this time, 
since $\y$ is discontinuous, contrary to $f$.

\subsubsection*{Link between perturbation and regularization}

As shown in \citep[Proposition 2.2]{berthet_2020},
assuming
$\cY$ is a convex polytope with non-empty interior
and
$p$ has a strictly positive density,
the function
\begin{equation*}
f_\sigma(\muv) 
\coloneqq \EE_Z [f(\muv + \sigma \cdot Z)]
= \EE_Z [\max_{\y \in \cY} \langle \muv + \sigma \cdot Z, \y \rangle]
\end{equation*}
is strictly convex and its convex conjugate $f^*_\sigma(\y)$ is
Legendre-type. 
We can therefore rewrite
$f_\sigma(\muv)$
from the regularization perspective as
\begin{equation*}
f_\sigma(\muv) 
=
\max_{\y \in \cY} \langle \muv, \y \rangle - f^*_\sigma(\y).
\end{equation*}
and $\nabla f_\sigma(\muv) = \y_\sigma(\muv)$ is a \textbf{mirror map},
a one-to-one mapping from $\RR^M$ to the interior of $\cY$.
Unfortunately,
$f^*_\sigma(\y)$
does not enjoy a closed form in general.
Conversely, does any regularization have a corresponding noise distribution?
The reciprocal is not true.

\subsection{Gumbel-softargmax}

Suppose we want to smooth out the \textbf{composition} $h(\u) \coloneqq
g(\y(\u))$ between some function 
$g \colon \{\e_1, \dots, \e_M\} \to \RR$ and the argmax
\begin{equation*}
    \y(\u) \coloneqq \argmax_{\y \in \{\e_1, \dots, \e_M\}} \langle \y, \u \rangle.
\end{equation*}
We can do so by
\begin{equation*}
h_\sigma(\muv) 
\coloneqq \EE_Z \left[g(\y(\muv + \sigma Z))\right].
\end{equation*}
This is useful for instance to compute the expectation of a loss (instead of the
loss of an expectation).
To compute the gradient of
$h_\sigma(\muv)$, we can readily use the SFE.
However, we saw that it suffers from high variance.
Unfortunately, we cannot swap differentiation and integration (expectation)
here, since $\y(\u)$ is a discontinuous function.
See \cref{grad_est:rem:swap_derivative_integration} for more details 
regarding differentiation under the integral sign.

The key idea of the Gumbel-softargmax \citep{jang_2016,maddison_2016} is to
replace $\y(\u)$ with a softargmax (with temperature parameter $\tau$) to define
\begin{equation*}
h_{\sigma,\tau}(\muv) 
\coloneqq \EE_Z \left[g(\mathrm{softargmax}_\tau(\muv +
\sigma Z))\right].
\end{equation*}
Since the softargmax is a regularized argmax,
we can see the Gumbel-softargmax approach as using \textbf{both} regularization
and perturbation.
Note that the approach is also known as Gumbel-softmax. 
We use the name Gumbel-softargmax for consistency with the terminology of this
book.

The key benefit is that we can now swap differentiation and integration (expectation)
to get an unbiased estimator of $\nabla h_{\sigma,\tau}(\muv)$.
However, this will be a \textbf{biased} estimator of 
$\nabla h_{\sigma}(\muv)$, the amount of bias being controlled by the
temperature $\tau$. 
In particular, in the limit case $\tau \to 0$, 
we have $h_{\sigma,\tau}(\muv) \to h_\sigma(\muv)$.
One caveat, however, is that the function $g$ needs to
be well defined on $\triangle^M$, instead of $\{\e_1,\dots,\e_M\}$.

The use of the softargmax transformation defines a continuous distribution
\citep{jang_2016,maddison_2016},
that we now explain with $\sigma=1$.
\begin{boxprop}{Gumbel-softargmax / Concrete distributions}
Let us define the \textbf{continuous} random variable
\begin{equation*}
T 
\coloneqq \mathrm{softargmax}_\tau(\muv + Z) \in \triangle^M,
\end{equation*}
where $Z$ is a Gumbel random variable. Then $T$ is
distributed according to a distribution with density
\begin{equation*}
p_{\muv,\tau}(\t) \coloneqq \Gamma(M) \tau^{M-1}
\left( \sum_{i=1}^M \frac{\pi_i}{t_i^\tau} \right)^{-M}
\prod_{i=1}^M \frac{\pi_i}{t_i^{\tau +1}},
\end{equation*}
where $\piv \coloneqq \mathrm{softargmax}(\muv)$.
\end{boxprop}
We can extend the Gumbel softargmax to the structured setting by replacing
\begin{equation*}
\y(\u) \coloneqq \argmax_{\y \in \cY} \langle \y, \u \rangle,
\end{equation*}
with its regularized variant \citep{paulus_2020}.
Similarly as before, one caveat is that $g$ needs to be well defined on
$\conv(\cY)$ instead of $\cY$. Moreover, regularizing $\y$ is not always easy
computationally.

\section{Summary}

\begin{itemize}

\item We studied smoothing techniques based on function convolution with a kernel.
Due to the commutativity of the convolution,
we can alternatively see these as the expectation of the function,
perturbed with noise, assuming the kernel corresponds to the PDF of some noise
distribution.

\item Their gradients can be estimated using the path gradient estimator (PGE) or
score function estimator (SFE), depending on whether the gradient of the
original function is available or not.

\item We saw that
Stein's lemma is a special case of SFE used with Gaussian noise.
The so-called ``evolution strategies'' are just a variant of that with variance
reduction and can be interpreted as randomized finite difference. 

\item When using Gumbel noise, we were able to derive closed-form expressions for the
expectation in specific cases: perturbed comparison, perturbed argmax and
perturbed max.

\item We also studied the connections between smoothing by optimization and smoothing
by integration.  Infimal convolution is the counterpart of convolution, and the
Legendre-Fenchel transform is the counterpart of Fourier and Laplace's
transforms.
Infimal convolution uses a min-plus algebra in the log domain, 
while the convolution uses a sum-product algebra in the exponential domain.

\end{itemize}

%% file: chapters/optim/basics.tex
\chapter{Optimization basics}

\section{Objective functions}

Consider a function $L$, for example evaluating the error or ``loss'' $L(\w)$
achieved by a model with parameters $\w \in \cW$, where $\cW=\RR^P$. 
To find the best
possible model parameterization, we seek to minimize $L(\w)$, that is, to
compute approximately
\[
L^\star \coloneqq \inf_{\w \in \cW} L(\w),
\]
assuming that the infimum exists (i.e., $L(\w)$ is lower bounded). We will
denote a solution, if it exists, by
\begin{equation*}
\w^\star \in \argmin_{\w \in \cW} L(\w) 
\coloneqq \left\{\w \in \cW: L(\w) = \min_{\w' \in \cW} L(\w')\right\}.
\end{equation*}
In general, an analytical solution is not available
and computing such a minimum approximately requires an optimization algorithm. 
An optimization algorithm is an iterative procedure, which, starting from an
initial point $\w \pow 0$, outputs after $t$ iterations a point $\w \pow t$
that approximates the minimum of $L$ up to some accuracy $\varepsilon$, i.e.,
\begin{equation}
L(\w \pow t) - L^\star \leq \varepsilon.
\label{optim:eq:epsilon_accurate}
\end{equation}

\section{Oracles}

To produce iterates $\w \pow 1, \w \pow 2, \dots$ that converge to a minimum,
the algorithm naturally needs to have access to information about $L$. 
For example, the algorithm needs a priori to be able to evaluate $L$
to know if it decreased its value or not. Such information about the function  
is formalized by the notion of
\textbf{oracles}~\citep{nemirovski1983problem}.
Formally, oracles are procedures that an algorithm can call to access
information about the objective $L(\w)$ at any given point $\w \in \cW$. 
We usually mainly consider the following three oracles.
\begin{itemize}[nosep]
    \item \textbf{Zero-order oracle:} evaluating the function
    $L(\w) \in \RR$.
    \item \textbf{First-order oracle:} evaluating the gradient $\nabla L(\w) \in
        \cW$ for $L$ differentiable.
    \item \textbf{Second-order oracle:} evaluating the Hessian matrix $\nabla^2
        L(\w)$, or evaluating the Hessian-vector product (HVP) $\nabla^2 L(\w)
        \v \in \cW$, for $L$ twice differentiable and any vector $\v \in
        \cW$.
\end{itemize}
Given an oracle $\mathcal{O}$ for a function $L$, we can formally define an
optimization algorithm as a procedure which computes the next iterate as a
function of all past and current information. Formally, an algorithm
$\mathcal{A}$ builds a sequence $\w \pow 1, \ldots, \w \pow t$ from a starting
point $\w \pow 0$ as 
\[
  \w \pow {t+1} 
  \coloneqq \mathcal{A}(\w \pow 0, \ldots, \w \pow t, 
    \mathcal{O}(\w \pow 0), \ldots, \mathcal{O}(\w \pow t), \lambdav),
\]
where $\lambdav \in \Lambda \subseteq \RR^Q$ 
encapsulates some hyperparameters of the algorithm,
such as the stepsize. Oftentimes, algorithms build the next iterate simply from
the information collected at the current iterate, without using all past
iterates. That is, they take the form $\w
\pow {t+1} = \mathcal{A}(\w \pow t, \mathcal{O}(\w \pow t), \lambdav)$. A
classical example is the gradient descent algorithm, 
which uses a first-order oracle to compute iterates of the form 
\[
  \w \pow {t+1} \coloneqq \w \pow t - \gamma \nabla L(\w \pow t),  
\]
where the stepsize $\gamma$ is a hyperparameter of the algorithm. The notion of
oracle therefore delineates different classes of algorithms. 
For instance, we may consider zero-order algorithms or first-order algorithms.

\section{Variational perspective of optimization algorithms}

One of the most basic optimization algorithms is the \textbf{proximal point}
method, which produces $\w \pow {t+1}$ from $\w \pow t$ by
\begin{equation*}
\w \pow {t+1} \coloneqq
\argmin_{\w \in \cW} L(\w) + \frac{1}{2\gamma} \|\w - \w \pow t\|_2^2.
\end{equation*}
In words, the next iterate is produced by solving
a trade-off between minimizing the function $L$ and staying close to $\w^t$.
Unfortunately, the optimization problem involved in performing this parameter
update is as difficult as the original optimization problem,
making the proximal point method impractical.

As we shall see in \cref{chap:optim} and \cref{chap:optim2}, 
many optimization algorithms can be seen as an approximation of
the proximal point method, in the sense that they solve
\begin{equation*}
\w \pow {t+1} \coloneqq
\argmin_{\w \in \cW} \tilde{L}(\w, \w \pow t) + 
\frac{1}{2\gamma} \|\w - \w \pow t\|_2^2.
\end{equation*}
or more generally
\begin{equation*}
\w \pow {t+1} \coloneqq
\argmin_{\w \in \cW} \tilde{L}(\w, \w \pow t) + 
\frac{1}{\gamma} d(\w,  \w \pow t),
\end{equation*}
where $\tilde{L}(\w, \w \pow t)$ is an approximation of $L(\w)$ around $\w \pow
t$ and $d(\w, \w')$ is some form of distance between $\w$ and $\w'$.
Different choices of $\tilde{L}$ and $d$ lead to different optimization
algorithms, and to different trade-offs.

\section{Classes of functions}
\label{optim:sec:function_classes}

When studying algorithms theoretically, stronger results can often be stated by
restricting to certain classes of functions. We already covered continuous and
differentiable functions in \cref{chap:diff}. We review a few important other
classes in this section.

\subsection{Lipschitz functions}

Lipschitz continuity is a stronger form of continuity.
Intuitively, a Lipschitz continuous function is limited in how fast it can
change.
\begin{boxdef}{Lipschitz-continuous functions}
A function $g \colon \cW \to \cF$ is $\beta$-\textbf{Lipschitz continuous}
 if for all $\w, \v \in \cW$
\begin{equation*}
\|g(\w) - g(\v)\|_2 \le \beta \|\w - \v\|_2.
\end{equation*}
\label{optim:def:lipschitz}
\end{boxdef}
Note that the definition is valid even for vector-valued functions.

\subsubsection*{With respect to arbitrary norms}

Thanks to dual norms reviewed in \cref{duality:sec:dual_norms}, we can state a
more general definition of Lipschitz continuity based on arbitrary norms,
instead of the $2$-norm. Moreover, we may consider Lipschitz-continuity over a
subset of the input domain.
\begin{boxdef}{Lipschitz continuous functions \wrt a norm}
A function $g \colon \cW \to \cF$ is said to be $\beta$-Lipschitz w.r.t. 
a norm $\|\cdot\|$ over a set $\cC \subseteq \cW$ if for all $\w, \v \in \cC$
\begin{equation*}
\|g(\w) - g(\v)\|_* \le \beta \|\w - \v\|.
\end{equation*}
\end{boxdef}
When $\|\cdot\| = \|\cdot\|_2$, we recover \cref{optim:def:lipschitz},
since the $2$-norm is dual to itself.

\subsection{Smooth functions}

A differentiable function $L$ is said to be $\beta$-smooth if its gradients are
$\beta$-Lipschitz continuous. Setting $g(\w) = \nabla L(\w)$ in
\cref{optim:def:lipschitz}, we obtain the following definition.
\begin{boxdef}{Smooth functions}\label{optim:def:smooth}
  A differentiable function \\$L \colon \cW \to \RR$ 
  is $\beta$-\textbf{smooth} for $\beta > 0$ if 
  for all $\w, \v \in \cW$
  \[
    \|\nabla L(\w) - \nabla L(\v)\|_2 \leq \beta \|\w-\v\|_2.
  \]
\end{boxdef}
Smoothness ensures that the information provided by the gradient at some $\w$ is
meaningful in a neighborhood of $\w$, since its variations are upper-bounded. If
the variations were not bounded, the gradient at $\v$ arbitrarily close to $\w$
could drastically change, rendering the information provided by a first-order
oracle potentially useless. 

Smoothness of a function can be
interpreted as having a quadratic upper bound on the function as formalized
below.
\begin{boxprop}{Smooth functions}
If a differentiable function $L:\cW \to \RR$ is $\beta$-smooth then
for all $\w, \v \in \cW$, 
\[
  |L(\w) - L(\v) + \langle \nabla L(\v), \w-\v\rangle|
  \leq
  \frac{\beta}{2}\|\w-\v\|_2^2.  
\]
In particular, we have
\[
L(\w) 
\leq 
L(\v) + \langle \nabla L(\v), \w-\v\rangle  +
\frac{\beta}{2}\|\w-\v\|_2^2.
\]
\label{optim:prop:smoothness}
\end{boxprop}
\begin{proof}
This is shown by bounding 
$|L(\v)- L(\w) - \langle \nabla L(\w), \v-\w\rangle|$ 
using the integral representation of the objective along $\w-\v$, i.e., 
$|L(\v)- L(\w) - \langle \nabla L(\w), \v -\w\rangle| 
= |\int_0^1 \langle \nabla L(\w + s(\v-\w)), \v-\w\rangle ds 
- \langle \nabla L(\w), \v-\w\rangle|
\leq \int_0^1 \|\nabla L(\w + s(\v-\w)) -  \nabla L(\w)\|_2ds\|\v-\w\|_2
\leq \beta \|\w-\v\|_2^2/2$, where the last inequality follows from the
smoothness assumption and standard integration.
\end{proof}
In other words, $L(\w)$ is upper-bounded and lower-bounded around $\v$ by a
quadratic function of $\w$. We will see in \cref{optim:sec:gd} that this
characterization gives rise to a variational perspective on gradient descent.

\subsubsection*{With respect to arbitrary norms}

We can generalize the definition of smoothness in
\cref{optim:def:smooth} to arbitrary norms.
\begin{boxdef}{Smooth functions \wrt a norm}
A function $L:\cW \to \RR$ is $\beta$-smooth w.r.t. a norm $\|\cdot\|$ over a
set $\cC$ if for all $\w, \v \in \cC$
\begin{equation*}
\|\nabla L(\w) - \nabla L(\v)\|_* 
\le  
\frac{\beta}{2} \|\w - \v\|.
\end{equation*}
\end{boxdef}
An equivalent characterization,
generalizing \cref{optim:prop:smoothness} to arbitrary norms,
is given below (see, e.g. \citet[Theorem 5.8]{beck_2017}).
\begin{boxprop}{Smooth functions \wrt a norm}
If a differentiable function $L:\cW \to \RR$ is $\beta$-smooth w.r.t. a norm
$\|\cdot\|$ over a set $\cC$, then for all $\w, \v \in \cC$
\begin{equation*}
|\underbrace{L(\w) - L(\v) - \langle \nabla L(\v), \w - \v \rangle}_{B_L(\w, \v)}|
\le  \frac{\beta}{2} \|\w - \v\|^2,
\end{equation*}
where $B_L$ is the Bregman divergence generated by $L$
(\cref{duality:def:bregman_div}).
\end{boxprop}

\subsection{Convex functions}

A convex function is a function such that its value on the
average of two or more points is smaller than the average of the values of the
functions at these points. This is illustrated in Figure~\ref{optim:fig:cvx} and
formalized below.
\begin{boxdef}{Convex functions}
    A function $L: \cW \rightarrow \RR$ is said to be \textbf{convex} if 
      for all $\w, \v \in \cW$ and $\tau\in [0, 1]$
    \[
      L(\tau\w + (1-\tau)\v) \leq \tau L(\w) + (1-\tau)L(\v).
    \]
    The function $L$ is \textbf{strictly convex} if the above inequality is
    strict for all $\w \neq \v$.
\end{boxdef}
The above characterization can easily be generalized to multiple points. Namely,
for $\w_1, \ldots, \w_n \in \cW$ and $\tau_1, \ldots, \tau_n \geq 0$ such
that $\sum_{i=1}^n \tau_i = 1$ (that is, $\tau_1, \ldots, \tau_n$
defines a probability distribution over $[n])$, we have if $L$ is convex that
\[
  L\left(\sum_{i=1}^n \tau_i \w_i\right) \leq
  \sum_{i=1}^n \tau_i L(\w_i).
\]
The point $\sum_{i=1}^n \tau_i \w_i$ is called a convex combination.
This can be seen as comparing the function at the average point to the average
of the values at these points and can further be generalized to any random
variable.
\begin{boxprop}{Jensen's inequality}
    A function $L \colon \cW \rightarrow \RR$ is convex if it satisfies
    \textbf{Jensen's inequality}, that is, for any random variable $W$ on
    $\cW$,
    \[
      L(\mathbb{E}[W]) \leq \mathbb{E}[L(W)],
    \]
    provided that the expectations are well-defined.
\end{boxprop}

If the function considered is differentiable, an alternative characterization of
convexity is to observe how linear approximations of the function lower bound
the function. This is illustrated in Figure~\ref{optim:fig:cvx} and formalized
below.
\begin{boxdef}{Convex differentiable functions}\label{optim:def:convex_differentiable}
    A differentiable function $L:\cW \rightarrow \RR$ is convex if
    and only if for all $\w, \v \in \cW$
    \[
        L(\v) \geq L(\w) + \langle \nabla L(\w), \v-\w\rangle.
    \] 
    The function $L$ is strictly convex if and only if the above
    inequality is strict for any $\w \neq \v$.
\end{boxdef}
The above characterization pinpoints the relevance of convex functions in
optimization: if we can find a point $\hat\w$ with null gradient, then we know
that we have found the minimum as we have
\[
  \nabla L(\hat \w) = \zeros \implies  
  \forall \v \in \RR^P, \  L(\v) \geq L(\hat \w) \implies 
  L(\hat \w) = L^\star.
\]
This means that by having access to the gradient of the function or an
approximation thereof, we have access to a sufficient criterion to know whether
we found a global minimum. In the case of a gradient descent on a smooth
function, convexity ensures convergence to a minimum at a sublinear rate as
detailed below.

Finally, if the function is twice differentiable, convexity of a function can be
characterized in terms of the Hessian of the function.

\begin{boxprop}{Convex twice differentiable functions}\label{optim:prop:convex_twice_differentiable}
  A twice differentiable function $L: \cW \rightarrow \RR$ is convex if and
  only if its Hessian is positive semi-definite, 
  \[
  \forall  \w \in \cW, \ \nabla^2 L(\w) \succeq 0, 
  \ \mbox{i.e.}, \ 
  \forall \w, \v  \in \cW, \ 
  \langle \v, \nabla^2 L(\w) \v \rangle \geq 0.
  \]
  The function $L$ is strictly convex if and only if the Hessian is
  positive-definite, $\forall  \w \in \cW, \ \nabla^2 L(\w) \succ 0$, i.e.,
  $\forall \w, \v  \in \cW$,  $\langle \v, \nabla^2 L(\w) \v \rangle > 0$.
\end{boxprop}

\subsection{Strongly-convex functions}

Convexity can also be strengthened by considering $\mu$-strongly convex
functions.
\begin{boxdef}{Strongly-convex functions}\label{optim:def:strongly_convex}
A function $L: \cW \rightarrow \RR$ is $\mu$-\textbf{strongly
convex} for $\mu >0$ if for all $\w, \v \in \cW$ and $\tau\in [0, 1]$
\[
  L(\tau\w + (1-\tau)\v) \leq \tau L(\w) + (1-\tau)L(\v) -
  \frac{\mu}{2} \tau (1-\tau) \|\w - \v\|^2_2.
\]
A differentiable function $L$ is $\mu$-strongly convex if and only if
for all $\w, \v \in \cW$
\[
  L(\v) \geq L(\w) + \langle \nabla L(\w), \w-\v\rangle 
  + \frac{\mu}{2} \|\w-\v\|_2^2.
\]
A twice differentiable function is $\mu$-strongly convex if and only if its
Hessian satisfies
\[
  \forall \w \in \cW,  \ \nabla^2 L(\w) \succeq \mu \idm, 
  \ \mbox{i.e.}, \ 
  \forall \w, \v  \in \cW, \ 
  \langle \v, \nabla^2 L(\w) \v \rangle \geq \mu\|\v\|_2^2.
\]
\end{boxdef}
The characterization of strong convexity for
differentiable functions states that $L(\w)$ is lower-bounded by a quadratic.
This enables the design of linearly convergent algorithms as explained later. 
We naturally have the implications
\[
\mbox{$L$ strongly convex} 
\implies \mbox{$L$ strictly convex} 
\implies \mbox{$L$ convex}.
\]

\subsubsection*{With respect to arbitrary norms}

\begin{figure}[t]
\centering
\includegraphics[width=0.7\linewidth]{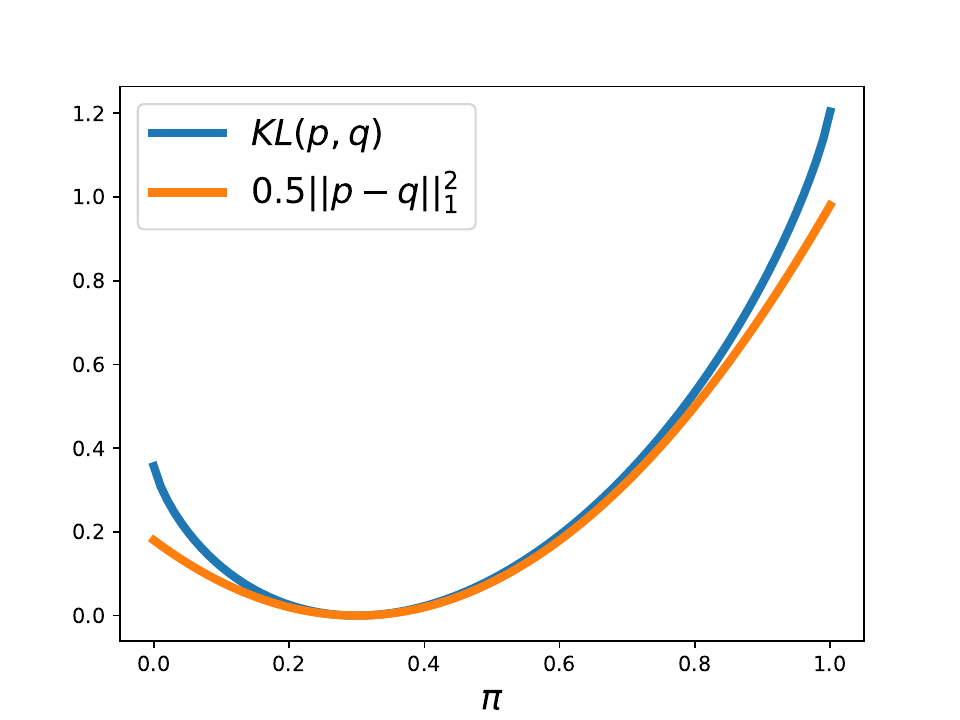}
\caption{Graphical verification of Pinsker's inequality,
$\frac{1}{2} \|\p - \q\|_1^2 \le \mathrm{KL}(\p, \q)$, 
with $\p \coloneqq (\pi, 1-\pi)$ and $\q \coloneqq (0.3, 0.7)$.}
\label{optim:fig:pinsker}
\end{figure}

A function can be strongly convex w.r.t. an arbitrary norm,
simply by replacing the $2$-norm in \cref{optim:def:strongly_convex} with that
norm. For differentiable strongly convex functions, we have the following 
alternative characterization,
generalizing \cref{optim:def:strongly_convex} to arbitrary norms.
\begin{boxprop}{Differentiable strongly-convex functions}
If a differentiable function $L:\cW \rightarrow \RR$ is $\mu$-strongly convex
w.r.t. a norm $\|\cdot\|$ over a set $\cC$, then for all $\w, \v \in \cC$
\begin{equation*}
\frac{\mu}{2} \|\w - \v\|^2 \le
\underbrace{L(\w) - L(\v) - \langle \nabla L(\v), \w - \v \rangle}_{B_L(\w,
\v)}.
\end{equation*}
\label{optim:prop:strongly_convex_any_norm}
\end{boxprop}
Obviously, if a function $L$ is $\mu$-strongly convex, then, $\lambda L$ is
$(\mu\lambda)$-strongly convex. Because all norms are equivalent, if a function
is strongly convex w.r.t. a norm, it is also strongly-convex w.r.t. another
norm. However, stating the norm w.r.t. which strong convexity holds can lead to
better constant $\mu$ (the higher, the better in terms of convergence rates of,
e.g., a gradient descent). We also emphasize that it is important to mention
over which set strong convexity holds. We give examples below.
\begin{boxexm}{Strongly convex functions}
The function $f(\u) = \frac{1}{2} \|\u\|^2_2$ is $1$-strongly convex 
w.r.t. $\|\cdot\|_2$ over $\RR^M$.

The function $f(\u) = \langle \u, \log \u \rangle$ is $1$-strongly convex w.r.t.
$\|\cdot\|_1$ over
$\triangle^M$.
Applying \cref{optim:prop:strongly_convex_any_norm},
we obtain for all $\p, \q \in \triangle^M$
\begin{equation*}
\frac{1}{2} \|\p - \q\|_1^2 \le B_f(\p, \q) = \mathrm{KL}(\p, \q), 
\end{equation*}
which is known as \textbf{Pinsker's inequality}.
We empirically verify the inequality in \cref{optim:fig:pinsker}.

More generally, $f(\u)$ is $\frac{1}{\mu}$-strongly convex w.r.t. $\|\cdot\|_1$
over any bounded set $\cC \subset \RR^M_+$ such that $\mu = \sup_{\u \in \cC}
\|\u\|_1$ \citep{blondel_2019}.  
However, it is not strongly convex over $\RR^M_+$, as it is not bounded.
\end{boxexm}

\subsection{Nonconvex functions}

\begin{figure}
	\begin{minipage}{0.58\linewidth}
		\begin{center}
			\includegraphics[width=\linewidth]{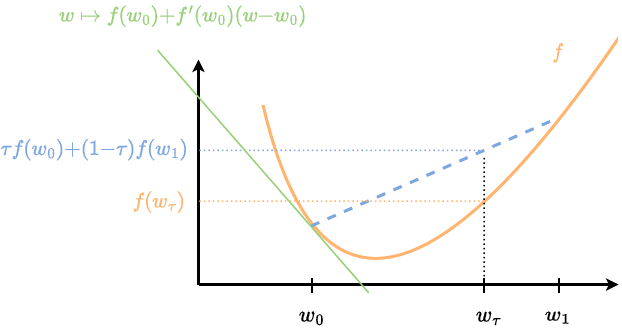}
		\end{center}
		\caption{Convex function: any secant is above the function, any tangent is
		below the function, a point with zero gradient is a minimum.
		\label{optim:fig:cvx}}
	\end{minipage}
	\begin{minipage}{0.38\linewidth}
        \vspace*{15pt}
		\begin{center}
			\includegraphics[width=\linewidth]{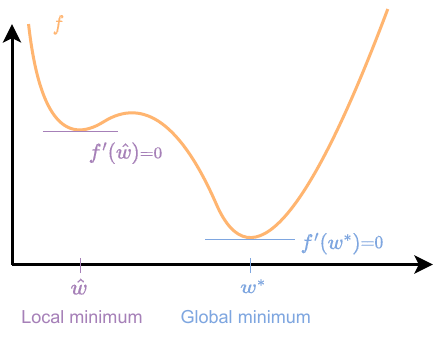}
		\end{center}
		\caption{Non-convex function: a point with zero gradient is not
        necessarily the global minimum. \label{optim:fig:noncvx}}
	\end{minipage}
\end{figure}

In general, the minimum of a function necessarily has a null gradient, that is, 
\[
  \w^\star \in \argmin_{\w \in \cW} L(\w) \implies \nabla L(\w^\star) = 0.
\]
To see this, consider the function $F: t \rightarrow L(\w^\star - t \nabla
L(\w^\star))$. If $\nabla L(\w^\star)\neq 0$, then $F'(0) = -\|\nabla
L(\w^\star)\|_2^2 \neq 0$. Therefore, there exists a small $t >0$ such that $F(t) <
F(0)$, i.e., $L(\w^\star)$ is not the minimum. However, if the function is not
convex, the converse is a priori not true: finding a point that has a null
gradient does not ensure that we have found a global minimum as illustrated in
Figure~\ref{optim:fig:noncvx}. 

For non-convex functions, a point with null gradient is called a
\textbf{stationary point}. A stationary point may define a {\bf local maximum} or a
{\bf local minimum}. Formally, $\hat \w$ is a local minimum if
\[
  \exists r>0, \ \mbox{s.t.}\ 
  \forall \v \in \cW\ \mbox{satisfying} \ \|\v-\hat \w\|\leq r, \ 
  \mbox{we have} \  L(\v) \geq L(\hat\w).  
\]
A local maximum is defined similarly, except that $ L(\v) \leq L(\hat\w)$ in a
neighborhood of $\hat \w$. For non-convex functions, convergence rates are
therefore generally expressed in terms of convergence of the norm of the
gradient $\|\nabla L(\w \pow t)\|_2$ towards $0$. Such theoretical results do not ensure
convergence to the global minimum but rather convergence to a point where no
further progress may a priori be possible with just gradient information.

\section{Performance guarantees}
\label{optim:sec:perf}

For a given class of functions, we can define the performance of an 
algorithm as the number of iterations the algorithm would need to find an
$\varepsilon$-accurate solution as in
\cref{optim:eq:epsilon_accurate}. This is called the \textbf{computational
complexity} of the algorithm, denoted
\begin{equation*}
t = T(\varepsilon). 
\end{equation*}
Alternatively, the performance of an algorithm can be stated in terms of
\textbf{convergence rate}, i.e., the accuracy that the algorithm reaches 
after $t$ iterations,
\begin{equation*}
\varepsilon = R(t),
\end{equation*}
where $R$ is a decreasing positive function vanishing as $t \rightarrow
+\infty$. Usually, $R$ incorporates properties of the function minimized, such
as its smoothness constant $\beta$ and information on the initial point, such as
its function value. The corresponding computational complexity $T(\varepsilon)$
is then given as the minimum number of iterations $t$ such that $R(t) \leq
\varepsilon$,
\begin{equation*}
T(\varepsilon) = \min \{t \in \NN \colon R(t) \le \varepsilon\}.
\end{equation*}

\begin{figure}[t]
\centering
\includegraphics[width=0.48\linewidth]{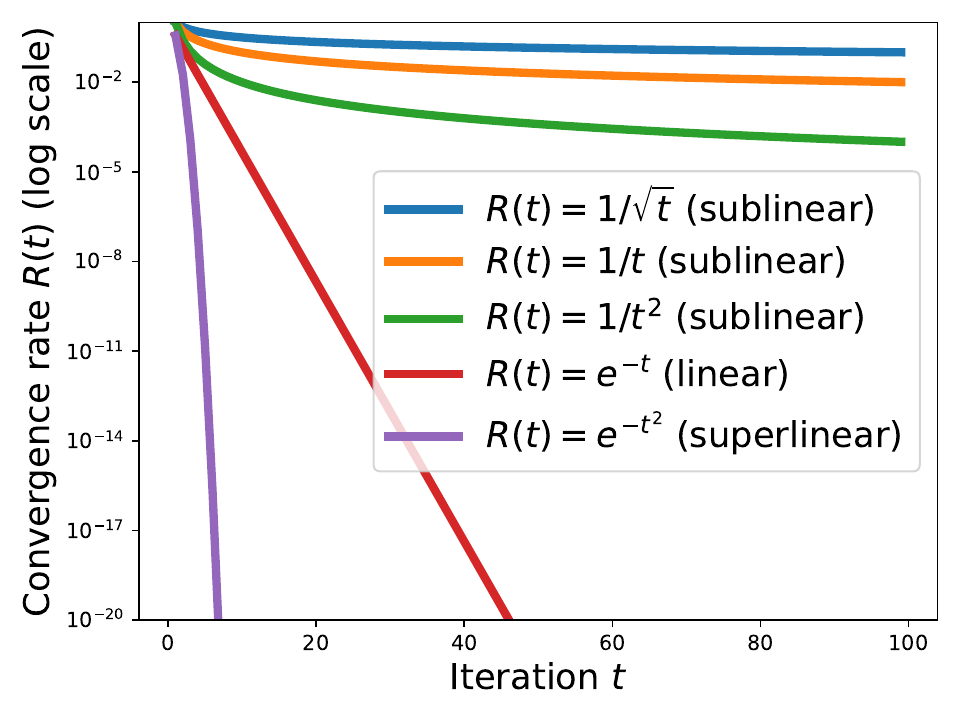}
\includegraphics[width=0.48\linewidth]{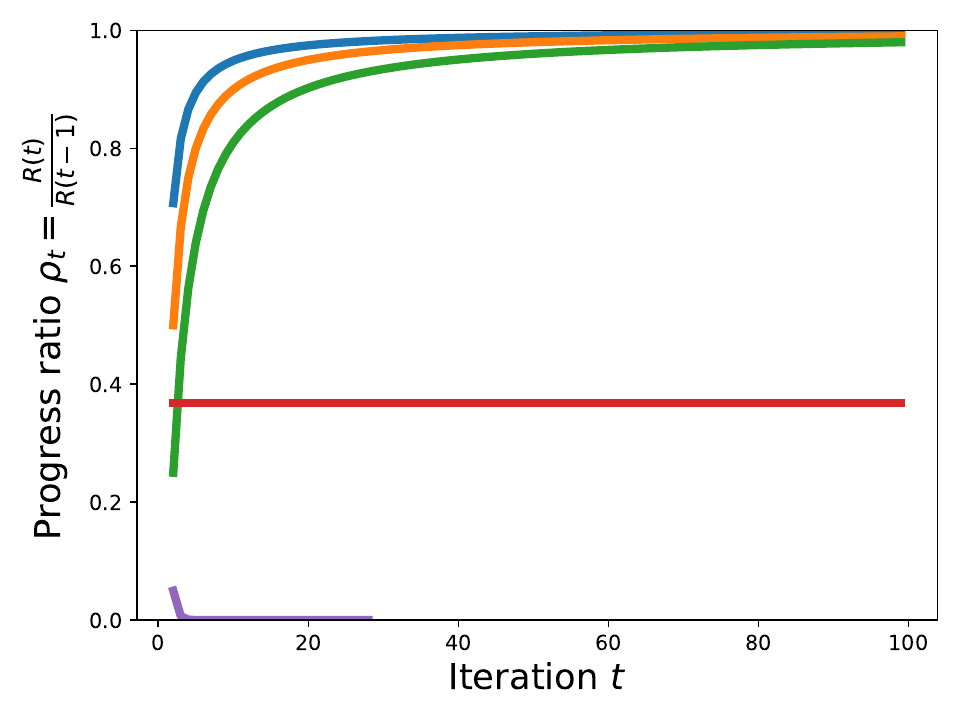}
\caption{{\bf Left:} convergence rates. {\bf Right:} progress ratios. An
    algorithm with sublinear convergence rates eventually stops
    making progress.
An algorithm with linear convergence rate eventually reaches a state of constant
progress.
An algorithm with superlinear convergence rate makes faster progress after each
iteration.
}
\label{optim:fig:rates}
\end{figure}

Convergence rates can generally be classified by considering the progress ratio
on iteration $t$, defined by
\begin{equation*}
\rho_t \coloneqq \frac{R(t)}{R(t-1)}.
\end{equation*}
The asymptotic convergence rate is then defined by
\[
\rho_{\infty} \coloneqq \lim_{t \rightarrow + \infty} \rho_t.
\]
We can classify the rates as follows.
\begin{enumerate}[nosep]

    \item \textbf{Sublinear convergence rates, $\rho_\infty =1$:} the longer the
        algorithm runs, the slower it makes progress. That is, the relative
    progress eventually tends to stall as $t \rightarrow + \infty$. Examples
    of $R(t)$ in this category include $O(1/t), O(1/t^2)$ or more generally
    $O(1/t^\alpha)$ for some $\alpha>0$. This is equivalent to 
    $T(\varepsilon) = O(\varepsilon^{-1/\alpha})$.

    \item \textbf{Linear convergence rates, $\rho_\infty = c \in (0, 1)$:} 
    the algorithm eventually reaches a state of constant relative progress at
    each iteration, leading to an overall
    rate $R(t) = O(\exp(- c t))$ for $c$ depending on the properties of the
    objective. This corresponds to $T(\varepsilon) = O (c^{-1}\ln \varepsilon^{-1})$.

    \item \textbf{Superlinear convergence rates, $\rho_\infty = 0$:} the
    relative progress is better at each new iteration. This can happen for,
    e.g., $R(t) = O(\exp(-t^2))$, leading to $T(\varepsilon) = O(
    \sqrt{\ln\varepsilon^{-1}})$, or $R(t) = O(\exp(-2^t))$, which is
    specifically called a \textbf{quadratic rate}, leading to 
    $T(\varepsilon) = O (\ln\ln\varepsilon^{-1})$.
\end{enumerate}
This is illustrated in \cref{optim:fig:rates}.

Note that the term ``linear'' may be misleading as the rates are in fact
exponential.  They are called ``linear'' because of their behavior in log scale.

\subsubsection*{Upper and lower bounds}

The best performance of a class of algorithms equipped with a given oracle (e.g.
first-order oracle) can be upper-bounded or lower-bounded. This allows to show
that an algorithm with access limited to a certain type of oracle cannot
theoretically do better than a certain number. For example, the computational
complexity to minimize $\beta$-smooth functions restricted on $[0, 1]^P$ with
first-order oracles is lower bounded by $\frac{c}{\varepsilon^P}$
~\citep[1.1.7]{nemirovski1983problem}. For example, with $P = 10$ and
$\varepsilon=10^{-3}$, this gives $10^{30}$ iterations. Note that these results
are pessimistic by construction. The actual performance of an algorithm on a
specific instance of this function class may be much better than this worst-case
scenario, as it is the case with popular algorithms such as quasi-Newton
methods. Better computational complexities can be achieved by further restricting
the class of functions to the set of convex functions, which play a
central role in optimization and many other fields.

\subsubsection*{Zero-order vs. first-order}

For the class of smooth strongly convex functions, the computational complexity
of the best first-order algorithm is (up to constant and logarithmic factors)
$P$ times better than that of the best zero-order algorithm
\citep{nesterov2018lectures, nesterov2017random}.
This theoretical comparison shows that, while zero-order optimization
algorithms may perform on par with first-order optimization algorithms for
problems with a low dimension $P$, they can be much slower for high dimensional
problems, i.e., $P\gg 1$.

In different settings, for example with stochastic
oracles~\citep{duchi2015optimal} or for different classes of functions, slightly
different comparisons may be achieved, such as a $\sqrt{P}$ factor instead of
$P$. However, the same conclusion holds in the current frameworks considered:
first-order optimization algorithms can provide fast rates that are dimension
independent while the rates of zero-order optimization algorithms generally
depend on the dimension of the problem, making them unfit for
high-dimensional problems. 

This explains the immense success of first-order algorithms for training neural
networks.  Fortunately, using reverse-mode autodiff, as studied in
\cref{chap:auto_diff}, it can be shown that computing a gradient has roughly the
same complexity as evaluating the function
itself~\cref{auto_diff:sec:complexity_dag}.

\section{Summary}

\begin{itemize}

\item The information available to us on a function can be formalized by the notion of
\textbf{oracle}. Zero-order oracles can only evaluate the function; first-order
oracles can also compute the gradient; second-order oracles can also compute the
Hessian or the Hessian-vector product (HVP).

\item Most optimization algorithms reviewed in this book can be viewed from a
\textbf{variational perspective}, in which the next iteration is produced by
optimizing a trade-off between an approximation of the function and a proximity
term. Different approximations and different proximity terms lead to different
algorithms.

\item We also reviewed different classes of functions, and performance guarantees.

\end{itemize}

%% file: chapters/optim/first_order.tex
\chapter{First-order optimization} \label{chap:optim}

\section{Gradient descent}
\label{optim:sec:gd}

\begin{figure}[t]
  \centering
  \includegraphics[width=0.49\linewidth]{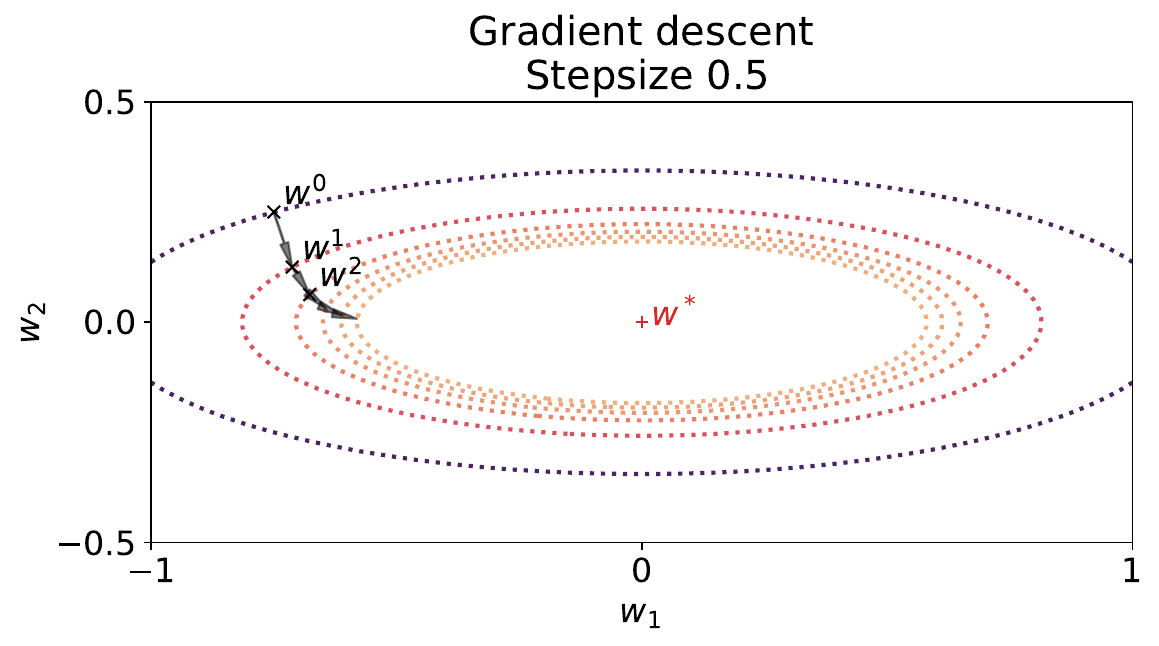}
  \includegraphics[width=0.49\linewidth]{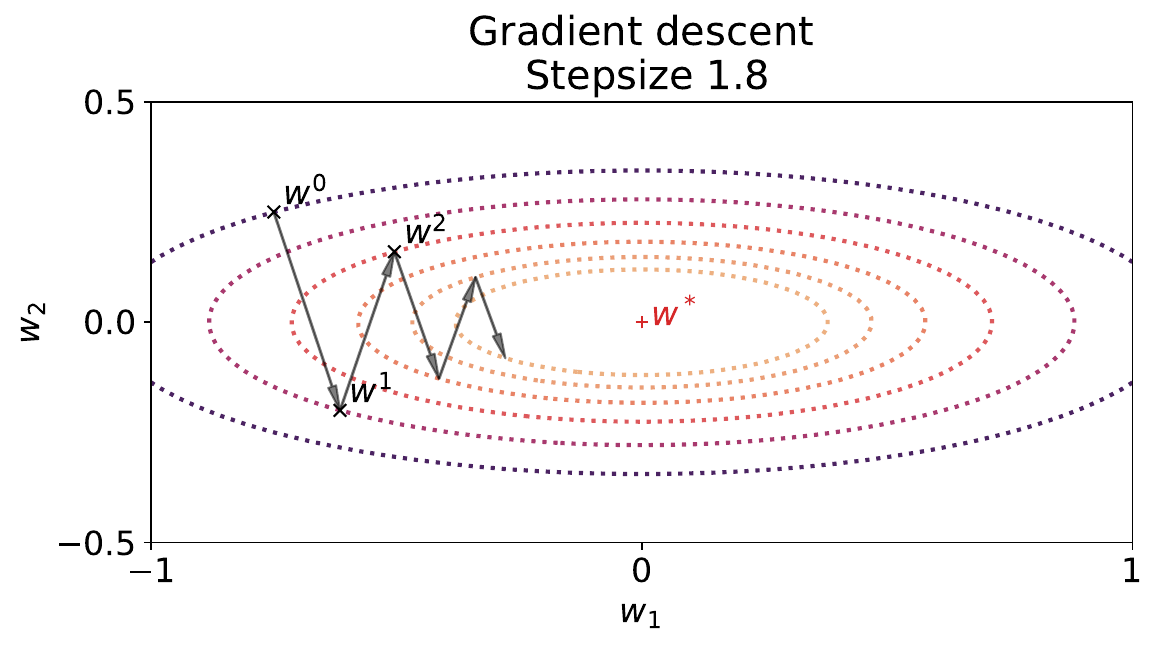}
  \caption{Trajectory taken by gradient descent on the objective $f(w) \coloneqq
  0.05w_1^2 + 0.5 w_2^2$ with a small (left) or large (right) stepsize. In each
  case, the iterates follow the normal vectors to the contour lines (dashed
  lines): the negative gradients. A small stepsize leads to slow
  convergence but a larger stepsize induces oscillations.
  \label{optim:fig:grad_descent}}
\end{figure}

Gradient descent is one of the simplest algorithms in our toolbox to minimize 
a function. At each iteration, it moves along the negative gradient direction, 
scaled by a stepsize $\gamma$: 
\begin{equation}
    \w \pow {t+1} = \w \pow t - \gamma \nabla L(\w\pow t). 
\label{optim:eq:gradient_descent}
\end{equation}
The path taken by a gradient descent on a simple quadratic is illustrated in
\cref{optim:fig:grad_descent} for different choices of the stepsize.

\subsection{Variational perspective}

Consider the linear approximation of $L(\w)$ around $\w \pow t$,
\begin{equation*}
L(\w) 
\approx 
L(\w \pow t) + \langle \nabla L(\w \pow t), \w - \w \pow t \rangle.
\end{equation*}
One can easily check that the gradient descent update
in \cref{optim:eq:gradient_descent} can be rewritten
as the solution of a minimization problem, namely,
\begin{equation}
\w \pow {t+1} 
= \argmin_{\w \in \cW} 
L(\w \pow t) + \langle \nabla L(\w \pow t), \w - \w \pow t \rangle 
+ \frac{1}{2\gamma} \|\w - \w \pow t\|_2^2.
\label{optim:eq:gd_variational}
\end{equation}
In words, a gradient descent update optimizes a trade-off between staying close
to the current $\w \pow t$, thanks to the proximity term $\frac{1}{2\gamma} \|\w
- \w \pow t\|_2^2$, and minimizing the linearization of $L$ around $\w \pow t$.
Intuitively, by choosing $\gamma$ sufficiently small, we ensure that the
minimizer of the regularized linear approximation stays in a neighborhood where
the linear approximation is valid. This viewpoint is useful to motivate
gradient descent extensions.

\subsection{Convergence for smooth functions}

As long as $\nabla L(\w\pow t) \neq \zeros$, the function 
$L_t(\gamma) \coloneqq L(\w \pow
t - \gamma \nabla L(\w\pow t))$ has a negative derivative at 0, i.e.,
$L'_t(0)=-\|\nabla L(\w\pow t)\|_2^2$. Hence, as long as $\nabla L(\w\pow t) \neq
\zeros$, there exists a stepsize ensuring a decrease in objective values at each
iterate. However, without further assumptions, such a stepsize may depend on
each iterate and may be infinitesimally small. To quantify the convergence of 
gradient descent with a constant stepsize, we restrict to the class of smooth
functions. By applying \cref{optim:prop:smoothness} on the iterate of gradient
descent, we obtain that 
\[
  L(\w\pow {t+1}) 
  \leq L(\w\pow t) 
  - \gamma \|\nabla L(\w\pow t)\|_2^2
  + \frac{\beta\gamma^2}{2} \|\nabla L(\w \pow t)\|_2^2.
\]
Therefore, for $\beta$-smooth functions, 
by selecting $\gamma \leq \frac{1}{\beta}$, we get that
\[
  L(\w\pow {t+1}) - L(\w\pow t) \leq 
  - \frac{\gamma}{2}\|\nabla L(\w\pow t)\|_2^2,
\]
which illustrates the main mechanism behind gradient descent: each iteration
decreases the objective by a constant times the norm of the gradient of the
current iterate. This equation can further be summed over all iterates up to
$T$. This telescopes the objective values, leading to
\begin{align*}
  \min_{t\in \{0, \ldots, T-1\}} \|\nabla L(\w \pow t)\|_2^2 
  & \leq \frac{1}{T}\sum_{t=0}^{T-1} \|\nabla L(\w\pow t)\|_2^2 \\
  & \leq \frac{2}{\gamma T}\left(L(\w \pow 0) - L(\w \pow {T})\right) \\
  & \leq \frac{2}{\gamma T}\left(L(\w \pow 0) - L^\star\right),
\end{align*}
where we recall that $L^\star$ is the infimum of $L$. Therefore, after
sufficiently many iterations, gradient descent finds a point whose gradient norm
is arbitrarily small.

\subsubsection*{Non-convex case}

Without further assumptions, i.e., in the non-convex
case, the above result (i.e., convergence to a stationary point, measured by
the gradient norm) is the best we may get in theory.
Denoting $T_s(\varepsilon)$ the number of iterations needed for a gradient
descent to output a point that is $\varepsilon$-stationary, i.e., $\|\nabla
L(\hat \w)\|_2\leq \varepsilon$, we have $T_s(\varepsilon) \leq
O(\varepsilon^{-2})$.

\subsubsection*{Convex case}

By adding a convexity assumption on the objective, we can use the lower bound
provided by the convexity assumption to ensure convergence to a minimum. Namely,
for a $\beta$-smooth and convex function $f$, and with stepsize 
$\gamma \leq 1/\beta$, we have that~\citep{nesterov2018lectures}
\[
    L(\w \pow T) - L^\star 
    \leq \frac{1}{\gamma T}\|\w \pow 0 - \w^\star\|_2^2.
\]
That is, we get a sublinear
convergence rate, and the associated computational complexity to find a minimum
is $T(\varepsilon)= O(1/\varepsilon)$.

\subsubsection*{Strongly convex case}

If we further strengthen the assumptions by considering $\beta$-smooth,
$\mu$-strongly convex functions, the convergence rate of a gradient descent can
be shown to be~\citep{nesterov2018lectures}, for any stepsize $\gamma \leq 1/\beta$,
\begin{align*}
  L(\w \pow T) - L^\star
  & \leq \left(1 - \gamma \mu\right)^T 
  \left(L(\w\pow 0) - L^\star\right) \\
  & \leq \exp\left(- \gamma \mu T\right)
  \left(L(\w\pow 0) - L^\star\right).
\end{align*}
That is, we obtain a linear convergence rate and the associated computational
complexity is $T(\varepsilon) = O(\ln\varepsilon^{-1})$. The above convergence
rates may be further refined~\citep{nesterov2018lectures}; we focused above on
the simplest result for clarity.

Strong convexity can also be replaced by
a weaker assumption, gradient-dominating property~\citep{polyak1963gradient}, i.e., 
$\|\nabla L(\v)\|_2^2 \geq c(L(\v) - L^\star)$ for some constant $c$
and any $\v \in \cW$. A convex, gradient-dominating function can also be
minimized at a linear rate.

\subsection{Momentum and accelerated variants}

\begin{figure}[t]
  \centering
  \includegraphics[width=0.49\linewidth]{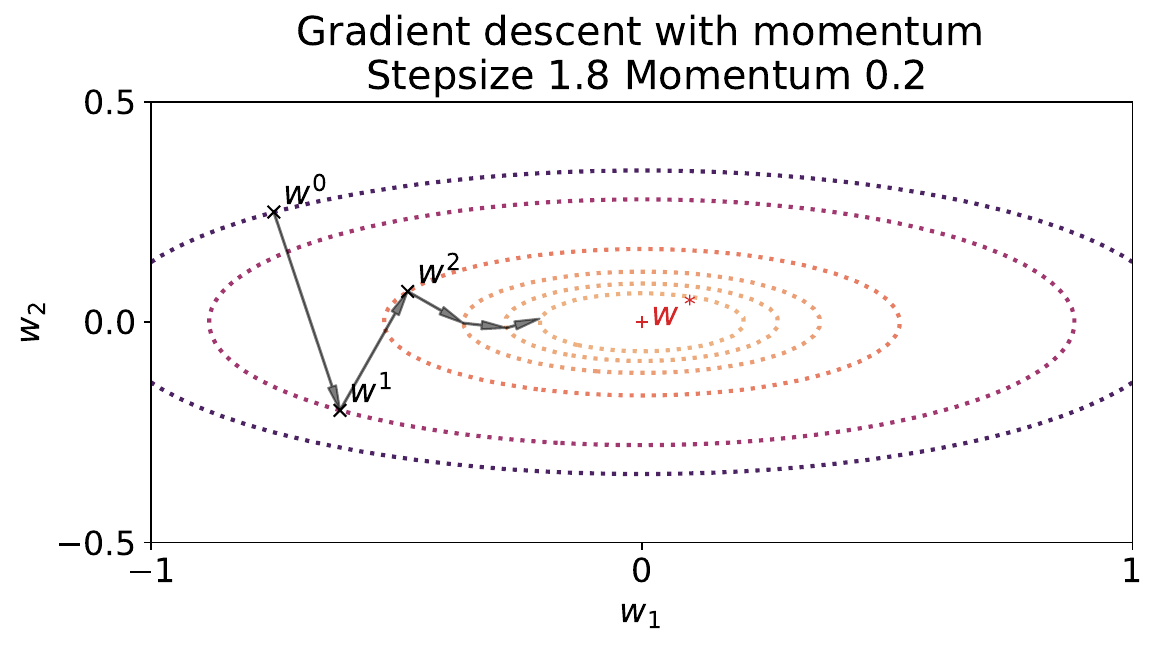}
  \caption{Trajectory taken by gradient descent with momentum. Compared to
  gradient descent without momentum, for the same stepsize, the oscillations
  previously observed in \cref{optim:fig:grad_descent} are no longer present,
  and the algorithm converges faster to the minimum.
  \label{optim:fig:grad_descent_mom}
  }
\end{figure}

We started with gradient descent as a simple example of first-order
optimization algorithm. However, different optimization algorithms can be
designed from the access to first-order oracles and the knowledge of the class
of functions considered. For example, consider quadratic convex functions $\w
\mapsto \frac{1}{2} \w^\top \A \w + \b^\top \w$, that are a basic example of
smooth strongly convex functions if $\A$ is positive definite. An optimal method
in this case is the heavy-ball method of \citet{polyak1964some}, that can be
written as
\begin{align*}
  \v \pow {t+1} &\coloneqq \nu \v \pow t - \gamma \nabla L(\w \pow t) \\
  \w \pow {t+1} &\coloneqq \w \pow t + \v \pow {t+1}.
\end{align*}
The heavy-ball method uses an additional variable $\v\pow t$, that can be
interpreted as the velocity of a ball driven by the negative gradient to
converge towards a minimum. Intuitively, this additional velocity circumvents
the oscillations that a gradient descent may present as illustrated in
\cref{optim:fig:grad_descent_mom} compared to \cref{optim:fig:grad_descent}.
For $\nu=0$, we recover usual gradient
descent. For $\nu>0$, the velocities accumulate a form of an inertia
momentum, where $\nu$ is interpreted as the ``mass'' of the ball. In
terms of convergence rates, the heavy-ball method can be shown to converge 
linearly similarly to gradient descent, 
but with a rate $O(\exp(- T \sqrt{\mu /\beta}))$ for appropriate
choices of $\nu, \gamma$. In comparison, by choosing an optimal stepsize
for the gradient descent, its convergence rate is $O(\exp(-T\mu/\beta))$ which is
provably worse, as we always have $\mu/\beta \leq 1$.

Beyond the case of quadratic functions, accelerated variants of gradient descent
for convex or strongly convex functions have been developed
by~\citet{nesterov2018lectures}. Such variants have inspired the design of
optimization algorithms in stochastic settings presented below.

\section{Stochastic gradient descent}
\label{optim:sec:sgd}

In machine learning, we are usually interested in minimizing
the \textbf{expected loss} of the model over the data distribution $\rho$:
\[
\min_{\w \in \cW} \ 
L(\w) \coloneqq \mathbb{E}_{S \sim \rho} \left[L(\w; S)\right].
\]
For example, $L$ is often set to 
$L(\w; S) \coloneqq \ell(Y, f(X, \w))$, 
where $\ell$ is a loss function,
$f$ is a neural network and 
$S=(X, Y)$ is a random pair, composed of an input $X$ and an associated
target $Y$, sampled from $\rho$.
In this setting, since the data distribution $\rho$ is generally unknown and may be
infinite, we cannot exactly
evaluate the expected loss $L(\w)$ or its gradient $\nabla L(\w)$.

In practice, we are often given a fixed
dataset of $n$ pairs $\s_i = (\x_i, \y_i)$. 
This is a special case of the expected loss setting, 
since this can be seen as an empirical distribution $\rho = \rho_n$
\[
L(\w)
= \mathbb{E}_{S \sim \rho_n} \left[L(\w; S)\right] 
= \frac{1}{n} \sum_{i=1}^n L(\w; (X_i, Y_i)).
\]
The gradient of $L(\w)$ is then
\begin{equation*}
\nabla L(\w)
\coloneqq
\frac{1}{n} \sum_{i=1}^n \nabla L(\w; (\x_i, \y_i)).
\end{equation*}
In this case, we see that the \textbf{full} gradient $\nabla L(\w)$, as needed
by gradient descent, is the average of the \textbf{individual} gradients. That
is, the cost of computing $\nabla L(\w)$ is proportional to the number of
training points $n$. For $n$ very large, that is a very large amount of samples,
this computational cost can be prohibitive. Stochastic gradients circumvent this
issue.

\subsection{Stochastic gradients}

Usually, even if we do not know $\rho$, we can sample from it, i.e.,
we have access to samples $S \sim \rho$. 
We can then use a \textbf{stochastic gradient} of the form
$
  \nabla L(\w; S)
$
as a random estimate of $\nabla L(\w)$. 
This may look like a rough estimate
but, on average, this is a valid approximation since
\[
  \mathbb{E}_{S \sim \rho}\left[ \nabla L(\w; S)\right] = \nabla L(\w).
\] 
We say that $\nabla L(\w; S)$ is an 
\textbf{unbiased estimator} of $\nabla L(\w)$.
To further improve the approximation, we may also consider 
\textbf{mini-batch} estimates by sampling $m \ll n$
data points $S_i \coloneqq (X_i, Y_i)$ and using 
$\frac{1}{m} \sum_{i=1}^m \nabla L(\w;S_i)$, 
whose expectation still matches $\nabla L(\w)$,
while potentially reducing the approximation error by averaging multiple
stochastic gradients. Computationally, the main advantage is that the cost is
now proportional to $m$ instead of $n$.

In whole generality, one can consider stochastic first-order oracles defined
below.
\begin{boxdef}{Stochastic first-order oracles}
A \textbf{stochastic first-order oracle} of an expected objective $L(\w)$ is a
random estimate $g(\w; S)$ of $\nabla L(\w)$ with $S$ sampled
according to some distribution $q$. 
A stochastic gradient is said to be an \textbf{unbiased estimator} if 
\[
\mathbb{E}_{S \sim q}\left[g(\w; S)\right] = \nabla L(\w).
\]
The \textbf{variance} of a stochastic gradient is
  \[
    \mathbb{E}_{S \sim q}\left[\left\| g(\w; S) 
    - \nabla L(\w) \right\|_2^2\right].
  \]
\end{boxdef}
When $q = \rho$, we recover stochastic gradients.
When $q$ is the product of $m$ independent samples according to $p$, we recover
mini-batch stochastic gradients. 
First-order stochastic optimization algorithms build upon
stochastic first-order oracles to approximately find the minimum of the expected
objective. In such a setting, the iterates of the algorithm are by definition
random.
Convergence rates therefore need to be expressed in probabilistic terms
by considering for example the expected objective value according to the
randomness of the oracles.

\subsection{Vanilla SGD}

Equipped with a stochastic first-order oracle, such as (mini-batch) 
stochastic gradients,
we can define \textbf{stochastic gradient descent} as
\[
\w \pow {t+1} = \w \pow t - \gamma g(\w \pow t; S \pow t) 
  \quad \mbox{where} \
  S \pow t \sim q.
\]
We assume that $S \pow t$ is independent of $\w \pow t$.
Compared to the usual gradient descent, the main impediment of the stochastic
setting is the additional noise induced by the stochastic estimates:
their variance. 

For example, consider applying a stochastic gradient descent on the expectation
of $\beta$-smooth convex functions $L(\w; \s)$ with unbiased oracles. To harness
the randomness of the iterates, consider after $T$ iterations outputting the
average of the first $T$ iterates, that is $\bar \w \pow T \coloneqq \frac{1}{T}
\sum_{t=1}^T \w\pow t$. Moreover, suppose that the variance of the stochastic
first-order oracles is bounded by $\sigma^2$ for all minimizers $\w^\star$ of
$L$. Denoting by $\mathbb{E}_{S_0, \ldots, S_{T-1}}$ the randomness associated
to the stochastic oracles, we have then that for a stepsize $\gamma \leq
1/(4\beta)$ ~\citep{lan2012optimal},
\[
  \mathbb{E}_{S_0, \ldots, S_{T-1}}[ L(\bar \w \pow T)] 
  - L^\star \\
  \leq \frac{1}{\gamma T}\|\w_0 - \w^\star\|_2^2 + 2\gamma \sigma^2.
\]
The resulting convergence rate illustrates that a stochastic gradient descent
converges to the minimum of the expected objective up to a constant term
depending on the variance of the oracle and the stepsize. One can diminish the
variance by considering mini-batches: if the variance of a single stochastic
gradient is $\sigma_1^2$, considering a mini-batch of $m$ gradients reduces the
variance of the corresponding oracle to $\sigma^2_m = \sigma_1^2/m$. To decrease
the additional term, one may also decrease the stepsizes over the iterations.
For example,
by choosing a decreasing stepsize like $\gamma^t = t^{-1/2}$, the convergence
rate is then of the order $O((\|\w_0-\w^\star\|_2^2 + \sigma^2\ln t)/\sqrt{t})$.
The stepsize can also be selected as a constant $\gamma_0$ that decreases the
average objective for the first $T_0$ iterations and is reduced by a
multiplicative factor at regular intervals like $\gamma_j = \rho \gamma_{j-1}$
for $\rho \in (0, 1)$ to handle iterations between $T_j, T_{j+1}$. Alternative
stepsize schedules such as a cosine decay \citep{loshchilov2016sgdr} have
recently become popular.

The literature on alternative optimization schemes for stochastic
optimization is still rapidly evolving, with new heuristics regularly proposed.
We present below two popular techniques.

\subsection{Momentum variants}

Accelerated optimization algorithms developed in the deterministic setting may
be extended to the stochastic setting. For example, the heavy-ball method can be
adapted to the stochastic setting, leading to stochastic gradient descent with
\textbf{momentum}~\citep{sutskever2013importance} generally implemented as
\begin{align*}
  \v \pow {t+1} &\coloneqq \nu \v\pow t + \g(\w \pow t; S \pow t) \\
  \w \pow {t+1} &\coloneqq \w \pow t - \gamma \v \pow {t+1}. 
\end{align*}
As mentioned earlier the momentum method can be modified to handle non-quadratic
smooth strongly convex functions. This leads to Nesterov's accelerated method in
the deterministic setting. This has been adapted to the stochastic setting with
a so-called \textbf{Nesterov momentum}~\citep{sutskever2013importance}
\begin{align*}
  \v \pow {t+1} &\coloneqq \nu \v\pow t 
  + \g(\w \pow t - \gamma \nu \v\pow t; S \pow t) \\
  \w \pow {t+1} &\coloneqq \w \pow t -  \gamma \v \pow {t+1}.
\end{align*}

\subsection{Adaptive variants}

In any gradient descent-like algorithm, selecting the stepsize is key for good
performance. While a constant stepsize may be used if the function is smooth, we
may not know in advance the smoothness constant of the objective, which means
that additional procedures may be required to select appropriately the stepsize.
In the deterministic case, line-searches such as the Armijo or Wolfe's
rules~\citep{wright1999numerical} can be used to check whether the selected
stepsize sufficiently decreases the objective at each iteration. Such rules have
been adapted in the stochastic setting~\citep{vaswani2019painless}.

Another way to decrease the sensitivity of the algorithm with respect to the
stepsize has been to estimate first- and second-order moments of the gradients
and use the latter as a form of preconditioning to smooth the trajectory of the
iterates. This led to the popular \textbf{Adam}
optimizer~\citep{kingma2014adam}. It takes the form, 
\begin{align*}
\m\pow {t+1} &\coloneqq \nu_1  \m \pow t + (1-\nu_1) \g \pow t \\
\v\pow {t+1} &\coloneqq \nu_2  \v \pow t + (1-\nu_2)  (\g\pow t)^2 \\
\hat{\m}\pow t &\coloneqq \m\pow t / {(1-\nu_1^t)} \\
\hat{\v}\pow t &\coloneqq \v\pow t / {(1-\nu_2^t)} \\
\w \pow {t+1} &\coloneqq \w \pow t - \gamma \ 
\hat{\m}\pow {t+1} / \left({\sqrt{\hat{\v}\pow {t+1}} + \varepsilon} \right),
\end{align*}
where
$\g \pow t \coloneqq \g(\w\pow t; S \pow t)$,
$(\g\pow t)^2$ denotes the element-wise square of $\g \pow t$ and 
$\nu_1, \nu_2, \gamma, \varepsilon$ are hyper-parameters of the algorithm. 
Numerous variants exist, such as varying the stepsize $\gamma$ above along the
iterations.

\section{Projected gradient descent}
\label{optim:sec:proj_grad}

Oftentimes, we seek to find the solution of a minimization problem subject to
\textbf{constraints} on the variables, of the form
\begin{equation}
\min_{\w \in \cC} L(\w),
\label{optim:eq:constrained_pb}
\end{equation}
where $\cC \subseteq \cW = \RR^P$ is a set of constraints. 
We say that an approximate solution $\widehat \w$
to \cref{optim:eq:constrained_pb}
is \textbf{feasible} if $\widehat \w \in \cC$.
Naturally, the design of algorithms for the constrained setting now depends,
not only on information about $L$, but also on information about $\cC$.

Similarly to $L$, different \textbf{oracles} can be considered about $\cC$.
One of the most commonly used oracles is the \textbf{Euclidean projection} 
\begin{boxdef}{Euclidean projection}
The Euclidean projection onto the set $\cC$ is defined by
\[
\mathrm{proj}_{\cC}(\w) \coloneqq \argmin_{\v \in \cC} \|\w-\v\|_2^2.
\]
\label{optim:def:projection}
\end{boxdef}
This projection, which is well-defined when $\cC$ is a convex set,
can be used in projected gradient descent, that we briefly review below.
Typically, the projection on a particular set $\cC$ requires a dedicated
algorithm to compute it.

Other possible oracles are \textbf{linear maximization oracles} (LMO) used in
Frank-Wolfe algorithms and \textbf{Bregman projection oracles}, used in mirror
descent algorithms.  The algorithm choice can be dictated by what oracle about
$\cC$ is available.

\subsection{Variational perspective}

Projected gradient descent is a natural generalization of gradient descent,
based on the Euclidean projection oracle. Its iterates read 
\[
\w \pow {t+1} \coloneqq \mathrm{proj}_{\cC}(\w \pow t - \gamma \nabla L(\w \pow t)).
\]
At each iteration, we attempt to decrease the objective by moving along the
negative gradient direction, while ensuring that the next iterate remains
feasible, thanks to the projection step. 

Similarly to the variational perspective of gradient descent in
\cref{optim:eq:gd_variational}, the projected gradient descent update is
equivalent to
\begin{equation*}
\w \pow {t+1} 
= \argmin_{\w \in \cC} 
L(\w \pow t) + \langle \nabla L(\w \pow t), \w - \w \pow t \rangle 
+ \frac{1}{2\gamma} \|\w - \w \pow t\|_2^2.
\end{equation*}
This shows that projected gradient descent minimizes a trade-off between staying
close to $\w \pow t$ and minimizing the linearization of $L$ around $\w \pow t$,
while staying in $\cC$.

In terms of convergence rates, 
they remain the same as gradient descent \citep{nesterov2018lectures}. 
For example, projected gradient descent on a smooth convex function still
converges at a rate $R(T) = O(1/T)$. 

There are numerous extensions of vanilla projected gradient descent. Similarly
to gradient descent, the stepsize can be automatically adjusted using linesearch
techniques and there exists accelerated variants. If we replace $\nabla L(\w)$
with a stochastic gradient $\nabla L(\w; S)$, we obtain a stochastic projected
gradient descent.

\subsection{Optimality conditions}

In the unconstrained case, a minimum necessarily has a zero gradient. In the
constrained setting, there may not be any feasible parameters with zero
gradient. Instead, the optimality of a point is characterized by the fact that
no better solution can be found by moving along the gradient at that point,
while staying in the constraints. Formally, it means that for any $\gamma > 0$,
a minimizer $\w^\star$ of $L$ on $\cC$ satisfies
\[
\w^\star = \mathrm{proj}_{\cC}(\w^\star - \gamma \nabla L(\w^\star)).
\]
It can be shown that this condition is equivalent \citep{nesterov2018lectures}
to
\begin{equation*}
\langle \nabla L(\w^\star), \w - \w^\star \rangle \geq 0 
\quad \forall \w \in {\cC}.
\end{equation*}

\subsection{Commonly used projections}

We now briefly review a few useful Euclidean projections.
\begin{itemize}
    \item If $\cC = \RR^P$, we obviously have
\begin{equation*}
\mathrm{proj}_{\cC}(\w) = \w.
\end{equation*}
Therefore, in the unconstrained setting, projected gradient descent indeed
recovers gradient descent.

\item If $\cC = [a, b]^P$ (box constraints), we have
\begin{equation*}
\mathrm{proj}_{\cC}(\w) 
= \mathrm{clip}(\w, a, b)
\coloneqq \min\{\max\{\w, a\}, b\}.
\end{equation*}
where the $\min$ and $\max$ are applied coordinate-wise.

\item As a special case of the above, if $\cC = \RR_+^P$ (non-negative orthant),
\begin{equation*}
\mathrm{proj}_{\cC}(\w) 
= \max\{\w, 0\},
\end{equation*}
also known as non-negative part or ReLU.

\item If $\cC = \triangle^P$ (unit probability simplex),
\begin{equation*}
\mathrm{proj}_{\cC}(\w) 
= \max\{\w - \tau \ones, 0\},
\end{equation*}
where $\tau \in \RR$ is a constant ensuring that 
$\mathrm{proj}_{\cC}(\w)$ normalizes to $1$. 
It is known that $\tau$ can be found in $O(P \log P)$ using a sort.
This can be improved to $O(P)$ using a median-finding like algorithm.
\end{itemize}

\section{Proximal gradient method}
\label{optim:sec:prox_grad}

The constrained setting (with $\cC$ a convex set) 
can be recast as unconstrained optimization,
by extending our analysis to functions taking infinite
values. Let us denote the indicator function of the set $\cC$ by
\begin{equation*}
\indic_{\cC}(\w)
\coloneqq
\begin{cases}
0 &\mbox{if } \w \in \cC \\
+\infty &\mbox{otherwise}
\end{cases}.
\end{equation*}
Clearly, the constrained problem in \cref{optim:eq:constrained_pb} 
can then be rewritten as
\[
\min_{\w \in \cW} L(\w) + \indic_{\cC}(\w).
\]
This suggests that constrained optimization is a special case
of \textbf{composite objectives} of the form
\[
\min_{\w \in \cW} L(\w) + \Omega(\w),
\]
where $\Omega$ is a convex but potentially non-differentiable function.
We assume that we have access to an oracle associated with $\Omega$
called the \textbf{proximal operator}.
\begin{boxdef}{Proximal operator}
The proximal operator associated with $\Omega \colon \cW \to \RR$ is
\[
\mathrm{prox}_{\Omega}(\w) 
\coloneqq \argmin_{\v \in \cW} \frac{1}{2} \|\w -\v\|_2^2 + \Omega(\v).
\]
\label{optim:def:prox}
\end{boxdef}
This leads to the proximal gradient method, reviewed below.

\subsection{Variational perspective}

With this method, the update reads
\[
\w \pow {t+1} = 
\mathrm{prox}_{\gamma \Omega}(\w \pow t - \gamma \nabla L(\w \pow t)).
\]
This update again enjoys an intuitive variational perspective, namely,
\begin{equation*}
\w \pow {t+1} 
= \argmin_{\w \in \cW} 
L(\w \pow t) + \langle \nabla L(\w \pow t), \w - \w \pow t \rangle 
+ \frac{1}{2\gamma} \|\w - \w \pow t\|_2^2 + \Omega(\w).
\end{equation*}
That is, we linearize $L$ around $\w \pow t$, but keep $\Omega$ as is.

The proximal gradient method is popularly used when the objective function
contains a sparsity-inducing regularizer $\Omega$. For example, for the
LASSO~\citep{tibshirani1996regression}, 
which aims at predicting targets $\y = (y_1, \ldots, y_n)^\top \in \RR^N$ 
from observations $\X = (\x_1, \ldots, \x_n)^\top\in \RR^{N\times P}$, 
we set $L(\w) = \frac{1}{2}\|\X \w - \y\|_2^2$
and $\Omega(\w) = \lambda \|\w\|_1$, where $\lambda > 0$ controls the
regularization strength. In this case, $\mathrm{prox}_{\Omega}$ is
the so-called soft-thresholding operator (see below).

Convergence guarantees of the
proximal gradient method remain the same as for gradient descent,
such as a $O(1/T)$ rate for smooth convex functions.

\subsection{Optimality conditions}

An optimal solution of the problem is characterized by 
the \textbf{fixed point} equation
\[
\w^\star = \mathrm{prox}_{\gamma \Omega}(\w^\star - \gamma \nabla L(\w^\star)), 
\]
for all $\gamma > 0$~\citep{nesterov2018lectures}.  In other words, the proximal
gradient method (which includes gradient descent and projected gradient descent
as special cases), can be seen as fixed point iteration schemes. Such a
viewpoint suggests using acceleration methods from the fixed point literature
such as Anderson acceleration~\citep{pollock2021anderson}. It is also
useful when designing implicit differentiation schemes as presented in
\cref{chap:auto_diff}. 

\subsection{Commonly-used proximal operators}

We now briefly review a few useful proximal operators.
\begin{itemize}
    \item If $\Omega(\w) = 0$, we have
\begin{equation*}
\mathrm{prox}_{\gamma\Omega}(\w) = \w.
\end{equation*}
Therefore, with this proximal operator, the proximal gradient method recovers 
gradient descent.

    \item If $\Omega(\w) = \indic_{\cC}(\w)$, we have
\begin{equation*}
    \mathrm{prox}_{\gamma \Omega}(\w) = \mathrm{proj}_{\cC}(\w).
\end{equation*}
Therefore, with this proximal operator, the proximal gradient method recovers 
projected gradient descent.

    \item If $\Omega(\w) = \lambda \|\w\|_1$, we have
\begin{equation*}
\mathrm{prox}_{\gamma \Omega}(\w) 
= (\mathrm{sign}(\w) \cdot \max(|\w| - \gamma \lambda, 0)),
\end{equation*}
where the operations are applied coordinate-wise. This is the so-called
soft-thresholding operator.

    \item $\Omega(\w) = \lambda \sum_{g \in G} \|\w_g\|_2$ where $G$ is a
        partition of $[P]$ and $\w_g$ denotes the subvector restricted to $g$, 
        then we have
\begin{equation*}
\left[\mathrm{prox}_{\gamma \Omega}(\w)\right]_g
= 
\max(1 - \lambda \cdot \gamma / \|\w_g\|_2, 0) \w_g,
\end{equation*}
which is used in the group lasso \citep{yuan_2006}
and can be used to encourage group sparsity.

\end{itemize}
For a review of more proximal operators, see for instance 
\citep{bach_2012,parikh_2014}.

\section{Summary}

\begin{itemize}

\item From a variational perspective,
gradient descent is the algorithm obtained
when linearizing the objective function and using a quadratic regularization
term.

\item Projected gradient descent is the algorithm obtained when there is an additional
constraint (the Euclidean projection naturally appearing, due to the quadratic
regularization term).

\item When the objective is the sum of a differentiable function and a
non-differentiable function, proximal gradient is the algorithm obtained when
the differentiable function is linearized but the non-differentiable function is
kept as is. 

\item We also reviewed various stochastic gradient based algorithms,
including vanilla SGD, SGD with momentum and Adam.

\end{itemize}

%% file: chapters/optim/second_order.tex
\chapter{Second-order optimization}\label{chap:optim2}

We review in this chapter methods whose iterations take the form 
\[
  \w^{t+1} \coloneqq \w^t - \gamma^t B^t \nabla L(\w^t),
\]
where $\gamma^t$ is a stepsize and $B^t$ is a pre-conditioning matrix involving
second-order derivatives.

\section{Newton's method}
\label{optim:sec:newton}

\subsection{Variational perspective}

We saw in \cref{optim:eq:gd_variational} that gradient descent can be motivated
from a variational perspective, in which we use a linear approximation of the
objective around the current iterate, obtained from the current gradient. 
Similarly, if we have access not only to
the gradient but also to the Hessian of the objective, we can use a quadratic
approximation of the objective around the current iterate.  
More precisely, given a function $L(\w)$, we
may consider minimizing the second-order Taylor approximation of $L(\w)$ around
the current iterate $\w^t$,
\[
L(\w) \approx L(\w^t) + \langle
\nabla L(\w^t), \w-\w^t\rangle + \frac{1}{2}\langle \w-\w^t, \nabla^2 L(\w^t)(\w
-\w^t) \rangle. \]
\textbf{Newton's method} simply iteratively minimizes this quadratic
approximation around the current iteration $\w^t$, namely,
\begin{align}
  \w^{t+1} 
  = \argmin_{\w \in \cW} 
  L(\w^t) 
  + \langle \nabla L(\w^t), \w{-}\w^t\rangle  
  + \frac{1}{2}\langle \w{-}\w^t, \nabla^2 L(\w^t)(\w {-}\w^t) \rangle. 
  \label{optim:eq:newton}
\end{align}
If the Hessian is positive definite at $\w^t$, which we denote by $\nabla^2
L(\w^t) \succ 0$, then the minimum is well-defined and unique (this is for
example the case if $L$ is strictly convex). The iterates can then be written
analytically as
\[
  \w^{t+1} = \w^t - \nabla^2 L(\w^t)^{-1}\nabla L(\w^t).
\]
If the Hessian is not positive definite, the minimum may not be defined.
Ignoring this issue and taking the analytical formulation could be dangerous, as
it could amount to computing the maximum of the quadratic instead if, for
example, the quadratic was strictly concave (i.e., $\nabla^2 L(\w) \prec 0$).

\subsection{Regularized Newton method}

A simple technique to circumvent this issue consists in adding a
regularization term to the Hessian. Namely, from a variational viewpoint, we can
add a proximity term $\frac{1}{2} \|\w -\w^t\|_2^2$,
encouraging it to stay close to the current $\w^t$. The iterates of this
regularized Newton method then take the form
\begin{align*}
\w^{t+1} 
= \argmin_{\w \in \cW} ~
&L(\w^t) + \langle \nabla L(\w^t), \w{-}\w^t\rangle +
\frac{1}{2}\langle \w{-}\w^t, \nabla^2 L(\w^t)(\w {-}\w^t) \rangle  \\
&+ \frac{\eta^t}{2} \|\w-\w^t\|_2^2,
\end{align*}
where $\eta^t$ controls the regularization strength.
Assuming $\eta^t > 0$ is strong enough
to make $\nabla^2 L(\w^t) + \eta^t \idm$ positive-definite, we have
\begin{equation*}
\w^{t+1} = \w^t - \d^t,
\end{equation*}
where we defined the direction
\begin{equation}
\d^t \coloneqq (\nabla^2 L(\w^t) + \eta^t \idm)^{-1} \nabla L(\w^t).
\label{optim:eq:reg_newton_direction}
\end{equation}
Other techniques to circumvent this issue include using cubic
regularization and modifying the spectral decomposition of the Hessian, by
thresholding the eigenvalues or taking their absolute values. We refer the
interested reader to, e.g., \citep{nesterov2018lectures,wright1999numerical} for
more details. 

\subsection{Approximate direction}

We observe a main impediment for implementing such a second-order optimization
algorithm: even if we had access to the Hessian of the objective for free and
this Hessian was positive definite, computing the exact direction $\d^t$ in
\cref{optim:eq:reg_newton_direction}
requires computing an inverse-Hessian vector product (IHVP) with
the gradient $\nabla L(\w^t)$. Doing so exactly
requires solving a linear system
\begin{equation*}
(\nabla^2 L(\w^t) + \eta^t \idm) \d^t =  \nabla L(\w^t),
\end{equation*}
which a priori takes $O(P^3)$ time. In practice, however, we can 
compute IHVPs approximately, as explained in \cref{higher:sec:ihvp}.

\subsection{Convergence guarantees}

While implementing Newton's method comes at a higher computational cost, it can
also benefit from faster convergence rates. Briefly, if Newton's method is
initialized at a point $\w^0 \in \cW$ close enough to the minimizer
$\w^\star$ of a $\mu$-strongly convex function with $M$-Lipschitz continuous
Hessian (namely $\|\w^0 -\w^*\|_2\leq \frac{2\mu}{3M}$), then Newton's method
converges at a quadratic rate \citep{nesterov2018lectures}, that is, $R(t) \leq
O(\exp(-2^t))$ (see \cref{optim:sec:perf} for a brief introduction to
performance guarantees). This is far superior to
gradient descent. Such an efficiency motivated the development of interior point
methods, that have been a breakthrough in constrained optimization, thanks to
the use of log-barrier penalties~\citep{nesterov2018lectures}.

\subsection{Linesearch}

In practice, we may not have access to an initial point close enough to the
minimizer. In that case, even for strictly convex functions for which Newton's
steps are well-defined, taking $\w^{t+1} = \w^t - \d^t$ may not ensure a
decrease of the objective values. 
Nevertheless, the direction $\d^t$ may define a descent direction as defined
below.

\begin{boxdef}{Descent direction\label{optim:rem:descent_dir}}
  A point $\d \in \cW$ defines a \textbf{descent direction} $-\d$ for an objective $L$ at $\w$,
  if there exists a positive stepsize $\gamma >0$ such that 
  \[
  L(\w - \gamma \d) \leq L(\w).
  \]
  If $L$ is differentiable, $-\d$ is a descent direction if $\langle -\d,
  \nabla L(\w) \rangle <0$. 
\end{boxdef}
For Newton's method without regularization, 
$\d^t = \nabla^2 L(\w^t)^{-1}\nabla L(\w^t)$ is then a
descent direction at $\w^t$, as long as $\nabla L(\w^t) \neq 0$ and $\nabla^2
L(\w^t)\succ0$. If $\nabla^2 L(\w^t)\not \succ 0$, choosing $\eta^t >0$ such
that $\nabla^2 L(\w^t) + \eta^t \idm \succ 0$, also ensures that $\d^t = -
(\nabla^2 L(\w^t) +\eta^t\idm)^{-1}\nabla L(\w^t)$ is a descent direction (as
long as $\nabla L(\w^t) \neq 0$). Newton's method is then generally equipped
with a linesearch method that attempts to take steps of the form 
\[
  \w^{t+1} = \w^t - \gamma^t \d^t
\] 
with $\gamma^t$ chosen as the largest stepsize among $\{\rho^\tau, \tau \in
\mathbb{N}\}$ for $\rho \in (0, 1)$ until a sufficient
decrease of the objective is satisfied such as, for $c \in (0, 1)$,
\[
  L(\w^t - \gamma^t \d^t) \leq L(\w^t) 
  - c \gamma^t
  \langle \nabla L(\w^t), \nabla^2 L(\w^t)^{-1}\nabla L(\w^t) \rangle.
\]
For strongly convex functions, such an implementation exhibits two phases: a
first phase during which Newton's steps are ``damped'' by using a stepsize
$\gamma^t <1$ and a second phase of super-fast convergence during which
stepsizes $\gamma^t = 1$ are taken, and the objective
decreases very fast. Even far from the optimum, Newton directions can
advantageously adapt to the local geometry of the objective to speed up
convergence compared to a regular gradient descent as explained below.

\subsection{Geometric interpretation}

\begin{figure}
  \begin{center}
    \includegraphics[width=\linewidth]{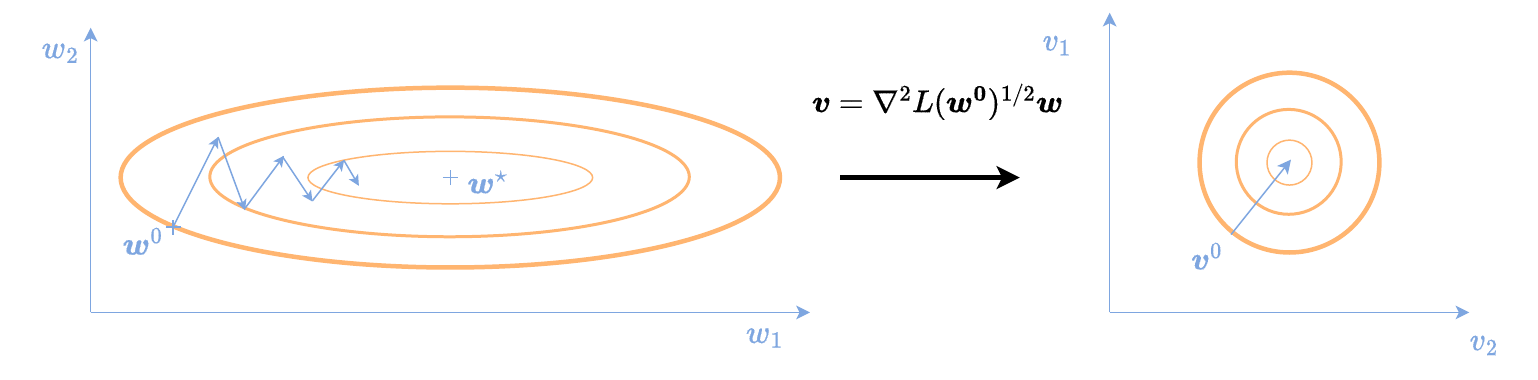}
    \caption{Left: Minimization of a quadratic $L(\w) \coloneqq \frac{1}{2}a w_1^2 +
    \frac{1}{2} b w_2^2$ by gradient descent. For $a\gg b \geq 0$, gradient
    descent typically oscillates. Right: minimization by Newton's method amounts
    to changing the geometry of the problem to avoid
    oscillations.\label{optim:fig:newton}}
  \end{center}
\end{figure}

To understand the efficiency of Newton's method compared to gradient
descent, consider the minimization of a simple quadratic 
\[
  L(\w) = \frac{1}{2} a
w_1^2 + \frac{1}{2} b w_2^2
\] 
for $a\gg b \geq 0$, as illustrated in \cref{optim:fig:newton}. A gradient descent
moves along the directions $\nabla L(\w) = (a w_1, b w_2)^\top$ and its stepsize
is limited by the variations in the first coordinate leading to some
oscillations. If we were simply rescaling the gradient by $(a, b)$, i.e., taking
steps of the form 
\[
  \w^{t+1} =
\w^t - \gamma \diag(a^{-1}, b^{-1})\nabla L(\w^t),
\]
the variations in both coordinates would be normalized to one and the stepsize
could simply be chosen to $\gamma = 1$ to directly get $\w^\star$. In other words,
by adapting the geometry of the directions with the geometry induced by the
objective, we can circumvent the oscillations. 

That's exactly what Newton's method does by modifying the gradient direction
using the inverse of the Hessian. Formally, at iteration $t$, consider the
modified objective 
\[ 
  \tilde L(\v) = L(A \v) 
  \ \mbox{for} \ 
  A = \nabla^2 L(\w^t)^{-1/2},
\]
with $L$ strictly convex and $A$ the inverse matrix square root of the Hessian.
One easily verifies that a Newton step is equivalent to a gradient step on
$\tilde L$, that is, 
\begin{equation*}
\v^{t+1}  = \v^t - \nabla \tilde L(\v^t) 
\ \iff \ 
\w^{t+1} = \w^t - (\nabla^2 L(\w^t))^{-1} \nabla L(\w^t)
\end{equation*}
where
\begin{equation*}
\w^t = A \v^t = \nabla^2 L(\w^t)^{-1/2} \v^t.
\end{equation*}
In the geometry induced by $A$, the objective is generally better conditioned as
illustrated in \cref{optim:fig:newton}. This explains the efficiency of
Newton's method. In particular for any strongly convex quadratic, a Newton step
reaches the optimum in one iteration, while a gradient step can take many more
iterations.

\subsection{Stochastic Newton's method}

Consider now an expected loss  
\[
\min_{\w \in \cW} \ 
L(\w) \coloneqq \mathbb{E}_{S \sim \rho} \left[L(\w; S)\right].
\]
In that case, an estimate of the Hessian can be constructed just like for the
gradient using that
\[
  \mathbb{E}_{S \sim \rho} \left[\nabla^2 L(\w; S)\right] = \nabla^2 L(\w).
\]
Les us then denote
\[
  g(\w; S) \approx \nabla L(\w), \quad H(\w; S') \approx \nabla^2 L(\w)
\]
some stochastic estimates of the gradient and the Hessian, respectively, with
$S, S'$ independently drawn from $\rho$ or from mini-batch approximations with
varying mini-batch sizes. One implementation of a \textbf{stochastic Newton
method} can then be
\[
  \w^{t+1} = \w^t - \gamma^t (H(\w^t; S') + \eta^t \idm)^{-1}g(\w^t; S),
\]
for $\eta^t \geq 0$ such that $(H(\w^t; S') + \eta^t \idm)^{-1} \succ 0$ and
$\gamma^t$ fixed or chosen to satisfy some sufficient decrease condition. We
refer the interested reader to, e.g., \citep{xu2020second}, for more details and
variants.

\section{Gauss-Newton method}
\label{optim:sec:gauss_newton}

Newton's method ~\eqref{optim:eq:newton} is usually not properly defined for
non-convex objective functions, since the Hessian may not be
positive definite at the current iterate. 
We saw in \cref{higher:sec:gnvp} that the Gauss-Newton matrix
can be used to define a positive-semidefinite approximation of the Hessian.
Here, we revisit the Gauss-Newton method from a variational and
\textbf{partial linearization} perspective. While the original
Gauss-Newton method originates from nonlinear least-squares, we will first
describe an extension to arbitrary convex loss functions, since it is both more
general and easier to explain.

\subsection{With exact outer function}

Consider a composite objective of the form
\[
  L(\w) \coloneqq \ell(f(\w)),
\]
where 
$\ell: \cM \rightarrow \RR$ 
is a \textbf{convex} function, such as a convex loss function applied on a given
sample, and 
$f: \cW \rightarrow \cM$ 
is a \textbf{nonlinear} function, such as a neural network with parameters 
$\w \in \cW$, evaluated on the same sample.
We saw that gradient descent and Newton's method amount to using \textbf{linear}
and \textbf{quadratic} approximations of $L(\w)$ around the current iterate
$\w^t$, respectively. As a middle ground between the two, the Gauss-Newton
method uses the linearization of $f$ around $\w^t$
\begin{equation*}
f(\w) \approx f(\w^t) + \partial f(\w^t) (\w -\w^t)
\end{equation*}
but keeps $\ell$ as is to obtain the objective
\begin{align*}
\w^{t+1} 
&\coloneqq \argmin_{\w \in \cW} \ell(f(\w^t) + \partial f(\w^t) (\w -\w^t)) \\
&= \argmin_{\w \in \cW} \ell(\partial f(\w^t) \w + f(\w^t) -\partial
f(\w^t)\w^t) \\
&= \argmin_{\w \in \cW} \ell(J^t \w + \deltav^t),
\end{align*}
where we defined the shorthands
$J^t \coloneqq \partial f(\w^t)$
and
$\deltav^t \coloneqq f(\w^t) - \partial f(\w^t)\w^t$.
We call
$\ell(J^t \w + \deltav^t)$
the \textbf{partial linearization} of $L = \ell \circ f$ at $\w^t$, 
as opposed to the full linearization of $L$ used in gradient descent. 

Since the composition of a convex function and of linear
function is convex,
this objective is \textbf{convex} even if $L(\w)$ is nonconvex.
In practice, we often add a proximity term as regularization to define
\begin{equation}
\w^{t+1} 
\coloneqq \argmin_{\w \in \cW} \ell(J^t \w + \deltav^t) 
+ \frac{\eta^t}{2} \|\w - \w^t\|_2^2.
\label{optim:eq:modified_gauss_newton}
\end{equation}
We can see this update as an approximation of the \textbf{proximal point} update
\begin{equation*}
\argmin_{\w \in \cW} L(\w) + \frac{\eta^t}{2} \|\w - \w^t\|_2^2,
\end{equation*}
where $L(\w)$ has been replaced by its partial linearization.
Solving \cref{optim:eq:modified_gauss_newton} using gradient-based solvers
requires computing the gradient of $\w \mapsto \ell(J^t \w + \deltav^t)$, which
is $\w \mapsto (J^t)^* \nabla \ell(J^t \w + \deltav^t)$.
Computing this gradient by autodiff therefore requires performing a forward pass
to compute the JVP $J^t \w$ and a backward pass to compute the VJP
$(J^t)^* \nabla \ell(\z)$. See \cref{diff:sec:jvp_vjp} for an introduction to
these operators and \cref{chap:auto_diff} for an introduction to autodiff.

The Gauss-Newton method with arbitrary convex outer loss is often called
modified Gauss-Newton \citep{nesterov2007modified} or prox-linear
\citep{drusvyatskiy2019efficiency}. The classical Gauss-Newton and
Levenberg-Marquardt
\citep{levenberg1944method, marquardt1963algorithm}
methods originate from
nonlinear least-squares and are recovered when $\ell(\z)$ is quadratic
\citep{kelley1995iterative}, such as
$\ell(\z) \coloneqq \frac{1}{2} \|\z - \y\|^2_2$, 
for $\y$ some reference target. 
The Gauss-Newton method corresponds classically to not using
regularization (i.e., $\eta^t = 0$) and the Levenberg-Marquardt
method uses regularization (usually called damping, potentially changing
$\eta^t$ across iterations).
See e.g.,~\citep{messerer2021survey}, for a survey of different variants.

\subsection{With approximate outer function}

Another variant of the Gauss-Newton method consists in replacing the convex loss
$\ell$ with its quadratic approximation around $\z^t \coloneqq f(\w^t)$,
\begin{equation*}
q^t(\z) \coloneqq \ell(\z^t) 
+ \langle \nabla \ell(\z^t), \z-\z^t\rangle 
+ \frac{1}{2}\langle \z-\z^t, \nabla^2 \ell(\z^t) (\z -\z^t)\rangle
\approx \ell(\z)
\end{equation*}
to define the update
\begin{equation*}
\w^{t+1} \coloneqq 
\argmin_{\w \in \cW} q^t(J^t \w + \deltav^t)
+ \frac{\eta^t}{2} \|\w - \w^t\|_2^2.
\end{equation*}
Notice that $\ell$ has been replaced by its quadratic approximation $q^t$. This
objective is always a \textbf{convex quadratic}, unlike the objective of the
Newton method in \cref{optim:eq:newton}, which is a priori a \textbf{nonconvex
quadratic}, if $f$ is nonlinear. Simple calculations show that
\begin{align*}
\w^{t+1} = \argmin_{\w \in \cW} ~
&L(\w^t) + \langle \q^t, J^t(\w {-} \w^t) \rangle \\
& + \frac{1}{2} \langle \w {-} \w^t, 
  (J^t)^* Q^t J^t (\w {-} \w^t) \rangle 
 +\frac{\eta^t}{2} \|\w - \w^t\|^2_2
\end{align*}
where
$
    \q^t \coloneqq \nabla \ell(f(\w^t)) \in \cM=\RR^Z, 
    Q^t \coloneqq \nabla^2 \ell(f(\w^t)) \in \RR^{Z \times Z}.
$
The closed form solution is
\begin{align*}
\w^{t+1} 
&= \w^t - ((J^t)^* Q^t J^t + \eta^t \idm)^{-1} (J^t)^* \q^t \\
&= \w^t - (\gn (\ell \circ f) (\w^t) + \eta^t \idm)^{-1} \nabla L(\w^t),
\end{align*}
where we used the (generalized) \textbf{Gauss-Newton matrix} 
of $L = \ell \circ f$, defined in \cref{higher:sec:gnvp}.

\subsection{Linesearch}

Similarly to Newton's method, the iterates of a Gauss-Newton method may diverge
when used alone. However, the direction $-(\gn(\ell\circ f)(\w^t) + \eta^t
\idm)^{-1} \nabla L(\w^t)$ defines a descent direction for any $\eta^t >0$ and
can be combined with a stepsize $\gamma^t$ (typically chosen using a linesearch)
to obtain iterates of the form
\[
\w^{t+1} = 
\w^t - \gamma^t(\gn(\ell\circ f)(\w^t) + \eta^t \idm)^{-1} \nabla L(\w^t).
\]

\subsection{Stochastic Gauss-Newton}

In deep learning, the objective generally consists in an expectation over
samples of the composition between a loss function and a network function:
\[
L(\w) 
= \mathbb{E}_{S \sim \rho} [L(\w; S)] 
= \mathbb{E}_{(X, Y) \sim \rho}[\ell(f(\w; X); Y)]
\] 
where $S = (X, Y)$ denotes a sample pair of input $X$ with associated label $Y$.
In that case, 
as already studied in \cref{higher:sec:gnvp},
the Gauss-Newton matrix $\gn L$ is the expectation
of the individual Gauss-Newton matrices 
\begin{align*}
\gn L(\w; \x, \y) 
& \coloneqq \partial f(\w; \x)^\top \nabla^2 \ell(f(\w; \x)) \partial f(\w; \x), \\
\gn L (\w) & \coloneqq \mathbb{E}_{(X, Y)\sim
  \rho}[\gn L(\w; \x, \y)]. \nonumber
\end{align*}
We can estimate the gradient and the Gauss-Newton matrix by, respectively, $\g(\w;
S) \approx \nabla L(\w)$, and $G(\w; S') \approx \gn L(\w)$ for $S, S' \sim \rho$
or using mini-batch approximations. A \textbf{stochastic} Gauss-Newton method
therefore performs iterates
\[
\w^{t+1} \coloneqq \w^t - \gamma^t(G(\w^t; S') + \eta^t \idm)^{-1} \g(\w^t; S),
\]
for $\eta^t \geq 0$ and $\gamma^t$ fixed or selected to satisfy some criterion.

\section{Natural gradient descent}\label{optim:sec:nat_grad}

Natural gradient descent~\citep{amari1998natural} follows a similar
principle as gradient descent: linearize the objective around the current
iterate and minimize this approximation together with a proximity term. 
It differs from gradient descent in the choice of the proximity term:
rather than using a squared Euclidean distance between the \textbf{parameters},
it uses a Kullback-Leibler divergence between the \textbf{probability
distributions} these parameters define. 

\subsubsection*{Negative log-likelihood}

We consider objectives of the form
\[
\min_{\w \in \cW} L(\w) 
= \EE_{S \sim \rho} \left[L(\w; S)\right]
= \EE_{S \sim \rho} \left[-\log q_\w(S)\right],
\]
where $\rho$ is an unknown data distribution (but from which we can sample) 
and
where $q_\w$ is a probability distribution parameterized by $\w$. 
As reviewed in \cref{chap:proba_learn}, the negative log-likelihood can be
used as a loss function (many loss functions can be seen from this perspective,
including the squared and logistic loss functions).
In the unsupervised setting, where $S=Y$, we simply use $q_\w(Y)$ as is.
In the supervised setting, where $S=(X,Y)$, we use the product rule
$\PP(X, Y) = \PP(X) \PP(Y|X)$
to parameterize $q_\w(S)$ as
\begin{equation*}
q_\w(\x, \y) \coloneqq \rho_X(\x) p_\thetav(\y),
\end{equation*}
where $\rho_X$ is the marginal distribution for $X$,
$p_\thetav(\y)$ is the PMF/PDF of a probability distribution 
and $\thetav = f(\w; \x)$ is for instance a neural network with parameters $\w
\in \cW$ and input $\x \in \cX$.

\subsection{Variational perspective}

Natural gradient descent is motivated by updates of the form
\[
  \w^{t+1} = 
  \argmin_{\w \in \cW} L(\w^t) 
  + \langle \nabla L(\w^t), \w - \w^t\rangle 
  + \mathrm{KL}(q_{\w^t},q_\w),
\]
where $\mathrm{KL}(p,q) \coloneqq \int p(\z) \log \frac{p(\z)}{q(\z)} d\z$ is
the Kullback-Leibler (KL) divergence.
Unlike gradient descent, the proximity term is therefore
between the current distribution $q_{\w^t}$ and a candidate
probability distribution $q_{\w}$.
The above problem is intractable in general, as the KL may not have a closed
form. Nevertheless, its quadratic approximation can be shown
\citep{amari1998natural} to admit a simple form,
\begin{equation*}
\mathrm{KL}(q_{\w^t}, q_\w) \approx 
\frac{1}{2}\langle \w - \w^t, \fisher L (\w^t)(\w -\w^t)\rangle
\end{equation*}
where we used the Fisher information matrix $\fisher L (\w)$,
studied in \cref{higher:sec:fisher}.
Equipped with this quadratic approximation of the KL divergence, natural
gradient descent amounts to computing iterates as 
\begin{align*}
\w^{t+1} \coloneqq 
\argmin_{\w \in \cW} ~
&L(\w^t) 
+ \langle \nabla L(\w^t), \w - \w^t\rangle \\
& + \frac{1}{2}\langle \w - \w^t, \fisher L (\w^t)(\w -\w^t)\rangle
  + \frac{\eta^t}{2}\|\w-\w^t\|^2_2,
\end{align*}
where a quadratic proximity term was added to ensure a unique solution.
This is a strictly convex problem as $\fisher L(\w^t)$ is positive semi-definite.
The closed-form solution is
\[
  \w^{t+1} = \w^t - (\fisher L (\w^t) + \eta^t \idm )^{-1}\nabla L(\w^t).
\]
Because the Gauss-Newton and Fisher information matrices are equivalent when
$p_\thetav$ is an exponential family distribution
(\cref{higher:prop:equiv_fisher_gn}), the Gauss-Newton and natural gradient
methods coincide in this case.

\subsection{Stochastic natural gradient descent}

In practice, we may not have access to $\nabla L(\w^t)$ in closed form as it is
an expectation over $\rho$.
Moreover, $\fisher L(\w^t)$ may not be
computable in closed form either. To estimate the Fisher information matrix, 
we can use that (see \cref{higher:sec:fisher})
using the shorthand $\thetav \coloneqq f(\w, X)$,
\begin{equation*}
\fisher L(\w) = \EE_{X \sim \rho_\cX} \EE_{Y \sim p_\thetav}
[\nabla L(\w; X, Y) \otimes \nabla L(\w; X, Y)].
\end{equation*}
We can then build estimates
$\g(\w^t; S) \approx \nabla L(\w^t, S)$ and $F(\w^t; S') \approx \fisher L(\w^t)$
for $S$ sampled from $\rho$ and $S'$ sampled from $q_{\w^t}(\x, \y) = p_\cX(\x)
\rho_{\thetav}(\y)$.
A stochastic natural gradient descent can then be implemented as
\[
  \w^{t+1} = \w^t - \gamma^t(F(\w^t; S') + \eta^t \idm )^{-1}\g(\w^t; S),
\]
where $\gamma^t$ is a stepsize, possibly chosen by linesearch.

In deep learning, the product with the inverse Fisher or Gauss-Newton matrices
can remain costly to compute. Several approximations have been proposed,
such as KFAC \citep{martens2015optimizing, botev2017practical},
which uses a computationally efficient structural approximation to these
matrices.

\section{Quasi-Newton methods}

\subsection{BFGS}

A celebrated example of \textbf{quasi-Newton} method is the BFGS method
\citep{broyden1970convergence,fletcher1970new,goldfarb1970family,shanno1970conditioning},
whose acronym follows from its authors' names. The rationale of the BFGS update
stems once again from a variational viewpoint. We wish to build a simple
quadratic model of the objective $h^t(\w) = L(\w^t) + \langle \nabla L(\w^t), \w
- \w^t\rangle + \frac{1}{2} \langle \w - \w^t, Q^t (\w -\w^t) \rangle$ for some
$Q^t$ built along the iterations rather than taken as $\nabla^2 L(\w^t)$. One
desirable property of such a quadratic model would be that its gradients at consecutive
iterates match the gradients of the original function, i.e., $\nabla h^t(\w^t) =
\nabla L(\w^t)$ and $\nabla h^t(\w^{t-1})= \nabla L(\w^{t-1})$. A simpler condition,
called the \textbf{secant condition} consists in considering the differences of
these vectors, that is, ensuring that
\begin{align*}
  &\nabla h^t(\w^t) - \nabla h^t(\w^{t-1}) = \nabla L(\w^t) - \nabla L(\w^{t-1})\\
    \iff& Q^t (\w^t - \w^{t-1}) = \nabla L(\w^t) - \nabla L(\w^{t-1})\\
    \iff& \w^t - \w^{t-1} = B^t(\nabla L(\w^t) - \nabla L(\w^{t-1})),
\end{align*}
for $B^t = (Q^t)^{-1}$. Building $B^t$, a surrogate of the inverse of the
Hessian satisfying the secant equation, can then be done as 
\begin{equation*}
B^{t+1} \coloneqq 
\left(\idm - \rho^t \s^t (\y^t)^\top \right)
B^t 
\left(\idm - \rho^t \s^t (\y^t)^\top\right)
+ \rho^t \s^t (\s^t)^\top
\end{equation*}
where
\begin{align*}
  \s^t &\coloneqq \w^{t+1} - \w^t \\
  \y^t &\coloneqq \nabla L(\w^{t+1}) - \nabla L(\w^t) \\
  \rho^t &\coloneqq \frac{1}{\langle \s^t, \y^t \rangle}.
\end{align*}
A typical implementation of BFGS stores $B^t \in \RR^{P \times P}$ in memory,
which is prohibitive when $P$ is large.

\subsection{Limited-memory BFGS}

In practice, the limited-memory counterpart of BFGS, called
LBFGS~\citep{liu1989limited}, is often preferred.  The key observation of LBFGS
is that we do not need to materialize $B^t$ in memory: we only need to multiply
it with the gradient $\nabla L(\w^t)$. That is, we can see $B^t$ as a linear
map.  Fortunately, the product between $B^t$ and any vector $\v$ can be computed
efficiently if we store $(\s^1, \y^1, \rho^1), \dots, (\s^t, \y^t, \rho^t)$ in
memory.  In practice, a small history of past values is used to reduce memory
and computational cost.  Because LBFGS has the benefits of second-order-like
methods with much reduced cost, it has become a de-facto algorithm,
outperforming most other algorithms for medium-scale problems without particular
structure~\citep{liu1989limited}. 

\section{Approximate Hessian diagonal inverse preconditioners}

One application of the approximations of the Hessian diagonal developed in
\cref{higher:sec:diag_approx} is to obtain cheap approximations of the Hessian
diagonal inverse,
\[
B^t \coloneqq \diag(|H_{11}^t|^{-1}, \ldots, |H_{PP}^t|^{-1}).
\]
Such a scaling would for instance be sufficient to make the
quadratic example presented in~\cref{optim:fig:newton} work. 
Many optimization algorithms, including the popular ADAM,
can be viewed as using a preconditioner that approximates the inverse of the
Hessian's diagonal.

\section{Summary}

\begin{itemize}

\item We reviewed Newton's method, the Gauss-Newton method, natural
gradient descent, quasi-Newton methods and preconditioning methods.

\item We adopted a variational viewpoint, where the method's next iterate is computed
as the solution of a trade-off between minimizing an approximation of the
function (linear, partially linear, quadratic) and a proximity term (squared
Euclidean, KL). 

\item All methods were shown to use iterates of the form
\begin{equation*}
\w^{t+1} \coloneqq \w^t - \gamma^t B^t \nabla L(\w^t)
\end{equation*}
but have different trade-offs between the cost it takes to evaluate
$B^t \nabla L(\w^t)$ and the richness of the information used about $L$.

\end{itemize}

%% file: chapters/duality/duality.tex
\chapter{Duality}
\label{chap:duality}

In this chapter, we review duality principles in optimization.

\section{Dual norms}
\label{duality:sec:dual_norms}

We introduce in this section dual norms, since they are useful in
this book.
\begin{boxdef}{Dual norms}
Given a norm $\|\u\|$, its dual is
\begin{equation*}
\|\v\|_* 
\coloneqq 
\max_{\|\u\| \le 1} \langle \u, \v \rangle.
\end{equation*}
\end{boxdef}
Therefore, the dual norm of $\|\cdot\|$ is the \textbf{support function} of the 
unit ball induced by the norm $\|\cdot\|$,
\begin{equation*}
B_{\|\cdot\|} \coloneqq \{\u \in \RR^D \colon \|\u\| \le 1\}.
\end{equation*}
We give examples of pairs of dual norms below.
\begin{boxexm}{Dual norm of $p$-norms}
The $p$-norm is defined by
\begin{equation*}
\|\u\|_p \coloneqq 
\left(
\sum_{j=1}^D |u_j|^p
\right)^{1/p}.
\end{equation*}
Its dual is $\|\v\|_q$ where $q$ is such that $\frac{1}{p} + \frac{1}{q} = 1$.
For instance, the dual norm of the $2$-norm is itself, since $\frac{1}{2} +
\frac{1}{2} = 1$. The $1$-norm and the $\infty$-norm are duals of each other,
since $\frac{1}{1} + \frac{1}{\infty} = 1$.
\end{boxexm}
The definition of dual norm implies a generalization
of the \textbf{Cauchy–Schwarz inequality}: for all $\u, \v \in \RR^D$
\begin{equation*}
|\langle \u, \v \rangle|
\le
\|\u\|_* \|\v\|.
\end{equation*}
See, e.g., \citet[Lemma 1.4]{beck_2017}.
\begin{boxprop}{Conjugate of norms and squared norms}
We know that the conjugate of the support function is the indicator
function. Therefore, if $f(\u) = \|\u\|$, then
\begin{equation*}
f^*(\v) 
= \indic_{B_{\|\cdot\|}}(\v)
= \begin{cases}
    0 &\mbox{ if } \|\v\|_* \le 1 \\
    \infty &\mbox { otherwise }
\end{cases}.
\end{equation*}
On the other hand, if $f(\u) = \frac{1}{2} \|\u\|^2$, then
\begin{equation*}
f^*(\v) = \frac{1}{2} \|\v\|^2_*.
\end{equation*}
\end{boxprop}

\section{Fenchel duality}

We consider in this section standard objectives of the form
\begin{equation*}
\min_{\w \in \cW} L(\w) \coloneqq
\min_{\w \in \cW} \ell(f(\w)) + R(\w),
\end{equation*}
where 
$f \colon \cW \to \cM$,
$\ell \colon \cM \to \RR$ 
and $R \colon \cW \to \RR$.
We first show that the minimization of this objective, called the
\textbf{primal}, can be lower bounded by a \textbf{concave} maximization
objective, called the \textbf{dual}, even if the primal is nonconvex.
\begin{boxprop}{Weak duality}
Let 
$f \colon \cW \to \cM$ (potentially nonlinear),
$\ell \colon \cM \to \RR$ (potentially nonconvex)
and 
$R \colon \cW \to \RR$ (potentially nonconvex).
Then
\begin{equation*}
\min_{\w \in \cW} \ell(f(\w)) + R(\w)
\ge
\max_{\alphav \in \cM}
-R^f(\alphav) - \ell^*(-\alphav),
\end{equation*}
where we used the conjugate
\begin{equation*}
\ell^*(-\alphav) \coloneqq 
\max_{\thetav \in \cM} \langle -\alphav, \thetav \rangle -
\ell(\thetav)
\end{equation*}
and the ``generalized conjugate''
\begin{equation*}
R^{f}(\alphav) \coloneqq 
\max_{\w \in \cW}
\langle \alphav, f(\w) \rangle - R(\w).
\end{equation*}
Moreover, $\ell^*$ and $R^f$ are both convex functions.
\label{duality:prop:weak_duality}
\end{boxprop}
We emphasize that the result in \cref{duality:prop:weak_duality} is fully
general, in the sense that it does not assume the linearity of $f$ or the
convexity of $\ell$ and $R$. The caveat, of course, is that $R^f$ 
and $\ell^*$ are difficult to compute in general, 
if $f$ is nonlinear, and if $\ell$ and $R$ are nonconvex.
\begin{proof}
\begin{align*}
&\min_{\w \in \cW} \ell(f(\w)) + R(\w)\\
=& \min_{\substack{\w \in \cW\\\thetav \in \cM}}
\ell(\thetav) + R(\w)
\quad \text{s.t.} \quad \thetav = f(\w) \\
=& \min_{\substack{\w \in \cW\\\thetav \in \cM}}
\max_{\alphav \in \cM}
\ell(\thetav) + R(\w) + \langle \alphav, \thetav - f(\w) \rangle \\
\ge&
\max_{\alphav \in \cM}
\min_{\substack{\w \in \cW\\\thetav \in \cM}}
\ell(\thetav) + R(\w) + \langle \alphav, \thetav - f(\w) \rangle \\
=&
\max_{\alphav \in \cM}
\min_{\w \in \cW}
\langle \alphav, -f(\w) \rangle + R(\w)
+
\min_{\thetav \in \cM}
\ell(\thetav) + \langle \alphav, \thetav\rangle \\
=&
\max_{\alphav \in \cM}
-\max_{\w \in \cW}
\langle \alphav, f(\w) \rangle - R(\w)
-
\max_{\thetav \in \cM}
\langle -\alphav, \thetav\rangle - \ell(\thetav) \\
=&
\max_{\alphav \in \cM}
-R^f(\alphav) - \ell^*(-\alphav).
\end{align*}
\end{proof}
In the case when $f(\w) = A \w$, where $A$ is a linear map, 
and when both $\ell$ and $R$ are convex,
we can state a much stronger result.
\begin{boxprop}{Strong duality}
Let $A$ be a linear map from $\cW$ to $\cM$.
Let
$\ell \colon \cM \to \RR$
and 
$R \colon \cW \to \RR$ 
be convex functions.
Let $A^*$ denote the adjoint of $A$ (\cref{diff:sec:jvp_vjp}).
Then,
\begin{equation*}
\min_{\w \in \cW} \ell(A\w) + R(\w)
=
\max_{\alphav \in \cM}
-R^*(A^*\alphav) - \ell^*(-\alphav).
\end{equation*}
Furthermore, the primal solution satisfies
\begin{equation*}
\w^\star 
\in \argmax_{\w \in \cW} \langle A \alphav^\star, \w \rangle - R(\w).
\end{equation*}
When $R$ is strictly convex, the primal solution is uniquely determined by
\begin{equation*}
\w^\star = \nabla R^*(A^* \alphav^\star). 
\end{equation*}
\label{duality:prop:strong_duality}
\end{boxprop}
\begin{proof}
Since $f(\w) = A\w$, we have
\begin{align*}
R^{f}(\alphav) 
&\coloneqq 
\max_{\w \in \cW}
\langle \alphav, f(\w) \rangle - R(\w) \\
&= 
\max_{\w \in \cW}
\langle \alphav, A\w \rangle - R(\w) \\
&=\max_{\w \in \cW}
\langle A^*\alphav, \w \rangle - R(\w) \\
&=R^*(A^*\alphav).
\end{align*}
Furthermore, the inequality in the proof of \cref{duality:prop:weak_duality}
is an equality, since the $\min \max$ is that of a convex-concave function.
\end{proof}
The maximization problem in \cref{duality:prop:strong_duality} 
is called the \textbf{Fenchel dual}. By strong duality, the value of the maximum
and the value of the minimum are equal. We can therefore choose to equivalently
solve the dual instead of the primal. This can be advantageous when the space
$\cM$ is smaller than $\cW$.

We now apply the Fenchel dual to obtain the dual of 
regularized multiclass linear classification. 
\begin{boxexm}{Sum of separable loss functions}
When the loss is\\ $\ell(\thetav) \coloneqq \sum_{i=1}^N \ell_i(\thetav_i)$,
where $\thetav = A\w = (A_1 \w, \dots, A_N \w) \in \cM^N$
and $A_i$ is a linear map from $\cW$ to $\cM$,
we obtain
\begin{equation*}
\min_{\w \in \cW} \sum_{i=1}^N \ell_i(A_i \w) + R(\w)
=
\max_{\alphav \in \cM^N}
-R^*(A^*\alphav) - \sum_{i=1}^N \ell_i^*(-\alphav_i),
\end{equation*}
where $A^* \alphav = (A_1^* \alphav_1, \dots, A_N^* \alphav_N)$.
Typically, we define
\begin{equation*}
A_i \w \coloneqq \W \x_i,
\end{equation*}
where $\W \in \RR^{M \times D}$ is a reshaped version of $\w \in \cW$,
$\x_i \in \RR^D$ is a training sample, and $M$ is the number of classes. 
In this case, we then have
\begin{equation*}
A_i^* \alphav_i = \alphav_i \x_i^\top. 
\end{equation*}
\end{boxexm}
Examples of loss function conjugates are given 
in \cref{duality:tab:loss_conjugates}.

\begin{table}[t]
\caption{Examples of loss conjugates. 
For regression losses (squared, absolute),
where $\y_i \in \RR^M$,
we define $\t_i = \phi(\y_i) = \y_i$.
For classification losses (logistic, perceptron, hinge),
where $y_i \in [M]$,
we define $\t_i = \phi(y_i) = \e_{y_i}$.
To simplify some expressions, we defined
the change of variable $\muv_i \coloneqq \y_i - \alphav_i$.}
\centering
\begin{tabular}{lcc}
\toprule
~ & $\ell_i(\thetav_i)$ & $\ell_i^*(-\alphav_i)$ \\
\midrule
Squared 
& $\frac{1}{2} \|\thetav_i - \t_i\|^2_2$ 
& $\frac{1}{2}\|\alphav_i\|^2_2 - \langle \t_i, \alphav_i \rangle$ \\
Absolute & $\|\thetav_i - \t_i\|_1$ & 
$\indic_{[-1,1]^M}(\alphav_i) - \langle \t_i, \alphav_i \rangle$\\
Logistic & $\mathrm{LSE}(\thetav_i) - \langle \thetav_i, \t_i \rangle$ & 
$\langle \muv_i, \log \muv_i \rangle + \indic_{\triangle^M}(\muv_i)$ \\
Perceptron
     & $\max_{j \in [M]} \theta_{i,j} - \theta_{i, y_i}$
& $\indic_{\triangle^M}(\muv_i)$ \\
Hinge & $\max_{j \in [M]} [j \neq y_i] + \theta_{i,j} - \theta_{i,y_i}$ 
      & $\indic_{\triangle^M}(\muv_i) - \langle \ones - \t_i, \muv_i \rangle$ \\
\bottomrule
\end{tabular}
\label{duality:tab:loss_conjugates}
\end{table}

\section{Bregman divergences}

\begin{figure}
\centering
\includegraphics{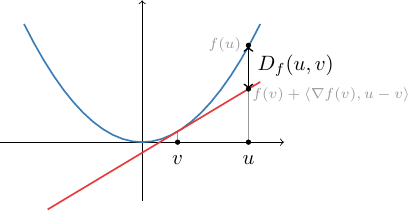}
\caption{The Bregman divergence generated by $f$ 
is the difference between $f(\u)$ and its linearization around $\v$.}
\label{duality:fig:bregman}
\end{figure}

Bregman divergences are a measure of difference between two points.
\begin{boxdef}{Bregman divergence}
The Bregman divergence generated by a differentiable convex function 
$f \colon \RR^D \to \RR$ is
\begin{align*}
B_f(\u, \v)
&\coloneqq f(\u) - f(\v) - \langle \nabla f(\v), \u - \v \rangle \\
&= \langle \nabla f(\v), \v \rangle - f(\v) - 
\left[\langle \nabla f(\v), \u \rangle - f(\u)\right],
\end{align*}
where $\u, \v \in \dom(f)$.
\label{duality:def:bregman_div}
\end{boxdef}
Intuitively, the Bregman divergence is the difference between $f(\u)$
and its linearization
$\u \mapsto f(\v) + \langle \nabla f(\v), \u - \v \rangle$ around $\v$.
This is illustrated in \cref{duality:fig:bregman}.

\begin{boxexm}{Examples of Bregman divergences}
If $f(\u) = \frac{1}{2} \|\u\|^2_2$, 
where $\dom(f) = \RR^D$, then
\begin{equation*}
B_f(\u, \v) = \frac{1}{2} \|\u - \v\|^2_2,
\end{equation*}
the \textbf{squared Euclidean distance}.
If $f(\u) = \langle \u, \log \u \rangle$, 
where $\dom(f) = \RR_+^D$, then
\begin{equation*}
B_f(\u, \v) = 
\sum_{j=1}^D u_j \log \frac{u_j}{v_j}
- \sum_{j=1}^D u_j + \sum_{j=1}^D v_j,
\end{equation*}
the (generalized) \textbf{Kullback-Leibler divergence}.

\end{boxexm}

\subsubsection*{Properties}

Bregman divergences enjoy several useful properties.
\begin{boxprop}{Properties of Bregman divergences}
Let $f \colon \RR^D \to \RR$ be a differentiable convex function.
\begin{enumerate}

\item \textbf{Non-negativity:}
$B_f(\u, \v) \ge 0$ for all $\u, \v \in \dom(f)$.

\item \textbf{Positivity:}
$B_f(\u, \v) = 0$ if and only if $\u = \v$ 
(when $f$ is strictly convex).

\item \textbf{Convexity:}
$B_f(\u, \v)$ is convex in $\u$.

\item \textbf{Dual-space form:}
$B_f(\u, \v) = B_{f^*}(\b, \a)$,
where $\b = \nabla f(\v) \in \dom(f^*)$ and $\a = \nabla f(\u) \in \dom(f^*)$.

\end{enumerate}
\end{boxprop}
\begin{proof}
The properties follow immediately from the convexity of $f(\u)$.
\begin{enumerate}
\item From \cref{optim:def:convex_differentiable}.

\item From the uniqueness of minimizers.

\item From the fact that $\u \mapsto B_f(\u, \v)$ is the sum of $f(\u)$ and
    a linear function of $\u$.

\end{enumerate}
\end{proof}

The Bregman divergence can be used to define natural generalizations of the
Euclidean projection and proximal operators, reviewed
in \cref{optim:sec:proj_grad} and \cref{optim:sec:prox_grad}.
\begin{boxdef}{Bregman proximal and projection operators}
Let $\v \in \dom(f)$.
The Bregman proximal operator is
\begin{equation*}
\mathrm{bprox}_{f,g}(\v) 
\coloneqq
\argmin_{\u \in \dom(f) \cap \dom(g)}
B_f(\u, \v) + g(\u).
\end{equation*}
In particular, the Bregman projection onto $\cC \subseteq \dom(f)$ is
\begin{equation*}
\mathrm{bproj}_{f,\cC}(\v) 
\coloneqq
\argmin_{\u \in \cC} B_f(\u, \v).
\end{equation*}
\end{boxdef}
It turns out that these operators are intimately connected to the gradient
mapping of the convex conjugate.
\begin{boxprop}{Link with conjugate's gradient}
If $\Omega = f + g$, 
then for all $\thetav \in \dom(f^*)$
\begin{equation*}
\nabla \Omega^*(\thetav) 
= \mathrm{bprox}_{f,g}(\nabla f^*(\thetav)).
\end{equation*}
In particular, if $\Omega = f + \indic_{\cC}$,
then for all $\thetav \in \dom(f^*)$
\begin{equation*}
\nabla \Omega^*(\thetav) 
= \mathrm{bproj}_{f,\cC}(\nabla f^*(\thetav)).
\end{equation*}
\end{boxprop}
We give two examples below.
\begin{boxexm}{Bregman projections on the simplex}
If $f(\u) = \frac{1}{2}\|\u\|^2_2$, then
\begin{align*}
\mathrm{bproj}_{f, \triangle^D}(\v) 
= \argmin_{\u \in \triangle^D} \frac{1}{2} \|\u - \v\|^2_2.
\end{align*}
If $f(\u) = \langle \u, \log \u - \ones \rangle$, then
\begin{align*}
\mathrm{bproj}_{f, \triangle^D}(\v) 
= \argmin_{\u \in \triangle^D} \mathrm{KL}(\u, \v)
= \mathrm{softargmax}(\thetav),
\end{align*}
where
$\v = \nabla f^*(\thetav) = \exp(\thetav)$.
\end{boxexm}
Therefore, the softmax can be seen as a projection onto the probability simplex
in the Kullback-Leibler divergence sense!

\section{Fenchel-Young loss functions}

We end this chapter with a brief review of the Fenchel-Young family of loss
functions \citep{blondel_2020}, which includes all loss functions in
\cref{duality:tab:loss_conjugates}.
\begin{boxdef}{Fenchel-Young loss}
The Fenchel-Young loss function generated by $\Omega$ is
\begin{equation*}
\ell_\Omega(\thetav, \t)
\coloneqq
\Omega^*(\thetav) + \Omega(\t) - \langle \thetav, \t \rangle
\end{equation*}
where 
$\thetav \in \dom(\Omega^*)$
and
$\t \in \dom(\Omega)$.
\end{boxdef}
Typically, we set 
$\thetav = f(\x, \w)$, 
where $f$ is a model prediction function with parameters $\w$
and
$\t = \phi(\y)$, where $\phi \colon \cY \to \dom(\Omega)$.
For instance, suppose we work with categorical outputs $y \in [M]$.
Then, we can set $\phi(y) = \e_y$, where $\e_y$ is the one-hot encoding of $y$.

The important point to notice is that the Fenchel-Young loss is defined over
arguments in \textbf{mixed spaces}: $\thetav$ belongs to the dual space, while
$\t$ belongs to the primal space. In fact, the Fenchel-Young loss is intimately
connected to the Bregman divergence, since
$B_\Omega(\t, \v) 
= \Omega^*(\thetav) + \Omega(\t) - \langle \thetav, \t \rangle$,
if we set $\thetav = \nabla \Omega(\v)$. 
The key properties of Fenchel-Young loss functions are summarized below.
\begin{boxprop}{Properties of Fenchel-Young loss functions}
~
\begin{enumerate}

\item \textbf{Non-negativity:}
$\ell_\Omega(\thetav, \t) \ge 0$
for all $\thetav \in \dom(\Omega^*)$ and $\t \in \dom(\Omega)$.

\item \textbf{Positivity:}
$\ell_\Omega(\thetav, \t) = 0$ 
if and only if 
$\nabla \Omega^*(\thetav) = \t$,
assuming $\Omega$ is strictly convex.

\item \textbf{Convexity:} 
$\ell_\Omega(\thetav, \t)$ is convex in $\thetav$ (regardless of $\Omega$) 
and in $\t$ (if $\Omega$ is convex).

\item \textbf{Relation with composite Bregman divergence:}
\begin{equation*}
0 
\le 
\underbrace{B_\Omega(\t, \nabla \Omega^*(\thetav))}_{\text{possibly nonconvex
in } \thetav}
\le 
\underbrace{\ell_\Omega(\thetav, \t)}_{\text{convex in } \thetav}. 
\end{equation*}

\end{enumerate}
\end{boxprop}
See \citet{blondel_2020} for an in-depth study of more properties.

\section{Summary}

\begin{itemize}

\item The convex conjugate serves as a powerful
abstraction in \textbf{Fenchel duality}, decoupling the dual expression and
function-specific terms.  

\item The convex conjugate is also tightly connected to
\textbf{Bregman divergences} and can be used to derive the family of
\textbf{Fenchel-Young loss} functions, which can be seen as primal-dual
Bregman divergences.

\end{itemize}